\documentclass[a4paper,11pt,oneside,times,authoryear,print,index]{Classes/PhDThesisPSnPDF}

\input{Preamble/preamble}
\newcommand\I[1]{\mathbf{I}_{#1}}
\newcommand\Zero[1]{\mathbf{0}_{#1}}
\newcommand\Zeros[2]{\mathbf{0}_{#1 \times #2}}

\newcommand\X{\mathbf{X}}
\newcommand\Y{\mathbf{Y}}
\newcommand\Z{\mathbf{Z}}

\newcommand\A{\mathbf{A}}
\newcommand\B{\mathbf{B}}
\newcommand\Cbold{\mathbf{C}}

\newcommand\J{\mathbf{J}}

\newcommand\R{\mathbb{R}}

\newcommand\ghat[2]{\hm{\left[} #2 \hm{\right]}^\wedge_{#1}}
\newcommand\gvee[2]{\hm{\left[} #2 \hm{\right]}^\vee_{#1}}
\newcommand\gexp[1]{\exp_{#1}}
\newcommand\glog[1]{\log_{#1}}
\newcommand\gexphat[1]{\exp_{#1}^\wedge}
\newcommand\gExp[1]{\textup{Exp}_{#1}}
\newcommand\gLog[1]{\textup{Log}_{#1}}
\newcommand\glogvee[1]{\log_{#1}^\vee}
\newcommand\gadj[1]{\textup{Ad}_{#1}}
\newcommand\gsmalladj[1]{\textup{ad}_{#1}}
\newcommand\gljac[1]{\mathbf{J}^l_{#1}}
\newcommand\grjac[1]{\mathbf{J}^r_{#1}}
\newcommand\liebracket[1]{\hm{[} #1 \hm{]}}

\newcommand\SO[1]{\textup{SO}(#1)}
\newcommand\so[1]{\mathfrak{so}(#1)}
\newcommand\SE[1]{\textup{SE}(#1)}
\newcommand\se[1]{\mathfrak{se}(#1)}
\newcommand\SEk[2]{\textup{SE}_{#1}(#2)}
\newcommand\sek[2]{\mathfrak{se}_{#1}(#2)}
\newcommand\T[1]{\textup{T}(#1)}
\newcommand\SL[1]{\textup{SL}(#1)}

\newcommand\Exp{\textup{Exp}}

\newcommand\bias{{\mathbf{b}}}
\newcommand\biasHat{{\mathbf{\hat{b}}}}
\newcommand\noiseBias[1]{{\mathbf{w}_{{\mathbf{b}}}^{#1}}}

\newcommand\yRot[2]{{\prescript{{#1}}{}{\mathbf{\tilde{R}}}_{#2}}}

\newcommand\accBar[2]{{\prescript{{#2}}{}{\bar{\alpha}}^{\text{lin}}_{#1, #2}}}
\newcommand\yAcc[2]{ { { \tilde{\alpha} }^{g, \text{lin}}_{{#1, #2}}  } }
\newcommand\biasAcc{{{\mathbf{b}}_{a}}}
\newcommand\biasAccHat{{\mathbf{\hat{b}}_a}}
\newcommand\noiseAcc[1]{{\mathbf{w}_{a}^{#1}}}

\newcommand\yGyro[2]{{\prescript{{#2}}{}{\tilde{\omega}}_{#1, #2}}}
\newcommand\yGyroBar[2]{{\prescript{{#2}}{}{\bar{\omega}}_{#1, #2}}}
\newcommand\biasGyro{{{\mathbf{b}}_{g}}}
\newcommand\biasGyroHat{{\mathbf{\hat{b}}_g}}
\newcommand\noiseGyro[1]{{\mathbf{w}_{g}^{#1}}}

\newcommand\noiseLinVel[1]{{\mathbf{w}_{v}^{#1}}}
\newcommand\noiseAngVel[1]{{\mathbf{w}_{\omega}^{#1}}}

\newcommand\fkNoiseLinVel[1]{{\mathbf{n}_{v}^{#1}}}
\newcommand\fkNoiseAngVel[1]{{\mathbf{n}_{\omega}^{#1}}}

\newcommand\encoders{\mathbf{\tilde{s}}}
\newcommand\encoderSpeeds{\mathbf{\dot{\tilde{s}}}}
\newcommand\encoderNoise{\mathbf{w}_\mathbf{s}}

\newcommand\TransformMeasured[2]{{\prescript{{#1}}{}{\mathbf{\tilde{H}}}_{#2}}}

\newcommand{\cov}{\mathbf{P}}
\newcommand{\Q}{\mathbf{Q}}
\newcommand{\N}{\mathbf{N}}

\newcommand{\F}{\mathbf{F}}

\newcommand{\Xhat}{\mathbf{\hat{X}}}
\newcommand{\err}{\epsilon}
\newcommand{\errG}{\eta}

\newcommand\kpred{{k+1 \mid k}}
\newcommand\kprior{{k \mid k}}
\newcommand\kest{{k+1 \mid k+1}}
\newcommand\kcurr{{k}}
\newcommand\knext{{k+1}}
\newcommand\knextgivenl{{k+1 \mid l}}

\newcommand{\expectation}[1]{\mathbb{E}\hm{[} {#1} \hm{]}}

\newcommand\gravity[1]{{\prescript{{#1}}{}{\mathbf{g}}}}
\newcommand\Transform[2]{{\prescript{{#1}}{}{\mathbf{H}}_{#2}}}

\newcommand\Xtwist[2]{{\prescript{{#1}}{}{\mathbf{X}}_{#2}}}

\newcommand\PointTilde[2]{{\prescript{{#1}}{}{\mathbf{\tilde{p}}}_{#2}}}
\newcommand\Point[2]{{\prescript{{#1}}{}{\mathbf{p}}_{#2}}}
\newcommand\Rot[2]{{\prescript{{#1}}{}{\mathbf{R}}_{#2}}}
\newcommand\Pos[2]{{\prescript{{#1}}{}{\mathbf{o}}_{#2}}}
\newcommand\oDot[2]{{\prescript{{#1}}{}{\mathbf{\dot{o}}}_{#2}}}
\newcommand\oDoubleDot[2]{{\prescript{{#1}}{}{\mathbf{\ddot{o}}}_{#2}}}

\newcommand\jointSubspaceLeftTriv[2]{{\prescript{{#2}}{}{\mathbf{s}}_{#1, #2}}}
\newcommand\jointSubspaceRightTriv[2]{{\prescript{{#1}}{}{\mathbf{s}}_{#1, #2}}}
\newcommand\jointSubspaceMixedTriv[2]{{\prescript{{#2[#1]}}{}{\mathbf{s}}_{#1, #2}}}

\newcommand\relativeJacobianLeftTrivLinIn[2]{{\prescript{{#2}}{}{\mathbf{S}}^{\text{lin}}_{#2, #1}}}

\newcommand\relativeJacobianLeftTrivIn[2]{{\prescript{{#2}}{}{\mathbf{S}}_{#2, #1}}}

\newcommand\relativeJacobianLeftTrivAng[2]{{\prescript{{#2}}{}{\mathbf{S}}^{\text{ang}}_{#1, #2}}}
\newcommand\relativeJacobianLeftTriv[2]{{\prescript{{#2}}{}{\mathbf{S}}_{#1, #2}}}

\newcommand\twist{\textbf{v}}
\newcommand\twistLeftTriv[2]{{\prescript{{#2}}{}{\textbf{v}}_{#1, #2}}}
\newcommand\twistRightTriv[2]{{\prescript{{#1}}{}{\textbf{v}}_{#1, #2}}}
\newcommand\twistMixedTriv[2]{{\prescript{{#2[#1]}}{}{\textbf{v}}_{#1, #2}}}

\newcommand\vLeftTriv[2]{{\prescript{{#2}}{}{\bm{v}}_{#1, #2}}}
\newcommand\vRightTriv[2]{{\prescript{{#1}}{}{\bm{v}}_{#1, #2}}}
\newcommand\vMixedTriv[2]{{\prescript{{#2[#1]}}{}{\bm{v}}_{#1, #2}}}

\newcommand\omegaLeftTriv[2]{{\prescript{{#2}}{}{\omega}_{#1, #2}}}
\newcommand\omegaRightTriv[2]{{\prescript{{#1}}{}{\omega}_{#1, #2}}}

\newcommand\sensorAcc[2]{{\alpha}_{#1, #2}}
\newcommand\properAcc[2]{{\alpha}^g_{#1, #2}}
\newcommand\twistDotMixedTriv[2]{{\prescript{{#2[#1]}}{}{\dot{\textbf{v}}}_{#1, #2}}}
\newcommand\omegaDotLeftTriv[2]{{\prescript{{#2}}{}{\dot{\omega}}_{#1, #2}}}

\newcommand\Rdot[2]{{\prescript{{#1}}{}{\mathbf{\dot{R}}}_{#2}}}
\newcommand\Hdot[2]{{\prescript{{#1}}{}{\mathbf{\dot{H}}}_{#2}}}
\newcommand\Rhat[2]{{\prescript{{#1}}{}{\mathbf{\hat{R}}}_{#2}}}
\newcommand\Hhat[2]{{\prescript{{#1}}{}{\mathbf{\hat{H}}}_{#2}}}
\newcommand\Rtilde[2]{{\prescript{{#1}}{}{\mathbf{\tilde{R}}}_{#2}}}

\newcommand\jointPos{\mathbf{{s}}}
\newcommand\jointVel{\mathbf{\dot{s}}}
\newcommand\jointAcc{\mathbf{\ddot{s}}}

\newcommand\freeFloatingJacobian[2]{{\mathbf{J}}_{{#1, #2}}}

\newcommand\freeFloatingJacobianAng[2]{{\mathbf{J}}^{\text{ang}}_{{#1, #2}}}
\newcommand\leftTrivializedJacobian[2]{{\mathbf{J}}_{{#1, #2}/#2}}

\newcommand\mixedTrivializedJacobian[2]{{\mathbf{J}}_{{#1, #2}/#2[#1]}}

\newcommand\force[1]{{}_{{#1}} \bm{f}}
\newcommand\wrenchExt[1]{{}_{{#1}} \mathbf{f}^{\mathbf{x}}}
\newcommand\forceExt[1]{{}_{{#1}} \bm{f}^{\mathbf{x}}}
\newcommand\torqueExt[1]{{}_{{#1}} {\tau}^{\mathbf{x}}} 

\title{State Estimation for Human Motion and Humanoid Locomotion}

\subtitle{Using the Latex thesis template}

\author{Your name}

\dept{DIBRIS}

\university{University of Genova}

\degreetitle{Doctor of Philosophy}

\subject{LaTeX} \keywords{{LaTeX} {PhD Thesis} {Bioengineering and Robotics} {University of
Genova}}

\begin{document}


\thispagestyle{empty}
\begin{figure}[h!]
 \centering
 \includegraphics[scale=0.20]{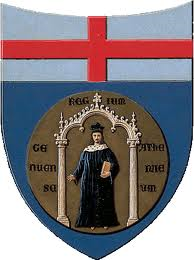} 
	\begin{center} 
		\Large
		{\textsc{University of Genova}}\\
		  \vspace{0.5em}
		  \large
	         \textsc{PhD Program in Bioengineering and Robotics}\\

	\end{center}
\end{figure}

\begin{center} 
	

		\LARGE
		\textbf{State Estimation for Human Motion and Humanoid Locomotion} \\
\end{center}

 	\begin{center} 
		\vspace{0.5em}
		\textbf{Prashanth Ramadoss}\\
		\vspace{0.5em}

		\normalsize
		Thesis submitted for the degree of  \\
		\textit{Doctor of Philosophy} ($34^\circ$ cycle) \\
	\vspace{1cm}	
		\normalsize
		\textbf{Supervisor:} Daniele Pucci \\
		\textbf{Co-Advisors:} Silvio Traversaro, Stefano Dafarra \\
		\textbf{Head of the PhD Program}: Giorgio Cannata
	\end{center}

\vfill

	\vspace{0.7em} 
		\noindent {\textbf{\textit{Thesis Jury and Reviewers*:}}}			
	\\
	\noindent {Joan Sola*} \hfill {Tenured Scientist at IRI CSIC-UPC, Barcelona, Spain}	 \\
	\noindent {Mehdi Benallegue*} \hfill {Permanent Researcher at AIST, Tsubaka, Japan} \\
	\noindent {Olivier Stasse} \hfill {Senior Researcher at LAAS-CNRS, Toulouse, France} \\
	\noindent {Lorenzo Rosasco} \hfill {Professor at DIBRIS-UNIGE, Genova, Italy} \\
	\noindent {Giorgio Cannata} \hfill {Professor at DIBRIS-UNIGE, Genova, Italy} \\


\begin{figure}[h!]
\centering
\begin{subfigure}{0.45 \textwidth}
\centering
\captionsetup{justification=centering}
  \includegraphics[scale=0.3]{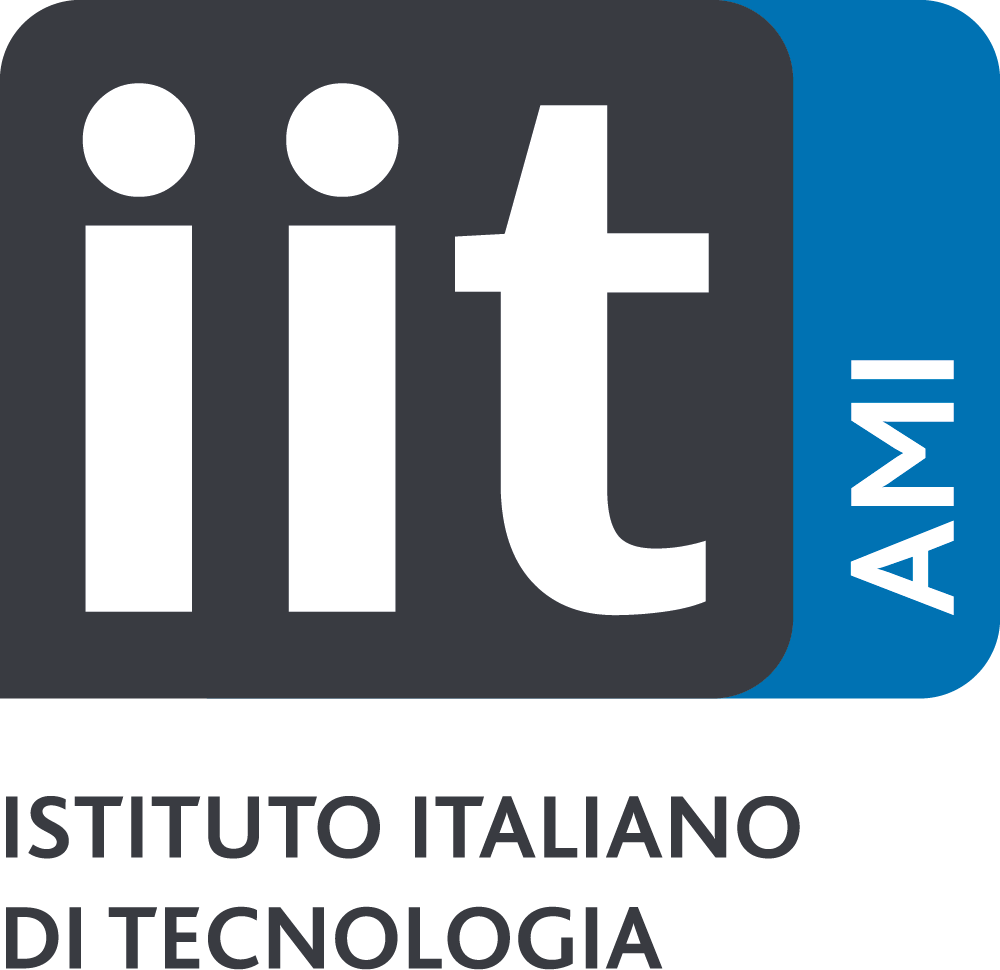}
  \caption*{ Artificial and Mechanical Intelligence, \\ Fondazione Istituto Italiano di Tecnologia}
\end{subfigure}
\begin{subfigure}{0.45 \textwidth}
 \centering
 \captionsetup{justification=centering}
 \vspace{1.5em}
\includegraphics[scale=0.3]{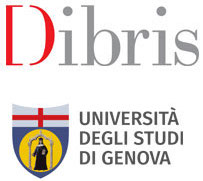}
\caption*{ Dipartimento di Informatica, Bioingegneria, Robotica e Ingegneria dei Sistemi, \\  Universit\`a di Genova}
\end{subfigure}
\end{figure}

\begin{center}
    \normalsize
    Genova, Italy \\
    2022
\end{center}


\frontmatter



\begin{dedication} 


\end{dedication}

\begin{abstract}
The future where the industrial shop-floors witness humans and robots working in unison and the domestic households becoming a shared space for both these agents is not very far.
The scientific community has been accelerating towards that future by extending their research efforts in human-robot interaction towards human-robot collaboration.
It is possible that the anthropomorphic nature of the humanoid robots could deem the most suitable for such collaborations in semi-structured, human-centered environments.
Wearable sensing technologies for human agents and efficient human-aware control strategies for the humanoid robot will be key in achieving a seamless human-humanoid collaboration.
This is where reliable state estimation strategies become crucial in making sense of the information coming from multiple distributed sensors attached to the human and those on the robot to augment the feedback controllers designed for the humanoid robot to aid their human counterparts.

In this context, this thesis investigates the theory of Lie groups for designing state estimation techniques aimed towards humanoid locomotion and human motion estimation.
The abstract nature of Lie theory provides a unified approach to handle the three-dimensional machinery and the complex geometry required for modeling free-floating, highly articulated multi-body systems. 
It enables suitably appropriate methods to perform rigorous calculus over complex nonlinear spaces and to handle the notion of uncertainties in such spaces, which are important for an estimator design. \looseness=-1

Methods for loosely-coupled and tightly-coupled sensor fusion for floating base estimation of a humanoid robot are presented through the theory of \emph{averaging} and \emph{filtering} on Lie groups.
The problem of human motion estimation through wearable sensing technologies is tackled through a combination of dynamical systems' theory-based Inverse Kinematics and \emph{filtering} on Lie groups, demonstrated to be directly applicable also for humanoid state estimation.
Experimental validations of the estimators for humanoid base estimation and human motion estimation have been carried out on simulated datasets and datasets collected from real-world experiments conducted on the iCub humanoid robot and Xsens Motion capture technology, respectively.


\end{abstract}


\begin{acknowledgements}
\emph{On the shoulders of giants, I stand}; (paraphrasing Sir Isaac Newton).

I started my doctoral program with the aspirations of becoming someone capable of conducting independent research. 
I began this pursuit in na\"ivety with only intellectual curiosity and a deep-rooted interest in robotics.
The last three years, however hard and challenging, have been an incredibly formative and fruitfully molding journey owing to the support of my colleagues, friends, and family.

My first and foremost thanks are to my supervisor, Daniele Pucci, for providing me the opportunity to conduct research in the \emph{Artifical and Mechanical Intelligence} laboratory.
My efforts in the last three years have culminated in this research work thanks to his guidance and support.
I am grateful to him for enabling me with the freedom to explore the field while providing his valuable counsel at critical points of my research term which has helped me grow both personally and professionally. \looseness=-1

I can't thank Silvio enough, who has not only been my immediate scientific advisor during my Ph.D. years and earlier, but also a friend and a mentor I look up to for several reasons. 
I truly appreciate all the time and effort you have taken to guide me on several topics.
Your selfless helping and teaching nature never cease to inspire me. 

I must thank Stefano, the scrum master of our Telexistence team, who amidst handling the huge responsibility of managing our team proceedings for the ANA Avatar XPrize competition, has always managed to find time to participate in discussions and guide me in research directions for this thesis. 

My sincere thanks to the reviewers, Joan Sola, and Mehdi Benallegue, for their time and effort in accepting to review my thesis.
I must not forget to thank Lorenzo Natale, Marco Maggiali, Ugo Pattacini, Vadim Tikhanoff, and Alberto Parmiggiani for their valuable feedback during the yearly evaluations of my Ph.D. research.

I take this moment to express my thanks also to Naveen Kuppuswamy and Dibyajyoti Das, my mentors from the past, who have been important sources of inspiration in my career.
I appreciate the moral support and counsel provided by my colleagues/friends from the past: Aiko, Anand, Yue, Jorhabib, and importantly, Nuno who is a colleague also in the present. \looseness=-1

My hearty thanks to Yeshasvi, Giulio, Giuseppe, Hosam, Lorenzo, Cisco, Punith, Fabio B., and Emilio who have been kind enough to help me in developing and conducting several necessary aspects of research during my Ph.D. years.
A special mention to Nicola Piga for all the insightful brainstorming discussions we have had regarding state estimation and robotics in general.
I also thank other members (current and past) of our lab: Ines, Paolo, Diego, Kourosh, Raffaello, Antonello, Riccardo, Gabriele, Claudia, Giovanni, Carlotta, Italo, Affaf, Gianluca, Enrico, Tong, Milad, Liana, Anqing, Prajval, Venus, Valentino, Luis, Luca F., Luca T., Ahmad and Mohamed, with whom I have shared many valuable, enriching and memorable experiences.

I am sincerely thankful for the administrative support from Marta Caracalli, Lucia Betro, Francesco Roda and Valentina Scanarotti.

My special thanks to all the dear friends I have made on the football field and big cheers for all those precious football moments. \emph{Pirates of iCub}, for the win!

I would like to extend my sincerest thanks to all the members of the \emph{iCub Facility} for creating a wonderful infrastructure that allows people to perform effective research at a cutting-edge level.
My gratitude will only be complete by thanking once again the founding members of the \emph{Dynamic Interaction Control} lab, now renamed as the \emph{Artificial and Mechanical Intelligence} lab.
I am happy being a part of this team amidst many brilliant and gracious minds with whom almost every other day is a day to learn something new and lasting. \looseness=-1

Finally, I would like to express my everlasting gratitude to my dearest friends and family.
Any virtues in me are from the gifts you have bestowed upon me through your unique and unconditionally loving ways. \looseness=-1

\end{acknowledgements}

\tableofcontents

\listoffigures

\listoftables
\newpage

\mainmatter

\chapter*{Prologue}

\addcontentsline{toc}{chapter}{Prologue} 
\chaptermark{Prologue}
\lhead{\leftmark}

Humans have marveled upon multitudes of science fiction depictions portraying cyber-physical beings that would resemble our anthropomorphic structures but have a distinct bio-mechanical schema, either aiding us towards a utopian future or haunting us in a dystopian alternative.
We are in a time where these cyber-physical systems have already long-crawled out of our imaginations to coexist with us in our physical reality during this post-Industrial Revolution era. \looseness=-1

These mechanical systems, we now call \emph{robots}, were initially developed on a large commercial scale by \emph{Unimation}, in the 1960s, for reducing human labor along the assembly lines, in the manufacturing process of the automotive industry.
However, the recent decades have seen a myriad of venues that can be beneficial through the application of cutting-edge robotic technologies, from medical sciences and healthcare to disaster-response and interplanetary operations.
The term robot would now imply physically and cognitively embodied machines capable of performing tasks autonomously, either mundane or super-human.
Our next step seems to be towards a future where robots collaborate with humans in a productive and ethical manner, enriching our societal lifestyles.

In the 21st century, a field of study in robotics is focused on developing technologies for Human-Robot Collaboration (HRC) that will allow the robots to operate safely through effective interaction, cooperation, and collaboration with humans in an industrial or a domestic setting.
While robots of several morphologies exist, humanoid robots that have a similarity with our anthropomorphic structure may have the advantage of being well-suited for human-centered environments and might also be useful in complementing the studies in human biomechanics and ergonomics.

Despite the advantages of being well suited for human-robot collaboration, humanoid robots are hard, in the sense that formulating control strategies for achieving efficient locomotion for a humanoid robot still remains an active and challenging research problem.
A humanoid robot in collaboration with a human counterpart should not only regulate its own balance and locomotion but also actively control the interaction forces that it might want to exchange with the human in order to perform a task.
Further, such robots do not usually remain fixed to their environment. 
Their capability to freely move and interact with their surroundings classifies them as \emph{floating base systems} contrary to their fixed-base counterparts, where the term floating base reflects their freely moving or free-floating nature.
In order to fully describe the motion of the robot over time, it is necessary to track some physical quantities such as its position and orientation (collectively called as \emph{pose}) in its environment along with its velocity.
These properties referring to the kinematic description of the robot constitute its \emph{state}.
More generally, the state may contain any variable that describes a desired behavior of the system. 
In a collaborative scenario, the robot should thus be aware of its own state and that of its human counterparts in order to regulate itself through feedback-control strategies. \looseness=-1

Owing to their highly-articulated and free-floating nature, the underlying geometry for modeling both humans and humanoid robots is not straightforward.
As mentioned already, the robot state is partly composed of its orientation (often called as \emph{rotation}) or its pose, which are geometric objects evolving over a space different from the well-known Euclidean vector space and thus, require careful attention to enable a proper representation.
Further, the state, of such a \emph{floating-base} system, reflecting their geometric configuration, usually cannot be measured directly through sensors but needs to be estimated.
With the recent advances in sensing technologies, it has become a common approach to equip both the robots and humans with multiple, distributed sensors to measure quantities that might be relevant in perceiving their state; not only kinematic and dynamic states but also physiological states in the case of humans.
Such an approach by itself poses deeper questions of how to fuse multiple sources of information to obtain a reliable estimate.
Appropriate state estimation strategies, relying on a proper representation of both the state to be estimated and the observations measured by the multiple sensors along with their associated uncertainties, thus, become crucial for both humans and humanoid robots.  \looseness=-1

In this thesis, we apply state estimation techniques for humanoid locomotion and human motion estimation using the theory of Lie groups focusing mainly on the kinematics reconstruction for the involved agents. 
The theory of Lie groups enables the construction of estimators that capture the underlying geometry of these agents effectively while providing insights, through its abstract nature, into the three-dimensional machinery that a roboticist might use on a daily-life basis to make the robots move.
We explore estimator designs employing \emph{averaging} and \emph{filtering}  methods for an appropriate sensor fusion in complicated state and measurement spaces for achieving rigorous nonlinear state estimation.
This research work has been carried out during my tenure as a Ph.D. candidate in the \emph{Artificial and Mechanical Intelligence} laboratory at the \emph{Italian Institute of Technology} in Genova, Italy.
The doctoral program has been carried out in accordance with the requirements of \emph{University
of Genoa, Italy} in order to obtain a Ph.D. title.\looseness=-1

\subsection*{}
\noindent This thesis is divided into two parts.
\subsubsection*{Part I: Background and Thesis Context}
\noindent This part provides an introduction to the fundamental concepts required for understanding the developments made in this thesis and reviews the state-of-the-art methods relevant to the context of the thesis. \looseness=-1
\begin{itemize}
    \item Chapter \ref{chap:intro} begins by motivating the need for state estimation in the application context of this thesis.
    It also introduces, in brief, the underlying technologies used for implementing the algorithms presented in this thesis.
    \item Chapter \ref{chap:estimation} provides a highly condensed account of matrix Lie groups and their application within the theoretical framework of estimation on Lie groups.
    The notion of uncertainty over matrix Lie groups is introduced with the concept of Concentrated Gaussian Distributions (CGD).
    This chapter provides a self-contained derivation of Extended Kalman Filtering on matrix Lie groups using the concept of CGD. 
    Appendix \ref{appendix:chapter02:examples-matrix-lie-groups} provides examples of matrix Lie groups that are most commonly used in robotics. \looseness=-1
    \item Chapter \ref{chap:modeling} presents a unified modeling approach for representing human and humanoid robot systems within the framework of rigid multibody systems.
    \item Chapter \ref{chap:context} provides a detailed literature review on the state-of-the-art methods used in state estimation for humanoid robots and human motion estimation.
    In order to gain a broader perspective on the landscape of intersecting domains, the review also surveys the body of literature from the SLAM and the legged robots community.
\end{itemize}

\subsubsection*{Part II: Thesis Contributions}
\noindent This part provides a detailed account of the contributions from the research work carried out for developing this thesis.
\begin{itemize}
    \item Chapter \ref{chap:floating-base-swa} presents a loosely-coupled sensor fusion approach for the problem of floating base estimation of a humanoid robot.
    In particular, we exploit the theory of averaging on matrix Lie groups to perform the fusion of rotations and poses.
    We apply the averaging approach to formulate contact-aided, kinematic-inertial odometry and demonstrate its extension for localization in a known environment.
    Such an approach with its foundation based on a powerful toolkit of Lie groups is mainly motivated to provide practitioners with a simple-yet-effective sensor fusion approach for humanoid robots that maintains a scalable algorithmic structure to combine multimodal measurements, while providing reliable estimates for closed-loop control.
    \item Chapter \ref{chap:floating-base-diligent-kio} presents the development of a tightly-coupled sensor fusion approach for the problem of floating base estimation of a humanoid robot.
    For applications demanding fully-coupled inference of relevant states from multimodal measurements, we motivate ourselves in formulating a filtering based estimation approach that generalizes the current standard approach of quaternion based EKF for humanoid base estimation using the theory of matrix Lie groups.    
    The estimator design is based on extended Kalman filtering on Lie groups with the consideration that both the states and measurements evolve over distinct matrix Lie groups.
    The design is mainly focused on formulating a proprioceptive floating base estimator for a humanoid robot, however, its straightforward extension to absolute humanoid localization is also discussed.
    Further, multiple variants of the proposed estimator are analyzed as a consequence of the choice of Lie group error and the time representation of the system dynamics used for designing the estimator.  \looseness=-1
    \item Chapter \ref{chap:human-motion} demonstrates a proprioceptive estimator design for human motion estimation exploiting the use of wearable sensing technologies such as a motion capture suit with distributed inertial sensors and sensorized shoes attached with force-torque sensors. 
    This work is motivated from the standpoint of achieving full-body motion tracking in the absence of position sensors while being able to meet the computational requirements for time-critical applications.
    The estimator design uses a cascading architecture of a dynamical Inverse Kinematics optimization followed by invariant extended Kalman filtering on Lie groups to solve joint state estimation and floating base estimation using inertial measurements and ground reaction force measurements.
    A Center of Pressure based contact detection strategy is used assuming a simplified geometry of the foot which allows accounting for dynamic heel-to-toe motions of the human.
    This approach is demonstrated to be applicable for both human and humanoid robot estimation.
\end{itemize}

\section*{Research publications}
\nobibliography*

\noindent The findings from the research work within the scope of this thesis have been (or will be) disseminated as peer-reviewed research publications.
Supplementary material including, video presentation and software repository for the associated publication, are listed wherever available, aiming for the reproducibility of conducted research.

\noindent The contents of Chapter \ref{chap:floating-base-diligent-kio} appear in a conference publication,
\begin{leftbar}
\begin{quote}
    \noindent{P. Ramadoss, G. Romualdi, S. Dafarra, F. J. Andrade Chavez, S. Traversaro and D. Pucci, "DILIGENT-KIO: A Proprioceptive Base Estimator for Humanoid Robots using Extended Kalman Filtering on Matrix Lie Groups," \textit{2021 IEEE International Conference on Robotics and Automation (ICRA)}, 2021, pp. 2904-2910, doi: 10.1109/ICRA48506.2021.9561248.}\newline
    \noindent {\small \textbf{Video:}  \href{https://www.youtube.com/watch?v=CaEZvbR9ZcA}{click here}}, {\small \textbf{Github:}  \href{https://github.com/ami-iit/ramadoss-2021-icra-proprioceptive_base_estimator}{click here}}.
\end{quote}
\end{leftbar}

\noindent The dissemination effort for a few contributions of this thesis has been planned after the submission of this PhD thesis.
\noindent The extended contents of Chapter \ref{chap:floating-base-diligent-kio} will be combined for a journal publication tentatively titled as, \looseness=-1
\begin{leftbar}
\begin{quote}
    \noindent{Prashanth Ramadoss, Stefano Dafarra, Silvio Traversaro, and Daniele Pucci. An Experimental Comparison of Floating Base Estimators for Humanoid Robots with Flat Feet. \textit{IEEE Robotics and Automation Letters (2022)}, (Submitted, Under Review)}.
    \newline
    {\small \textbf{Github:}  \href{https://github.com/ami-iit/paper_ramadoss-2022-ral-humanoid-base-estimation}{click here}, \small \textbf{Preprint:}  \href{https://arxiv.org/abs/2205.07765}{click here}}.
\end{quote}
\end{leftbar}
\noindent The contents of Chapter \ref{chap:human-motion} will be submitted for a journal publication tentatively titled as,
\begin{leftbar}
\begin{quote}
    \noindent{Prashanth Ramadoss, Lorenzo Rapetti, Yeshasvi Tirupachuri, Stefano Dafarra, Silvio Traversaro, and Daniele Pucci.Whole-Body Human Kinematics Estimation using Dynamical Inverse Kinematics and Contact-Aided Lie Group Kalman Filter. \textit{IEEE Robotics and Automation Letters (2022)}, (To be Submitted)}.
    \newline
    {\small \textbf{Preprint:}  \href{https://arxiv.org/abs/2205.07835}{click here}}.
\end{quote}
\end{leftbar}

\noindent Apart from the publications directly related to the theoretical framework of this thesis, contributions have also been made in the dissemination of other closely related research.

The following journal publication describes a framework for dynamics estimation of a human through a cascading architecture of simplified models capturing the centroidal dynamics and complete models describing the whole body dynamics of the human through the use of wearable sensing technologies.
I have contributed in this dissemination by aiding the first author with the validation, software testing and experimental analysis for the proposed theoretical formulation.
The dynamics estimation becomes fundamental in human-robot collaboration scenarios and relies heavily on the kinematic state of the human, thus having a direct implications from the developments of this thesis. \looseness=-1
\begin{leftbar}
\begin{quote}
    \noindent{Yeshasvi Tirupachuri, \textbf{Prashanth Ramadoss}, Lorenzo Rapetti, Claudia Latella, Kourosh Darvish, Silvio Traversaro, and Daniele Pucci. Online non-collocated estimation of payload and articular stress for real-time human ergonomy assessment. \textit{IEEE Access}, 9:123260-123279, 2021}\newline
    \noindent {\small \textbf{Video:}  \href{https://www.youtube.com/watch?v=oPTiWrlKRJ0}{click here}}, {\small \textbf{Github:}  \href{https://github.com/ami-iit/tirupachuri-2021-access-estimation_payload_stresses}{click here}}
\end{quote}
\end{leftbar}

The journal publication mentioned below, reports on a benchmarking evaluation of different control strategies for achieving locomotion on a humanoid robot. 
My contribution towards this publication was concerned with the development of a simple state-estimation sub-module to close the loop for the feedback controllers using estimates of robot pose and velocity  
Further, I supported the first author in the experimental procedures for the proposed architectures.
The context of this publication is a motivating application scenario for the theoretical developments of this thesis.
\begin{leftbar}
\begin{quote}
    \noindent{Giulio Romualdi, Stefano Dafarra, Yue Hu, \textbf{Prashanth Ramadoss}, Francisco Javier Andrade Chazvez, Silvio Traversaro, and Daniele Pucci. A benchmarking of dcm-based architectures for position, velocity and torque-controlled humanoid robots. \textit{International Journal of Humanoid Robotics}, 17(01):1950034, 2020}
\end{quote}
\end{leftbar}

\vspace{5mm}
\noindent The four subsequent chapters begin by building the foundational context for the developments of this thesis. 
The last three chapters report, in detail, the contributions made by this thesis.

\part{Background and Thesis Context} 

\chapter{Introduction}
\label{chap:intro}
\section{Unified Control Architecture for Human-Humanoid Collaboration}

Among many innovation challenges foreseen by the European robotics community in employing robotic technologies for the betterment of societal well-being and environmental aspects, interaction technologies that allow for a safe, human-compatible physical interaction and collaboration over a range of tasks are one of the expected outcomes that would create significant economic, social and environmental value (\cite{zillner2020strategic}).
In this context, effective human-robot collaboration is remarked as a key technology enabler that will allow a broad deployment of robots to augment human capabilities across several sectors, 
"\textit{Human-robot collaboration will require high-level decision making to be part of the safety concept for robots, e.g. by predicting the movement of a human in direct interaction. Safety will no longer solely be accomplished by a dedicated layer but will rely on various types of sensors for perception, high-level algorithms for the interpretation of the sensor data, and trustworthy decision-making to act accordingly. Distributing the safety to different layers will make robots more flexible and reactive while maintaining an acceptable level of risk}  (\cite{zillner2020strategic})." \looseness=-1
The demand for human-robot collaborative scenarios in several sectors such as manufacturing, domestic assistance, and healthcare require to endow robots with sensing capabilities and control strategies that allow for a partner-aware robot collaboration with the human.
The success of a safe and effective collaborative task, however, relies heavily on the robot being able to predict complex motions of the human and respond accordingly to improve human ergonomics in such a task (\cite{latella2020human}).

\begin{figure}[!h]
\centering
\includegraphics[width=\textwidth]{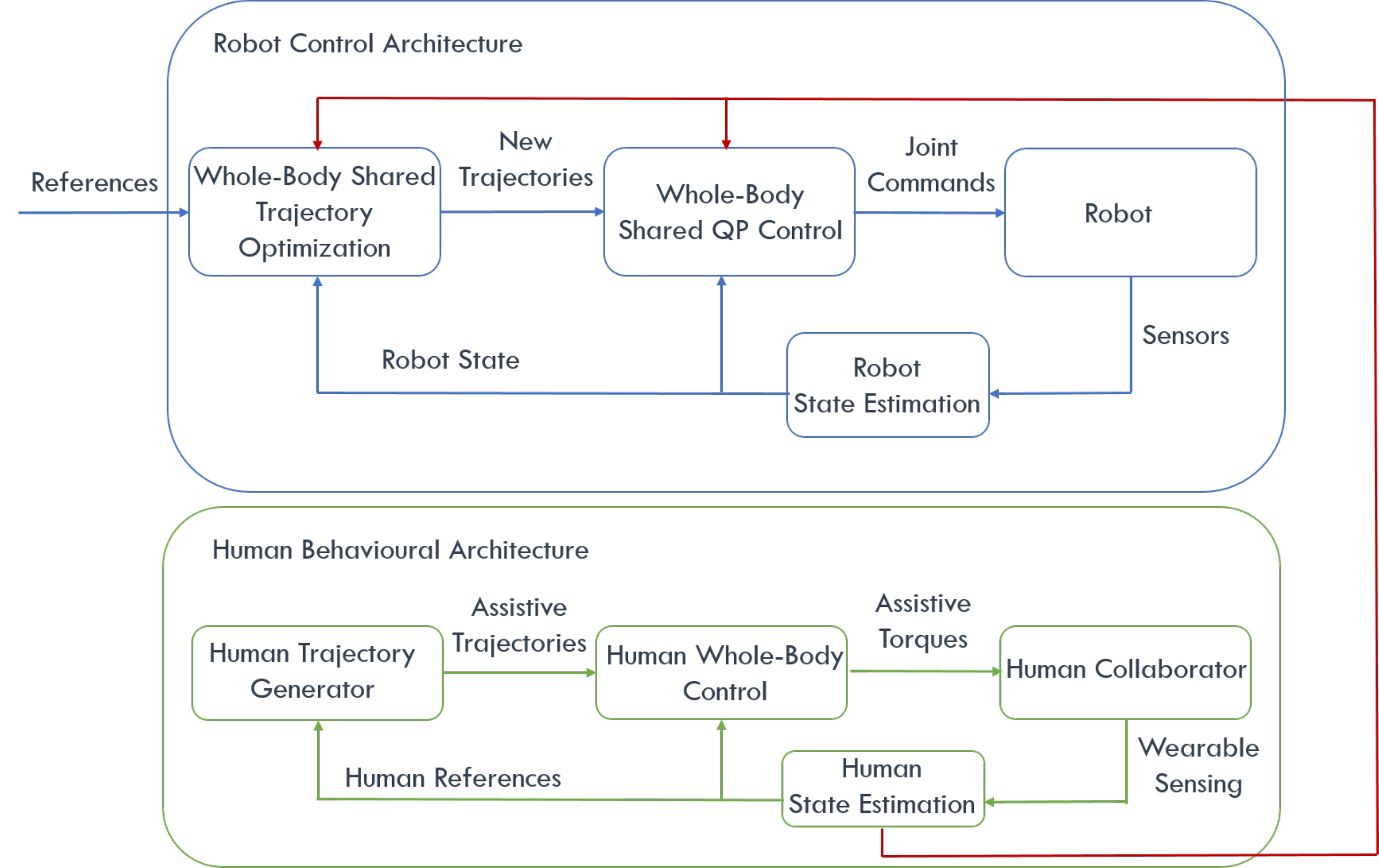}
\caption{A unified control architecture for human-humanoid collaboration composed of robot control architecture and human behavioural architecture components.}
\label{fig:chap:intro:hrc-arch}
\end{figure}

The development of an interaction technology that allows for a reliable physical Human-Robot Interaction (pHRI) can be benefited through the realization of a unifying control architecture for human-humanoid collaboration that will account for the several complexities and challenges within the problem.
Such a high-level collaborative control architecture is depicted in Figure \ref{fig:chap:intro:hrc-arch} combining cascaded and nested layers of trajectory optimization, control, and estimation for both robots and humans.
This architecture combines the capabilities of two distinct components namely \emph{robot control} and \emph{human behaviour architectures}, each of which can be seen as engines driving the robot motions and assisting human actions respectively to bring a \emph{collaborative task} into fruition.

The robot control architecture observed as a standalone component is composed of three main high-level layers communicating with the robot in order to generate the desired motion.
These layers are called as \emph{whole-body shared trajectory optimization}, \emph{whole-body shared Quadratic Programming (QP) control} and \emph{robot state estimation}.
The adjective \emph{shared} comes into play when the robot is in active interaction with another agent, either a robot or a human, and needs to regulate its motion considering the agent's interactions (\cite{rapetti2021shared}).
The trajectory optimization layer is in charge of generating desired trajectories, such as end-effector and Center of Mass (CoM) trajectories, for the robot by accounting for any reference trajectories as inputs.
The whole-body QP control layer is then used to produce the necessary joint commands to control the robot's motion by exploiting its known model to guarantee the tracking of desired high-level trajectories.
Oftentimes, it is also necessary to introduce an intermediate layer known as \emph{simplified model control} between the trajectory optimization and whole-body control layers to ensure the tracking of trajectories generated by the former layer in a computationally efficient manner with the help of reduced-order models that approximate the full model of the robot effectively (\cite{romualdi2020benchmarking}).
Both the trajectory optimization and whole-body control layers rely on continuous feedback from the robot to generate, regulate and track the motions on the robot.
The state estimation layer is then used to make sense of information obtained from the multiple sensors on the robot to estimate the robot state which is then passed as feedback to the other layers. \looseness=-1

The human behavioural architecture is composed of three layers, namely, \emph{human state estimation}, \emph{human trajectory generator} and \emph{human whole-body control}, acting in synergy with the human collaborator equipped with wearable sensing and actuation technologies.
The sensing technologies are used to retrieve some useful information from the human agent such as quantities describing the kinematic, dynamic, and physiological state required to understand and predict human motion and fatigue.
The \emph{human state estimation} layer is then used to determine these states of the human which are usually not directly measurable and can only be estimated.
The human state is then passed as feedback to a trajectory generator layer that is used to generate \emph{recommended} high-level trajectories called as \emph{assistive trajectories}, such as postures and hand trajectories, required for the completion of the collaborative task.
The whole-body control layer then uses the \emph{assistive trajectories} to compute \emph{assistive torques} by optimizing for human ergonomics. These assistive torques then act as recommended stress exertions for the human during the task.  
Wearable actuation technologies are used to actively assist or guide human actions to effectively coordinate their interactions with their partner or the environment.
While the human behavioral model provides active assistance for the human agent to augment their capabilities in the collaborative task, the human state feedback is also passed on to the robot control architecture to allow the robot to regulate itself suitably, for better human working conditions.


\begin{figure}[!h]
\centering
\includegraphics[scale=0.45]{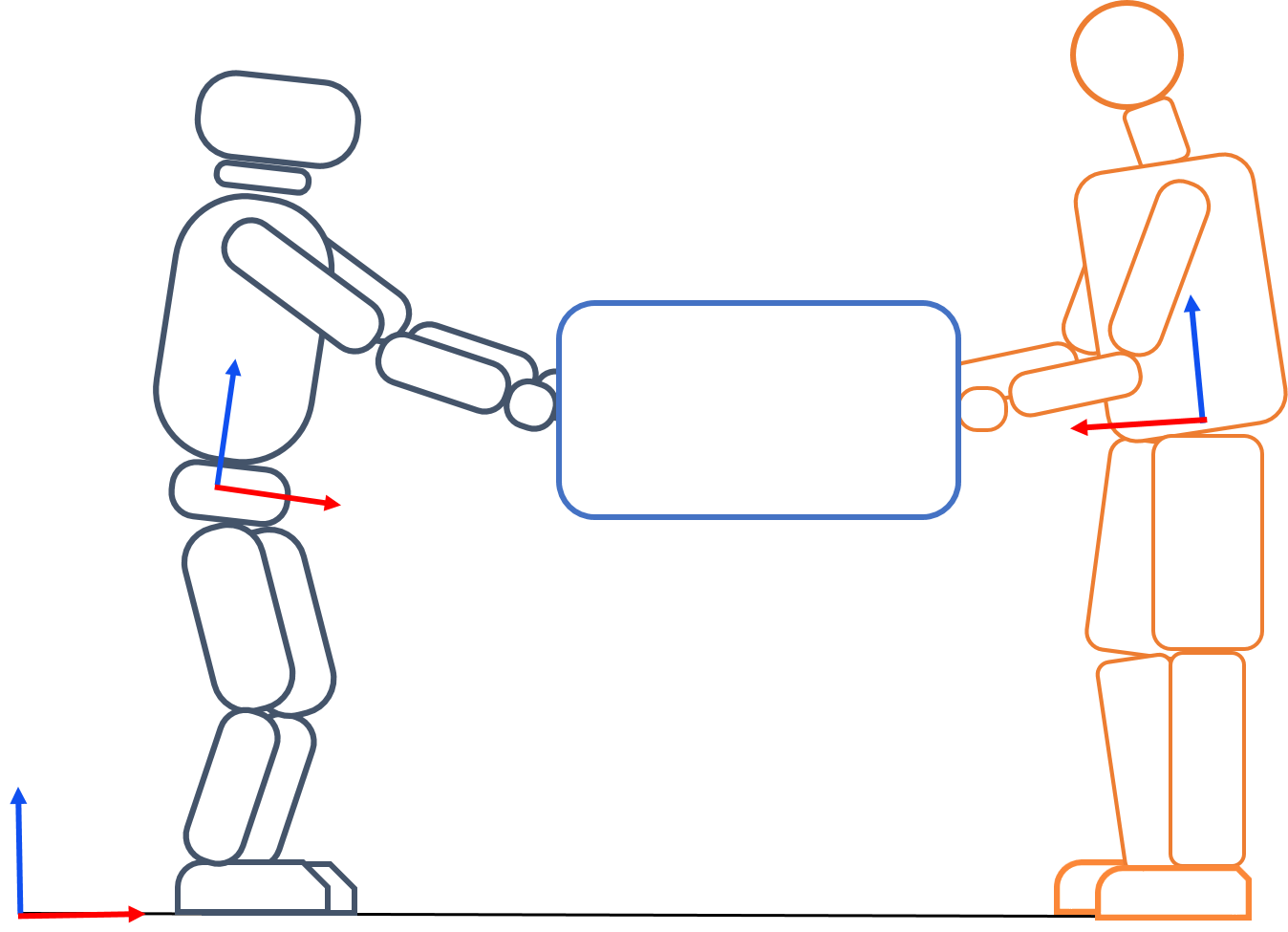}
\caption{Example scenario where a human subject performs collaborative work with the robot. \looseness=-1}
\label{fig:chap:intro:hri}
\end{figure}

In this thesis, we narrow our focus onto the \emph{state estimation algorithms for human motion and humanoid locomotion} that play an important role of providing the necessary feedback for the control layers in the unified control architecture.
We motivate the context of the state estimation algorithms developed in this thesis along the basis of a primary application scenario depicted in Figure \ref{fig:chap:intro:hri}.
The figure depicts a robot involved in active collaboration with a human carrying a load to the desired location, for instance, in a warehouse environment.
The first and foremost objective in this scenario is to augment feedback controllers developed for locomotion and collaborative strategies on the humanoid robot. 
To be robust towards model uncertainties and unpredictable environments, a multi-modal sensor fusion-based floating base estimation becomes crucial. 
Moreover, the importance of the floating base estimation for the humanoid robot and whole-body motion estimation for the human becomes directly apparent from such a scenario where the robot needs to understand both its system configuration and that of the human counterpart in an absolute reference frame within the warehouse.
This is necessary to augment the planning and control strategies designed for human safety and ergonomics.
This can, in turn, lead to a reliable and shared floating base estimation algorithm that can allow a proprioceptive terrain mapping of the environment local to these agents which might be necessary for efficient navigation in cases where the perception systems all have an occluded field of view and become obsolete for mapping or landmark association.
In particular, we take a few preliminary steps to investigate state estimation strategies for a humanoid robot and a human subject equipped with distributed wearable sensors, aimed towards effective human-humanoid collaboration. \looseness=-1

\section{Enabling Technologies}
In this section, we briefly describe the technological resources used for the experimental validation of the thesis developments.
We first introduce the iCub humanoid robot on which the humanoid state estimation algorithms are tested, followed by a description of the wearable sensing technologies constituting a whole-body human perception system with which the human motion estimation experiments are verified. \looseness=-1
\subsection{Description of iCub}
\label{sec:chap:intro:iCub}
\begin{figure}[!t]
\centering
\includegraphics{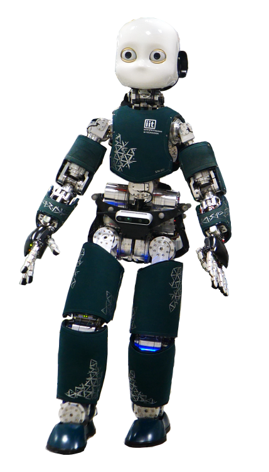}
\caption{iCub v2.5, additionally equipped with a Realsense camera and a Vicon marker mount on its waist, making them base-collocated sensors. \looseness=-1}
\label{fig:chap:intro:icub}
\end{figure}

The iCub humanoid robot is a child-size robot, born out of the EU funded \emph{RobotCub} project, developed by the \emph{iCub Tech} in the \emph{Italian Institute of Technology}.
The robot was mainly created to support research in embodied artificial intelligence. 
One of the primary research goals focuses on providing iCub with capabilities that allows it to indulge in active physical interactions with the environment and human counterparts (\cite{metta2008icub, natale2017icub, natale2021icub}).
In 2021, 42 of these robots built into different versions have been distributed to several laboratories worldwide. 
The robotic platform used for experimental validation of the algorithms developed in this thesis is the iCub version 2.5, depicted in Figure \ref{fig:chap:intro:icub}.
\looseness=-1

iCub v2.5 is a $1.04$ meter tall robot weighing close to $33$ kilograms.
It consists of $53$ degrees of freedom (DoFs) of motion with the distribution of $6$ DoFs for each leg, $7$ and $9$ DoFs for each arm and hand respectively, $3$ DoFs for the torso, and $6$ DoFs for the head and the eyes.
A reduced subset of $32$ (or sometimes $26$) DoFs for the neck, torso, legs, and arms, which are most relevant for locomotion purposes, are considered in this thesis.
Each of these DoFs is electrically actuated with Brushless DC (BLDC) motors and a harmonic drive transmission (\cite{parmiggiani2012design}).
This subset of DoFs is also equipped with joint encoders and hall effect sensors that are used to measure the joint angles and velocities at 1000 Hz. \looseness=-1

\begin{figure}[!t]
\centering
\begin{subfigure}{0.3 \textwidth}
\centering
\includegraphics[scale=0.4]{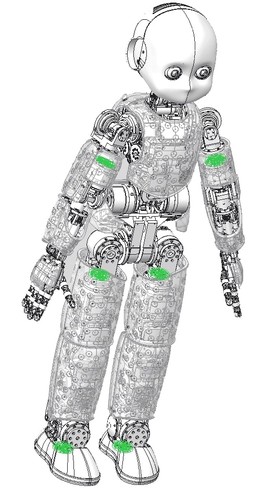}
\caption{}
\end{subfigure}
\begin{subfigure} {0.3 \textwidth}
\centering
\includegraphics[scale=0.4]{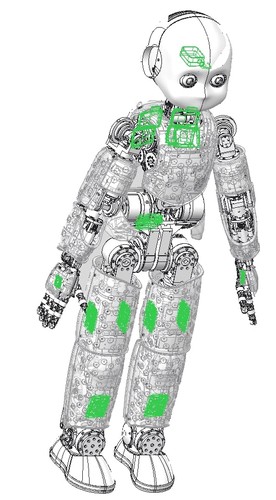} 
\caption{}
\end{subfigure}
\begin{subfigure}{0.3 \textwidth}
\includegraphics[scale=0.4]{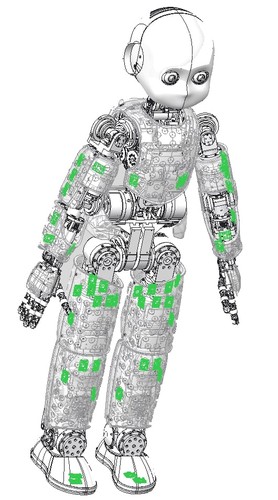} 
\caption{}
\end{subfigure}
\caption{Distribution of (a) six-axis force-torque sensors and (b, c) inertial sensors i.e gyroscopes (left) and accelerometers (right) in iCub v2.5 (\cite{traversaro2017modelling}).}
\label{fig:chap:intro:sensors}
\end{figure}

\begin{figure}[!t]
\centering
\begin{subfigure}{0.18\textwidth}
\centering
\includegraphics[scale=0.15]{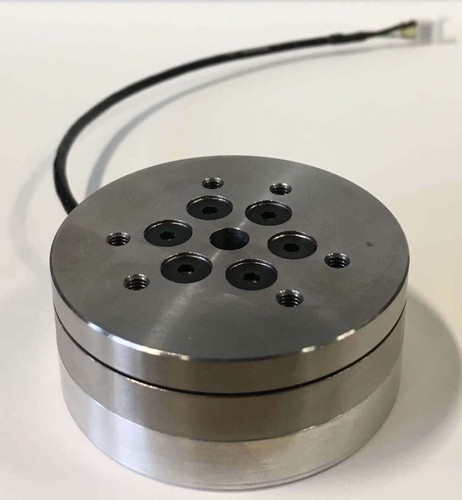} 
\caption{}
\end{subfigure}
\begin{subfigure}{0.18\textwidth}
\centering
\includegraphics[scale=0.15]{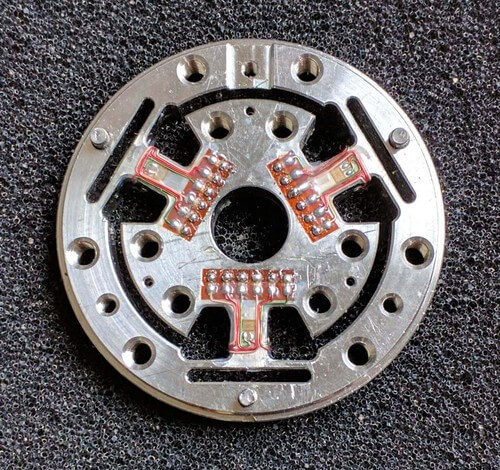} 
\caption{}
\end{subfigure}
\begin{subfigure}{0.18\textwidth}
\centering
\includegraphics[scale=0.15]{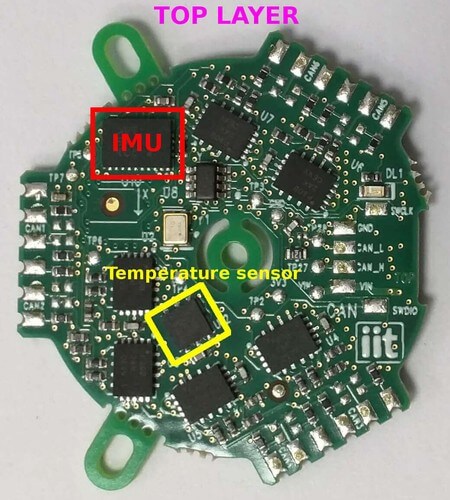}
\caption{}
\end{subfigure}
\begin{subfigure}{0.18\textwidth}
\centering
\includegraphics[scale=0.15]{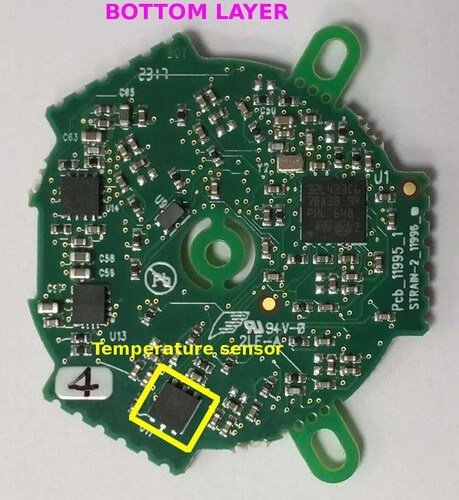} 
\caption{}
\end{subfigure}
\begin{subfigure}{0.18\textwidth}
\centering
\includegraphics[scale=0.15]{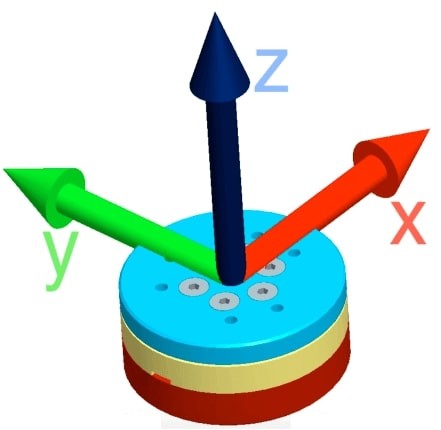} 
\caption{}
\end{subfigure}
\caption{(a) Six-axis force-torque sensors developed at the Italian Institute of Technology (IIT); (b) Strain gauge; (c, d) Internal electronics highlighting an Inertial Measurement Unit (IMU) and a temperature sensor; (e) Sensor reference frame (\cite{tirupachuri2020enabling}).}
\label{fig:chap:intro:ft}
\end{figure}

iCub is a sensor-rich platform with numerous sensing modalities distributed across the entire body of the robot.
We will describe only those modalities which are relevant to the scope of this thesis.
To support advanced physical interaction with its environments, the robot needs to be capable of measuring and controlling the forces to be exchanged with the environment.
iCub is equipped with six internal six-axis force-torque sensors as shown in Figure \ref{fig:chap:intro:sensors}, of which four of them are attached at the base of each limb while two of them are attached on the foot, close to the ankle of the robot.
These force-torque sensing technologies are developed in-house at the \emph{Italian Institute of Technology, Genova} (see Figure \ref{fig:chap:intro:ft}) based on the principles of strain-gauge sensing.
They also consist of an Inertial Measurement Unit (IMU) and temperature sensor on-board for effective non-linear modeling required for the calibration of these sensors  (\cite{chavez2016model}).
These force-sensing technologies along with a Bosch BN055 IMU placed on the head of the robot is used for the implementation of whole-body dynamics estimation algorithms that allow retrieving contact wrenches at the end-effectors (feet and hands) and joint torque estimates on the robot by exploiting the model information (\cite{nori2015icub, nori2015simultaneous}).
These calibration and estimation algorithms have been used successfully in the development of whole-body controllers for balancing and locomotion algorithms for achieving highly dynamic motions on the robot (\cite{dafarra2016torque, romualdi2020benchmarking, pucci2016highly}). 
Besides that, a vast array of three-axis accelerometers and three-axis gyroscopes are distributed in various parts of the body as depicted in Figure \ref{fig:chap:intro:sensors}. Table \ref{table:chap:intro-sensors-spec} provides a consolidated list of sensors used in the development of the state estimation strategies proposed in this thesis. \looseness=-1

\begin{table}[t]
\caption{Sensor specifications of iCub; C.M. denotes Custom Modifications on the original platform. \looseness=-1}
\centering
\scalebox{0.8}{
\begin{tabular}{|p{0.2\textwidth} |p{0.3\textwidth} | p{0.1\textwidth} | p{0.4\textwidth} | }
\hline
Sensor  &    Model    &  Operating Frequency (\si{\hertz})     &     Experimental Use Case        \\
\hline
Encoders   &  AES absolute encoder (joint side), incremental encoder (motor side)     &    1000    &   Generic         \\
Head IMU  &    Bosch BNO055    &   100     &  Whole-body dynamics estimation          \\
Base IMU (C.M.) &     XSens MTi-300      &   100     &   Floating base estimation     \\
F/T Sensors &   IIT STRAIN2      &   100     &   Whole-body dynamics estimation     \\
F/T Onboard IMU &   Bosch BNO055      &   100     &   Feet IMUs for improving floating base estimation     \\
\hline
\end{tabular}}
\label{table:chap:intro-sensors-spec}
\end{table}

\subsubsection{Custom Modifications to iCub}
Our particular version of the  robot in the \emph{Artificial and Mechanical Intelligence} laboratory is equipped with an XSens MTi-300 series IMU mounted in its base link providing linear accelerometer, gyroscope, magnetometer and orientation measurements streaming at 100Hz.
This custom modification was done on our robot to be able to have an IMU collocated on the base link for enabling floating base estimation algorithms beyond legged odometry.

In order to obtain ground truth of base link trajectories from a state-of-the-art motion capture system such as Vicon, we developed custom mounts equipped with reflective markers that are directly mountable on the base link of the robot (see Figure \ref{fig:chap:intro:icub}).
Additionally, in order to support an extension towards vision based floating base estimation, this support allows to attach a Realsense D435i camera which can complement the stereo vision cameras mounted in the eyes of the robot.

\subsection{Description of Wearable Sensing Technology}
\label{sec:chap:intro:wearable}

\begin{figure}[!h]
\centering
\includegraphics[scale=0.5]{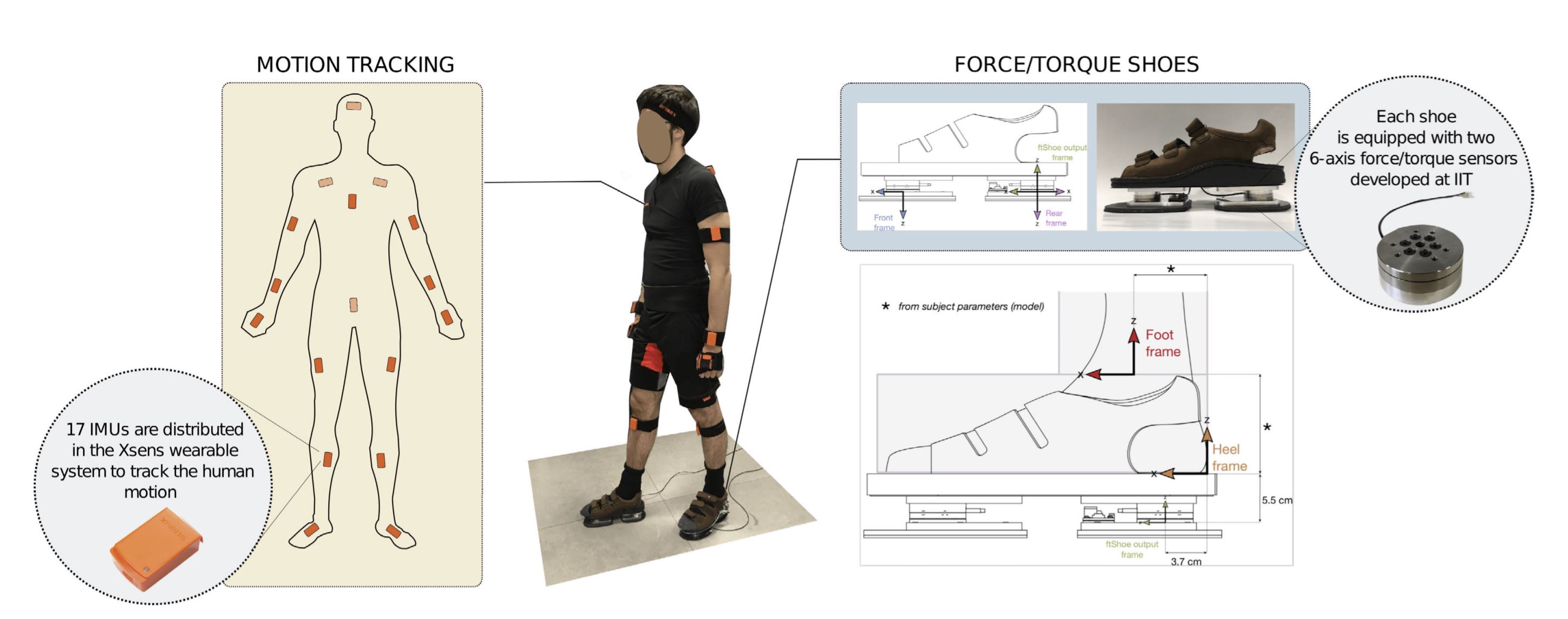}
\caption{A concise visualization of wearable technologies used: Human subject wears Xsens inertial motion tracking system and a pair of sensorized shoes equipped with force/torque
sensors developed at IIT (\cite{tirupachuri2021online}).}
\label{fig:chap:intro:wearable}
\end{figure}

A concise description of the wearable sensing technologies used within the context of this thesis is depicted in Figure \ref{fig:chap:intro:wearable}.
It comprises of a whole-body motion tracking suit and sensorized shoes.

\subsubsection{iFeel Sensorized Shoes}
\begin{figure}[!t]
\centering
\begin{subfigure}{0.4\textwidth}
\includegraphics[scale=0.325]{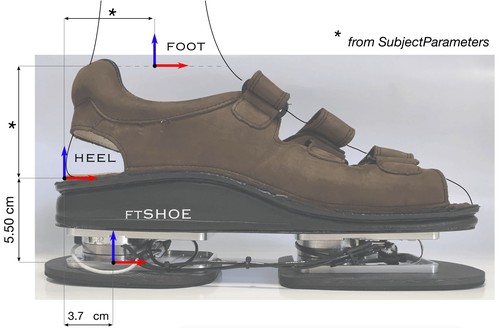}
\caption{}
\end{subfigure}
\begin{subfigure}{0.4\textwidth}
\includegraphics[scale=0.2]{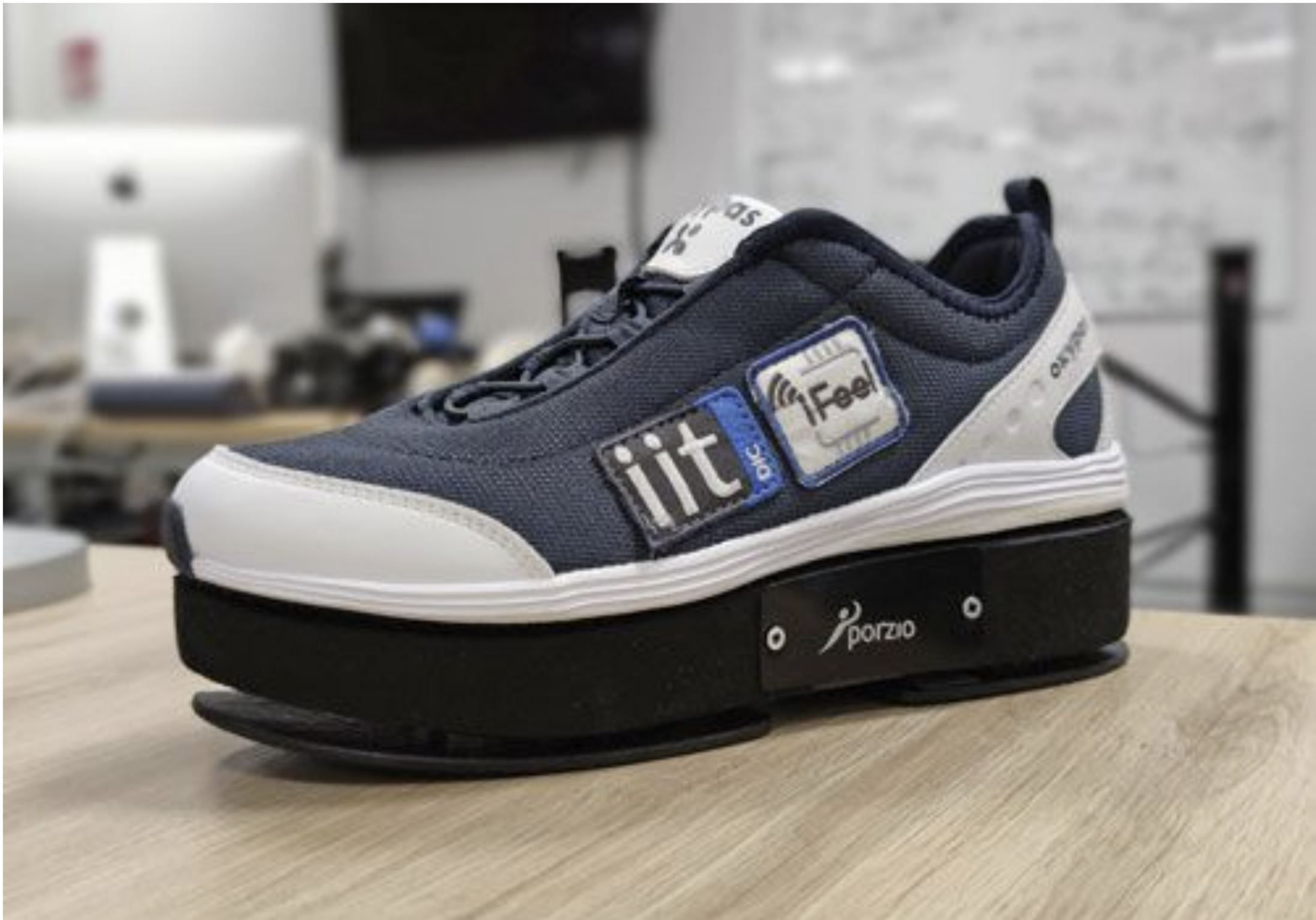}
\caption{}
\end{subfigure}
\caption{(a) An early prototype of sensorized shoes equipped with six-axis force torque sensors at its front and rear parts of the sole along with the visualization of coordinate system relationships for the sensor and the human foot. (\cite{latella2019human}). (b) Latest version of the iFeel shoes.}
\label{fig:chap:intro:ftShoes}
\end{figure}

The sensorized shoes depicted in Figure \ref{fig:chap:intro:ftShoes} is a wearable technology developed by \emph{iFeel}\footnote{\url{https://ifeeltech.eu/}} which is a part of a whole-body wearable technology suite aimed at real-time human motion tracking, articular stress analysis and fatigue assessment. 
Each shoe is mounted with two six-axis force-torque sensors at the front and rear balls of the foot (near the toe and the heel).
These force-torque sensors rely on the force sensing technologies described earlier (Figure \ref{fig:chap:intro:ft}).
The wrench measurements from these force-torque sensors are measured in their local frames which are then transformed into a common heel frame and streamed as a net wrench at the rate of nearly $100$ Hz.
Depending on the foot geometry of the human-subject, another relevant transformation is required to express these measurements in the coordinate frame associated to the foot.
These coordinate frames are depicted in Figure \ref{fig:chap:intro:ftShoes}.

\subsubsection{Xsens Motion Capture Suit}
The whole-body motion tracking suit is an \emph{Xsens}\footnote{\url{https://www.xsens.com/}} suit consisting of a set of $17$ inertial measurement units distributed in selected locations on the human body. 
This motion capture suit is capable of providing several kinematic and dynamic quantities relevant for human motion estimation.
Within the scope of this thesis, we mostly rely on the absolute orientation and the angular velocity measurements provided by the distributed IMUs streamed at a rate of $60$ Hz frequency.
Xsens suit offers a thorough calibration procedure to align all the measurements in a single frame of reference.
We additionally carry out a secondary calibration using some known conditions at the first instant of operation to align these measurements into our desired frame of reference.

\vspace{5em}
In the subsequent chapters, we will look at the mathematical background and familiarize with the tools used for the development of state estimation techniques in this thesis.


\chapter{State Estimation using Matrix Lie Groups}
\label{chap:estimation}

This chapter presents the mathematical background for the estimation machinery used within the scope of this thesis. 
Most of the material discussed in this chapter is based on the works of \cite{fang2018nonlinear, sola2018micro, barfoot2014associating, bourmaud2015estimation, barrau2015non, hall2013lie}.
Section \ref{sec:chapter02:intro-matrix-lie-groups} presents a brief introduction to matrix Lie groups and some useful matrix Lie group identities.
Examples of most commonly used matrix Lie groups in state estimation for robotic applications are described in the Appendix \ref{appendix:chapter02:examples-matrix-lie-groups}.
Section \ref{sec:chapter02:uncertainty-matrix-lie-groups} describes a notion of uncertainty representation and propagation over matrix Lie groups.
Sections \ref{sec:chapter02:ekf-matrix-lie-groups} and \ref{sec:chapter02:invekf-matrix-lie-groups} provide an overview of two common filtering techniques using matrix Lie groups which are used as the foundations for deriving the state estimation methods presented in this thesis.
Finally, averaging on matrix Lie groups is presented in Section \ref{sec:chap:loosely-couple-averaging}.

\section{A Primer on State Estimation}
\label{sec:chapter02:state-estimation-primer}

In order to understand the evolution of many real-world phenomena, we tend to use the theory of dynamical systems to model them into a mathematical description.
The behavior of these systems can be inferred through a minimal set of variables, called the state variables.
 The evolution of these variables over time is obtained as the result of the considered system dynamics.
When we are not fully certain about the system dynamics, sensor measurements may be used to complement the information obtained from the system dynamics.
However, in most practical applications, either the measurement observations of the system state are very noisy or having a complete observation of the system is not possible.
In the latter case, some of the states are not directly measurable but can only be estimated.
State estimation is the process of inferring the evolving state of the system from a complete or a partial set of noisy observations. 

One of the common approaches to state estimation follows a Bayesian way of thinking.
This approach provides a framework to understand how an existing belief of a variable in consideration is changed or updated when new evidence about this variable is available.
Within this framework, both the system state and the measurements are considered as \emph{random variables}. 
While the system dynamics provides a prediction of the unknown state variable, the measurement-based update is used to correct these predicted state estimates.

Consider a nonlinear, dynamical system,
\begin{align}
\label{eq:chap02:estimation-dynamical-system}
\mathbf{x}_\knext &=  f\left(\mathbf{x}_\kcurr, \mathbf{u}_\kcurr\right) + \mathbf{w}_\kcurr, \\
\label{eq:chap02:estimation-measurement}
\mathbf{z}_\kcurr &= h \left(\mathbf{x}_\kcurr\right) + \mathbf{n}_\kcurr.
\end{align}
Here, we denote $\mathbf{x}_\kcurr \in \R^p$ as the unknown state vector of dimensions $p$,\ $\mathbf{u}_\kcurr \in  \R^m$ as the control input and $\mathbf{z}_\kcurr \in \R^q$ is the vector of measured system outputs with dimensions $q$.
The nonlinear mapping $f:\R^p \mapsto \R^p$ describes the system dynamics, usually used as the prediction model in the state estimation problem (\cite{barfoot2017state}). 
The function $h: \R^p \mapsto \R^q$ is the measurement model that maps the system state to expected observations. \looseness=-1
For stochastic systems, the prediction and the measurement models are assumed to be affected by noise to model the uncertainty in these models.
The process noise $\mathbf{w}_\kcurr$ and the measurement noise $\mathbf{n}_\kcurr$ are usually assumed to be zero-mean white Gaussian noise with covariance matrices $\mathbf{Q}_k \in \R^{p\times p}$ and $\mathbf{N}_k  \in \R^{q\times q}$ respectively.
Given a set of observations $Z_\kcurr = \left\{\mathbf{z}_1, \mathbf{z}_2,  \dots \mathbf{z}_k\right\}$ at time instant $k$, the problem of state estimation is then to obtain an estimate of $\mathbf{x}_\kcurr$ using the conditional probability distribution $p\left(\mathbf{x}_\kcurr \mid Z_\kcurr \right)$.

\cite{fang2018nonlinear} presents a concise tutorial review of various state-of-the-art techniques used for nonlinear Bayesian estimation.
Computing and tracking the conditional probability density function for systems with strong nonlinearity and non-Gaussian probability distributions of the stochastic variables is often intractable due to the lack of a closed-form solution.
This led to the development of a simplified and tractable solution called Kalman filtering.
Under the assumption that the state and measurements are characterized by Gaussian distributions, this approach tracks only the mean and covariance of the state estimates conditioned over the measurements while being subject to the nonlinear transformations. 
In contrast to filtering methods, there exist great alternatives such as smoothing methods and optimization-based state estimation approaches that can be used for applications with varying degrees of complexity.
In this thesis, we only focus on Kalman filtering approaches, however a brief comparison with other methods is done in Chapter \ref{chap:context}.
\looseness=-1

\subsection{Extended Kalman Filter}
\label{sec:chapter02:state-estimation-primer-ekf}
While the Kalman filter is an optimal state estimation approach when applied to linear systems (\cite{sarkka2013bayesian, fang2018nonlinear, simon2006optimal}) with state and measurements characterized as Gaussian variables, it is however not directly applicable for nonlinear systems. 
A nonlinear version of the Kalman filter called the Extended Kalman Filter (EKF) is a widely used state estimation technique that linearizes the nonlinear functions about the mean and the covariance of the current state estimate.
The EKF can be viewed as a two-step process involving a \emph{propagation} (or a \emph{prediction}) step and an \emph{update} step. 
The propagation step uses the state estimate from the previous time instant to produce a prior estimate of the state at the current time instant.  
An innovation or a residual is then computed, in the update step, as the difference between the expected observation obtained from the current prior prediction and the actual observation. 
This residual error is multiplied by an optimal Kalman gain and combined with the previous state estimate to obtain a refined posterior state estimate. 
The outcome of the EKF at the end of the propagation and the update step is a conditional probability of the state $\mathbf{x}_\knext$ given a set of observations $Z_l$, where $l = \kcurr$ for the propagation and $l = \knext$ for the update.
Given the prior estimate $\mathbf{\hat{x}}_\kprior$ of the state $\mathbf{x}_\kcurr$ and the exogenous input vector $\mathbf{u}_\kcurr$, the prediction model is linearized as,  \looseness=-1
$$
f\left(\mathbf{x}_\kcurr, \mathbf{u}_\kcurr\right) \approx f\left(\mathbf{\hat{x}}_\kprior, \mathbf{u}_\kcurr\right)\ +\ \mathbf{F}_\kcurr\ (\mathbf{x}_\kcurr\ -\ \mathbf{\hat{x}}_\kprior), \quad \mathbf{F}_\kcurr = \frac{\partial f}{\partial \mathbf{x}}|_{\mathbf{x} = \mathbf{\hat{x}}_\kprior, \mathbf{u}_k} \in \R^{p \times p},
$$
leading to the computation of the estimated state $\mathbf{\hat{x}}_\kpred$ and the covariance $\cov_\kpred$ of the state prediction error in the prediction step as, \looseness=-1
\begin{align}
\begin{split}
     &\mathbf{\hat{x}}_\kpred = f\left(\mathbf{\hat{x}}_\kprior, \mathbf{u}_\kcurr\right), \\
     &\cov_\kpred = \mathbf{F}_\kcurr\ \cov_\kprior\ \mathbf{F}_\kcurr^T + \mathbf{Q}_\kcurr,
 \end{split}
\end{align}
where, $\mathbf{\hat{x}}_\kpred \in \R^p, \cov_\kpred \in \R^{p \times p}$.
The measurement model can then be linearized around a new operating point $\mathbf{\hat{x}}_\kpred$ as,
$$
h\left(\mathbf{x}_\knext\right) \approx h\left(\mathbf{\hat{x}}_\kpred\right) + \mathbf{H}_\knext\  (\mathbf{x}_\kcurr\ -\ \mathbf{\hat{x}}_\kpred), \quad \mathbf{H}_\knext = \frac{\partial h}{\partial \mathbf{x}}|_{\mathbf{x} = \mathbf{\hat{x}}_\kpred} \in  \R^{q \times p},
$$
leading to the innovation covariance $\mathbf{S}_\knext = \mathbf{H}_\knext\ \cov_{\kpred}\ \mathbf{H}_\knext^T\ +\ \mathbf{\hat{N}}_\knext  \in  \R^{q \times q}$. 
The innovation, $\mathbf{\tilde{z}}_\knext = \mathbf{z}_\knext - h\left(\mathbf{\hat{x}}_\kpred\right) \in \R^q$, is defined as the difference between the measured observation and the expected observation from the predicted state estimate.
The Kalman gain is computed as $\mathbf{K}_\knext =\cov_\kpred \ \mathbf{H}_\knext^T  \mathbf{S}^{-1}_\knext  \in \R^{p \times q}$, where $\cov_\kpred \ \mathbf{H}_\knext^T$ is the cross-covariance between the predicted state $\mathbf{\hat{x}}_\kpred$ and the measurement $\mathbf{z}_\knext$.

An updated state estimated after incorporating the measurements is given as,
\begin{align}
\begin{split}
     &\mathbf{\hat{x}}_\kest = \mathbf{\hat{x}}_\kpred\ +\ \mathbf{K}_\knext \mathbf{\tilde{z}}_\knext, \\
     &\cov_\kest = \left(\I{p} - \mathbf{K}_\knext\ \mathbf{H}_\knext\right) \cov_\kpred,
 \end{split}
\end{align}
where, $\mathbf{\hat{x}}_\kest \in \R^p, \cov_\kest \in \R^{p \times p}$ and  $\I{p} \in \R^{p \times p}$ is an identity matrix. 

For a more detailed understanding of the Extended Kalman Filter and other state estimation methods, one may refer to \cite{simon2006optimal, sarkka2013bayesian, fang2018nonlinear}.
In this thesis, Kalman filtering on special mathematical structures called the matrix Lie groups is used to formulate the state estimation problems. \looseness=-1


\section{Matrix Lie Groups}
\label{sec:chapter02:intro-matrix-lie-groups}

The theory of Lie groups serves as a powerful mathematical toolkit that allows tackling problems related to mechanical systems with complex underlying geometries in a manner rigorous for handling uncertainties, derivatives, and integrals for such systems more faithfully.
A subset of Lie groups realized as a group of matrices, known as the matrix Lie groups, can be used to provide appropriate representations of spatial relationships for such mechanical systems.

\subsection{Lie Group and Lie Algebra}
A Lie group is a mathematical structure that is a combination of two mathematical objects known as groups and differentiable manifolds.
Specifically, a Lie group $G$ is a differentiable manifold whose elements obey the axioms of a group.
The concept of a differentiable manifold has its origin in \emph{set theory} where it is related to the point-set topology that describes the continuity or closeness of a space (\cite{weisstein2021pointsettopology}).
At each point on the manifold, a vector space can be defined that best approximates the manifold structure within the local neighborhood of the point.
This vector space is called as the tangent space defined at the point on the manifold.

\begin{remark}
\label{remark:chap02-smooth-manifold}
A differentiable or a smooth manifold is a topological space in which every point is associated with a unique tangent space, usually characterized as a local linear or a vector space, in which calculus operations can be performed (\cite{sola2018micro}). It is also called as analytic manifold that are infinitely differentiable and has a convergent Taylor series (\cite{weisstein2021convergentseries, weisstein2021taylorseries}).
\end{remark}

\begin{definition}[Group]
\label{def:chap02-group}
A group is a mathematical structure that constitutes a pair $(G, \circ)$ consisting of a set $G$ and a binary operation $\circ$ such that for any elements $g, g_1, g_2 \in G$, the following properties always hold, \looseness=-1
\begin{itemize}
    \item \textbf{Closure:} For every element $g_1,\ g_2 \in G$, their operation $g_1 \circ g_2$ also belongs to $G$.
    \item \textbf{Associativity:} For every element $g_1,\ g_2,\ g_3 \in G$, $(g_1 \circ g_2) \circ g_3\ =\  g_1 \circ (g_2 \circ g_3)$.
    \item \textbf{Identity:} There exists an element $e$ called identity, such that $\forall g \in G,\ e \circ g\ =\ g \circ e\ =\ g$.
    \item \textbf{Inverse:} For every element $g \in G$, there exists an element $g^{-1}$ called inverse of $g$, such that $g \circ g^{-1}\ =\ g^{-1} \circ g\ =\ e$.
\end{itemize}
\end{definition}

\begin{definition}[Matrix Lie Group]
\label{def:chap02-matrix-lie-group}
A matrix Lie group $(G, \circ)$ is a group for which the set $G$ is a differentiable manifold in which,
\begin{itemize}
    \item each element $\X \in G$ is an $n \times n$ matrix, 
    \item the group operation "$\circ$" is the matrix multiplication,
    \item the identity element is the identity matrix $\I{} \in \R^{n \times n}$,
    \item the inverse is the matrix inversion,
    \item the mappings $a(\X_1, \X_2)\ =\ \X_1 \circ \X_2$ and $b(\X)\ =\ \X^{-1}$ are both analytic i.e. mappings for which a convergent Taylor series can be defined.
\end{itemize}
\end{definition}

\begin{remark}
\label{remark:chap02-matrix-lie-group}
A Lie group in which the group elements are matrices, but the group operation is not the matrix multiplication is not a matrix Lie group.
\end{remark}

\begin{definition}[Action of a Lie group, \cite{sola2018micro}, Section II.B]
\label{def:chap02-group-action}
Given a Lie group $G$ and a set $H$, the action of $\ \X \in G$  on $h \in H$ is denoted as $\X. h$, if for an identity element $\I{} \in G$ and an arbitrary element $\Y \in G$, \looseness=-1
\begin{align}
&\I{}. h = h,\\
& (\X \circ \Y).h = \X.(\Y.h).
\end{align}
\end{definition}

\begin{definition}[Tangent Space of a Lie Group, \cite{sola2018micro}]
\label{def:chap02-tangent-space-matrix-lie-group}
Given a point $\gamma(t): \R \mapsto G$ moving on the manifold of Lie group $G$, it's velocity $\dot{\gamma} = \frac{d \gamma(t)}{dt}$ belongs to the space $T_\gamma G$ tangent to $G$ at the point $\gamma$. This is the space of all matrices $\A \in \R^{n \times n}$ related to the smooth function $\gamma(t)$ for which: \looseness=-1
\begin{equation}
   \A = \left( \frac{d \gamma(t)}{dt} \right)_{t = T}, \ \gamma(T) \in G.
\end{equation}
\end{definition}

Intuitively, the tangent space at a point can be viewed as the space of possible velocities at a given element on the group manifold.

\begin{definition}[Lie algebra]
\label{def:chap02-lie-algebra}
Given a Lie group $G$, the tangent space $T_{\I{}}G$ at the identity element $\I{} \in G$  is called as the Lie algebra $\mathfrak{g}$ of $G$.
\end{definition}

The Lie algebra captures all the properties of the Lie group.
Since the Lie algebra is a vector space, its elements can be equivalently identified by vectors in $\R^p$ through bijective mappings $\gvee{G}{.}$ and $\ghat{G}{.}$, called the \emph{vee} and \emph{hat} operators respectively, where $p$ is the dimension of the matrix Lie group.
Specifically, the Lie algebra $\mathfrak{g} \subset \R^{n \times n}$ is a $p$-dimensional vector space defined by a basis of $p$ real matrices $\ghat{G}{\mathbf{e}_i}$ called as generators, where $\mathbf{e}_i$ defines the natural basis of $\R^p$. 
The Lie algebra defines an open neighborhood around $\Zeros{p}{1}$ in its associated vector space, over which the notions of calculus are applicable.
Let $\A \in \mathfrak{g}$ and $\mathbf{a} \in \R^p$, the Lie algebra and its associated vector space are related as follows, \looseness=-1
\begin{equation}
\label{eq:chap02:liegroups-generators}
\A = \ghat{G}{\mathbf{a}} = \sum_{i = 1}^p \mathbf{a}_i \ghat{G}{\mathbf{e}_i} \in \mathfrak{g}, \quad \quad \gvee{G}{.}: \mathfrak{g} \mapsto \R^p, \quad \quad \ghat{G}{.}: \R^p \mapsto \mathfrak{g}.
\end{equation}


\subsection{Exponential and Logarithm Mapping}
The Lie group and the Lie algebra can be related through two important maps which transport the elements from one space to the other.
These are the exponential and the logarithm map of the Lie group. \looseness=-1

\begin{definition}[Exponential Map]
\label{def:chap02:exponential-map}
The exponential map of the Lie group $\exp_G: \mathfrak{g} \to G$ maps an element $\ghat{G}{\mathbf{a}}$ at the Lie algebra to an element $\X$ in the Lie group. 
\end{definition}

\begin{definition}[Logarithm Map]
\label{def:chap02:log-map}
The logarithm map of the Lie group $\log_G: G \to \mathfrak{g}$, is the inverse of the exponential map that takes an element of the Lie group and produces an element of the Lie algebra. \looseness=-1
\end{definition}

The relationship between the exponential map and the logarithm map is straightforward,
\begin{equation}
    \label{eq:chap02:liegroups-exp-log-relation}
    \glog{G}\left(\gexp{G}\left(\ghat{G}{\mathbf{a}}\right) \right) = \ghat{G}{\mathbf{a}}.
\end{equation}

Interestingly, for matrix Lie groups, these mappings are just the matrix exponential and matrix logarithm, respectively.
\begin{definition}[Matrix Exponential]
\label{def:chap02:matrix-exponential}
Given a matrix $\A \in \R^{n \times n}$, the \emph{exponential} of the matrix $\A$, indicated with $\exp{(\A)}$, is defined as:
\begin{equation}
\exp{(\A)} = \sum_{k = 0}^{\infty} \frac{\A^k}{k!} . 
\end{equation}
\end{definition}

\begin{definition}[Matrix Logarithm]
\label{def:chap02:matrix-logarithm}
Given a matrix $\X \in \R^{n \times n}$, the matrix logarithm is defined by the Taylor series about the identity matrix:
\begin{equation}
\log{(\X)} = \log(\I{n} - (\X - \I{n})) = \sum_{k = 1}^{\infty} \left(-1\right)^{k+1} \frac{\left(\X - \I{n}\right)^k}{k} . 
\end{equation}
\end{definition}

A short-hand notation composing the hat operation with the exponential map and the logarithm mapping with the vee operation can be introduced,
\begin{center}
 \begin{tabular}{c c} 
 \label{table:chap02:liegroup-operators}
$\gexp{G}: \mathfrak{g} \to G$,  & $\glog{G}: G \to \mathfrak{g}$, \\
$\gexphat{G}: \R^p \to G$, & $\glogvee{G}: G \to \R^p$.
\end{tabular}
\end{center}
Furthermore, depending on the context, these composition operators are sometimes overloaded as $\gExp{G} \triangleq \gexphat{G}$ and $\gLog{G} \triangleq \glogvee{G}$.
In contexts where the underlying group is clear, the subscript $G$ may also be dropped. \looseness=-1

Since the exponential map is a continuous group homomorphism of the one parameter subgroup of $G$, indicated as $\gamma: \mathbb{R} \to G$, it has some useful identities.
\begin{property}[Homomorphism of exponential map]
\label{prop:chap02:exp-map-homomorphism}
$\forall\; t, \;t_1, \;t_2 \in \mathbb{R}, \; \mathbf{a} \in \mathbb{R}^p, \; \ghat{G}{\mathbf{a}} \in \mathfrak{g}$,
\begin{align}
    \begin{split}
        & \gamma(t)\ =\ \gexphat{G}\left(t\; \mathbf{a}\right), \\
        & \gamma(-t)\ =\ \gexphat{G}\left(-\;t\; \mathbf{a}\right)\ =\ \left(\gexphat{G}\left(t \; \mathbf{a}\right)\right)^{-1}\ =\ \gamma^{-1}(t), \\
        & \gamma(0) = \I{n}, \\
        & \gamma(t_1 + t_2)\  =\  \gexphat{G}\left((t_1\; + t_2)\; \mathbf{a}\right)\ =\ \gexphat{G}\left(t_1 \; \mathbf{a}\right) \;\gexphat{G}\left(t_2 \; \mathbf{a}\right)\ =\ \gamma(t_1)\;\gamma(t_2).
    \end{split}
\end{align}
\end{property}

\subsection{Adjoint Representation}
Adjoint representation is an important map which is an invertible linear transformation that operates on the vectors of the tangent space $T_{\X}G$ at a certain element $\X \in G$ and transports them to Lie algebra. \looseness=-1

\begin{definition}[Adjoint action of a matrix Lie group on its Lie algebra, \citep{hall2013lie} Definition 3.32]
\label{def:chap02:adjointActionOfGroupOnAlgebra}
For each element $\X$ in the Lie group $G$ and a corresponding element $\A$ in its Lie algebra $\mathfrak{g}$, the adjoint map of the group ${\gadj{}}_\X : \mathfrak{g} \mapsto \mathfrak{g}$ is defined as:
\begin{equation}
  {\gadj{}}_\X\ \left(\A\right) = \X\ \A\ \X^{-1}.
\end{equation}
\end{definition}

Along with the exponential map, the adjoint map captures the non-commutativity of a matrix Lie Group.
An equivalent matrix operator called the Adjoint matrix can be defined that acts directly on the elements in the vector space,
\begin{align}
\label{eq:chap02-liegroups-adjoint-repr}
    \begin{split}
        & \forall\; \X \in G, \; \mathbf{a} \in \R^p, \\
        & \gadj{\X}\ \mathbf{a} = \gvee{G}{\X\ \ghat{G}{\mathbf{a}}\ \X^{-1}}, \\
        & \X\ \gexp{G}\left(\ghat{G}{\mathbf{a}}\right) = \gexp{G}\left(\ghat{G}{\gadj{\X}\ \mathbf{a}} \right)\ \X, \\
        & \X\; \gexphat{G}\left(\mathbf{a}\right) = \gexphat{G}\left(\gadj{\X}\ \mathbf{a}\right) \X.
    \end{split}
\end{align}

The Lie bracket is obtained as a consequence of the derivative of the adjoint transformation of a one-parameter subgroup (\cite{chirikjian2011stochastic}), $$ \frac{d}{dt} \left(\gadj{\gexp{G}\left(t\ \B\right)}\ \Cbold\right)_{t = 0} = \liebracket{\B, \Cbold} = \B\Cbold - \Cbold\B,$$
$\forall \; \mathbf{b}, \mathbf{c} \in \R^p, \quad \B = \ghat{G}{\mathbf{b}}, \; \Cbold = \ghat{G}{\mathbf{c}} \in \mathfrak{g}$. The Lie bracket satisfies the Jacobi identity \cite[Definition 3.1]{hall2013lie} and $\liebracket{\B, \B} = 0$ .

\begin{definition}[Adjoint action of a Lie algebra on itself, \citep{hall2013lie} Definition 3.7]
\label{def:chap02:adjointActionOfAlgebraOnAlgebra}
For each element $\A, \B \in \mathfrak{g}$, the adjoint map of the Lie algebra $\gsmalladj{\A} : \mathfrak{g} \mapsto \mathfrak{g}$, sometimes called as the \emph{small} adjoint operator, is defined as:
\begin{equation}
  \gsmalladj{\A} \B = \liebracket{\A, \B} .
\end{equation}
\end{definition}

\noindent Consequently, we have $\gsmalladj{\B}\ \mathbf{c} = \liebracket{\B, \Cbold}$, $\gsmalladj{\Cbold}\ \mathbf{b} = \liebracket{\Cbold, \B} = -\liebracket{\B, \Cbold}$ and $\gsmalladj{\B}\ \mathbf{b} = 0$.

\begin{figure}[tpb]
	\centering
	\includegraphics[scale=0.85]{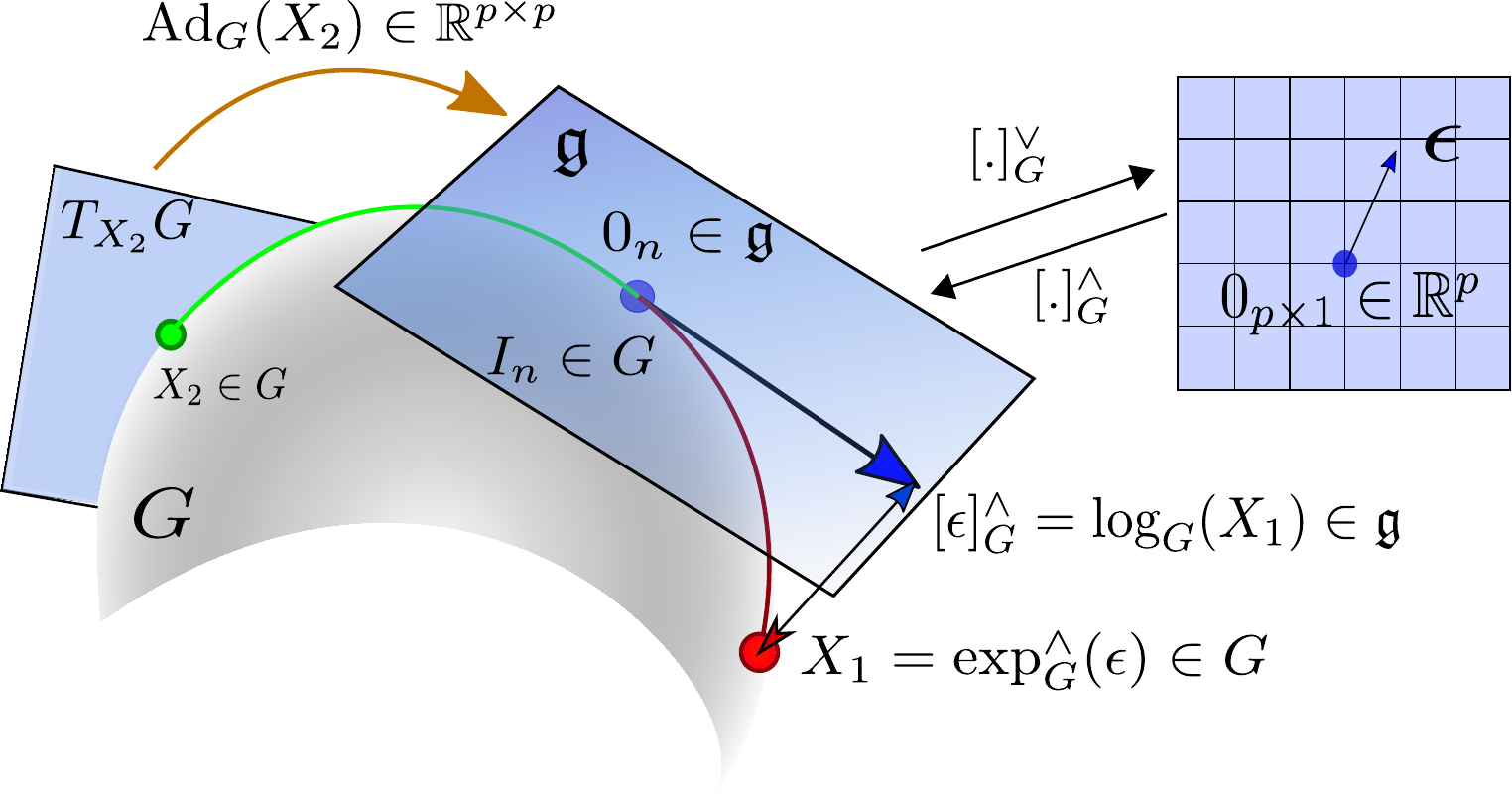}
	\caption{An illustration of a Lie group and its operators transporting elements between the group, Lie algebra and the associated vector space.}
\label{fig:chap02-matrixliegroup-liegroup}
\vspace{-2mm}
\end{figure}

A concise visual description of a matrix Lie group is provided in Figure \ref{fig:chap02-matrixliegroup-liegroup}.
In summary, a Lie group is a mathematical construct that has the structure of a smooth, differentiable manifold $G$ represented by the grey surface in the figure; it can simply be thought of as a collection of entities with similar characteristics where at each entity the notion of a derivative can exist and every entity satisfies the properties of a group. 
Its associated Lie algebra is the tangent space represented by the blue plane at the identity element $\I{n}$. 
The exponential and the logarithmic mapping of the Lie group can intuitively be thought of as moving from a curved, nonlinear space to a flat, linear space, and vice-versa, respectively. 
An analogous example of such operations applied to scalars are the exponential function that converts an "addition to multiplication" ($a, b \in \mathbb{R}, \; e^{a+b} = e^a e^b$) and the  (natural) logarithm function that has a "multiplication-to-addition" property ($\glog{}(ab) = \glog{}(a) + \glog{}(b)$). 
The linear space of the Lie algebra can further be mapped to a space of vectors (and vice-versa), a familiar space where we could perform calculus. 
Further, for Lie groups that define a non-commutative, bilinear operation between its elements, the adjoint representation quantifies how much an element fails to commute with another element in the group. It can also be viewed as a coordinate transformation of vectors from one tangent space to another.

\subsection{Left and Right Jacobians}
The Jacobians of the matrix Lie group map any changes in the local coordinates of the Lie group to the generalized velocity of the corresponding element in the Lie group.

\begin{definition}[Jacobian of matrix Lie group, \cite{chirikjian2011stochastic}]
Given $\mathbf{q} \in \R^p$ is a vector of local coordinates and $\X(t) = \mathbf{\tilde{X}}\left(\mathbf{q}(t)\right)$ is a curve in the Lie group $G$, where $\mathbf{\tilde{X}}: \R^p \mapsto G$ is the local parametrization of the Lie group $G$, 
\begin{itemize}
    \item the left Jacobian matrix is defined as the matrix $\gljac{G}\left(\mathbf{q}\right)$ that relates rates of change $\mathbf{\dot{q}}$ to $\mathbf{\dot{X}}\ \X^{-1}$, \looseness=-1
    \item the right Jacobian matrix is defined as the matrix $\grjac{G}\left(\mathbf{q}\right)$ that relates rates of change $\mathbf{\dot{q}}$ to $\X^{-1} \mathbf{\dot{X}}$.
\end{itemize}

\end{definition}

Another notion of the left and right Jacobians of the matrix Lie group can be obtained from the so-called  Baker-Campbell-Hausdorff formula which will be an important tool to handle uncertainties on matrix Lie groups.
The Baker-Campbell-Hausdorff (BCH) formula defines an important relationship between the Lie bracket, the matrix exponential, and the matrix logarithm. \looseness=-1
\begin{definition}[Baker Campbell Hausdorff formula, (\cite{chirikjian2011stochastic}), Section 10.2.7]
\label{def:chap02:BCH-formula}
For each element $\B, \Cbold \in \mathfrak{g}$, the logarithm of the product of two Lie group elements written as the exponential of the Lie algebra elements is related to Lie bracket as, 
\begin{equation}
  \log\left(\exp\left(\B \right) \exp\left(\Cbold \right)\right) = \B + \Cbold + \liebracket{\B, \Cbold} + \frac{1}{12}\left(\liebracket{\B, \liebracket{\B, \Cbold}\ } + \liebracket{\Cbold, \liebracket{\Cbold, \B}\ }\right) + \dots.
\end{equation}
\end{definition}

Since the Lie bracket is related to the small adjoint operator, the Baker-Campbell-Hausdorff (BCH) formula can be equivalently written as, \looseness=-1
\begin{equation}
\label{eq:chap02:liegroups-bch-adj}
  \glogvee{G}\left(\gexphat{G}\left(\mathbf{b} \right) \gexphat{G}\left(\mathbf{c} \right)\right) = \mathbf{b} + \mathbf{c} + \frac{1}{2}\gsmalladj{\B}\ \mathbf{c}\ +\ \frac{1}{12}\left(\gsmalladj{\B}\ \gsmalladj{\B}\ \mathbf{c} + \gsmalladj{\Cbold}\ \gsmalladj{\Cbold}\ \mathbf{b}\right) + \dots.
\end{equation}

A slight manipulation of the BCH formula leads to the Jacobian of the matrix Lie group, \looseness=-1
\vspace{-2mm}
\begin{equation}
\label{eq:chap02:liegroups-bch-jacobians}
\begin{split}
  \glogvee{G}\left(\gexphat{G}\left(-\mathbf{a} \right) \gexphat{G}\left(\mathbf{a} + \mathbf{b} \right)\right) &= -\mathbf{a}\ +\ (\mathbf{a} + \mathbf{b}) + \frac{1}{2}\ \gsmalladj{\ghat{G}{-\mathbf{a}}}\ (\mathbf{a} + \mathbf{b})\ +\ \dots \\
  &= \mathbf{b}\ +\ \frac{1}{2}\ \gsmalladj{\ghat{G}{-\mathbf{a}}}\ \mathbf{b}\ +\ \frac{1}{6}\ \gsmalladj{\ghat{G}{-\mathbf{a}}}^2\ \mathbf{b}\ +\ \dots \\
  &= \sum_{k=0}^\infty \frac{1}{(k+1)!}\  \gsmalladj{\ghat{G}{-\mathbf{a}}}^k\ \mathbf{b} \\
  &= \gljac{G}\left(-\mathbf{a}\right)\ \mathbf{b}\ =\ \grjac{G}\left(\mathbf{a}\right)\ \mathbf{b},
\end{split}
\end{equation}
where, we have used the identity $\gsmalladj{\ghat{G}{-\mathbf{a}}}\ \mathbf{a} = 0$, and  $\grjac{G}\left(\mathbf{a}\right) = \sum_{k=0}^\infty \frac{1}{(k+1)!}\  \gsmalladj{\ghat{G}{-\mathbf{a}}}^k$ denotes the right Jacobian operator of the matrix Lie group $G$.
As a consequence, we have the so-called push-forward identity of the matrix Lie group which captures the effect of any additive perturbations in the tangent space onto the multiplicative perturbations on the group manifold.
\begin{align}
\label{eq:chap02:liegroups-pre-push-forward}
  &\gexphat{G}\left(-\mathbf{a} \right) \gexphat{G}\left(\mathbf{a} + \mathbf{b} \right)\ =\ \gexphat{G}\left(\grjac{G}\left(\mathbf{a}\right)\ \mathbf{b}\right), \\
\label{eq:chap02:liegroups-push-forward}
  &\gexphat{G}\left(\mathbf{a} + \mathbf{b} \right)\ =\ \gexphat{G}\left(\mathbf{a} \right)\ \gexphat{G}\left(\grjac{G}\left(\mathbf{a}\right)\ \mathbf{b}\right).
\end{align}

The right Jacobian inverse is related to the BCH formula approximated to the first order as, \looseness=-1
\begin{equation}
\label{eq:chap02:liegroups-rjac-bch}
\glogvee{G}\left(\gexphat{G}\left(\mathbf{a} \right) \gexphat{G}\left(\mathbf{b} \right)\right) \approx \mathbf{a}\ +\ \left(\grjac{G}\left(\mathbf{a}\right)\right)^{-1}\ \mathbf{b}.
\end{equation}

Similar identities can be obtained for the left Jacobian,
\begin{align}
\label{eq:chap02:liegroups-ljac-pushforward}
  &\gexphat{G}\left(\mathbf{a} + \mathbf{b} \right)\ =\  \gexphat{G}\left(\gljac{G}\left(\mathbf{a}\right)\ \mathbf{b}\right) \gexphat{G}\left(\mathbf{a} \right),\\
 \label{eq:chap02:liegroups-ljac-bch}
  &\glogvee{G}\left(\gexphat{G}\left(\mathbf{b} \right) \gexphat{G}\left(\mathbf{a} \right) \right)\ \approx\ \mathbf{a}\ +\ \left(\gljac{G}\left(\mathbf{a}\right)\right)^{-1}\ \mathbf{b}.
\end{align}

\subsection{Invariant Error}
In control and estimation, we often define error residuals or cost functions as a difference between the true state and the desired or observed state, respectively.
Such error definitions are then useful, in control, for driving the system towards the desired state using the feedback of the current state, while it is used in estimation to get an estimate that closely describes a prediction or an observation of the current state.
This is usually done so by the minimizing the error residuals or the cost functions.
An appropriate choice of these errors is then often useful for formulating non-linear observers and control algorithms. \looseness=-1
It is straightforward to construct errors for variables evolving in a vector space due to its additive nature, 
$
\mathbf{e} = \mathbf{x} - \hat{\mathbf{x}},
$
where, $\mathbf{e} \in \R^p$ is a linear state-error between the true state $\mathbf{x} \in \R^p$ and the estimated state $\mathbf{\hat{x}} \in \R^p$.

The errors for variables evolving in matrix Lie groups are constructed with the help of the composition operator which is the matrix multiplication and the inverse operator which is the matrix inversion.
For matrix Lie groups, there is a possibility of constructing \emph{invariant errors} to compare trajectories. 
These are errors that remain the same even for a transformed system under a left- or right-group action, i.e. the error is invariant to a transformation or changes to the system introduced by left- or right-translations.

\begin{definition}[Left and right invariant error of type 1]
For $\mathbf{S}, \Y \in G$, given the left translation $L_\mathbf{S}: G \mapsto G, \ L_\mathbf{S} \Y = \mathbf{S}\ \Y$ and the right translation $R_\mathbf{S}: G \mapsto G, \ R_\mathbf{S} \Y = \Y\ \mathbf{S}$, the invariant errors between a true state $\X \in G$ and an estimated state $\Xhat \in G$ are,
\begin{align}
\label{def:left-invariant-error}
&\errG^L = \Xhat^{-1}\ \X = \left(L_\mathbf{S}\Xhat \right)^{-1} \left(L_\mathbf{S} \X \right) = \left(\mathbf{S}\Xhat \right)^{-1} \left(\mathbf{S} \X \right) = \Xhat^{-1} \mathbf{S}^{-1} \mathbf{S} \X \quad \text{(left-invariant)}, \\
\label{def:right-invariant-error}
&\errG^R = \X\ \Xhat^{-1} = \left(R_\mathbf{S}\X \right) \left(R_\mathbf{S} \Xhat \right)^{-1} = \left(\X \mathbf{S} \right) \left(\Xhat \mathbf{S} \right)^{-1} = \X \mathbf{S} \mathbf{S}^{-1} \Xhat^{-1} \quad \text{(right-invariant)}.
\end{align}
\end{definition}

\begin{definition}[Left and right invariant error of type 2]
The invariant errors between a true state $\X \in G$ and an estimated state $\Xhat \in G$ can also be defined as,
\begin{align}
\label{def:left-invariant-error-type2}
&\errG^L = \X^{-1}\ \Xhat = \left(L_\mathbf{S}\X \right)^{-1} \left(L_\mathbf{S} \Xhat \right) = \left(\mathbf{S}\X \right)^{-1} \left(\mathbf{S} \Xhat \right) = \X^{-1} \mathbf{S}^{-1} \mathbf{S} \Xhat \quad \text{(left-invariant)}, \\
\label{def:right-invariant-error-type2}
&\errG^R = \Xhat\ \X^{-1} = \left(R_\mathbf{S}\Xhat \right) \left(R_\mathbf{S} \X \right)^{-1} = \left(\Xhat \mathbf{S} \right) \left(\X \mathbf{S} \right)^{-1} = \Xhat \mathbf{S} \mathbf{S}^{-1} \X^{-1} \quad \text{(right-invariant)}.
\end{align}
\end{definition}

In the context of Kalman filtering on Lie groups, the choice of error affects the evolution of linearized error variables through the filter equations, thereby affecting the manner in which the covariance is propagated through time-updates and measurement-updates. 
Thus, a proper choice of error often helps in formulating an estimator with strong convergence and consistency properties. \looseness=-1

The Lie group operators for most commonly used matrix Lie groups in robotics are described in Appendix \ref{appendix:chapter02:examples-matrix-lie-groups}. 
Having defined the notion of an error in matrix Lie groups, in the following section we define the notion of uncertainty over a matrix Lie group using the so-called \emph{Concentrated Gaussian Distributions} (CGD) on matrix Lie groups.\looseness=-1


\section{Uncertainty Representation in Matrix Lie Groups}
\label{sec:chapter02:uncertainty-matrix-lie-groups}

The group of rotations and rigid body transformations are particularly important groups that provide a representation of spatial relationships for a rigid body moving in three-dimensional space (for definitions, see Appendix \ref{appendix:chapter02:examples-matrix-lie-groups}).
Usually in robotics, estimating uncertain spatial relationships is a significant problem which requires establishing the notion of uncertainty over such matrix Lie groups.
In general, the notions of uncertainty over matrix Lie groups are different from the standard notion of uncertainty described in vector spaces.
The notion of uncertainty can be introduced with the concepts of random variables and their associated mean and covariance.

A random variable in the vector space $\mathbf{x} \in \R^p$ can be associated with a probability density function (pdf),  $p(\mathbf{x})$ that satisfies,
$$
\forall \mathbf{x} \in \R^p, \quad p(\mathbf{x}) \geq 0,\ \int_{\R^p} p(\mathbf{x}) d\mathbf{x} = 1,
$$
which is also characterized by the mean,
$$
\mu = \expectation{\mathbf{x}} = \int_{\R^p} \mathbf{x}  p(\mathbf{x}) d\mathbf{x},
$$ 
and the covariance about the mean,
$$\hm{\Sigma} = \expectation{\err\ \err^T} = \int_{\R^p} \left(\mathbf{x} - \mu\right)\ \left(\mathbf{x} - \mu\right)^T p(\mathbf{x}) d\mathbf{x},
$$
where, we use $\expectation{.}$ to denote the expectation operator, and $\err = \mathbf{x} - \mu$ as a zero-mean perturbation acting on the noise-free value $\mu$.
Focusing solely on Gaussian distributions, the density function in $\R^p$ can be defined as,
$$
p(\mathbf{x}) = \frac{1}{\sqrt{(2 \pi)^p \det \hm{\Sigma}}} \text{exp}\left(-\frac{1}{2} \err^T \hm{\Sigma}^{-1} \err\right)
$$
\looseness=-1
This notion of additive uncertainty given by $\mathbf{x} = \mu + \err$, where $\err \sim \mathscr{N}_{\R^p}(\Zeros{p}{1}, \hm{\Sigma})$ may not be directly applicable for matrix Lie groups.
This is because Lie groups are characterized by constrained nonlinear spaces as seen in Section \ref{sec:chapter02:intro-matrix-lie-groups} and directly applying the above-mentioned definition of uncertainty might break the group structure, i.e. the evolution of random variable might no longer belong to the group.

The random variables in a matrix Lie group can be defined as,
\begin{align}
 \X = \Xhat \gexphat{G}\ \left(\err\right), \quad \text{(perturbations $\err$ applied locally)}, \\ 
 \X = \gexphat{G}\ \left(\err\right) \Xhat,   \quad \text{(perturbations $\err$ applied globally)},
\end{align}
 where, $\Xhat \in G$ is a large, noise-free value and $\err \in \R^p$ is a small perturbation induced into the group.

A simple and commonly used method to express uncertainty on matrix Lie groups is to use the concept of Concentrated Gaussian Distribution (CGD).
A CGD can be defined for a connected, unimodular matrix Lie group.
The property of connectedness for a matrix Lie group shows the existence of continuous paths in $G$ from any element $\X \in G$ to the identity element in $G$ (\cite{hall2013lie}, Def. 1.9).
The property of unimodularity (\cite{wolfe2011bayesian}) defines a bi-invariant integration measure $d\X$ for the matrix Lie group such that the concept of probability densities $p(\X)$ makes sense in the integral,
$$
\int_G  p(\X) d\X\ =\ \int_G  p(\Y \circ \X) d\X\ =\ \int_G  p(\X \circ \Y) d\X\ =\ 1,
$$
for any fixed $\Y \in G$ and $\circ$ is the group operation (matrix multiplication in this case).

Consider the small perturbation $\err \sim \mathscr{N}_{\R^p}\left(\mathbf{0}_{p \times 1}, \cov\right)$ to be tightly focused around a region $\err \in S \subset \R^p$ with a zero-mean Gaussian distribution with covariance $\cov$.
When the probability mass of $\err$ is tightly focused or close to one within the region $S$, the following approximation remains valid,
$$
\glogvee{G}\left(\gexphat{G}\left(\err\right)\right) = \err.
$$

\begin{figure}[!t]
\centering
\includegraphics[scale=0.5]{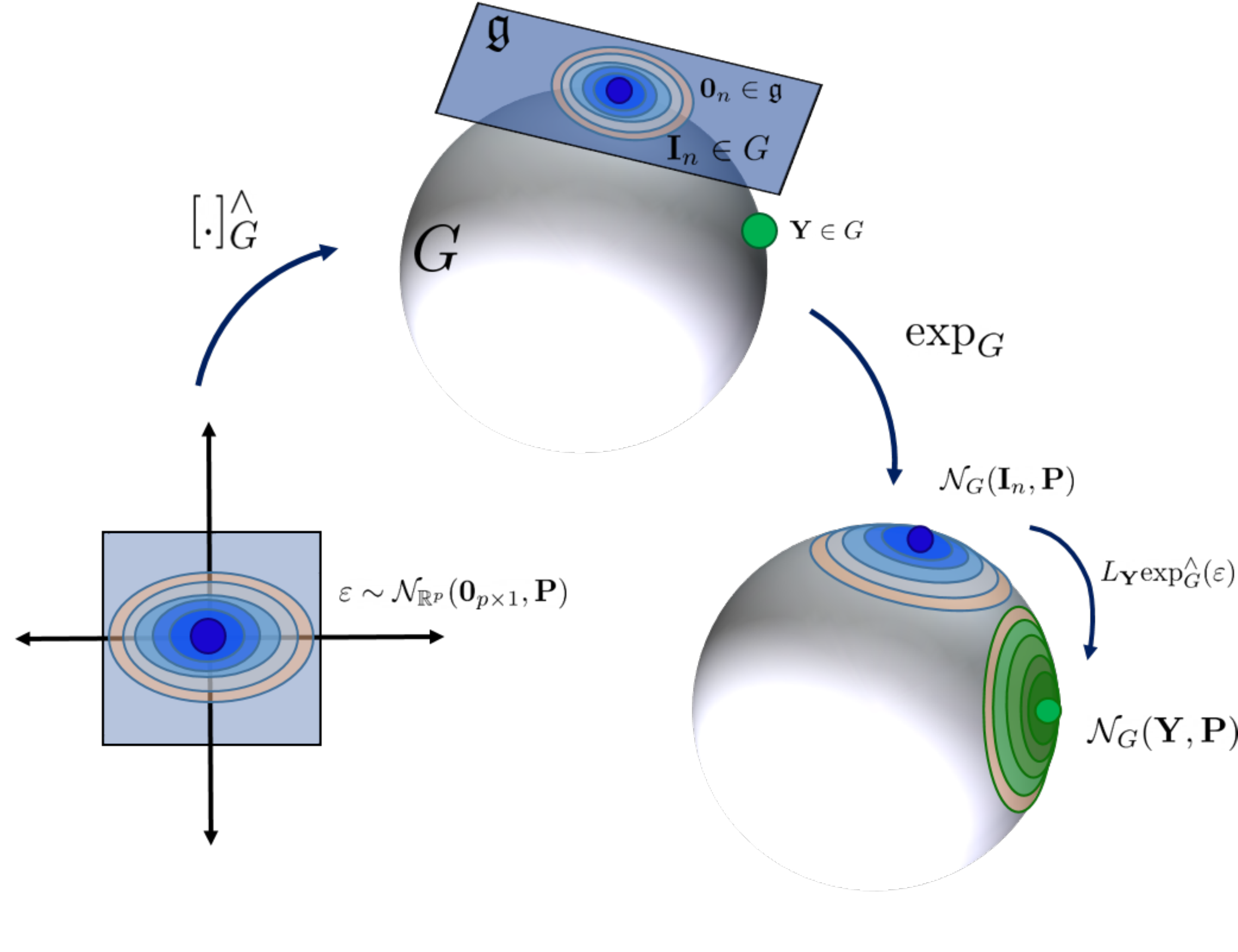}
\caption{Concentrated Gaussian distributions on matrix Lie groups. This figure demonstrates how a distribution in the vector space induces a distribution in group G.}
\label{fig:chap:chap:state-est:uncertainty-cgd}
\end{figure}

As described in Figure \ref{fig:chap:chap:state-est:uncertainty-cgd}, the probability distribution of $\err \in \R^p$ induces a CGD on $G$ around its identity element, which can be transported around $\Xhat \in G$ using the left action 
$$
L_{\Xhat} \gexphat{G}\left(\err\right) = \Xhat \gexphat{G}\left(\err\right),
$$ 
producing a CGD around $\Xhat$, where $\Xhat$ is the mean of $\X$ and $\err$ is the zero-mean Lie algebraic error with covariance $\cov$.
The eigenvalues of the covariance matrix $\cov$ can be used to characterize the uncertainty related to the variable, where these values represent the variance of the distribution in the direction of eigenvectors of $\cov$.
The major assumption in the definition of CGD is that the maximum eigenvalues of $\cov$ is required to be sufficiently small, such that the Gaussian distribution is not very spread out within/beyond the region $S$.

\begin{definition}[Concentrated Gaussian Distribution on Matrix Lie groups]
\label{def:chap02:CGD}
Concentrated Gaussian distribution on a matrix Lie group $G$ is a distribution $\mathscr{N}_G(\Xhat, \cov)$ for which the mean $\Xhat$ is defined on the group $G$ and the covariance $\cov$ is defined on the Lie algebra $\mathfrak{g}$, such that the distribution defined by perturbation vector $\err \sim \mathscr{N}_{\R^p}(\Zeros{n}{1}, \cov)$ in the vector space associated to the Lie algebra remains sufficiently close to zero.
\end{definition}

For such a definition, the mean lies on the Lie group while the covariance is strictly defined on the Lie algebra. 
Furthermore, here we have considered a left-trivialized perturbation (counterintuitively, the perturbation is applied on the right), meaning that the perturbation is acting locally, which leads to the use of the left action $\mathbf{L}_\Xhat$ for transporting the distribution in $G$ to $\Xhat$.
The choice of right-trivialized perturbation (perturbation applied on the left), meaning the perturbation is acting globally, leads to the use of the right action for transporting the distribution in $G$.

To summarize the notion of CGD in simple words, a zero-mean Gaussian distribution can be defined on the vector space such that the spread of this distribution is sufficiently small.
This distribution can be expressed straightforwardly around the zero element of the Lie algebra using the \emph{hat} operator (See Eq. \ref{eq:chap02:liegroups-generators} for the relation between hat operator and Lie algebra).
Samples obtained from the distribution on the Lie algebra produce perturbation elements on the group $G$, through the exponential map, which remain closed to the group identity (see Fig. \ref{fig:chap:chap:state-est:uncertainty-cgd}). 
These perturbation elements can then be transported from the identity element to the mean using the left- or right-action of the mean element on the perturbation element.


In order to change variables from $\err$ and express a pdf over $\X$, it is necessary to define an infinitesimal volume $d\X$ of $\X$.
The infinitesimal volume element $d\err$ is related to $d\X$ through the determinant of the right Jacobian of matrix Lie group, $d\X = \lvert \det\ \grjac{G}\rvert\ d\err$  (\cite{chirikjian2011stochastic, park1991optimal}). \looseness=-1
Considering the left-invariant error of type 1 $\errG = \Xhat^{-1} \X$ (see Eq. \ref{def:left-invariant-error}), the pdf induced in $G$ by the pdf in $\R^p$ can be observed through a change of coordinates using the relation $\err = \glogvee{G}(\Xhat^{-1} \X)$,
\begin{align}
\label{eq:chap02:induced-cgd}
    \begin{split}
        1 &= \int_{\R^p} \alpha \text{exp}\left(-\frac{1}{2} \err^T \cov^{-1} \err \right) d\err \\
        &= \int_G \alpha \exp\left(-\frac{1}{2} \glogvee{G}\left(\Xhat^{-1}\ \X\right)^T \cov^{-1}\glogvee{G}\left(\Xhat^{-1}\ \X\right) \right) \frac{d\X}{\lvert \det\ \grjac{G}\rvert} \\
        &= \int_G \beta \gexphat{G}\left(-\frac{1}{2} \glogvee{G}\left(\Xhat^{-1}\ \X\right)^T \cov^{-1} \glogvee{G}\left(\Xhat^{-1}\ \X\right) \right) d\X \\
        &= \int_G p(\X) d\X,
    \end{split}
\end{align}
where, the normalizing factors are $\alpha = \frac{1}{\sqrt{(2 \pi)^p \det \cov}}$ and 
$$
\beta = \frac{\alpha}{\lvert \det\ \grjac{G}\rvert} = \frac{1}{\sqrt{(2 \pi)^p \det\left(\grjac{G}\ \cov\ {\grjac{G}}^T \right)}},
$$
with $\grjac{G} \triangleq \grjac{G}\left(\err\right)$ being the right Jacobian of the matrix Lie group (due to the choice of the error as a consequence of BCH formula).
The dependence of the normalization factor $\beta$ on $\X$ through $\grjac{G}$ prevents  $p(\X)$ to be a valid Gaussian distribution. 
However, when the perturbation $\err$ is small, the Jacobian of the matrix Lie group remains close to identity resulting in a reasonable approximation of the Gaussian distribution. 
Interestingly. unimodular matrix Lie groups have the property that $\lvert \det\ \grjac{G}\rvert = \lvert \det\ \gljac{G}\rvert$ (\cite{chirikjian2011stochastic}), which is true for the case of most matrix Lie groups of interest in robotic applications, for example $\SO{3}, \SE{3}, \SL{3}$.

\begin{remark}
It must be noted, if we define the concentrated Gaussian distribution starting from the perturbation vector $\err$ which consequently induces the distribution in $G$ as shown in Eq. \eqref{eq:chap02:induced-cgd}, then the assumption of unimodularity of the Lie group is relaxed due to the configuration-dependent normalization factor $\beta$.
\end{remark}

\cite{barfoot2014associating}, \cite{wolfe2011bayesian}, \cite{wang2006error} and  \cite{su1992manipulation} all provide slightly modified accounts of this distribution for describing the uncertainty on matrix Lie groups. 
\cite{barfoot2014associating} and \cite{su1992manipulation} both start by defining $p(\epsilon)$ directly in $\R^p$ and then inducing $p(\X)$ through $\X = \Xhat \gexphat{G}\left(\epsilon\right)$ leading to the normalization factor $\beta$ dependent on $\X$. 
The other works define $p(\X)$ directly and choose a normalization factor $\beta^\prime$ which is constant in order to obtain a valid pdf. 
It must be noted that the definition by \cite{barfoot2014associating} is a more intuitive and sufficient representation for most robotic applications while being computationally efficient.
The definition by \cite{barfoot2014associating} is used within this thesis.
\section{Extended Kalman Filtering using Matrix Lie Groups}
\label{sec:chapter02:ekf-matrix-lie-groups}

This section follows the derivation of a discrete Extended Kalman Filter (EKF) over Lie groups described in \cite{bourmaud2013discrete}, in which both the state and the observations are evolving over different Lie groups $G$ and $G^\prime$, respectively.

Consider a discrete dynamical system for which both the state and the  measurements are evolving over distinct matrix Lie groups, \looseness=-1
\vspace{-2mm}
\begin{align}
\label{eq:chap02:ekf-lie-group-system}
\X_\knext &=  \X_\kcurr \; \gexphat{G}\left(\Omega\left(\X_\kcurr, \mathbf{u}_\kcurr\right) + \mathbf{w}_\kcurr\right), \\
\label{eq:chap02:ekf-lie-group-meas}
\Z_\kcurr &= h (\X_\kcurr) \; \gexphat{G^\prime}\left(\mathbf{n}_\kcurr\right).
\end{align}

\noindent The function $\Omega : G \times \R^m \to \R^p$ is the left trivialized velocity (or the motion model) of the matrix Lie group expressed as a function of the state $\X_\kcurr \in G$ and an exogenous control input $\mathbf{u}_\kcurr \in \R^m$ at time instant $\kcurr$.
We use $\mathbf{w}_\kcurr\sim \mathscr{N}_{\R^p}(\mathbf{0}_{p \times 1},  \Q_\kcurr)$ to denote discrete-time Gaussian white noise with covariance $\Q_\kcurr \in \R^{p \times p}$ acting on the motion model. \looseness=-1

The observations $\Z_\kcurr$ are considered to be evolving over a matrix Lie group $G^\prime$ of dimensions $q$  distinct from the state space $G$. $h: G \to G^\prime$ is the measurement model mapping the states $\X \in G$ to the space of observations $G^\prime$. We use $\mathbf{n}_\kcurr\sim\mathscr{N}_{\R^{q}}(\mathbf{0}_{q\times 1}, \; \N_\kcurr)$ to denote the measurement noise described as a discrete-time Gaussian white noise with covariance $\N_\kcurr$ defined in the $q$-dimensional vector space of the observations.\looseness=-1

The procedure for deriving the EKF on matrix Lie groups is similar to the two-step process described for Euclidean vectors, as already seen in the Section \ref{sec:chapter02:state-estimation-primer-ekf}.
This problem involves finding the optimal estimate $\Xhat$ given a set of observations $Z_l = {\Z_1,\dots, \Z_l}$ at time instants $l = \kcurr$ for the propagation and $l = \knext$ for the update.
The estimated state at each step is given by the mean $\Xhat_\knextgivenl$ of the conditional probability and its covariance is $\cov_\knextgivenl$. \looseness=-1 \looseness=-1
\begin{align}
\label{eq:chap02:ekf-posterior-distribution}
\begin{split}
    &p(\X_\knext\; \lvert\; \Z_1 \dots \Z_l) \approx \mathscr{N}_G(\Xhat_\knextgivenl, \;\cov_\knextgivenl), \\
    &\X_\knextgivenl = \Xhat_\knextgivenl\; \gexphat{G}\left({\err_\knextgivenl}\right), \\
    &\err_\knextgivenl \sim \mathscr{N}_{\R^p}(\mathbf{m}_\knextgivenl = \mathbf{0}_{p \times 1}, \cov_\knextgivenl), 
\end{split}
\end{align}
where, $p(\X_\knext\; \lvert\; \Z_1 \dots \Z_l)$ is the conditional probability distribution characterized as a concentrated Gaussian distribution.

\begin{remark}
The invariant errors considered in this section are of type 1 (Eq. \eqref{def:left-invariant-error} and Eq. \eqref{def:right-invariant-error}).
\end{remark}

In the following subsections, we will derive the discrete extended Kalman filter equations by considering left-invariant error formulation of type 1 ($\errG^L = \Xhat^{-1} \X$) and right-invariant error formulation of type 1 ($\errG^R = \X \Xhat^{-1}$) respectively, for systems whose state and measurements are evolving over distinct matrix Lie groups.

\subsection{EKF using left-invariant error formulation of type 1}
In this subsection, we will understand how the mean and covariance of the CGD are propagated using the system dynamics in the propagation step and updated through measurements in the update step, considering  the left-invariant error formulation of type 1 denoted as $\errG = \errG^L = \Xhat^{-1} \X$ (see Eq. \ref{def:left-invariant-error}) for defining the error between the true state $\X$ and the estimate $\Xhat$.
This leads to the \emph{Discrete Lie Group Extended Kalman Filter} (abbreviated as DILIGENT) whose filter equations are summarized in Algorithm \ref{algo:chap02:dlgekf}.
Even before getting into the details of the derivation, it can already be seen that the filter structure of DILIGENT resembles that of the standard EKF. 

\begin{remark}
The algorithmic structure reduces to the regular EKF algorithm if $G$ and $G^\prime$ are considered as Euclidean spaces.
\end{remark}

\begin{algorithm}[tpb]
\small
  \caption{Discrete extended Kalman filter on matrix Lie groups with left-invariant error formulation of type 1 \looseness=-1}
  \label{algo:chap02:dlgekf}
  \begin{algorithmic}
    \State \textbf{Input:} $\Xhat_\kprior, \cov_\kprior, \Z_\knext, \mathbf{u}_\kcurr$ \\ 
    \textbf{Output:} $\Xhat_\kest, \cov_\kest$ \\
    \State \textbf{Propagation:}\\
    $\Xhat_\kpred = \Xhat_\kprior\; \gexphat{G}\left(\hat{\Omega}_\kcurr\right)$ \\
    $\cov_\kpred =\ \F_\kcurr\ \cov_\kprior\ \F_\kcurr^T\ +\ \grjac{G}\left(\hat{\Omega}_\kcurr\right)\ \Q_\kcurr\ \grjac{G}\left(\hat{\Omega}_\kcurr\right)^T$ \\
    \State \textbf{Update:} \\
    $\mathbf{\tilde{z}}_\knext =  \glogvee{G^\prime}\left( h^{-1}\left(\Xhat_\kpred\right)\ \Z_\knext\right)$ \\
    $\mathbf{K}_\knext = \cov_\kpred\ \mathbf{H}_\knext^T\left(\mathbf{H}_\knext\;\cov_\kpred\;\mathbf{H}_\knext^T \;+\; \mathbf{N}_\knext\right)^{-1}$ \\
    $\mathbf{m}_\knext^{-} = \mathbf{K}_\knext\; \mathbf{\tilde{z}}_\knext$ \\
    $\Xhat_\kest\ =\ \Xhat_\kpred \gexphat{G}\left(\mathbf{m}_\knext^{-}\right)$ \\
    $\cov_\kest =  \grjac{G}\left(\mathbf{m}_\knext^{-}\right) \left(\I{p} \;-\; \mathbf{K}_\knext\;\mathbf{H}_\knext\right)\;\cov_\kpred\;  \grjac{G}\left(\mathbf{m}_\knext^{-}\right)^T$\\

    \State where, \\
    $\hat{\Omega}_\kcurr = \Omega\left(\Xhat_\kprior, \mathbf{u}_\kcurr\right)$ \\
    $\F_\kcurr\ = \gadj{\gexphat{G}\left(-\hat{\Omega}_\kcurr\right)} + \grjac{G}\left(\hat{\Omega}_\kcurr\right)\ \mathfrak{F}_\kcurr$ \\
    $\mathfrak{F}_\kcurr = \frac{\partial}{\partial \err} \Omega\left(\Xhat_\kprior\ \gexphat{G}\left({\err_\kprior}\right), \mathbf{u}_\kcurr\right)_{\err_\kprior = 0}$ \\
    $\mathbf{H}_\knext = \frac{\partial}{\partial \err} \glogvee{G^\prime}\left( h^{-1}\left(\Xhat_\kpred\right)\ h\left(\Xhat_\kpred\ \gexphat{G}\left(\err_\kpred\right)\right)\right)_{\err=0}$ \\ \\

    \vspace{-1mm}
  \end{algorithmic}
  \vspace{-2mm}
\end{algorithm}

\subsubsection{Propagation}
We first derive the propagation of the mean and covariance through the prediction model described by the system dynamics.
As already discussed in Section \ref{sec:chapter02:state-estimation-primer-ekf}, this preliminary step involves using the state estimate
from the previous time instant to produce a predicted estimate of the state at the current time instant with the help of the prediction model.
The distribution of the state in the propagation step  with discrete time step $l = \kcurr$ is given by $\mathscr{N}_G(\Xhat_\kprior, \;\cov_\kprior)$.
The mean and the covariance of the posterior distribution is propagated to obtain the distribution $\mathscr{N}_G(\Xhat_\kpred, \;\cov_\kpred)$ between two sensor measurements in the interval $[\kcurr, \knext)$.

The propagation of the mean can be directly obtained from the system dynamics in Eq. \eqref{eq:chap02:ekf-lie-group-system}. \looseness=-1
\vspace{-2mm}
\begin{align}
\label{eq:chap02:ekf-mean-propagation}
\begin{split}
    &\Xhat_\kpred = \Xhat_\kprior\; \gexphat{G}\left(\hat{\Omega}_\kcurr\right), \quad \hat{\Omega}_\kcurr = \Omega\left(\Xhat_\kprior, \mathbf{u}_\kcurr\right).
\end{split}        
\end{align}

We will now derive the linearized error dynamics and propagation of covariance using the prediction model which depends on the choice of error.
The linearized error or the error in the vector representation of Lie algebra is given as, $\err = \glogvee{G}\left(\errG\right)$, from which the covariance can be directly computed as $\expectation{\err \err^T}$.
At the end of this step, the aim is to have the mean $\mathbf{m}_\kpred =  \expectation{\err_\kpred}$ of propagated linearized error which must be equivalent to a zero vector $\Zeros{p}{1}$ and a state-error covariance $\cov_\kpred = \expectation{\err_\kpred \err_\kpred^T} $. \looseness=-1

Given the true state $\X_\knext$ and the predicted state estimate $\Xhat_\kpred$. we know that, 
$$
\errG_\kpred = \Xhat^{-1}_\kpred \; \X_\knext = \gexphat{G}\left({\err_\kpred}\right).
$$

Substituting Eq. \eqref{eq:chap02:ekf-lie-group-system} and Eq. \eqref{eq:chap02:ekf-mean-propagation} in $\errG_\kpred$, we have,
\begin{align}
\begin{split}
    \errG_\kpred &= \left(\Xhat_\kprior\; \gexphat{G}\left(\hat{\Omega}_\kcurr\right)\right)^{-1} \; \X_\knext \\
    & = \left(\gexphat{G}\left(\hat{\Omega}_\kcurr\right)\right)^{-1} \left(\Xhat_\kprior\right)^{-1}  \; \X_\kcurr \; \gexphat{G}\left(\Omega\left(\X_\kcurr, \mathbf{u}_\kcurr\right) + \mathbf{w}_\kcurr\right) \\
    & = \gexphat{G}\left(-\hat{\Omega}_\kcurr\right) \errG_\kprior\;\gexphat{G}\left(\Omega_\kcurr + \mathbf{w}_\kcurr\right) \\
    &  = \gexphat{G}\left(-\hat{\Omega}_\kcurr\right) \; \gexphat{G}\left({\err_\kprior}\right)\;\gexphat{G}\left(\Omega_\kcurr + \mathbf{w}_\kcurr\right),
\end{split}
    \label{eq:chap02:ekf-predicted-error}
\end{align}
where, $\Omega_\kcurr \triangleq \Omega\left(\X_\kcurr, \mathbf{u}_\kcurr\right)$.
Here, the identities, $\forall\ \X, \Y \in \R^{n\times n}, (\X \Y)^{-1} =\; \Y^{-1}\X^{-1}$ and $\forall\ t\in \R,\ \A \in \mathfrak{g},\ \left(\exp(t \A)\right)^{-1} =\ \exp(- t \A)$ have been used.

Using the adjoint identity from Eq. \eqref{eq:chap02-liegroups-adjoint-repr}, 
\begin{align}
\label{eq:chap02:ekf-err-as-poe}
\begin{split}
     \errG_\kpred &= \gexphat{G}\left(\gadj{\gexphat{G}\left(-\hat{\Omega}_\kcurr\right)} \err_\kprior \right)\; \gexphat{G}\left(-\hat{\Omega}_\kcurr\right)\;\gexphat{G}\left(\Omega_\kcurr + \mathbf{w}_\kcurr\right).
\end{split}
\end{align}

Given $\X_\kcurr = \Xhat_\kprior \gexphat{G}\left(\err_\kprior\right)$, $\Omega_\kcurr$ can be linearized around $\Xhat_\kprior$ using the Taylor series expansion up to the first-order as,
\begin{align}
\label{eq:chap02:ekf-omega-linearized}
\begin{split}
     \Omega_\kcurr &= \Omega\left(\Xhat_\kprior\ \gexphat{G}\left({\err_\kprior}\right), \mathbf{u}_\kcurr\right) \\
     &\approx \Omega\left(\Xhat_\kprior, \mathbf{u}_\kcurr\right)  + \mathfrak{F}_\kcurr\; \err_\kprior\ =\ \hat{\Omega}_\kcurr + \mathfrak{F}_\kcurr\; \err_\kprior,
\end{split}
\end{align}
where, $\mathfrak{F}_\kcurr = \frac{\partial}{\partial \err} \Omega\left(\Xhat_\kprior\ \gexphat{G}\left({\err_\kprior}\right), \mathbf{u}_\kcurr\right)_{\err_\kprior = 0}$ is the Jacobian of the left trivialized motion model $\Omega_\kcurr$ at the current state estimate $\Xhat_\kprior$ with an infinitesimal additive perturbation $\err_\kprior$ in the vector space. \looseness=-1

On substituting Eq. \eqref{eq:chap02:ekf-omega-linearized} in Eq. \eqref{eq:chap02:ekf-err-as-poe} and using the Jacobian identity from Eq. \eqref{eq:chap02:liegroups-pre-push-forward}, we have, \looseness=-1
\begin{align}
    \begin{split}
        \errG_\kpred &= \gexphat{G}\left({\err_\kpred}\right) \\
        &= \gexphat{G}\left(\gadj{\gexphat{G}\left(-\hat{\Omega}_\kcurr\right)} \err_\kprior \right)\; \gexphat{G}\left(-\hat{\Omega}_\kcurr\right)\;\gexphat{G}\left(\hat{\Omega}_\kcurr + \mathfrak{F}_\kcurr\; \err_\kprior  + \mathbf{w}_\kcurr\right) \\
        &= \gexphat{G}\left(\gadj{\gexphat{G}\left(-\hat{\Omega}_\kcurr\right)} \err_\kprior \right)\; \gexphat{G}\left(\grjac{G}\left(\hat{\Omega}_\kcurr\right)\ \left( \mathfrak{F}_\kcurr\; \err_\kprior  + \mathbf{w}_\kcurr\right) \right).
    \end{split}
    \label{eq:chap02:ekf-err-prop-bch-applied}
\end{align}

On applying logarithm map on both sides and using the truncated BCH formula (Def. \ref{def:chap02:BCH-formula}) up to first-order terms, we obtain,
\begin{align}
\begin{split}
    \err_\kpred &= \gadj{\gexphat{G}\left(-\hat{\Omega}_\kcurr\right)} \err_\kprior + \grjac{G}\left(\hat{\Omega}_\kcurr\right)\  \left( \mathfrak{F}_\kcurr\; \err_\kprior  + \mathbf{w}_\kcurr\right) \\
    &= \left(\gadj{\gexphat{G}\left(-\hat{\Omega}_\kcurr\right)} + \grjac{G}\left(\hat{\Omega}_\kcurr\right)\ \mathfrak{F}_\kcurr \right) \err_\kprior + \grjac{G}\left(\hat{\Omega}_\kcurr\right) \mathbf{w}_\kcurr \\
    &= \F_\kcurr\ \err_\kprior + \grjac{G}\left(\hat{\Omega}_\kcurr\right) \mathbf{w}_\kcurr,
\end{split}
\label{eq:chap02:ekf-err-prop-final}
\end{align}
where, $\F_\kcurr\ = \gadj{\gexphat{G}\left(-\hat{\Omega}_\kcurr\right)} + \grjac{G}\left(\hat{\Omega}_\kcurr\right)\ \mathfrak{F}_\kcurr$ is the linearized error-propagation matrix accounting for the proper transport of the error in the Lie algebra to an appropriate tangent space for the associated error propagation, while considering any effects on the error due to perturbations on the motion model.

The mean and the covariance of the predicted error can be obtained by applying the $\expectation{.}$ operator and by ignoring negligible terms such as the product of noises and linearized error,
\begin{align}
\begin{split}
    \mathbf{m}_\kpred &= \expectation{\err_\kpred}\ =\ \mathbf{0}_{p \times 1},\\
    \cov_\kpred &= \expectation{\err_\kpred\ \err_\kpred^T}\ =\ \F_\kcurr\ \cov_\kprior\ \F_\kcurr^T\ +\ \grjac{G}\left(\hat{\Omega}_\kcurr\right)\ \Q_\kcurr\ \grjac{G}\left(\hat{\Omega}_\kcurr\right)^T.
\end{split}
\label{eq:chap02:ekf-errordyn}
\end{align}

\subsubsection{Update}
Having computed the predicted mean $\Xhat_\kpred$ and $\cov_\kpred$ in Eqs. \eqref{eq:chap02:ekf-mean-propagation} and \eqref{eq:chap02:ekf-errordyn} in the propagation step, the incoming measurements evolving over the group $G^\prime$ are incorporated in the update step to correct these quantities if there are any errors between the predicted state and the observed state.
This is done by defining an innovation term which is the error between the actual measurements and the expected measurements obtained by passing the predicted estimates into the measurement model.
The innovation term along with the adaptively tuned Kalman gain is then used to correct the measurement updates.
At the end of the update step, the aim is then to have the updated state estimate $\Xhat_\kest$,  its associated state error covariance $\cov_\kest$ for which the mean of linearized state error $\mathbf{m}^{-}_\knext = \Zeros{p}{1}$.

An innovation term is defined as the error between the expected and the actual measurements that can be written as,
\begin{align}
\begin{split}
    & \mathbf{\tilde{z}}_\knext =  \glogvee{G^\prime}\left( h^{-1}\left(\Xhat_\kpred\right)\ \Z_\knext\right).
\end{split}
\label{eq:chap02:ekf-innovation-defn}
\end{align}

Upon substituting Eq, \eqref{eq:chap02:ekf-lie-group-meas} in the innovation term defined in Eq. \eqref{eq:chap02:ekf-innovation-defn} and applying the left Jacobian-logarithm identity from Eq. \eqref{eq:chap02:liegroups-ljac-bch} with the consideration that the left Jacobian of zero-mean measurement noise is identity ($\gljac{G^\prime}\left(\mathbf{n}_\knext\right) \approx \I{q}$), the innovation is reduced as,
\begin{align}
\begin{split}
    \mathbf{\tilde{z}}_\knext &=  \glogvee{G^\prime}\left( h^{-1}\left(\Xhat_\kpred\right)\ h\left(\Xhat_\kpred\ \gexphat{G}\left(\err_\kpred\right)\right)\ \gexphat{G^\prime}\left(\mathbf{n}_\knext\right) \right) \\
    &= \mathbf{n}_\knext +  \glogvee{G^\prime}\left( h^{-1}\left(\Xhat_\kpred\right)\ h\left(\Xhat_\kpred\ \gexphat{G}\left(\err_\kpred\right)\right)\right).
\end{split}
\end{align}

By linearizing the measurement model about the state estimate $\Xhat_\kpred$ and neglecting higher-order terms, the innovation term can be approximated as,
\begin{align}
\begin{split}
    \mathbf{\tilde{z}}_\knext  &= \frac{\partial}{\partial \err} \glogvee{G^\prime}\left( h^{-1}\left(\Xhat_\kpred\right)\ h\left(\Xhat_\kpred\ \gexphat{G}\left(\err_\kpred\right)\right)\right)_{\err=0}  \err_\kpred\ +\ \mathbf{n}_\knext  \\
    &= \mathbf{H}_\knext\ \err_\kpred + \mathbf{n}_\knext.
\end{split}
\label{eq:chap02:ekf-innovation-update}
\end{align}
where, the measurement model Jacobian $\mathbf{H}_\knext$ is defined as,
\begin{align}
\begin{split}
    \mathbf{H}_\knext &= \frac{\partial}{\partial \err} \glogvee{G^\prime}\left( h^{-1}\left(\Xhat_\kpred\right)\ h\left(\Xhat_\kpred\ \gexphat{G}\left(\err_\kpred\right)\right)\right)_{\err=0}.
\end{split}
\end{align}

Since, the Eq. \eqref{eq:chap02:ekf-innovation-update} is linear in $\err_\kpred$,  the classical EKF update equations can be applied to obtain the posterior distribution $\err_\kest^{-} \sim \mathscr{N}_{\R^{p}}(\mathbf{m}_\knext^{-}, \cov_\kest^{-})$,
\begin{align}
\begin{split}
    & \mathbf{K}_\knext = \cov_\kpred\ \mathbf{H}_\knext^T\left(\mathbf{H}_\knext\;\cov_\kpred\;\mathbf{H}_\knext^T \;+\; \mathbf{N}_\knext\right)^{-1}, \\
    & \mathbf{m}_\knext^{-} = \mathbf{0}_{p \times 1} + \mathbf{K}_\knext\; \left(\mathbf{\tilde{z}}_\knext - \mathbf{H}_\knext\; \mathbf{0}_{p \times 1}\right), \\
    & \cov_\kest^{-} = (\I{p} \;-\; \mathbf{K}_\knext\;\mathbf{H}_\knext)\;\cov_\kpred.
\end{split}
\label{eq:chap02-ekf-update-eqns}
\end{align}

At the end of the update step, the mean of the conditional probability distribution $\expectation{\err_\kest^{-}} = \mathbf{m}_\knext^{-} \neq \mathbf{0}_{p \times 1}$, while the expected value needs to be $\expectation{\err_\kest} = \mathbf{0}_{p \times 1}$ to satisfy the definition of the concentrated Gaussian distribution.
Thus, a state reparametrization is required to satisfy this condition.
We have,
\begin{align}
\begin{split}
    \X_\knext &=\ \Xhat_\kpred \gexphat{G}\left(\err_\kest^{-}\right) \\
    &=\ \Xhat_\kpred \gexphat{G}\left(\mathbf{m}_\knext^{-}\ +\ \mathbf{r}_\knext^{-}\right),
\end{split}
\end{align}
where, $ \mathbf{r}_\knext^{-} \sim \mathscr{N}_{\R^p}\left(\mathbf{0}_{p \times 1}, \cov_\kest^{-}\right)$.
On using Eq. \eqref{eq:chap02:liegroups-rjac-bch},
\begin{align}
\begin{split}
    \X_\knext &=\  \Xhat_\kpred \gexphat{G}\left(\mathbf{m}_\knext^{-}\right) \gexphat{G}\left( \grjac{G}\left(\mathbf{m}_\knext^{-}\right) \mathbf{r}_\knext^{-} \right) \\
    &= \Xhat_\kest \gexphat{G}\left(\err_\kest\right),
\end{split}
\end{align}

where, $\Xhat_\kest\ =\ \Xhat_\kpred \gexphat{G}\left(\mathbf{m}_\knext^{-}\right)$ and $\err_\kest = \grjac{G}\left(\mathbf{m}_\knext^{-}\right) \mathbf{r}_\knext^{-}$. \looseness=-1
On reparametrization, the probability distribution is characterized by the mean and covariance,
\begin{align}
\begin{split}
    & \mathbf{m}_\kest = \expectation{\err_\kest} = \mathbf{0}_{p \times 1}, \\
    & \cov_\kest\ =\ \grjac{G}\left(\mathbf{m}_\knext^{-}\right)\ \cov_\kest^{-}\ \grjac{G}\left(\mathbf{m}_\knext^{-}\right)^T.
\end{split}
\end{align}

Thus, gathering the estimated states and covariances at the end of the propagation step and the updates steps along with a few intermediate Jacobian computations results in the overall algorithm of Discrete Lie Group Extended Kalman Filter as described in Algorithm \ref{algo:chap02:dlgekf}.

\subsection{Changes in the EKF equations for right-invariant error of type 1}
In this subsection, we will understand how the mean and covariance of the CGD are propagated and updated while considering  the right-invariant error formulation of type 1 denoted as $\errG = \errG^R = \X \Xhat^{-1}$ (see Eq. \ref{def:right-invariant-error}) for defining the error between the true state $\X$ and the estimate $\Xhat$.
This leads to the \emph{Discrete Lie Group Extended Kalman Filter with Right Invariant Error} (abbreviated as DILIGENT-RIE) whose filter equations are summarized in Algorithm \ref{algo:chap02:dlgekf-rie}.

\begin{algorithm}[tpb]
\small
  \caption{Discrete extended Kalman filter on matrix Lie groups with right-invariant error formulation of type 1 \looseness=-1}
  \label{algo:chap02:dlgekf-rie}
  \begin{algorithmic}
    \State \textbf{Input:} $\Xhat_\kprior, \cov_\kprior, \Z_\knext, \mathbf{u}_\kcurr$ \\ 
    \textbf{Output:} $\Xhat_\kest, \cov_\kest$ \\
    \State \textbf{Propagation:}\\
    $\Xhat_\kpred = \Xhat_\kprior\; \gexphat{G}\left(\hat{\Omega}_\kcurr\right)$ \\
    $\cov_\kpred =\ \F_\kcurr\ \cov_\kprior\ \F_\kcurr^T\ +\ \gadj{\Xhat_\kprior} \gljac{G}\left(\hat{\Omega}_\kcurr\right)\ \Q_\kcurr\ \gljac{G}\left(\hat{\Omega}_\kcurr\right)^T \gadj{\Xhat_\kprior}^T$ \\
    \State \textbf{Update:} \\
    $\mathbf{\tilde{z}}_\knext =  \glogvee{G^\prime}\left( h^{-1}\left(\Xhat_\kpred\right)\ \Z_\knext\right)$ \\
    $\mathbf{K}_\knext = \cov_\kpred\ \mathbf{H}_\knext^T\left(\mathbf{H}_\knext\;\cov_\kpred\;\mathbf{H}_\knext^T \;+\; \mathbf{N}_\knext\right)^{-1}$ \\
    $\mathbf{m}_\knext^{-} = \mathbf{K}_\knext\; \mathbf{\tilde{z}}_\knext$ \\
    $\Xhat_\kest\ =\ \gexphat{G}\left(\mathbf{m}_\knext^{-}\right) \Xhat_\kpred$   \\
    $\cov_\kest =  \gljac{G}\left(\mathbf{m}_\knext^{-}\right) \left(\I{p} \;-\; \mathbf{K}_\knext\;\mathbf{H}_\knext\right)\;\cov_\kpred\;  \gljac{G}\left(\mathbf{m}_\knext^{-}\right)^T$\\

    \State where, \\
    $\hat{\Omega}_\kcurr = \Omega\left(\Xhat_\kprior, \mathbf{u}_\kcurr\right)$ \\
    $\F_\kcurr\ = \I{p} + \gadj{\Xhat_\kprior} \gljac{G}\left(\hat{\Omega}_\kcurr\right)\ \mathfrak{F}_\kcurr$ \\
    $\mathfrak{F}_\kcurr = \frac{\partial}{\partial \err} \Omega\left(\gexphat{G}\left({\err_\kprior}\right) \Xhat_\kprior, \mathbf{u}_\kcurr\right)_{\err_\kprior = 0}$ \\
    $\mathbf{H}_\knext = \frac{\partial}{\partial \err} \glogvee{G^\prime}\left( h^{-1}\left(\Xhat_\kpred\right)\ h\left(\gexphat{G}\left(\err_\kpred\right) \Xhat_\kpred\ \right)\right)_{\err=0}$ \\ \\

    \vspace{-1mm}
  \end{algorithmic}
  \vspace{-2mm}
\end{algorithm}

When the right-invariant error is chosen, the error propagation, update and the state reparametrization are changed from the left-invariant error variant accordingly.
The error propagation equation can be written as,
\begin{align}
\begin{split}
    \errG_\kpred &= \X_\knext\ \Xhat_\kpred^{-1} \\ 
    & =\X_\kcurr \; \gexphat{G}\left(\Omega\left(\X_\kcurr, \mathbf{u}_\kcurr\right) + \mathbf{w}_\kcurr\right) \left(\Xhat_\kprior\; \gexphat{G}\left(\hat{\Omega}_\kcurr\right)\right)^{-1} \\
    & = \gexphat{G}\left(\err_\kprior\right)\ \Xhat_\kprior\ \gexphat{G}\left(\hat{\Omega}_\kcurr + \mathfrak{F}_\kcurr + \mathbf{w}_\kcurr\right)\ \gexphat{G}\left(-\hat{\Omega}_\kcurr\right)\ \Xhat_\kprior^{-1} \\
    & = \gexphat{G}\left(\err_\kprior\right)\ \Xhat_\kprior\ \gexphat{G}\left(\gljac{G}\left(\hat{\Omega}_\kcurr\right) \left(\mathfrak{F}_\kcurr\ \err_\kprior + \mathbf{w}_\kcurr\right)\right)\ \Xhat_\kprior^{-1} \\
    & = \gexphat{G}\left(\err_\kprior\right)\ \gexphat{G}\left(\gadj{\Xhat_\kprior}\gljac{G}\left(\hat{\Omega}_\kcurr\right) \left(\mathfrak{F}_\kcurr\ \err_\kprior + \mathbf{w}_\kcurr\right) \right).
\end{split}
    \label{eq:chap02:ekf-predicted-right-invariant-error}
\end{align}
where, Eqs. \eqref{eq:chap02:liegroups-ljac-pushforward} and \eqref{eq:chap02-liegroups-adjoint-repr} have been used along with the definition, 
$$\mathfrak{F}_\kcurr = \frac{\partial}{\partial \err} \Omega\left(\gexphat{G}\left({\err_\kprior}\right)\ \Xhat_\kprior, \mathbf{u}_\kcurr\right)_{\err_\kprior = 0}.$$
The linearized error propagation becomes,
\begin{align}
    \err_\kpred = \left(\I{p}\ +\ \gadj{\Xhat_\kprior}\gljac{G}\left(\hat{\Omega}_\kcurr \right) \mathfrak{F}_\kcurr \right) \err_\kprior\ +\ \gadj{\Xhat_\kprior}\gljac{G}\left(\hat{\Omega}_\kcurr \right)\ \mathbf{w}_\kcurr,
\end{align}
leading to the linearized error-propagation matrix $\mathbf{F}_\kcurr = \I{p}\ +\ \gadj{\Xhat_\kprior}\gljac{G}\left(\hat{\Omega}_\kcurr \right) \mathfrak{F}_\kcurr$.

With the choice of the innvoation in Eq. \eqref{eq:chap02:ekf-innovation-defn}, the only modification in the update equations is with the measurement model Jacobian which is defined as, $$\mathbf{H}_\knext = \frac{\partial}{\partial \err} \glogvee{G^\prime}\left( h^{-1}\left(\Xhat_\kpred\right)\ h\left(\gexphat{G}\left(\err_\kpred\right)\ \Xhat_\kpred\ \right)\right)_{\err=0}. $$

Finally, the state reparametrization follows,
\begin{align}
\begin{split}
    \X_\knext &=\ \gexphat{G}\left(\err_\kest^{-}\right)\ \Xhat_\kpred \\
    &=\ \gexphat{G}\left(\mathbf{m}_\knext^{-}\ +\ \mathbf{r}_\knext^{-}\right)\ \Xhat_\kpred \\
    &=\ \gexphat{G}\left(\gljac{G}\left(\mathbf{m}_\knext^{-}\right) \mathbf{r}_\knext^{-} \right)\ \gexphat{G}\left(\mathbf{m}_\knext^{-}\right)\  \Xhat_\kpred \\
    &= \gexphat{G}\left(\err_\kest\right)\ \Xhat_\kest,
\end{split}
\end{align}
leading to the update of mean and covariance as,
\begin{align}
\label{eq:chap02:ekf-rie-state-reparam}
\begin{split}
    &\Xhat_\kest\ =\ \gexphat{G}\left(\mathbf{m}_\knext^{-}\right)\ \Xhat_\kpred,\\ 
    &\err_\kest = \gljac{G}\left(\mathbf{m}_\knext^{-}\right) \mathbf{r}_\knext^{-},\\
    &\cov_\kest\ =\ \gljac{G}\left(\mathbf{m}_\knext^{-}\right)\ \cov_\kest^{-}\ \gljac{G}\left(\mathbf{m}_\knext^{-}\right)^T.
\end{split}
\end{align}

It must be noted that the left Jacobians are acting on the covariance matrix in this case differently from the right Jacobians from the case of left-invariant error formulation.
This is clearly due to the consequence of the expression of the perturbation vector, i.e. the perturbation is expressed in a global frame for a right-invariant error while the perturbation is expressed locally for the left-invariant error.

\section{Invariant Extended Kalman Filtering using Matrix Lie Groups}
\label{sec:chapter02:invekf-matrix-lie-groups}
For a particular class of continuous-time systems on matrix Lie groups with discrete vector observations, it is possible to derive a non-linear observer, known as the \emph{Invariant Extended Kalman Filter} (InvEKF), which often has stronger convergence properties than the standard EKF. 
In this section, we review the InvEKF derived for systems with continuous-time system dynamics and discrete observations, introduced in \cite{barrau2017invariant}.

It must be noted that for the derivation of InvEKF, we will use a definition of the dynamical system that is different from the definition in Eqs. \eqref{eq:chap02:ekf-lie-group-system} and \eqref{eq:chap02:ekf-lie-group-meas}.
The purpose and benefits of such a change in the design choice is described later in this section.
The state is modeled to evolve over matrix Lie groups along a continuous-time trajectory while the observations obtained at discrete time-instants evolve over vector spaces.
The considered class of dynamical systems evolving over matrix Lie groups is given as, \looseness=-1
\begin{align}
\label{eq:chap02:invekf-dynamical-system}
\frac{d}{dt} \X_t = f_{\mathbf{u}_t}\left(\X_t\right),
\end{align}
where, the state evolves over the Lie group $\X_t \in G$, $\mathbf{u}_t \in \R^m$ is the exogenous control input and the system dynamics is given by $f_{\mathbf{u}_t}\left(\X_t\right) \triangleq f\left(\X_t, \mathbf{u}_t\right)$.
If the system dynamics defined in Eq. \eqref{eq:chap02:invekf-dynamical-system} obeys a so-called \emph{group-affine} property (Eq. \eqref{eq:chap02:invekf-group-affine-dynamics}), then the resulting filter design has a key property related to the propagation of the state error through the linearized error dynamics associated with the system. 
The estimation error obeys an autonomous equation, i.e., the error propagation is independent of the state trajectory, leading to guaranteed local convergence properties.

\begin{remark}
The invariant errors considered in this section are of type 2, $\errG^L = \X^{-1}\ \Xhat$ and  $\errG^R = \Xhat\ \X^{-1}$ (See Eq. \ref{def:left-invariant-error-type2} and \ref{def:right-invariant-error-type2}).
\end{remark}

The choice of error, right-invariant or left-invariant, affects the observation structure for the discrete measurements.
The measurement models may fall into to the family of vector observations,
\begin{align}
    \label{eq:chap02:invekf-left-invariant-observation}
    & \mathbf{z}_{t_\kcurr} = \X_{t_\kcurr}\ \mathbf{b}\ +\ \mathbf{n}_{t_\kcurr} \quad \text{(left-invariant observation)}, \\
    \label{eq:chap02:invekf-right-invariant-observation}
    & \mathbf{z}_{t_\kcurr} = \X_{t_\kcurr}^{-1}\ \mathbf{b}\ +\ \mathbf{n}_{t_\kcurr} \quad \text{(right-invariant observation)},
\end{align}
where, $\mathbf{b} \in \R^n$ is a known, constant vector and $\mathbf{n}_{t_\kcurr} \in \R^n$ is a vector containing continuous-time white Gaussian noise with covariance $\mathbf{N}_\kcurr$ affecting the measurements.
The choice of right-invariant or left-invariant observation for filter design leads to the name Right-Invariant EKF (RIEKF) and Left-Invariant EKF (LIEKF), respectively in the literature (\cite{barrau2017invariant}).

The complete set of equations for the Invariant EKF on matrix Lie groups, using invariant error formulations of type 2 (see Eq. \ref{def:left-invariant-error-type2} and \ref{def:right-invariant-error-type2}) is summarized in Algorithm \ref{algo:chap02:dlgekf-rie}. 

\begin{algorithm}[tpb]
\small
  \caption{Invariant Extended Kalman Filter with invariant error formulation of type 2 \looseness=-1}
  \label{algo:chap02:invekf-rie}
  \begin{algorithmic}
    \State \textbf{Input:} $\Xhat_t,\ \cov_t,\ \mathbf{Z}_{t_\kcurr},\ \mathbf{u}_t$ \\ 
    \textbf{Output:} $\Xhat_{t_\kcurr}^{+}, \cov_{t_\kcurr}^{+}$ \\
    \State \textbf{Propagation:}\\
    $\frac{d}{dt}\Xhat_t =  f_{\mathbf{u}_t}\left(\Xhat_t\right)$ \\
    $\frac{d}{dt}\cov_t = \mathbf{A}_{\mathbf{u}_t}\ \cov_t\ +\ \cov_t\ \mathbf{A}_{\mathbf{u}_t}^T\ +\ \mathbf{\hat{Q}}_t$ \\
    \State \textbf{Update:} \\
    \textbf{if} Right Invariant EKF \textbf{do}:  \\
        $\quad \Xhat_{t_\kcurr}^{+} = \gexphat{G} \left(\mathbf{L}_\kcurr \left(\Xhat_{t_\kcurr}\ \mathbf{z}_{t_\kcurr} - \mathbf{b}\right) \right)\ \Xhat_{t_\kcurr}$  \\
    \textbf{else if} Left Invariant EKF \textbf{do}:  \\
        $\quad \Xhat_{t_\kcurr}^{+} = \Xhat_{t_\kcurr}\ \gexphat{G} \left(\mathbf{L}_\kcurr \left(\Xhat^{-1}_{t_\kcurr}\ \mathbf{z}_{t_\kcurr} - \mathbf{b}\right) \right)$ \\
    \textbf{end} \\
    $ \mathbf{S}_\kcurr = \mathbf{H}\ \cov_{t_\kcurr}\ \mathbf{H}^T\ +\ \mathbf{\hat{N}}_\kcurr$ \\
    $ \mathbf{L}_\kcurr = \cov_{t_\kcurr}\ \mathbf{H}^T\ \left(\mathbf{S}_\kcurr\right)^{-1}$ \\
    $ \cov_{t_\kcurr}^{+} = \left(\I{p} - \mathbf{L}_\kcurr\ \mathbf{H}\right)\ \cov_{t_\kcurr}$\\
    \vspace{-1mm}
  \end{algorithmic}
  \vspace{-2mm}
\end{algorithm}

Having drawn the outline of the InvEKF, we will take a detailed look into the filter design to understand a few properties that allow for better convergence of the estimates.
The invariant EKF design relies on two important properties such as autonomous error dynamics and log-linearity of the error variable.
These properties lead to an EKF design generalizing those of linear systems.
With such a design, a wide range of nonlinear problems can lead to linear error equations through a correct parameterization of the error. \looseness=-1 

\begin{definition}[Autonomous error, (\cite{barrau2017invariant}, Def. 1)]
\label{def:chap02:automonous-error}
The left-invariant and right-invariant errors are said to have a state-trajectory independent propagation if they satisfy a differential equation of the form $\frac{d}{dt} \errG_t = g_{\mathbf{u}_t}\left(\errG_t\right)$.
\end{definition}

\begin{theorem}[Autonomous error dynamics, (\cite{barrau2017invariant}, Theorem 1)]
\label{theorem:chap02:autonomous-error-dynamics}
The three following conditions are equivalent for the dynamics \eqref{eq:chap02:invekf-dynamical-system},
\begin{itemize}
    \item the left-invariant error $\errG^L = \X^{-1}\ \Xhat$ is state-trajectory independent,
    \item the right-invariant error $\errG^R = \Xhat\ \X^{-1}$ is state-trajectory independent, and
    \item $\forall t > 0$ and $\X_1, \X_2 \in G$, 
    \begin{equation}
    \label{eq:chap02:invekf-group-affine-dynamics}
    f_{\mathbf{u}_t}\left(\X_1\ \X_2\right) = f_{\mathbf{u}_t}\left(\X_1\right) \X_2 + \X_1\ f_{\mathbf{u}_t}\left(\X_2\right) - \X_1 f_{\mathbf{u}_t}\left(\I{p}\right) \X_2,
    \end{equation}
\end{itemize}
where, $\I{p}$ denotes of the identity element of the group $G$, and Eq. \eqref{eq:chap02:invekf-group-affine-dynamics} describes the dynamics of a group-affine system. Moreover, if one of these conditions is satisfied then we have,
\begin{align}
        & \frac{d}{dt} \errG_t^L = g^L_{\mathbf{u}_t}\left(\errG_t^L\right), \quad g^L_{\mathbf{u}_t}\left(\errG_t^L\right) = f_{\mathbf{u}_t}\left(\errG^L\right) - f_{\mathbf{u}_t}\left(\I{p}\right) \errG^L, \\
        & \frac{d}{dt} \errG_t^R = g^R_{\mathbf{u}_t}\left(\errG_t^R\right), \quad g^R_{\mathbf{u}_t}\left(\errG_t^R\right) = f_{\mathbf{u}_t}\left(\errG^R\right) - \errG^R\  f_{\mathbf{u}_t}\left(\I{p}\right).
\end{align}
\end{theorem}

\begin{theorem}[Log-linear property of the error, (\cite{barrau2017invariant}, Theorem 2)]
\label{theorem:chap02:log-linear-prop-of-error}
Consider the left or right invariant error $\errG_t$ as defined by \eqref{def:left-invariant-error-type2} or \eqref{def:right-invariant-error-type2} between two arbitrarily far trajectories of form \eqref{eq:chap02:invekf-dynamical-system} satisfying \eqref{eq:chap02:invekf-group-affine-dynamics}. 
Let $\err_0 \in \R^p$ be such that initially $\gexphat{G}\left(\err_0\right) = \errG_0$.
Let $\A_{\mathbf{u}_t}$ be defined by $g_{\mathbf{u}_t}\left(\gexphat{G}\left(\err\right)\right) = \ghat{G}{\A_{\mathbf{u}_t}\ \err} + O\left(\left|\left| \err\right|\right|^2\right)$.
If $\err_t$ is defined for $t > 0$ by the differential equation in $\R^p$
\begin{equation}
    \frac{d}{dt} \err_t = \A_{\mathbf{u}_t}\ \err_t,
\end{equation}
then, we have for the true non-linear error $\errG_t$, the correspondence at all times and for arbitrarily large errors
\begin{equation}
    \forall t \geq 0 \quad \errG_t = \gexphat{G}\left(\err_t\right).
\end{equation}
\end{theorem}

\begin{remark}
The proofs for Theorems \ref{theorem:chap02:autonomous-error-dynamics} and \ref{theorem:chap02:log-linear-prop-of-error} can be found in \cite{barrau2017invariant}.
\end{remark}

Assuming the error to be small ($\errG_t = \gexphat{G}\left(\err_t\right) \approx \I{n}$), the results of Theorem \ref{theorem:chap02:autonomous-error-dynamics} can be extended to a noisy model of the form,
\begin{align}
\label{eq:chap02:invekf-noisy-dynamical-system}
\frac{d}{dt} \X_t = f_{\mathbf{u}_t}\left(\X_t\right) + \X_t\ \ghat{G}{\mathbf{w}_t},
\end{align}
where, $\mathbf{w}_t \sim \mathscr{N}_{\R^p}(\mathbf{0}_{p \times 1},  \Q_t)$ is a continuous-time white noise belonging to $\mathfrak{g}$ with covariance $\Q_t$ defined in $\R^p$.
The error propagation becomes,
\begin{align}
    & \frac{d}{dt} \errG_t^L = g^L_{\mathbf{u}_t}\left(\errG_t^L\right) - \ghat{G}{\mathbf{w}_t}\ \errG^L_t, \\
    & \frac{d}{dt} \errG_t^R = g^R_{\mathbf{u}_t}\left(\errG_t^R\right) - \left(\Xhat_t\ \ghat{G}{\mathbf{w}_t}\ \Xhat_t^{-1}\right) \errG^R_t,
\end{align}
with the linearized error equations,
\begin{align}
    \label{eq:chap02:invekf-linearized-lie-dynamics}
    & \frac{d}{dt} \err^L_t = \A^L_{\mathbf{u}_t}\ \err^L_t -  \mathbf{w}_t, \\
    \label{eq:chap02:invekf-linearized-rie-dynamics}
    & \frac{d}{dt} \err^R_t = \A^R_{\mathbf{u}_t}\ \err^R_t - \gadj{\Xhat_t} \mathbf{w}_t.
\end{align}

For systems whose dynamics satisfy the group-affine property, these results of autonomous error and log-linear error equations play a significant role in the propagation step of the invariant EKF leading to state-independent linearized error systems.

Given the underlying geometrical structure of the invariant EKF, autonomous innovation equations can be formulated with an appropriate choice of measurement models.
This allows the error updated from the set of incoming observations to remain independent of the true state.
The autonomous innovations for the left- and right-invariant error (Eqs. \eqref{def:left-invariant-error-type2} and \eqref{def:right-invariant-error-type2} resp.) become,
\begin{align}
    & \mathbf{\tilde{z}}^L_{t_\kcurr} = \Xhat^{-1}_{t_\kcurr}\ \mathbf{z}_{t_\kcurr} \quad \text{(innovation for left-invariant error)}, \\
    &\mathbf{\tilde{z}}^R_{t_\kcurr} = \Xhat_{t_\kcurr}\ \mathbf{z}_{t_\kcurr} \quad \text{(innovation for right-invariant error)}.
\end{align}

Therefore the update equations can be computed as,
\begin{align}
    & \Xhat_{t_\kcurr}^{+} = \Xhat_{t_\kcurr}\ \gexphat{G} \left(\mathbf{L}_\kcurr \left(\Xhat^{-1}_{t_\kcurr}\ \mathbf{z}_{t_\kcurr} - \mathbf{b}\right) \right) \quad \text{(left-invariant update)}, \\
    &\Xhat_{t_\kcurr}^{+} = \gexphat{G} \left(\mathbf{L}_\kcurr \left(\Xhat_{t_\kcurr}\ \mathbf{z}_{t_\kcurr} - \mathbf{b}\right) \right)\ \Xhat_{t_\kcurr} \quad \text{(right-invariant update)},
\end{align}
where, $\mathbf{L}_\kcurr: \R^n \mapsto \R^p$ is a gain function defined using the linearization of the error system.

The error-update equations can be computed to show that the evolution of the invariant error at the update step is independent of the state trajectory when the measurement noise is not considered,
\begin{align}
    & \errG^{L+}_{t_\kcurr} = \X_{t_\kcurr}^{-1}\ \Xhat_{t_\kcurr}^{+}  = \errG^{L}_{t_\kcurr}\ \gexphat{G} \left(\mathbf{L}_\kcurr \left( \left(\errG^{L}_{t_\kcurr}\right)^{-1}\mathbf{b}\ -\ \mathbf{b} + \Xhat^{-1}_{t_\kcurr}\mathbf{n}_{t_\kcurr} \ \right) \right), \\
    & \errG^{R+}_{t_\kcurr} = \Xhat_{t_\kcurr}^{+}\ \X_{t_\kcurr}^{-1}  = \gexphat{G} \left(\mathbf{L}_\kcurr \left( \errG^{R}_{t_\kcurr}\ \mathbf{b}\ -\ \mathbf{b} + \Xhat_{t_\kcurr}\mathbf{n}_{t_\kcurr} \ \right) \right) \errG^{R}_{t_\kcurr}.
\end{align}

Linearizing the error using the Taylor's series expansion of the exponential map upto the first order leads to the linearized error equation in $\R^p$,
\begin{align}
	\label{eq:chap02:invekf-linearized-update-lie}
	\begin{split}
	& \I{} + \ghat{G}{\err^{L+}_{t_\kcurr}} \approx \left(\I{} +  \ \ghat{G}{\err^{L}_{t_\kcurr}}\right) \left(\I{} + \ghat{G}{\mathbf{L}_\kcurr \left( \left(\I{}  -  \ \ghat{G}{\err^{L}_{t_\kcurr}}\right)\ \mathbf{b}\ -\ \mathbf{b} + \Xhat^{-1}_{t_\kcurr}\mathbf{n}_{t_\kcurr} \ \right)}\right)  \\
	& \I{} + \ghat{G}{\err^{L+}_{t_\kcurr}} \approx \I{} +  \ \ghat{G}{\err^{L}_{t_\kcurr}} + \ghat{G}{\mathbf{L}_\kcurr \left(-\ghat{G}{\err^{L}_{t_\kcurr}}\ \mathbf{b}\ + \Xhat^{-1}_{t_\kcurr}\mathbf{n}_{t_\kcurr}\right) }  \\
	&\ghat{G}{\err^{R+}_{t_\kcurr}} = \ \ghat{G}{\err^{R}_{t_\kcurr}} + \ghat{G}{\mathbf{L}_\kcurr \left(-\ghat{G}{\err^{R}_{t_\kcurr}}\ \mathbf{b}\ + \Xhat^{-1}_{t_\kcurr}\mathbf{n}_{t_\kcurr}\right) }  \\
    & \err^{L+}_{t_\kcurr} =\ \err^{L}_{t_\kcurr} + \mathbf{L}_\kcurr \left(-\ghat{G}{\err^{L}_{t_\kcurr}}\ \mathbf{b} + \Xhat^{-1}_{t_\kcurr}\mathbf{n}_{t_\kcurr}\right)\ =\ \err^{L}_{t_\kcurr} - \mathbf{L}_\kcurr \left(\mathbf{H}\ \err^{L}_{t_\kcurr} - \mathbf{\hat{n}}_\kcurr\right).
	\end{split}
\end{align}
where, we have used $\mathbf{H}\ \err^{L}_{t_\kcurr} =  \ghat{G}{\err^{L}_{t_\kcurr}}\ \mathbf{b},\; \mathbf{\hat{n}}_\kcurr=  \Xhat^{-1}_{t_\kcurr}\mathbf{n}_{t_\kcurr}$ and $\mathbf{H}$ forms a time-invariant measurement model Jacobian.
Similarly, on linearizing the right-invariant error update equations, we have,
\begin{align} 
	\label{eq:chap02:invekf-linearized-update-rie}
	\begin{split}
	& \I{} + \ghat{G}{\err^{R+}_{t_\kcurr}} \approx \left(\I{} +  \ \ghat{G}{\err^{R}_{t_\kcurr}}\right) \left(\I{} + \ghat{G}{\mathbf{L}_\kcurr \left( \left(\I{} +  \ \ghat{G}{\err^{R}_{t_\kcurr}}\right)\ \mathbf{b}\ -\ \mathbf{b} + \Xhat_{t_\kcurr}\mathbf{n}_{t_\kcurr} \ \right)}\right)  \\
	& \I{} + \ghat{G}{\err^{R+}_{t_\kcurr}} \approx \I{} +  \ \ghat{G}{\err^{R}_{t_\kcurr}} + \ghat{G}{\mathbf{L}_\kcurr \left(\ghat{G}{\err^{R}_{t_\kcurr}}\ \mathbf{b}\ + \Xhat_{t_\kcurr}\mathbf{n}_{t_\kcurr}\right) }  \\
	&\ghat{G}{\err^{R+}_{t_\kcurr}} =  \ \ghat{G}{\err^{R}_{t_\kcurr}} + \ghat{G}{\mathbf{L}_\kcurr \left(\ghat{G}{\err^{R}_{t_\kcurr}}\ \mathbf{b}\ + \Xhat_{t_\kcurr}\mathbf{n}_{t_\kcurr}\right) }  \\
	& \err^{R+}_{t_\kcurr} = \ \err^{R}_{t_\kcurr} + \mathbf{L}_\kcurr \left(\ghat{G}{\err^{R}_{t_\kcurr}}\ \mathbf{b} + \Xhat_{t_\kcurr}\mathbf{n}_{t_\kcurr}\right)\ =\ \err^{R}_{t_\kcurr} - \mathbf{L}_\kcurr \left(\mathbf{H}\ \err^{R}_{t_\kcurr} - \mathbf{\hat{n}}_\kcurr\right).
	\end{split}
\end{align}

where, we have used $\mathbf{H}\ \err^{R}_{t_\kcurr} = - \ghat{G}{\err^{R}_{t_\kcurr}}\ \mathbf{b},\; \mathbf{\hat{n}}_\kcurr=  \Xhat_{t_\kcurr}\mathbf{n}_{t_\kcurr}$ and $\mathbf{H}$ forms a time-invariant measurement model Jacobian.
Since, the linearized error propagation and update resemble that of standard EKF equations, the gain $\mathbf{L}_\kcurr$ can be computed using the Ricatti equations, similar to EKF's Kalman gain ($\mathbf{L}_\kcurr \triangleq \mathbf{K}_k$) and the measurement update step is given by Kalman filter update equations. \looseness=-1
\begin{align}
    \begin{split}
        & \frac{d}{dt}\cov_t = \mathbf{A}_{\mathbf{u}_t}\ \cov_t\ +\ \cov_t\ \mathbf{A}_{\mathbf{u}_t}^T\ +\ \mathbf{\hat{Q}}_t, \\
        & \mathbf{S}_\kcurr = \mathbf{H}\ \cov_{t_\kcurr}\ \mathbf{H}^T\ +\ \mathbf{\hat{N}}_\kcurr, \\
        & \mathbf{L}_\kcurr = \cov_{t_\kcurr}\ \mathbf{H}^T\ \left(\mathbf{S}_\kcurr\right)^{-1}, \\
        & \cov_{t_\kcurr}^{+} = \left(\I{p} - \mathbf{L}_\kcurr\ \mathbf{H}\right)\ \cov_{t_\kcurr} \left(\I{p} - \mathbf{L}_\kcurr\ \mathbf{H}\right)^T + \mathbf{L}_\kcurr\ \mathbf{\hat{N}}_\kcurr \mathbf{L}_\kcurr^T.
    \end{split}
\end{align}
where, $\mathbf{\hat{Q}}_t$ and $\mathbf{\hat{N}}_\kcurr$ are the modified process noise and measurement noise covariance matrices depending on the choice of the invariant error. In order to decrease computational complexity, the covariance update can be reduced to,
$$ 
\cov_{t_\kcurr}^{+} = \left(\I{p} - \mathbf{L}_\kcurr\ \mathbf{H}\right)\ \cov_{t_\kcurr}.
$$

\section{Averaging on matrix Lie groups}
\label{sec:chap:loosely-couple-averaging}
Having already seen filtering on Lie groups in the previous section, we now consider the operation of averaging on matrix Lie groups.
Averaging is a fundamental operation that can be used to obtain a descriptive statistic for the distribution of quantities, such as an arithmetic mean for a set of real numbers.
Geometrically, it might describe the center of the mass of a scattered set of points in a given space.
Averaging of points distributed in a Euclidean space is usually straightforward, however, such a computation on a complex space such as a matrix Lie group is usually much more complicated. \looseness=-1

An operator called the Karcher mean was introduced by \cite{karcher1977riemannian} to describe the center of mass of elements on a \emph{Riemannian} manifold that can be visualized as points distributed over a curved surface given that these elements lie close to one another.
\begin{definition}[Karcher Mean \cite{manton2004globally}]
If $d: G \times G \mapsto \R$ is the distance function defined for a Lie group $G$, the Karcher mean of points $\X_1, \dots, \X_k \in G$ is the point $\X$ minimizing,
\begin{equation}
\label{eq:chap:loosely-coupled:karcher-mean}
    f(\X) = \frac{1}{2k} \sum_{i = 1}^k d^2(\X, \X_i).
\end{equation}
\end{definition}
The global minimum of $f(\X)$ is unique as long as the elements $\X_1, \dots, \X_k$ lie close to each other, which is a necessary condition for the Karcher mean to be well-defined.

Another condition for the Karcher mean to reflect the structure of a Lie group (due to its non-commutativity) is that the mean needs to be left- and right-translation invariant.
This means that $\forall\ \Y \in G$, if $\X$ is the Karcher mean of $\X_1, \dots \X_k$, then $L_\Y(\X)$ must be the Karcher mean of $L_\Y(\X_1), \dots L_\Y(\X_k)$, and analogously for the right-translation function.
This means that the distance function must be a bi-invariant metric and such a distance function can be chosen naturally for elements of $\X, \Y$ of matrix Lie group $G$ if they are sufficiently close as,
\begin{equation}
\label{eq:chap:loosely-coupled:distance-lie-group}
    d^2(\X, \Y) = \norm{\glogvee{G}(\X^{-1} \Y)}^2.
\end{equation}

\cite{manton2004globally} provides a globally convergent algorithm (Algorithm \ref{algo:chap:loosely-coupled:manton-algo}) for the computation of the Karcher mean for a Lie group $G$, which can be seen as a Riemannian gradient descent algorithm for minimizing $\X$ in Eq. \eqref{eq:chap:loosely-coupled:karcher-mean}.

\begin{algorithm}[!h]
\small
  \caption{Manton's convergent algorithm for computation of Karcher mean \looseness=-1}
  \label{algo:chap:loosely-coupled:manton-algo}
  \begin{algorithmic}
    \State \textbf{Input:} $\X_1, \dots, \X_k \in G$ \\ 
    \textbf{Output:} $\X \in G$ \\
    \State \textbf{Initialize:}\\
    $\X = \X_1$ \\
    Desired tolerance $\tau > 0$ \\
    \State \textbf{Iterate until convergence:} \\
    \quad $\mathbf{a} = \frac{1}{k} \sum_{i = 1}^k \glogvee{G}(\X^{-1} \X_i)$ \\
    \quad \textbf{if} $\norm{\mathbf{a}} < \tau$ \textbf{break}.  \\
    \quad \textbf{else, do}:  \\
    \quad  \quad $\X = \X \gexphat{G}(\mathbf{a}).$ \\
    \vspace{-1mm}
  \end{algorithmic}
  \vspace{-2mm}
\end{algorithm}

\begin{figure}[!h]
\centering
\includegraphics[scale=0.7]{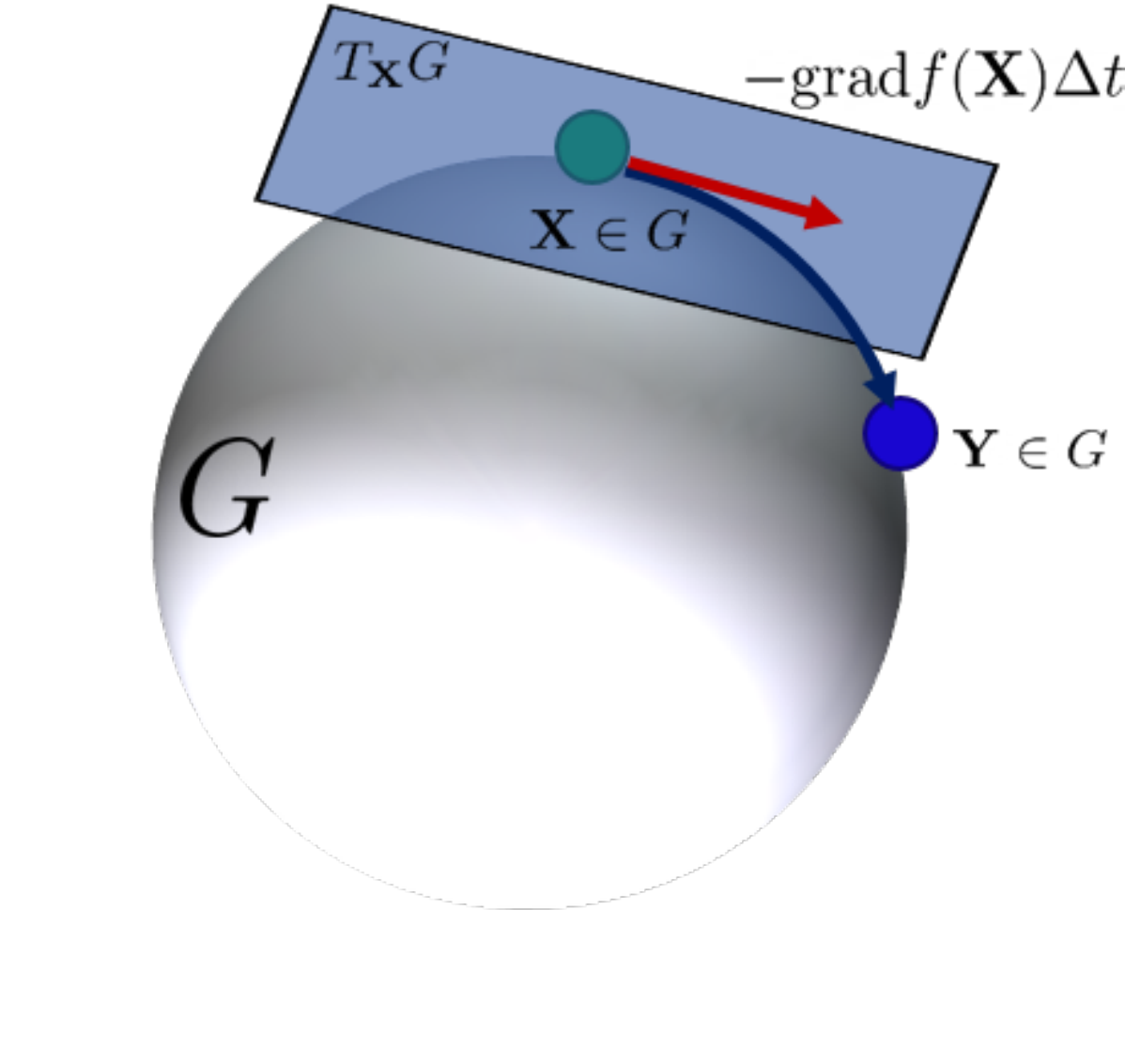}
\vspace*{-3mm}
\caption{Riemannian gradient descent. The gradient shown in red arrow depicts the direction of descent in the tangent space at $\X \in G$ scaled by the learning rate $\Delta t$ and the resulting change is projected onto the manifold to result in the element $\Y \in G$.}
\label{fig:chap:loosely-coupled:riemann-grad}
\end{figure}

The gradient of the Karcher mean with the distance function from Eq. \eqref{eq:chap:loosely-coupled:distance-lie-group} can be obtained as, \looseness=-1
$$
- \frac{1}{k} \X  \sum_{i = 1}^k \glogvee{G}(\X^{-1} \X_i),
$$
which needs to be zero as per the necessary condition of the elements lying sufficiently close to each other.
Geometrically, this would mean that the center of mass of the points $\glogvee{G}(\X^{-1} \X_i)$ for $i = 1, \dots, k$ must lie at the origin of the Lie algebra, around the zero elements.
Figure \ref{fig:chap:loosely-coupled:riemann-grad} depicts an iteration of the gradient descent  and how a step will be taken in the tangent space then projected onto the group manifold using the Lie group operators.
For a much more detailed explanation of the global convergence proofs for the Riemannian gradient descent algorithm to compute the Karcher mean, it is suggested to refer to the original paper by \cite{manton2004globally}.

\subsection{A Probabilistic View}
\label{sec:chap:loosely-coupled=bayesian-fusion}
A direct extension of averaging towards a probabilistic perspective can be obtained by considering the problem of fusion on Lie groups.
The works from \cite{barfoot2014associating, wolfe2011bayesian, chirikjian2011stochastic} account for a rigorous treatment of Bayesian fusion on Lie groups assuming Concentrated Gaussian Distribution (CGD).
It must be noted that, by definition, the \emph{tightly focused} assumption of the CGD (see Sec. \ref{sec:chapter02:uncertainty-matrix-lie-groups}) seems quite similar to the necessary condition used for computing the Karcher mean (elements lying close to each other).
In fact, the fusion approach proposed by \cite{barfoot2014associating} for combining $k$ estimates of poses and their uncertainties is analogous to computing a Karcher mean of poses weighted with the associated uncertainties.
Here, we describe briefly the fusion generalized to any matrix Lie groups.

Given the estimates of a set of elements and their associated covariances, \looseness=-1
$$
\left\{ (\Xhat_1, \cov_1), \dots, (\Xhat_k, \cov_k) \right\},
$$
the problem of fusion is then to determine a single estimate $(\Xhat, \cov)$ that optimally combines all the estimates from the given set.

Considering the optimal estimate $\Xhat^{*}$ and individual estimate $\Xhat_k$, we can initialize the optimal estimate to $\Xhat$ affected by a perturbation $\err$.  
An error $\mathbf{e}_k \sim \mathscr{N}(\Zeros{p \times 1}, \cov_k)$ between $\Xhat^{*}$ and $\Xhat_k$ can be written as, \looseness=-1
\begin{align}
    \begin{split}
        \mathbf{e}_k &= \glogvee{G}\left( (\Xhat^{*})^{-1} \Xhat_k\right)  \\
        &= \glogvee{G}\left( (\Xhat\ \gexphat{G}(\err))^{-1} \Xhat_k\right) \\
        &= \glogvee{G}\left( \gexphat{G}(-\err) \Xhat^{-1} \Xhat_k\right) \\
        &= \glogvee{G}\left( \gexphat{G}(-\err) \ \gexphat{G}(\err_k)\right) \\
        &= \err_k + (\gljac{G}(\err_k))^{-1} \err,
    \end{split}
\end{align}
where, $\err_k = \glogvee{G}\left(\Xhat^{-1} \Xhat_k\right)$ is the error between the current best guess and $k$-th individual estimate and the BCH formula relation with the left Jacobian of Lie group from Eq. \eqref{eq:chap02:liegroups-ljac-bch} has been used in the last step.
It must be noted that the definition of the error $\mathbf{e}_k$ depends on the choice of error used for the distance function. \looseness=-1

The distance function defined in Eq. \eqref{eq:chap:loosely-coupled:distance-lie-group} can be modified into a Mahalanobis distance based cost function (\cite{barfoot2014associating}) with inverse covariance matrices as the weighting matrices,
\begin{align}
    \begin{split}
V &= \frac{1}{2} \sum_{i = 1}^{k} \mathbf{e}_k^T \cov_k^{-1} \mathbf{e}_k  \\
&\approx \frac{1}{2} \sum_{i = 1}^{k} (\err_k + (\gljac{G}(\err_k))^{-1} \err)^T \cov_k^{-1} (\err_k + (\gljac{G}(\err_k))^{-1} \err).
    \end{split}
\end{align}
Thus, determining an optimally fused estimate boils down to minimizing the cost function $V$ with respect to $\err$ in an unconstrained optimization problem which can be alternatively seen as maximizing the joint likelihood of individual estimates.
The system of linear equations for the optimal value of $\err$ is given by, \looseness=-1
$$
\left( \sum_{i = 1}^{k} (\gljac{G}(\err_k))^{-T} \cov_k^{-1} (\gljac{G}(\err_k))^{-1} \right) \err = - \sum_{i = 1}^{k} (\gljac{G}(\err_k))^{-T} \cov_k^{-1} \err_k.
$$
The optimal perturbation $\err$ is then obtained to the current best guess as, $\Xhat = \Xhat \gexphat{G}(\err)$.
The optimally fused covariance matrix can be obtained in the last step of the minimization procedure as,
$$
\cov = \left( \sum_{i = 1}^{k} (\gljac{G}(\err_k))^{-T} \cov_k^{-1} (\gljac{G}(\err_k))^{-1} \right)^{-1}.
$$

While the convergent algorithm proposed by \cite{manton2004globally} is seen as a Riemannian gradient descent since it computes the gradient and takes a step along the tangent space in the direction of the gradient, the fusion described here resembles a Gauss-Newton minimization problem, given the occurrence of the Hessian matrix in the form 
$$\left( \sum_{i = 1}^{k} (\gljac{G}(\err_k))^{-T} \cov_k^{-1} (\gljac{G}(\err_k))^{-1} \right).$$
The latter resembles the procedure described by \cite{barfoot2014associating} for fusing uncertain poses. \looseness=-1

\begin{remark}
\cite{barfoot2014associating} uses a form of error different from what is described in this section and considers the multiplication of perturbation on the left, contrary to what is done here.
However, the underlying procedure is the same as that described by those authors.
\end{remark}

For a more in-depth review of averaging techniques specifically applied to rotations, it is recommended to refer to \cite{hartley2013rotation} which can offer insights into averaging on matrix Lie groups. \looseness=-1


\chapter{Modeling of Floating Base Systems}
\label{chap:modeling}
In this chapter, we review some fundamental concepts of rigid body and rigid multibody systems that form the crux for formulating state estimation methods for tracking relevant physical quantities for a humanoid robot or a human.
Most of the material in this chapter is based on the works of \cite{featherstone2014rigid, siciliano2010robotics, park1995lie, muller2018screw, lynch2017modern, murray2017mathematical, traversaro2017modelling} and \cite{latella2019human}.
Section \ref{sec:chap03:modeling-rigid-body-kin} presents an overview of rigid body kinematics from a mechanical systems' perspective while drawing connections with the Lie groups formalism of rigid body transformations.
This is followed by the description of the kinematics of free-floating rigid multibody systems in Section \ref{sec:chap03:modeling-multi-body-kin}.
Finally, Section \ref{sec:chap03:modeling-human} demonstrates how the framework of the multibody system is used for modeling humans as a highly-articulated, free-floating, rigid multibody system. \looseness=-1

\section{Rigid Body Kinematics}
\label{sec:chap03:modeling-rigid-body-kin}


\subsubsection{Rigid body}
A rigid body is a body that is not subject to any internal deformations. 
In other words, the body is assumed to be non-deformable.
This means that the distance between any two points on the body remains the same at all times even when subjected to any external disturbances, in the form of forces or moments.
Although the rigid body is only an ideal model of a physical object from the real world, it is suitable in the study of most mechanical systems whose dominant kinematics and dynamics can be captured through the rigid body assumption.

\subsubsection{Frames of Reference}
The kinematics or the study of the motion of a rigid body is usually described by attaching a frame $A = (\Pos{}{A}, [A])$ to the rigid body which is a combination of a point on the body called the origin $\Pos{}{A}$ and an orientation frame $[A]$ with an orthonormal basis in the three-dimensional space $\R^3$.
Since the motion of a body can be described only relative to another, a frame of particular importance is the so-called inertial frame \cite[Chapter 3]{lynch2017modern}.
This is a frame of reference that does not undergo any acceleration, and any object in rest or constant rectilinear motion, described in reference with the inertial frame, will remain to be so until it is subjected to an external force.
This concept of an inertial frame or an absolute frame (denoted by $A$ in this thesis) is particularly useful in understanding the spatiotemporal evolution of rigid bodies described with non-inertial frames, i.e. the position or orientation of the moving body in interest described by frame $B$  can be specified with respect to this fixed frame $A$. \looseness=-1

\subsection{Rotations and Homogeneous Transformations}
Given unit vectors $\vec{\mathbf{x}}_A, \vec{\mathbf{y}}_A$ and $\vec{\mathbf{z}}_A$ defining the orientation frame $[A]$, the coordinate vector of a point $\Point{}{}$ with respect to the frame $A$,
\begin{equation}
    \Point{A}{} = 
    \begin{bmatrix} 
        \vec{\mathbf{r}}_{{\Pos{}{A}}, \Point{}{}} . \vec{\mathbf{x}}_A  \\
        \vec{\mathbf{r}}_{{\Pos{}{A}}, \Point{}{}} . \vec{\mathbf{y}}_A \\
        \vec{\mathbf{r}}_{{\Pos{}{A}}, \Point{}{}} . \vec{\mathbf{z}}_A
    \end{bmatrix},
\end{equation}
is defined as the coordinates of the three-dimensional vector $\vec{\mathbf{r}}_{{\Pos{}{A}}, \Point{}{}}$ from origin $\Pos{}{A}$ to the point $\Point{}{}$, expressed in the orientation frame $[A]$.

The coordinates of the point $\Point{}{}$ expressed in the frame $A$ as $\Point{A}{}$ can be transformed to be expressed in a different frame $B$ through a rotation $\Rot{B}{A} \in \SO{3}$ as,
\begin{equation}
    \Point{B}{} = \Rot{B}{A}\ \Point{A}{}.
\end{equation}
The rotation matrix $\Rot{B}{A}$ denotes the coordinate transformation from frame $A$ to frame $B$.
Through the lens of Lie theory, this transformation can be seen as an action of an element of the $\SO{3}$ group on a three dimensional vector (see Def. \ref{def:chap02-group-action}), transforming the original vector to produce a rotated vector.

Given a frame $F = (\Pos{}{F}, [F])$, and the position $\Pos{B}{F}$ of the origin $\Pos{}{F}$ expressed in the frame $B$, then we can define a homogeneous transformation matrix $\Transform{B}{F}$ that also maps the coordinate vector $\Point{F}{}$ to $\Point{B}{}$ through a homogeneous representation of the coordinate vectors, $\PointTilde{B}{} = \Transform{B}{F}\  \PointTilde{F}{}$ or,
\begin{equation}
\label{chap03:rigid-body-transform-act}
\begin{bmatrix}
    \Point{B}{} \\ 1
\end{bmatrix}   =
\begin{bmatrix}
    \Rot{B}{F}  & \Pos{B}{F}\\ \Zeros{1}{3} & 1
\end{bmatrix} 
\begin{bmatrix}
    \Point{F}{} \\ 1
\end{bmatrix},
\end{equation}
which is the same as $\Point{B}{} = \Rot{B}{F}\ \Point{F}{} + \Pos{B}{F}$ in matrix representation and $\PointTilde{B}{} \triangleq\ \text{vec}(\Point{B}{},\ 1) \in \R^4$ is the homogeneous representation of $\Point{B}{}$ ($\PointTilde{F}{}$ for $\Point{F}{}$ accordingly).
The homogeneous transform $\Transform{B}{F}$ belongs to the $\SE{3}$ group of rigid body transformations thereby representing the relative position-and-orientation, referred to as pose, between the frames $B$ and $F$ rigidly attached to same or different rigid bodies in space. 
Eq. \eqref{chap03:rigid-body-transform-act} simply denotes a $\SE{3}$ group action on vectors from $\R^3$. 

\subsection{Rigid Body Velocity}
Given frames $A$ and $B$, the relative velocity between these frames can be obtained by starting from the the time-derivative of the homogeneous transformation matrix $\Transform{A}{B} \in \SE{3}$,
\begin{equation}
\label{eq:chap03:rigid-body-Hdot}
    \Hdot{A}{B} = \frac{d}{dt}(\Transform{A}{B}) = \begin{bmatrix}
        \Rdot{A}{B} & \oDot{A}{B} \\ \Zeros{1}{3} &  0
    \end{bmatrix}.
\end{equation}

On multiplying $\Hdot{A}{B}$ by the inverse of $\Transform{A}{B}$ either on its left- or the right-side leads different representations of 6D velocities belonging to the elements of $\se{3}$. 

\begin{remark}
The different representation of the 6D velocities are referred to as trivialization throughout this thesis as described in \cite[Section 5]{traversaro2019multibody}.
\end{remark}

\subsubsection{Left Trivialized Rigid Body Velocity}
Premultiplying Eq. \eqref{eq:chap03:rigid-body-Hdot} on the left with the inverse of the transform leads to the left-trivialized velocity $\twistLeftTriv{A}{B}$ of $B$ with respect to the frame $A$,
\begin{equation}
\label{chap03:rigid-body-Hdot-left-triv}
    \begin{split}
        \Transform{A}{B}^{-1}\ \Hdot{A}{B} &= \begin{bmatrix}
            \Rot{A}{B}^T & - \Rot{A}{B}^T\ \Pos{A}{B} \\ \Zeros{1}{3} & 1
        \end{bmatrix}\ \begin{bmatrix}
            \Rdot{A}{B} & \oDot{A}{B} \\ \Zeros{1}{3} & 1
        \end{bmatrix} \\
        &= \begin{bmatrix}
            \Rot{A}{B}^T \Rdot{A}{B} & \Rot{A}{B}^T \oDot{A}{B}  \\ \Zeros{1}{3} & 0
        \end{bmatrix} \in \se{3}.
    \end{split}
\end{equation}

The term $\Rot{A}{B}^T \Rdot{A}{B}$ is skew-symmetric and belongs to the Lie algebra of $\SO{3}$.
The skew-symmetric nature of $\Rot{A}{B}^T \Rdot{A}{B}$ can be proved by starting from the equality $\Rot{A}{B}^T \Rot{A}{B} = \I{3}$ given by the definition of the orthonormality of rotation matrices.
On differentiation, we have,
\begin{equation}
    \Rdot{A}{B}^T\ \Rot{A}{B} + \Rot{A}{B}^T\ \Rdot{A}{B} = \Zero{3},
\end{equation}
leading to,
\begin{equation}
    \Rdot{A}{B}^T\ \Rot{A}{B} = - (\Rdot{A}{B}^T\ \Rot{A}{B})^T,
\end{equation}
through the transpose identity of square matrices ($\forall\ \X, \Y \in \R^{n \times n}, (\X \Y)^T = \Y^T \X^T$).
The skew-symmetric matrices are those matrices $\A \in \R^{n \times n}$ for which, by definition $\A = -\A^T$.

The vectors $\vLeftTriv{A}{B} \in \R^3$ and $\omegaLeftTriv{A}{B} \in \R^3$ can be defined such that,
\begin{align}
    & \vLeftTriv{A}{B} \triangleq \Rot{A}{B}^T \oDot{A}{B}, \\
    & \ghat{\SO{3}}{\omegaLeftTriv{A}{B}} \triangleq \Rot{A}{B}^T \Rdot{A}{B}.
\end{align}
It can be seen that the $\ghat{\SO{3}}{\omegaLeftTriv{A}{B}}: \R^3 \mapsto \so{3} \subset \R^{3 \times 3}$ is the \emph{hat} operator of $\SO{3}$ that moves the 3D vector $\omegaLeftTriv{A}{B}$ to the space of skew-symmetric matrices or the Lie algebra $\so{3}$.
Thus, the left trivialized 6D rigid body velocity of frame $B$ with respect to frame $A$ is given as, \looseness=-1
\begin{equation}
\twistLeftTriv{A}{B} \triangleq \begin{bmatrix}
    \vLeftTriv{A}{B} \\ \omegaLeftTriv{A}{B}
\end{bmatrix}    \in \R^6.
\end{equation}
leading to the definition,
\begin{equation}
\label{chap03:rigid-body-left-triv-construction}
    \ghat{\SE{3}}{\twistLeftTriv{A}{B}} = \Transform{A}{B}^{-1}\ \Hdot{A}{B},
\end{equation}
where,  $\ghat{\SE{3}}{\twistLeftTriv{A}{B}}: \R^6 \mapsto \se{3} \subset \R^{4 \times 4}$ maps the 6D vectors to the Lie algebra of $\SE{3}$.

\begin{remark}
It must be noted that $\ghat{\SO{3}}{.}$ and $S(.)$ are used interchangeably in this thesis.\looseness=-1
\end{remark}

\subsubsection{Right Trivialized Rigid Body Velocity}
Similar to Eq. \eqref{chap03:rigid-body-Hdot-left-triv}, the right-trivialized velocity $\twistRightTriv{A}{B}$ of $B$ with respect to $A$ can be obtained by post-multiplying Eq. \eqref{eq:chap03:rigid-body-Hdot} on the right by the inverse of homogeneous transform,
\begin{equation}
\label{chap03:rigid-body-Hdot-right-triv}
    \begin{split}
         \Hdot{A}{B}\  \Transform{A}{B}^{-1} &=  \begin{bmatrix}
            \Rdot{A}{B} & \oDot{A}{B} \\ \Zeros{1}{3} & 1
        \end{bmatrix}\ \begin{bmatrix}
            \Rot{A}{B}^T & - \Rot{A}{B}^T\ \Pos{A}{B} \\ \Zeros{1}{3} & 1
        \end{bmatrix} \\
        &= \begin{bmatrix}
            \Rdot{A}{B}\ \Rot{A}{B}^T &  \oDot{A}{B} - \Rdot{A}{B}\ \Rot{A}{B}^T \Pos{A}{B}  \\ \Zeros{1}{3} & 0
        \end{bmatrix} \in \se{3}.
    \end{split}
\end{equation}
It must be noted that the term $\Rdot{A}{B}\ \Rot{A}{B}^T$ is also a skew-symmetric matrix. The vectors $\vRightTriv{A}{B} \in \R^3$ and $\omegaRightTriv{A}{B} \in \R^3$ can be defined such that,
\begin{align}
    & \vRightTriv{A}{B} \triangleq \oDot{A}{B} - \Rdot{A}{B}\ \Rot{A}{B}^T \Pos{A}{B}, \\
    & \ghat{\SO{3}}{\omegaRightTriv{A}{B}} \triangleq \Rdot{A}{B}\ \Rot{A}{B}^T,
\end{align}
leading to the definition of the right trivialized velocity as,
\begin{equation}
\twistRightTriv{A}{B} \triangleq \begin{bmatrix}
    \vRightTriv{A}{B} \\  \omegaRightTriv{A}{B}
\end{bmatrix}    \in \R^6.
\end{equation}
and consequently by construction, 
\begin{equation}
\label{chap03:rigid-body-right-triv-construction}
    \ghat{\SE{3}}{\twistRightTriv{A}{B}} =  \Hdot{A}{B}\ \Transform{A}{B}^{-1}.
\end{equation}
The linear part of right trivialized velocity can also be written as,
$$
\vRightTriv{A}{B} = \oDot{A}{B} + S(\Pos{A}{B}) \omegaRightTriv{A}{B},
$$
by exploiting the skew-symmetric nature of $\ghat{\SO{3}}{\omegaRightTriv{A}{B}}$.

In connection with the Lie theory, the time derivative $\Hdot{A}{B} \subset \R^{4 \times 4}$ of a rigid body trajectory $\Transform{A}{B}(t): \R \mapsto \SE{3}$ lies in the tangent space $T_{\Transform{A}{B}}\SE{3}$, while the right- and left- trivialized 6D velocities (or twists) $\ghat{\SE{3}}{\twistRightTriv{A}{B}}$ and $\ghat{\SE{3}}{\twistLeftTriv{A}{B}}$ respectively both belong to the Lie algebra of $\SE{3}$ denoted by $\se{3}$.

\subsubsection{Left and Right Trivialized Angular Velocities}
Similar analogy applies also to a trajectory $\Rot{A}{B}(t) \in \SO{3}$, for which the time derivative $\Rdot{A}{B}$ lies in the tangent space $T_{\Rot{A}{B}}\SO{3}$ of the rotation group, while the right- and left- velocities $\ghat{\SO{3}}{\omegaRightTriv{A}{B}} \in \R^{3 \times 3}$ and $\ghat{\SO{3}}{\omegaLeftTriv{A}{B}} \in \R^{3 \times 3}$ belong in the Lie algebra $\so{3}$ of the $\SO{3}$ group.
\begin{align}
    \ghat{\SO{3}}{\omegaLeftTriv{A}{B}} =  \Rot{A}{B}^{T}\ \Rdot{A}{B},  \\
    \ghat{\SO{3}}{\omegaRightTriv{A}{B}} =  \Rdot{A}{B}\ \Rot{A}{B}^{T}.
\end{align}
Therefore, $\omegaLeftTriv{A}{B}$ and $\omegaRightTriv{A}{B}$ then become the left- and right- trivialized velocities of $\SO{3}$ also called as angular velocities of the frame $B$ with respect to $A$, respectively.

\begin{remark}
In cases where the frames $A$ and $B$ correspond to the inertial frame and a body-fixed frame $B$, the left-trivialized velocity of $B$ with respect to $A$ is also called as body-fixed velocity or velocity in the local frame $B$, while the right-trivialized velocity of $B$ with respect to $A$ can also be called as the spatial velocity or the velocity in the inertial frame.
\end{remark}

\subsubsection{Adjoint matrix for frame transformation of velocities}
\label{sec:chap03:rigid-body-adjoint-matrix}
The left- and right-trivialization arise naturally in these Lie groups due to the underlying non-commutative nature of the matrix multiplication used as the composition operator.
The adjoint representation of the group (see Eq. \eqref{eq:chap02-liegroups-adjoint-repr}) is then used to capture the commutativity in these groups.
From Eqs \eqref{chap03:rigid-body-left-triv-construction} and \eqref{chap03:rigid-body-right-triv-construction}, we have,
\begin{equation}
\label{eq:chap03:rigid-body-adjoint-repr}
    \ghat{\SE{3}}{\twistRightTriv{A}{B}} = \Transform{A}{B}\ \ghat{\SE{3}}{\twistLeftTriv{A}{B}}\ \Transform{A}{B}^{-1}.
\end{equation}
which is in the form of Def. \ref{def:chap02:adjointActionOfGroupOnAlgebra}. 
The relationship between the right-trivialized rigid body velocity and the left-trivialized rigid body velocity can then be captured through a linear mapping,
\begin{equation}
\label{eq:chap03:rigid-body-adjoint-map}
    \twistRightTriv{A}{B} = \Xtwist{A}{B}\ \twistLeftTriv{A}{B},
\end{equation}
where, the adjoint matrix of $\SE{3}$, denoted as $\Xtwist{A}{B}$, allows the coordinate transformation of velocities expressed in a frame $B$ to another frame $A$,
\begin{equation}
    \Xtwist{A}{B} = \Xtwist{A}{B}(\Transform{A}{B}) = \begin{bmatrix}
        \Rot{A}{B} & S(\Pos{A}{B})\ \Rot{A}{B} \\ \Zero{3} & \Rot{A}{B}
    \end{bmatrix}.
\end{equation}
$\Xtwist{A}{B}$ is usually called only as "adjoint matrix" by dropping the context of $\SE{3}$, in reference with the spatial algebra notation described in \cite{featherstone2014rigid}. 
However, it is important to emphasize that the term adjoint matrix is applicable for all Lie groups and not only $\SE{3}$, and each adjoint matrix acts on the elements from one tangent space and transforms them to a different tangent space. \looseness=-1

\begin{remark}
The 6D velocity vectors are constructed by serializing the linear part first followed by the angular part throughout this thesis. It must be noted that many works in literature (for example, \cite{featherstone2014rigid}) follow the opposite serialization of angular part first followed by linear part. The choice of the serialization affects how the adjoint matrices used for the coordinate transformation of these velocity vectors are constructed.\looseness=-1
\end{remark}

Analogously, the relationship between the right- and left- trivialized angular velocity can be obtained through the adjoint matrix of $\SO{3}$, which is the rotation itself.
\begin{equation}
\label{eq:chap03:rigid-body-adjoint-rotation}
\omegaRightTriv{A}{B} = \Rot{A}{B}\ \omegaLeftTriv{A}{B}.    
\end{equation}

This can be proven as,
\begin{align}
\begin{split}
    \ghat{\SO{3}}{\omegaRightTriv{A}{B}} &=  \Rdot{A}{B}\ \Rot{A}{B}^{T} \\
    &= \Rot{A}{B}\ \Rot{A}{B}^{T}\ \Rdot{A}{B}\ \Rot{A}{B}^{T} \\
    &= \Rot{A}{B}\  \ghat{\SO{3}}{\omegaLeftTriv{A}{B}}\ \Rot{A}{B}^{T} \\
    &= \ghat{\SO{3}}{\Rot{A}{B} \omegaLeftTriv{A}{B}}.
\end{split}    
\end{align}
Applying the vee operators, $\gvee{\SO{3}}{.}$ on both sides gives us the equation in Eq \eqref{eq:chap03:rigid-body-adjoint-rotation}.

\subsubsection{Mixed Trivialized Rigid Body Velocity}
\begin{figure}[!t]
\centering
\includegraphics[scale=0.5]{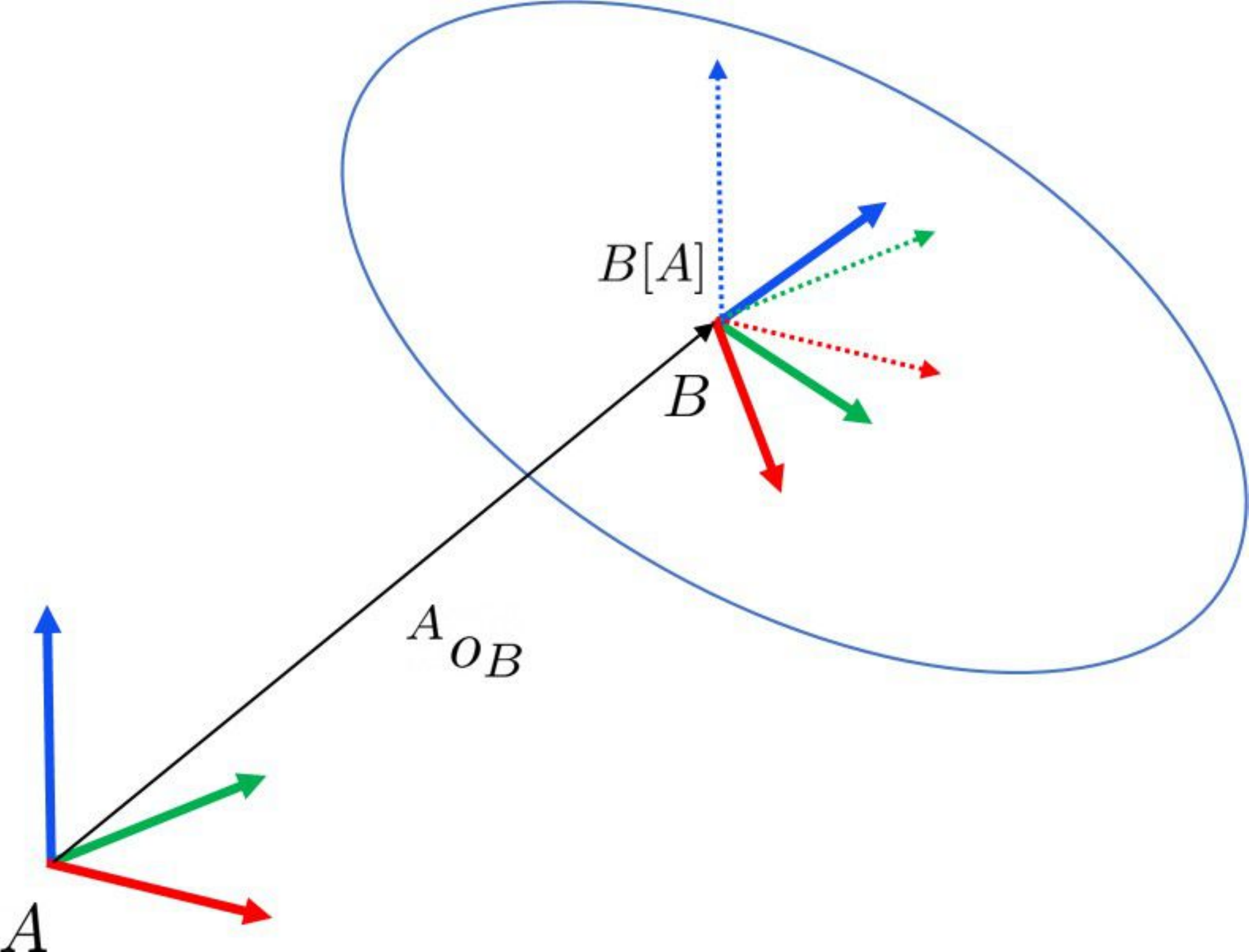}
\caption{Demonstration of a mixed frame $B[A]$ shown in dotted lines with origin at the body frame $B$ and the orientation of the inertial frame $A$. XYZ axes of the coordinate frames follow the RGB color convention. \looseness=-1}
\label{fig:chap:modeling:rigid-body}
\end{figure}

A particular representation of the rigid body velocity known as the \emph{mixed-trivialized rigid body velocity} (\cite{traversaro2019multibody}) becomes useful in control algorithms when representing the state associated with a system described by \emph{Newton-Euler} equations, expressed in a linear-angular momentum serialization (\cite{siciliano2010robotics}). 
Further, in many applications, it is most intuitive and computationally efficient to express the rigid body velocity of $B$ with respect to $A$ in the form,
\begin{equation}
    \begin{bmatrix}
        \oDot{A}{B} \\
        \omegaRightTriv{A}{B}
    \end{bmatrix},
\end{equation}
where, the linear velocity corresponds directly to the time derivative of the position of $B$ with respect to $A$.
This velocity can be associated with the left- or right-trivialized velocity in a straight-forward manner by the introduction of a so-called mixed frame $B[A] = (\Pos{}{B}, [A])$ as shown in Figure \ref{fig:chap:modeling:rigid-body}, where the origin of the frame is coincident with frame $B$, while it has the same orientation as the frame $A$.
This leads to a relative pose between frames $B[A]$ and $B$,
\begin{equation}
    \Transform{B[A]}{B} = \begin{bmatrix}
        \Rot{A}{B} & \Zeros{3}{1} \\ \Zeros{1}{3} & 1
    \end{bmatrix}.
\end{equation}

Thus, the rigid body velocity of $B$ with respect to $A$ is expressed in $B[A]$ as,
\begin{equation}
\begin{split}
    \twistMixedTriv{A}{B} &= \Xtwist{B[A]}{B} \twistLeftTriv{A}{B} \\
    &= \begin{bmatrix}
        \Rot{A}{B} & \Zero{3} \\ \Zero{3} & \Rot{A}{B} 
    \end{bmatrix}\ \begin{bmatrix}
        \Rot{A}{B}^T \oDot{A}{B} \\ \omegaLeftTriv{A}{B}
    \end{bmatrix} = \begin{bmatrix}
        \oDot{A}{B} \\
        \omegaRightTriv{A}{B}
    \end{bmatrix},
\end{split}
\end{equation}
and is called the \emph{mixed} velocity of frame $B$ with respect to $A$ \cite[Section 2.3]{traversaro2017modelling}.

\subsection{Rigid Body Acceleration}

The left-, right- and mixed-trivialized rigid body acceleration can be obtained by a straight-forward time derivative of the associated rigid body velocities, respectively.
Of particular interest within the scope of this thesis is the relationship between the linear part of the mixed-trivialized rigid body acceleration and the measurement of a three-axis linear accelerometer from an Inertial Measurement Unit.
To understand this relationship, we may first define what is called as the \emph{sensor} acceleration \cite[Section 2.4]{traversaro2017modelling} as, \looseness=-1
\begin{equation}
    \sensorAcc{A}{B} = \Xtwist{B}{B[A]}\ \twistDotMixedTriv{A}{B} = \begin{bmatrix}
        \Rot{A}{B}^T \oDoubleDot{A}{B} \\ \omegaDotLeftTriv{A}{B}
    \end{bmatrix}
\end{equation}
Subsequently, the \emph{proper} acceleration is defined using the sensor acceleration as,
\begin{equation}
    \properAcc{A}{B} = \sensorAcc{A}{B} - \begin{bmatrix}
        \Rot{A}{B}^T \gravity{A} \\ \Zeros{3}{1}
    \end{bmatrix}
\end{equation}
The linear part of the proper acceleration can be directly obtained as an accelerometer measurement $\yAcc{A}{B}$ from an inertial measurement unit whose coordinate frame is aligned with the frame $B$, while the angular part of the acceleration can be obtained by a time-derivative of the gyroscope measurement $\yGyro{A}{B}$. 
These measurements are discussed in much detail in the Section \ref{sec:chap:diligent-kio-sys-dyn} of this thesis where an IMU based model is used to predict the state evolution of a floating base system.

\section{Multi Body Kinematics}
\label{sec:chap03:modeling-multi-body-kin}
In this section, we will review the concepts of a free-floating, multi-body mechanical system which is a commonly used mathematical model for describing humanoid robots.

\subsection{Modeling of Multi Body Systems}
\begin{figure}[!t]
\centering
\includegraphics[scale=0.5]{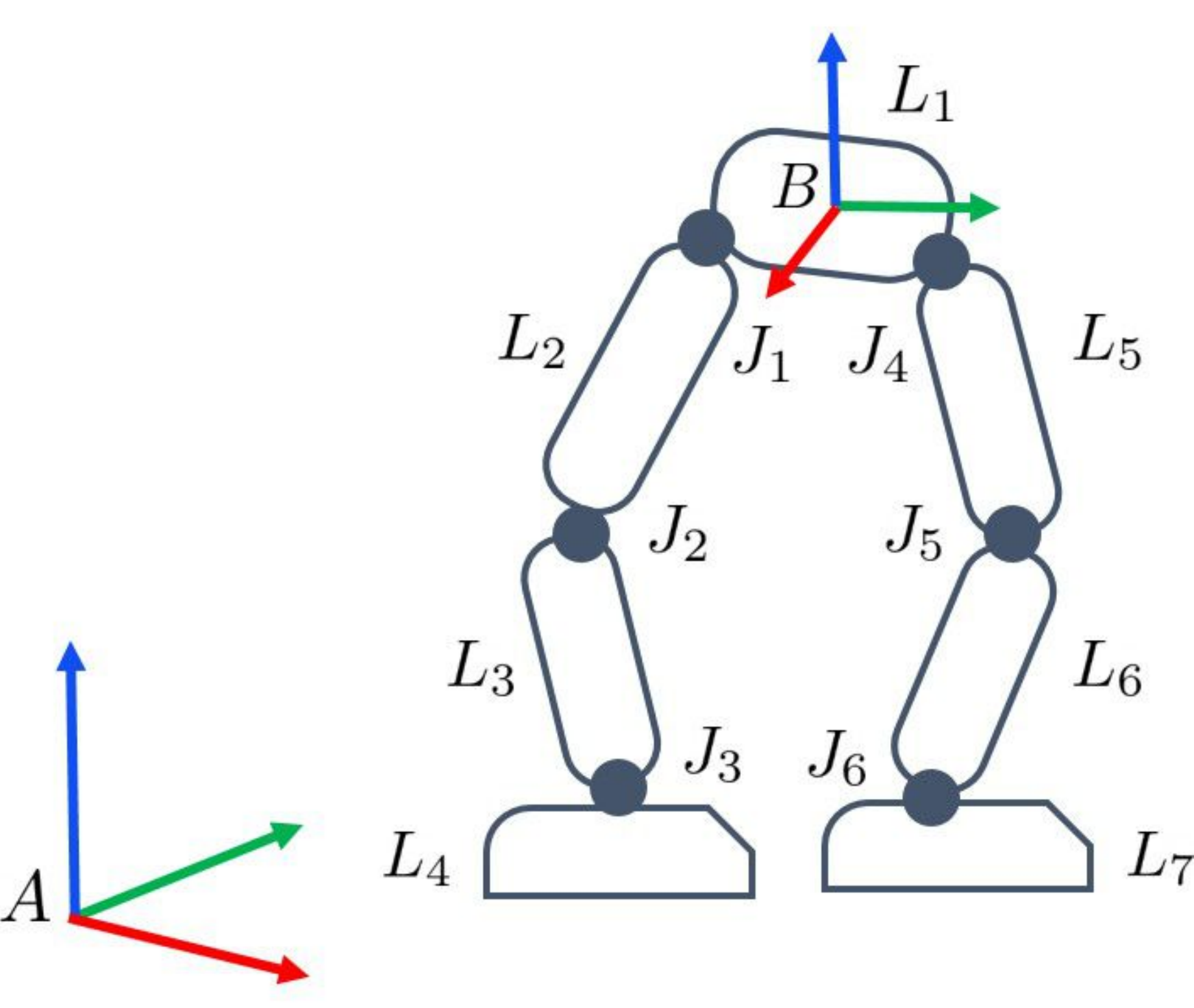}
\caption{A multibody system consisting of links $L_1, \dots, L_7$ (shown as outlined ovals) connected by one DOF joints $J_1, \dots, J_6$ (shown as solid circles). The link $L_1$ is the root link which is associated with the floating base frame $B$ connected virtually to the inertial frame $A$ with a 6 DOF joint.}
\label{fig:chap:modeling:multi-body}
\end{figure}

Consider a multi body system which is composed of $n_L$ rigid bodies called \emph{links} interconnected by $n_J = n_L - 1$ mechanisms called \emph{joints}.
An example of a multibody system is shown in Figure \ref{fig:chap:modeling:multi-body}.
It can be represented by a couple $(\mathfrak{L}, \mathfrak{J})$, where $\mathfrak{L}$ is the set of all the links $\{L_1, L_2, \dots, L_{n_L}\}$ and $\mathfrak{J}$ is the set of all joints $\{J_1, J_2 \dots, J_{n_J}\}$.
The joint constrains the relative motion between pairs of links and the number of degrees of freedom (DOFs) of a joint is the number of unconstrained relative degrees of freedom between two links.
The joints are thus modeled as sets containing two distinct links $\{L_i, L_{i+1}\} \in \mathfrak{J}$.
\begin{definition}[Path connecting link $D$ with link $B$]
\label{def:chap03:multi-body-path}
A path $\pi_B(D)$ between links $B$ and $D$ of length $d$ is defined as the unique ordered sequence of links, $(L_1, L_2, \dots, L_d)$ such that $L_1 = B,\ L_2 = D$ and  $\forall i \in 1,2, \dots, d\ \{L_i, L_{i+1}\} \in \mathfrak{J}$.
\end{definition}
The readers are recommended to refer to \cite{traversaro2017modelling, featherstone2014rigid} for a detailed description of multi body systems and its relationship with graph theory.

\begin{assumption}[Multi Body with only revolute joints]
\label{assumption:chap03:mbs-revolute}
In this thesis we consider only revolute joints which is a $1$-dof joint whose configuration lies in $\R$.
\end{assumption}

The consideration of Assumption \ref{assumption:chap03:mbs-revolute} is a commonly used approach in the design and model of humanoid robots due to it's simplicity, nevertheless, enabling a wide range of anthropomorphic motions on the robot. The revolute joint with an axis $\mathbf{a} \in \R^3,\ \norm{a} = 1$, induces a joint transform between two links $B$ and $D$ through the joint position $\theta \in \R$,
\begin{equation}
\label{eq:chap03:multi-body-joint-transform}
    \Transform{B}{D}(\theta) = \begin{bmatrix}\Rot{B}{D}(\theta) & \Zeros{3}{1} \\ \Zeros{1}{3} & 1 \end{bmatrix},
\end{equation}
with $\Transform{B}{D}(0) = \I{4}$.
The rotation is given by the Rodrigues' formula using axis-angle representation, \looseness=-1
$$
\Rot{B}{D}(\theta) = \I{3} + \cos(\theta) S(\mathbf{a}) + \sin(\theta) S(\mathbf{a})^2.
$$
The relative velocity of $D$ with respect to $B$ is given by its relationship with joint velocity $\dot{\theta}$,
\begin{equation}
\label{eq:chap03:multi-body-one-dof-jvel}
\frac{d}{dt}\Transform{B}{D}(\theta) = \left(\frac{d}{d\theta}  \Transform{B}{D}(\theta)\right) \dot{\theta}.
\end{equation}
The left-, right- and mixed trivialized relative velocity is then given by,
\begin{align}
\label{eq:chap03:multi-body-left-triv-joint-subspace}
    \twistLeftTriv{B}{D} = \jointSubspaceLeftTriv{B}{D}(\theta) \dot{\theta}, \quad \jointSubspaceLeftTriv{B}{D}(\theta) = \gvee{\SE{3}}{\Transform{B}{D}^{-1}(\theta)  \left(\frac{d}{d\theta}  \Transform{B}{D}(\theta)\right)} \\
\label{eq:chap03:multi-body-right-triv-joint-subspace}
    \twistRightTriv{B}{D} = \jointSubspaceRightTriv{B}{D}(\theta) \dot{\theta}, \quad \jointSubspaceRightTriv{B}{D}(\theta) = \gvee{\SE{3}}{\left(\frac{d}{d\theta}  \Transform{B}{D}(\theta)\right)\ \Transform{B}{D}^{-1}(\theta)} \\
\label{eq:chap03:multi-body-mixed-triv-joint-subspace}
    \twistMixedTriv{B}{D} = \jointSubspaceMixedTriv{B}{D}(\theta) \dot{\theta}, \quad \jointSubspaceMixedTriv{B}{D}(\theta) = \begin{bmatrix} 
    \frac{d}{d\theta} \Pos{B}{D}(\theta) \\ \gvee{\SO{3}}{\left(\frac{d}{d\theta}  \Rot{B}{D}(\theta)\right)\ \Rot{B}{D}^{T}(\theta)}
    \end{bmatrix}
\end{align}
The vector $\jointSubspaceLeftTriv{B}{D} \in \R^6$ is called the (left-trivialized) joint motion subspace vector.
This vector is useful in constructing the relative Jacobian that maps the velocities of all the joints of the system to a relative velocity between two links in a path.

The internal dofs of the multi body system, is then the sum of all the dofs of all its joints, $n = \sum_{J \in \mathfrak{J}} \text{DOFs}(J)$.
The shape of the system is defined as the position of all the joints in the system, $\jointPos \in \R^n$.
It is also called the joint configuration of the system.
The velocity and acceleration of the joints are given as $\jointVel \in R^n$ and $\jointAcc \in \R^n$, respectively.
Usually, a link is chosen to be used as the base link $B$ to induce directionality in the multi-body system representation.
With the choice of the base link, the multi body system can be represented as a directed tree (when no loops are present) with base link as the root of the tree.
Given that the base link is the root, it has no parent link. 
The parent link for other links can be defined as the function $\lambda_B: \mathfrak{L} - B \mapsto \mathfrak{L}$ that maps every link except the base link to it's unique parent.
Similarly, the set of children of a link $L$ can be defined as all the links whose unique parent is $L$, $\mu_B(L) = \{D \in \mathfrak{L} \mid \lambda_B(D) = L\}$.
It must be noted that the origin of the coordinate frame associated with the joint coincides with that of the child link.

\subsection{Relative Forward Kinematics}
The relative forward kinematics is used to compute the relative pose $\Transform{L}{D}(\jointPos)$ between two arbitrary links $L$ and $D$ as a function of the shape $\jointPos$ of the multi body system.
The relative forward kinematics $\text{FK}: \R^n \mapsto \SE{3}$ is then defined as,
\begin{equation}
\label{eq:chap03:multi-body-rel-fk}
    \text{FK}(\jointPos) = \Transform{L}{D}(\jointPos) = \Transform{L}{\lambda_D(L)}\ \Transform{\lambda_D(L)}{\lambda^2_D(L)}\dots\Transform{\lambda_D^{d-1}(L)}{\lambda^{d}_D(L)}.
\end{equation}
This is done so by using the definition of the path from links $L$ to $D$ in Def. \ref{def:chap03:multi-body-path} and inducing the directionality by choosing link $D$ as the root of the chosen path, then starting from the leaf link $L$, post-multiplying by the transform between the $i$-th link and it's parent in an iterative fashion.
The relative transform $\Transform{L}{\lambda_D(L)}(s_j)$ between adjacent links $L$ and it's parent $\lambda_D(L)$ in the path is given by the definition of joint transform of the joint $J = \{L, \lambda_D(L)\}$, in our case by Eq. \eqref{eq:chap03:multi-body-joint-transform}. \looseness=-1
The relationship between the relative velocity of links $L$ and $D$ is related to the joint velocities $\jointVel$ through a relative Jacobian matrix.
Consider the left-trivialized velocity of $D$ with respect to $L$, 
$$\twistLeftTriv{L}{D} = \gvee{\SE{3}}{\Transform{D}{L}^{-1}\ \Hdot{L}{D}}.$$
Using the time derivative of the relative forward kinematics map defined in Eq. \eqref{eq:chap03:multi-body-rel-fk}, we have,
\begin{equation}
\label{eq:chap03:multi-body-=left-triv-relative-jac-deriv}
    \begin{split}
        \Transform{D}{L}^{-1}\ \Hdot{L}{D} &= \sum_{i \in \pi_L^{\text{DOF}}(D)} \Transform{D}{L} \Transform{L}{E} \frac{\partial}{\partial s_i} \Transform{E}{F}(s_i) \Transform{F}{D}\ \dot{s}_i \\
        &= \sum_{i \in \pi_L^{\text{DOF}}(D)} \Transform{D}{F} \Transform{F}{E} \frac{\partial}{\partial s_i} \Transform{E}{F}(s_i) \Transform{F}{D}\ \dot{s}_i \\
        &= \sum_{i \in \pi_L^{\text{DOF}}(D)} \Transform{D}{E} \ghat{\SE{3}}{\jointSubspaceLeftTriv{E}{F}} \Transform{F}{D}\ \dot{s}_i \\
        &= \sum_{i \in \pi_L^{\text{DOF}}(D)} \ghat{\SE{3}}{\Xtwist{D}{F} \jointSubspaceLeftTriv{E}{F}}\ \dot{s_i},
    \end{split}
\end{equation}
where, $E$ and $F$ have been introduced to denote adjacent links along the path and we have used $\ghat{\SE{3}}{\jointSubspaceLeftTriv{E}{F}} = \Transform{F}{E} \frac{\partial}{\partial s_i} \Transform{E}{F}(s_i)$ from Eq. \eqref{eq:chap03:multi-body-left-triv-joint-subspace} and the adjoint definition of rigid body velocity from Eqs. \eqref{eq:chap03:rigid-body-adjoint-repr} and \eqref{eq:chap03:rigid-body-adjoint-map}.
Further, $\pi_L^{\text{DOF}}(D)$ is defined as the set of DOFs that belong to the path $\pi_L(D)$ connecting links $L$ with $D$.
Therefore, the left-trivialized velocity of $D$ with respect to $L$ is written as, \looseness=-1
\begin{equation}
    \twistLeftTriv{L}{D} = \sum_{i \in \pi_L^{\text{DOF}}(D)} \left(\Xtwist{D}{F} \jointSubspaceLeftTriv{E}{F}\right)\ \dot{s_i} =\ \relativeJacobianLeftTriv{L}{D}(\jointPos)\  \jointVel,
\end{equation}
where $\relativeJacobianLeftTriv{L}{D}(\jointPos)$ is the left-trivialized relative Jacobian  that maps the joint velocities to relative rigid body velocities constrained by the joints along the path connecting these bodies. The right- and the mixed-trivialized relative Jacobians can be obtained by following a derivation similar to Eq. \eqref{eq:chap03:multi-body-=left-triv-relative-jac-deriv}.

\subsection{Free Floating Kinematics}
So far, we have seen how the topology of the multibody system defined by its shape or joint configuration dictates the relative pose of each link with respect to any other link in the kinematic tree.
However, it is not necessary that the base link is rigidly attached to its environment, i.e. the base link need not be a fixed link.
For floating-base mechanical systems, the base link is a free-floating link, and the overall configuration of the system with respect to an inertial frame $A$ is determined by the pose $\Transform{A}{B}$ of the base link with respect to the inertial frame along with the joint positions $\jointPos \in \R^n$.
The overall configuration of the \emph{free-floating system position} is considered to be evolving in the space of $\mathbb{Q} = \SE{3} \times \R^{n}$ and is defined as, \looseness=-1
\begin{equation}
\label{eq:chap03:multi-body-system-pos}
    \mathbf{q}^B = (\Transform{A}{B}, \jointPos) \in \mathbb{Q}.
\end{equation}
The system configuration of a free-floating multi body system is dependent on the base link $B$.
The pose of any link $L \in \mathfrak{L}$ with respect to the inertial frame $A$ can then be determined in a straight-forward manner from $\mathbf{q}^B$.
Let $\Transform{A}{L}(\mathbf{q}^B):\mathbb{Q} \mapsto \SE{3}$, then
\begin{equation}
\label{eq:chap03:multi-body-link-pose}
     \Transform{A}{L}(\mathbf{q}^B) = \Transform{A}{B} \Transform{B}{L}(\jointPos) = \begin{bmatrix}\Rot{A}{B} & \Pos{A}{B} \\ \Zeros{1}{3} & 1 \end{bmatrix} \Transform{B}{L}(\jointPos) \in \SE{3}.
\end{equation}

The \emph{free-floating system velocity} $\nu^B$ can be obtained by taking the time derivative of the system position, $\dot{\mathbf{q}}^B \triangleq (\Hdot{A}{B}, \jointVel)$.
The trivialization of the base link velocity naturally induces a trivialization on the system velocity as well.
The left-trivialized system velocity $\nu^{B/B} \in \R^{6+{n}}$ is defined as,
\begin{equation}
\label{eq:chap03:multi-body-system-vel-left-triv}
    \nu^{B/B} = \begin{bmatrix} \twistLeftTriv{A}{B} \\ \jointVel \end{bmatrix} = \begin{bmatrix} \Rot{A}{B}^T\ \oDot{A}{B} \\ \gvee{\SO{3}}{\Rot{A}{B}^T\ \Rdot{A}{B}} \\ \jointVel \end{bmatrix} \in \R^{n+6},
\end{equation}
where, $\twistLeftTriv{A}{B}$ is the left-trivialized base velocity and $\jointVel$ is the joint velocities.
The right trivialized base velocity induces a right-trivialized system velocity $\nu^{B/A} \in \R^{6+{n}}$,
\begin{equation}
\label{eq:chap03:multi-body-system-vel-right-triv}
    \nu^{B/A} = \begin{bmatrix} \twistRightTriv{A}{B} \\ \jointVel \end{bmatrix} = \begin{bmatrix} \oDot{A}{B} - \Rdot{A}{B}\ \Rot{A}{B}^T \Pos{A}{B} \\ \gvee{\SO{3}}{\Rdot{A}{B}\ \Rot{A}{B}^T} \\ \jointVel \end{bmatrix} \in \R^{n+6}.
\end{equation}
The mixed-trivialized system velocity $\nu^{B/B[A]} \in \R^{6+{n}}$ is defined as,
\begin{equation}
\label{eq:chap03:multi-body-system-vel-mixed-triv}
    \nu^{B/B[A]} = \begin{bmatrix} \twistMixedTriv{A}{B} \\ \jointVel \end{bmatrix} = \begin{bmatrix} \oDot{A}{B} \\ \gvee{\SO{3}}{\Rot{A}{B}^T\ \Rdot{A}{B}} \\ \jointVel \end{bmatrix} \in \R^{n+6}.
\end{equation}

The system velocity can be used to determine the velocity of any link $L$ with respect to the inertial frame $A$ using the so-called \emph{free-floating Jacobian} matrix $\freeFloatingJacobian{L}{B}$.
The free-floating Jacobian is dependent on the system position $\mathbf{q}^B$.
The velocity of the link $L$ using the left-trivialized system velocity is is given by,
\begin{equation}
\label{eq:chap03:multi-body-link-vel-left-triv}
    \twistLeftTriv{A}{L}(\mathbf{q}^B, \nu^B) = \begin{bmatrix}\Xtwist{L}{B} & \relativeJacobianLeftTriv{L}{B}(\jointPos) \end{bmatrix} \nu^{B/B} = \leftTrivializedJacobian{L}{B}(\mathbf{q}^B)\ \nu^{B/B}.
\end{equation}

The system position and the system velocity described in the Eqs. \eqref{eq:chap03:multi-body-system-pos}, \eqref{eq:chap03:multi-body-system-vel-left-triv}, \eqref{eq:chap03:multi-body-system-vel-right-triv} and \eqref{eq:chap03:multi-body-system-vel-mixed-triv} define the overall system state while Eqs. \eqref{eq:chap03:multi-body-link-pose} and \eqref{eq:chap03:multi-body-link-vel-left-triv} together define the forward kinematics of a floating base mechanical system.

\subsection{Inverse Kinematics}
\label{sec:modeling:ik}
Inverse Kinematics (IK) (\cite{waldron2016kinematics}) is referred to as the process of estimating the configuration of a mechanical system when provided with task space measurements.
Consider a set of $n_p$ frames $P =\{P_1, P_2, \dots, P_{N_p}\}$ associated with target position measurements $\Pos{A}{{P_i}}(t) \in \R^3$ and target linear velocity measurements $\oDot{A}{P_i}(t) \in \R^3$.
Similarly, consider a set of $n_o$ frames $O = \{O_1, O_2, \dots, O_{N_o}\}$ associated with target orientations $\Rot{A}{{O_j}}(t) \in \SO{3}$ and target angular velocity measurements $\omegaRightTriv{A}{O_j}(t) \in \R^3$.
Given the kinematic description of the model, IK is used to find the state configuration $(\mathbf{q}(t), \nu(t))$, such that
\begin{equation}
\label{chap:human-motion:ik-problem}
    \begin{cases}
        \Pos{A}{{P_i}}(t) = h^p_{P_i}(\mathbf{q}(t)), & \forall\ i = 1, \dots, n_p \\
        \Rot{A}{{O_j}}(t) = h^o_{O_j}(\mathbf{q}(t)), & \forall\ j = 1, \dots, n_o \\
        \oDot{A}{{P_i}}(t) = \mathbf{J}^\text{lin}_{P_i}(\mathbf{q}(t)) \nu(t), & \forall\ i = 1, \dots, n_p \\
        \omegaRightTriv{A}{{O_j}}(t) = \mathbf{J}^\text{ang}_{O_j}(\mathbf{q}(t)) \nu(t), & \forall\ j = 1, \dots, n_o \\
        \A^\mathbf{q}\ \jointPos(t) \leq \mathbf{b}^\mathbf{q}, \\
        \A^\nu\ \jointVel(t) \leq \mathbf{b}^\nu,
    \end{cases}
\end{equation}
where, $\mathbf{q}(t) \triangleq \mathbf{q}^B(t)$ and $\nu(t) \triangleq \nu^{B/B[A]}(t)$ is the position and velocity of the mechanical system, $h^p_{P_i}(\mathbf{q}(t))$ and $h^o_{O_j}(\mathbf{q}(t))$ are the position and orientation selection functions from the forward kinematics of frames $P_i$ and $O_j$ respectively, while $\mathbf{J}^\text{lin}_{P_i}(\mathbf{q}(t))$ is the linear part of the mixed-trivialized Jacobian matrix $\mixedTrivializedJacobian{B}{{P_i}}(\mathbf{q}^B)$ and $\mathbf{J}^\text{ang}_{O_j}(\mathbf{q}(t))$ is the angular part of the Jacobian matrix $\mixedTrivializedJacobian{B}{O_j}(\mathbf{q}^B)$.
$\A^\mathbf{q}$ and $\mathbf{b}^\mathbf{q}$ are constant parameters representing the limits for joint configuration $\jointPos(t)$, while the constant parameters $\A^\nu$ and $\mathbf{b}^\nu$ represent the limits for joint velocities $\jointVel(t)$.
The targets can be collected in a pose target tuple $\mathbf{x}(t) \in (\R^3)^{n_p} \times (\SO{3})^{n_o}$ and a velocity target vector $\mathbf{v}(t) \in (\R^3)^{n_p} \times (\R^3)^{n_o}$ for having a compact representation.
\begin{equation}
\label{eq:ik:tuples}
\mathbf{x}(t) \triangleq 
\begin{pmatrix} 
\Pos{A}{{P_1}}(t) \\ \vdots \\ \Pos{A}{{P_{n_p}}}(t) \\ \Rot{A}{{O_1}}(t) \\ \vdots \\ \Rot{A}{{O_{n_o}}}(t)
\end{pmatrix}, \quad
\mathbf{v}(t) \triangleq 
\begin{bmatrix} 
\oDot{A}{{P_1}}(t) \\ \vdots \\ \oDot{A}{{P_{n_p}}}(t) \\ \omegaRightTriv{A}{{O_1}}(t) \\ \vdots \\ \omegaRightTriv{A}{{O_{n_o}}}(t)
\end{bmatrix}.
\end{equation}
Similarly, the forward kinematics map and the Jacobians necessary for the differential kinematics can be stacked as follows,
\begin{equation}
h(\mathbf{q}(t)) \triangleq 
\begin{pmatrix} 
h^p_{P_1}(\mathbf{q}(t)) \\ \vdots \\ h^p_{P_{n_p}}(\mathbf{q}(t)) \\ h^o_{O_1}(\mathbf{q}(t)) \\ \vdots \\ h^o_{O_{n_o}}(\mathbf{q}(t))
\end{pmatrix}, \quad
\mathbf{J}(\mathbf{q}(t)) \triangleq 
\begin{bmatrix} 
\mathbf{J}^\text{lin}_{P_1}(\mathbf{q}(t)) \\ \vdots \\ \mathbf{J}^\text{lin}_{P_{n_p}}(\mathbf{q}(t)) \\ \mathbf{J}^\text{ang}_{O_1}(\mathbf{q}(t)) \\ \vdots \\ \mathbf{J}^\text{ang}_{O_{n_o}}(\mathbf{q}(t))
\end{bmatrix},
\end{equation}
leading to the set of equations,
\begin{equation}
\label{chap:human-motion:state-configuration}
\begin{split}
    &\mathbf{x}(t) = h(\mathbf{q}(t)) \\
    &\mathbf{v}(t) = \mathbf{J}(\mathbf{q}(t)) \nu(t).
\end{split}
\end{equation}
When the targets are not consistent, a solution to the system of equations described in Eq. \eqref{chap:human-motion:state-configuration} may not exist at all.
However, when such a solution exists, it is usually challenging to find an analytical solutions for the system state configuration $(\mathbf{q}(t), \nu(t))$ that satisfies the Eq. \eqref{chap:human-motion:state-configuration}, especially for highly articulated systems like humans.
Thus, a numerical approach through a non-linear optimization is usually preferred to find an optimal solution for the system state configuration,
\begin{equation}
\label{chap:human-motion:nlopt-prob}
\begin{split} 
    \underset{\mathbf{q}(t_\kcurr), \nu(t_\kcurr)}{\text{minimize}} & \qquad \norm{\mathbf{K}_q\ \mathbf{r}_\mathbf{q}(\mathbf{q}(t_\kcurr), \mathbf{x}(t_\kcurr))} _2  + \norm{\mathbf{K}_\nu\ \mathbf{r}_\nu(\nu(t_\kcurr), \mathbf{v}(t_\kcurr))} _2 \\
    \text{subject to} & \qquad \mathbf{A} \begin{bmatrix} \jointPos(t_\kcurr) \\ \jointVel(t_\kcurr) \end{bmatrix} \leq \mathbf{b},
\end{split}
\end{equation}
where $\mathbf{r}_\mathbf{q}$ and $\mathbf{r}_\nu$ are residual cost functions for target position and velocities with $\mathbf{K}_q$ and $\mathbf{K}_\nu$ being the associated gain matrices for the target quantities.
\section{Human Modeling}
\label{sec:chap03:modeling-human}
In this section, we recall the basic concepts for human modeling based on a more detailed development presented by \cite{latella2019human} and \cite{tirupachuri2020enabling}.
This becomes important in the context of human motion estimation since our algorithms rely heavily on the model representations for the human body.
Human modeling has remained an active topic of research for many years focusing on varying degrees of model representation such as skeletal modeling, muscle modeling, non-deformable models, data-driven models, and so on (\cite{seth2011opensim,loper2015smpl}).
The human model, in this thesis, is developed within the formalism of rigid multibody systems described in the previous sections.
Such a representation does not fully capture the biomechanics of the human body which is a much more complex structure containing muscles, tendons, soft tissues, cartilages, and bones functioning in harmony to generate a wide range of dynamic motions and static exertions.
Nevertheless, it sufficiently captures the skeletal geometries and inertial properties of the several body parts and joint articulations to represent realistic motions on the human in a simplified manner.  \looseness=-1
\begin{remark}
The terms \emph{body}, \emph{segment} or \emph{links} are used interchangeably in this thesis.
\end{remark}
\begin{assumption}[Simple Geometric Shapes of Human Model]
\label{assumption:chap:background:human-links}
The human model is represented as a set of rigid bodies with simple geometrical shapes such as solid parallelopipeds, solid cylinders and solid spheres. \looseness=-1
\end{assumption}
\begin{assumption}[Link Density, \cite{hanavan1964mathematical}]
\label{assumption:chap:background:human-link-density}
The density of each link of the human model is uniform and isotropic.
\end{assumption}

\begin{figure}[!t]
\centering
\includegraphics[scale=0.3]{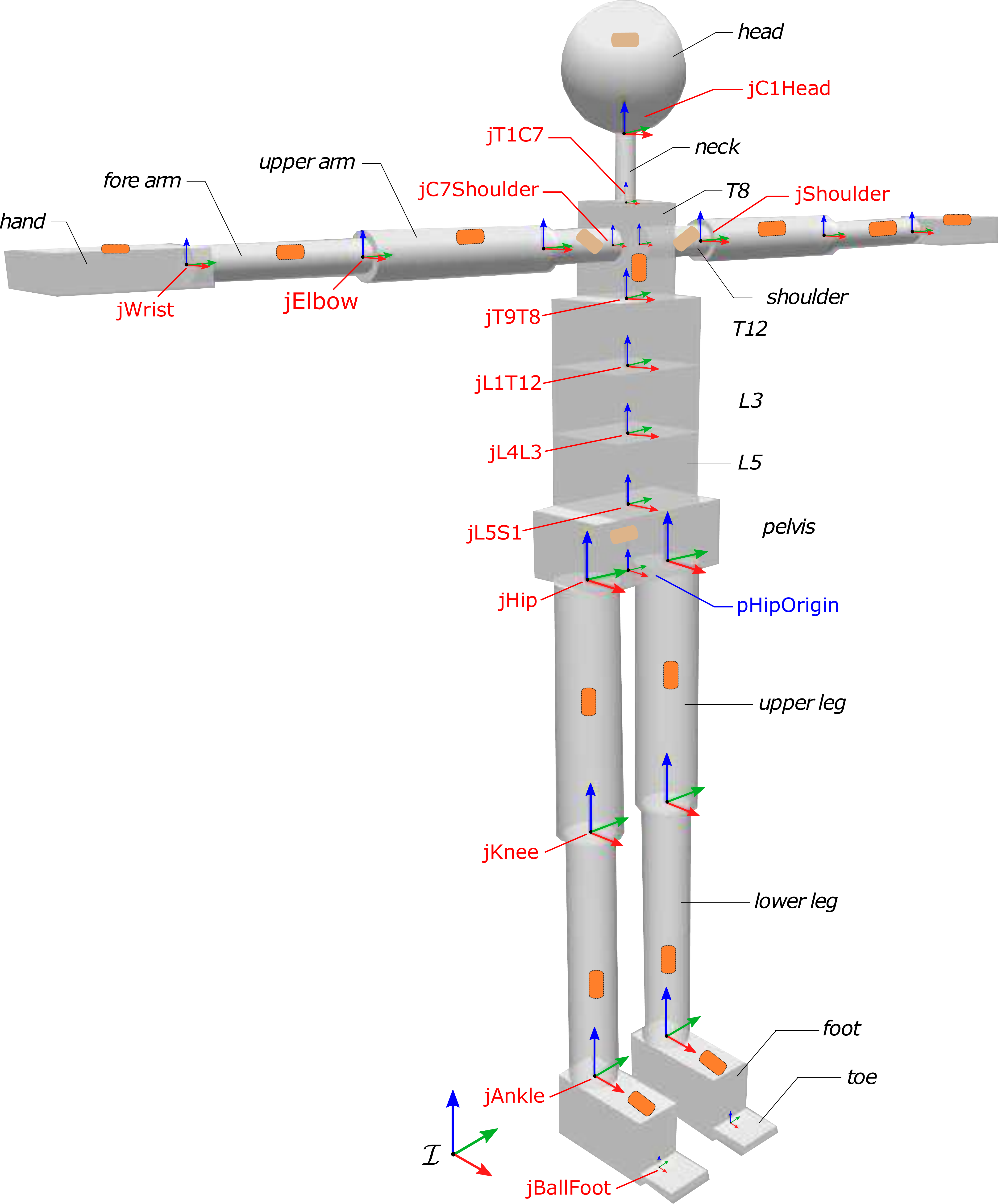}
\caption{URDF representation of human model (\cite{latella2019human}).}
\label{fig:chap:background:human-model-urdf}
\end{figure}

The model developed for the Xsens MVN motion capture system \cite{roetenberg2009xsens} is used as a baseline for the human model developed in this thesis.
We use a description format commonly used in the robotics community called the \emph{Universal Robot Description Format} (URDF) to represent human models.
It is tailored for describing highly-articulated robots with rigid links as kinematic trees.
Given its possibility to handle complex tree structures and the possibility to support the idea of representing both human and robot models using the same representation, we choose the URDF as a unifying representation for describing human models. Figure \ref{fig:chap:background:human-model-urdf} shows a graphical visualization of the human model described using URDF representation.

In the following subsections, we briefly go through the modeling of links and joints of the human body that are the fundamental building blocks for representing a human model. \looseness=-1

\subsection{Modeling of Links}
\begin{figure}[!t]
\begin{subfigure} {0.42 \textwidth}
\centering
\includegraphics[scale=0.5]{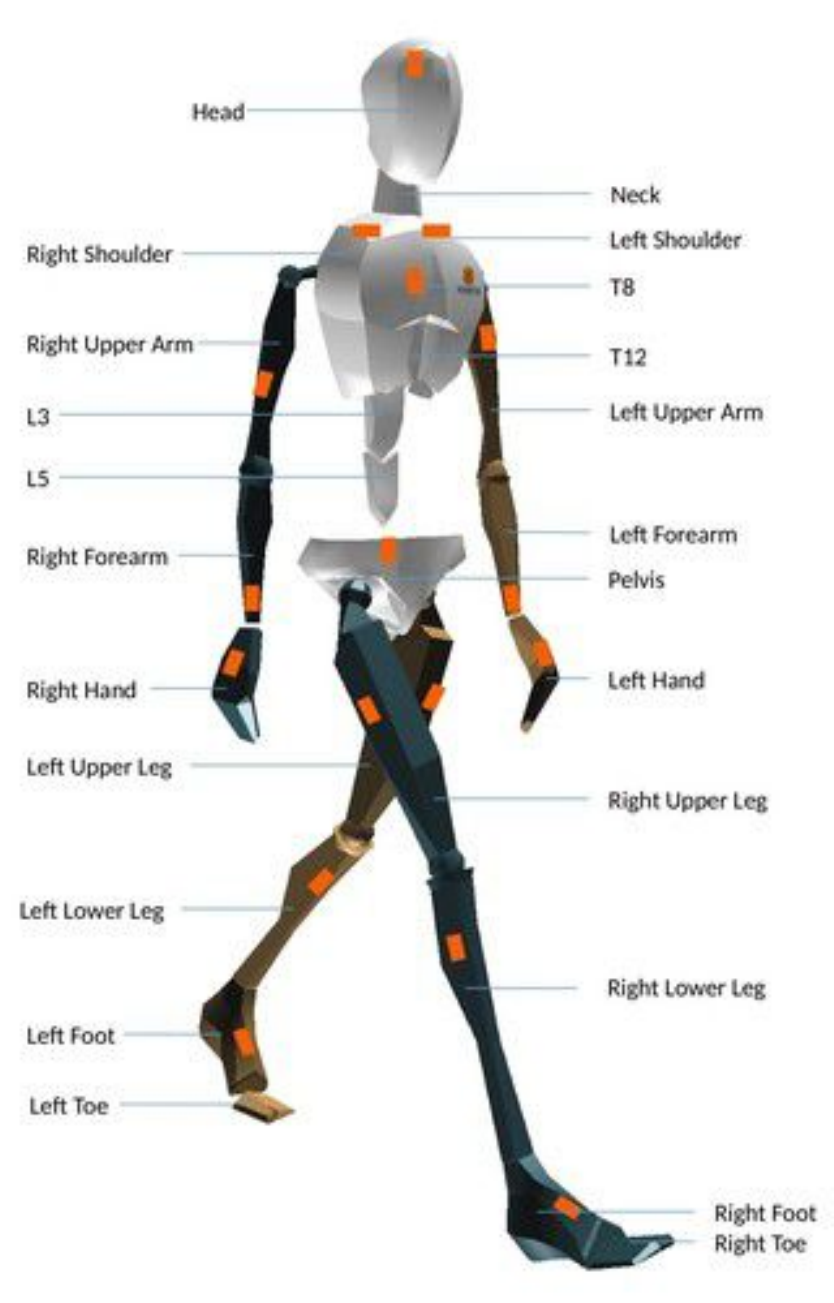}
\end{subfigure}
\begin{subfigure}{0.38 \textwidth}
\centering
\includegraphics[scale=0.5]{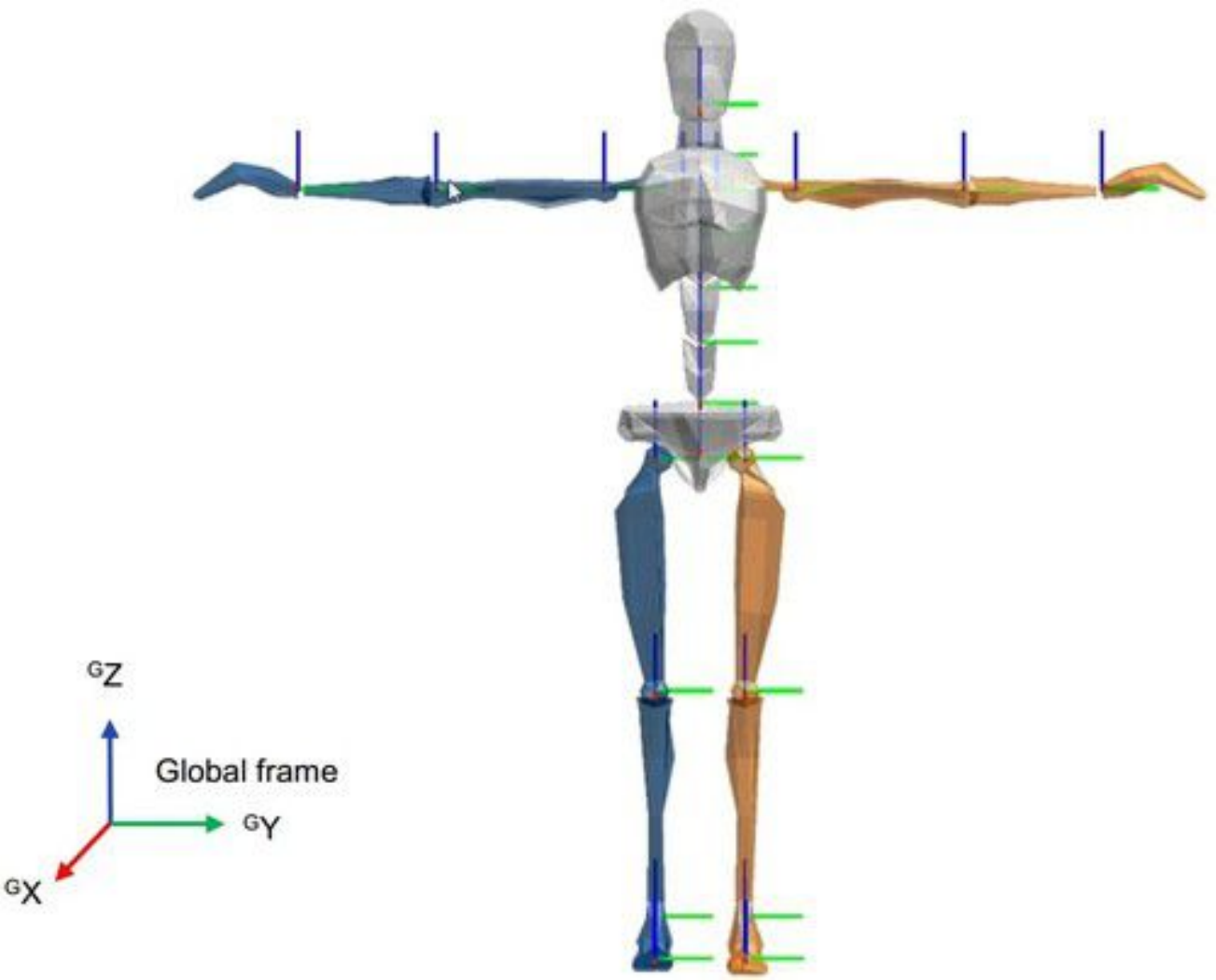}
\end{subfigure}
\caption{Xsens MVN representation of human model (\cite{karatsidis2017estimation}).}
\label{fig:chap:background:human-model-xsens}
\end{figure}

The kinematic tree of the human model is considered to have $n_L = 23$ moving links with labels inherited from the Xsens model as seen in Figure \ref{fig:chap:background:human-model-xsens}.
The choice of origin and orientation of the coordinate frames associated with each of the links is detailed by \cite{latella2019human}.
According to Assumption \ref{assumption:chap:background:human-links}, the link geometries are simplified to be either a parallelopiped, cylinder or a sphere.
The dimensions of the solid shapes are calibrated using measurements from the Xsens motion capture system to remain scalable with subject proportions.
The mass of each link is obtained as a ratio of the total human mass according to a lookup table generated from anthropometric studies (\cite{winter2009biomechanics}).
The list of links in the human model, its associated shape, and the mass ratio is given in Table \ref{table:human-links}.


\begin{table}[!h]
	\caption{Link properties of the human model: link labels, shapes and mass as a percentage of total human subject mass (extracted from \cite{winter2009biomechanics})  (\cite{latella2019human}).  \looseness=-1 }
	\label{table:human-links}
	\centering
	\scalebox{0.8}{
		\begin{tabular}{|c|c|c|}
			\hline & & \\
			\textbf{Label}  & \textbf{Shape}  & \textbf{Link mass to total mass ratio}  \\
			& &  \\
			\hline 
			Pelvis & parallelopiped & $0.08$  \\
			\hline 
			L5 & parallelopiped & $0.102$ \\
			\hline 
			L3 & parallelopiped & $0.102$ \\
			\hline 
			T12 & parallelopiped & $0.102$ \\
			\hline 
			T8 & parallelopiped & $0.04$ \\
			\hline 
			Neck & cylinder & $0.012$ \\
			\hline 
			Head & sphere & $0.036$ \\
			\hline 
			RightShoulder & cylinder & $0.031$ \\
			\hline 
			RightUpperArm & cylinder & $0.030$ \\
			\hline 
			RightForeArm & cylinder & $0.020$ \\
			\hline 
			RightHand & parallelopiped & $0.006$ \\
			\hline 
			LeftShoulder & cylinder & $0.031$ \\
			\hline 
			LeftUpperArm & cylinder & $0.030$ \\
			\hline 
			LeftForeArm & cylinder & $0.020$ \\
			\hline 
			LeftHand & parallelopiped & $0.006$ \\
			\hline 
			RightUpperLeg & cylinder & $0.125$ \\
			\hline 
			RightLowerLeg & cylinder & $0.0365$ \\
			\hline 
			RightFoot & parallelopiped & $0.013$ \\
			\hline 
			RightToe & parallelopiped & $0.015$ \\
			\hline 
			LeftUpperLeg & cylinder & $0.125$ \\
			\hline 
			LeftLowerLeg & cylinder & $0.0365$ \\
			\hline 
			LeftFoot & parallelopiped & $0.013$ \\
			\hline 
			LeftToe & parallelopiped & $0.015$ \\
			\hline 
		\end{tabular}}
\end{table}

\subsubsection{Inertial Properties}
The inertial parameters of mass, the center of mass, and moments of inertia capture the rigid body dynamics faithfully enabling the prediction of realistic motion patterns.
With the assumptions \ref{assumption:chap:background:human-link-density} and \ref{assumption:chap:background:human-link-density}, simple analytical formulas can be used to represent the inertia tensor as a diagonal matrix with principal moments of inertia as the elements.
The moments of inertia can be easily computed with the help of the link masses registered with the help of anthropometric evaluations.
The inertia tensor thus takes the form, \looseness=-1
\begin{equation}
\mathbf{I} = \begin{bmatrix} I_{xx} & 0 & 0 \\
0 & I_{yy} & 0 \\
0 & 0 & I_{zz} 
\end{bmatrix}
\end{equation}
where, $I_{xx}, I_{yy}$ and $I_{zz}$ are the principal moments of inertia and the moments of inertia for each shape listed in Table \ref{table:human-inertias}.

\begin{table}[!h]
	\caption{Moments of inertia for a solid rectangular parallelopiped with width $w$, height $h$ and depth $d$, a solid cylinder of radius $r$ and height $h$ and a sphere of radius $r$, each having mass $m$ (\cite{latella2019human}). \looseness=-1 }
	\label{table:human-inertias}
	\centering
	\scalebox{0.8}{
		\begin{tabular}{|c|c|c|c|}
			\hline & & & \\
			\textbf{Moment of Inertia}  & \textbf{Parallelopiped}  & \textbf{Cylinder} & \textbf{Sphere} \\
			& & &  \\
			\hline 
             & & & \\
			$I_{xx}$ & $\frac{1}{12} m (w^2 + h^2)$ & $\frac{1}{12} m (3r^2 + h^2)$ & $\frac{2}{5}mr^2$ \\
			& & & \\
			\hline 
			& & & \\
			$I_{yy}$ & $\frac{1}{12} m (h^2 + d^2)$ & $\frac{1}{2}mr^2$ & $\frac{2}{5}mr^2$ \\
			& & & \\
			\hline 
			& & & \\
			$I_{zz}$ & $\frac{1}{12} m (d^2 + w^2)$ & $\frac{1}{12} m (3r^2 + h^2)$ & $\frac{2}{5}mr^2$ \\
			& & & \\
			\hline 
		\end{tabular}}
\end{table}

\subsection{Modeling of Joints}
The links are coupled by joints and the modeling of joints requires information about the motion type described by the number of DoFs, links connected by the joints, and the joint coordinate system.
The joint labeling is similar to that of the links are also derived from Xsens nomenclature
The origin of the frame associated with the joint coincides with that of the child link.
Rotational joints with one, two, or three DoFs are considered for modeling different joints and the number of DoFs is chosen based on the relative biological joint.
The relative biological joint represents a point on the human body where relevant bones articulate.
For the sake of generality, three DoFs are sometimes assumed for all joints leading to the total number of DoFs in the human model as  $3 (n_L- 1) = 66$ DoFs but with proper classification using the information about relative biological joints, the reduced model has $48$ DoFs.
The list of joints, DoFs per joint, and the links connected through each joint are detailed in Table \ref{table:human-joints}. \looseness=-1

\begin{table}[!h]
	\caption{Joint properties of the human model: joint labels, DoFs per joint (along with reduced DoFs) and links connected through the given joint (\cite{latella2019human}). \looseness=-1 }
	\label{table:human-joints}
	\centering
	\scalebox{0.8}{
		\begin{tabular}{|c|c|r c l|}
			\hline & & & & \\
			\textbf{Label}  & \textbf{DoFs (Reduced DoFs)}  & \multicolumn{3}{|c|}{\textbf{Connected links}}   \\
			& & & & \\
			\hline 
			jL5S1 & 3(2) & Pelvis & $\longleftrightarrow$ & L5 \\
			\hline 
			jL4L3 & 3(2) & L5 & $\longleftrightarrow$ & L3 \\
			\hline 
			jL1T12 & 3(2) & L3 & $\longleftrightarrow$ & T12 \\
			\hline 
			jT9T8 & 3 & T12 & $\longleftrightarrow$ & T8 \\
			\hline 
			jT1C7 & 3 & T8 & $\longleftrightarrow$ & Neck \\
			\hline 
			jC1Head & 3(2) & Neck & $\longleftrightarrow$ & Head \\
			\hline 
			jRightHip & 3 & Pelvis & $\longleftrightarrow$ & RightUpperLeg \\
			\hline 
			jRightKnee & 3(2) & RightUpperLeg & $\longleftrightarrow$ & RightLowerLeg \\
			\hline 
			jRightAnkle & 3 & RightLowerLeg & $\longleftrightarrow$ & RightFoot \\
			\hline 
			jRightBallFoot & 3(1) & RightFoot & $\longleftrightarrow$ & RightToe \\
			\hline 
			jLeftHip & 3 & Pelvis & $\longleftrightarrow$ & LeftUpperLeg \\
			\hline 
			jLeftKnee & 3(2) & LeftUpperLeg & $\longleftrightarrow$ & LeftLowerLeg \\
			\hline 
			jLeftAnkle & 3 & LeftLowerLeg & $\longleftrightarrow$ & LeftFoot \\
			\hline 
			jLeftBallFoot & 3(1) & LeftFoot & $\longleftrightarrow$ & LeftToe \\
			\hline 
			jRightC7Shoulder & 3(1) & T8 & $\longleftrightarrow$ & RightShoulder \\
			\hline 
			jRightShoulder & 3 & RightShoulder & $\longleftrightarrow$ & RightUpperArm \\
			\hline 
			jRightElbow & 3(2) & RightUpperArm & $\longleftrightarrow$ & RightForeArm \\
			\hline 
			jRightWrist & 3(2) & RightForeArm & $\longleftrightarrow$ & RightHand \\
			\hline 
			jLeftC7Shoulder & 3(1) & T8 & $\longleftrightarrow$ & LeftShoulder \\
			\hline 
			jLeftShoulder & 3 & LeftShoulder & $\longleftrightarrow$ & LeftUpperArm \\
			\hline 
			jLeftElbow & 3(2) & LeftUpperArm & $\longleftrightarrow$ & LeftForeArm \\
			\hline 
			jLeftWrist & 3(2) & LeftForeArm & $\longleftrightarrow$ & LeftHand \\
			\hline 
		\end{tabular}}
\end{table}



The modeling tools described in this chapter will be used in unison with the Lie-theoretic, estimation machinery described in Chapter \ref{chap:estimation} to develop state estimation algorithms for humans and humanoid robots described as \emph{floating-base, rigid, multibody mechanical systems}.

\chapter{State of the Art and Thesis Context}
\label{chap:context}

The technological objective of this thesis is the development of state estimation algorithms for floating base estimation of humanoid robots and whole-body human motion reconstruction.
A reliable estimation of the humanoid robot's floating base state and the motion of the human is useful in the development of a shared human-humanoid collaborative control framework. 

\begin{figure}[!h]
\centering
\includegraphics[scale=0.35]{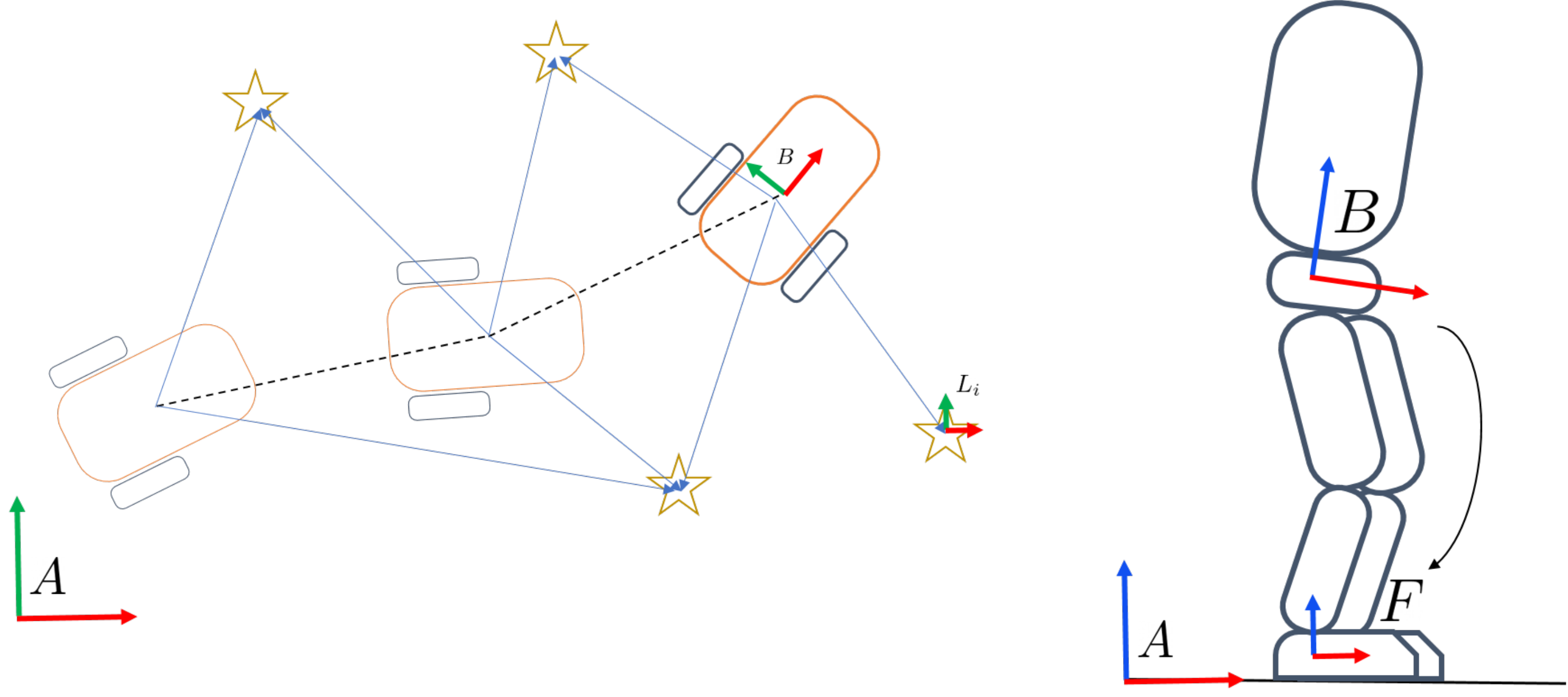}
\caption{A demonstration of the fundamental problem of estimating spatial relationships in SLAM and humanoid base estimation.}
\label{fig:chap:soa:slam-base-sim}
\end{figure}

The primary problem of state estimation for humanoid robots requires determining the spatial relationship of the many links of the robot in its environment.
For the objective of humanoid base estimation, we survey the state-of-the-art in Simultaneous Localization and Mapping (SLAM), Lie-theoretic estimation, legged robots and humanoids community. 
Since SLAM, at its fundamental level, tries to solve a more general problem that is usually faced in robotics: understanding uncertain spatial relationships between the robot and other objects in its environment and their motion in the environment, it becomes an important topic of study in state estimation.
Further, the theory of Lie groups offers a toolkit for useful representations of these spatial relationships.
Drawing connections between the SLAM community and the legged robots community allows us to understand the landscape of theoretical developments made in both these communities in the context of state estimation, which might help further the existing body of literature for humanoid robots.
Many of the promising state-of-the-art techniques that have been applied to the state estimation of highly-articulated structures like legged robots and humanoids have originally been ideas developed for general mobile robot systems, mainly in the context of navigation systems.
This is due to an inherent similarity in the underlying problem, as seen in Figure \ref{fig:chap:soa:slam-base-sim}.
However, it must be emphasized that application of such techniques on humanoid robots and legged robots are usually more challenging than the other mobile robot systems due to many inherent complexities in the former that might usually not be faced by a generic mobile robot.

The latter objective aims at the reconstruction of the whole-body motion of a human using measurements obtained from a distributed set of sensors worn by the human.
These measurements usually describe the motion of body parts indirectly, which when incorporated with the knowledge of the skeletal model of the human, can enable achieving whole-body motion reconstruction.
For this purpose, we look at the state-of-the-art estimation methods employed for human motion estimation using wearable sensing technologies such as distributed inertial sensors and sensorized shoes providing ground reaction force measurements.

The aim of this chapter is to first draw out relevant works done in related fields of state estimation before diving into the survey for humanoid robot and human state estimation.
Having painted a big picture, we then zoom into the context of this thesis to highlight where the contributions fit in, and finally review the design decisions taken for the proposed contributions.

The chapter begins with a review of the state estimation methods used for SLAM and navigation systems, in Section \ref{sec:chap:soa:mobile-robot-state-estimation}.
Then, Section \ref{sec:chap:soa:state-estimation-on-lie-groups} highlights the trends in the application of Lie theoretical tools for filter design and how some issues in SLAM are solved through such a filter design.
Section \ref{sec:chap:soa:humanoid-state-estimation} covers the advances made in the topic of state estimation for legged and humanoid robots and their connections with the topics of SLAM.
Then, a brief review related to the different approaches for achieving human motion estimation using motion capture technologies is presented in Section \ref{sec:chap:soa:human-motion-estimation}.
Finally, we introduce the context of this thesis and its contributions to the topics of state estimation in humanoid robots and human motion estimation in relevance to the cutting-edge technologies developed in both fields.

\section{SLAM and Navigation Systems}
\label{sec:chap:soa:mobile-robot-state-estimation}

In the context of robotics, state estimation finds itself focusing mainly on the problem of estimating a set of quantities such as position, orientation, and velocity of the robot that faithfully describe its motion over time using measurements coming from a variety of sensors mounted on the robot (\cite{barfoot2017state, kok2017using, schwendner2014using}).
In plain words, it is necessary to answer the following questions; "what does the world look like?", "where am I in this world?" and "where should I go next?".
These questions fall under the general problem of \emph{simultaneous localization and mapping} (SLAM, \cite{thrun2002probabilistic, cadena2016past, fernandez2012simultaneous, bresson2017simultaneous}), where the robot estimates its location in the environment (\emph{localization}) while simultaneously building a map of its environment (\emph{mapping}).

Of particular interest are navigation systems that focus on a subset of the SLAM problem relying mainly on light-weight and low-cost sensors for motion estimation.
Ubiquitously-used sensors such as inertial and visual sensors enable the state estimation of robots even in situations lacking the availability of absolute position sensors or in GPS-denied environments (for example, underwater and indoor environments).
These systems are usually referred to as Inertial Navigation Systems (INS) which rely on an Inertial Measurement Unit (IMU, \cite{chatfield1997fundamentals, titterton2004strapdown}) and Visual Inertial Navigation System (VINS, \cite{huang2019visual}) which additionally includes a visual source such as a camera.
The camera provides rich information about the surrounding environment complementing the IMU measurements constituting the angular velocity and linear acceleration.
Several decades of continuing research efforts are being put into the field of INS, VINS, and SLAM making them a significant body of literature consisting of several theoretical and computational developments relevant to state estimation problems. \looseness=-1


\subsection{Navigation Systems}
Advances in sensor technologies are crucial for achieving a reliable state estimation for our robots.
While Global Positioning Systems (GPS) remained an important technology of the twentieth century mainly used for aerospace and navigation purposes, inertial navigation systems have been a cornerstone technology in tracking the motion of a body in a self-contained or a "\emph{proprioceptive}" manner, in the absence of global information.
Inertial sensors composed of a three-axis linear accelerometer and a three-axis gyroscope were primarily used for attitude estimation of spacecraft vehicles (\cite{crassidis2007survey}) and eventually became frequently used in navigation.
The gyroscope measures the angular velocity and the accelerometer measures the specific force acting on the sensor constituting the sensor acceleration and Earth's gravity.
These quantities are implicitly related to the position and orientation of the body equipped with such a sensor.
The technique of having the inertial sensors rigidly attached or strapped down to the body of the host vehicles (\cite{titterton2004strapdown}) allowed the emergence of small, light-weight, and more accurate inertial navigation systems known as the strap-down inertial navigation systems.
This technology had a dramatic impact on navigation leading to highly efficient and cost-effective methods. \looseness=-1

\begin{figure}[!h]
\centering
\includegraphics[scale=0.35]{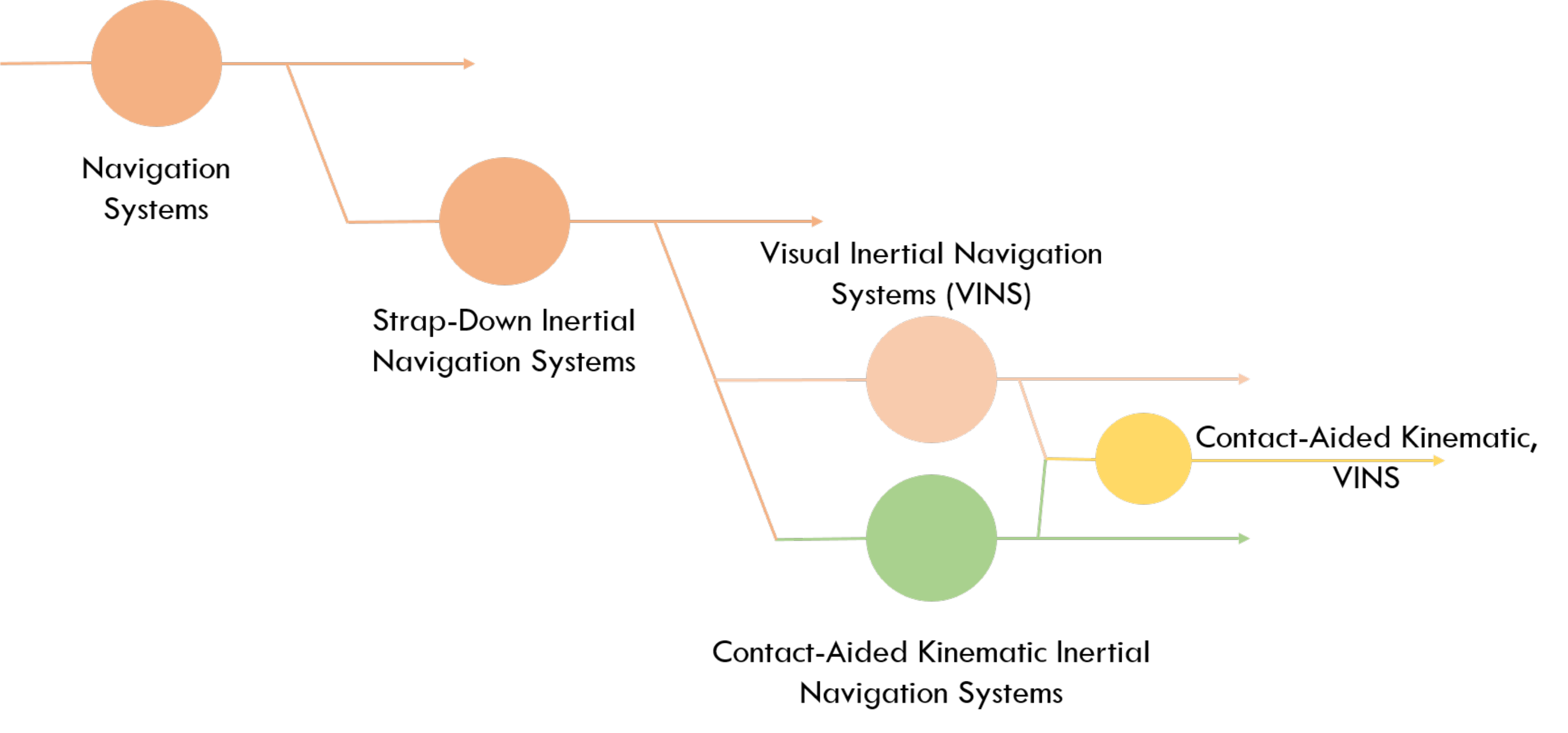}
\caption{Sensor-based classification of navigation systems and its connection to humanoid base estimation.}
\label{fig:chap:soa:ins}
\end{figure}

Although inertial sensors provide measurements at high sampling rates, the accuracy of the estimation using inertial sensors depends on the sensor characteristics like precision, biases affecting the sensor measurements, and their noise levels.
These effects lead to significant drifts in the estimated position and orientation over a long duration of operation, thereby limiting their use only for a short duration.
The more expensive an inertial sensor, the more resilient it is towards these degrading effects resulting in better motion estimates with lesser drifts.
However, such an expensive sensor becomes a hindrance for widespread, commercial deployment.
A standard approach relies on complementing the inertial sensors with other sensors for enabling inexpensive and reliable motion estimation.
The most sought-after sensor is a camera which is a low-cost, energy-efficient sensor that provides rich information about the environment.
Using a camera to complement the inertial sensor for motion estimation results in what is referred to as Visual Inertial Navigation Systems (VINS, \cite{huang2019visual}). \looseness=-1

Many other relevant sensor fusion techniques in the SLAM community try to include sensors like a laser rangefinder, LIDAR, SONAR, infrared sensors, Ultra Wide Band (UWB) position sensors, Wifi networks, and auditory sensors for state estimation. 
Since we are mostly interested in the theoretical development perspectives of the state estimation algorithms from the SLAM community which are usually common for a variety of sensors used, we will narrow our focus on VINS algorithms developed for Visual-Inertial Odometry (VIO) systems in subsequent sections.
These algorithms can provide insights into the algorithms used in the state-of-the-art methods for humanoid state estimation (see Figure \ref{fig:chap:soa:ins}). \looseness=-1

\begin{figure}[!h]
\centering
\includegraphics[scale=0.5]{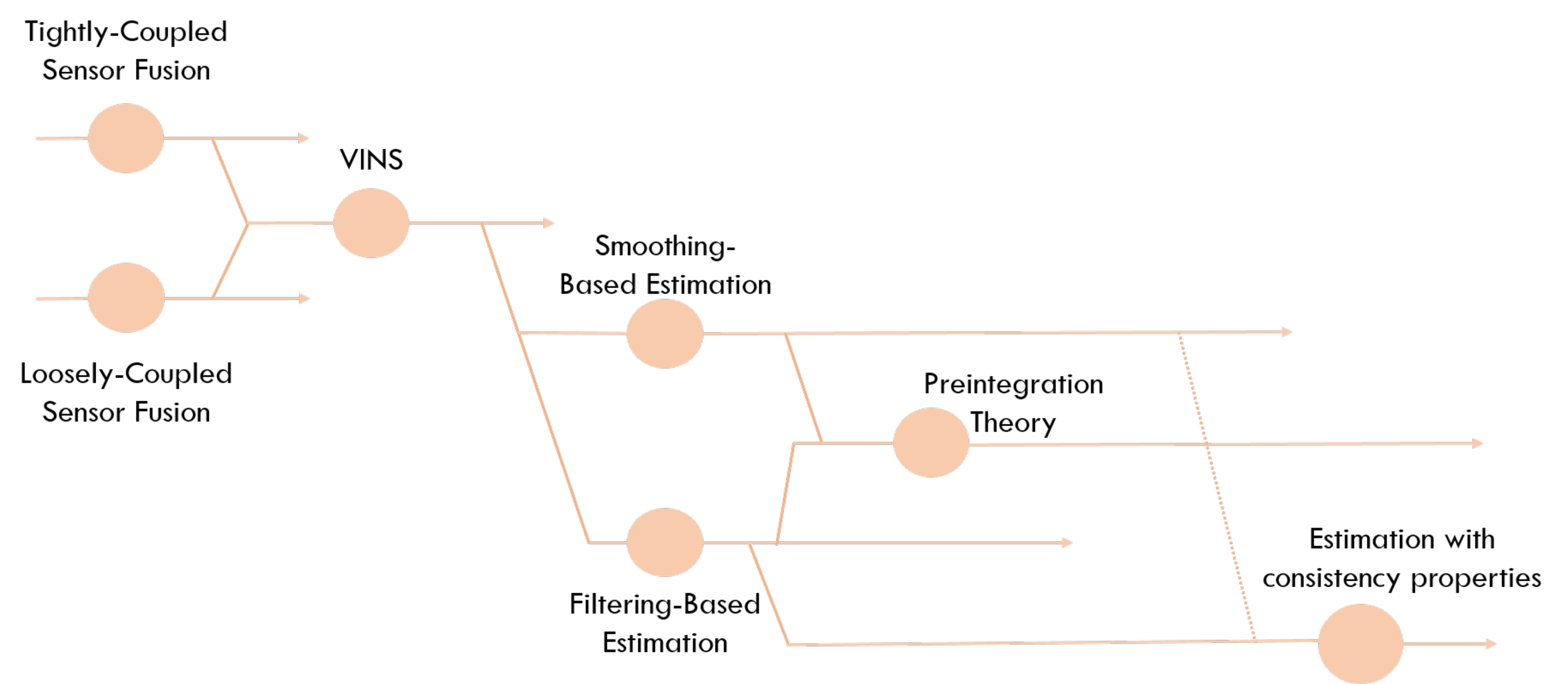}
\caption{Evolution of theoretical tools and approaches for solving the VINS problem.}
\label{fig:chap:soa:ins}
\end{figure}

 \subsection{Tightly-Coupled vs Loosely-Coupled Sensor Fusion}
A broad classification of VINS algorithms arises depending on how different sets of measurements are combined within the state estimation algorithm.
One such classification is based on the loosely-coupled sensor fusion and tight-coupled sensor fusion methods.
Within a loosely coupled sensor fusion approach, the visual and inertial measurements are pre-processed separately to infer the motion constraints established by each of these measurements on their own and then are combined to obtain a fused estimate  (\cite{weiss2011real, lynen2013robust}). 
On the contrary, tightly-coupled sensor fusion combines the measurements within a single process resulting in higher accuracy and less information loss (\cite{shen2015tightly, eckenhoff2019tightly}).
The type of sensor-fusion used to implement an estimator is usually a design decision depending mainly on the computational budget and simplicity-to-performance trade-offs.  

 \subsection{Filtering-based vs Optimization-based Estimation}
 \label{subsec:chap:soa:filter-vs-optim}
Under a certain taxonomy, VINS algorithms can be broadly classified into filtering-based and optimization-based methods.
Most filtering-based estimation algorithms are commonly based on the EKF, where the IMU measurements are used for state propagation and the vision-based measurements are used for measurement updates (\cite{mourikis2007multi, bloesch2014fusion}).
EKF relies on the linearization of nonlinear models using a first-order Taylor expansion around the current estimate. 
As long as the linearization is made around the true state, EKF is shown to demonstrate optimality and theoretical guarantees for convergence. 
Due to its simplicity and ease of implementation, it has been a very common tool in developing SLAM algorithms. 

EKF methods although suffer from limitations when handling highly nonlinear measurement models due to the fact that the linearization of these measurement models is done only once at the measurement update step.
This might potentially lead to large linearization errors leading to the degradation of the estimator performance and eventually the divergence of the filter. 
UKF offers a derivative-free solution;
however, the usage of UKF for SLAM algorithms is limited due to its computational complexity. \looseness=-1 

In contrast to filtering methods, optimization-based state estimation methods have been proposed that solved a non-linear least squares problem over a set of measurements aiming to minimize the residual error between the expected measurement given the current estimated state and the actual measurement (\cite{qin2018vins, nisar2019vimo, wisth2021unified, ding2021vid}).
This approach can be viewed as maximizing the conditional probability of the state estimate given a set of observations.
This approach allowed the reduction of linearization errors through repeated linearizations at the expense of computations. 
Nevertheless, a strong dependence on the linearization points for both the filtering and the optimization methods makes them vulnerable to linearization errors. 
The choice for opting either a filtering-based or an optimization-based estimation framework is a crucial design decision that needs to be taken upfront based on the requirements of modularity, performance, computational complexity and implementation issues.
\looseness=-1

 \subsubsection{Filtering vs Smoothing}
 A common debate or a design decision that arises while constructing estimators is the choice between filtering and smoothing (\cite{strasdat2010real}).
 The filtering approach fuses measurements in a sequential manner without considering the complete history of states while smoothing methods perform a batch optimization by considering either the complete state trajectory, or a selected keyframe of the past states or a sliding window of a number of previous states.
 The latter allows for a drift-free long term operation given the continuous re-optimization of the estimated states given the history and incoming measurements.
 Such a dichotomy in the estimation approaches clearly begs the question of why and whether one approach is more suitable for the application in hand.
 At a conceptual level, both these approaches boil down to a nonlinear least-squares problem that minimizes a desired cost function or a Maximum a Posteriori (MAP) inference that maximizes the posterior density of the states given a set of measurements (\cite{dellaert2017factor}).
 Thinking visually in terms of a factor graph, for the filtering approaches all the previous states are marginalized, thus keeping the graph compact and not growing with time. 
 Whereas, for smoothing approaches such a factor graph might grow arbitrarily causing computational cost to rocket rapidly.
 However, in terms of modularity in implementation and scalability to several cost functions and constraints within the estimation problem, smoothing based methods might offer some advantage also allowing us to also choose from several off-the-shelf numerical optimizers to solve for the least-square solution.
 The relative merits of smoothing- and filtering-based approaches in terms of accuracy, computational cost, scalability and modular implementation relies mainly on the application in hand and their efficiency and performance must be considered as a key-performance indicator for designing estimators.
  \looseness=-1

 \subsubsection{Preintegration Theory}
\begin{figure}[!h]
\centering
\includegraphics[scale=0.35]{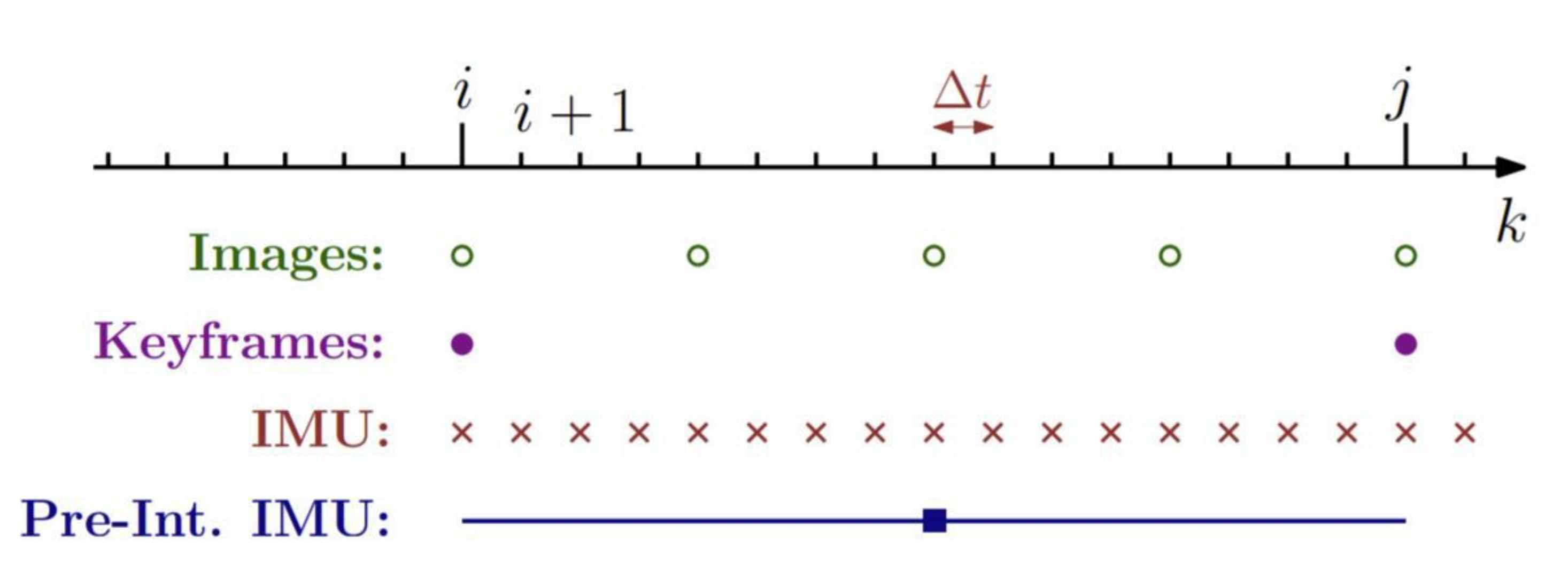}
\caption{IMU preintegration in a VINS problem summarizing high frequnecy IMU measurements as a preintegrated IMU measurement between the image keyframes. (\cite{forster2016manifold})}
\label{fig:chap:soa:preint}
\end{figure}

Preintegration theory is an important tool that was first introduced in the seminal work of \cite{lupton2011visual} that allowed to summarize a large number of high-frequency measurements into a single motion constraint between the interval of low-frequency measurements, in a manner independent of initial conditions. 
The summarized motion constraint is then combined with the low-frequency measurement enabling fast computations of the filter-based or optimization-based approaches.
Figure \ref{fig:chap:soa:preint} shows the preintegration of high-frequency measurments between the low-frequency keyframes of visual measurements.
This computationally efficient alternative for integrating inertial measurements within a VINS framework is being adopted as a standard in many state-of-the-art estimation approaches (\cite{forster2016manifold, eckenhoff2020high, qin2018vins}).
Further, the preintegration theory is being generalized to consider other high-frequency measurements apart from the inertial sensor measurements (\cite{nisar2019vimo, ding2021vid}). 
The concept of preintegration is being popularized also in the context of legged robot state estimation using factor-graph optimization based state estimation methods (see Section \ref{sec:chap:soa:humanoid-state-estimation} for a detailed review).
However, it must be noted that the idea was originally introduced with filtering strategies, hence can be effectively applied with easily implementable filtering or constrained filtering methods, well suited for legged robot state estimation, especially dealing with IMU based prediction models during flight phase or absence of contacts.
\looseness=-1

 \subsection{Effects of Inconsistencies and Drifts}
 \label{sec:chap:soa:inconsistency}
Most VINS estimators by construction do not consider any global measurements and rely on reconstructing the position and orientation of the robot along with its linear velocity from a partial set of observations from IMU and visual measurements.
Due to the nature of the measurement set, the VINS problem has an inherent unobservability along four directions corresponding to the global translation and the rotation about the gravity vector.
However, there might be cases where the dimensions of the unobservable subspace are reduced than its actual dimension of four.
This situation can arise in the presence of large linearization errors or due to the choice of the linearized system itself. 
Such an event causes the filter to gain spurious information along the unobservable directions resulting in an inconsistent estimate, which might eventually lead to filter divergence as well.
Several techniques such as first-estimates Jacobian EKFs (FEJ-EKF, \cite{huang2009first}) and observability-based rules for constructing consistent EKFs (OC-EKF, \cite{huang2010observability}) have been proposed to solve the inconsistency problems in the EKF.
The former approach relies on filter Jacobians computed around the first-ever available estimates for every state leading to the error-state system model having an observable subspace of the same dimension as the underlying nonlinear system.
While the latter relies on the selection of linearization points by imposing constraints on the observability matrix to always maintain the expected dimensions in its unobservable subspace.
The inconsistency problem is not specific to only EKF based estimators but also optimization-based estimators that rely on Jacobian matrices through the linearization of the non-linear models.
Similarly, this problem pertains not only to VINS but affects also legged robot state estimation and it is crucial to be considered during estimator design. \looseness=-1
 
 \subsubsection{Robot-Centric Estimators}
A common approach to constructing VINS estimators is based on a world-centric perspective where the estimates of the robot state are given with respect to a fixed world or inertial frame.
However, a few works (\cite{huai2018robocentric}) which construct robot-centric estimators have shown to preserve observability properties independent of linearization points and greatly reduce the filter divergence. 
In the robot-centric approach, the estimates are always provided with respect to the robot position and allow to maintain better consistency in the filter. 
This approach might also be useful for maintaining local maps of the environment thereby improving obstacle avoidance. \looseness=-1

\section{State Estimation using Lie Groups}
\label{sec:chap:soa:state-estimation-on-lie-groups}
\begin{figure}[!h]
\centering
\includegraphics[scale=0.5]{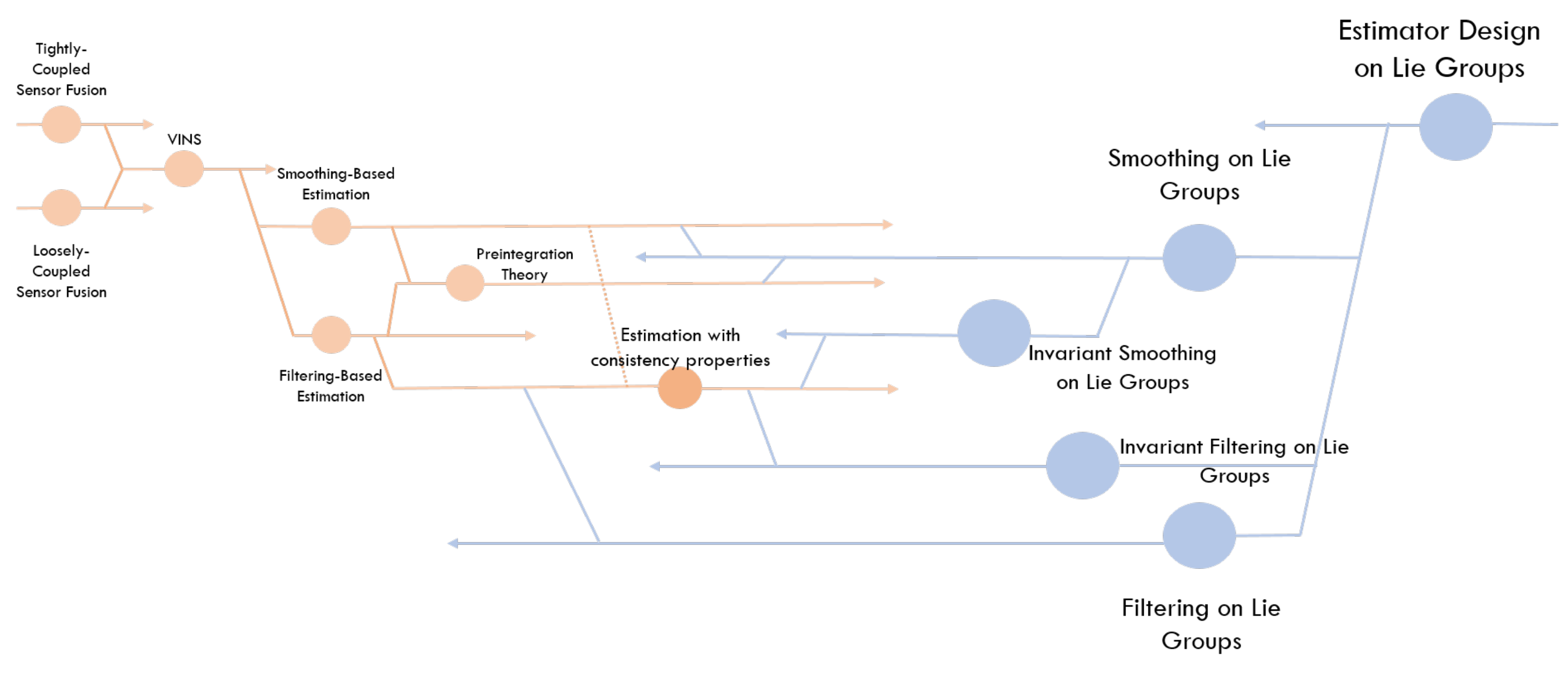}
\caption{Evolution of Lie-theoretic estimation tools and its connections to the SLAM community.}
\label{fig:chap:soa:slam-liegroups}
\end{figure}

So far in the previous sections, we have reviewed a few important developments that have been made to strengthen the state-of-the-art in VINS.
However in the review so far, we had not discussed the representation choice for the state variables and an important paradigm shift that was taking place, in parallel, for constructing efficient state estimation methods that considered different choices of state representations for the spatial relationships.
This paradigm shift was related to the construction of state estimation algorithms by employing the tools provided by the corpus of "Lie theory" which allows solving many problems discussed so far in an implicit manner (see Figure \ref{fig:chap:soa:slam-liegroups} depicting the connection with SLAM).

As noted already, the state estimation problem in robotics is focused on estimating the position and orientation of a body,  put together called as pose.
A common challenge then arises in choosing a proper representation for the pose variables, specifically the representation for the orientation embedded in the pose variable. 
This becomes a crucial modeling choice. 
A few common representations for orientation include rotation matrices, quaternions, Euler angles each of which is either an over-parametrized representation or prone to singularities.
Each of these representations has some constraints that need to be satisfied in order to be physically meaningful. 
Considering these variables for state estimation requires the design to respect the constraints they impose, which is rather difficult to establish innately, leading to complex estimator designs (some of which are reported by \cite{hertzberg2013integrating}). 
For example, Euler angle representation, which is a minimal representation for the rotations, is computationally efficient however is prone to singularities. 
Different types of Euler angle choices have singularities in different configurations.
There exists a few estimator designs which rely on switching the choice of the Euler angle when the current choice approaches its singularity configuration. 
This is a working idea, nevertheless, an unnecessarily complex design.
Many other works rely on the over-parametrized yet computationally efficient representation of unit quaternions which relies on a unit norm constraint and requires an orthonormalization trick whenever the constraint is violated. \looseness=-1

Further, associating uncertainties to such quantities is usually challenging, since it is not intuitive to define a probability distribution for these variables.
This is due to the fact that these variables are not vectors and establishing a notion of probability distribution with a valid definition of a probability density function (pdf) is not straightforward.

Lie theory, originally developed by the Norwegian mathematician Sophus Lie (1842-1899), offers tools to consider these types of variables in the state, without the need to to impose any constraints while allowing to establish an appropriate notion of uncertainty for such variables in a feasible manner.
A remarkable amount of theoretical developments assimilated in the works of
\cite{hall2013lie, mahony2008nonlinear,  bonnabel2009non, chirikjian2009stochastic, chirikjian2011stochastic, sola2018micro, barfoot2014associating, barfoot2017state, blanco2010tutorial} provides tools and methodologies for constructing estimators that allow for an appropriate representation and uncertainty management for complex non-linear quantities like rotations and poses. 
With a proper choice for state and its associated uncertainty representation, we can construct rigorous and accurate state estimators.
\looseness=-1

\subsection{Uncertainty Representation}
One of the earliest notable works that dealt with uncertainties associated with homogeneous transformation matrices was done by \cite{su1992manipulation} where the uncertainty representation is derived using the first-order approximation of the exponential map of $\SE{3}$. 
This work demonstrated propagation, fusion and compounding of spatial uncertainties. 
These techniques for uncertainty representation of rigid body transformations were then consolidated in \cite{wang2006error, chirikjian2009stochastic, chirikjian2011stochastic} with the application of the Concentrated Gaussian Distributions (CGD) on Special Euclidean groups, which allowed for the formulation of the so-called concentrated pdfs over $\SE{3}$.
The CGD formulation relied on the definition of the mean of the distribution in the Lie group, with the covariance defined in the Lie algebra(where calculus operations are possible).
\cite{wang2008nonparametric} considered a second-order theory. instead of the first-order approximation of the exponential map, for the propagation of mean and covariance of the pdfs over \SE{3}.
\cite{wolfe2011bayesian} generalized the notion of CGD for all Lie groups considering first and second-order theories for uncertainty propagation and a rigorous Bayesian fusion on Lie groups. 

\cite{barfoot2014associating} provides an accessible and specific implementation for handling spatial uncertainties that is only subtly different from the previous works but gains in terms of its ease in computations.
The previous works begin by computing the pdf directly over the group and switches to parametrizing the pose objects using exponential coordinates, while \cite{barfoot2014associating} begins by defining a pdf in the vector space and then induces a corresponding pdf in the Lie group using the definitions of several Lie group operators.
The latter is an easier approach to intuitively grasp and communicate the notion of uncertainties over Lie groups.

It must be noted that the notion of concentrated Gaussian distributions allows the definition of the uncertainty to be valid only locally which might cause the estimation algorithms to fail when the covariance associated with the variable of interest becomes large. 
A global representation of uncertainties over Lie groups has also been investigated in the context of rotation group of $\SO{3}$ by \cite{lee2008global, lee2018bayesian}.
Nevertheless, the uncertainty representation using the concept of CGD is a commonly used approach and has demonstrated its usefulness in various real-world applications (\cite{barfoot2017state}). \looseness=-1

\subsection{Estimator design on Lie Groups}
\begin{figure}[!h]
\centering
\includegraphics[scale=0.5]{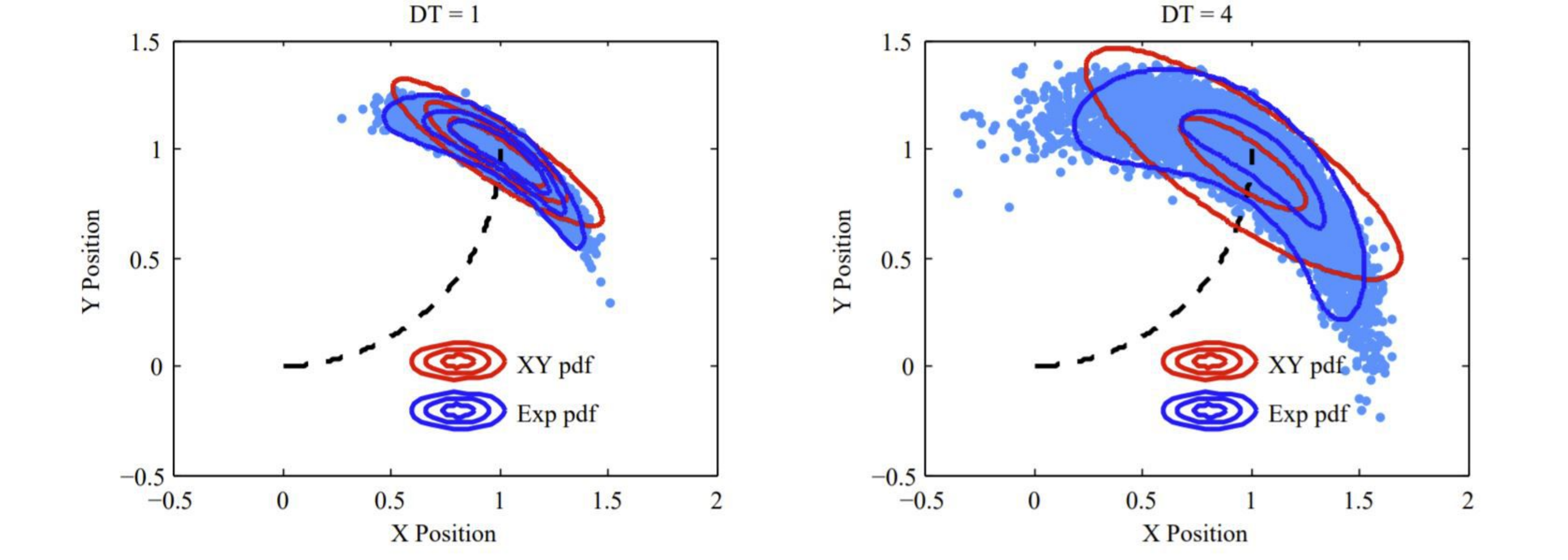}
\caption{The so-called banana distribution formed by a mobile robot moving on a plane along a constant-curvature trajectory. The pdf contours of Gaussian in exponential coordinates marginalized about the heading direction is seen to represent the true distribution more faithfully than the Gaussian in Cartesian coordinates (\cite{long2012banana}).}
\label{fig:chap:soa:banana}
\end{figure}

The viewpoint of designing state estimators on Lie groups has contributed significantly to the SLAM community in the recent decade.
\cite{long2012banana} applied the concept of Gaussians in exponential coordinates and Lie group representation for poses to the EKF-SLAM problems of a differential drive robot and demonstrated that such a definition can be used to faithfully describe the banana distribution that results from multiple sample paths obtained from the robot moving in planar space, as shown in Figure \ref{fig:chap:soa:banana}.
This banana distribution does not have elliptical probability density contours that are usually obtained from Gaussians in Cartesian coordinates and such a density usually introduces inconsistencies in the estimation when the true distribution becomes widely-spread. 
The former representation which captures the underlying distribution with more accuracy and precision allows for developing consistent estimators over the groups of $\SE{2}$.

While state estimators considering a probabilistic perspective were being developed, a considerable amount of work was done by \cite{bonnabel2009non} for developing the theory of non-linear observer design exploiting the geometry of the Lie group structure.
These designs focused on exploiting the so-called \emph{symmetry} of the system which are certain features in a system that remain unchanged when subjected to some transformation.
This work evolved into the topic of invariant filtering over Lie groups which exploited the dynamics structure for a specific class of systems called as \emph{group-affine systems}.
This property enables an autonomous error evolution, meaning the evolution of the error in the state variable with respect to the true system state is independent of the state variable itself (invariance).
Additionally, another property of these systems known as the \emph{log-linearity} property allowed the non-linear error of the state evolving over the Lie groups to be linearized in a straight-forward manner using a suitable change of coordinates.
These combination of properties resulted in an estimator design with strong convergence guarantees of the estimated state towards the true state (\cite{barrau2015non, barrau2017invariant}).
Further, measurement models evolving in vector space can be chosen in such a way that it leads to a definition of the so-called \emph{invariant observations}.
This results in the update equations of the propagated error using the available sensor measurement becoming independent of the system state.

With keen observation, it can be noticed that at a certain level the linearization through a suitable change of coordinates is quite similar to the notion of uncertainty using the CGD.
This is true due to the fact that the underlying formulation for both approaches depends on approximating the so-called Baker-Campbell-Hausdorff (BCH) formula which is used to compute the logarithm of a product of two exponentials (see Def. \ref{def:chap02:BCH-formula}).
However, the strength of the invariant filter design lies in the properties of autonomous error-evolution and autonomous error-updates which lead to time-invariant Jacobians for filter design.
This automatically implies a remedy for the consistency problems faced in EKF-SLAM or optimization-based methods which rely on the Jacobians that depend on the choice of linearization points around the current state estimate, as discussed in Section \ref{sec:chap:soa:inconsistency}.
The applications of Invariant Extended Kalman Filter (InvEKF) have been demonstrated in \cite{barrau2015non} for constructing consistent EKF-SLAM estimators.

\cite{barrau2015non} also introduces the group of double direct isometries or the group of extended poses $\SEk{2}{3}$ that extends the pose objects of $\SE{3}$ to also contain linear velocity within the same matrix in a semi-direct product with the rotation, similar to the position.
This representation of rotation, position and linear velocity embedded within the same matrix allows a direct relation relation of quantities of interest efficiently within the context of inertial navigation algorithms.
\cite{wu2017invariant} applied the InvEKF algorithm for the VINS problem accommodating the $\SEk{2}{3}$ representation to demonstrate the improved consistency properties within this context.

While InvEKF proposed by \cite{barrau2017invariant} was suited for continuous systems possessing symmetries such as the group-affine dynamics, it did not handle observations evolving over matrix Lie groups.
\cite{bourmaud2013discrete, bourmaud2015continuous, bourmaud2015estimation} developed EKF algorithms using the concept of CGD with both the state and observations evolving over distinct matrix Lie groups; Continuous-Discrete Lie Group EKF (CD-LG-EKF, otherwise renames as CODILIGENT within this thesis) for dealing with continuous system dynamics and discrete observations and a Discrete Lie Group EKF (D-LG-EKF, otherwise renamed as DILIGENT within this thesis) for handling discrete system dynamics and discrete observations.
This is an interesting improvement from the case where only vector observations were considered and facilitated a natural generalization of the EKF over vector spaces to EKF over matrix Lie groups (as a preview, we explore these methodologies within the scope of the thesis). 
However this filter formulation does not account for group-affine properties of the system dynamics, therefore might not implicitly lead to a consistent estimator.  
\looseness=-1

Very recent works of \cite{phogat2020discrete, phogat2020invariant} try to address the consideration of matrix Lie groups for the representation of both state and observations within the framework of invariant filtering.
This approach is different from the usual approaches of transporting the non-linear error to the vector space using the logarithm map for the necessary linearization.
This approach of invariant filtering on matrix Lie groups could provide an alternative that strengthens over the framework developed by \cite{bourmaud2013discrete, bourmaud2015continuous}. \looseness=-1

It is also worth mentioning some notable works along the line of estimator design on Lie groups.
\cite{brossard2017unscented, brossard2018unscented} develop an Unscented Kalman Filter on Lie groups by exploiting the concept of CGD applied to the problem of visual-inertial odometry.
The computational burden of UKF-based VINS is relaxed by combining the invariant propagation of the InvEKF along with the use of an unscented transform inferred Jacobian for the measurement update.
\cite{forster2016manifold} extended the IMU preintegration introduced in \cite{lupton2011visual} to the group of rotations $\SO{3}$, enabling singularity-free uncertainty management within the smoothing based approaches. 
Recent noteworthy work is from \cite{mahony2017geometric} which proposes a non-linear observer for the SLAM problem with the introduction of a so-called $\text{SLAM}_k(3)$ group which is a generalization of the group of extended poses ($\SEk{k}{3}$) to contain $n$ vector quantities in semi-direct product with the rotation (in this thesis, we will use $\SEk{k}{3}$ for denoting this group).
This observer does not require any linearization and demonstrates constant-time computational complexity along with global-asymptotic stability in estimating the robot motion along with sparse landmarks, owing to much more rigorous geometric formulation.
Some studies have also been made in \cite{chauchat2018invariant, huai2021consistent} to design invariant smoothing algorithms for building consistent optimization-based VINS estimators. \looseness=-1

\section{State of the art in Humanoid State Estimation}
\label{sec:chap:soa:humanoid-state-estimation}

\begin{figure}[!h]
\centering
\includegraphics[scale=0.5]{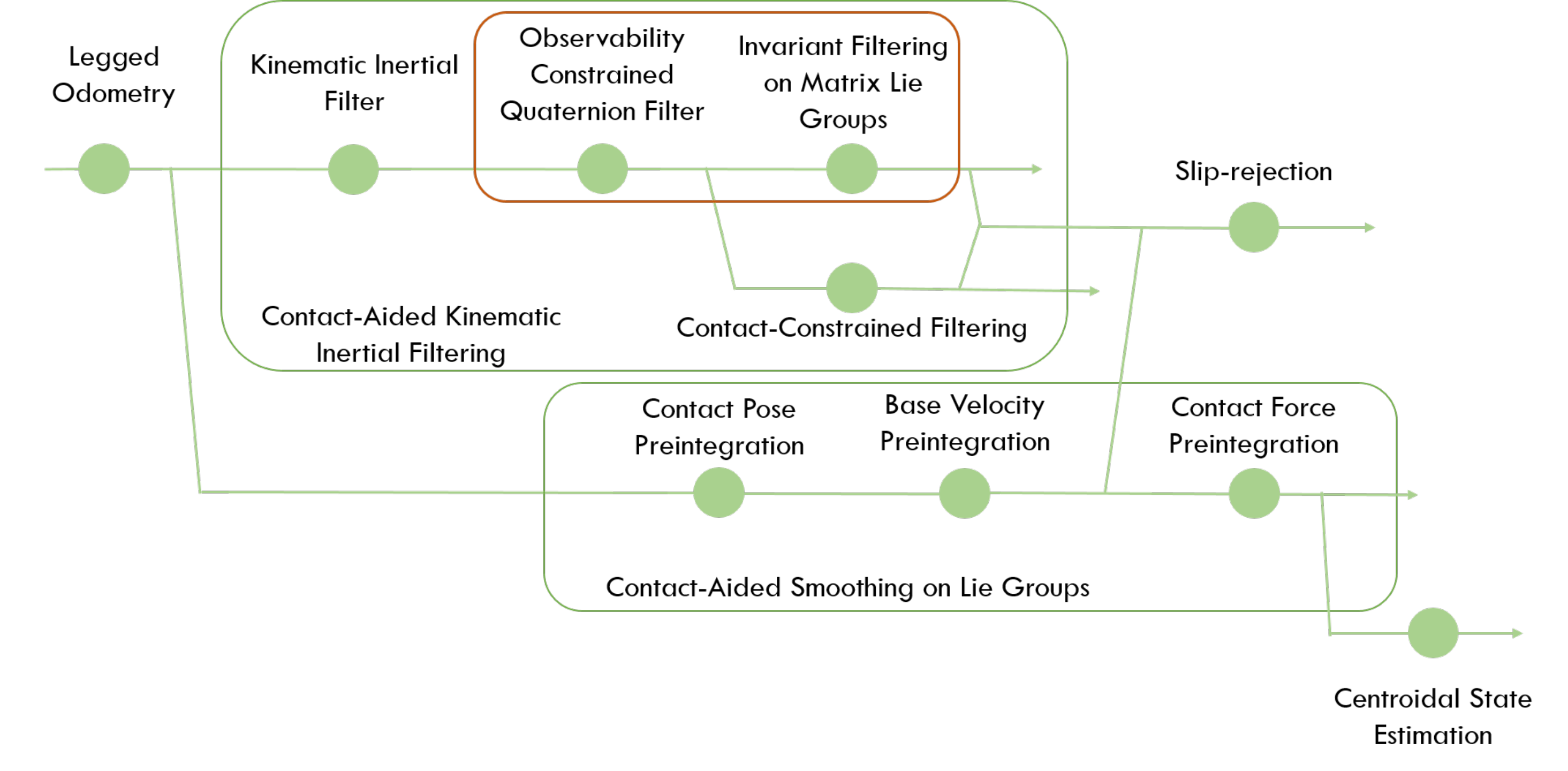}
\caption{Evolution of theoretical tools for proprioceptive state estimation of a legged robot. The methods proposed in this thesis are closely related to the tools depicted in the red box.}
\label{fig:chap:soa:proprio-legged}
\end{figure}

Simply taking a SLAM-based perspective of state estimation is not the most feasible way to achieve a reliable estimation on legged and humanoid robots.
When it comes to the latter, there is a clear distinction in the need of estimation aimed for autonomy and feedback control.
This is due to the fact that these types of robots are in persistent interactions with the environment and rely heavily on controlling the external wrenches exerted on the environment to generate a desirable motion.
Therefore, it becomes necessary to exploit the mechanical structure of these systems, their underlying geometry, and their intermittent interactions with the environment in order to design suitable state estimation technologies.
Further, these robots rely on high-frequency, low-latency state estimates for feedback control,
Rephrasing the words of \emph{Maurice Fallon} from one of his talks on state estimation for legged robots, "control guys do not want perfect estimates, they want good estimates faster" \footnote{\href{https://www.youtube.com/watch?v=Rqfnq6uZvK8&list=PLTzTxsG2Q0FKlsCRU6xT86Mw0jSph54Zl&index=11}{Walking robot navigation and state estimation: a talk by Maurice Fallon and Marco Camurri}}, it is evident from practical experience that proprioceptive estimation is fundamental for state estimation since such high rates are not yet feasible through estimation based on exteroceptive sensors (exception being the relatively young technology of event-driven cameras).
In this section, we will first survey the state-of-the-art state estimation for legged robots exploiting the point foot contact model focusing both on proprioceptive (see Figure \ref{fig:chap:soa:proprio-legged}) and their extension to exteroceptive approaches.
Subsequently, we will review the body of literature specific to humanoid base estimation.

\subsection{State Estimation for Legged Robots}
 \label{subsec:chap:soa:estimation-legged-robot}
One of the earliest works to estimate the body pose of a legged robot using its kinematic configuration in a legged odometry fashion was done by \cite{lin2006legged}, demonstrated for a walking hexapod robot operating on flat terrain.
Later, this approach was extended to fuse also inertial measurements within an EKF based approach in \cite{lin2006sensor}. \cite{chitta2007proprioceptive} proposed a proprioceptive estimator for global localization of a quadruped by combining kinematic and inertial measurements along with terrain information through a particle filter. 
This estimator was designed for robots performing statically stable gaits and allowed for global localization through data associated with the terrain characteristics.
The idea of global localization through proprioception is rather interesting, however, the uncertainty about its localization grows rapidly in the case of flat terrains due to feature scarcity for data association and there are also possibilities of localizing towards a wrong state in case of similar terrain characteristics.

\subsubsection{Contact-Aided Filtering}
An observability-constrained Quaternion EKF was developed by \cite{bloesch2013state, bloesch2017state} combining leg kinematics and inertial measurements to design a consistent proprioceptive estimator.
The linearization points for the Jacobian computations in the filter were chosen by constraining the unobservable subspace of the filter to always respect its ideal dimensions (translation and rotation about gravity). 
Quaternion EKF implies that the chosen representation for the base orientation within the state vector was a quaternion representation.
The linearized error corresponding to the base rotation was chosen in exponential coordinates.
Further, the state vector was augmented to maintain also feet positions to better constrain the problem. 
This estimator draws inspiration from the SLAM community where the robot would simultaneously estimate its state along with landmark positions from the environment. 
Analogously, here the landmarks are replaced by foot positions with relative measurements coming from the leg kinematics.
\cite{yang2019state} proposed a derivative-free alternative for \cite{bloesch2013state} using an SR-UKF through a contact-centric estimator design differently from the IMU-centric approach used in the latter.
Unlike \cite{bloesch2013state}, they consider using accelerometer measurements in the observation model instead of passing them as inputs to the prediction model.
This allows the estimator to be resilient from the sensitivity of the accelerometer measurements with foot-ground impacts.

\cite{hartley2019contact, hartley2020contact} combines the efforts made in legged robots community (by \cite{bloesch2013state}) and SLAM community (by \cite{barrau2017invariant}).
The invariant filtering approach allows to maintain consistency and observability properties of the filter while enabling fast convergence from random initial states.
This estimator uses $\SEk{k}{3}$ Lie group for state representation to estimate robot base state along with foot positions, easily extensible to consider also landmark positions from the environment.
Alternately, a proprioceptive estimator using a cascade architecture is proposed by \cite{fink2020proprioceptive} which decouples the overall state estimation through a non-linear attitude estimation followed by a linear estimation of position and velocity. 
A globally exponentially stable non-linear attitude observer on $\SO{3}$ is to handle large initialization errors.
This estimator provides a simple yet effective loosely-coupled sensor fusion approach.
Such contrasting filter designs beg the question of choice between tightly-coupled and loosely-coupled fusion approaches for legged robot state estimation, in the context of simplicity-performance trade-offs.

\subsubsection{Contact-Aided Smoothing}
Several works using the optimization-based state estimation approach and exploiting the theory of Lie groups have also been proposed recently drawing inspiration from the trends in the SLAM community.
\cite{hartley2018legged} developed a loosely-coupled smoother to complement the pose measurements coming from a perception system with forward kinematics, IMU, and contact measurements.
The preintegration theory is used to design a pre-integrated contact factors in \cite{hartley2018legged, hartley2018hybrid} that summarizes the evolution of a contact frame pose between successive low-frequency measurements assuming a rigid point contact model and accounting for slip through Gaussian noise models.
\cite{fourmy2021contact} consolidates this theory through the concepts of Lie groups to provide a unified framework for preintegration of arbitrary proprioceptive measurements.
They proposed a contact force preintegration factor based on centroidal dynamics to be used with pre-integrated IMU factors, legged odometry, and centroidal kinematics factors within the optimization framework.
This is a tightly-coupled estimation of both centroidal and floating base states crucial for the control of legged robots.
These methods become more sensible when extended towards constrained optimization frameworks that are well-suited for legged robot state estimation.
Although, these tools provide powerful, scalable frameworks for legged robot state estimation, the amount of groundwork, degree of complexity and time-to-deployment for such systems are relatively higher than filtering based methods. \looseness=-1

\subsubsection{Contact-Constrained Estimation}
The intermittent breaking and making of contacts by these articulated systems are quite challenging for state estimation.
Within the context of state estimation, contact kinematics and dynamics play a crucial role in achieving reliable base estimation. 
Usually, a strong assumption of a rigid contact is made for determining base velocities from the null velocities of the stance foot imposed by its holonomic constraints.
This restricts the filter to be used only conservatively within no-slip scenarios.
\cite{varin2018constrained} proposed a contact-constrained Kalman filter that takes the complementarity constraints of the contact force and a polyhedral friction approximation within the filter design allowing for the estimation to be valid also for slipping and sliding scenarios.
This is done by a simple reformulation of the EKF equations into a Maximum A Posteriori estimation through a constrained optimization problem.
A factor graph-based smoothing alternative for this problem is proposed by \cite{wisth2020preintegrated}, where pre-integrated velocity factors are introduced that allow accounting for velocity biases in estimated velocities that may be introduced in scenarios of slipping or sliding, which might otherwise result in position drifts when not considered.
\cite{lim2021walk} introduces gait-specific factors that are adaptive to different walking patterns of the robot and allow to compensate for errors due to vibrating or wobbling motion behaviors on the robot.
This approach can be seen as dynamically scaling the covariances of the factors based on the motion of the features observed in the consecutive camera images to regulate the effect of constraints introduced by leg kinematics.
\cite{teng2021legged} tackles the same problem by replacing the heavy computation of the constrained Kalman filter with augmented sensing technologies. This is done by extending the invariant EKF on Lie groups proposed by \cite{hartley2020contact} to consider base velocity measurements from a camera mounted on the robot as non-invariant observations. 
It can be seen that a suitable sensor augmentation with a simple yet mathematically rigorous filtering approach can allow to tackle complex situations that might be faced by a legged robot.
However, in the absence of such sensors, more rigorous algorithms such as above must be formulated in order to handle the complexities of contacts.
\looseness=-1

\subsubsection{Multimodal sensor fusion}
\cite{nobili2017heterogeneous} present a multi-modal sensor fusion framework for the base estimation of a quadrupedal robot combining legged odometry, inertial measurements, visual and LIDAR measurements in a loosely-coupled modular EKF.
The term loosely-coupled comes from the fact that a visual odometry module is used to obtain pose increments in parallel to point cloud registration from the LIDAR which provides another estimate of the pose, which are then combined with the high-frequency estimates through the EKF.
The performance of this estimator is benchmarked for various real-world experiments and different types of IMU sensing hardware.
A similar work that performs a tightly-coupled multi-modal sensor fusion within the context of optimization-based state estimation is presented in \cite{wisth2021vilens, wisth2019robust}.
The difference between these works is that the latter exploits the theory of Lie groups for state representation with uncertainty representation using CGDs and considers an unconstrained optimization over manifolds for the underlying MAP estimator.

The research interest for invariant filtering and optimization-based approaches has been increasing in parallel within the legged robots community.
While the former allows the construction of filters with consistency properties the latter provides a framework to include constraints from the contact and impact events seamlessly within the estimators.
Many very recent works within the SLAM community have started looking into invariant smoothing methods (\cite{chauchat2018invariant, huai2018robocentric}).
Meanwhile, in the legged robotics community, research in hybrid dynamical systems has led to the development of the so-called Salted Kalman filter by \cite{kong2021salted} that allows for proper uncertainty management in discontinuous scenarios through the so-called saltation matrix.
This accounts for the propagation of uncertainties through well-defined jump maps. 
Hybrid Invariant Kalman filter has also been recently studied in an unpublished work (2021).

All of these methods present a vast landscape of design tools and design decisions that need to be considered while formulating an estimator for legged robots. 
Positioning our own estimator designs in this vast landscape requires to analyze these choices critically, which is reviewed in Section \ref{sec:chap:soa:thesis-context}, after presenting the context of the thesis contributions. \looseness=-1

\subsection{State Estimation for Humanoid Robots}
 \label{subsec:chap:soa:estimation-humanoid-robot}
\begin{figure}[!h]
\centering
\includegraphics[width=\textwidth]{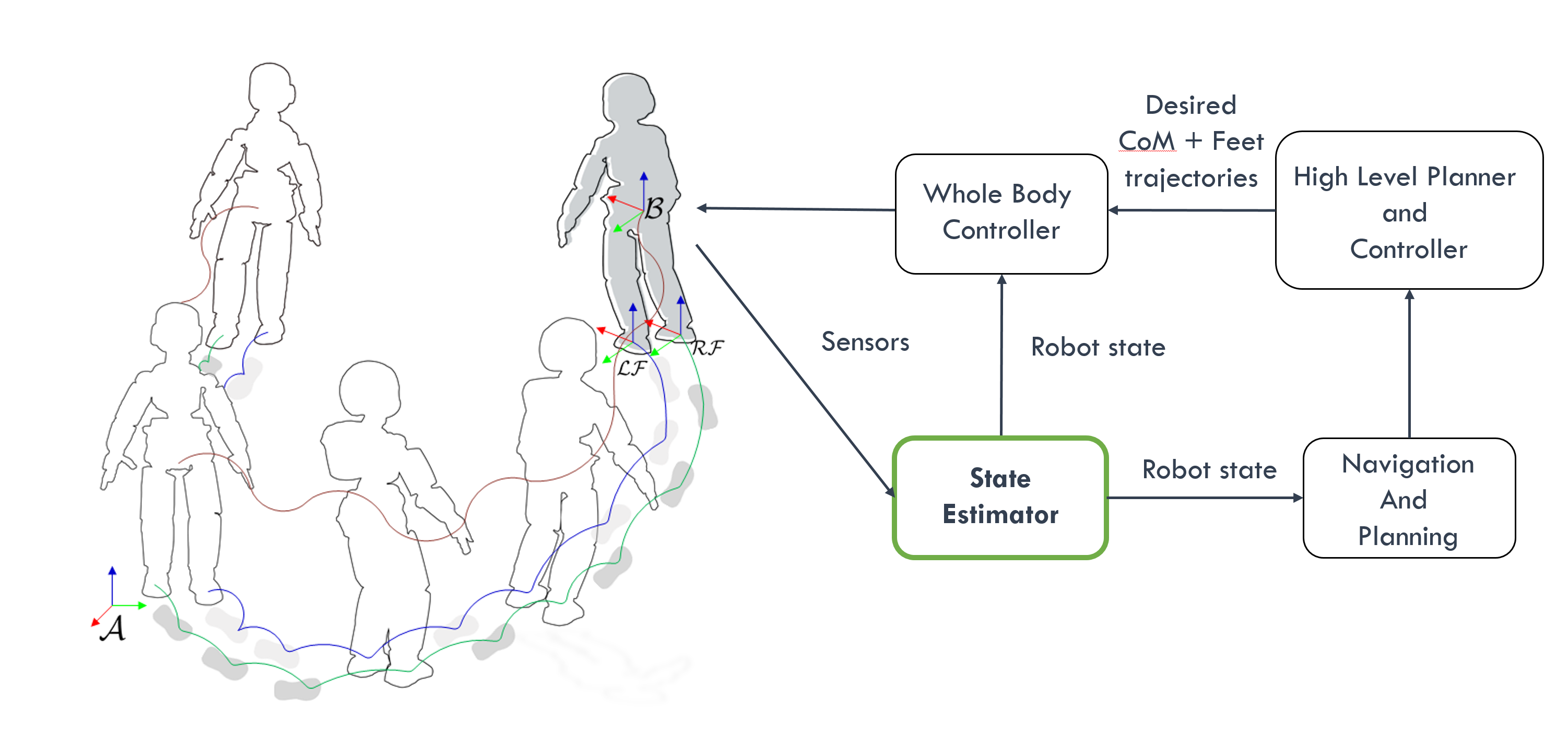}
\caption{A generic control architecture for humanoid locomotion.}
\label{fig:chap:soa:loco-arch}
\end{figure}

Although state estimation for humanoid robots may seem quite similar to that of legged robots given their corresponding characteristics of establishing contacts with the environment, the influence of complex articulated structures in humanoid robots is not negligible.
Figure \ref{fig:chap:soa:loco-arch} depicts a common architecture for developing humanoid locomotion strategies. 
This architecture involves a high-level planner and controller generating the desired center of mass and feet trajectories which are then passed as inputs to a whole-body controller.
This low-level controller then creates the necessary actuation commands for the generation of motion on the robot.
The controllers for humanoid robots rely heavily on the feedback of centroidal states such as Center of Mass (COM), COM velocity, and centroidal momentum (specifically angular momentum plays a key role) in order to realize stable walking motions while establishing contacts with their environments using their flat foot structures.
A reliable floating base estimation deems crucial for estimating the necessary centroidal states of the humanoid robot. 
\cite{masuya2020review} provides an excellent review of the developments done for state estimation of humanoid robots.
In this section, we highlight a few relevant works from their review and augment them with more relevant works in the context of base estimation.

\subsubsection{Proprioceptive Estimation}
The importance of a reliable COM estimation for humanoid locomotion is reported by \cite{xinjilefu2015center} which was developed in the context of DARPA Robotics Challenge (2013).
A reliable estimation of the center of mass allowed the robot to prevent itself from falling throughout the competition.
Many works in recent years have focused specifically on the estimation of the center of mass and angular momentum derivative for human motion analysis and improving humanoid locomotion. (\cite{carpentier2015kinematics, carpentier2016center, piperakis2016non, bailly2019estimating}) based on different estimation techniques combining different sources of proprioceptive measurements.

\cite{piperakis2018nonlinear} highlights the need for an accurate floating base estimation which is important for obtaining the inputs necessary for the centroidal estimation, usually required to be expressed in an inertial frame.
A cascading architecture of floating base estimation and COM estimation is presented by \cite{bae2017new} within the framework of Moving Horizon Estimation (MHE).
This is a flexible optimization-based framework that allows seamless integration of constraints within the estimation problem and consideration of non-Gaussian noise models.
This paradigm of constrained optimization-based estimation in a cascaded fashion could be translated into a stack of tasks estimation approach as done for human dynamics estimation by \cite{tirupachuri2021online}, inspired from its original framework used for humanoid control and also has its similarities with factor-graph optimization based state estimation approaches presented in the previous section.
A decoupled approach for whole-body state estimation was proposed by \cite{xinjilefu2014decoupled} where an inertial-kinematic EKF is used for base state estimation while steady-state Kalman filter is used for joint state estimation exploiting the full-body dynamics for joint velocity estimation.
Simple estimators fusing proprioceptive sensors in a cascading fashion are proposed by \cite{flayols2017experimental} for a decoupled humanoid base estimation, first estimating the base orientation then estimating the base position to avoid having a lack of convergence guarantees that might occur as a  result of improper linearizations within a tightly-coupled EKF framework. 
It is evident that for deploying locomotion controllers which more recently rely on reduced models based on the centroidal dynamics, a reliable floating base estimation is necessary. 
This again brings forth the question on the degree of complexity for design such estimators for humanoid robots.
\looseness=-1

\subsubsection{Exploiting the Flat Foot}
Estimating the rotation of the humanoid robot's flat foot is crucial for achieving accurate estimates of the base pose in order to also account for position displacements from the foot rotations.
\cite{rotella2014state} extend the Observability Constrained Quaternion EKF developed by \cite{bloesch2013state} for humanoid robot's flat foot scenarios in order to consider feet orientations within the state, allowing to better constrain the estimation problem resulting in improved accuracy of the estimates.
\cite{masuya2015dead} combines base orientation estimated by an attitude estimator using IMU measurements with a decoupled, fast estimation of position and velocity estimates using a dead reckoning approach.
The position and velocity estimation is based on a complementary filter on kinematic, force, and inertial measurements. 
The position displacements caused by the rotation of the foot are taken into account using a concept called Anchoring Pivot (AP).
This is a point in proximity to the point of action of the net ground reaction force of the supporting foot.
\cite{benallegue2020lyapunov} expands over the idea of modeling the contact as the anchor point to present a globally exponentially stable non-linear observer for velocity and tilt estimation.
\cite{benallegue2015fusion} and  \cite{mifsud2015estimation} exploit a flexibility deformation in the robot's feet to estimate base kinematics through the fusion of kinematic, inertial, and force-torque sensing capabilities within an EKF framework.
These methods show that when designing either loosely-coupled or tightly-coupled sensor fusion approaches, it is useful in augmenting the base state with the information about the feet of the robot.
This is important because the feet acts as an interface with the environment and through constant exchange of interaction forces with the environment, it affects the centroidal states of the robot thereby relating to a change in the floating base state of the robot.

Recent work by \cite{qin2020novel} tries to extend the invariant EKF proposed by \cite{hartley2020contact} for point foot models towards humanoid flat foot model.
Further, the authors rely on a foot contact probability estimator (similar to \cite{camurri2017probabilistic}) using wrench measurements from a force-torque sensor attached to the foot for dynamically scaling the covariances of the filter corresponding to a constant motion model assumed for the feet position and orientation.

\subsubsection{Combining Proprioception and Exteroception}
An early effort to bridge the gap between the SLAM and humanoid community has been made by \cite{stasse2006real}. 
The authors proposed to combine information about planned Zero Moment Point (ZMP) and robot waist trajectory from a walking pattern generator as observations within a monocular vision-based EKF SLAM framework.
\cite{hornung2010humanoid} combines kinematics and measurements from a laser rangefinder for a global localization of the robot in a known environment demonstrating the potential for autonomous humanoid robot operation within complex indoor environments.
Contact aided kinematic-visual-inertial odometry approaches are presented by \cite{oriolo2016humanoid} and \cite{piperakis2019robust} within an EKF framework and an estimator robust to the presence of outliers in visual measurements has been developed in \cite{piperakis2019outlier}, all of which demonstrate the benefits of heterogeneous sensor fusion for humanoid state estimation. \looseness=-1

A full-stack multi-modal sensor fusion framework called \emph{Pronto} was developed by \cite{fallon2014drift} which combines kinematics, inertial, stereo vision, and LIDAR data in a cascading stage of estimators for joint state estimation, base estimation, and global localization architected in a manner suitable for control, planning, and navigation.
High-frequency state estimates from an EKF combining foot placements with kinematic and inertial measurements are used as feedback to the control loop while a drift-free alignment is provided by visual processing along with global position corrections coming from a particle filter on LIDAR data.
The performance of this estimator has been demonstrated in the experiments conducted for the DARPA Robotics Challenge (2013).
\cite{camurri2020pronto} extends and generalizes the \emph{Pronto} framework for heterogeneous sensor fusion using filtering and smoothing methods applicable for legged and humanoid robots and the performance of this framework is benchmarked across several robots for several real-world experiments.
\cite{raghavan2018study} combine the EKF proposed in \cite{fallon2014drift} with a LIDAR Odometry And Mapping (LOAM) module in a closed-loop architecture to achieve a high accuracy state estimation with reduced drifts. 
An optimization-based smoothing approach for absolute humanoid localization combining inertial and fiducial marker pose measurements is proposed by \cite{fourmy2019absolute}.
The authors introduce a novel IMU deltas Lie group that is well-suited for IMU preintegration.

Although, the humanoids community has been progressing in constructing advanced state estimation strategies and architectures necessary for handling the overall complexity of these systems, the body of literature showcases a limited number of approaches focusing on rigorous nonlinear estimator designs using the theory of Lie groups.
An interesting future direction can combine the theoretical aspects of constrained optimization and hybrid dynamical systems-based invariant filtering approaches through the lens of Lie theory to achieve a full-fledged state estimator for humanoid robots. 
In this thesis, we mainly look into the development of filters on matrix Lie groups exploiting the flat foot nature of the humanoid robot with the high degree of focus relying on proprioceptive sensing, nevertheless extensible towards exteroceptive sensing as well.
\looseness=-1

\section{State of the Art in Human Motion Estimation}
\label{sec:chap:soa:human-motion-estimation}
\begin{figure}[!h]
\centering
\includegraphics[scale=0.5]{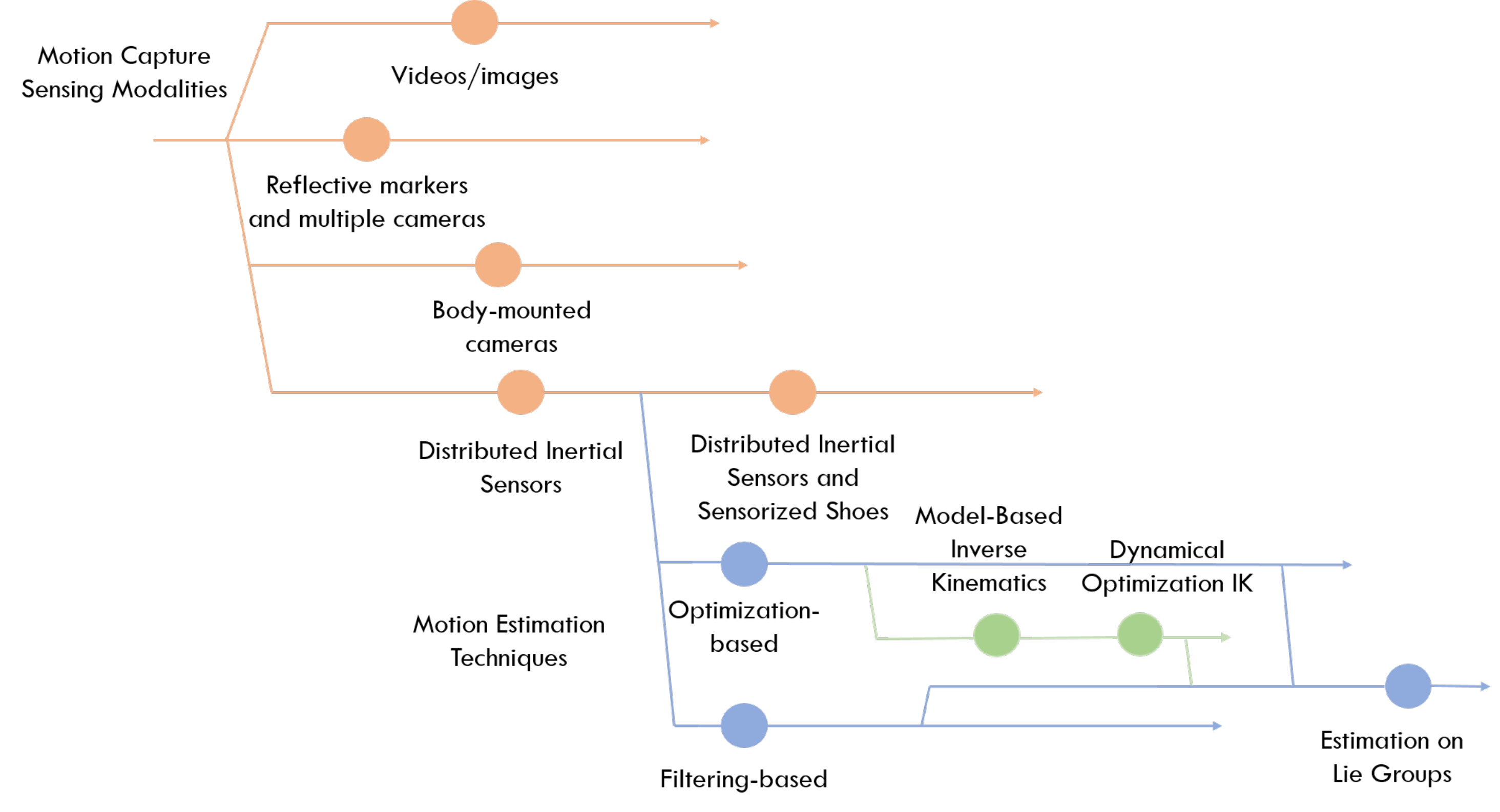}
\caption{Trends in Human Motion Estimation.}
\label{fig:chap:soa:human-motion}
\end{figure}

A reliable human motion estimation can be a crucial technology for accelerating the benefits of human-robot interaction.
The utility of human motion estimation can apply to a wide range of applications like gait analysis (\cite{tao2012gait}), allowing robots to be used as walking-aids for humans (\cite{huang2015posture}), assessment of human ergonomics (\cite{yan2017wearable}), augmenting human capabilities using exoskeletons (\cite{vigne2019state}), and human-robot collaborative work in industrial scenarios (\cite{kyrkjebo2018inertial}), robot collision avoidance for locomotion in shared spaces.
Suiting these purposes, real-time human motion capture is becoming a commonplace technology in various fields ranging from entertainment to medicine.
Further, it is proving to be an essential tool in robotics especially in the context of teleoperating robots in remote environments suited for human morphology (\cite{darvish2019whole}). 
In this section, we will review different sensing modalities used for motion capture and focus on the estimation techniques used for motion estimation through distributed wearable inertial sensors (see Figure \ref{fig:chap:soa:human-motion}). \looseness=-1

\subsection{Sensing Modalities for Human Motion Estimation}
The most common approach for human motion estimation has been using optical motion capture systems where a human subject wears multiple reflective markers on different parts of their body and performs desired motions in a dedicated motion capture environment within the field of view of multiple cameras (\cite{guerra2005optical}).
Such a setup is usually expensive and is not applicable for real-world scenarios. 
This drove the research community to estimate human motion from videos, single or multiple images from a camera (\cite{cao2019openpose, xie2021physics, ganapathi2010real, bogo2016keep}).
For a practical deployment, these approaches require the intervention of an active camera following the human subject or the subject required to be within the camera's field of view.

Although using cameras is the most common approach, other sensing modalities have been explored for the localization of a human within an indoor environment such as Wifi networks, ultra-wideband position sensors, infrared sensing, RFID tags, and so on (\cite{pham2018sensor}), all of which still rely on sensing modalities mounted at different places within a controlled environment. \looseness=-1

Owing to the advances in wearable sensing technologies, there have been several developments in the use of distributed wearable sensors that allows human motion to be estimated in several outdoor and indoor environments without any active intervention.
\cite{wong2015wearable} notices in their survey that the most common wearable technologies have been mobile force plates, pressure insoles for shoes, inertial sensors, EMG sensors, and body-mounted cameras (\cite{sorrentino2020novel, mitjans2021visual}).
The progress in MEMS technology has seen an increased use of IMU-based wearable sensing that enable high-frequency and low-latency data well-suited for real-time motion tracking applications. This technology deems to be a promising alternative over marker-based Optical Motion Capture (OMC) systems and vision-based motion estimation mainly due to its inherent ability to be proprioceptive.
In combination with the other proprioceptive wearable sensing technologies, a full-fledged, reliable motion capture suit may allow the tracking of most relevant signals for embodied human state estimation, i.e. kinematic, dynamic and physiological states. \looseness=-1

\subsection{Human Motion Estimation using Wearable Sensors}
Sensor fusion of measurements from distributed IMU sensors has been used to estimate human motion assuming varying degrees of biomechanical models for the human body with the help of coarsely known location of these sensors in the highly articulated kinematic chain of the model (see \cite{ lopez2016wearable, filippeschi2017survey} for a detailed review). 
\cite{roetenberg2009xsens} presented a commercially available full-body wearable sensor suit that was capable of tracking the 6-DOF human motion enabling full-body motion capture without the need for any external sensor and relying only on wearable sensing technologies.
Several EKF-based and optimization-based methods for body segment-to-sensor calibration and motion estimation based on a biomechanical model exploiting different physical constraints and IMU measurements in the presence/absence of magnetometer measurements have been reported in \cite{zhu2004real, kok2014optimization, qiu2016using, mcgrath2020body, li2021real}.
\cite{miezal2016inertial} compares Extended Kalman Filter (EKF)-based and optimization-based sensor fusion methods for inertial body tracking accounting for robustness to model calibration errors and absence of magnetometer measurements for yaw corrections. The same authors extend their work to ground contact estimation and improved lower-body kinematics estimation in \cite{miezal2017real}.
It was noted that the common source of errors inducing drifts in human motion estimation was related to the segment-to-sensor calibration and the non-rigidity of sensor placement.
Furthermore, considering a Zero Velocity Update information within the measurement model effectively reduces drifts in motion estimation.

An approach to solve the human estimation problem has been to rely on the model-based inverse kinematics.
Inverse Kinematics (IK) is the problem of finding the joint space motions when the target motions or the end-effector motions are given.
This approach has found its application in fields ranging from computer graphics, gaming, and robotics.
Several methods for solving the IK problem exist in the literature such as analytical methods, numerical methods, data-driven methods, and hybrid methods (a detailed survey in \cite{aristidou2018inverse}).
Usually, for highly articulated systems, analytical methods are not easily applicable to find closed-form solutions for the joint motion, hence a numerical approach is preferred.
An instantaneous optimization-based IK on kinematic chains has been successfully applied by \cite{monzani2000using} for reconstructing human motion estimation in the computer graphics community.
Such an approach was also followed by \cite{tagliapietra2018validation} for validating model-based IK using measurements from distributed IMUs as IK targets.
Recently, \cite{rapetti2020model} proposed to apply a model-based IK combining dynamical systems' theory with the IK optimization, thereby calling it \emph{Dynamical Optimization-based IK}, to human motion estimation and demonstrated full-motion reconstruction on a human model with $48$-DoFs and a $66$-DoFs in real-time.
Contrary to the instantaneous methods, this approach converges to a solution that maximizes the likelihood of IMU measurements in multiple steps rather than a single step.

\subsection{Human Motion Estimation using Lie Groups}
A human moving in space is usually modeled as a floating base system which implies the necessity to consider the evolution of the configuration space of joint angles and base pose over a differentiable manifold.
The theory of EKF over Lie groups proposed by \cite{bourmaud2013discrete} is exploited by \cite{cesic2016full} for the problem of reconstructing a whole-body human motion using 3D position measurements of motion-capture markers distributed across the body.
The estimator design exploits a state representation evolving over a complex manifold of the product of $\SO{3}$ and $\SO{2}$ rotations and is compared with an EKF using Euler angles.
The former is shown to be robust even when approaching configurations close to the \emph{gimbal lock} while the performance of the latter degrades in such situations.
Using a similar approach, \cite{joukov2017human} presents a markerless alternative using wearable IMU sensors by explicitly considering the non-Euclidean geometry of the state space while maintaining the required computational efficiency for real-time applications.
An observability analysis for the filtering on Lie groups is also reported by the same authors \cite{joukov2019estimation} for the problem of human motion estimation.
A full-body reconstruction with reduced IMU count through optimization-based estimation on Lie groups is proposed by \cite{von2017sparse} to maintain a history of measurements for achieving an improved position estimation by exploiting the accelerometer measurements effectively into a suitable motion constraint. 
\cite{sy2020estimating, sy2020estimatinga} propose to estimate lower limb kinematics with reduced wearable sensor count using an EKF over Lie groups accounting for physical constraints from the human body. \looseness=-1

The trends in human motion estimation using distributed inertial sensors are seen to be increasingly benefiting from the use of Lie-theoretic tools for estimator design.
Suitable choice of rigorous estimation methods along with low-cost MEMS sensors can enable a democratization of the motion capture technology.
In this thesis, we look into the development of a cascading framework of joint state and floating base estimation of a human based on dynamical optimization based inverse kinematics in series with a filter on matrix Lie groups.
\looseness=-1

\section{Thesis context}
\label{sec:chap:soa:thesis-context}
Having painted a landscape of the research efforts made in the field of humanoid floating base estimation and human motion estimation, before delving into the details for the contributions of the thesis, we would like to reiterate the motivational standpoint for the developments in this dissertation by formulating the following thesis statement,

\textit{Seamless human-humanoid collaboration requires reliable state estimation strategies for both the human and the humanoid robot combining measurements from multiple distributed sensors on the robot and wearable sensing technologies for the human while dealing with the underlying complex geometry related to the free-floating, highly articulated, multi-rigid body nature of these systems. Can the design of floating base estimators exploiting the theory of matrix Lie groups to represent the state, measurements and their associated uncertainties aid in the development of appropriate fusion methods over complex nonlinear spaces to obtain reliable state estimates for these systems which will further be used to augment the control architectures for human-humanoid collaboration?}

\begin{figure}[!h]
\centering
\includegraphics[scale=0.5]{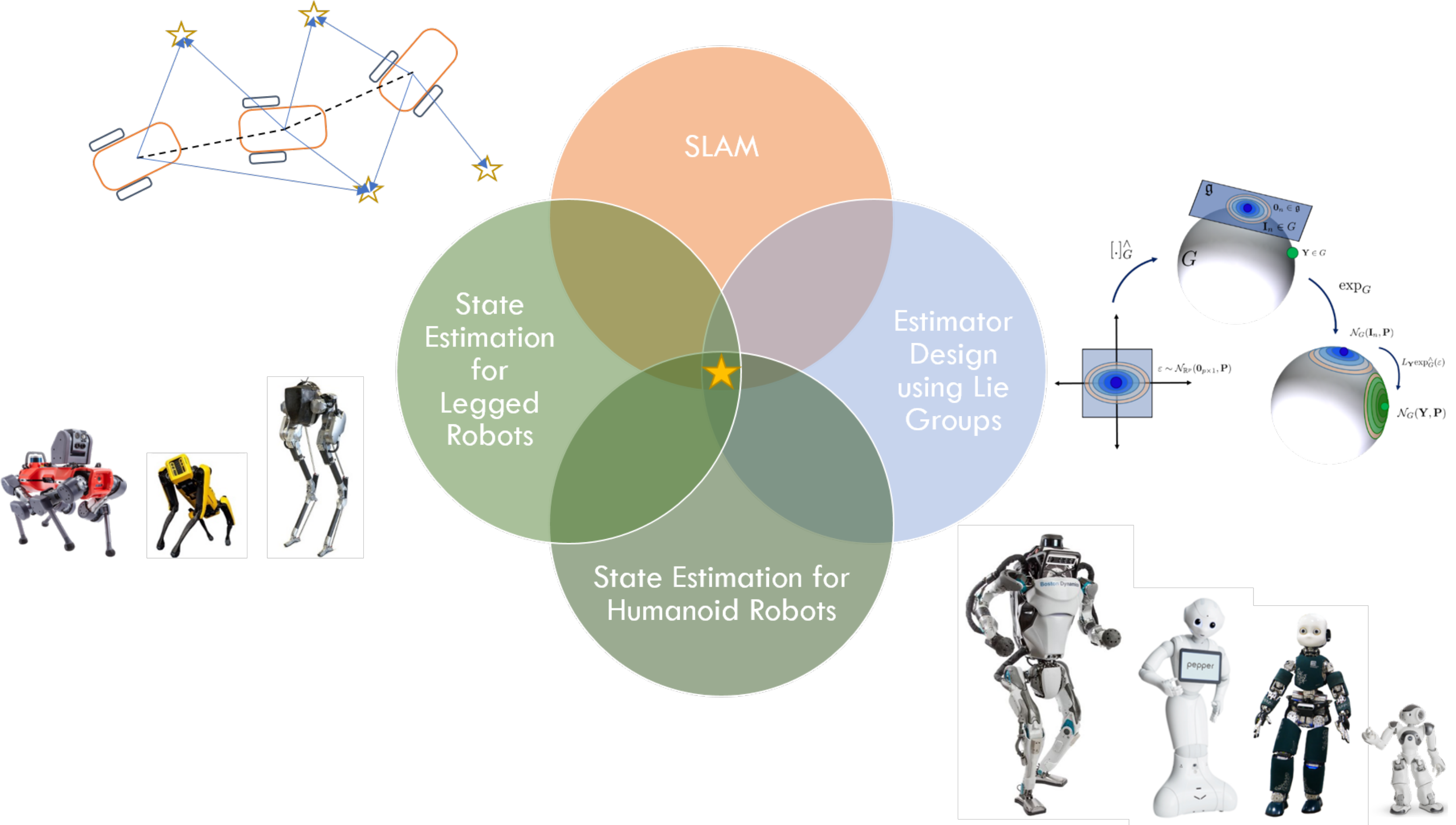}
\caption{The context of humanoid base estimation approaches presented in this thesis (denoted by star) at the intersecting domains of SLAM, Lie-theoretic estimation techniques, and estimation methods in legged robots and humanoids community. \looseness=-1}
\label{fig:chap:soa:intersect}
\end{figure}

\begin{figure}[!h]
\centering
\includegraphics[scale=0.5]{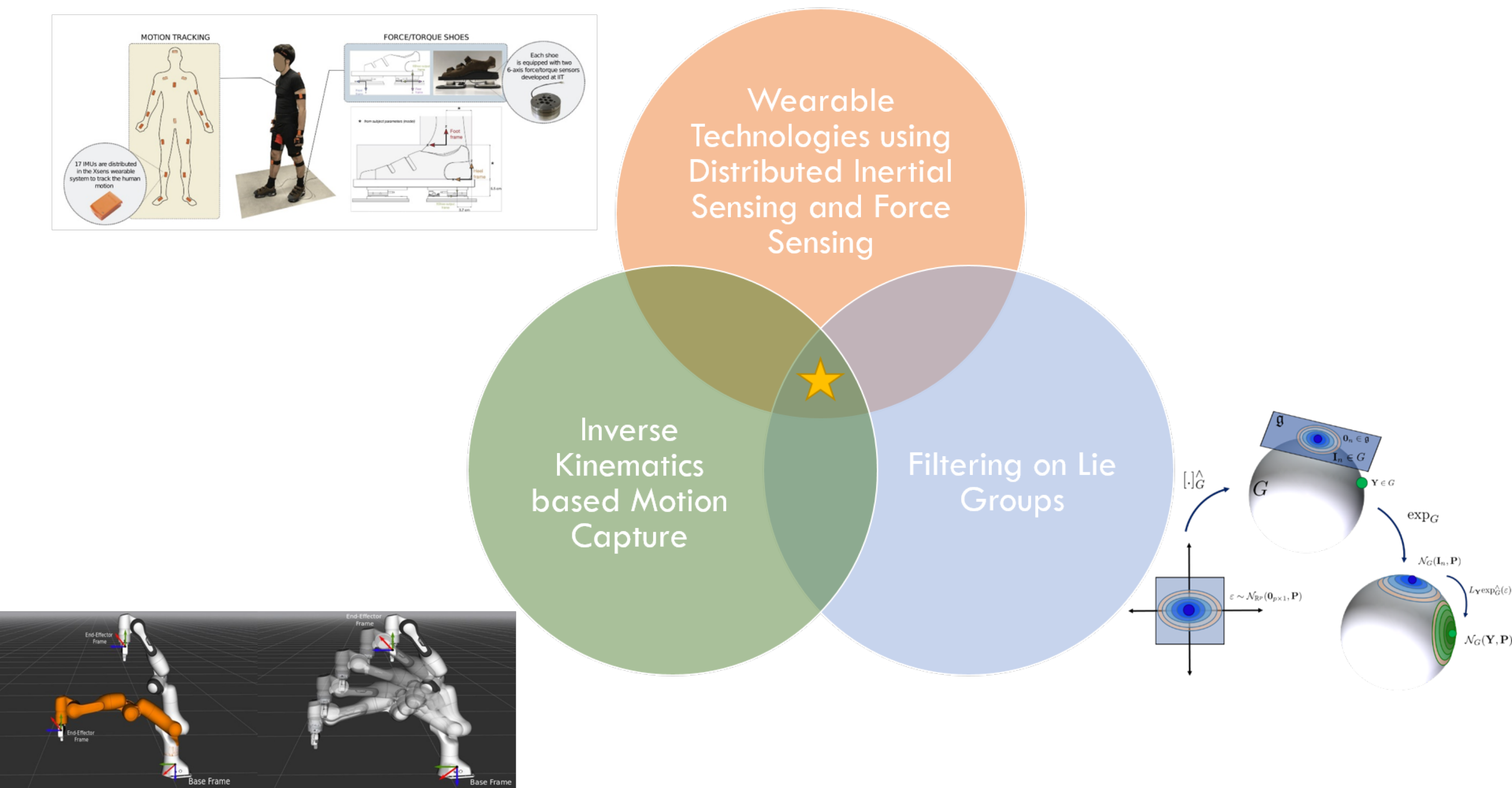}
\caption{The context of proposed approach for Human motion estimation in this thesis (denoted by star) combining wearable sensing technologies, inverse kinematics based motion capture and Lie-theoretic estimator design.}
\label{fig:chap:soa:humanik}
\end{figure}

With that standpoint, we now detail the context of the thesis contributions in the following subsections.
The contributions of Chapter \ref{chap:floating-base-swa} and \ref{chap:floating-base-diligent-kio} fall in the intersecting domain of the SLAM, estimator design on Lie groups, legged robots and humanoid robots community, as seen in Figure \ref{fig:chap:soa:intersect}.
The contributions of Chapter \ref{chap:human-motion} fall in the intersection of wearable sensing technologies, inverse kinematics based motion capture and estimator design on Lie groups, as seen in Figure \ref{fig:chap:soa:humanik}.

\subsection*{Chapter 5: Loosely-Coupled Sensor Fusion for Floating Base Estimation of Humanoid Robots}
\paragraph{Literature gap:} When put forth with the problem of base estimation, a natural question arises on the choice of representation for orientations and how to efficiently fuse orientations from different measurement sources.
A common loosely-coupled fusion approach uses a simple averaging on Euler angles owing to its ease of use.
Although, the issue of singularities with Euler angles is well-known, the literature in humanoids community still lacks in reporting methods that demonstrate simple averaging approaches for quantities like orientations and poses.
Therefore, the averaging of such quantities for a simple fusion to achieve humanoid robot base estimation in a loosely-coupled sensor fusion approach is reported in this chapter. \looseness=-1

\paragraph{Motivation:}
Practitioners often debate only about filtering and smoothing dichotomies when designing estimators for floating base estimation. 
Although such formulations lead to high-accuracy estimation, they usually require tedious formulations and suitable computational budget.
Although state estimation for a humanoid robot is a challenging task due to several complexities, we must sometimes ask whether such inference approaches are really crucial for high-performance humanoid control or if decoupling the estimation of position, orientation and velocity states inferable from multiple sensing modalities significantly impacts the estimation quality.
Moreover, nowadays most sensing capabilities come with dedicated on-board computing  and state-of-the-art estimation algorithms providing us with some motion estimates.
It is worth understanding if we are able to formulate a simple-yet-effective loosely-coupled sensor fusion that allows us to estimate the floating base state by combining multimodal measurements outsourced to existing off-the-shelf methods and how can these complex quantities such as pose and rotations be fused in a meaningful manner to obtain physically consistent estimates.
In this regard, we motivate ourselves with the following question for an estimator design based on a loosely-coupled sensor fusion approach,

\textit{How much can the design of a floating base estimator for a humanoid robot be simplified while maintaining a degree of algorithmic structure that is scalable to multiple sensing modalities, offering performances comparable with established state-of-the-art methods, and simultaneously remaining mathematically rigorous, and intuitive to implement, tune and debug?} 

The formulation of such an estimator design is shown to be feasible using Lie-theoretic tools presented in Chapter \ref{chap:estimation}. \looseness=-1

\paragraph{Context of Contribution:} This chapter presents contact-aided kinematic-inertial odometry for the floating base estimation of humanoid in a loosely-coupled sensor fusion approach.
We take inspiration from the two simple estimators proposed by \cite{flayols2017experimental} as alternatives for the commonly used extended Kalman filtering strategies. 
These simple estimators decouple the nonlinear estimation of base orientation from the linear estimation of base position and velocity in a cascading manner using only proprioceptive measurements such as a base collocated IMU, encoders, and six-axis force-torque sensors attached on the feet.
A similar cascading architecture of decoupled orientation estimation and position-velocity estimation is proposed by \cite{fink2020proprioceptive} for quadrupedal robot base estimation where they use a globally exponentially stable non-linear observer for the attitude estimation. \looseness=-1

Differently from \cite{flayols2017experimental} who chooses to fuse orientation estimates from kinematics and base collocated IMU in the roll-pitch-yaw based Euler angles representations, we propose a fusion in the  $\SO{3}$ space of rotations through the theory of averaging on matrix Lie groups to combine legged odometry and IMU orientation estimates.
This approach allows a singularity-free fusion of estimates and its abstract nature allows a straightforward extension to pose-averaging for the localization of the robot in a known environment.
In comparison with \cite{fink2020proprioceptive}, we only use off-the-shelf attitude estimators such as a locally stable quaternion Extended Kalman filter or a nonlinear observer on $\SO{3}$ proposed by \cite{mahony2008nonlinear}.

For the fusion of velocity estimates of the base link, the joint velocity measurements, gyroscope measurements from the base collocated IMU, and null velocity measurements of the stance foot are combined in a regularized weighted-pseudo inverse based differential inverse kinematics approach using a rigid contact assumption for the foot (\cite{englsberger2018torque}).
This approach is further improved to support measurements from IMUs attached to the feet of the robot along with contact wrench measurements to handle scenarios of foot rotations during locomotion.

\paragraph{Design Choices:}
A natural question that arises regarding the applicability of such an estimator design is "\textit{when to opt for such an estimator?}".
Clearly, this approach can be used, instead of designing dedicated estimators from scratch, when the practitioner wants to use several existing, off-the-shelf methods in a modular fashion and want to combine the estimates from these methods using a reliable fusion approach in an agnostic manner.
A straightfoward extension of the averaging over Lie groups to the Bayesian fusion over Lie groups as seen in Section \ref{sec:chap:loosely-couple-averaging} can allow to handle  uncertainties in the estimates effectively, thus bringing this approach on par with the tightly-coupled sensor fusion based filtering and smoothing approaches.
The algorithmic scalability from rotation-averaging to pose-averaging paves the way to multimodal sensor fusion where pose estimates from exteroceptive localization modules such as visual odometry, point-cloud registration based odometry can be combined in a modular fashion within the proposed method.
Moreover, application of domain-specific knowledge such as corrections from anchoring pivot point of the foot can be applied directly into the estimator formulation without having to go through tedious computations.

\subsection*{Chapter 6: Floating Base Estimation for Humanoid Robots using Filtering on Matrix Lie Groups}
\paragraph{Literature gap:} Accounting for the constraints imposed by the rotations of the foot within the estimator design leads to the \emph{flat-foot filter} that improves the accuracy of floating base estimation of humanoid robot.
This chapter tries to fill the gap in which the design of the flat-foot filter can consider the evolution of state and observations over distinct matrix Lie groups, contrary to the commonly used vector-space representations. \looseness=-1

\paragraph{Motivation:}
Comparing the current standard approach of quaternion-based Observability Constrained EKF (OCEKF) used for proprioceptive humanoid base estimation (\cite{rotella2014state}) with the invariant Lie group filtering methods employed recently for legged robot state estimation (\cite{hartley2020contact}), it is evident that there is room for improvement in filtering-based humanoid state estimation.
Eventhough several approaches different from filtering have been proposed already in the past for humanoid base estimation as seen in Section \ref{subsec:chap:soa:estimation-humanoid-robot}, filtering based methods still seems to be preferable in the community for it's relative ease of use and implementation. 
With such requirements in mind, we motivate ourselves with the following question for a tightly-coupled proprioceptive estimator design, \looseness=-1

\textit{Given the measurement set from contacts, kinematics and base-link collocated IMU, is it possible to formulate proprioceptive estimator with a degree of accuracy suited for humanoid locomotion while considering both the state and the measurements evolving over complex nonlinear spaces such as distinct matrix Lie groups? Can such an estimator be implemented  as a computationally-effective generalization of the commonly used flat foot filter design based on the quaternion EKF while remaining extensible to handle also exteroceptive measurements?}

In this chapter, we formulate a few variations of such a generalizable filter applied for contact-aided kinematic-inertial using the theory of filtering on matrix Lie groups.

\paragraph{Context of Contribution:} A commonly used approach for proprioceptive floating base estimation in both the legged robots' and the humanoid robots' community uses an extended Kalman filter based on a strap-down IMU-based system dynamics and relative forward kinematics-based measurement models.
In this chapter, we investigate this approach for tightly-coupled sensor fusion for floating base estimation. \looseness=-1

This work is closely related to those of \cite{bloesch2013state}, \cite{rotella2014state}, \cite{hartley2020contact} and \cite{qin2020novel} who present similar methods each improving over the other.
As already noted, \cite{bloesch2013state} used an observability-constrained quaternion EKF for a consistent fusion of IMU and encoder measurements for the base estimation of a quadrupedal robot considering the augmentation of the base state with feet positions through point foot contact models.
This was then extended by \cite{rotella2014state} for the flat-foot contact scenario of humanoid robots who chose to incorporate also the feet rotations within the state to better constrain the estimation problem. 
This resulted in improved accuracy for humanoid base estimation.
Recently, \cite{hartley2020contact} proposed an invariant extended Kalman filtering approach using the theory of matrix Lie groups as an alternative to the consistent estimator proposed by \cite{bloesch2013state} for robots with point foot contact models.
This estimator promises strong convergence properties along with inherently consistent estimation owing to the properties of the Lie group-based error evolution and group-affine system dynamics.
The matrix representation chosen by \cite{hartley2020contact} couples the foot positions with the base orientation in a semi-direct product along with base position and linear velocity in a $\SEk{k+2}{3}$ matrix Lie group, where $k$ may denote the number of feet or even the number of landmarks in the case of exteroceptive state estimation.
Such a state representation along with the group-affine dynamics property allows for autonomous error propagation and updates resulting in an extended Kalman filter in the form of a locally asymptotically stable observer.
The work of \cite{hartley2020contact} has been extended to the flat-foot scenario of the humanoid robots to consider the foot rotations within the state in \cite{qin2020novel}.
They do so by considering two axes of the overall foot rotation and coupling it with the base orientation in the semi-direct product.

Our approach is the most similar to \cite{qin2020novel} which was based on the important work by \cite{hartley2020contact}. 
We also develop a flat-foot filter for the humanoid robot by incorporating foot rotations within the state while exploiting matrix Lie group representations.
Differently from \cite{qin2020novel}, we choose a matrix Lie group representation that completely decouples the foot pose from the base state within the state representation and they become related only through the measurement model.
We consider a discrete Lie group Extended Kalman filter for which both the state and the observations evolve over distinct matrix Lie groups, contrary to the case of only vector observations in all the previous works.
This is done at the cost of the stability guarantees that the invariant filtering approach offers straightforwardly.
Experimental validation of the proposed estimator is performed in comparison with the SWA estimator approach proposed in  Chapter \ref{chap:floating-base-swa} and the state-of-the-art methods of \cite{rotella2014state} and \cite{hartley2020contact} for walking and center-of-mass swinging motion experiments conducted on the iCub humanoid robot.
Further, different variants of the filters based on the choice of error and time-representation for the system dynamics are also investigated.
A straightforward extension of the proprioceptive estimator for absolute humanoid localization is also discussed. \looseness=-1

\paragraph{Design choices:}
If a fully coupled inference is crucial for a contact-aided kinematic-inertial odometry problem, one may choose from filter-based or smoothing-based tightly-coupled sensor fusion both of which solve an unconstrained non-linear least squares problem, as seen in Sections \ref{subsec:chap:soa:filter-vs-optim} and \ref{subsec:chap:soa:estimation-legged-robot}. 
However, it maybe possible that filtering based approaches are sufficient enough to obtain reliable estimates with an appropriate choice of state and observation representation, given the \emph{exact linearization} tools offered by Lie-theoretic filtering methodologies,  while providing established theoretical guarantees and remaining relatively simple to implement.
Such approaches reduces the computational burden on the hardware while remaining relatively easier to deploy and debug.
The argument of dealing with constraints within a smoothing based optimization problem in a straightforward manner can also be tackled by extending the EKF towards constrained filtering approaches, which might be suitable for more rigorous handling of contact events.
Further, when dealing with high-frequency measurements which is often the case for proprioceptive estimation relying on IMUs, joint sensors and force-torque sensors, the temporal budget for estimator computations is limited which might often be a bottleneck for smoothing-based estimation approaches.
Besides that, the running frequency of the estimator must be such that it can be used to close the loop with the locomotion controllers at high rates and low latency.
This rationale drives us in the favor of developing an EKF-based estimator.
However, when extending towards a visual information aided kinematic-inertial odometry to consider landmark information within the estimation, the filtering costs to maintain the map of landmarks might grow rapidly, in which case one may opt for incremental smoothing methods which are computational efficient promising alternative to the filtering method proposed here for long-term, drift-free estimation. 
Cascading architectures of concurrent filtering and smoothing approaches can also be formulated, however in this thesis we focus mainly on filtering methods. \looseness=-1

\subsection*{Chapter 7: Human Motion Estimation using Wearable Sensing Technologies}
\paragraph{Literature gap:} A computationally effective whole-body joint state and floating base estimation for a highly articulated system such as a human, using distributed inertial sensing and ground reaction forces seems to be lacking in the human motion estimation literature. 
Most of the existing works in the literature are either not wholly proprioceptive or demonstrate only reduced motion reconstruction, such as pertaining only to limb kinematics.
We try to formulate a fully proprioceptive approach that allows for a whole-body motion estimation using only wearable sensors that can be used in a unifying manner on humans and humanoid robots. \looseness=-1

\paragraph{Motivation:}
Physically meaningful human motion reconstruction can be achieved by formulating a contact-aware whole body kinematics estimation where the information about the contacts constrains the lower limb kinematics effectively and also aids in the propagation of the 6D pose of floating base link.
With the consideration of incorporating the contact information effectively within a motion-tracking framework based on proprioceptive sensing, we motivate ourselves with the following question that enables an extension of motion tracking to also physically consistent pose estimation,

\textit{Is it possible to recover whole body kinematic motion of a human solely from densely equipped IMUs and contact wrench measurements through a reliable joint state and 6D floating base pose estimation? In particular, can a reliable reconstruction of also the linear quantities such as base position and linear velocity be obtained by sequentially cascading an Inverse Kinematics optimization approach with Lie group filtering based contact-aided kinematic inertial odometry?}

\paragraph{Context of Contribution:} In this chapter, we combine dynamical optimization-based inverse kinematics, center-of-pressure-based contact detection, and invariant filtering on Lie groups for floating base estimation to achieve a whole-body human motion estimation using wearable sensing technologies such as a distributed IMUs based motion capture suit and sensorized shoes mounted with force-torque sensors.
This work comes as an extension of the model-based inverse kinematics proposed by \cite{rapetti2020model} for real-time human motion estimation to account for a contact-aided floating base estimation.

Contrary to the EKF and instantaneous optimization approaches proposed in \cite{miezal2017real, tagliapietra2018validation, kok2014optimization}, here we use a dynamical system based inverse kinematics optimization to recover the joint states from the distributed inertial sensors.
This can be seen as a control loop converging towards the desired solution minimizing the error between the expected model-based kinematics and the target measurements obtained from the several IMUs.
The recovered joint states along with the contact information are then used within an EKF over Lie groups for floating base estimation.
The contact detection strategy relies on an approximation of the foot with a simplified rectangular geometry and considering the vertices as candidate contact points whose contact status is inferred based on the local center of pressure information (\cite{dafarra2020predictive}).
For the floating base estimation, instead of directly translating the IMU measurements into $6$-DoFs floating base state as done by \cite{von2017sparse, sy2020estimating}, we borrow the techniques of the robotics community to use contact-aided kinematic odometry.
We use constant motion models for the system dynamics and propose to combine right-invariant, left-invariant, and non-invariant observations within the same filter structure.
The performance of this approach is first demonstrated for a walking experiment conducted on the iCub humanoid robot and further applied to human walking experiments.

\paragraph{Design choices:}
The cascading architecture of dynamical system based IK optimization and filtering over Lie groups aims primarily to reduce the computational burden.
The dynamical system based IK optimization is shown to be a computationally efficient approach for the reconstruction of motion for highly articulated systems with a large number of degrees of freedom in comparison with other optimization based methods while accounting for densely equipped IMU sensors.
The dynamical system based IK optimization stands out in comparison with other instantaneous optimization methods. 
The latter although converges fast towards a desired solution for most robotic applications, it suffers to achieve high rates while finding solutions for an overly complex human model during time-critical applications.
Instead, the dynamical system based IK reformulates IK as a control problem and the model configuration is controlled to dynamically converge to sensor measurements over time.
Thanks to this dynamical nature of the model configuration, it is not required for the solver to perform repeated iterations at one time step to find a suitable solution.
Since the error dynamics of the control problem is guaranteed to decay, we obtain a suitable method for solving whole-body inverse kinematics for time-critical motion tracking of complex models.
Beyond motion tracking and reconstruction, this method can also benefit motion imitation applications especially aimed at retargeting human motion on the robot.
However, its advantages for motion tracking is particularly important for achieving reliable floating base estimation of the human that rely on contact-aided kinematics to propagate the base state through time.
If the orientation measurements from multiple IMUs are expressed in a common calibrated frame and other IMU measurements can still be expressed in local frames, this implies that the joint state estimation can be made invariant of the floating base pose. 
The joint state estimates can then be used for floating base estimation only through relative kinematics which along with contact information is sufficient to reconstruct the 6D pose of the human.
Further, the joint state estimation from the IK is decoupled from the base state estimation also with the perspective to include the advantages of invariant filtering and extending the base pose estimation with different sensing modalities.
This approach clearly relies on a known model for reconstructing the human motion and the estimator performance relies on joint state reconstruction based on densely equipped IMUs while being most suitable only with the availability of ground contact wrench measurements.

\section*{Basic Notation}
This section describes the basic notation which is commonly used throughout this thesis.

\begin{itemize}
    \item Matrices are denoted in bold capital letters such as $\X, \Y$, and vectors are denoted in bold small letters $\mathbf{u}, \mathbf{v}$.
    \item Scalars and function names  are denoted by small letters, $p, q$ and $f(.), h(.)$ respectively. 
    \item Greek symbols $\omega, \theta$ may represent either a vector or a scalar depending on the context.
    \item $\I{n}$ denotes an identity matrix of dimensions $n \times n$.  
    \item $\Zero{n}$ denotes a zero matrix of dimensions $n \times n$, while $\Zeros{n}{m}$ denotes a zero matrix with $n$ rows and $m$ columns.
    \item Coordinate frames are denoted by capital letters $A, B, F$.
    \item The elements of a Lie group are usually denoted as $\X, \Y, \Z$, while the elements of a Lie algebra are denoted as $\A, \B, \Cbold$ and corresponding vectors as $\mathbf{a}, \mathbf{b}, \mathbf{c}$.
\end{itemize} 

The next part of the thesis will describe the contributions of this thesis in detail, with the help of the mathematical background and the context built thus far.

\part{Thesis Contributions}
\chapter{Loosely-Coupled Sensor Fusion for Floating Base Estimation of Humanoid Robots}
\label{chap:floating-base-swa}

\begin{figure}
\centering
\includegraphics[scale=0.5, width=\textwidth]{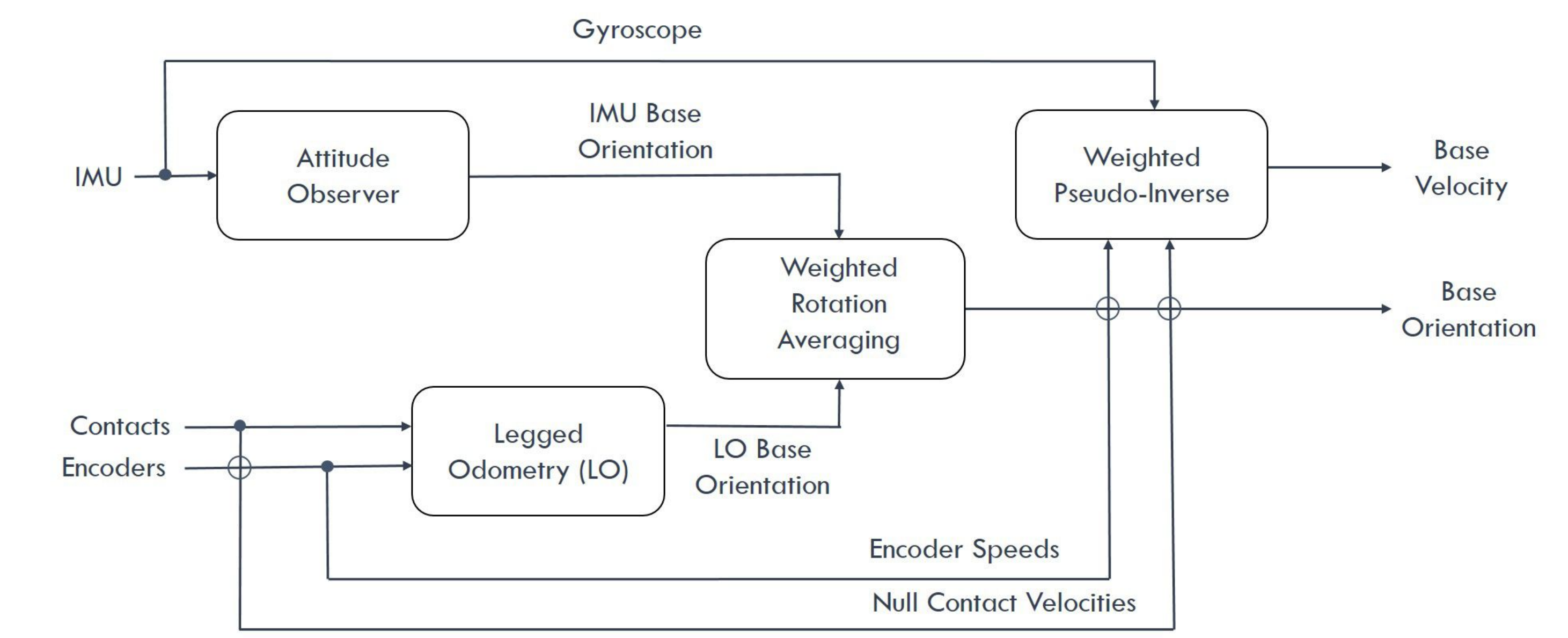}
\caption{A block diagram representation of Simple Weighted Averaging Estimator.}
\label{fig:chap:loosely-coupled:swa-blk-diag}
\end{figure}

The simplest way to design a floating base estimator for a humanoid robot is to fuse available estimates from multiple sources of information.
In a loosely-coupled sensor fusion approach, the measurements from different sensing modalities are processed separately to infer the state to be estimated which are then combined to obtain a fused estimate.
In this chapter, we develop a loosely-coupled sensor fusion approach for the floating base estimation of a humanoid robot.
A method decoupling the non-linear quantity of base orientation with the linear quantity of base velocity is proposed, using a singularity-free fusion for orientations, cascaded by a linear estimation of velocity of the base link, as shown in Figure \ref{fig:chap:loosely-coupled:swa-blk-diag}.
We employ the technique of averaging on matrix Lie groups to average rotations estimated by legged odometry and an IMU for a floating base orientation estimation along with a decoupled base velocity estimation using a least-squares approach in Section \ref{sec:chap:loosely-coupled-swa}.
Experimental results validated on a humanoid robot platform is discussed in Section \ref{sec:chap:loosely-coupled:experiments}.
Further, Section \ref{sec:chap:loosely-coupled-localization} presents a pose averaging approach for demonstrating localization of the robot in a known environment. \looseness=-1

\section{SWA: Simple Weighted Averaging Estimator}
\label{sec:chap:loosely-coupled-swa}
In this section, we describe the Simple Weighted Averaging (SWA) estimator used for a loosely-coupled floating base estimation of a humanoid robot.
The rotation estimates from an IMU and the legged odometry module are fused using the concept of averaging on Lie groups.
While, a fused velocity estimate is obtained by combining gyroscope measurements and the legged odometry based velocity estimate obtained using the encoder measurements.
The methods closest to this estimator are described by \cite{flayols2017experimental, fink2020proprioceptive, masuya2015dead} where estimates from different proprioceptive sources of information are combined in a cascading manner.
The main distinction from the other methods is the consideration of averaging for the rotation estimates in the space of $\SO{3}$.
This method remains an easy-to-implement and simple method to obtain a fused estimate on a robot equipped with an IMU on the base link and encoders. \looseness=-1

\subsection{Legged Odometry}
\label{sec:chap:loosely-coupled:swa-legged-odom}
We can compute the floating base pose and velocity of the robot with respect to an inertial frame assuming that at least one link of the robot is in rigid contact with the environment at any time instant.
This is done by propagating the relative forward kinematics between the link currently in contact with the environment and the link that will establish a new contact with the environment.  
Such computations require a combination of information coming from the joint encoders measuring the joint angles $\jointPos$ and information from contact switching deciding the link in contact with the environment.

\begin{figure}[!h]
\centering
\includegraphics[scale=0.37]{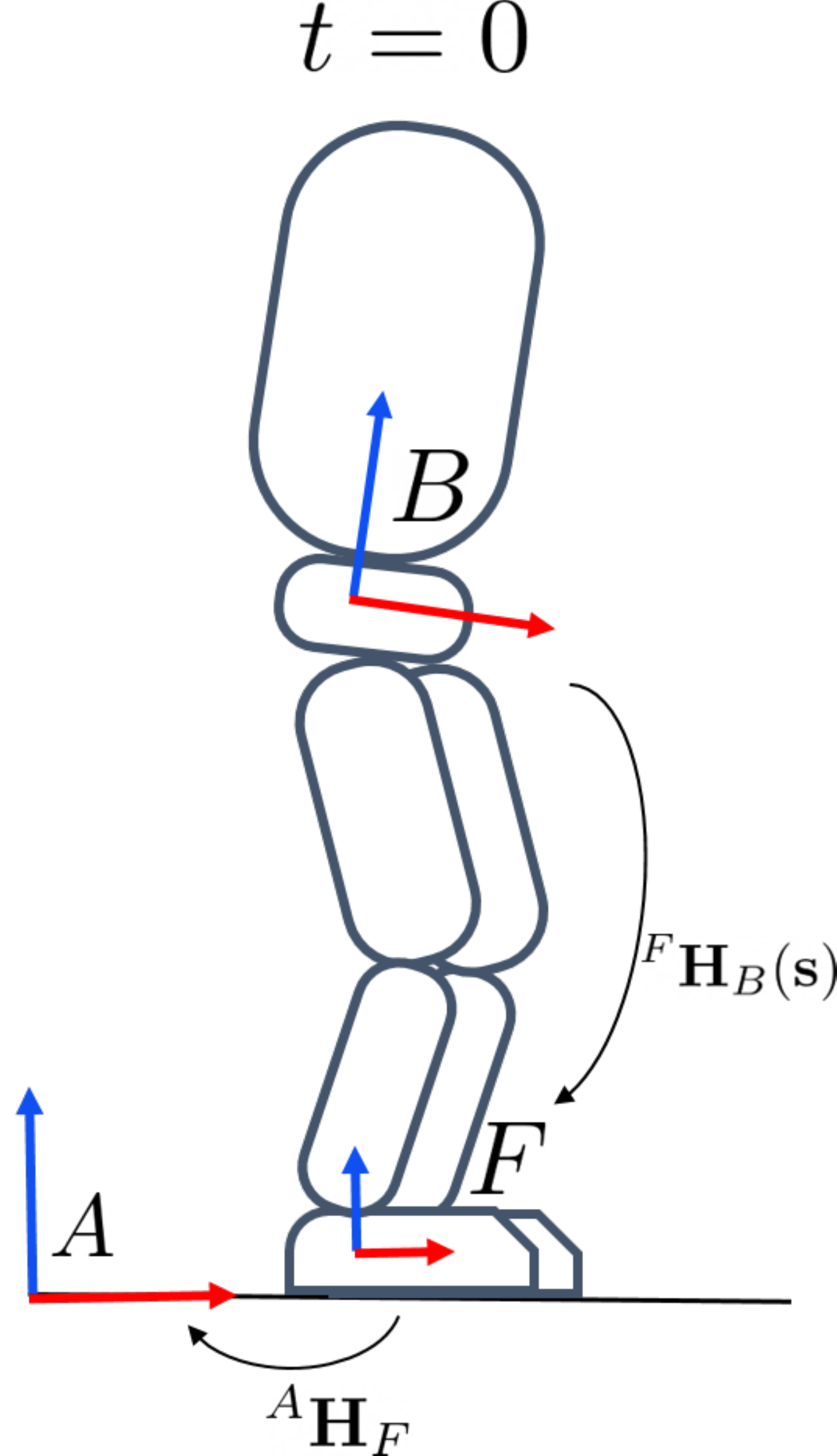}
\caption{Initial known coordinate transformations for initializing legged odometry computations assuming a fixed transform between inertial frame $A$ and fixed frame $F$.}
\label{fig:chap:loosely-coupled:legged-odom-t-zero}
\end{figure}

\begin{figure}[!t]
\centering
\includegraphics[scale=0.5]{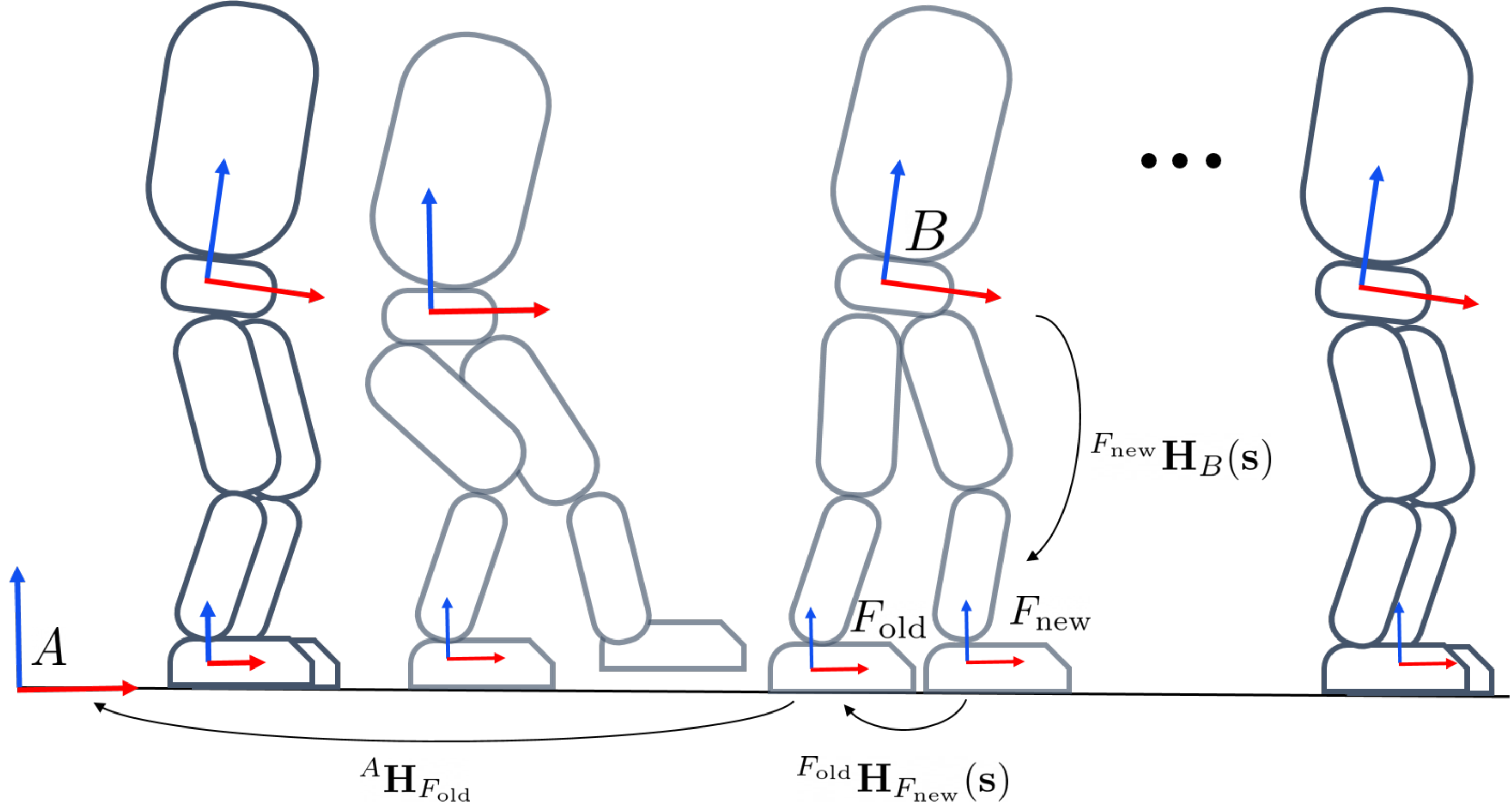}
\caption{A visual description of Legged Odometry. As the robot makes a new contact at $F_\text{new}$, the relative transforms with the base link $B$ and old contact $F_\text{old}$ computed through relative forward kinematics as a function of joint positions $\jointPos$ are used to update the base pose of the robot.}
\label{fig:chap:loosely-coupled:legged-odom}
\end{figure}

As shown in Figure \ref{fig:chap:loosely-coupled:legged-odom-t-zero}, given the pose of a fixed link $F$ w.r.t. the inertial frame $A$ described by the homogeneous transform   $\Transform{A}{F}$  at the first time instant, $t= 0$, the base pose at $t = 0 $ can be computed as, \looseness=-1
$$ \Transform{A}{B} =\ \Transform{A}{F}\ \Transform{F}{B}(\jointPos), $$
where,  $\Transform{F}{B}(\jointPos)$ is the homogeneous transform between the fixed link and the base link obtained through relative forward kinematics.  
The pose of the fixed link $F$ is assumed to be constant until a contact switch is triggered. 
As soon as a new link $F_{\text{new}}$ makes a rigid contact with the environment,  the old fixed link $F_{\text{old}}$ is discarded and $F$ is changed to be the new link $F_{\text{new}}$. 
The base pose is updated accordingly as depicted in Figure \ref{fig:chap:loosely-coupled:legged-odom},
$$  
\Transform{A}{B} = \ \Transform{A}{F_{\text{old}}} \  \Transform{F_{\text{old}}}{F_{\text{new}}}(\jointPos) \  \Transform{F_{\text{new}}}{B}(\jointPos). $$

The decision about the contact switching can be obtained through a contact detection strategy, such as, a Schmitt Trigger thresholding on the contact normal forces. 
The contact normal forces, in turn, can be obtained from force-sensing technologies attached to the end-effector or estimated using the external wrench estimation method exploiting the whole-body dynamics of the robot. 
The Schmitt trigger thresholding, is based on two parameters for force thresholds to infer events of contact making and breaking while two other parameters for timing thresholds are chosen to judge a stable switching of rising and falling edges of the force signals.
If the contact normal force of the foot exceeds the make threshold over a rising period of time, then a contact switch is made from off to on.
A contact is said to be lost if the force falls below the break threshold over a falling period of time.
This choice of parameters allow for a contact switching robust to noisy force measurements obtained from the estimated contact wrenches.
\looseness=-1

The floating base velocity, $ \twistMixedTriv{A}{B} $, is computed considering the holonomic constraint of a link in contact with the environment.
When a link is rigidly attached to the environment, the velocity of the link is zero. 
As a consequence, the floating system velocity $\nu^{B/B[A]}$ can be computed through the free-floating Jacobian $\freeFloatingJacobian{A}{F/F[A]}$ of the link $F$ in contact,
\begin{equation}
\label{eq:chap:loosely-coupled:null-feet-vel}
    \twistMixedTriv{A}{F}= \freeFloatingJacobian{A}{F/F[A]} \ \nu^{B/B[A]} = \Zeros{6}{1}.
\end{equation}
By expressing the Jacobian $\freeFloatingJacobian{A}{F/F[A]}$, in terms of base velocity and joint velocities as $\J_{F_B}$  and $\J_{F_\jointPos}$ respectively, the floating base velocity $ \twistMixedTriv{A}{B}$ can be computed as,
\begin{align}
    \label{eq:chap:loosely-coupled:base-vel-rigid}
    & \begin{bmatrix}
    \J_{F_{B}} & \J_{F_{\jointPos}}
    \end{bmatrix} 
    \begin{bmatrix}
       \ \twistMixedTriv{A}{B} \\ \jointVel
    \end{bmatrix} = \twistMixedTriv{A}{F}, \\
    & \J_{F_{B}} \ \twistMixedTriv{A}{B} + \J_{F_{\jointPos}}\ \jointVel = \Zeros{6}{1}, \\
    & \twistMixedTriv{A}{B} = - (\J_{F_{B}})^{-1} \J_{F_{\jointPos}}\ \jointVel.
\end{align}
It can noted that $\J_{F_{B}}$ is an adjoint matrix operator of $\SE{3}$ which acts as a coordinate transformation for the twists.
Hence, it is a $6 \times 6$ square matrix and is always invertible. 
While, $ \J_{F_{\jointPos}}$ is a relative Jacobian matrix that maps joint subspace velocities to the task space velocities, in this case, null foot velocities.
The structure of these matrices depends on the chosen trivialization for the base link and foot link twists.
Here, we have chosen a mixed-trivialized representation for the velocities.

\subsection{Fusion with a Base Collocated IMU}
Assuming that the robot is equipped with an IMU collocated on the base link of the robot, we can exploit the orientation estimates obtained from IMU measurements along with the gyroscope measurements to get a fused estimate in combination with the legged odometry estimates.

The orientation estimate is obtained using an off-the-shelf attitude estimator, nonlinear observer proposed by \cite{mahony2008nonlinear} or a quaternion Extended Kalman Filter \cite{kok2017using}.
Such attitude estimators typically fuse accelerometer, gyroscope and magnetometer measurements to estimate the absolute orientation of the body equipped with the IMU.
The gyroscope and accelerometer measurements together allows a reliable estimation of roll and pitch angles while magnetometer measurements are relied upon to make the heading or the yaw angle observable.
The magnetometer measurements mainly rely on the measure of Earth's magnetic field to infer the heading or the yaw angle and are prone to be affected by ambient magnetic disturbances, thus inducing vulnerability to yaw drifts.
This makes the yaw estimates to suffer from low robustness with respect to reproducibility based on local environmental conditions, and hence for this reason we neglect the magnetometer measurements from the IMU.
When magnetometer measurements are not considered, the orientation estimates from an IMU have an unobservable direction about the gravity vector described by the heading/yaw of the IMU.
To resolve this unobservability issue in a trivial manner, we project the heading estimated by the legged odometry as the heading estimated by the IMU through a Roll-Pitch-Yaw Euler angle decomposition of both the estimates.
This is done so assuming an alignment of the inertial frames assumed by IMU and legged odometry estimator. \looseness=-1

For the fusion of orientation estimates, a deterministic weighted averaging over $\SO{3}$ is proposed as a fusion method for rotations (\cite{hartley2013rotation}).
An averaged rotation is obtained from the definition of Karcher's mean in Section \ref{sec:chap:loosely-couple-averaging} and Manton's algorithm (Algorithm \ref{algo:chap:loosely-coupled:manton-algo}) as the corresponding averaging algorithm.
As already seen in Eq. \eqref{eq:chap:loosely-coupled:distance-lie-group}, a bi-invariant distance metric based on the geodesic length between two points on the group of rotations,
$$
d^2(\Rot{}{}_1, \Rot{}{}_2) = \norm{\glogvee{\SO{3}}(\Rot{}{}_1^{-1} \Rot{}{}_2)}^2 = \norm{\glogvee{\SO{3}}(\Rot{}{}_1^{T} \Rot{}{}_2)}^2, \forall\  \Rot{}{}_1, \Rot{}{}_2 \in \SO{3},
$$
is chosen as the metric with respect to which the averaging is performed.
It is straightforward to show that $d^2(\Rot{}{}_1, \Rot{}{}_2)$ is a bi-invariant metric on $\SO{3}$. Consider elements of the rotation group, $\Rot{}{}_1, \Rot{}{}_2, \Rot{}{} \in \SO{3}$.  
We can make use of the fact that the error between two rotations $\Rot{}{}_1$ and $\Rot{}{}_2$ given by $\Rot{}{}_1^T \Rot{}{}_2$, $\Rot{}{}_1 \Rot{}{}_2^T$, $\Rot{}{}_2^T \Rot{}{}_1$ and $\Rot{}{}_2 \Rot{}{}_1^T$ all represent a rotation through same angle to show that $d^2(\Rot{}{}_1, \Rot{}{}_2)$ is right-invariant,
\begin{align*}
    \begin{split}
        d^2(\Rot{}{}_1 \Rot{}{}, \Rot{}{}_2 \Rot{}{}) &= \norm{\glogvee{\SO{3}}( (\Rot{}{}_1 \Rot{}{})^{T} \Rot{}{}_2\Rot{}{})}^2 \\
        &= \norm{\glogvee{\SO{3}}( \Rot{}{}_1 \Rot{}{} (\Rot{}{}_2\Rot{}{})^{T})}^2 \\
        &= \norm{\glogvee{\SO{3}}( \Rot{}{}_1 \Rot{}{} \Rot{}{}^T\Rot{}{}_2^{T})}^2 \\
        &= \norm{\glogvee{\SO{3}}( \Rot{}{}_1 \Rot{}{}_2^{T})}^2 \\
        &= \norm{\glogvee{\SO{3}}( \Rot{}{}_1^T \Rot{}{}_2)}^2 = d^2(\Rot{}{}_1, \Rot{}{}_2).
    \end{split}
\end{align*}
Similarly, we can show that $d^2(\Rot{}{}_1, \Rot{}{}_2)$ is left-invariant as follows,
\begin{align*}
    \begin{split}
        d^2(\Rot{}{} \Rot{}{}_1, \Rot{}{} \Rot{}{}_2) &= \norm{\glogvee{\SO{3}}( (\Rot{}{} \Rot{}{}_1 )^{T} \Rot{}{} \Rot{}{}_2)}^2 \\
        &= \norm{\glogvee{\SO{3}}( \Rot{}{}_1^T \Rot{}{}^T \Rot{}{} \Rot{}{}_2)}^2 \\
        &= \norm{\glogvee{\SO{3}}( \Rot{}{}_1^T \Rot{}{}_2)}^2 = d^2(\Rot{}{}_1, \Rot{}{}_2).
    \end{split}
\end{align*}

 The base link rotations estimated individually from the legged odometry and the IMU, ${\Rot{A}{B}}^\text{LO}$ and ${\Rot{A}{B}}^\text{IMU}$ respectively,  are considered as inputs.
 Normalized weights $w^\text{LO}$ and $w^\text{IMU}$ are assumed for the legged odometry and IMU estimate respectively for a weighted averaging in the tangent space of $\SO{3}$.
 These weights can be tuned manually with the intuition of a linear interpolation between the legged odometry estimate and the IMU estimate or can be directly obtained from the inverse of estimated covariances obtained from encoder and IMU measurements.
 The optimization problem is initialized with the rotation estimated by the legged odometry and the Riemannian gradient descent is performed considering a step size $\Delta t$ and an error tolerance $\tau$ as a termination condition to obtain the fused estimate $\Rhat{A}{B}$.
The fusion of rotation based on Manton's convergent algorithm is described in Algorithm \ref{algo:chap:loosely-coupled:rotation-averaging}.
In this case, where only two rotation quantities are averaged, this can simply be viewed as a weighted linear interpolation performed in the tangent space of rotations estimated by the legged odometry and the IMU measurements. 

\begin{algorithm}[!h]
\small
  \caption{Weighted Rotation Averaging \looseness=-1}
  \label{algo:chap:loosely-coupled:rotation-averaging}
  \begin{algorithmic}
    \State \textbf{Input:} ${\Rot{A}{B}}^\text{LO}, {\Rot{A}{B}}^\text{IMU} \in \SO{3}$ \\ 
    \quad  $w^\text{LO}, w^\text{IMU} \in \R, \quad w^\text{LO} + w^\text{IMU} = 1$ \\
    \textbf{Output:} $\Rhat{A}{B} \in \SO{3}$ \\
    \State \textbf{Initialize:}\\
    $\Rhat{A}{B} = {\Rot{A}{B}}^\text{LO}$ \\
    Desired tolerance $\tau > 0$ \\
    Step Size $\Delta t > 0$\\
    
    \State \textbf{Iterate until convergence:} \\
    \quad $\mathbf{r} = \sum_{i = 1}^2 w^i \glogvee{\SO{3}}( (\Rhat{A}{B})^T {\Rot{A}{B}}^i) \quad \text{where}, i = \{1, 2\} \triangleq \{\text{LO}, \text{IMU}\}$ \\
    \quad \textbf{if} $\norm{\mathbf{r}} < \tau$ \textbf{break}.  \\
    \quad \textbf{else, do}:  \\
    \quad  \quad $\Rhat{A}{B} = \Rhat{A}{B} \gexphat{\SO{3}}(\mathbf{r} \Delta t).$ \\
    \vspace{-1mm}
  \end{algorithmic}
  \vspace{-2mm}
\end{algorithm}

Followed by the orientation estimation, a weighted pseudo-inverse method is used to obtain a fused base velocity estimate (\cite{englsberger2018torque}).
This is done so by defining a constrained velocity vector composed of the null velocity $\twistMixedTriv{A}{F}$ of the stance foot assuming holonomic constraints, gyroscope measurements $\yGyro{A}{\text{IMU}}$ obtained from the base collocated IMU and joint velocities $\encoderSpeeds$ measured by joint encoders and defining a linear system of equations with respect to the robot velocity $\nu^{B/B[A]}$, \looseness=-1
\begin{equation}
\begin{bmatrix}
    \twistMixedTriv{A}{F} \\ \Rot{A}{\text{IMU}}\ \yGyro{A}{\text{IMU}} \\ \encoderSpeeds
\end{bmatrix}
   = \begin{bmatrix}
    \freeFloatingJacobian{A}{F/F[A]} \\ \freeFloatingJacobianAng{A}{\text{IMU}/\text{IMU}[A]} \\ \mathbf{B}_c
\end{bmatrix}
\nu^{B/B[A]}.
\end{equation}
This is in the form of an over-constrained system of equations $\mathbf{y} = \mathbf{A}\mathbf{x}$, where $\mathbf{y} \in \R^m$ is the constrained velocity vector, $\mathbf{A} \in \R^{m \times (n+6)}$ is the constrained velocity matrix constructed using the configuration-dependent, free-floating Jacobians of the frames associated with the stance foot and the IMU.
The matrix $\mathbf{B}_c = \begin{bmatrix}
\Zeros{n}{6} & \I{n}
\end{bmatrix}$ is a selector matrix for the joint velocities in order to consider the measured joint velocities as a regularizing term for the system of linear equations.
The solution for this system of linear equations can be obtained as regularized weighted pseudo-inverse,
$$\mathbf{x}^{*} \ =\ \left( \mathbf{A}^{T} \mathbf{W} \mathbf{A}\ +\ \Delta \right)^{-1} \mathbf{A}^{T} \mathbf{W} \mathbf{y},$$ 
where, $\mathbf{W} \in \R^{m \times m}$ is a weighting matrix for the considered measurements, $\Delta \in \R^{m \times m}$ is a regularizer term that dampens the least-square problem in order to avoid singularities during matrix inversion.
The estimate $\mathbf{x}^{*}$ is the optimally fused estimate of the system velocity $\nu^{B}$, from which the fused base velocity estimate $\twistMixedTriv{A}{B}$ can be obtained.

\subsection{Exploiting Feet IMUs and Contact Wrenches}
\label{sec:chap:loosely-coupled-swa-feet-imu}
The assumption of the stance foot velocity to be null for the computation of the base velocity is rather a strong one, assuming perfectly rigid contacts with the environment.
However, there might be cases when the foot might be rotating or slipping when in contact with the ground, and enforcing the zero-velocity constraint to compute the base velocity usually results in poor estimates of the base velocity.
An alternative approach, assuming that the humanoid robots are equipped with an IMU in their end-effectors, is to exploit the measurements from the feet IMUs and contact wrench estimates to improve the base velocity estimates.
In the case of a rotating foot, the gyroscope measurement from the foot IMU can be used to detect a rotation while it is in contact and enforce that only the linear velocity of the point in contact with the ground is zero, instead of considering the entire velocity of the foot is zero. \looseness=-1

In case of a non-zero stance foot velocity, the base velocity can be computed from Eq. \eqref{eq:chap:loosely-coupled:base-vel-rigid} as, \looseness=-1
\begin{equation}
\label{eq:chap:loosely-coupled:base-vel-feet-imu}
    \twistMixedTriv{A}{B} = - (\J_{F_{B}})^{-1} \left(\twistMixedTriv{A}{F}\ -\ \J_{F_{\jointPos}}\ \jointVel\right),
\end{equation}
where, the stance foot velocity can be expressed in linear and angular parts as, 
$$
\twistMixedTriv{A}{F} = \begin{bmatrix} \vMixedTriv{A}{F} \\ \omegaRightTriv{A}{F} \end{bmatrix}.
$$

The feet IMU provides the angular velocity measurement $\yGyro{A}{{F_{\text{IMU}}}}$ in its local frame and this can be used to obtained the angular velocity of the foot $\omegaRightTriv{A}{F}$ directly, since the angular velocity remains the same for the rigid body, 
$$
\omegaLeftTriv{A}{F} = \Rot{F}{{F_{\text{IMU}}}} \yGyro{A}{{F_{\text{IMU}}}}, 
$$ 
where, $\Rot{F}{{F_{\text{IMU}}}}$ is a known, fixed rotation between the foot link frame $F$ and the rigidly attached IMU sensor frame ${F_{\text{IMU}}}$ obtained from the robot model. 
Thus, the right-trivialized angular velocity of the foot is given as,
\begin{equation}
\label{eq:chap:loosely-coupled:right-triv-foot-omega}
\omegaRightTriv{A}{F} = \Rhat{A}{F} \omegaLeftTriv{A}{F},
\end{equation}
where, $\Rhat{A}{F}$ is obtained as an estimate combining base orientation estimate and relative kinematics or as an absolute orientation measurement from the IMU.

Obtaining the linear velocity $\vMixedTriv{A}{F}$ of the rotating foot is not straightforward.
A possible approach can rely on approximating the foot to have a rectangular geometry, using the vertices of the foot as candidate contact points, and enforcing the velocity of the candidate contact point to be zero when actually in contact.
This information about the candidate point in strongest contact with the environment can in turn be used to compute the linear velocity of the foot.

\begin{figure}[!h]
\centering
\includegraphics[scale=0.75, width=\textwidth]{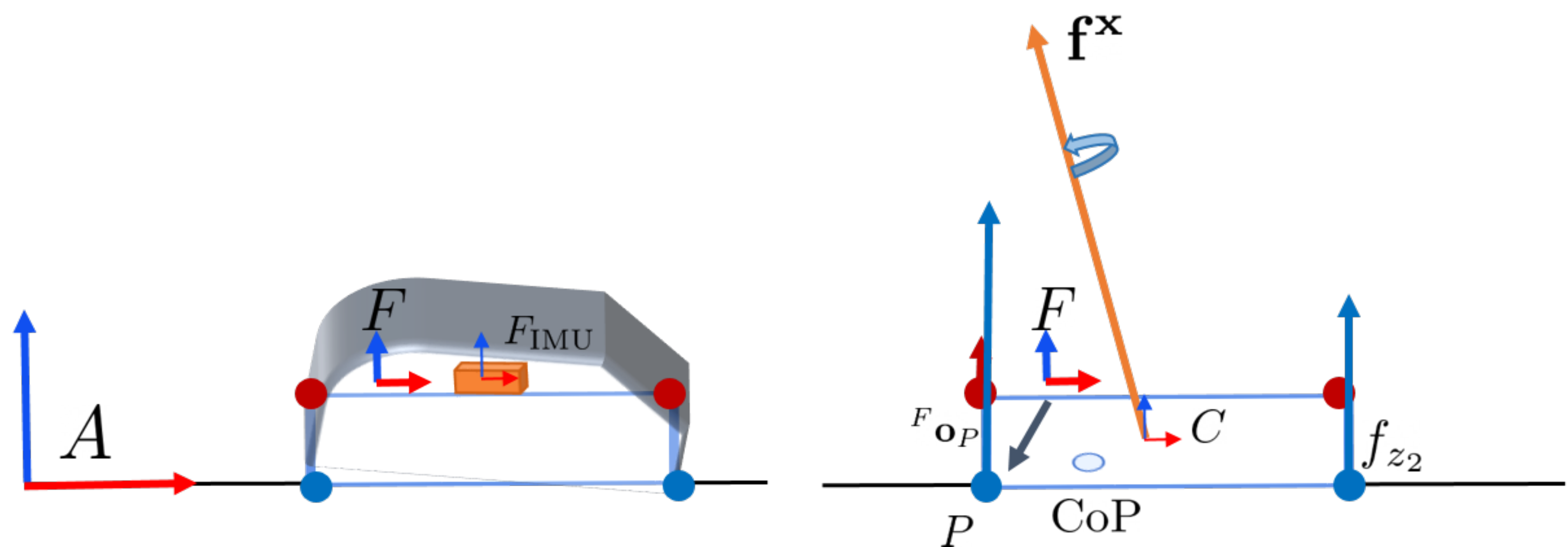}
\caption{Incorporating the velocity of the strongest contact point during foot rotations. The 6D wrench $\wrenchExt{}$ shown as a force-moment pair is decomposed in contact normal forces shown as red/blue arrows at the vertices of the rectangle. This decomposition is based on the position of the center of pressure (CoP) within the support polygon. Blue arrow means the vertex is in contact, while red means loss of contact. The vertex $P$ with highest magnitude force is chosen as the contact point. \looseness=-1}
\label{fig:chap:loosely-coupled:swa-feet-rotation}
\end{figure}

As depicted in Figure \ref{fig:chap:loosely-coupled:swa-feet-rotation}, the strongest contact point is found with the help of contact wrenches acting on the foot.
The wrench is decomposed into contact normal forces weighted on the local Center of Pressure (CoP).
Using the decomposed contact normal forces, the points in contact are found based on a threshold force for inferring contact and the point that is in strongest contact with the environment.
More details about the rectangular foot approximation and contact wrench decomposition into contact normal forces are provided in Section \ref{sec:chap:human-motion:cop-contact}, which is done in the context of human motion estimation, but nevertheless applicable also in the context of the robot given the assumption of a rigid multi-body model consideration for both humanoid robots and humans. \looseness=-1

Using the composition rule of velocities, the left-trivialized linear velocity $\vLeftTriv{A}{F}$ of the foot $F$ in relation with the linear velocity $\vLeftTriv{A}{P}$ with the strongest contact point $P$ is given as,
$$
\vLeftTriv{A}{F} = \Rot{F}{P}\ \vLeftTriv{A}{P} + S(\Pos{F}{P})\ \Rot{F}{P} \omegaLeftTriv{A}{P} + \vLeftTriv{P}{F}.
$$
By the definition of a rigid body, the velocity $\vLeftTriv{P}{F}$ of point $P$ on a rigid body with frame $F$ is zero ($\vLeftTriv{P}{F} = \Zeros{3}{1}$).
Given that point $P$ is the point in strongest contact with the environment inferred from the contact normal forces, the velocity $\vLeftTriv{A}{P}$ of point $P$ is zero  ($\vLeftTriv{A}{P} = \Zeros{3}{1}$).
Thus, the left-trivialized linear velocity of the foot can be obtained as,
\begin{equation}
	\label{eq:chap:loosely-coupled:left-triv-foot-linv}
\vLeftTriv{A}{F} = S(\Pos{F}{P})\ \omegaLeftTriv{A}{F}.
\end{equation}
The mixed-trivialized linear velocity of the foot can be obtained as,
\begin{equation}
\label{eq:chap:loosely-coupled:right-triv-foot-v}
\vMixedTriv{A}{F} = \Rhat{A}{F}\ \vLeftTriv{A}{F},
\end{equation}
where we have all the necessary quantities from the IMU measurements and the model.
The non-zero stance foot velocity obtained from Eqs. \eqref{eq:chap:loosely-coupled:right-triv-foot-omega} and \eqref{eq:chap:loosely-coupled:right-triv-foot-v} can be used to compute an improved base-velocity estimate using Eq. \eqref{eq:chap:loosely-coupled:base-vel-feet-imu}.

As a concluding remark, it must be noted that the method presented in this subsection is quite similar to the concept of Anchoring Pivot (AP) described by \cite{masuya2015dead} which was used to correct the estimated base position with the position displacements due to foot rotations. \looseness=-1
Similarly, the method used to get the strongest point in contact using the CoP of the foot in contact is closely related to the method described in \cite{flayols2017experimental} to weight the base pose estimates obtained using the relative kinematics of the feet (left and right) in the contact.
The method by \cite{flayols2017experimental} uses a truncated bivariate Gaussian distribution with mean at the center of the foot and spread until the boundaries and the weight for the pose estimated from the corresponding limb is obtained using the probability of the location of the Zero Moment Point (ZMP) within the foot. 
While the method described by those authors is used for inter-foot weighting, the method described in this section is used for intra-foot weighting.

\section{Extension to Absolute Localization in a Known Environment}
\label{sec:chap:loosely-coupled-localization}
A straightforward application of the averaging on the Lie groups is proposed as an approach to localize the robot in a known environment of landmarks.
A multiple landmark consensus algorithm is used to obtain the absolute pose of the robot in the inertial frame. This follows a weighted averaging over $\SE{3}$ to obtain the pose of the base link from a set of measured relative poses and known landmark poses, where the weights are decided by the inverse distance of the landmarks with respect to the base frame at the current time instant. \looseness=-1

An environment with known landmark poses is described by the set of poses, $$\mathbf{H}_{A, L} = \{\Transform{A}{{L_1}}, \dots, \Transform{A}{{L_k}}\}.$$
As the robot moves through this environment, the observed landmark measurements at each time-instant describe the relative poses $\mathbf{H}_{B, L} = \{\Transform{B}{{L_1}}, \dots, \Transform{B}{{L_k}}\}$.
The absolute base pose measurements then can simply be obtained by a composition of the a priori known absolute landmark pose and the relative poses between the robot base and the landmark:
$$
\mathbf{H}_{A, B} = \{\Transform{A}{{L_1}}\ (\Transform{B}{{L_1}})^{-1}, \dots, \Transform{A}{{L_k}}\ (\Transform{B}{{L_k}})^{-1} \}.
$$
The base pose estimated from the several landmark observations might not be the same, assuming the noisy nature of the observations.
Thus, a weighted average of the base pose estimates is chosen to represent the optimal estimate of the base pose in the known environment.
The weight of each landmark is computed using a normalized inverse distance metric,
$$
w_i = \frac{\frac{1}{d_i}}{\frac{1}{d_1} + \dots + \frac{1}{d_k}},
$$
where $d_i$ is the distance of the $i$-th landmark from the base link frame.
From the set of absolute base poses $\mathbf{H}_{A, B}$ estimated from the known landmarks and the associated inverse distance weights $w = \{w_1 , \dots, w_k\}$, the weighted averaging of $\SE{3}$ gives the optimal base pose $\Hhat{A}{B}$.
The fusion method is described in Algorithm \ref{algo:chap:loosely-coupled:pose-averaging}, whose underlying structure is the same as Algorithm \ref{algo:chap:loosely-coupled:rotation-averaging}.
\begin{algorithm}[htbp]
\small
  \caption{Weighted Pose Averaging \looseness=-1}
  \label{algo:chap:loosely-coupled:pose-averaging}
  \begin{algorithmic}
    \State \textbf{Input:} $\{\Transform{A}{{L_1}}\ (\Transform{B}{{L_1}})^{-1}, \dots, \Transform{A}{{L_k}}\ (\Transform{B}{{L_k}})^{-1} \} \in \SE{3}$ \\ 
    \quad \quad $\{w_1 , \dots, w_k\} \in \R,\ \text{s.t.} \sum_{i=1}^k w_i = 1$ \\
    \textbf{Output:} $\Hhat{A}{B} \in \SE{3}$ \\
    \State \textbf{Initialize:}\\
    $\Hhat{A}{B} = {\Transform{A}{B}}^1$ \\
    Desired tolerance $\tau > 0$ \\
    Step Size $\Delta t > 0$\\
    
    \State \textbf{Iterate until convergence:} \\
    \quad $\mathbf{r} = \sum_{i = 1}^k w^i \glogvee{\SE{3}}( (\Hhat{A}{B})^{-1} {\Transform{A}{B}}^i)$ \\
    \quad \textbf{if} $\norm{\mathbf{r}} < \tau$ \textbf{break}.  \\
    \quad \textbf{else, do}:  \\
    \quad  \quad $\Hhat{A}{B} = \Hhat{A}{B} \gexphat{\SE{3}}(\mathbf{r} \Delta t).$ \\
    \vspace{-1mm}
  \end{algorithmic}
  \vspace{-2mm}
\end{algorithm}


\section{Experimental Results}
\label{sec:chap:loosely-coupled:experiments}
In this section, a preliminary validation of the existing Manton's convergent algorithm (\cite{manton2004globally}) for weighted of averaging rotations and poses is performed first to simply demonstrate the computational utility of such an averaging approach.
This is followed by the experimental evaluation of the proposed SWA estimator (which is the main contribution of this chapter) for a robot walking experiment. \looseness=-1

\subsection{Rotation Averaging}
\begin{figure}[!h]
\centering
\includegraphics[scale =0.5]{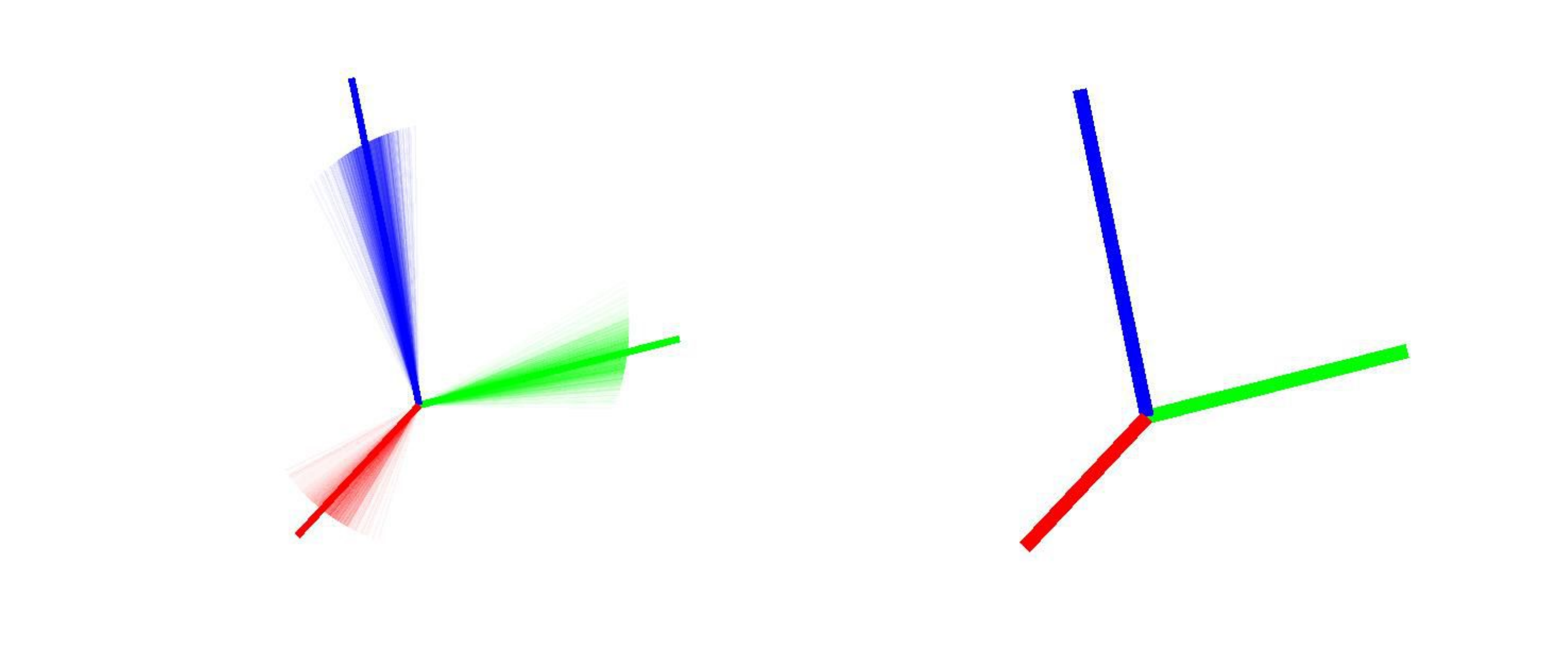}
\caption{Rotation averaging: Samples of rotation elements drawn from a concentrated Gaussian distribution (left); mean rotation computed by Manton's averaging algorithm (right).}
\label{fig:chap:loosely-coupled:rot-avg}
\end{figure}

In this subsection, Manton's averaging algorithm is validated for finding the average of $1000$ samples of rotations.
The rotation elements are sampled from a concentrated Gaussian distribution with mean rotation $\mathbf{R}$ and a perturbation in the vector space $\err$.
When parametrized with roll-pitch-yaw Euler angles' representation, the mean rotation has the value $(10, 10, 10)$ degrees. 
The corresponding rotation matrix $\mathbf{R}$ is then,
$$
\mathbf{R} = 
\begin{bmatrix}
0.9698 &  -0.1413 &    0.1986 \\
    0.1710 &    0.9751  & -0.1413 \\ 
   -0.1736  &  0.1710 &    0.9698
\end{bmatrix}.
$$
This matrix is denoted by the coordinate frame with the longer axes on the left side of Figure \ref{fig:chap:loosely-coupled:rot-avg}.
The perturbation vector in the tangent space of $\SO{3}$ is chosen to be $\err = \begin{bmatrix}
0.05 & 0.05 & 0.05
\end{bmatrix}^T$,
The samples drawn from the concentrated Gaussian distribution are depicted by the coordinate frames with axes that are thin and shorter than the mean rotation.
These samples are then passed as inputs to the averaging algorithm described in Algorithm \ref{algo:chap:loosely-coupled:pose-averaging}, excluding the mean rotation.
Equal weights are assigned for each rotation.
The gradient descent is initialized with a step size, $\Delta t = 0.1$ and the error tolerance $\tau = 10^{-4}$ as the termination condition.
The rate of convergence and the accuracy of the algorithm depend on these hyper-parameters.

The mean from the averaging algorithm for $1000$ samples, depicted on the right in Figure \ref{fig:chap:loosely-coupled:rot-avg}, is computed in $0.1$ seconds using MATLAB without any code optimization and is given as, \looseness=-1
$$
\Rhat{}{} = 
\begin{bmatrix}
0.9704  & -0.1401 &    0.1968 \\
    0.1693 &   0.9756  & -0.1400 \\
   -0.1724  &  0.1692 &   0.9704
\end{bmatrix}.
$$
The error between the true mean and the computed mean in RPY angles in degrees equals $$\begin{bmatrix}
0.1086 & 0.0733 & 0.1060
\end{bmatrix}^T.$$

\subsection{Pose Averaging}
\begin{figure}[!h]
\centering
\includegraphics[scale =0.4]{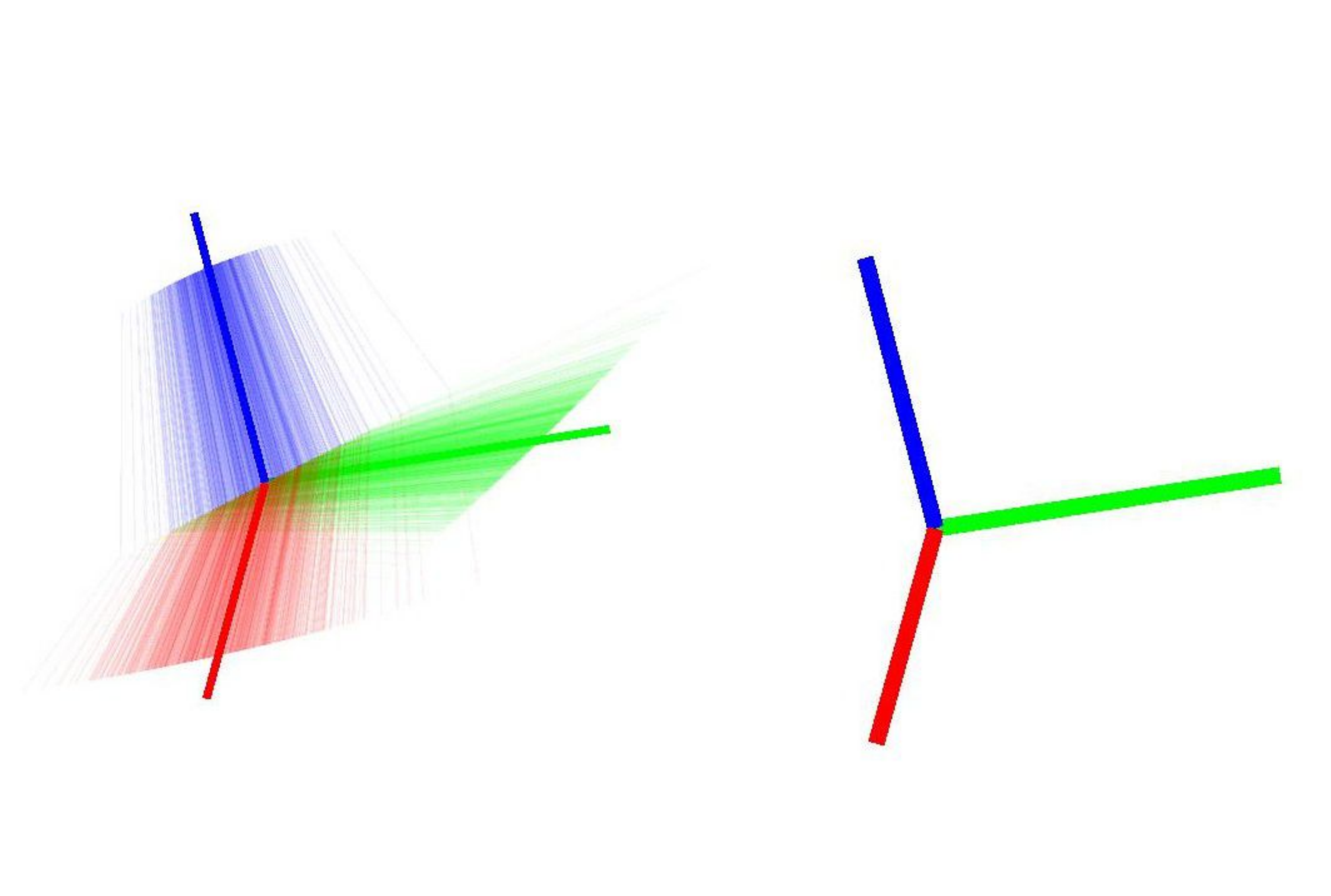}
\caption{Pose averaging: Samples of pose elements drawn from a concentrated Gaussian distribution (left); mean pose computed by Manton's averaging algorithm (right).}
\label{fig:chap:loosely-coupled:pose-avg}
\end{figure}

A similar validation is performed for pose averaging.
The rotation and position mean for constructing the mean pose are chosen as $(10, 10, 10)$ degrees in roll-pitch-yaw parametrization and $(0, 0.5, 0)$ meters in Cartesian coordinates.
The perturbation in the tangent space of $\SE{3}$ is chosen as $\err = \begin{bmatrix}
0.05 & 0.05 & 0.05 & 0.05 & 0.05 & 0.05
\end{bmatrix}^T.$
The mean and the samples obtained from the concentrated Gaussian distribution are depicted in Figure \ref{fig:chap:loosely-coupled:pose-avg} on the left.
The samples are passed as inputs to the averaging algorithm described in Algorithm \ref{algo:chap:loosely-coupled:pose-averaging} with equal weights.
The gradient descent is initialized with the same parameters described in the rotation averaging experiment.

The mean from the averaging algorithm for $1000$ samples, depicted on the right in Figure \ref{fig:chap:loosely-coupled:pose-avg}, is computed in $1.11$ seconds using MATLAB and is given as, \looseness=-1
$$
\Hhat{}{} = 
\begin{bmatrix}
0.9710 &  -0.1386  &  0.1947 &  -0.0035 \\
0.1672  &  0.9761 &  -0.1385  &  0.4965 \\
-0.1709 &   0.1671  &  0.9710 &  -0.0033 \\
0    &     0    &     0  &  1.0000
\end{bmatrix}.
$$
The mean pose is depicted in Figure \ref{fig:chap:loosely-coupled:pose-avg} on the right side.
The error between the true mean rotation and the computed mean rotation in RPY angles in degrees equals $$\begin{bmatrix}
0.2373 & 0.1606 & 0.2318
\end{bmatrix}^T.$$
The position error in meters equals $\begin{bmatrix}
0.0035 & 0.0035 & 0.0033
\end{bmatrix}^T.$


\subsection{Evaluation of SWA Estimator}

\subsubsection{Experiments in Simulation}
\begin{figure}[!h]
	\centering
	\includegraphics[scale =0.5]{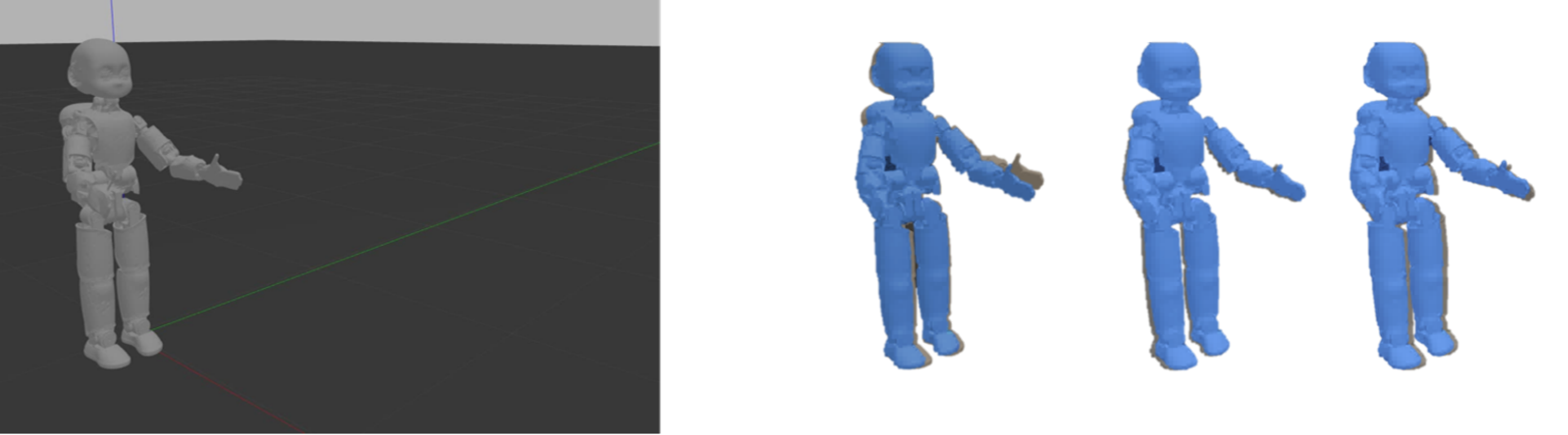}
	\caption{iCub in stationary, position-controlled two-feet standing configuration in Gazebo simulation environment (left). The three images from the right show an overlay of the robot configuration while being perturbed when in position-controlled mode. There are changes in the base state due to these perturbations which are very small in magnitude.}
	\label{fig:chap:loosely-coupled:swa-low-perturb}
\end{figure}

\begin{figure}[!h]
	\centering
	\includegraphics[scale =0.5]{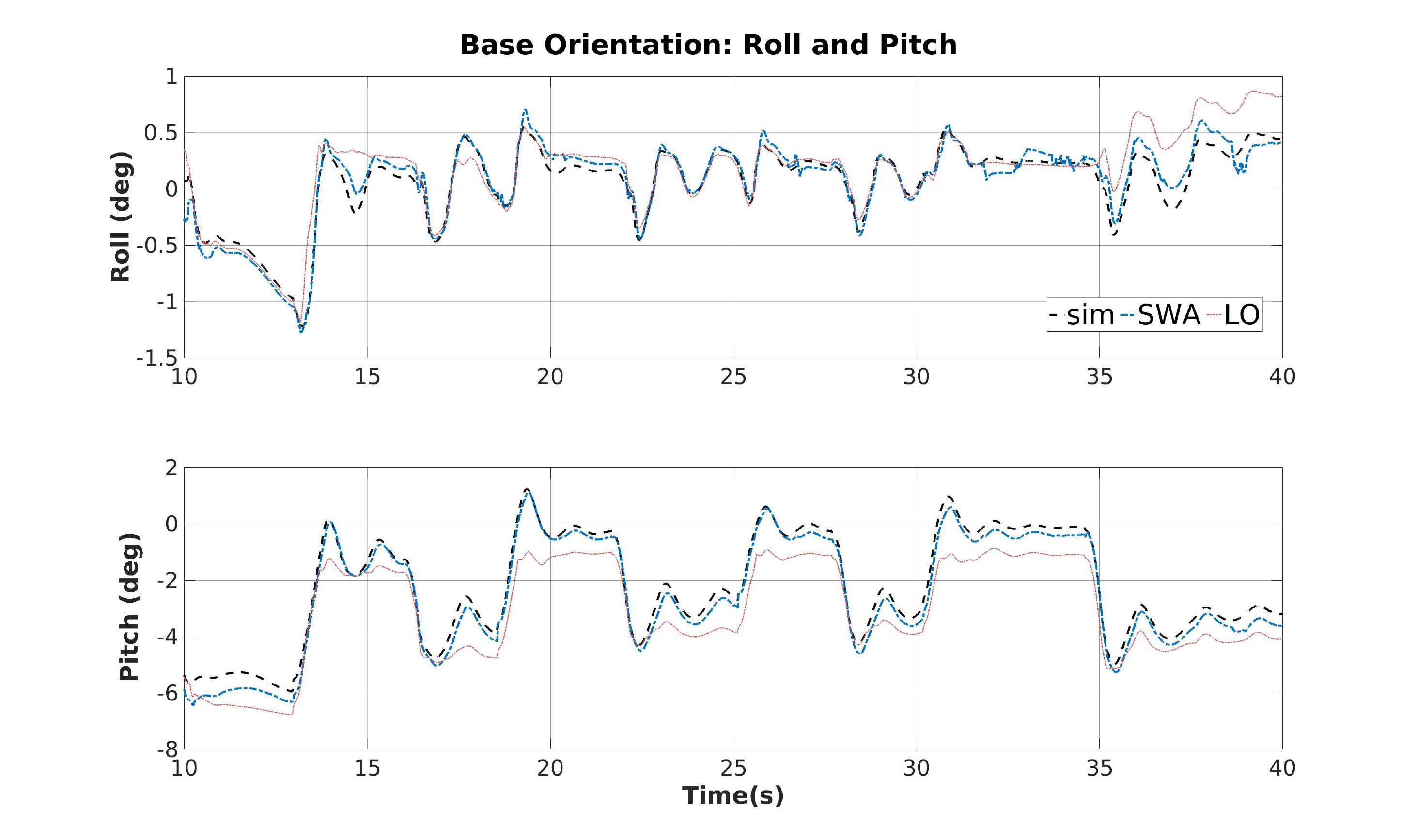}
	\caption{Base orientation estimates from the legged odometry (in red) and SWA (in blue) for the experiment where arbitrary perturbations are applied to the position-controlled robot in stationary configuration causing low magnitude of change in the base state. \looseness=-1}
	\label{fig:chap:loosely-coupled:swa-low-perturb-rotation}
\end{figure}

Firstly, we motivate the use of multimodal fusion-based approaches floating base estimation of the humanoid robot through an experiment in simulation.
This is demonstrated by a simple experiment where the robot is standing stationary while all the actuators of the robot are position-controlled to maintain the standing configuration.
Arbitrary low-magnitude perturbations are induced to the robot causing it to move without having to change the internal joint configuration of the robot. 
This can be seen in Figure \ref{fig:chap:loosely-coupled:swa-low-perturb} which shows the snapshots of the robot motion where we have overlaid two sequential instants in time around which the perturbations act on the robot.
Figure \ref{fig:chap:loosely-coupled:swa-low-perturb-rotation} shows the orientation estimates of the base link during this experiment for the vanilla legged odometry approach and SWA estimator.

It is evident that high frequency changes in the robot state during the application fo the external perturbation is not captured effectively by legged odometry relying solely on the contact-aided kinematics, whereas SWA is able to track these changes owing to the fusion with IMU measurements.
This shows a clear advantage of exploiting IMU measurements for floating base estimation which will particularly be useful in tracking base orientation even in cases when the robot falls due a high-frequency perturbation and is unable to establish rigid contacts with the environment.

Clearly, the floating base estimation can benefit in fusing information from the IMU.
Further, the inclusion of feet IMU measurements to account for non-zero feet velocities in the base velocity computation as seen in Section \ref{sec:chap:loosely-coupled-swa-feet-imu} shows an improvement in the velocity estimates.
This is particularly important in motions from high frequency perturbations, where usually the foot rotations are not explicitly considered in base state estimation.
In the case of SWA, this is handled using the gyroscope measurement and forcing null velocity for the point on the foot in strongest contact with the environment.

\begin{figure}[!h]
	\centering
	\includegraphics[scale =0.5]{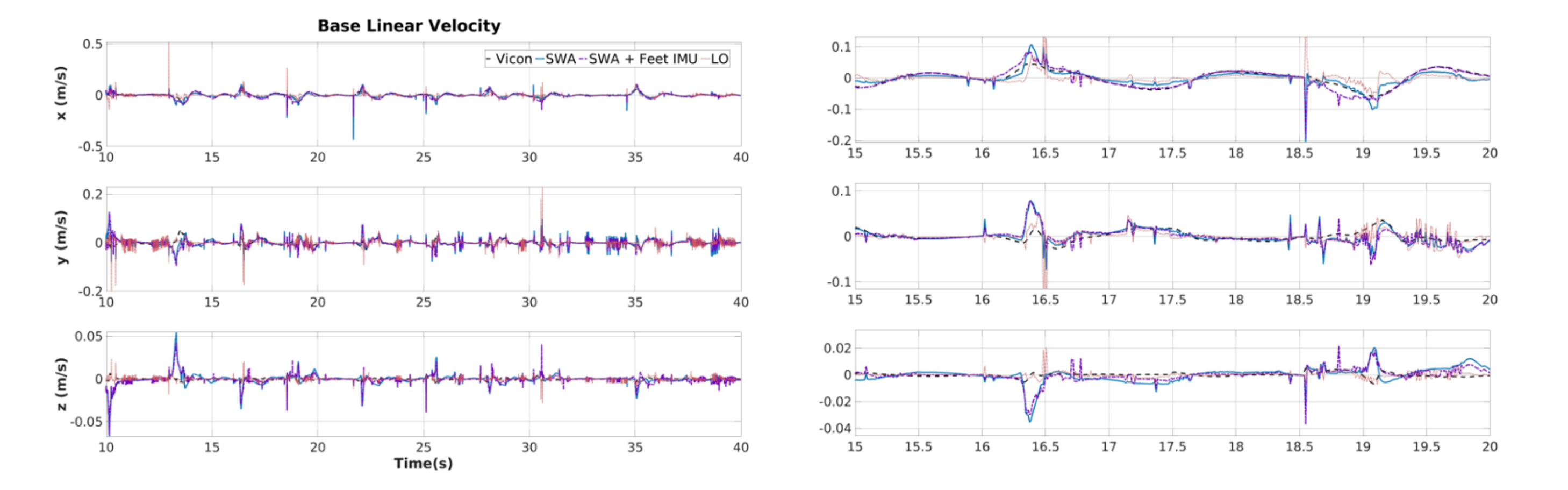}
	\caption{(left) Base velocity estimates from the legged odometry (in red), SWA (in blue) and SWA with feet IMU measurements (in violet) for the experiment where arbitrary perturbations are applied to the position-controlled robot in stationary configuration. (right) A zoomed-in snapshot of the plot in the left between time $t=15$ to $t=20$ seconds. \looseness=-1}
	\label{fig:chap:loosely-coupled:swa-low-perturb-basevel}
\end{figure}

\begin{table}[ht] 
	\caption{Errors comparison of Simple Weighted Averaging estimator (SWA), SWA with feet IMU, and Legged Odometry (LO) for arbitrary high-frequency perturbation experiment. \looseness=-1}
	\centering 
	\begin{tabular}{|c c c c c|} 
		\hline
		{Filter}	& \multicolumn{4}{c|}{\emph{Arbitrary Perturbation}}  \\
		\hline
		& RMSE & \multicolumn{3}{c|}{Max. Error} \\
		\hline
		& vel [$\si{\meter/\second}$] & vel x [$\si{\meter/\second}$] & vel y [$\si{\meter/\second}$] & vel z [$\si{\meter/\second}$]  \\
		\hline
		{LO} & 0.0254 & 0.5156 & 0.2131 & \textbf{0.0213}    \\
		{SWA} & 0.0225 & 0.1190 &  0.0825 & 0.0453  \\
		{SWA + Feet IMU} & \textbf{0.0201} & \textbf{0.0683} &  \textbf{0.0787} & 0.0563  \\
		\hline
	\end{tabular} 
	\label{table:swa-feet-imu-errors} 
\end{table}

Figure \ref{fig:chap:loosely-coupled:swa-low-perturb-basevel} shows the comparison of base velocity estimates from legged odometry, SWA estimator and SWA estimator incorporating feet IMU measurements and the corresponding RMSE and maximum errors are tabulated in Table \ref{table:swa-feet-imu-errors}.
On a first look, all the three estimators seem to be perform similarly in the base velocity computations and suffering regularly from the high magnitude noise introduced by noisy joint velocities.
However, zooming into the plot, it can be observed that the SWA estimator incorporating the feet IMU measurements follow the ground truth trajectory of the velocity estimates more closely than its counterparts.
This is also evident from Table \ref{table:swa-feet-imu-errors} that the SWA estimator incorporating the feet IMU measurements suffers from the least RMSE and maximum error in all directions except for $\mathbf{z}$.
The offsets between the estimates and the ground truth throughout the time-series are at instances when the perturbation is induced on the robot.
During these instances, it is noticed that the Gazebo simulated gyroscope measurements from the feet IMU suffer from high peak noise, which when accounted in the linear velocity computation of the feet as seen in Eq. \eqref{eq:chap:loosely-coupled:left-triv-foot-linv}, causes these errors in the resulting estimates.
Nevertheless, even while accounting for these errors, the maximum error of the SWA estimator incorporating feet IMU measurements remain lower than others. \looseness=-1

Next, we mainly want to motivate why simple estimators such as SWA will benefit from rigorous mathematical operations such as averaging on Lie groups while fusing orientation estimates from different measurement sources.
For this experiment, we introduce a modification in the SWA related to the fusion of orientation estimates coming from the IMU and the legged odometry.
The original SWA estimator follows an averaging on $\SO{3}$ while the modified uses an averaging on the Roll-Pitch-Yaw (RPY) angles directly.

\begin{figure}[!h]
	\centering
	\includegraphics[scale =0.5]{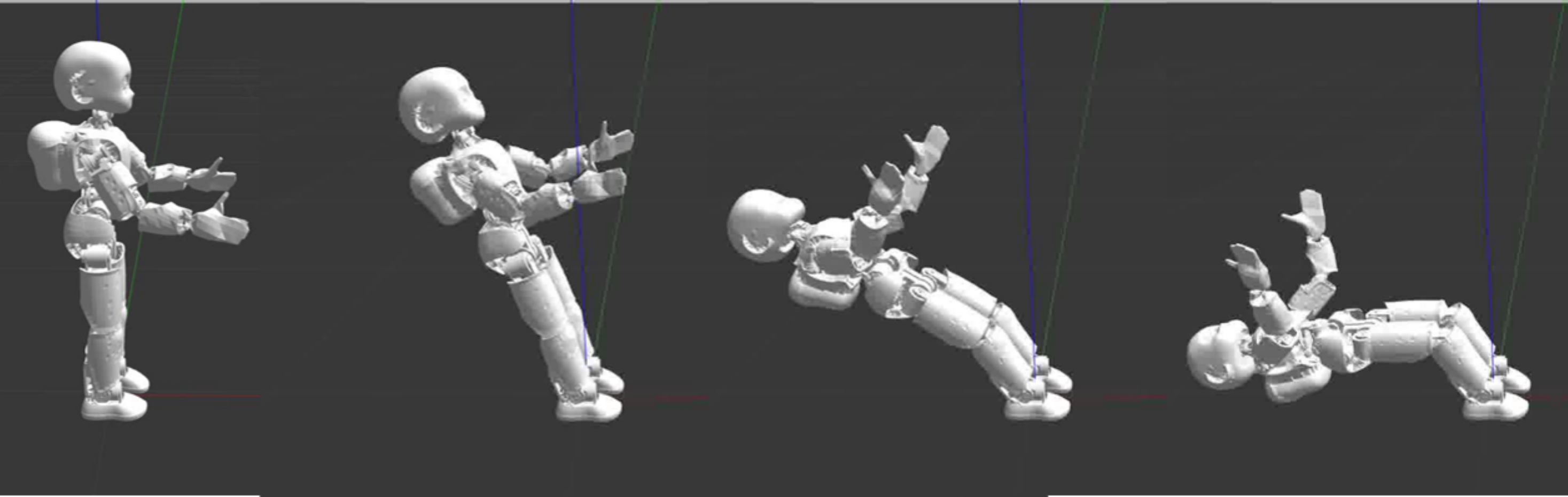}
	\caption{iCub with feet fixed to the ground in Gazebo simulation environment is made to go through a falling motion by commanding the ankle joints to its software limits, while the feet remain fixed to its initial pose. \looseness=-1}
	\label{fig:chap:loosely-coupled:swa-falling}
\end{figure}

\begin{figure}[!h]
	\centering
	\includegraphics[scale =0.3]{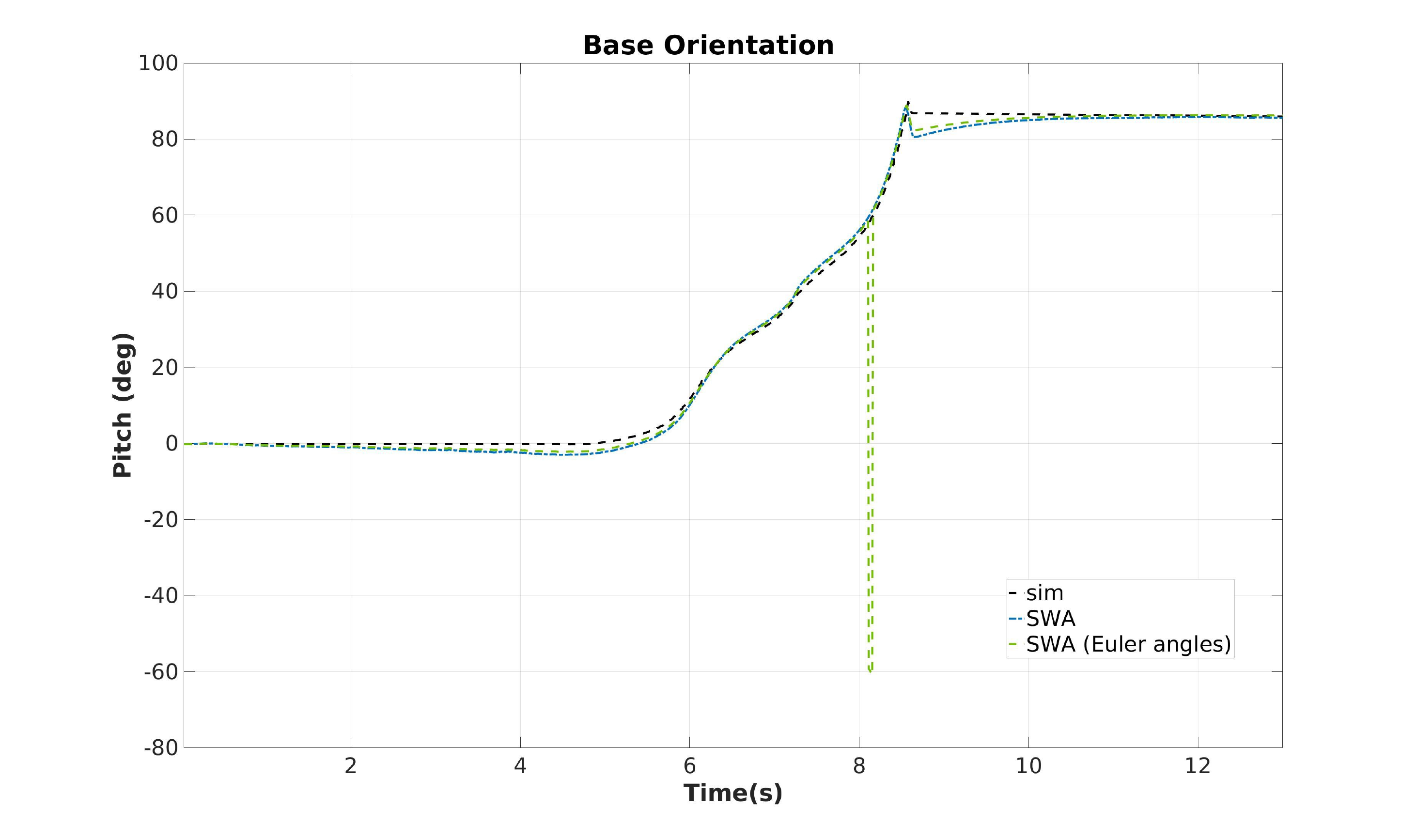}
	\caption{Comparison of pitch orientation estimates from SWA estimator employing fusion methods based on averaging Euler angles (green dashed) and averaging on $\SO{3}$ (blue dash-dotted) for a Gazebo simulated falling experiment. \looseness=-1}
	\label{fig:chap:loosely-coupled:swa-rotation}
\end{figure}

We simulate an experiment in which the change of orientation during the experiment will be large enough that we may encounter a wrap-around situation of the Euler angles.
For this purpose, we consider a position-controlled iCub in Gazebo simulation with its feet fixed to the environment. 
The ankle joints are commanded to reach its software limits in order to emulate a falling motion on the robot while being fixed at its feet as shown in Figure \ref{fig:chap:loosely-coupled:swa-falling}.
Such a falling motion induces a huge variation in the base link orientation as seen in Figure \ref{fig:chap:loosely-coupled:swa-rotation}.
On comparing the orientation estimates obtained from the original SWA and modified SWA, it can be seen that the latter which performs a fusion on the Euler angles suffers from a clear discontinuity.
This is due to the fact that the orientation of the IMU rigidly attached to the base link undergoes a wrap around between $-180$ and $180$ degrees in its roll angle.
It must be noted that the coordinate frame of the IMU does not coincide with that of the base link and the fusion is primarily done in reference to the IMU frame and is transformed later to the base link for outputs.
Clearly, such discontinuities in the base orientation are troublesome especially when closing the loop with balancing and locomotion controllers on the humanoid robot.
It is evident that this problem is suitably tackled by handling the fusion appropriately in the $\SO{3}$ space as noticed with the original SWA estimates, thereby showing a clear advantage of the proposed method. \looseness=-1

\subsubsection{Walking experiment on the real robot}
\label{sec:chap:swa-walking-exp}
In this subsection, the proposed SWA estimator is further validated for locomotion experiments on the real robot.
For this experimental evaluation, we do not exploit the use of IMUs on the feet and the contact wrench decomposition to improve the velocity estimation.
A position-controlled walking (\cite{romualdi2020benchmarking}) experiment is used for the evaluation of the SWA estimator on the real robotic platform, in an open-loop fashion, in comparison with the traditional legged odometry estimator.\looseness=-1

\begin{figure}[!h]
\centering
\includegraphics[scale=0.25, width=\textwidth]{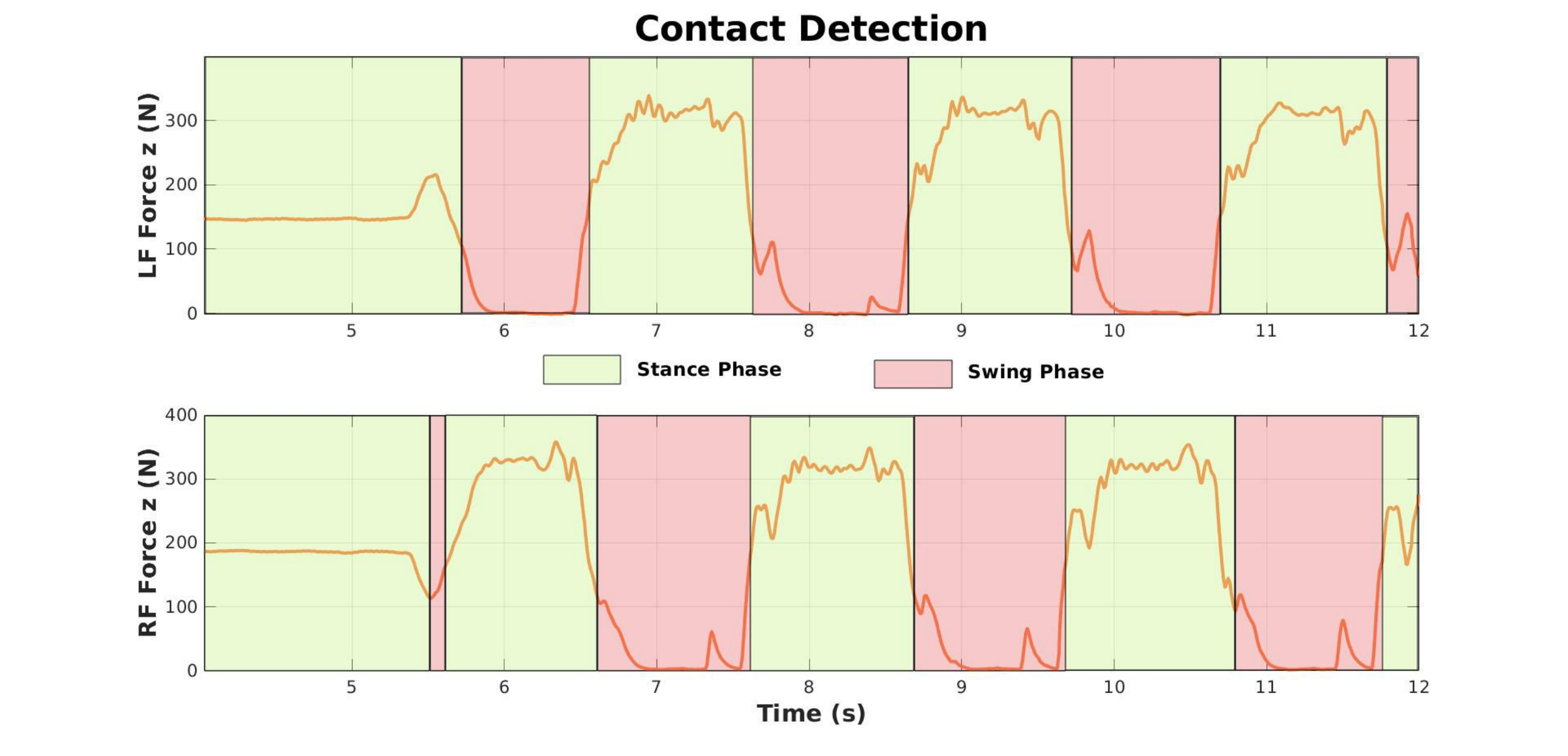}
\caption{Schmitt trigger thresholding based contact detection}
\label{fig:chap:loosely-coupled:swa-contact}
\end{figure}

The robot is made to walk in a forward direction for $1$ meter in a room mounted with a Vicon motion capture system.
The joint encoder measurements which are available at $1000$ Hz are sub-sampled to $100$ Hz to match the other proprioceptive measurements.
The Xsens MTi-300 series IMU mounted on the base link of the robot streams linear accelerometer and gyroscope measurements at $100$ Hz.
The feet contact wrenches estimated from a whole-body dynamics estimation algorithm (\cite{nori2015icub}) is available at $100$ Hz and is passed through a Schmitt trigger based contact thresholding for inferring contact states of the feet, as depicted in Figure \ref{fig:chap:loosely-coupled:swa-contact}.
The Schmitt trigger is tuned with contact make and break thresholds as $150$ N and $120$ N respectively with stable switching time parameters as $0.01$ seconds for both making and breaking contacts.
These values allow the contact detection to be robust to the high-frequency noise of the wrench estimates.
The base estimation algorithms are run at $100$ Hz, given the limits of the sensor rates.

\begin{table}
	\begin{center}
						\begin{tabular}{c|c}
						\hline 
							Sensor & noise std dev. \\
								\hline	
							Lin. Accelerometer & $0.0089\    (\si{(\metre \per {\second^2})/\sqrt{\hertz}})$	 \\
							Gyroscope & $3.16 \times 10^{-4}\  \si{(\radian \per \second)/\sqrt{\hertz}}$ \\
							Gyro. bias & $10^{-5}\  \si{(\radian \per \second^2)/\sqrt{\hertz}}$					\\	\hline 
						\end{tabular}
	\caption{IMU Noise parameters from Allan Variance}
	\label{table:swa-allan-var-params}
	\end{center}
\end{table}

The attitude estimation algorithm used for retrieving rotation estimates from the IMU is a Quaternion Extended Kalman Filter (QEKF).
Only accelerometer and gyroscope measurements are used, thus making the roll and pitch estimates of the base link alone reliable.
This is the reason why we project the yaw angle estimated by the legged odometry onto the IMU estimate before averaging in the $\SO{3}$ space.
The noise parameters for the IMU measurements required to tune the QEKF covariances are obtained through the computation of Allan variance.
The noise parameters are listed in Table \ref{table:swa-allan-var-params}. \looseness=-1
Equal weights are chosen for fusing the legged odometry estimate and the IMU estimate of the base orientation.
Equal weights are chosen also for the gyroscope, joint velocity, and stance-foot velocity measurements used for the computation of the base velocity.


\begin{figure}[!ht]
\centering
\begin{subfigure} { \textwidth}
\centering
\includegraphics[scale=0.4]{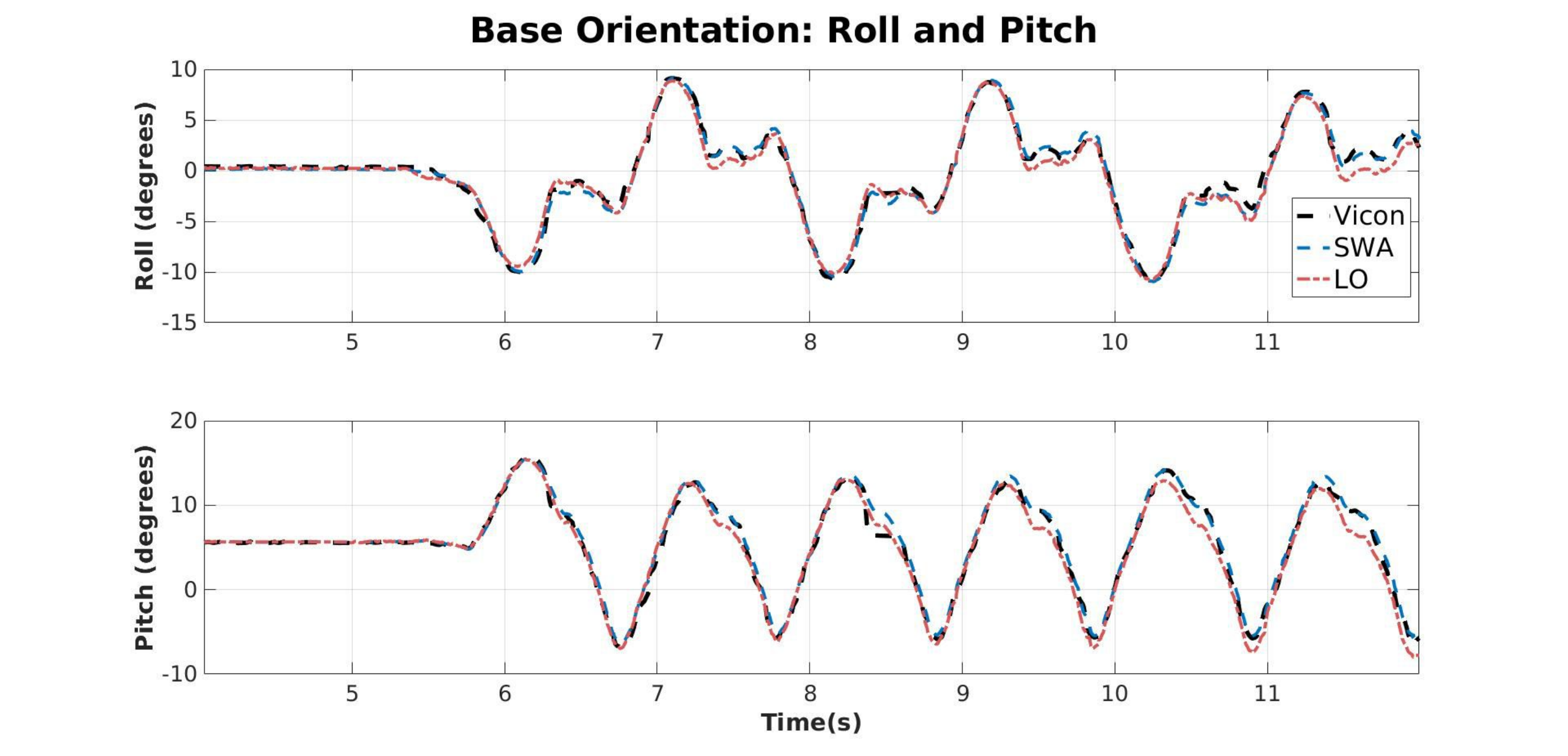}
\caption{}
\end{subfigure}
\begin{subfigure}{ \textwidth}
\centering
\includegraphics[scale=0.4]{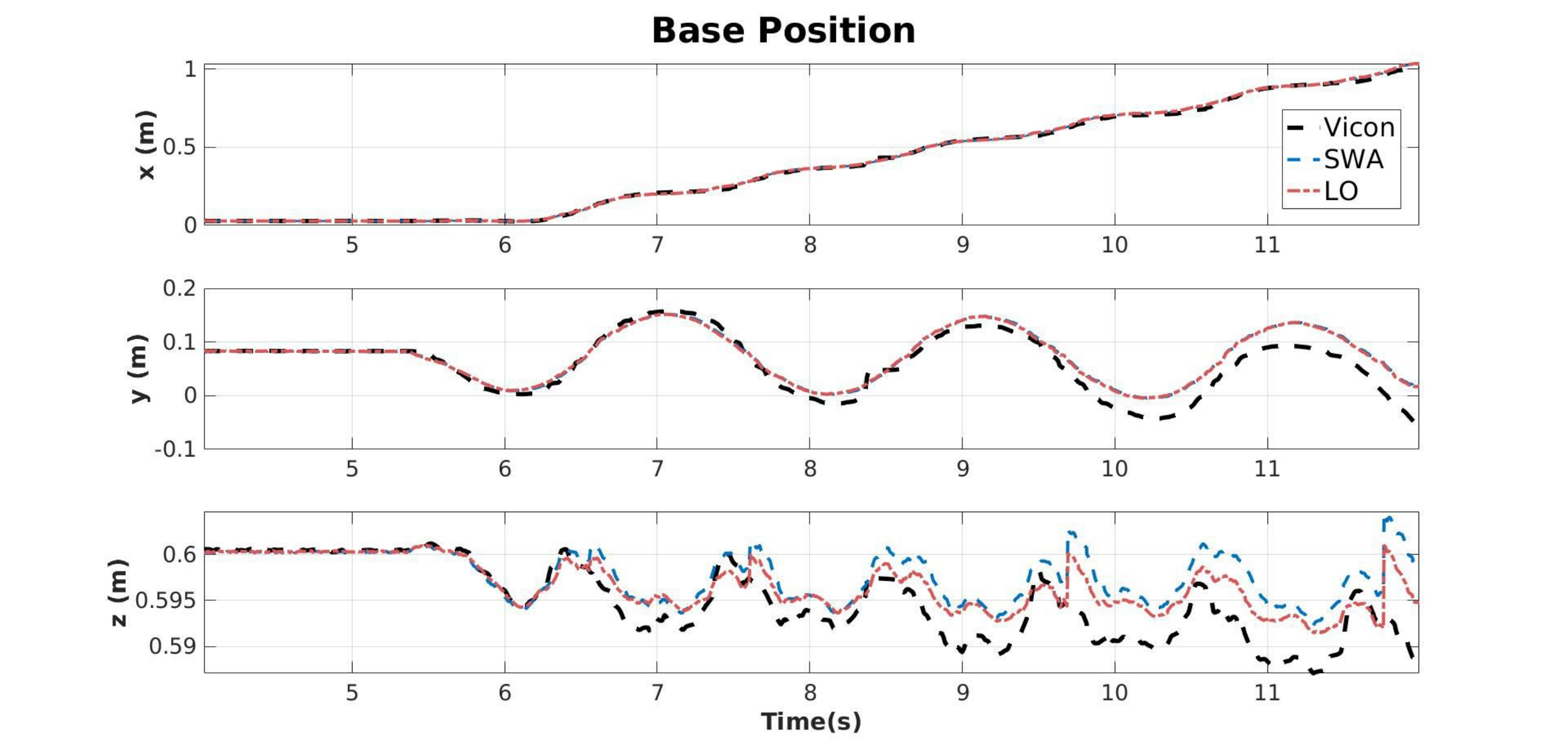}
\caption{}
\end{subfigure}
\begin{subfigure}{\textwidth}
\centering
\includegraphics[scale=0.4]{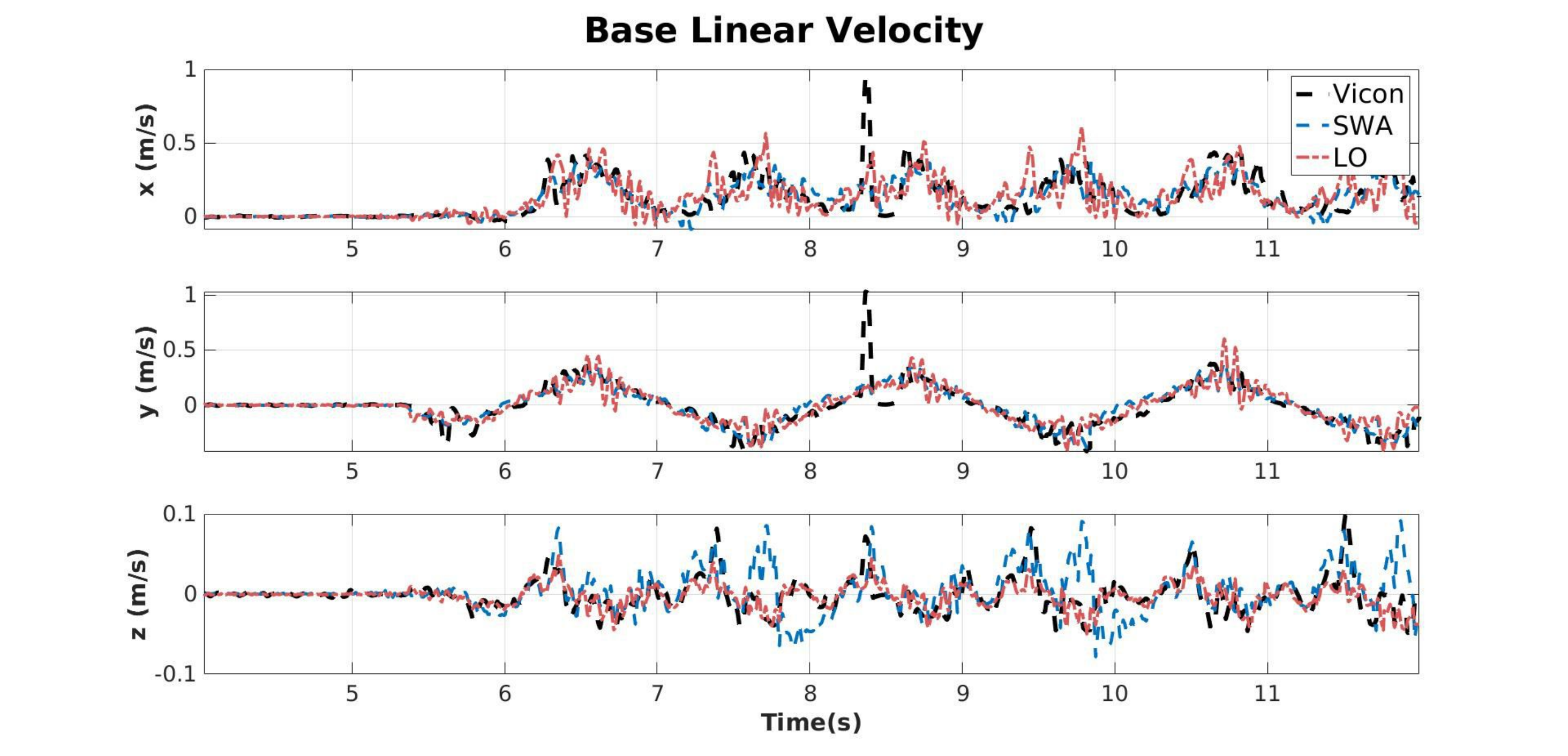}
\caption{}
\end{subfigure}
\caption{Orientation, position and linear velocity estimates of the base link from SWA (blue) and legged odoemtry (red) in comparison with ground-truth trajectory (black) retrieved from Vicon motion capture system. \looseness=-1}
\label{fig:chap:loosely-coupled:swa-estimates}
\end{figure}

The estimated trajectories and ground truth trajectories are available in their own coordinate systems.
While rigorous trajectory alignment methods such as \emph{Umeyama} method (\cite{zhang2018tutorial}) or the \emph{Cashbaugh and Kitts} method (\cite{cashbaugh2018automatic}) exist which find the closest points in trajectories to compute an alignment transform (analogous to Iterative Closest Point algorithm), we use a single-state trajectory alignment to bring the trajectories in a common reference frame. 
This single state is chosen as the very first instant of operation for the estimators when the robot is in stationary condition and initial conditions are known.

The ground truth linear velocity is obtained by Savitzky-Golay filtering of the ground truth position trajectory with the $3$rd order polynomial differentiation using a window size of $51$ samples.
A discontinuity in the ground truth trajectories is noticed between $t=8$ and $t=9$ seconds in Figure \ref{fig:chap:loosely-coupled:swa-estimates} which is due to the due to occlusion of reflective markers as the robot moved in the space-constrained Vicon environment. 
This discontinuity is realized as a peak in the ground truth velocities during the same time interval.
This discontinuity is the reason for higher velocity errors observed in Table \ref{table:swa-errors}.

\begin{figure}[H]
\centering
\includegraphics[scale=0.4]{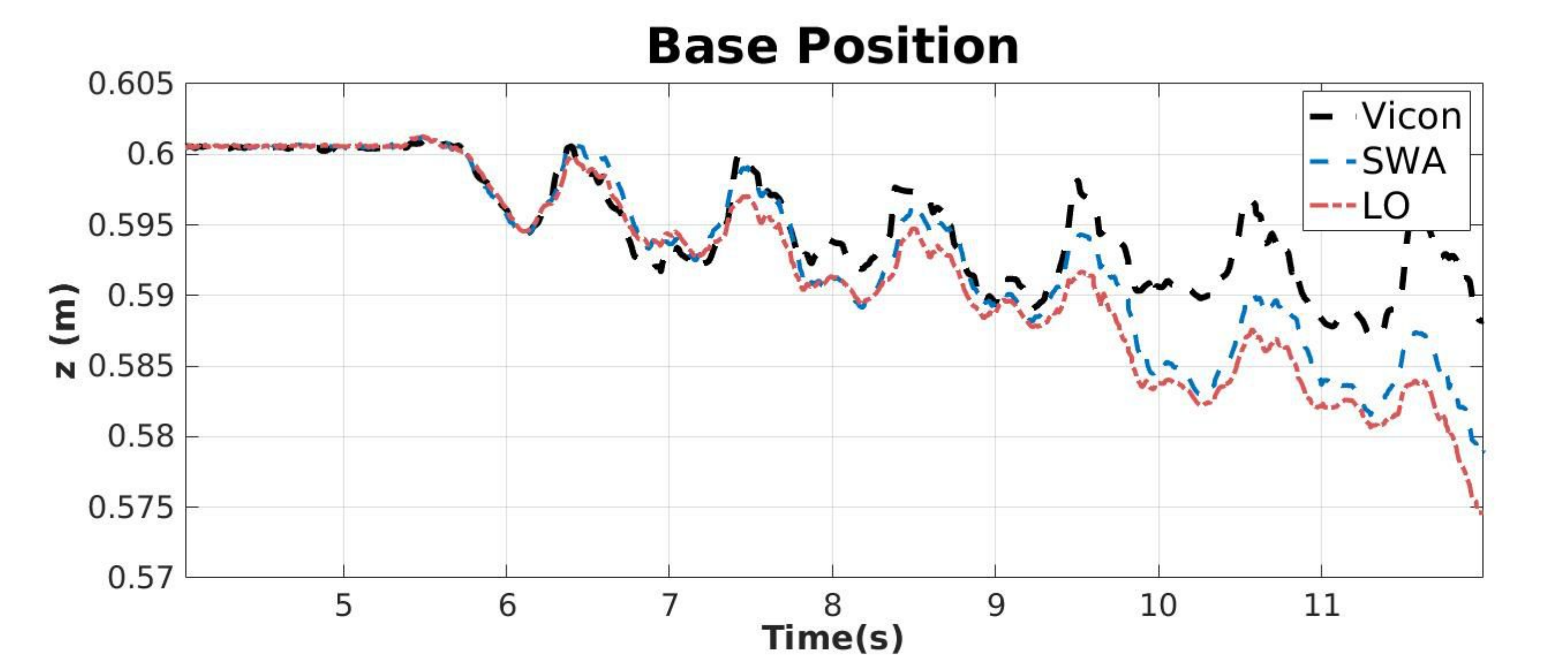}
\caption{Position estimate along the z-direction from SWA estimator and legged odometry (LO) without imposing flat floor constraints in comparison with ground-truth trajectory retrieved from Vicon motion capture system.}
\label{fig:chap:loosely-coupled:swa-height}
\end{figure}

Figure \ref{fig:chap:loosely-coupled:swa-estimates} shows the orientation, position and velocity estimates obtained from the SWA estimator (\emph{blue dashed line}) and legged odometry (\emph{red dash-dotted line}) in comparison with the Vicon ground truth trajectory (\emph{black dashed line}) for the position-controlled walking experiment.
It can be observed that the SWA estimator and legged odometry estimator perform comparably in this qualitative comparison.
Both the estimators are subject to position drifts in $y$- and $z$-directions.
We specifically enforce a flat-floor constraint every time the foot makes a rigid contact with the environment to bound the position drift in the z-direction within the order of millimeters.
This is done so by assuming a known value of the height of the foot frame and resetting the pose of the fixed frame to have this height when the contact is made.
This is rather a strong constraint that might not be suitable even for slightly rough terrains and might also lead to discontinuities in the estimates during contact switching in the context of highly dynamic motions, like human walking.
Without imposing the flat floor constraints the position drifts in the $z$-direction are in the order of $2$-$3$ centimeters for a $1$ meter walking, as shown in Figure \ref{fig:chap:loosely-coupled:swa-height}.
The drifts in the position estimates are caused mainly due to imprecise timings of the contact detection and unmodeled slipping while making contact with the ground. \looseness=-1

\begin{table}[ht] 
\caption{Absolute Trajectory Error (ATE) and Relative Pose Error (RPE) comparison of Simple Weighted Averaging estimator (SWA) and Legged Odometry (LO) for walking experiment. \looseness=-1}
\centering 
\begin{tabular}{|c c c c c c|} 
\hline
{Filter}	& \multicolumn{5}{c|}{\emph{Walking} $1\si{\meter}$}  \\
\hline
& \multicolumn{3}{c}{ATE} & \multicolumn{2}{c|}{RPE} \\
\hline
& rot [$\si{\degree}$] & pos [$\si{\meter}$] & vel [$\si{\meter/\second}$] & rot [$\si{\degree}$] & pos [$\si{\meter}$]  \\
\hline
{SWA} & \textbf{2.8897} & 0.0202 &  \textbf{0.1087} & \textbf{2.8360} & \textbf{0.0244}  \\
{LO} & 2.9614 & 0.0203 & 0.1373 & 2.8925 & 0.0272   \\
\hline
\end{tabular} 
\label{table:swa-errors} 
\end{table}

For a quantitative comparison of the legged odometry and the SWA estimators, we use the Absolute Trajectory Error (ATE) and the Relative Pose Error (RPE) as the error metrics reported in Table \ref{table:swa-errors}.
While the former is useful in evaluating the overall performance of the estimator, the latter is useful for understanding drifts in the estimates.
A detailed description of these error metrics is discussed in Section \ref{sec:chap:diligent-kio-error-metrics}.
It can be seen that for this particular experiment of forward-walking, the SWA estimator performs better than the legged odometry.
Although the percentage difference in errors does not seem very significant, this nevertheless demonstrates the SWA estimator to be a useful fusion-based alternative for the simple kinematic-based odometry which might not be reliable in the presence of modeling errors.
In particular, the orientation and velocity estimates, which are the most relevant feedback that needs to be passed onto the controllers at higher rates, seem to suffer from slightly lesser errors than the legged odometry.
Further, for this experimental evaluation, it must be noted that we do not exploit the use of IMUs on the feet and the contact wrench decomposition to improve the velocity estimation.\looseness=-1





\section{Conclusion}
This chapter presented a simple estimator for floating base estimation of a humanoid robot within a cascading framework of decoupled base orientation estimation and position-velocity estimation.
Such simple estimators, while being easy to prototype and being more intuitive to analyze in real-world experiments, also allow for rigorous representations of geometric objects like pose and rotations in a straightforward manner.
The decoupled estimation enables the flexibility of employing better non-linear observers and linear estimators independently from each other.
Several improvements can be already suggested from the current implementation of the SWA estimator. 
Firstly, the Bayesian fusion approach discussed in Section \ref{sec:chap:loosely-coupled=bayesian-fusion} could be used instead of the deterministic averaging approach to account for uncertainties in the estimates explicitly.
Then, a QP-based alternative can be used for the base velocity estimation to account for model constraints.
Finally, a precise contact detection strategy would significantly help such simple estimators to become a consistent choice for quick prototypes for relevant application scenarios.
In this context, the incorporation of the feet IMUs which can suitably robustify the base velocity estimation and the contact detection needs to be analyzed with appropriate experimental evaluations.\looseness=-1

It must be noted that the performance of the SWA estimator can be considerably increased by properly tuning the hyper-parameters for the rotation averaging, the weights for the averaging, and the fusion of the base velocity.
Further, relying on a Quadratic Programming (QP) solver instead of the regularized weighted pseudo inverse for the fusion of base velocity, one may be able to account for joint velocity and position limits as constraints within the optimization problem, which might significantly improve the base velocity estimates from the SWA estimator method. \looseness=-1

A flat-floor assumption was used to reduce the drifts in $z$-position of the base link.
It must be noted that the flat floor assumption can be relaxed if the terrain information is known, which can usually be constructed using probabilistic grid maps or Gaussian process regression using the contact locations obtained from the kinematics or using 3D point cloud information from exteroceptive sensors.
However, such mapping again is tightly coupled with reliable base estimation.
A possible idea could be to rely on a robot-centric base estimator which suffers from low drift and can enable a reliable local terrain mapping (\cite{fankhauser2018probabilistic}) that could be in turn used within a dual loop framework to correct the world-centric estimates. \looseness=-1

Alternately, in the context of absolute humanoid localization, we have presented a na\"ive application of localization in a known environment to demonstrate the notion of averaging over Lie groups, however, a more practical approach would require removal of the assumption of a known environment and estimate the absolute robot pose using only relative poses using the special Euclidean synchronization problem. \looseness=-1

\chapter{Floating Base Estimation for Humanoid Robots using Filtering on Matrix Lie Groups}
\label{chap:floating-base-diligent-kio}

\begin{figure}
\centering
\includegraphics[width=\textwidth]{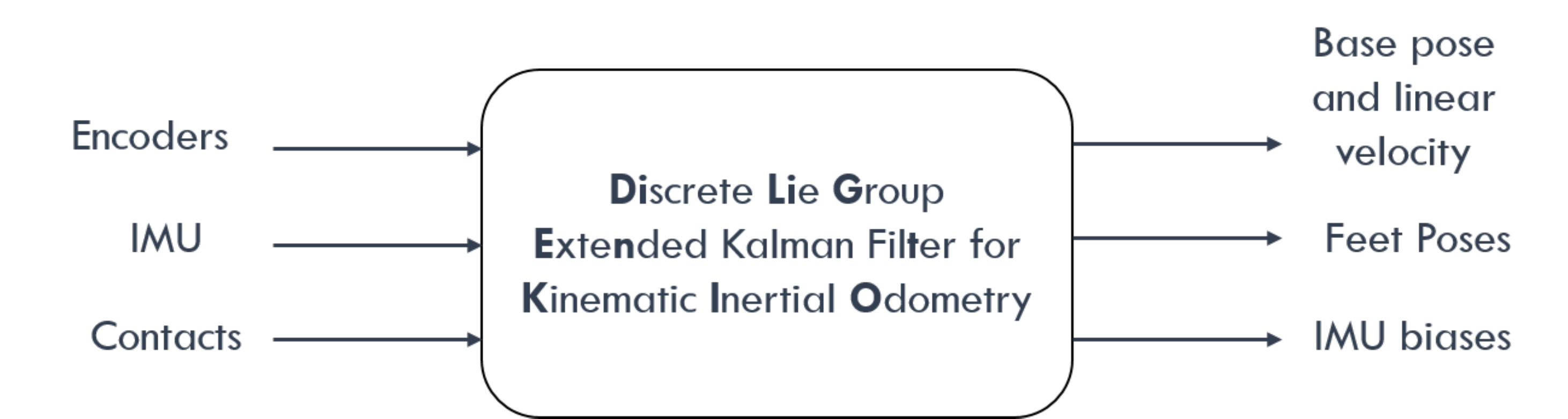}
\caption{High level block diagram for DILIGENT-KIO.}
\label{fig:diligent-blk-diag}
\end{figure}

It was understood that a loosely coupled sensor fusion approach combines pre-processed estimate from each measurement source independently to obtain a fused estimate while a tightly-coupled sensor fusion approach combines the measurements within a single process.
Although loosely-coupled fusion approaches may be easy to implement, tightly-coupled approaches directly combine raw measurements to produce estimates that might not be affected by any information loss.
This chapter presents a tightly-coupled sensor fusion approach using filtering over matrix Lie groups.
Section \ref{sec:chapter05:diligent-kio} presents a proprioceptive estimator for humanoid robots (see Figure \ref{fig:diligent-blk-diag}) using filtering over matrix Lie groups exploiting the theoretical background discussed in Section \ref{sec:chapter02:ekf-matrix-lie-groups}.
The extension of the proposed estimator to absolute humanoid localization is presented in Section \ref{sec:chap:diligent-kio:humanoid-localization}.
Section \ref{sec:chap:diligent-kio:alternate-formulations} describes a few variants for the formulations of the proposed approach.

\section{DILIGENT-KIO: Discrete Lie Group Extended Kalman Filter for Kinematic Inertial Odometry}
\label{sec:chapter05:diligent-kio}

In this section, we follow the development of a Discrete Lie Group Extended Kalman Filter for Kinematic Inertial Odometry, DILIGENT-KIO in short, for tackling the proprioceptive floating base estimation problem with the consideration that both the state and observations evolve over distinct matrix Lie groups.
DILIGENT-KIO is designed for the estimation of the humanoid robot's floating base and feet poses along with the biases for the IMU sensor rigidly mounted on the base link, all encapsulated within a matrix Lie group representation. 
The measurement updates for the estimator are provided as relative pose measurements of the foot in contact with respect to the base link, evolving over the group of rigid body transformations, obtained from the forward kinematics of the robot.
The proposed modeling choice leads to an improved uncertainty handling, leading to an estimator with good convergence properties.
In a sense, this estimator is formulated using the theory of EKF over matrix Lie groups using concentrated Gaussian distributions (as discussed in Section \ref{sec:chapter02:ekf-matrix-lie-groups}) to extend the application scenarios of~\cite{hartley2020contact} to flat contact surfaces while retaining \emph{fast} estimation convergence. Table \ref{table:chap:diligent-kio-soa} shows a comparison of the proposed estimator design with respect to the state-of-the-art methods.

\begin{table}[t]
\caption{Comparison of DILIGENT-KIO with the state-of-the-art}
\centering
\scalebox{0.8}{
\begin{tabular}{|p{0.3\textwidth} |p{0.2\textwidth} | p{0.2\textwidth} | p{0.1\textwidth} | p{0.1\textwidth}| }
\hline
Author, Year  &    State    &   Kinematic measurement     &   Support for flat contact surfaces    &    Fast Convergence        \\
\hline
\cite{rotella2014state} (OCEKF)   &  Euclidean + Unit quaternion      &    Euclidean + Unit quaternion    &   \checkmark    &    \ding{55}      \\
\cite{hartley2020contact} (InvEKF)  &    Matrix Lie group    &   Euclidean     &  \ding{55}     &    \checkmark        \\
\cite{qin2020novel}  &   Matrix Lie group      &   Euclidean     &   \checkmark    &            \checkmark \\
\cite{ramadoss2021diligent} (DILIGENT-KIO) &   Matrix Lie group      &   Matrix Lie group     &   \checkmark    &            \checkmark \\
\hline
\end{tabular}}
\label{table:chap:diligent-kio-soa}
\end{table}

\subsection{Matrix Lie Group Representation for State Space}

\begin{remark}
It must be noted that the short-hand notation introduced here remains specific for this chapter for the sake of readability of the forthcoming filter derivations.
\end{remark}

For floating base estimation of the humanoid robot, we wish to estimate the following quantities, \looseness=-1
\begin{itemize}
    \item the position $\Pos{A}{B}$ (or shorthand $\mathbf{p}$) of the base link in the inertial frame,
    \item the orientation $\Rot{A}{B}$ (or shorthand $\mathbf{R}$) of the base link in the inertial frame, and
    \item the linear velocity $\oDot{A}{B}$ (or shorthand $\mathbf{v}$) of the base link in the inertial frame.
\end{itemize}
The angular velocity of the base link can be directly obtained from the IMU rigidly attached to the base link of the robot. 
So we do not include it in the set of estimated quantities.
Additionally, we are also interested in estimating,
\begin{itemize}
    \item the position $\Pos{A}{F}$ (or shorthand $\mathbf{d}_F$) of the foot link in the inertial frame, and
    \item the orientation $\Rot{A}{F}$ (or shorthand $\mathbf{Z}_F$) of the foot link in the inertial frame,
\end{itemize}
where, $F = \{ {LF}, {RF} \}$ is the set containing left foot and right foot frames.
This augmentation of the state with the feet poses constrains the estimation problem for better accuracy.
In addition, a fused estimate of the foot pose can be useful to re-plan contact placements in the event of external perturbations. 
Since we also rely on the base-collocated IMU for the state estimation, it is necessary to also estimate its slowly time-varying biases,
\begin{itemize}
    \item bias $\biasAcc$ affecting the accelerometer measurement, and
    \item bias $\biasGyro$ affecting the gyroscope measurement.
\end{itemize}
The consideration of the IMU biases within the estimator design extends its practical usability for real-world hardware applications. 
These biases are always expressed in the local IMU frame. \looseness=-1

The following property of non-interacting manifolds is useful in constructing the matrix Lie group for the representation of the state space where we can establish an innate coupling between some state variables and a distinct decoupling between a few state variables.
\begin{property}[Non-Interacting Manifolds, (\cite{sola2018micro}, Section IV)]
\label{prop:chap:non-interacting-manif}
A composite state-space obtained through direct products results in non-interacting manifolds and corresponding non-interacting operators.
\end{property}

The tuple $ \X_B =  (\mathbf{p}, \mathbf{R}, \mathbf{v})$ represents the base link quantities, while the feet quantities are represented as $\X_F = (\mathbf{d}_F, \mathbf{Z}_F)$.
The biases are sometimes included in a single vector as $\bias = \text{vec}(\biasAcc, \biasGyro)$ within the scope of this section.
Using the Property \ref{prop:chap:non-interacting-manif}, the quantities needed to be estimated can be encapsulated within a matrix Lie group. \looseness=-1

\begin{proposition}[DILIGENT-KIO State Representation]
\label{proposition:chap:diligent-kio-state-repr}
A matrix Lie group embedded in the space of $\SEk{2}{3} \times \SE{3}^2 \times \T{6} \subset \R^{20 \times 20}$ is used to represent the state space $\mathscr{M}$ constituting the extended base pose, the left and the right foot poses and the IMU biases, respectively.
\end{proposition}

The Proposition \ref{proposition:chap:diligent-kio-state-repr} is valid owing to the property that the product of Lie groups is a Lie group (\cite{bourmaud2015continuous}).
It must be noted that this state representation follows a combination of \emph{direct} and \emph{semi-direct} products of Lie groups.
A well-known example of such products to two distinct combination of rotations and translations is given below,
\begin{align*}
& \SO{3} \times \R^3: (\mathbf{p}_1, {\Rot{}{}}_1) \circ (\mathbf{p}_2, {\Rot{}{}}_2) = (\mathbf{p}_1 + \mathbf{p}_2,\ {\Rot{}{}}_1\ {\Rot{}{}}_2) \ \text{(direct)}, \\
& \SE{3}: (\mathbf{p}_1, {\Rot{}{}}_1) \circ (\mathbf{p}_2, {\Rot{}{}}_2) = ({\Rot{}{}}_1\ \mathbf{p}_2 + \mathbf{p}_1,\ {\Rot{}{}}_1\ {\Rot{}{}}_2) \ \text{(semi-direct)}.
\end{align*}

A matrix Lie group element from the state space $\mathscr{M}$ is then defined as, 
\setcounter{MaxMatrixCols}{20}
\begin{align}
\label{eq:chap:diligent-kio-state}
\X &=
\begin{bmatrix}
\Rot{}{} & \mathbf{p} & \mathbf{v} & \Zero{3}  & \Zeros{3}{1} & \Zero{3} & \Zeros{3}{1} & \Zeros{3}{6} &  \Zeros{3}{1} \\
\Zeros{1}{3} & 1 & 0 & \Zeros{1}{3} & 0  & \Zeros{1}{3}  & 0  & \Zeros{1}{6}  &  0  \\
\Zeros{1}{3} & 0 & 1 & \Zeros{1}{3} & 0  & \Zeros{1}{3}  & 0  & \Zeros{1}{6}  &  0   \\
\Zero{3} & \Zeros{3}{1}  & \Zeros{3}{1}  & \mathbf{Z}_{LF} & \mathbf{d}_{LF} & \Zero{3}  & \Zeros{3}{1} & \Zeros{3}{6} & \Zeros{3}{1}  \\
\Zeros{1}{3} & 0 & 0 & \Zeros{1}{3} & 1  & \Zeros{1}{3}  & 0  & \Zeros{1}{6}  &  0 \\
\Zero{3} & \Zeros{3}{1} & \Zeros{3}{1}  & \Zero{3}  & \Zeros{3}{1}  & \mathbf{Z}_{RF} & \mathbf{d}_{RF} &\Zeros{3}{6} & \Zeros{3}{1} \\
\Zeros{1}{3} & 0 & 0 & \Zeros{1}{3} & 0  & \Zeros{1}{3}  & 1  & \Zeros{1}{6}  &  0 \\
\Zeros{6}{3} & \Zeros{6}{1}  &  \Zeros{6}{1} & \Zeros{6}{3} &  \Zeros{6}{1} &  \Zeros{6}{3} & \Zeros{6}{1}  & \I{6} & \bias  \\
\Zeros{1}{3} & 0 & 0 & \Zeros{1}{3} & 0  & \Zeros{1}{3}  & 0  & \Zeros{1}{6}  &  1 
\end{bmatrix} \in \mathscr{M} \subset \R^{20 \times 20}.
\end{align}

For convenience, the following tuple representation can be used as a short-hand notation for denoting an element from the state space,
$$
\X = (\X_B, \X_{LF}, \X_{RF}, \bias) = \left(\mathbf{p}, \mathbf{R}, \mathbf{v}, \mathbf{d}_{LF}, \mathbf{Z}_{LF}, \mathbf{d}_{RF}, \mathbf{Z}_{RF}, \bias \right)_{\mathscr{M}}.
$$
Although the group element $\X$ seems to be a matrix of high dimensions, the dense elements of zeros can be avoided to be stored computationally by exploiting the Property \ref{prop:chap:non-interacting-manif}.

Since the state space is a composite manifold, block operations can be performed because each of the sub-manifold is non-interacting by the implications of the direct product.
Exploiting this property allows us to define the Lie algebra and derive the exponential, logarithm, and Jacobians of the considered matrix Lie group.
The operator definitions for each sub-manifold is provided in Appendix \ref{appendix:chapter02:examples-matrix-lie-groups}. \looseness=-1

The hat operator $\ghat{\mathscr{M}}{\epsilon}$ mapping the vectors $\epsilon \in \R^{27}$ to the Lie algebra $\mathfrak{m}$ becomes, 

{
\centering
\scalebox{0.97}{\parbox{\textwidth}{
\begin{align}
\label{eq:chap:diligent-kio-state-lie-algebra}
\ghat{\mathscr{M}}{\epsilon} &=
			\begin{bmatrix}
			S(\epsilon_{\Rot{}{}}) & \epsilon_\mathbf{p} & \epsilon_\mathbf{v} & \Zero{3}  & \Zeros{3}{1} & \Zero{3} & \Zeros{3}{1} & \Zeros{3}{6} &  \Zeros{3}{1} \\
			\Zeros{1}{3} & 0 & 0 & \Zeros{1}{3} & 0  & \Zeros{1}{3}  & 0  & \Zeros{1}{6}  &  0  \\
			\Zeros{1}{3} & 0 & 0 & \Zeros{1}{3} & 0  & \Zeros{1}{3}  & 0  & \Zeros{1}{6}  &  0   \\
			\Zero{3} & \Zeros{3}{1}  & \Zeros{3}{1}  & S(\epsilon_{\mathbf{Z}_{LF}}) & \epsilon_{\mathbf{d}_{LF}} & \Zero{3}  & \Zeros{3}{1} & \Zeros{3}{6} & \Zeros{3}{1}  \\
			\Zeros{1}{3} & 0 & 0 & \Zeros{1}{3} & 0  & \Zeros{1}{3}  & 0  & \Zeros{1}{6}  &  0 \\
			\Zero{3} & \Zeros{3}{1} & \Zeros{3}{1}  & \Zero{3}  & \Zeros{3}{1}  & S(\epsilon_{\mathbf{Z}_{RF}}) & \epsilon_{\mathbf{d}_{RF}} &\Zeros{3}{6} & \Zeros{3}{1} \\
			\Zeros{1}{3} & 0 & 0 & \Zeros{1}{3} & 0  & \Zeros{1}{3}  & 0  & \Zeros{1}{6}  &  0 \\
			\Zeros{6}{3} & \Zeros{6}{1}  &  \Zeros{6}{1} & \Zeros{6}{3} &  \Zeros{6}{1} &  \Zeros{6}{3} & \Zeros{6}{1}  & \Zero{6} & \epsilon_{\mathbf{b}}  \\
			\Zeros{1}{3} & 0 & 0 & \Zeros{1}{3} & 0  & \Zeros{1}{3}  & 0  & \Zeros{1}{6}  &  0 
			\end{bmatrix} \in \mathfrak{m} \subset \R^{20 \times 20},
\end{align}
}}
}

\noindent with the vector $\epsilon \in \R^{27}$ defined as,
\begin{align}
\label{eq:chap:diligent-kio-state-vector}
\epsilon = \text{vec}(\epsilon_B, \epsilon_{LF}, \epsilon_{RF}, \epsilon_\bias)  = \text{vec}\left(\epsilon_\mathbf{p},\ \epsilon_\mathbf{R},\ \epsilon_\mathbf{v},\ \epsilon_{\mathbf{d}_{LF}},\ \epsilon_{\mathbf{Z}_{LF}},\ \epsilon_{\mathbf{d}_{RF}},\ \epsilon_{\mathbf{Z}_{RF}}, \epsilon_\bias \right).
\end{align}

The exponential mapping of the state space expressed in a tuple representation is given by, \looseness=-1
\begin{equation}
\label{eq:chap:diligent-kio-state-exp-map}
\begin{split}
\gexphat{\mathscr{M}}\left(\epsilon\right) &= \big(\J\left(\epsilon_\mathbf{R}\right)\ \epsilon_\mathbf{p},\ \Exp\left(\epsilon_\mathbf{R}\right),\ \J\left(\epsilon_\mathbf{R}\right)\ \epsilon_\mathbf{v}, \\
& \quad \ \ \ \J\left(\epsilon_{\mathbf{Z}_{LF}}\right)\ \epsilon_{\mathbf{d}_{LF}},\ 
\Exp\left(\epsilon_{\mathbf{Z}_{LF}}\right),\\
& \quad \ \ \ \J\left(\epsilon_{\mathbf{Z}_{RF}}\right)\ \epsilon_{\mathbf{d}_{RF}},\ 
\Exp\left(\epsilon_{\mathbf{Z}_{RF}}\right),\ \epsilon_{\mathbf{b}}\big)_{\mathscr{M}}.
\end{split}
\end{equation}
where, we have used $\Exp(.) \triangleq\ \gexphat{\SO{3}}(.)$ to denote the exponential map of $\SO{3}$ and $\J(.) \triangleq \gljac{\SO{3}}(.)$ for the left Jacobian of $\SO{3}$ which arises as a consequence of the series expansion of the exponential map while considering the semi-direct products in $\SEk{2}{3}$ and $\SE{3}$.

The adjoint matrix of the considered state space $\mathscr{M}$ obtained by applying Eq. \eqref{eq:chap02-liegroups-adjoint-repr} is given as,
\begin{align}
\label{eq:chap:diligent-kio-state-adj}
\gadj{\X} =
\begin{split}\begin{bmatrix}
\Rot{}{} & S(\mathbf{p}) \Rot{}{} & \Zero{3}  & \Zero{3} & \Zero{3} & \Zero{3} & \Zero{3} & \Zero{6} \\
\Zero{3} & \Rot{}{} & \Zero{3}  & \Zero{3} & \Zero{3} & \Zero{3} & \Zero{3} & \Zero{6} \\
\Zero{3} &  S(\mathbf{v}) \Rot{}{} & \Rot{}{}  & \Zero{3} & \Zero{3} & \Zero{3} & \Zero{3} & \Zero{6} \\
\Zero{3} &  \Zero{3} & \Zero{3}  &  \mathbf{Z}_{LF} &   S(\mathbf{d}_{LF})\mathbf{Z}_{LF} & \Zero{3} & \Zero{3} & \Zero{6} \\
\Zero{3} &  \Zero{3} & \Zero{3}  & \Zero{3} &  \mathbf{Z}_{LF} & \Zero{3} & \Zero{3} & \Zero{6} \\
\Zero{3} &  \Zero{3} & \Zero{3}  & \Zero{3} & \Zero{3} & \mathbf{Z}_{RF} & S(\mathbf{d}_{RF})\mathbf{Z}_{RF} & \Zero{6} \\
\Zero{3} &  \Zero{3} & \Zero{3}  & \Zero{3} & \Zero{3} & \Zero{3} & \mathbf{Z}_{RF} & \Zero{6} \\
\Zero{3} &  \Zero{3} & \Zero{3}  & \Zero{3} & \Zero{3} & \Zero{3} & \Zero{3} & \I{6} 
\end{bmatrix} \in \R^{27 \times 27}.
\end{split}
\end{align}
The structure of the adjoint matrix depends on the serialization of the vector chosen for the tangent space.
Here, we have chosen a \emph{linear-angular} serialization for the twist vectors of $\SE{3}$, differently from the choice of \emph{angular-linear} serialization.

The left Jacobian of the matrix Lie group $\mathscr{M}$ is obtained as a block-diagonal form of the left Jacobians of the constituting matrix Lie groups given as,
\begin{align}
\label{eq:chap:diligent-kio-state-ljac}
\begin{split}
\gljac{\mathscr{M}}(\epsilon) = \text{blkdiag} \left( \gljac{\SEk{2}{3}}\left(\epsilon_B\right), \; \gljac{\SE{3}}\left(\epsilon_{LF}\right),  \gljac{\SE{3}}\left(\epsilon_{RF}\right),\ \I{6} \right)  \in \R^{27 \times 27}.
\end{split}
\end{align}
It can be seen that the left Jacobian for the translation group of IMU biases is the identity matrix, since the underlying representation of the translation group is the Euclidean vector space.
The right Jacobian of the matrix Lie group can be constructed in a similar fashion. \looseness=-1 

\subsection{Discrete System Dynamics}
\label{sec:chap:diligent-kio-sys-dyn}
In this subsection, we develop the process model to predict the evolution of the system states using a discrete-time system dynamics model.

\begin{assumption}[Collocation of the IMU on the base link]
\label{assumption:chap:diligent-kio-imu-collocation}
An inertial measurement unit (IMU) is rigidly attached to the base link of the robot.
\end{assumption}
\begin{assumption}[Coincidence of the IMU and the base link coordinate systems]
\label{assumption:chap:diligent-kio-imu-coincide}
For simplicity, the coordinate frames of the base link and the IMU coincide.
\end{assumption}

If Assumption \ref{assumption:chap:diligent-kio-imu-coincide} is not valid, then proper care must be taken to perform the coordinate transformations that are required to express the quantities from the IMU sensor frame $S$ in the base link frame $B$.
Usually, it is convenient to handle all the estimator computations in the IMU sensor frame $S$ and then transform the estimated quantities to the base link, whenever necessary.
With this approach, the effect of the Coriolis terms while trying to transform the accelerometer measurements into the base link can be avoided.
This could occur in cases where the IMU frame is placed far away from the base link. \looseness=-1


A commonly used motion model for a rigid body equipped with an IMU sensor is the strap-down IMU-based rigid body kinematics model. 
This model uses accelerometer measurements $\yAcc{A}{B}$ and the gyroscope measurements $\yGyro{A}{B}$ as exogenous inputs.
Both the accelerometer measurements and the gyroscope measurements are modeled to be affected by slowly time-varying biases $\biasAcc$ and $\biasGyro$, respectively, along with additive white Gaussian noise $\noiseAcc{B}$ and $\noiseGyro{B}$, respectively, as seen in Eq. \eqref{eq:chap:diligent-kio-imu-sensor-model},
\begin{align}
\label{eq:chap:diligent-kio-imu-sensor-model}
\begin{split}
&\yAcc{A}{B} = {\Rot{}{}}^T \left(\oDoubleDot{A}{B} - \gravity{A} \right) + \biasAcc + \noiseAcc{B}, \\
&\yGyro{A}{B} = \omegaLeftTriv{A}{B} + \biasGyro + \noiseGyro{B}.
\end{split}
\end{align}
Ideally, this model tracks the body orientation $\Rot{}{}$ with the help of the gyroscope measurements, while an integration on the body acceleration $\oDoubleDot{A}{B}$  obtained by compensating gravity terms from the accelerometer measurements is used to track the evolution of the body position and linear velocity.
However, a reasonable estimate of the extended base pose cannot be obtained simply by using the accelerometer and gyroscope measurements, due to the noisy nature of the observations, inherent biases, and limits on the sampling rates of the sensor. 
It is required to complement this model with other sources of information to obtain a reliable estimate.

\begin{assumption}[Rigid contacts of the feet]
\label{assumption:chap:diligent-kio-rigid-foot-contact}
The feet of the robot always makes a rigid contact with the environment.
\end{assumption}

With the Assumption \ref{assumption:chap:diligent-kio-rigid-foot-contact}, holonomic constraints are imposed whenever the foot makes contact with the environment causing the foot velocity with respect to the inertial frame to be zero, considered to be affected by additive white Gaussian noise (see Eq. \eqref{eq:chap:loosely-coupled:null-feet-vel} to understand how base velocity computation can be achieved using null foot velocities).
When this noise is expressed in the inertial frame, \cite{rotella2014state} characterizes it as the translational or rotational slippage of the foot.
In our case, we will express the noises $\noiseLinVel{F}$ and $\noiseAngVel{F}$ affecting null foot velocities, with respect to the inertial frame, locally in the foot frame to be coherent with the matrix Lie group formulation.
A constant motion model is considered for the foot pose to respect the contact events. 
But, this motion model becomes invalid when the foot is in the swing phase. 
Thus, the contact states are assumed to be known and passed as inputs to the system and when a foot is not in contact, the variances related to the foot velocities are dynamically scaled to very high values causing the estimated foot pose from the prediction model to grow uncertain.
The measurement updates are then relied upon to update the foot pose causing it to reset to a more reliable estimate, whenever a contact is made.\looseness=-1

\begin{remark}
The absence of contacts is not handled rigorously within this formulation of the system dynamics, however, it can be tackled straightforwardly by passing the relative pose measurements obtained from the forward kinematics as inputs to the process model.
Alternately, one may also attach IMUs on the feet of the robot to improve foot orientation tracking. \looseness=-1
\end{remark}

The overall discrete system dynamics can be written in a form that reflect the motion integration equations of a discrete dynamical system on matrix Lie groups as defined in Eq. \eqref{eq:chap02:ekf-lie-group-system}, 
$$\X_\knext =  \X_\kcurr \; \gexphat{G}\left(\Omega\left(\X_\kcurr, \mathbf{u}_\kcurr\right) + \mathbf{w}_\kcurr\right).$$ 

One can choose a left trivialized motion model as a function of IMU measurements to reflect the group velocity of the base state and null velocities for the state of the foot and IMU biases.
The left trivialized motion model is written as, \looseness=-1
\begin{align}
	\begin{split}
		\label{eq:chap:diligent-kio-left-triv-motion-model}
		\Omega = \begin{bmatrix}
			{\Rot{}{} ^T}{\mathbf{v}} \Delta T +\ \frac{1}{2}\ \accBar{A}{B} \Delta T^2 \\
			\yGyroBar{A}{B}\ \Delta T \\ 
			\accBar{A}{B}\ \Delta T  \\ 
			\Zeros{18}{1}
		\end{bmatrix} \in \mathbb{R}^{27},
	\end{split}
\end{align}
\noindent where, we have used  $\accBar{A}{B}_{\kcurr} = \yAcc{A}{B}{_\kcurr} -\ \biasAcc{_\kcurr} +\ \Rot{}{}_{\kcurr}^T\ \gravity{A}$ to denote the computed linear acceleration of the base with respect to the inertial frame expressed in the base frame.
The computed linear acceleration is obtained directly from the unbiased accelerometer measurement after compensating for the gravity in the local frame.
Further, we have used $\yGyroBar{A}{B}_\kcurr = \yGyro{A}{B}_\kcurr -\ \biasGyro$ to denote the unbiased angular velocity measurement from the gyroscope expressed in the base frame. 
The left trivialized noise vector then becomes,
\begin{align}
	\begin{split}
		\label{eq:chap:diligent-kio-left-triv-motion-noise}
		\mathbf{w} = \text{vec}\left(-0.5\noiseAcc{B}\Delta T,\ -\noiseGyro{B},\ -\noiseAcc{B},\ -\noiseLinVel{LF},\ -\noiseAngVel{LF},\ -\noiseLinVel{RF},\ -\noiseAngVel{RF},\ \noiseBias{B} \right)\Delta T \in \mathbb{R}^{27},
	\end{split}
\end{align}
with the prediction noise covariance matrix $\mathbf{Q} = \expectation{\mathbf{w} \mathbf{w}^T}$.

The discrete system dynamics is obtained by assuming a zero-order hold on the inputs with a sampling period $\Delta T$ and the choice of left trivialized motion model in Eq. \eqref{eq:chap:diligent-kio-left-triv-motion-model} leads to the following evolution equations, \looseness=-1
\begin{align}
	\label{eq:chap:diligent-kio-sys-dynamics-full}
	\begin{split}
		\mathbf{p}_{\knext} =&\ \mathbf{p}_{\kcurr} + \ \Rot{}{}_{\kcurr}\ \mathbf{J}_b \left( \Rot{}{}_{\kcurr}^T \ \mathbf{v}_{\kcurr} \ \Delta T \ +  \frac{1}{2}\  \left(\accBar{A}{B}_\kcurr\ -\ \noiseAcc{B} \right) \ \Delta T^2 \right),\\
		\Rot{}{}_\knext =&\ \Rot{}{}_\kcurr\ \Exp\left(\left(\yGyroBar{A}{B}\ -\ \noiseGyro{B} \right)\ \Delta T\right), \\
		\mathbf{v}_{\knext} =& \; \mathbf{v}_{\kcurr} + \; \Rot{}{}_{\kcurr}\ \mathbf{J}_b \left(\left(\accBar{A}{B}_\kcurr\ -\ \noiseAcc{B} \right)\  \Delta T \right),\\
		\mathbf{d}_{F_{\knext}} =&\ \mathbf{d}_{F_{\kcurr}} +\ \mathbf{Z}_{F_{\kcurr}}\ (-\noiseLinVel{F} )\; \Delta T, \\
		\mathbf{Z}_{F_{\knext}} =&\ \mathbf{Z}_{F_{\kcurr}} \ \Exp\left(-\noiseAngVel{F} \; \Delta T\right), \\
		\bias_\knext =&\ \bias_\kcurr + \noiseBias{B}\ \Delta T,
	\end{split}
\end{align}
where, the Jacobian $\J_b = \gljac{\SO{3}}\left((\yGyroBar{A}{B}\ -\ \noiseGyro{B})\Delta T \right)$ acts on the evolution of base state quantities due to the natural definition of the exponential map of $\SEk{2}{3}$. 
It must be noted that there is no Jacobian present in the evolution equations of the foot state even though the exponential map of $\SE{3}$ gives rise to it. 
In this case, the Jacobian is an identity matrix due to null foot velocities considered in the left-trivialized motion model.
Nevertheless, the Jacobian $\J_b$ also reduces to an identity matrix when the sampling period $\Delta t$ tends to zero.
Although, Eq. \eqref{eq:chap:diligent-kio-sys-dynamics-full} leads to more accurate motion integration equations, we consider an assumption that the sampling period $\Delta t$ will always be considerably small, rendering the Jacobian $\J_b \approx  \I{3}$, and thereby leading to the following evolution equations,
\begin{align}
\label{eq:chap:diligent-kio-sys-dynamics}
\begin{split}
\mathbf{p}_{\knext} =&\ \mathbf{p}_{\kcurr} + \ \Rot{}{}_{\kcurr}\ \left( \Rot{}{}_{\kcurr}^T \ \mathbf{v}_{\kcurr} \ \Delta T \ +  \frac{1}{2}\  \left(\accBar{A}{B}_\kcurr\ -\ \noiseAcc{B} \right) \ \Delta T^2 \right),\\
\Rot{}{}_\knext =&\ \Rot{}{}_\kcurr\ \Exp\left(\left(\yGyroBar{A}{B}\ -\ \noiseGyro{B} \right)\ \Delta T\right), \\
\mathbf{v}_{\knext} =& \; \mathbf{v}_{\kcurr} + \; \Rot{}{}_{\kcurr} \left(\left(\accBar{A}{B}_\kcurr\ -\ \noiseAcc{B} \right)\  \Delta T \right),\\
\mathbf{d}_{F_{\knext}} =&\ \mathbf{d}_{F_{\kcurr}} +\ \mathbf{Z}_{F_{\kcurr}}\ (-\noiseLinVel{F} )\; \Delta T, \\
\mathbf{Z}_{F_{\knext}} =&\ \mathbf{Z}_{F_{\kcurr}} \ \Exp\left(-\noiseAngVel{F} \; \Delta T\right), \\
\bias_\knext =&\ \bias_\kcurr + \noiseBias{B}\ \Delta T.
\end{split}
\end{align}

\begin{remark}
The underlying algorithm of DILIGENT-KIO is the discrete EKF on matrix Lie groups with the left-invariant error formulation of type 1 as described in the Algorithm \ref{algo:chap02:dlgekf}. \looseness=-1
\end{remark}

In order to propagate the system state covariance, we need to compute the linearized error propagation matrix  $\F_\kcurr$  which requires the derivation of the left-trivialized motion model Jacobian  $\mathfrak{F}_\kcurr$.
This quantity $\mathfrak{F}_\kcurr$ along with the design covariance matrix $\mathbf{Q}_k$ and the adjoint and Jacobian matrices of the state-space Lie group $\mathscr{M}$ can be used to compute the linearized error propagation matrix  $\F_\kcurr$. \looseness=-1

\subsubsection{Left Trivialized Motion Model Jacobian}
The Jacobian of the left-trivialized motion model $\hat{\Omega}$ needs to be computed at the current state estimate with an infinitesimal additive perturbation in the vector space.
By inducing the perturbation $\err_\kprior$ on the current state estimate $\Xhat_\kprior$ and exploiting the definition of the concentrated Gaussian distribution (see Section \ref{sec:chapter02:uncertainty-matrix-lie-groups}), a first order approximation of the exponential map and the left Jacobian of $\SO{3}$ $\left(\Exp(\epsilon) \approx \I{3} + S(\epsilon),\ \J(\epsilon) \approx \I{3} + \frac{1}{2} S(\epsilon)\right)$ can be enforced to obtain the first-order approximation of the exponential map of the state-space group $\mathscr{M}$ in Eq. \eqref{eq:chap:diligent-kio-state-exp-map}. 
Neglecting the cross-product terms of infinitesimal perturbations leads to the perturbed state estimate in tuple representation as,
\begin{align}
\label{eq:chap:diligent-kio-perturbed-mean}
\begin{split}
\Xhat_\kprior\ \gexphat{\mathscr{M}}\left({\err_\kprior}\right) \approx& \big( \hat{\Rot{}{}}\epsilon_{\mathbf{p}} + \mathbf{\hat{p}}, \hat{\Rot{}{}} + \hat{\Rot{}{}}\ S(\epsilon_\Rot{}{}), \hat{\Rot{}{}}\epsilon_{\mathbf{v}} + \mathbf{\hat{v}} \\
&\ \mathbf{\hat{Z}}_{LF}\epsilon_{\mathbf{d}_{LF}} + \mathbf{\hat{d}}_{LF}, \mathbf{\hat{Z}}_{LF} + \mathbf{\hat{Z}}_{LF}\ S(\epsilon_{\mathbf{Z}_{LF}}) \\
&\ \mathbf{\hat{Z}}_{RF}\epsilon_{\mathbf{d}_{RF}} + \mathbf{\hat{d}}_{RF}, \mathbf{\hat{Z}}_{RF} + \mathbf{\hat{Z}}_{RF}\ S(\epsilon_{\mathbf{Z}_{RF}}), \biasHat + \epsilon_\bias
\big)_\mathscr{M}.
\end{split}
\end{align}

\begin{lemma}
\label{lemma:chap:diligent-kio:jacMotion}
The Jacobian $\mathfrak{F}_\kcurr$ of the left trivialized motion model $\hat{\Omega}$ with respect to the perturbation $\err$ is given as,
\begin{equation}
\label{eq:chap:diligent-kio-left-triv-motion-jacobian}
\begin{split}
\small
\mathfrak{F}_\kcurr &= \frac{\partial}{\partial \err} \Omega\left(\Xhat_\kprior\ \gexphat{\mathscr{M}}\left({\err_\kprior}\right), \mathbf{u}_\kcurr\right)_{\err_\kprior = 0} \\
&=
\begin{bmatrix}
	\Zero{3} & S(\Xi) & \I{3} \; \Delta T & \Zeros{3}{12} &  -\frac{1}{2}\I{3}\;\Delta T^2 & \Zero{3} \\
	\Zero{3} & \Zero{3} & \Zero{3} & \Zeros{3}{12} &  \Zero{3} & -\I{3}\;\Delta T \\
	\Zero{3} & S(\tilde{\mathbf{g}}) & \Zero{3} & \Zeros{3}{12} &  -\I{3}\;\Delta T & \Zero{3} \\
	\Zeros{18}{3} & \Zeros{18}{3} & \Zeros{18}{3} & \Zeros{18}{12} &  \Zeros{18}{3} & \Zeros{18}{3}
	\end{bmatrix} \in \R^{27 \times 27},
\end{split}
\end{equation}
where, we have used $ \Xi = {\hat{\Rot{}{}}^T_\kcurr \mathbf{\hat{v}}}_\kcurr \Delta T + \frac{1}{2}\; {\hat{\Rot{}{}}^T}_\kcurr \gravity{A} \; \Delta T^2 $ and $ \tilde{\mathbf{g}} = {\hat{\Rot{}{}}^T}_\kcurr \gravity{A} \; \Delta T$.
\end{lemma}


The proof of Lemma \ref{lemma:chap:diligent-kio:jacMotion} is provided in the Appendix \ref{appendix:chapter:left-triv-motion-model-jacobian-diligent-kio}.
The linearized error propagation matrix can be computed in a straight-forward manner  $\F_\kcurr\ = \gadj{\gexphat{\mathscr{M}}\left(-\hat{\Omega}_\kcurr\right)} + \grjac{\mathscr{M}}\left(\hat{\Omega}_\kcurr\right)\ \mathfrak{F}_\kcurr$.
At the end of the propagation step, we have the predicted state estimate $\Xhat_\kpred$ and state error covariance $\cov_\kpred$. \looseness=-1

\subsection{Measurement Model}
\label{sec:chap:diligent-kio-meas-model}
The measurement model producing the expected measurements as a function of the estimated state is constructed using the relative pose between the base link and the foot in contact with the environment.
The model observations are obtained by passing the joint encoder measurements through the forward kinematics map.
The encoder measurements $\encoders = \jointPos + \encoderNoise$ measure the joint positions $\jointPos$ affected by additive white Gaussian noise $\encoderNoise$.
The forward kinematics map $\text{FK}: \R^{\text{DOF}} \mapsto \SE{3}$ provide the relative pose measurements to construct the model observations $\Z^F_{\text{SS}}$ for single support evolving over $\SE{3}$ and $\Z^F_{\text{DS}}$ for double support evolving over $\SE{3} \times \SE{3}$.
\begin{equation}
\begin{split}
\label{eq:chap:diligent-kio-meas}
& \Z^F_{\text{SS}} =  \text{FK}(\tilde{s})  = \TransformMeasured{B}{F} \in \SE{3} \\
& \Z_{\text{DS}} = 
\begin{bmatrix}
     \text{FK}(\tilde{s}) & \Zero{4} \\ \Zero{4} &  \text{FK}(\tilde{s})
\end{bmatrix} =
\begin{bmatrix}
     \TransformMeasured{B}{LF} & \Zero{4} \\ \Zero{4} &  \TransformMeasured{B}{RF}
\end{bmatrix} \in \SE{3} \times \SE{3}.
\end{split}
\end{equation}

For the derivation of the measurement model Jacobian, we can focus only on the single support scenario, since the calculations extend also for the double support.
The measurement model can be written in the form of Eq. \eqref{eq:chap02:ekf-lie-group-meas} as $\TransformMeasured{B}{F} = \Transform{B}{F}\gexphat{\SE{3}}\left(\mathbf{n}_\text{FK}\right)$ where,
\begin{equation}
\label{eq:chap:diligent-kio-meas-model}
h^F_\text{SS}\left(\X\right)  = \Transform{B}{F} = \begin{bmatrix} {\Rot{}{}^T}{\mathbf{Z}_F} & {\Rot{}{}^T}({\mathbf{d}_F}\ -\ \mathbf{p})   \\
\Zeros{1}{3} & 1
 \end{bmatrix} \triangleq \left({\Rot{}{}^T}({\mathbf{d}_F}\ -\ \mathbf{p}),\  {\Rot{}{}^T}{\mathbf{Z}_F}\right)_{\SE{3}}.
\end{equation}
The inverse of the measurement model $h^{-1}\left(\X\right)$ is given as,
\begin{equation}
	\label{eq:chap:diligent-kio-meas-model}
	h^{-1}\left(\X\right)  = \Transform{F}{B} = \begin{bmatrix} {\mathbf{Z}_F^T}{\Rot{}{}} & {\mathbf{Z}_F^T}(\mathbf{p}\ -\ {\mathbf{d}_F})   \\
		\Zeros{1}{3} & 1
	\end{bmatrix} \triangleq \left({\mathbf{Z}_F^T}(\mathbf{p}\ -\ {\mathbf{d}_F}),\ {\mathbf{Z}_F^T}{\Rot{}{}}\right)_{\SE{3}}.
\end{equation}

The left-trivialized forward kinematic noise $\mathbf{n}_\text{FK} \triangleq \text{vec}\left(\fkNoiseLinVel{F}, \fkNoiseAngVel{F}\right) \in \R^6$, which is the relative foot velocity noise with respect to the base link, is related to the encoder noise $\encoderNoise$ through the left-trivialized relative Jacobian $\relativeJacobianLeftTriv{B}{F}$.
This can be shown either through differential kinematics or through the push-forward mapping (see Eq. \eqref{eq:chap02:liegroups-push-forward})  of $\SE{3}$, \looseness=-1
$$
\Transform{B}{F}\left(\encoders\right) = \Transform{B}{F}\left(\jointPos + \encoderNoise\right) = \Transform{B}{F}\left(\jointPos\right)\ \gexphat{\SE{3}}\left(\relativeJacobianLeftTriv{B}{F}(\jointPos)\ \encoderNoise\right).  $$
This leads to the measurement noise covariance $\mathbf{N} = \expectation{\mathbf{n}_\text{FK}\ \mathbf{n}_\text{FK}^T}$.

The innovation term $\mathbf{\tilde{z}}$ required to update the predicted state estimate using the FK measurements is computed using the logarithm mapping of $\SE{3}$.
In order to update the predicted covariance, we need to compute the measurement model error Jacobian $\mathbf{H}_\knext$.
This requires the computation of the term $\glogvee{\SE{3}}\left(h^{-1}\left(\Xhat_\kpred\right)\ h\left(\Xhat_\kpred\ \gexphat{G}\left(\err_\kpred\right)\right)\right)$.
The measurement model as a function of the predicted state estimate $\Xhat_\kpred$ induced with a small perturbation $\err_\kpred$ can be written in tuple representation obtained by substituting the first-order approximation from Eq. \eqref{eq:chap:diligent-kio-perturbed-mean} in the measurement model equation in Eq. \eqref{eq:chap:diligent-kio-meas-model} as,
\begin{align}
\begin{split}
h\left(\Xhat_\kpred\ \gexphat{\SE{3}}\left(\err_\kpred\right)\right) &=  \big(
\hat{\Rot{}{}}^T(\mathbf{\hat{d}}_F - \mathbf{\hat{p}}) + S\left(\hat{\Rot{}{}}^T(\mathbf{\hat{d}}_F - \mathbf{\hat{p}})\right)\err_\Rot{}{} - \err_\mathbf{p} + \hat{\Rot{}{}}^T\mathbf{\hat{Z}}_F\ \err_{\mathbf{d}_F}, \\
& \hat{\Rot{}{}}^T\mathbf{\hat{Z}}_F + \hat{\Rot{}{}}^T\mathbf{Z}_F\ S(\err_{\mathbf{Z}_F}) - S(\err_{\Rot{}{}})\ \hat{\Rot{}{}}^T\mathbf{\hat{Z}}_F \big)_{\SE{3}}.
\end{split}
\end{align}
The product term $h^{-1}\left(\Xhat_\kpred\right)\ h\left(\Xhat_\kpred\ \gexphat{G}\left(\err_\kpred\right)\right)$ can be computed as,
\begin{equation}
    \begin{bmatrix} 
\I{3} + S({\err}_{\mathbf{Z}_F})- S({\mathbf{\hat{Z}}_F^T}\;{\hat{\Rot{}{}}}\;{\err}_{\Rot{}{}}) &  {\err}_{\mathbf{d}_F} - {\mathbf{\hat{Z}}_F^T}\;{\hat{\Rot{}{}}}\;{\err}_{\mathbf{p}} + {\mathbf{\hat{Z}}_F^T}\; S({\hat{\Rot{}{}}}\;{\epsilon}_{\Rot{}{}})\; ({\mathbf{\hat{p}} - \mathbf{\hat{d}}_F}) \\
\Zeros{1}{3} & 1
\end{bmatrix}.
\end{equation}

\noindent Assuming that the perturbations are small, the matrix logarithm can be approximated as,
$$
\forall \X \in \mathbb{R}^{l \times l}, \; \text{log}( \X) \ \approx \X - \I{l}.
$$
This gives $\glog{\SE{3}}\left(h^{-1}\left(\Xhat_\kpred\right)\ h\left(\Xhat_\kpred\ \gexphat{G}\left(\err_\kpred\right)\right)\right)$ equal to,
$$
\begin{bmatrix} 
	S({\err}_{\mathbf{Z}_F})- S({\mathbf{\hat{Z}}_F^T}\;{\hat{\Rot{}{}}}\;{\err}_{\Rot{}{}}) &  {\err}_{\mathbf{d}_F} - {\mathbf{\hat{Z}}_F^T}\;{\hat{\Rot{}{}}}\;{\err}_{\mathbf{p}} + {\mathbf{\hat{Z}}_F^T}\; S({\hat{\Rot{}{}}}\;{\epsilon}_{\Rot{}{}})\; ({\mathbf{\hat{p}} - \mathbf{\hat{d}}_F}) \\
	\Zeros{1}{3} & 0
\end{bmatrix}.
$$

The translational part of the matrix can be rewritten using the identities $ \forall \mathbf{x}, \mathbf{y} \in \R^3, S(\mathbf{x}) \mathbf{y} = -S(\mathbf{y}) \mathbf{x}$ and the adjoint property of $\SO{3}$, $\forall \mathbf{u} \in \R^3, \Rot{}{} S(\mathbf{u}) \Rot{}{}^T = S(\Rot{}{} \mathbf{u})$ as follows,
\begin{align*}
	\begin{split}
		{\err}_{\mathbf{d}_F} - {\mathbf{\hat{Z}}_F^T}\;{\hat{\Rot{}{}}}\;{\err}_{\mathbf{p}} + {\mathbf{\hat{Z}}_F^T}\; S({\hat{\Rot{}{}}}\;{\epsilon}_{\Rot{}{}})\; ({\mathbf{\hat{p}} - \mathbf{\hat{d}}_F}) & = {\err}_{\mathbf{d}_F} - {\mathbf{\hat{Z}}_F^T}\;{\hat{\Rot{}{}}}\;{\err}_{\mathbf{p}} + {\mathbf{\hat{Z}}_F^T}\ \hat{\Rot{}{}}\; S({\epsilon}_{\Rot{}{}})\; {\hat{\Rot{}{}}}^T ({\mathbf{\hat{p}} - \mathbf{\hat{d}}_F}) \\
		& = {\err}_{\mathbf{d}_F} - {\mathbf{\hat{Z}}_F^T}\;{\hat{\Rot{}{}}}\;{\err}_{\mathbf{p}} + {\mathbf{\hat{Z}}_F^T}\ \hat{\Rot{}{}}\; S({\epsilon}_{\Rot{}{}})\; {\hat{\Rot{}{}}}^T\ {\mathbf{\hat{Z}}_F}\ {\mathbf{\hat{Z}}_F^T} ({\mathbf{\hat{p}} - \mathbf{\hat{d}}_F}) \\
		& = {\err}_{\mathbf{d}_F} - {\mathbf{\hat{Z}}_F^T}\;{\hat{\Rot{}{}}}\;{\err}_{\mathbf{p}} + \; S({\mathbf{\hat{Z}}_F^T}\ \hat{\Rot{}{}}\; {\epsilon}_{\Rot{}{}})\; {\mathbf{\hat{Z}}_F^T} ({\mathbf{\hat{p}} - \mathbf{\hat{d}}_F}) \\	
		& = {\err}_{\mathbf{d}_F} - {\mathbf{\hat{Z}}_F^T}\;{\hat{\Rot{}{}}}\;{\err}_{\mathbf{p}} - \; S({\mathbf{\hat{Z}}_F^T} ({\mathbf{\hat{p}} - \mathbf{\hat{d}}_F}))\; {\mathbf{\hat{Z}}_F^T}\ \hat{\Rot{}{}}\; {\epsilon}_{\Rot{}{}},
	\end{split}
\end{align*}
thus, leading to $\glog{\SE{3}}\left(h^{-1}\left(\Xhat_\kpred\right)\ h\left(\Xhat_\kpred\ \gexphat{G}\left(\err_\kpred\right)\right)\right)$ equal to,
$$
\begin{bmatrix} 
	S({\err}_{\mathbf{Z}_F})- S({\mathbf{\hat{Z}}_F^T}\;{\hat{\Rot{}{}}}\;{\err}_{\Rot{}{}}) &  {\err}_{\mathbf{d}_F} - {\mathbf{\hat{Z}}_F^T}\;{\hat{\Rot{}{}}}\;{\err}_{\mathbf{p}} - \; S({\mathbf{\hat{Z}}_F^T} ({\mathbf{\hat{p}} - \mathbf{\hat{d}}_F}))\; {\mathbf{\hat{Z}}_F^T}\ \hat{\Rot{}{}}\; {\epsilon}_{\Rot{}{}} \\
	\Zeros{1}{3} & 0
\end{bmatrix}.
$$

Applying the vee operator and using the adjoint property of $\SO{3}$, we have becomes,
$$
\glogvee{\SE{3}}\left(h^{-1}\left(\Xhat_\kpred\right)\ h\left(\Xhat_\kpred\ \gexphat{G}\left(\err_\kpred\right)\right)\right) =
\begin{bmatrix} 
 {\epsilon}_{\mathbf{d}_F} - \; {\mathbf{\hat{Z}}_F^T}{\hat{\Rot{}{}}}\;{\epsilon}_{\mathbf{p}} - \; S({\mathbf{\hat{Z}}_F^T(\mathbf{\hat{p}} -\ \mathbf{\hat{d}}_F)})\;{\mathbf{\hat{Z}}_F^T\hat{\Rot{}{}}}\;{\epsilon}_{\Rot{}{}}\\
 {\epsilon}_{\mathbf{Z}_F}- {\mathbf{\hat{Z}}_F^T}{\hat{\Rot{}{}}}\;{\epsilon}_{\Rot{}{}}
	\end{bmatrix}.
$$
The measurement model Jacobian during single support can be obtained by computing partial derivatives as,
\begin{equation}
\label{eq:chap:diligent-kio-measmodeljacobianLF}
\begin{split}
\mathbf{H}^{LF} = \begin{bmatrix} 
-{\mathbf{\hat{Z}}_{LF}^T \hat{\Rot{}{}}} & -{\mathbf{\hat{Z}}_{LF}^T}S({\mathbf{\hat{p}} - \mathbf{\hat{d}}_{LF}}){\hat{\Rot{}{}}} & \Zero{3} & \I{3} & \Zero{3} & \Zeros{3}{12} \\
\Zero{3} & -{\mathbf{\hat{Z}}_{LF}^T \hat{\Rot{}{}}}  & \Zero{3} & \Zero{3} & \I{3} & \Zeros{3}{12} 
\end{bmatrix}, \\
\mathbf{H}^{RF} = \begin{bmatrix} 
-{\mathbf{\hat{Z}}_{RF}^T \hat{\Rot{}{}}} & -{\mathbf{\hat{Z}}_{RF}^T}S({\mathbf{\hat{p}} - \mathbf{\hat{d}}_{RF}}){\hat{\Rot{}{}}} & \Zeros{3}{9} & \I{3} & \Zero{3} & \Zeros{3}{6} \\
\Zero{3} & -{\mathbf{\hat{Z}}_{RF}^T \hat{\Rot{}{}}}  & \Zeros{3}{9} & \Zero{3} & \I{3} & \Zeros{3}{6} 
\end{bmatrix}.
\end{split}
\end{equation}

These matrices can be stacked together during the double support.
The same procedure can be extended to handle an arbitrary number of contacts.
The Kalman gain $\mathbf{K}_\knext$ can be computed from predicted state covariance $\cov_\kpred$, measurement noise covariance $\mathbf{N}_\knext$ and the measurement model Jacobian $\mathbf{H}_\knext$.
These matrices are used to get an updated state estimate $\Xhat_\kest$ and covariance $\cov_\kest$ after an appropriate state reparametrization. (See Algorithm \ref{algo:chap02:dlgekf}). \looseness=-1

As a concluding remark for DILIGENT-KIO, while looking back at the state representation proposed in Proposition \ref{proposition:chap:diligent-kio-state-repr}, some similarities and differences can be drawn when comparing with the  $\SEk{N+2}{3}$ state representation used in the seminal work of \cite{hartley2020contact} for the invariant EKF developed for legged robot state estimation.
The latter is used for state-estimation of point-contact-based legged robots and includes base pose, base linear velocity, and the feet positions within the state representation, while the foot rotations are omitted given the point contacts.
The feet positions are introduced within the state to exploit the property of the semi-direct product with the base rotation. 
This state representation is exploited in building the so-called invariant observation (see Eq. \eqref{eq:chap02:invekf-right-invariant-observation}), further leading to an estimator design with stronger convergence guarantees.
However, since in Proposition \ref{proposition:chap:diligent-kio-state-repr}, we additionally consider foot rotations, the semi-direct product rule cannot be used effectively when characterizing the foot rotation with full rotation matrix, leading to a decoupling of the base link and foot link quantities. \looseness=-1

\begin{remark}
Nevertheless, the property of invariant updates is seamless only for vector observations and cannot be directly applied for humanoid robots' flat foot scenario if we consider the use of relative pose measurements as a whole.
One may choose to use invariant observations for relative position measurements while handling relative orientations through a non-invariant measurement update by decoupling the foot poses; foot position considered in semi-direct product with the base rotation while keeping the foot rotation isolated. 
This is the approach taken, in Chapter \ref{chap:human-motion} of this thesis in the estimator design  for human base estimation, which can also be used for humanoid base estimation.
Recent works of \cite{phogat2020invariant} and \cite{phogat2020discrete} extend the invariant filtering for observations evolving over matrix Lie groups.
\end{remark}

\subsection{Error Metrics for Trajectory Evaluation}
\label{sec:chap:diligent-kio-error-metrics}
As already noted in Section \ref{sec:chap:swa-walking-exp}, we rely on Absolute Trajectory Error (ATE) and Relative Pose Error (RPE) as the error metrics for evaluating the performance of the estimators. 
These metrics are suggested as common benchmark metrics by  \cite{sturm2012evaluating,zhang2018tutorial, schubert2018tum} for evaluating visual-inertial state estimation algorithms. \looseness=-1


A left invariant absolute error can be computed given $(\mathbf{R}, \mathbf{p}, \mathbf{v})$ as true states and $(\Rhat{}{}, \mathbf{\hat{p}}, \mathbf{\hat{p}})$ as estimates for a trajectory with discrete time steps $k =1:n$,
\begin{align}
    &\text{ATE}_\text{rot} = \sum_{k=1}^{n} \frac{1}{n}\norm{ \glogvee{\SO{3}}(\Rhat{}{}^T \Rot{}{}) }^2 \\
    &\text{ATE}_\text{pos} = \sum_{k=1}^{n} \frac{1}{n}\norm{ \Rhat{}{}^T (\mathbf{p} - \mathbf{\hat{p}}) }^2 \\
    &\text{ATE}_\text{vel} = \sum_{k=1}^{n} \frac{1}{n}\norm{ \Rhat{}{}^T (\mathbf{v} - \mathbf{\hat{v}}) }^2 
\end{align}

Instead of using the axis-angle representation error used here for computing the orientation errors through the logarithm mapping, one may also choose to convert to roll-pitch-yaw Euler angles.
The right-invariant ATE is similarly computed depending on the error-definition as described in Eq. \eqref{eq:eq:chap:cdekf-sys-rie}. \looseness=-1

The Relative Pose Error (RPE) in the left-invariant sense is used to analyze the local accuracy of the trajectories within fixed intervals of time $e$.
A left-invariant relative pose error at any time-instant $k$ can be computed as,
$$
\mathbf{E}_k = (\Hhat{}{}_k^{-1} \Hhat{}{}_{k + e})^{-1}  (\Transform{}{}_k^{-1} \Transform{}{}_{k + e} ) \in \SE{3}.
$$
The total relative pose error for the orientation and position trajectories can be obtained as,
\begin{align}
    &\text{RPE}_\text{rot} = \sum_{k=1}^{n} \frac{1}{n}\norm{ \glogvee{\SO{3}}(\text{rot}(\mathbf{E}_k) }^2 \\
    &\text{RPE}_\text{pos} = \sum_{k=1}^{n} \frac{1}{n}\norm{ \text{trans}(\mathbf{E}_k) }^2,
\end{align}
where, the operators $\text{rot}(\mathbf{E}_k)$ and $\text{trans}(\mathbf{E}_k)$ are used to select the rotation and the position elements from the error transformation matrix.
For computing the RPE in the right invariant sense, the order of multiplication changes accordingly. \looseness=-1

\subsection{Experimental Evaluation}
We use the state-of-the-art filters,  Observability Constrained quaternion based EKF (OCEKF) presented by \cite{rotella2014state} and Right Invariant EKF (InvEKF) proposed by \cite{hartley2020contact}, for a baseline comparison. 
It must be noted that while the former supports flat foot contacts, the latter considers only point foot contacts.
Additionally, we also use the Simple Weighted Averaging (SWA) estimator presented in Chapter \ref{chap:floating-base-swa} for the comparison.
It must be noted that, for the experiments in this section, the parameters were chosen differently from those described in Section \ref{sec:chap:swa-walking-exp} for the SWA estimator.
Along with the position-controlled walking experiment already described in Section \ref{sec:chap:swa-walking-exp}, we use also a torque-controlled Center of Mass (CoM) sinusoidal trajectory tracking experiment (\cite{nori2015icub}) on iCub for the evaluation of estimators in an open-loop manner.

\begin{remark}
In these experiments, we have not performed a comparison with the most relevant state-of-the-art estimator proposed by \cite{qin2020novel} which is a flat-foot filter within the invariant filtering framework on Lie groups.
This is because this piece of literature was not yet surveyed at the time of performing this experimental evaluation.
\end{remark}

Tables \ref{table:diligent-kio:noise-parameters} and \ref{table:diligent-kio:prior-parameters} show the noise parameters and initial state standard deviations used for the experiment. 
It must be noted that the noise parameters for the IMU measurements are different from those obtained using the Allan Variance analysis performed in Section \ref{sec:chap:swa-walking-exp}.
This is due to the sensitivity of the filter to the combination of chosen variance values for the contact foot velocity noise and the accelerometer noise.
The position estimates in particular are very sensitive to these noises especially because the accelerometer measurements are passed as exogenous inputs to the filter and the position estimates rely on double integration of the acceleration computed from accelerometer measurements for the prediction.

\begin{table}
\centering
						\begin{tabular}{|c|c|}
						\hline 
							Sensor & noise std dev. \\
								\hline	
							Lin. Accelerometer & 0.09 \si{\metre \per {\second^2}}	 \\
							Gyroscope & 0.01 \si{\radian \per \second} \\
							Acc. bias & 0.01 \si{\metre \per {\second^2}} \\
							Gyro. bias &  0.001 \si{\radian \per \second} \\
							Contact foot lin. velocity & 0.009 \si{\meter\per\second}\\
							Contact foot ang. velocity & 0.004 \si{\radian \per \second} \\							
							Joint encoders & 0.1 \si{\degree}						\\	\hline 
						\end{tabular}
						\caption{Noise parameters used for the estimators.}
	\label{table:diligent-kio:noise-parameters}
\end{table}

\begin{table}
\centering
					\begin{tabular}{|c|c|}
					\hline 
						State element & initial std dev. \\
						\hline	
						IMU \& feet position & 0.01 \si{\meter} \\
						IMU \& feet orientation & 10 \si{\degree} \\
						IMU linear velocity & 0.5 \si{\meter\per\second} \\
						Acc. bias & 0.01 \si{\metre \per {\second^2}} \\
						Gyro. bias & 0.002  \si{\radian \per \second}
						\\	\hline 
					\end{tabular}
	\caption{Prior deviations used for the estimators.}
	\label{table:diligent-kio:prior-parameters}
\end{table}

A convergence analysis similar to what is done in \cite{hartley2020contact} is first performed for the walking experiment, in order to compare the performance of DILIGENT-KIO with the OCEKF in estimating the observable states. 
These estimators are run for $25$ trials with the same measurements, noise parameters and prior deviations but with random initial orientations and linear velocities. 
The roll and pitch Euler angles for setting the initial IMU orientation were uniformly sampled from $-30 \si{\degree}$ to $30 \si{\degree}$, while the initial IMU linear velocities were sampled uniformly from $-0.5$ to $0.5 \si{\meter\per\second}$. 

\begin{figure}[!t]
\centering
\begin{subfigure}{\textwidth}
\centering
    \includegraphics[scale=0.4]{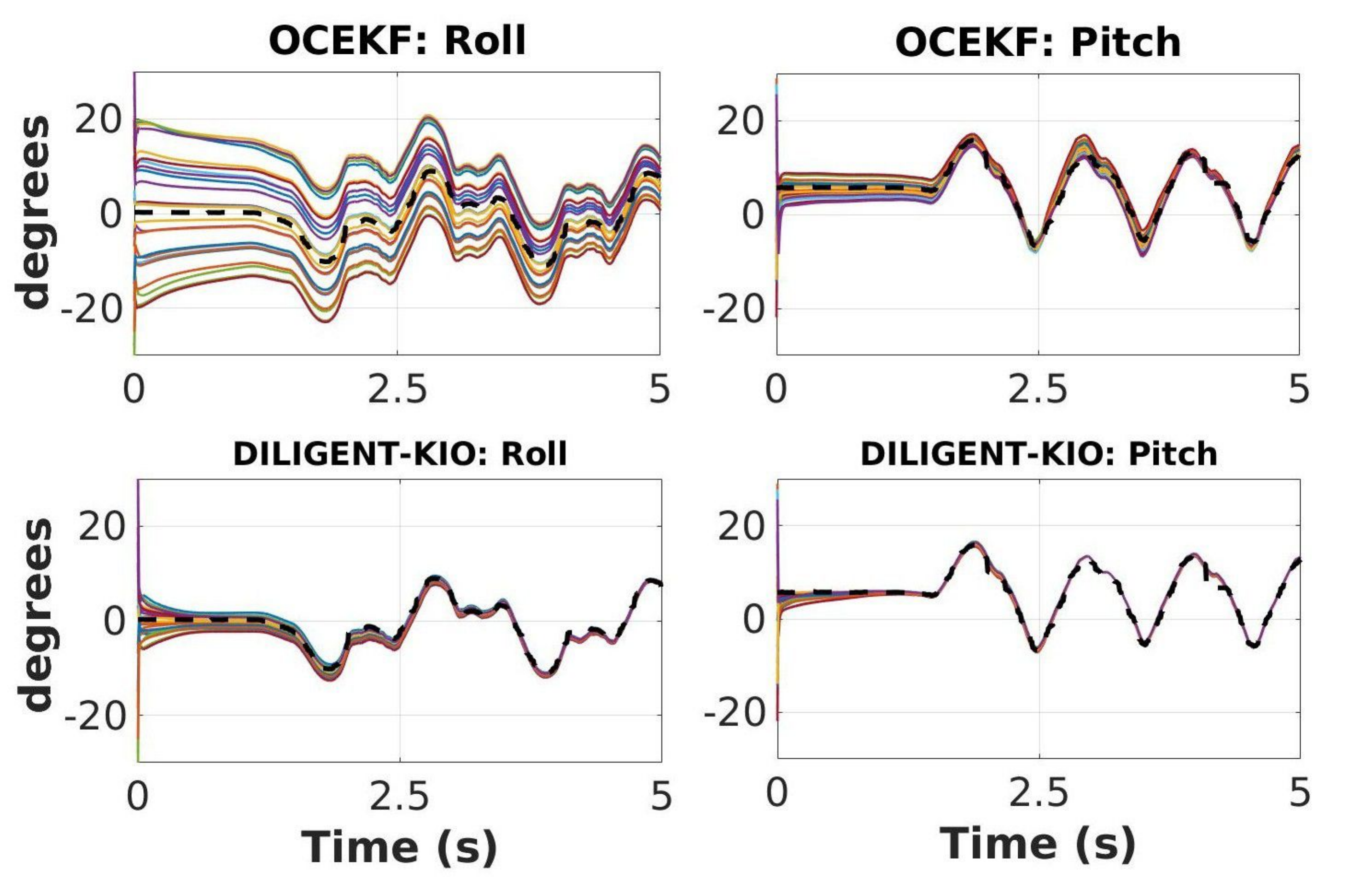}
\end{subfigure}
\begin{subfigure}{\textwidth}
\centering
    \includegraphics[scale=0.5]{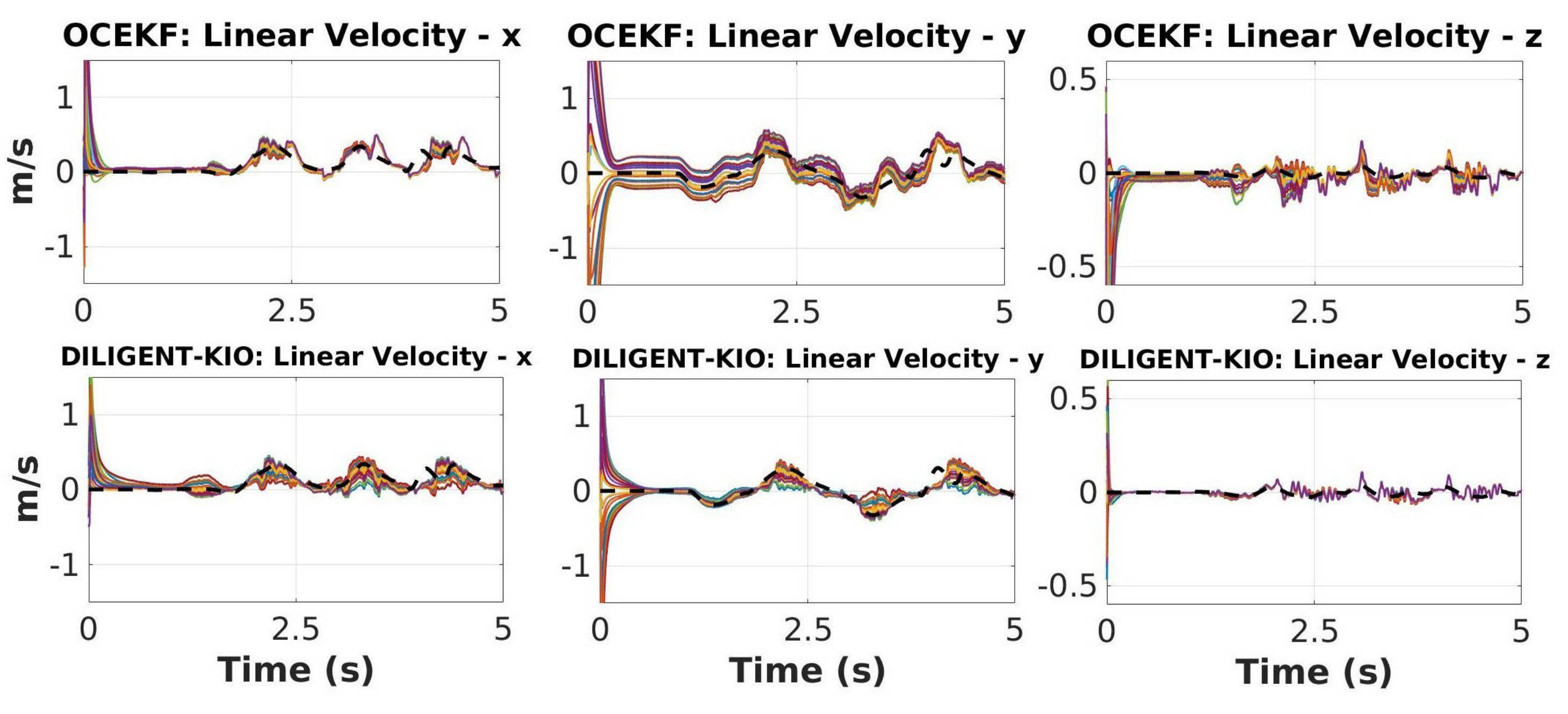}
\end{subfigure}
	\caption{ Orientation and velocity estimates from 25 trials of OCEKF and DILIGENT-KIO (proposed) for a forward walking experiment on the iCub humanoid platform using the same noisy measurements from the robotic hardware, noise statistics and initial covariances, but initialized with random orientations and velocities. The dashed black line is the ground truth trajectory from the Vicon motion capture system. DILIGENT-KIO (bottom row) is seen to converge considerably faster than OCEKF (top row) in almost all the directions. }
	\label{fig:diligent-kio:random_initialstates_real}
\end{figure}

The comparison of the estimates for OCEKF and DILIGENT-KIO is shown in Figure \ref{fig:diligent-kio:random_initialstates_real}. 
DILIGENT-KIO is seen to converge considerably faster than the OCEKF towards the ground truth measurements. 
The difference between the OCEKF and DILIGENT-KIO lies mainly in the uncertainty representation. The former uses non-interacting manifolds for all the state elements, while the latter uses interacting manifolds (such as $\SEk{2}{3}$ and $\SE{3}$) resulting in different tangent space parametrizations for the error state.
This causes the uncertainty representation to be more coupled and rigorous for the latter as seen in Eq. \eqref{eq:chap:diligent-kio-perturbed-mean} resulting in different Kalman gain computations for innovation updates.
It must be noted that OCEKF is also an EKF on Lie groups, however, differs from DILIGENT-KIO in terms of uncertainty propagation due to the decoupled state representation. \looseness=-1

\begin{figure}[!h]
	\begin{subfigure}{\textwidth}
		\centering
\includegraphics[scale=0.15]{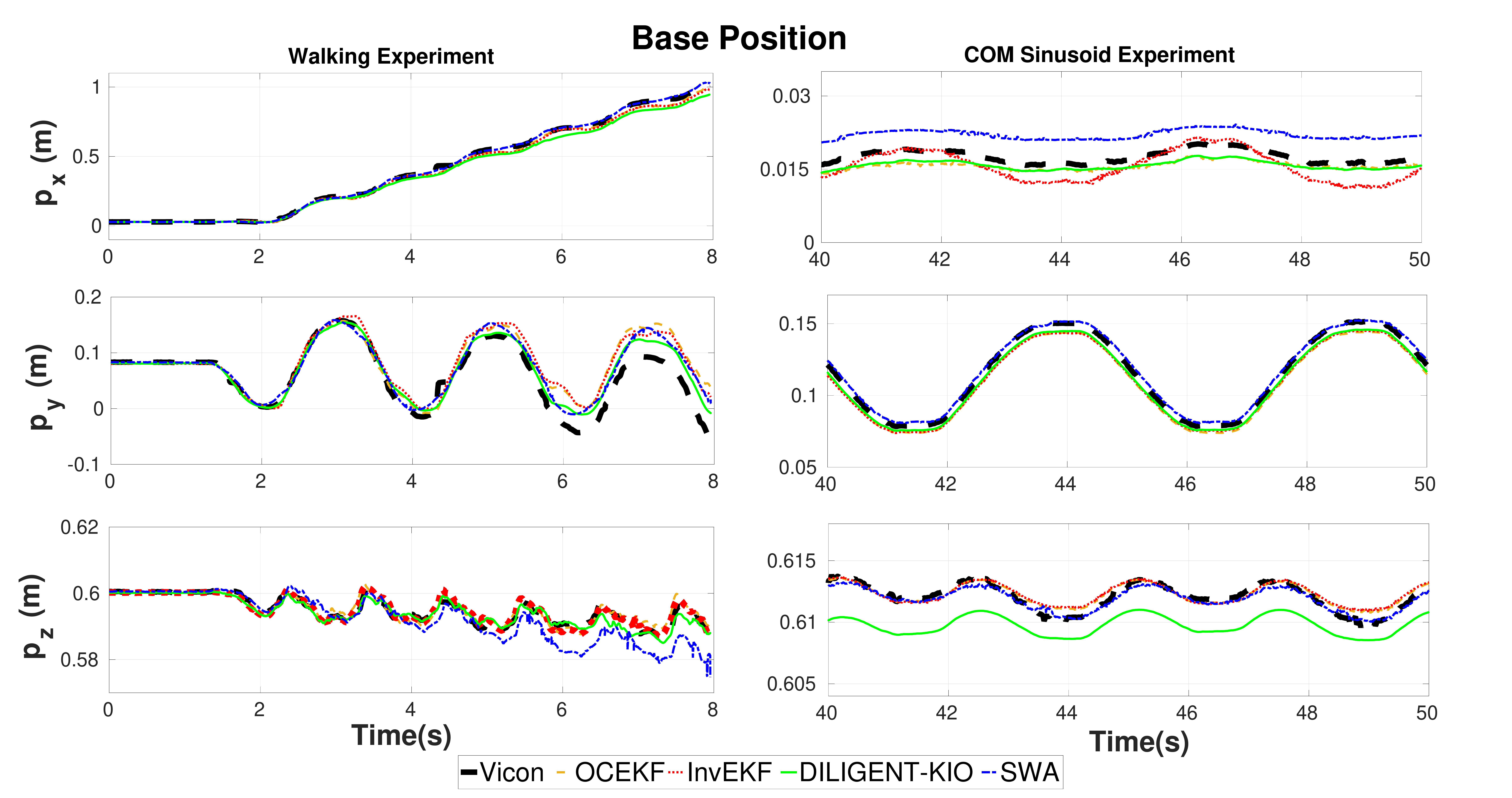}
\caption{Base position for walking (\emph{left}) and COM sinusoid experiment (\emph{right}).}
\label{fig:diligent-kio:basepos}
	\end{subfigure}
\begin{subfigure}{\textwidth}
	\centering
\includegraphics[scale=0.15]{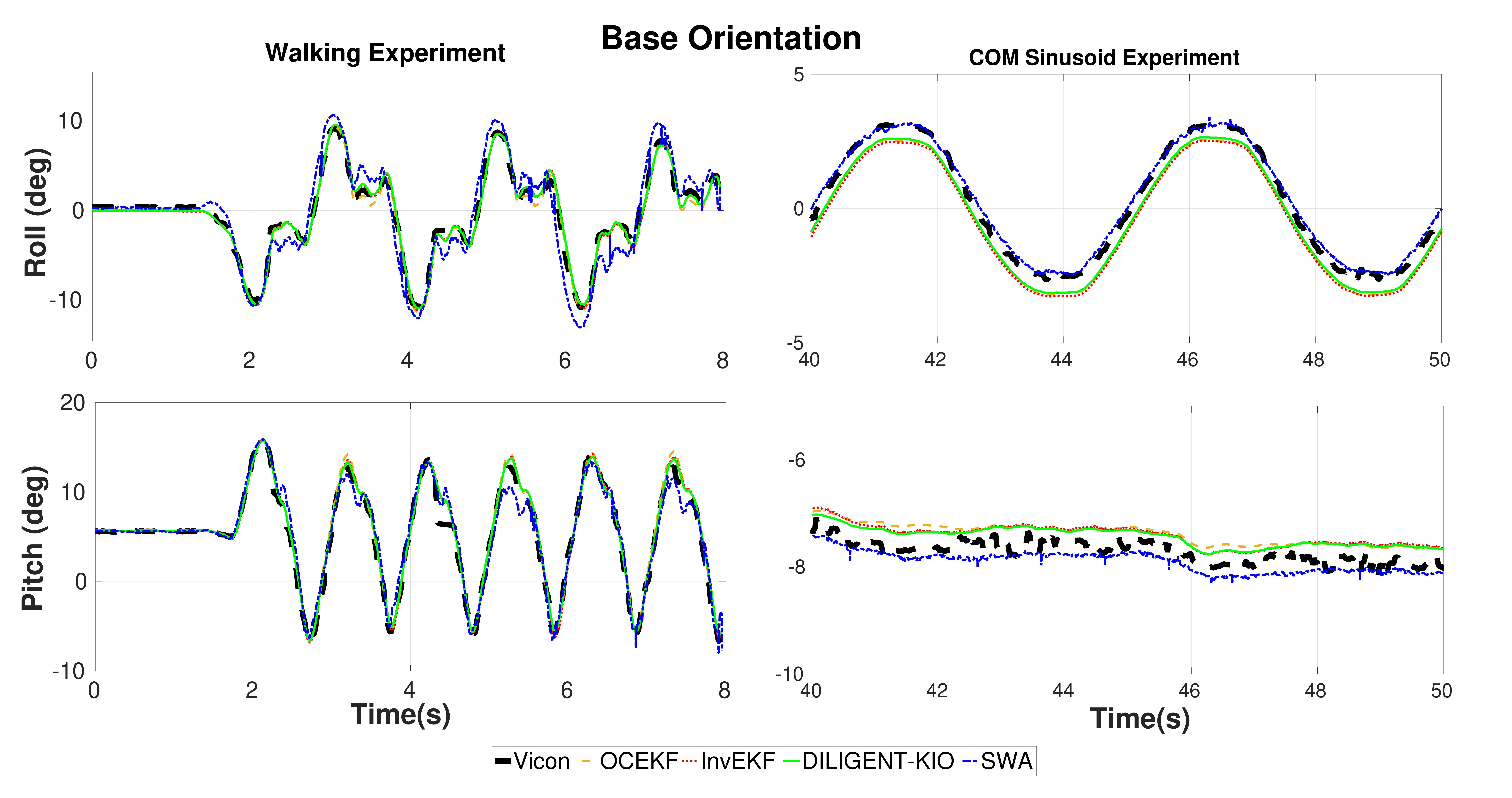}
\caption{Base roll and pitch for walking and COM sinusoid experiment. \looseness=-1}
\label{fig:diligent-kio:baserot} 
\end{subfigure}	
\caption{Experimental comparison of DILIGENT-KIO with state-of-the-art estimators.}
\end{figure}

Figures \ref{fig:diligent-kio:basepos} and \ref{fig:diligent-kio:baserot} show the comparison of estimates from OCEKF, InvEKF,  SWA and DILIGENT-KIO against the ground truth measurements for both the walking and the CoM sinusoid trajectory experiment. 
It can be noticed in Figures \ref{fig:diligent-kio:random_initialstates_real}, \ref{fig:diligent-kio:basepos}, and \ref{fig:diligent-kio:baserot}, there seems to be a discontinuity in the ground truth measurements between times $t = 4.3 \si{\second}$ and $t= 4.5 \si{\second}$ for the walking experiment,  which is caused by the occlusion of Vicon markers while the robot was walking in a cluttered environment.  \looseness=-1

\begin{table}[ht]
\centering 
\begin{tabular}[width=\textwidth]{c c c c c c}
\hline
Filter	& \multicolumn{5}{c}{Walking $1\si{\meter}$} \\
 												\hline
& \multicolumn{3}{c}{ATE} & \multicolumn{2}{c}{RPE}  \\
\hline
& rot [$\si{\degree}$] & pos [$\si{\meter}$] & vel [$\si{\meter/\second}$] & rot [$\si{\degree}$] & pos [$\si{\meter}$] \\
\hline
\footnotesize{SWA} & 3.98 & \textbf{0.025} &  0.291 & 3.93 & 0.029  \\
\footnotesize{InvEKF} & 3.34 & 0.038 &  0.133 & \textbf{1.90} & \textbf{0.025}  \\
\footnotesize{OCEKF} & 4.67 & 0.038 & 0.132 &  4.47 & 0.035  \\
\footnotesize{DILIGENT-KIO}  & \textbf{2.29} & 0.040 &  \textbf{0.130} & \textbf{1.90} & 0.039  \\
\hline
\end{tabular}
\caption{Left invariant Absolute Trajectory Error (ATE) and Relative Pose Error (RPE) comparison for walking and CoM sinusoid experiment. \looseness=-1}
\label{table:diligent-kio:walking-errors} 
\end{table}

\begin{table}[ht]
\centering 
\begin{tabular}[width=\textwidth]{c c c c c c}
\hline
Filter & \multicolumn{5}{c}{CoM sinusoid} \\
 												\hline
& \multicolumn{3}{c}{ATE} & \multicolumn{2}{c}{RPE}   \\
\hline
& rot [$\si{\degree}$] & pos [$\si{\meter}$] & vel [$\si{\meter/\second}$] & rot [$\si{\degree}$] & pos [$\si{\meter}$] \\
\hline
\footnotesize{SWA} & \textbf{0.39} & \textbf{0.005} & 0.0545 & 0.28 & \textbf{0.0013} \\
\footnotesize{InvEKF} & 4.27 & 0.006 & 0.0098 & 1.36 & 0.0018  \\
\footnotesize{OCEKF} & 0.68 & 0.006 & 0.0094 & 0.18 & 0.0019  \\
\footnotesize{DILIGENT-KIO} & 0.59 & 0.005 & \textbf{0.0089} & \textbf{0.16} & 0.0016 \\
\hline
\end{tabular}
\caption{Left invariant Absolute Trajectory Error (ATE) and Relative Pose Error (RPE) comparison for CoM sinusoid experiment. \looseness=-1}
\label{table:diligent-kio:com-errors} 
\end{table}

The ATE and RPE in the left-invariant sense are shown in Table \ref{table:diligent-kio:walking-errors} for the $1 \si{\meter}$ walking experiment and in Table \ref{table:diligent-kio:com-errors} for the CoM trajectory tracking experiment. 
DILIGENT-KIO is seen to perform comparably with the base-line estimators for both the experiments, especially in the observable directions of orientation and velocity.
For longer experimental duration, DILIGENT-KIO suffers more from position drifts than the other filters, while the orientation and velocity errors remain comparable.
The accuracy of the InvEKF in the rotation estimates seems to be low for both experiments as shown by the high errors in rotation part of absolute trajectory errors.
This can be accounted for the use of a point foot filter for the flat foot scenario of the humanoid robot.

\section{Extension to Humanoid Localization}
\label{sec:chap:diligent-kio:humanoid-localization}
In the previous sections, along with the base state and IMU biases, we have considered the state of both the feet of the robot. 
However, the presented method can be extended for an arbitrary number of contacts and also landmarks for performing localization using exteroceptive sensors.
A typical example would be the humanoid robot navigating in an environment detecting fiducial markers in its environments and estimating the pose of these markers relative to the camera frame.
These relative landmark poses are analogous to the relative feet poses with the difference that the intermediary transformation given by the kinematic chain is replaced with a camera-based transformation which depends on the intrinsic and extrinsic calibration parameters of the camera.
We can consider adding or removing contacts/static landmarks to/from the state in a straightforward manner as shown by \cite{hartley2020contact}. \looseness=-1

New landmarks or contacts can be added to the state directly using the information about the current state estimate, but the prior covariance for the landmark needs to be properly initialized.
With our choice of state representation, there is no dependence of the error-state of the landmark on the rest of the system states.
Thus, it can be initialized directly with the prediction noise covariance expressed in its local frame.
If we choose to add the state of a landmark $L$ whose position $\mathbf{d}_L$ and orientation $\mathbf{Z}_L$ we want to estimate as pose $\Hhat{A}{L} \in \SE{3}$, then we update the state error covariance matrix with the known noise parameters of the landmark as follows,
{
\small
\begin{align}
    \begin{split}
    &  \begin{bmatrix}
            \epsilon_\mathbf{p} \\ 
            \epsilon_\mathbf{R} \\ 
            \epsilon_\mathbf{v} \\ \epsilon_{\mathbf{d}_{LF}} \\ \epsilon_{\mathbf{Z}_{LF}} \\ \epsilon_{\mathbf{d}_{RF}} \\ \epsilon_{\mathbf{Z}_{RF}} \\ 
            \epsilon_{\mathbf{d}_{L}} \\ \epsilon_{\mathbf{Z}_{L}} \\
            \epsilon_{\biasAcc} \\
            \epsilon_{\biasGyro}
        \end{bmatrix} = \begin{bmatrix}
        \I{3} & \Zero{3} & \Zero{3} & \Zero{3} & \Zero{3} & \Zero{3} & \Zero{3} & \Zero{3} & \Zero{3} \\
        \Zero{3} & \I{3} & \Zero{3} & \Zero{3} & \Zero{3} & \Zero{3} & \Zero{3} & \Zero{3} & \Zero{3} \\
        \Zero{3} &  \Zero{3} & \I{3} & \Zero{3} & \Zero{3} & \Zero{3} & \Zero{3} & \Zero{3} & \Zero{3} \\
        \Zero{3} &  \Zero{3} & \Zero{3} & \I{3} & \Zero{3} & \Zero{3} & \Zero{3} & \Zero{3} & \Zero{3}  \\
        \Zero{3} & \Zero{3} & \Zero{3} & \Zero{3} & \I{3} & \Zero{3} & \Zero{3} & \Zero{3} & \Zero{3}  \\
        \Zero{3} & \Zero{3} & \Zero{3} & \Zero{3} & \Zero{3} & \I{3} & \Zero{3} & \Zero{3} & \Zero{3}\\
        \Zero{3} & \Zero{3} & \Zero{3} & \Zero{3} & \Zero{3} & \Zero{3} & \I{3} & \Zero{3} & \Zero{3} \\
        \Zero{3} & \Zero{3} & \Zero{3} & \Zero{3} & \Zero{3} & \Zero{3} & \Zero{3} & \Zero{3} & \Zero{3} \\
        \Zero{3} & \Zero{3} & \Zero{3} & \Zero{3} & \Zero{3} & \Zero{3} & \Zero{3} & \Zero{3} & \Zero{3} \\
        \Zero{3} & \Zero{3} & \Zero{3} & \Zero{3} & \Zero{3} & \Zero{3} & \Zero{3} & \I{3} & \Zero{3} \\
        \Zero{3} & \Zero{3} & \Zero{3} & \Zero{3} & \Zero{3} & \Zero{3} & \Zero{3} & \Zero{3} & \I{3}
        \end{bmatrix}\begin{bmatrix}
            \epsilon_\mathbf{p} \\ 
            \epsilon_\mathbf{R} \\ 
            \epsilon_\mathbf{v} \\ \epsilon_{\mathbf{d}_{LF}} \\ \epsilon_{\mathbf{Z}_{LF}} \\ \epsilon_{\mathbf{d}_{RF}} \\ \epsilon_{\mathbf{Z}_{RF}} \\ 
            \epsilon_{\biasAcc} \\
            \epsilon_{\biasGyro}
        \end{bmatrix} +  \begin{bmatrix}
            \Zeros{3}{1} \\ 
            \Zeros{3}{1} \\ 
            \Zeros{3}{1} \\ 
            \Zeros{3}{1} \\ 
            \Zeros{3}{1} \\  
            \Zeros{3}{1} \\ 
            \Zeros{3}{1} \\ 
            -\mathbf{w}_{\mathbf{d}_{L}} \\
            -\mathbf{w}_{\mathbf{Z}_{L}} \\
            \Zeros{3}{1} \\ 
            \Zeros{3}{1} 
        \end{bmatrix} \Delta T, \\
        & \err_\text{new} = \mathbf{F}_\text{new} \err + \mathbf{w}_\text{new} , \\
        & \cov{}_\text{new} = \mathbf{F}_\text{new}\ \cov{}\ \mathbf{F}_\text{new}^T + \mathbf{Q}_c \Delta T.
    \end{split}
\end{align}
}

While removing contacts or landmarks from the state, we need to be careful in properly marginalizing out the covariance related to the landmark/contact we want to remove, where marginalization allows us to factor out the probability distribution associated with the landmark from the joint probability distribution of the rest of the state and landmarks.
A landmark $L$ with position $\mathbf{d}_L$ and orientation $\mathbf{Z}_L$ can be removed by removing the corresponding rows in the covariance matrix, \looseness=-1
{\small
\begin{align}
    \begin{split}
    & \begin{bmatrix}
            \epsilon_\mathbf{p} \\ 
            \epsilon_\mathbf{R} \\ 
            \epsilon_\mathbf{v} \\ \epsilon_{\mathbf{d}_{LF}} \\ \epsilon_{\mathbf{Z}_{LF}} \\ \epsilon_{\mathbf{d}_{RF}} \\ \epsilon_{\mathbf{Z}_{RF}} \\ 
            \epsilon_{\biasAcc} \\
            \epsilon_{\biasGyro}
        \end{bmatrix} = 
        \begin{bmatrix}
        \I{3} & \Zero{3} & \Zero{3} & \Zero{3} & \Zero{3} & \Zero{3} & \Zero{3} & \Zero{3} & \Zero{3} & \Zero{3}  & \Zero{3} \\
        \Zero{3} & \I{3} & \Zero{3} & \Zero{3} & \Zero{3} & \Zero{3} & \Zero{3} & \Zero{3} & \Zero{3} & \Zero{3}  & \Zero{3} \\
        \Zero{3} &  \Zero{3} & \I{3} & \Zero{3} & \Zero{3} & \Zero{3} & \Zero{3} & \Zero{3} & \Zero{3} & \Zero{3}  & \Zero{3} \\
        \Zero{3} &  \Zero{3} & \Zero{3} & \I{3} & \Zero{3} & \Zero{3} & \Zero{3} & \Zero{3} & \Zero{3} & \Zero{3}  & \Zero{3} \\
        \Zero{3} & \Zero{3} & \Zero{3} & \Zero{3} & \I{3} & \Zero{3} & \Zero{3} & \Zero{3} & \Zero{3} & \Zero{3}  & \Zero{3} \\
        \Zero{3} & \Zero{3} & \Zero{3} & \Zero{3} & \Zero{3} & \I{3} & \Zero{3} & \Zero{3} & \Zero{3} & \Zero{3}  & \Zero{3} \\
        \Zero{3} & \Zero{3} & \Zero{3} & \Zero{3} & \Zero{3} & \Zero{3} & \I{3} & \Zero{3} & \Zero{3} & \Zero{3}  & \Zero{3} \\
        \Zero{3} & \Zero{3} & \Zero{3} & \Zero{3} & \Zero{3} & \Zero{3} & \Zero{3} & \Zero{3} & \Zero{3}  & \I{3} &  \Zero{3}\\
        \Zero{3} & \Zero{3} & \Zero{3} & \Zero{3} & \Zero{3} & \Zero{3} & \Zero{3} & \Zero{3} & \Zero{3}  & \Zero{3} & \I{3}\end{bmatrix}
        \begin{bmatrix}
            \epsilon_\mathbf{p} \\ 
            \epsilon_\mathbf{R} \\ 
            \epsilon_\mathbf{v} \\ \epsilon_{\mathbf{d}_{LF}} \\ \epsilon_{\mathbf{Z}_{LF}} \\ \epsilon_{\mathbf{d}_{RF}} \\ \epsilon_{\mathbf{Z}_{RF}} \\ 
            \epsilon_{\mathbf{d}_{L}} \\ \epsilon_{\mathbf{Z}_{L}} \\
            \epsilon_{\biasAcc} \\
            \epsilon_{\biasGyro}
        \end{bmatrix}, \\
        & \err_\text{new} = \mathbf{F}_\text{new} \err, \\
        & \cov{}_\text{new} = \mathbf{F}_\text{new}\ \cov{}\ \mathbf{F}_\text{new}^T.
    \end{split}
\end{align}
}

The measurement updates for the landmark poses follow the same equations as discussed in Section \ref{sec:chap:diligent-kio-meas-model}.
The measurement noise covariance matrix for the measured landmark poses can be computed starting from the pixel noise or manually tuned to a suitable value.

\section{Variants of DILIGENT-KIO}
\label{sec:chap:diligent-kio:alternate-formulations}

In this section, we will look at some alternate formulations of DILIGENT-KIO and investigate the differences between these formulations.
Different choices lead to different linearized error propagation and update matrices which in turn affect the covariance propagation in the estimator.
Depending on the choices, these matrices of the linearized error system may or may not be independent of the trajectory of the system state.

When designing an estimator with partial observations, proper covariance management is crucial for maintaining the consistency of the filter.
The availability of only partial measurements may lead to the existence of a few unobservable directions in the underlying nonlinear system. 
In case, the system matrices of the linearized error system are dependent on the state configuration, there is a possibility that, with a wrong choice of linearization point, the covariance matrix gains information in the unobservable directions, leading to an overconfident estimate along these directions.
This overconfident estimate of an unobservable direction causes an inconsistency in the filter which might also lead to divergence in the estimated states.

\begin{remark}
The choice of the invariant error (left or right) is pertained only to the invariant errors of type 1 throughout this section (See Eqs. \eqref{def:left-invariant-error}, \eqref{def:right-invariant-error}). 
\end{remark}

In the following section, we derive three variants of DILIGENT-KIO in order to analyze the linearized error system of each variant.
In particular a discrete filter is derived in Section \ref{sec:chap:diligent-kio-rie} using the right-invariant error $\errG^R = \X \Xhat^{-1}$ differently from the left-invariant error  $\errG^L = \Xhat^{-1} \X$ used for the original DILIGENT-KIO.
A continuous-discrete filter using the right-invariant error formulation is derived in Section \ref{sec:chap:cd-ekf-rie} which combines the property of autonomous error propagation of Invariant EKF with observations evolving on matrix Lie groups.
This is followed by the derivation of its left-invariant error counterpart in Section \ref{sec:chap:cd-ekf-lie}.


\subsection{DILIGENT-KIO-RIE}
\label{sec:chap:diligent-kio-rie}
DILIGENT-KIO with right-invariant error formulation, DILIGENT-KIO-RIE in short, differs from its original only in terms of the uncertainty management and state reparametrization.
The underlying algorithm is the discrete EKF over matrix Lie groups using the right invariant error of type 1 described in Algorithm \ref{algo:chap02:dlgekf-rie}.
The discrete system dynamics remain the same as in Eq. \eqref{eq:chap:diligent-kio-sys-dynamics}, consequently leading to the same left trivialized motion model and associated noise as in Eqs. \eqref{eq:chap:diligent-kio-left-triv-motion-model} and \eqref{eq:chap:diligent-kio-left-triv-motion-noise}, respectively.
It must be noted that the trivialization of the motion model is independent from the kind of error and solely depends on the design choice of the system states and their associated dynamics.  \looseness=-1

In the state propagation step, the difference with the original estimator design arises with the Jacobian of the left-trivialized motion model. 
Given the right-invariant error, the perturbation is induced on the current state estimate from the left side,
\begin{align}
\label{eq:chap:diligent-kio-rie-perturbed-mean}
\begin{split}
\gexphat{\mathscr{M}}\left({\err_\kprior}\right)\ \Xhat_\kprior \approx& \big(  
\mathbf{\hat{p}} + S(\epsilon_\Rot{}{})  \mathbf{\hat{p}} + \epsilon_{\mathbf{p}}, 
\hat{\Rot{}{}} + S(\epsilon_\Rot{}{})\ \hat{\Rot{}{}}, 
\mathbf{\hat{v}} + S(\epsilon_\Rot{}{})  \mathbf{\hat{v}} + \epsilon_{\mathbf{v}}, \\
&\ \mathbf{\hat{d}_{LF}} + S(\epsilon_{\mathbf{Z}_{LF}})  \mathbf{\hat{d}_{LF}} + \epsilon_{\mathbf{d}_{LF}},\
\hat{\mathbf{Z}}_{LF} + S(\epsilon_{\mathbf{Z}_{LF}})\ \hat{\mathbf{Z}}_{LF}, \\
&\ \mathbf{\hat{d}_{RF}} + S(\epsilon_{\mathbf{Z}_{RF}})  \mathbf{\hat{d}_{RF}} + \epsilon_{\mathbf{d}_{RF}},\
\hat{\mathbf{Z}}_{RF} + S(\epsilon_{\mathbf{Z}_{RF}})\ \hat{\mathbf{Z}}_{RF}, 
\biasHat + \epsilon_\bias
\big)_\mathscr{M}.
\end{split}
\end{align}

\begin{lemma}
\label{lemma:chap:diligent-kio:jacMotionRIE}
The Jacobian $\mathfrak{F}_\kcurr$ of the left trivialized motion model $\hat{\Omega}$ with respect to the perturbation $\err$ for DILIGENT-KIO-RIE is given as,
\begin{equation}
\label{eq:chap:diligent-kio-rie-left-triv-motion-jacobian}
\begin{split}
\small
\mathfrak{F}_\kcurr &= \frac{\partial}{\partial \err} \Omega\left(\gexphat{\mathscr{M}}\left({\err_\kprior}\right)\ \Xhat_\kprior, \mathbf{u}_\kcurr\right)_{\err_\kprior = 0} \\
&=
\begin{bmatrix}
	\Zero{3} & \frac{1}{2} \Xi_1 \Delta T & \hat{\Rot{}{}}^T \; \Delta T & \Zeros{3}{12} &  -\frac{1}{2}\I{3}\;\Delta T^2 & \Zero{3} \\
	\Zero{3} & \Zero{3} & \Zero{3} & \Zeros{3}{12} &  \Zero{3} & -\I{3}\;\Delta T \\
	\Zero{3} & \Xi_1 & \Zero{3} & \Zeros{3}{12} &  -\I{3}\;\Delta T & \Zero{3} \\
	\Zeros{18}{3} & \Zeros{18}{3} & \Zeros{18}{3} & \Zeros{18}{12} &  \Zeros{18}{3} & \Zeros{18}{3}
	\end{bmatrix} \in \R^{27 \times 27},
\end{split}
\end{equation}
where, we have used $\Xi_1 = \hat{\Rot{}{}}^T S\left(\gravity{A}\right) \Delta T$. 
\end{lemma}
The proof for Lemma \ref{lemma:chap:diligent-kio:jacMotionRIE} is provided in the Appendix \ref{appendix:chapter:left-triv-motion-model-jacobian-diligent-kio-rie}. 
The linearized error propagation matrix can therefore be computed as $\mathbf{F}_\kcurr = \I{p}\ +\ \gadj{\Xhat_\kprior}\gljac{\mathscr{M}}\left(\hat{\Omega}_\kcurr \right) \mathfrak{F}_\kcurr$.\looseness=-1

We use the same measurement model as in Eq. \eqref{eq:chap:diligent-kio-meas-model} and  the difference arises in the measurement model (error) Jacobian,
$$
\mathbf{H}_\knext = \frac{\partial}{\partial \err} \glogvee{\SE{3}}\left( h^{-1}\left(\Xhat_\kpred\right)\ h\left(\gexphat{\mathscr{M}}\left(\err_\kpred\right)\ \Xhat_\kpred\ \right)\right)_{\err=0}. 
$$
\begin{lemma}
\label{lemma:chap:diligent-kio:jacMeasRIE}
The measurement model Jacobian during single support for DILIGENT-KIO-RIE can be obtained as,
\begin{equation}
\label{eq:chap:diligent-kio-rie-measmodeljacobianLF}
\begin{split}
\mathbf{H}^{LF} = \begin{bmatrix} 
-{\mathbf{\hat{Z}}_{LF}^T} & {\mathbf{\hat{Z}}_{LF}^T}S({\mathbf{\hat{d}}_{LF}}) & \Zero{3} & {\mathbf{\hat{Z}}_{LF}^T} & -{\mathbf{\hat{Z}}_{LF}^T}S({\mathbf{\hat{d}}_{LF}}) & \Zeros{3}{12} \\
\Zero{3} & -{\mathbf{\hat{Z}}_{LF}^T}  & \Zero{3} & \Zero{3} & {\mathbf{\hat{Z}}_{LF}^T} & \Zeros{3}{12} 
\end{bmatrix}, \\
\mathbf{H}^{RF} = \begin{bmatrix} 
-{\mathbf{\hat{Z}}_{RF}^T} & {\mathbf{\hat{Z}}_{RF}^T}S({\mathbf{\hat{d}}_{RF}}) & \Zeros{3}{9} & {\mathbf{\hat{Z}}_{RF}^T}  & -{\mathbf{\hat{Z}}_{RF}^T}S({\mathbf{\hat{d}}_{RF}}) & \Zeros{3}{6} \\
\Zero{3} & -{\mathbf{\hat{Z}}_{RF}^T}  & \Zeros{3}{9} & \Zero{3} & {\mathbf{\hat{Z}}_{RF}^T} & \Zeros{3}{6} 
\end{bmatrix}.
\end{split}
\end{equation}
\end{lemma}
The proof for Lemma \ref{lemma:chap:diligent-kio:jacMeasRIE} is provided in the Appendix \ref{appendix:chapter:meas-model-jacobian-diligent-kio-rie}.
Followed by the correction of the predicted state and the covariance, the update step ends with a state reparametrization $\Xhat_\kest\ =\ \gexphat{G}\left(\mathbf{m}_\knext^{-}\right)\ \Xhat_\kpred$ and the covariance reparametrization $\cov_\kest\ =\ \gljac{G}\left(\mathbf{m}_\knext^{-}\right)\ \cov_\kest^{-}\ \gljac{G}\left(\mathbf{m}_\knext^{-}\right)^T$.


\subsection{CODILIGENT-KIO-RIE}
\label{sec:chap:cd-ekf-rie}
In this section, we derive CODILIGENT-KIO-RIE which is a continuous-discrete EKF on matrix Lie groups with right invariant error formulation and the consideration of non-invariant observations evolving over matrix Lie groups.
This estimator design allows us to combine the property of autonomous error propagation and observations over matrix Lie groups.
The overall system takes the form, \looseness=-1
\begin{equation}
    \label{eq:eq:chap:cdekf-rie}
    \begin{split}
    & \frac{d}{dt}\X_t = f_{\mathbf{u}_t}\left(\X_t\right) + \X_t \ghat{G}{\mathbf{w}_t}, \\
    & \Z_{t_\kcurr} = h (\X_{t_\kcurr}) \; \gexphat{G^\prime}\left(\mathbf{n}_{t_\kcurr}\right).
    \end{split}
\end{equation}

The continuous system dynamics take the form, \looseness=-1
\begin{equation}
    \label{eq:eq:chap:cdekf-sys-dynamics}
    \begin{split}
    & \mathbf{\dot{p}} = \mathbf{v},    \\
    & \mathbf{\dot{\Rot{}{}}} = \Rot{}{}\ S\left(\yGyro{A}{B} - \biasGyro - \noiseGyro{B}\right), \\
    & \mathbf{\dot{v}} = \Rot{}{}\ \left(\yAcc{A}{B}  - \biasAcc - \noiseAcc{B}\right) + \gravity{A}, \\
    & \mathbf{\dot{d}}_F = \mathbf{Z}_F\ (-\noiseLinVel{F}),  \\
    & \mathbf{\dot{Z}}_F = \mathbf{Z}_F\ S\left(-\noiseAngVel{F}\right), \\
    & \mathbf{\dot{\bias}} = \noiseBias{B}.
    \end{split}
\end{equation}

In the propagation step, the mean can be simply propagated using $\frac{d}{dt}\Xhat_t = f_{\mathbf{u}_t}\left(\Xhat_t\right) $.
For the covariance propagation, we begin by writing the dynamics of the right invariant error.
We derive the continuous dynamics of the linearized error in order to compute the error and noise propagation matrices required to compute the predicted state error covariance.

The right-invariant error $\eta^R_\mathscr{M} = \X \Xhat^{-1}$ for the state representation defined in Eq. \eqref{proposition:chap:diligent-kio-state-repr} can also be written in the form of following tuple representation,
\begin{equation}
    \label{eq:eq:chap:cdekf-sys-rie}
    \begin{split}
    \eta^R_\mathscr{M} &=\left(\mathbf{p}, \mathbf{R}, \mathbf{v}, \mathbf{d}_{LF}, \mathbf{Z}_{LF}, \mathbf{d}_{RF}, \mathbf{Z}_{RF}, \bias \right)_{\mathscr{M}} \circ \\
    & \left(-\mathbf{\hat{R}}^T\mathbf{p},\ \mathbf{\hat{R}}^T,\ -\mathbf{\hat{R}}^T\mathbf{v},\  -\mathbf{\hat{Z}}_{LF}\mathbf{\hat{d}}_{LF},\ \mathbf{\hat{Z}}_{LF}^T,\  -\mathbf{\hat{Z}}_{RF}^T\mathbf{\hat{d}}_{RF},\ \mathbf{\hat{Z}}_{RF}^T ,\ -\hat{\bias} \right)_{\mathscr{M}} \\
    &= \big(\mathbf{p} - \Rot{}{}\hat{\Rot{}{}}^T \mathbf{\hat{p}}, \
             \Rot{}{}\hat{\Rot{}{}}^T, \ 
             \mathbf{v} - \Rot{}{}\hat{\Rot{}{}}^T \mathbf{\hat{v}}, \ \\
             &  \quad \quad \mathbf{d}_{LF} - \mathbf{Z}_{LF}\mathbf{\hat{Z}}_{LF}^T  \mathbf{\hat{d}}_{LF}, \
             \mathbf{Z}_{LF} \mathbf{\hat{Z}}_{LF}^T, \ \\
             & \quad \quad \mathbf{d}_{RF} - \mathbf{Z}_{RF}\mathbf{\hat{Z}}_{RF}^T  \mathbf{\hat{d}}_{RF}, \
             \mathbf{Z}_{RF} \mathbf{\hat{Z}}_{RF}^T, \
             \bias - \hat{\bias}
             \big)_{\mathscr{M}}.
    \end{split}
\end{equation}

Given, $\eta^R_\mathscr{M} = \gexphat{\mathscr{M}}\left(\err\right) \approx \I{n} + \ghat{\mathscr{M}}{\err}$, the overall non-linear error dynamics $\frac{d}{dt} \eta^R_t = \frac{d}{dt} \left(\X \Xhat^{-1} \right)$ can be summarized as,
\begin{equation}
\label{eq:eq:chap:cdekf-sys-rie-dynamics}
    \begin{split}
    &  \dot{\eta}_\mathbf{p} =\ \err_\mathbf{v} - S(\mathbf{\hat{p}})\hat{\Rot{}{}}\left(\err_\biasGyro + \noiseGyro{B}\right), \\
    &  \dot{\eta}_\mathbf{R} =\ S(\hat{\Rot{}{}} \left(-\err_\biasGyro - \noiseGyro{B}\right) ) ,\\
    & \dot{\eta}_\mathbf{v} =\ S(\gravity{A}) \err_\mathbf{R} - S(\mathbf{\hat{v}})\hat{\Rot{}{}}\left(\err_\biasGyro + \noiseGyro{B}\right) -\hat{\Rot{}{}}\left(\err_\biasAcc + \noiseAcc{B}\right), \\
    & \dot{\eta}_{\mathbf{d}_F} =\ -S(\mathbf{\hat{d}}_F)\hat{\mathbf{Z}}_F\noiseAngVel{F} - \hat{\mathbf{Z}}_F \noiseLinVel{F},\\
    & \dot{\eta}_{\mathbf{Z}_F} =\ S(-\hat{\mathbf{Z}}_F\noiseAngVel{F}),\\
    & \dot{\eta}_\bias =\ \noiseBias{B}.
    \end{split}
\end{equation}

\begin{lemma}
\label{lemma:chap:diligent-kio:cdekf-rie-error-dyn}
With the non-linear error dynamics in Eq. \eqref{eq:eq:chap:cdekf-sys-rie-dynamics} and the definition of the error vector given in Eq. \eqref{eq:chap:diligent-kio-state-vector}, linearized error dynamics becomes $\dot{\err} = \mathbf{F}_c \err - \mathbf{L}_c \mathbf{w}$ (which is of the form described in Eq. \eqref{eq:chap02:invekf-linearized-rie-dynamics}), where,
\begin{equation}
\mathbf{F}_c =
\begin{bmatrix}
    \Zero{3} & \Zero{3} & \I{3} & \Zeros{3}{12} & \Zero{3} & - S(\mathbf{\hat{p}})\hat{\Rot{}{}} \\
    \Zero{3} & \Zero{3} & \Zero{3} &  \Zeros{3}{12} & \Zero{3} & -\hat{\Rot{}{}} \\
    \Zero{3} & S(\gravity{A}) & \Zero{3} &  \Zeros{3}{12} & -\hat{\Rot{}{}} & - S(\mathbf{\hat{v}})\hat{\Rot{}{}} \\
    \Zeros{18}{3} & \Zeros{18}{3} & \Zeros{18}{3} &  \Zeros{18}{12} & \Zeros{18}{3} &  \Zeros{18}{3}
\end{bmatrix}
\end{equation}

\begin{equation}
\mathbf{L}_c =
\begin{bmatrix}
\hat{\Rot{}{}} & S(\mathbf{\hat{p}})\hat{\Rot{}{}} & \Zero{3}  & \Zero{3} & \Zero{3} & \Zero{3} & \Zero{3} & \Zero{6} \\
\Zero{3} & \hat{\Rot{}{}} & \Zero{3}  & \Zero{3} & \Zero{3} & \Zero{3} & \Zero{3} & \Zero{6} \\
\Zero{3} &  S(\mathbf{\hat{v}}) \hat{\Rot{}{}} & \hat{\Rot{}{}}  & \Zero{3} & \Zero{3} & \Zero{3} & \Zero{3} & \Zero{6} \\
\Zero{3} &  \Zero{3} & \Zero{3}  &  \mathbf{\hat{Z}}_{LF} &   S(\mathbf{\hat{d}}_{LF})\mathbf{\hat{Z}}_{LF} & \Zero{3} & \Zero{3} & \Zero{6} \\
\Zero{3} &  \Zero{3} & \Zero{3}  & \Zero{3} &  \mathbf{\hat{Z}}_{LF} & \Zero{3} & \Zero{3} & \Zero{6} \\
\Zero{3} &  \Zero{3} & \Zero{3}  & \Zero{3} & \Zero{3} & \mathbf{\hat{Z}}_{RF} & S(\mathbf{\hat{d}}_{RF})\mathbf{\hat{Z}}_{RF} & \Zero{6} \\
\Zero{3} &  \Zero{3} & \Zero{3}  & \Zero{3} & \Zero{3} & \Zero{3} & \mathbf{\hat{Z}}_{RF} & \Zero{6} \\
\Zero{3} &  \Zero{3} & \Zero{3}  & \Zero{3} & \Zero{3} & \Zero{3} & \Zero{3} & \I{6} 
\end{bmatrix},
\end{equation}
with the noise vector is defined as, 
\begin{equation}
\label{eq:chap:cd-ekf-rie-prop-noise}
\mathbf{w} \triangleq \text{vec}(\Zeros{3}{1},\ \noiseGyro{B}, \noiseAcc{B}, \noiseLinVel{LF}, \noiseAngVel{LF}, \noiseLinVel{RF}, \noiseAngVel{RF}, -\noiseBias{B} ).
\end{equation}
\end{lemma}
A detailed derivation of the nonlinear error propagation in Eq. \eqref{eq:eq:chap:cdekf-sys-rie-dynamics} as a proof of Lemma \ref{lemma:chap:diligent-kio:cdekf-rie-error-dyn} is provided in the Appendix \ref{appendix:chapter:nonlinear-err-dyn-cdekf-rie}.
It must be noted that the matrix $\mathbf{L}_c$ is simply the adjoint matrix $\gadj{\Xhat}$ of the state representation $\mathscr{M}$.
When the biases are not considered within the system, it can be seen the error propagation $\mathbf{F}_c$ becomes time-invariant (first 21 rows and columns of $\mathbf{F}_c$).
This is due to fact that, when the biases are not considered, the overall system defined in Eq. \eqref{eq:eq:chap:cdekf-sys-dynamics} obeys the property of group-affine dynamics in Eq. \eqref{eq:chap02:invekf-group-affine-dynamics} leading to an autonomous error propagation described in Theorem \ref{theorem:chap02:autonomous-error-dynamics}, as seen in Section \ref{sec:chapter02:invekf-matrix-lie-groups}.
Nevertheless, the overall error system is only dependent on the state trajectory through the noise and bias errors.
The continuous time filter equations for the propagation step becomes, \looseness=-1
\begin{equation}
    \begin{split}
        & \frac{d}{dt} \Xhat_t = f_{\mathbf{u}_t}(\Xhat_t), \\
        &  \frac{d}{dt} \cov_t = \mathbf{F}_c \cov_t + \cov_t \mathbf{F}_c^T + \mathbf{\hat{Q}}_c.
    \end{split}
\end{equation}
where $\mathbf{\hat{Q}}_c = \mathbf{L}_c\ \text{Cov}(\mathbf{w})\ \mathbf{L}_c^T$. For the software implementation of the continuous-time equations, we discretize the continuous dynamics using a zero-order hold on the inputs with a sampling period $\Delta T$. The discrete system dynamics takes the form of Eq. \eqref{eq:chap:diligent-kio-sys-dynamics}, while the uncertainty propagation is approximated as, \looseness=-1
\begin{equation}
    \begin{split}
        & \mathbf{F}_k = \exp(\mathbf{F}_c \Delta T) \approx \I{p} + \mathbf{F}_c \Delta T, \\
        & \mathbf{Q}_k = \mathbf{F}_k \mathbf{\hat{Q}}_c \mathbf{F}_k^T \Delta T,\\
        & \mathbf{P}_\kpred = \F_\kcurr\ \cov_\kprior\ \F_\kcurr^T + \Q_\kcurr.
    \end{split}
\end{equation}

The filter observations are the same as those defined in Section \ref{sec:chap:diligent-kio-meas-model} however due to the choice of right invariant error, the measurement update follows the procedure defined in Section \ref{sec:chap:diligent-kio-rie}, with the measurement model $\mathbf{H}_\knext$ defined in Eq. \eqref{eq:chap:diligent-kio-rie-measmodeljacobianLF} and the state reparametrization defined in Eq. \eqref{eq:chap02:ekf-rie-state-reparam}.


\subsection{CODILIGENT-KIO}
\label{sec:chap:cd-ekf-lie}

CODILIGENT-KIO is a continuous-discrete EKF on matrix Lie groups with a left invariant error formulation and the consideration of non-invariant observations evolving over matrix Lie groups.
This estimator follows the same procedure from Section \ref{sec:chap:cd-ekf-rie} but varies in the definition of the linearized error system and the measurement model Jacobian.

The left-invariant error $\eta^L_\mathscr{M} = \Xhat^{-1} \X $ can be written as,
\begin{equation}
    \label{eq:eq:chap:cdekf-sys-lie}
    \begin{split}
    \eta^L_\mathscr{M} &= \left(-\mathbf{\hat{R}}^T\mathbf{p},\ \mathbf{\hat{R}}^T\Rot{}{},\ -\mathbf{\hat{R}}^T\mathbf{v},\  -\mathbf{\hat{Z}}_{LF}\mathbf{\hat{d}}_{LF},\ \mathbf{\hat{Z}}_{LF}^T,\  -\mathbf{\hat{Z}}_{RF}^T\mathbf{\hat{d}}_{RF},\ \mathbf{\hat{Z}}_{RF}^T ,\ -\hat{\bias} \right)_{\mathscr{M}}  \circ \\
    & \quad \quad \left(\mathbf{p}, \mathbf{R}, \mathbf{v}, \mathbf{d}_{LF}, \mathbf{Z}_{LF}, \mathbf{d}_{RF}, \mathbf{Z}_{RF}, \bias \right)_{\mathscr{M}}  \\
    &= \big( \hat{\Rot{}{}}^T(\mathbf{p} - \mathbf{\hat{p}}), \
             \hat{\Rot{}{}}^T \Rot{}{}, \ 
             \hat{\Rot{}{}}^T(\mathbf{v} - \mathbf{\hat{v}}), \ \\
             &  \quad \quad \mathbf{\hat{Z}}_{LF}^T(\mathbf{d}_{LF} - \mathbf{\hat{d}}_{LF}), \
             \mathbf{\hat{Z}}_{LF}^T \mathbf{Z}_{LF}, \ \\
             & \quad \quad \mathbf{\hat{Z}}_{RF}^T (\mathbf{d}_{RF} - \mathbf{\hat{d}}_{RF}), \
             \mathbf{\hat{Z}}_{RF}^T \mathbf{Z}_{RF} , \
             \bias - \hat{\bias}
             \big)_{\mathscr{M}}.
    \end{split}
\end{equation}

Given, $\eta^L_\mathscr{M} = \gexphat{\mathscr{M}}\left(\err\right) \approx \I{n} + \ghat{\mathscr{M}}{\err}$, the overall non-linear error dynamics $\frac{d}{dt} \eta^L_t = \frac{d}{dt} \left(\Xhat^{-1} \X \right)$ can be summarized as,
\begin{equation}
\label{eq:eq:chap:cdekf-sys-lie-dynamics}
    \begin{split}
    &  \dot{\eta}_\mathbf{p} =\ \err_\mathbf{v} - S(\yGyro{A}{B} - \mathbf{\hat{b}}_g)\err_\mathbf{p}, \\
    &  \dot{\eta}_\mathbf{R} =\ S(-\err_\biasGyro - \noiseGyro{B} ), \\
    & \dot{\eta}_\mathbf{v} =\   - S({\yGyro{A}{B}} - \biasGyroHat) \err_\mathbf{v} - S(\yAcc{A}{B} - \biasAccHat)  \err_\mathbf{R} -\err_\biasAcc -\noiseAcc{B}, \\
    & \dot{\eta}_{\mathbf{d}_F} =\ - \noiseLinVel{F},\\
    & \dot{\eta}_{\mathbf{Z}_F} =\ -S(\noiseAngVel{F}),\\
    & \dot{\eta}_\bias =\ \noiseBias{B}.
    \end{split}
\end{equation}

\begin{lemma}
\label{lemma:chap:diligent-kio:cdekf-lie-error-dyn}
With the non-linear error dynamics in Eq. \eqref{eq:eq:chap:cdekf-sys-lie-dynamics}, the definition of the error vector given in Eq. \eqref{eq:chap:diligent-kio-state-vector}, and the definition of noise vector in Eq. \eqref{eq:chap:cd-ekf-rie-prop-noise}, the linearized error dynamics becomes $\dot{\err} = \mathbf{F}_c \err - \mathbf{L}_c \mathbf{w}$ with $\mathbf{L}_c = \I{27}$ (which is of the form described in Eq. \eqref{eq:chap02:invekf-linearized-lie-dynamics}).
Considering $\yGyroBar{A}{B} = \yGyro{A}{B} - \biasGyroHat$ and $\accBar{A}{B} = \yAcc{A}{B} - \biasAccHat$, the linearized error propagation matrix $\mathbf{F}_c$ becomes,
\begin{equation}
\mathbf{F}_c =
\begin{bmatrix}
    -S(\yGyroBar{A}{B}) & \Zero{3} & \I{3} & \Zeros{3}{12} & \Zero{3} & \Zero{3} \\
    \Zero{3} & \Zero{3} & \Zero{3} &  \Zeros{3}{12} & \Zero{3} & -\I{3} \\
    \Zero{3} & -S(\accBar{A}{B}) & -S(\yGyroBar{A}{B}) &  \Zeros{3}{12} & -\I{3} & \Zero{3} \\
    \Zeros{18}{3} & \Zeros{18}{3} & \Zeros{18}{3} &  \Zeros{18}{12} & \Zeros{18}{3} &  \Zeros{18}{3}
\end{bmatrix}
\end{equation}
\end{lemma}
A detailed derivation of the non-linear error propagation in Eq. \eqref{eq:eq:chap:cdekf-sys-lie-dynamics} is provided as the proof of Lemma \ref{lemma:chap:diligent-kio:cdekf-lie-error-dyn} in the Appendix \ref{appendix:chapter:nonlinear-err-dyn-cdekf-lie}.
It can be observed that the error propagation depends not only on the bias errors and the noise terms, but also the IMU measurements compensated by estimated IMU biases, leading to a time-varying linearized error system.

The filter observations, the measurement model jacobian and the state reprametrization equations are the same as those defined in Section \ref{sec:chap:diligent-kio-meas-model}.

\subsection{Qualitative Comparison}
In this subsection, we describe the comparison of the DILIGENT-KIO and its variants with InvEKF (point foot filter, invariant EKF on matrix Lie groups proposed by \cite{hartley2020contact}) and OCEKF (flat foot filter, observability constrained EKF using a quaternion+Euclidean representation proposed by \cite{rotella2014state}).
The variants of DILIGENT-KIO include DILIGENT-KIO-RIE, CODILIGENT-KIO, and CODILIGENT-KIO-RIE.
Out of the 6 estimators, 4 have the properties to handle appropriate consistency management, retaining nominal observability conditions throughout their operation.
This implies that the filter will not become overconfident about estimates along the unobservable directions and not result in false observability.
This can be shown by visualizing the error and the uncertainty envelopes about the unobservable directions as seen in Figure \ref{fig:chap:diligent-variants-comparison}. 

\begin{itemize}
    \item OCEKF uses observability-based rules to impose constraints on the rank of the null space of the observability matrix to choose linearization points for the filter computations. This choice of linearization points allows maintaining the nominal rank value of the unobservable subspace.
    \item InvEKF exploits symmetries using the Lie group structure, providing an implicit consistency for the filter owing to the autonomous error evolution property and invariant observation structure. \looseness=-1
    \item DILIGENT-KIO and DILIGENT-KIO-RIE do not exploit the autonomous error evolution property, meaning the evolution error is not independent of the state trajectory. This leads to a lack of inherent consistency properties in the filter and requires to explicitly consider this in filter design either through observability rules or first estimate Jacobians to lead to a reliable filter design.
    \item CODILIGENT-KIO and CODILIGENT-KIO-RIE borrow the autonomous error propagation property from InvEKF while retaining a non-invariant observation structure to account for observations evolving over matrix Lie groups.
\end{itemize}

\begin{figure}[!b]
\begin{subfigure}{\textwidth}
\centering
\includegraphics[scale=0.45]{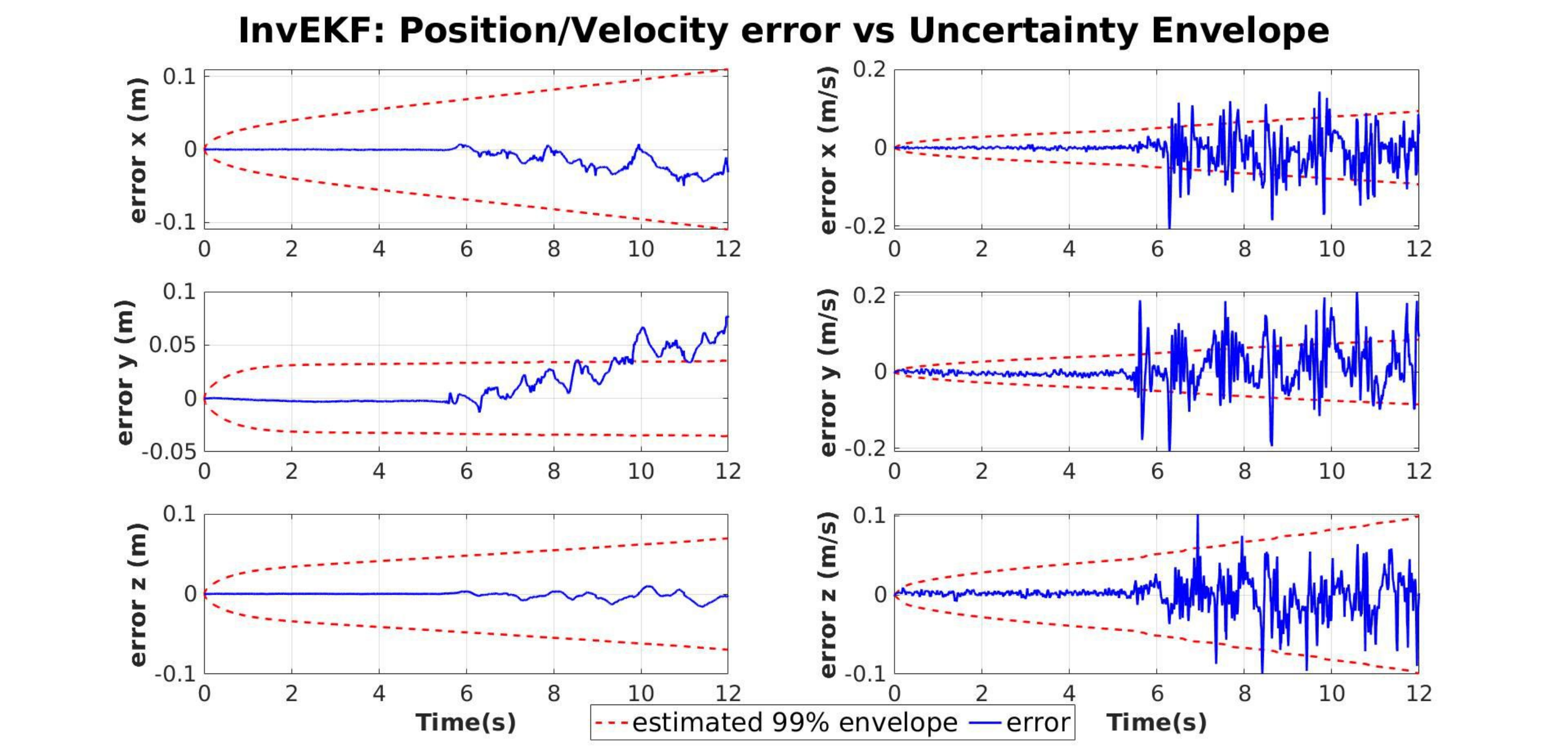}
	\end{subfigure}
\begin{subfigure}{\textwidth}
\centering
\includegraphics[scale=0.45]{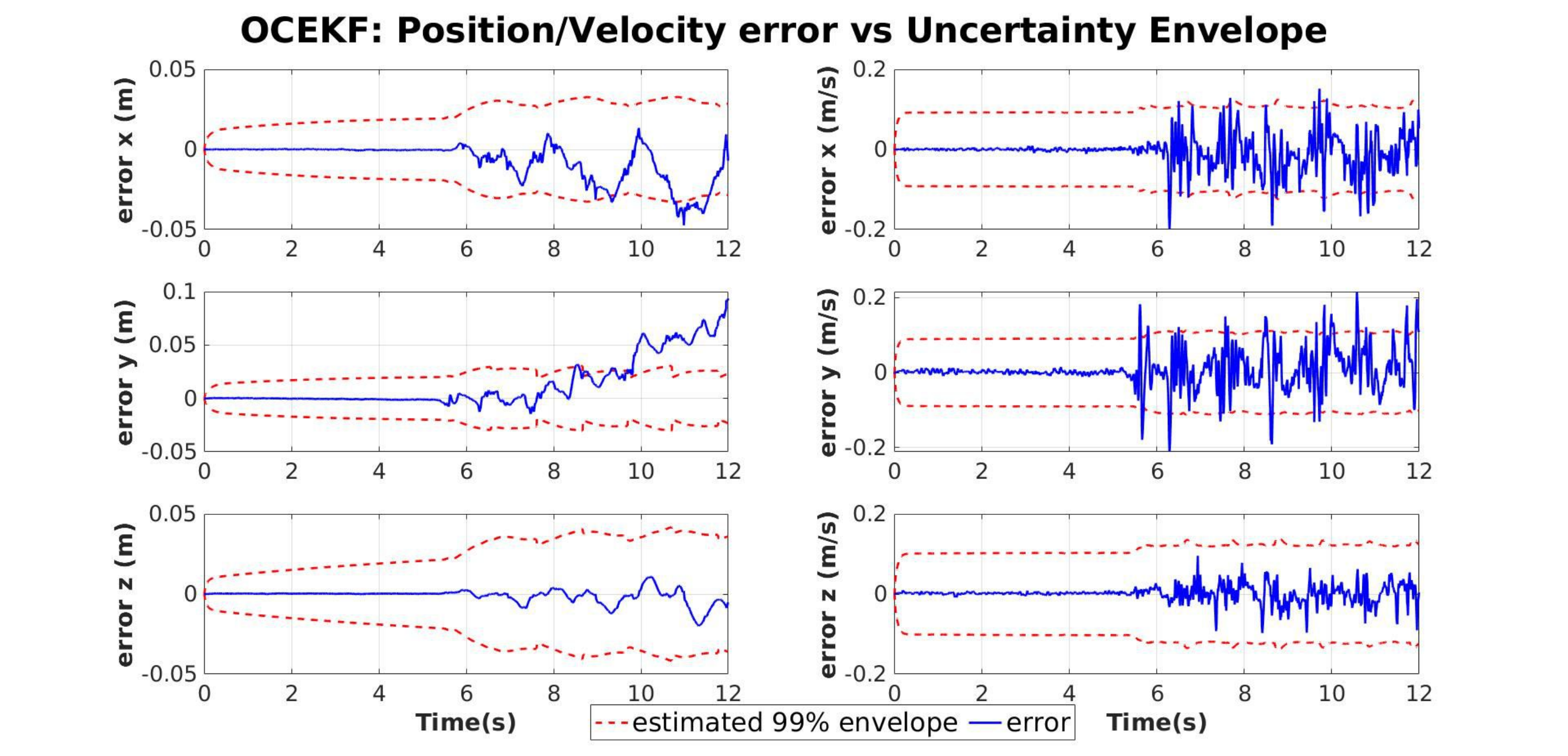}
	\end{subfigure}
\begin{subfigure}{\textwidth}
		\centering
\includegraphics[scale=0.45]{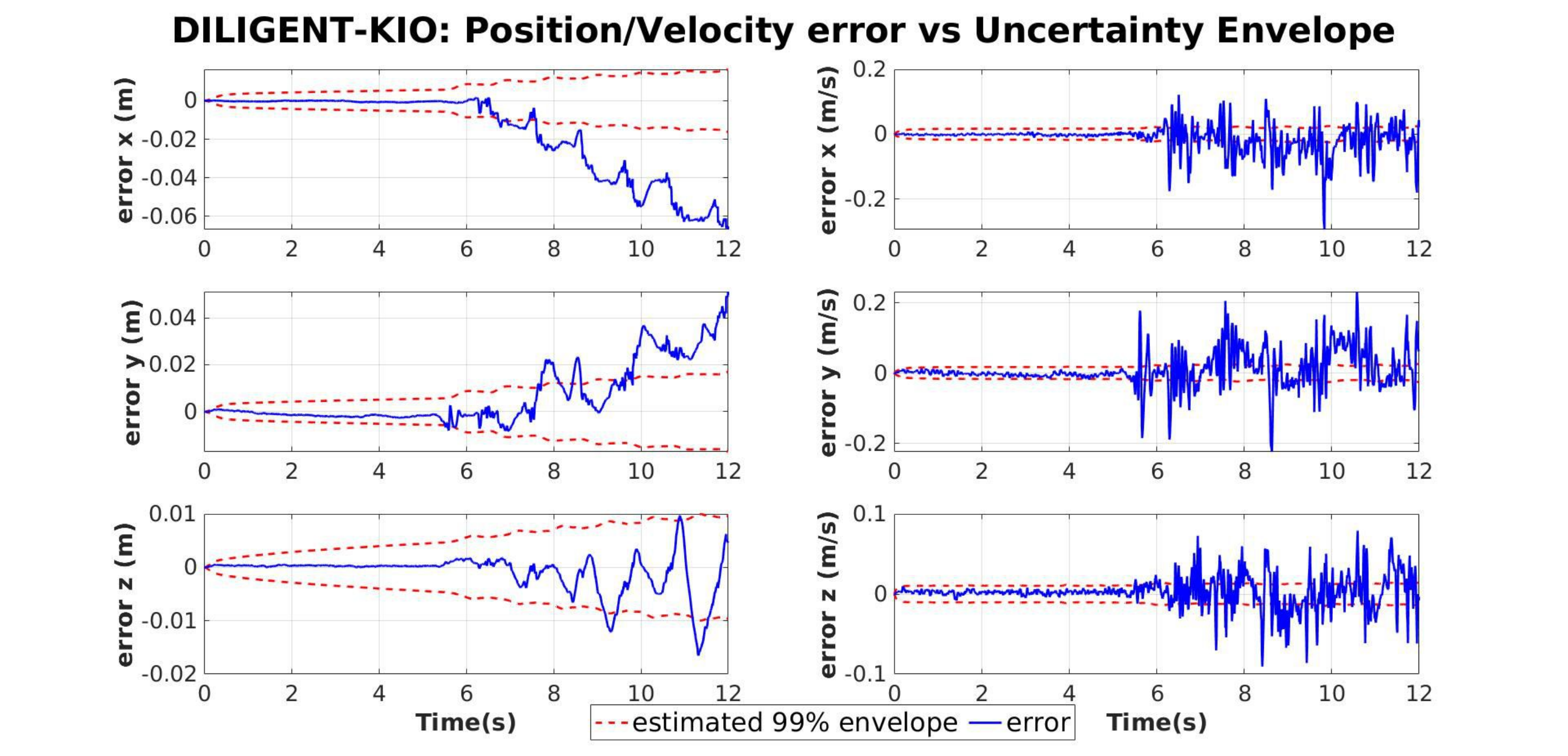}
	\end{subfigure}
\end{figure} 
\begin{figure}[tbp]\ContinuedFloat
\begin{subfigure}{\textwidth}
		\centering
\includegraphics[scale=0.45]{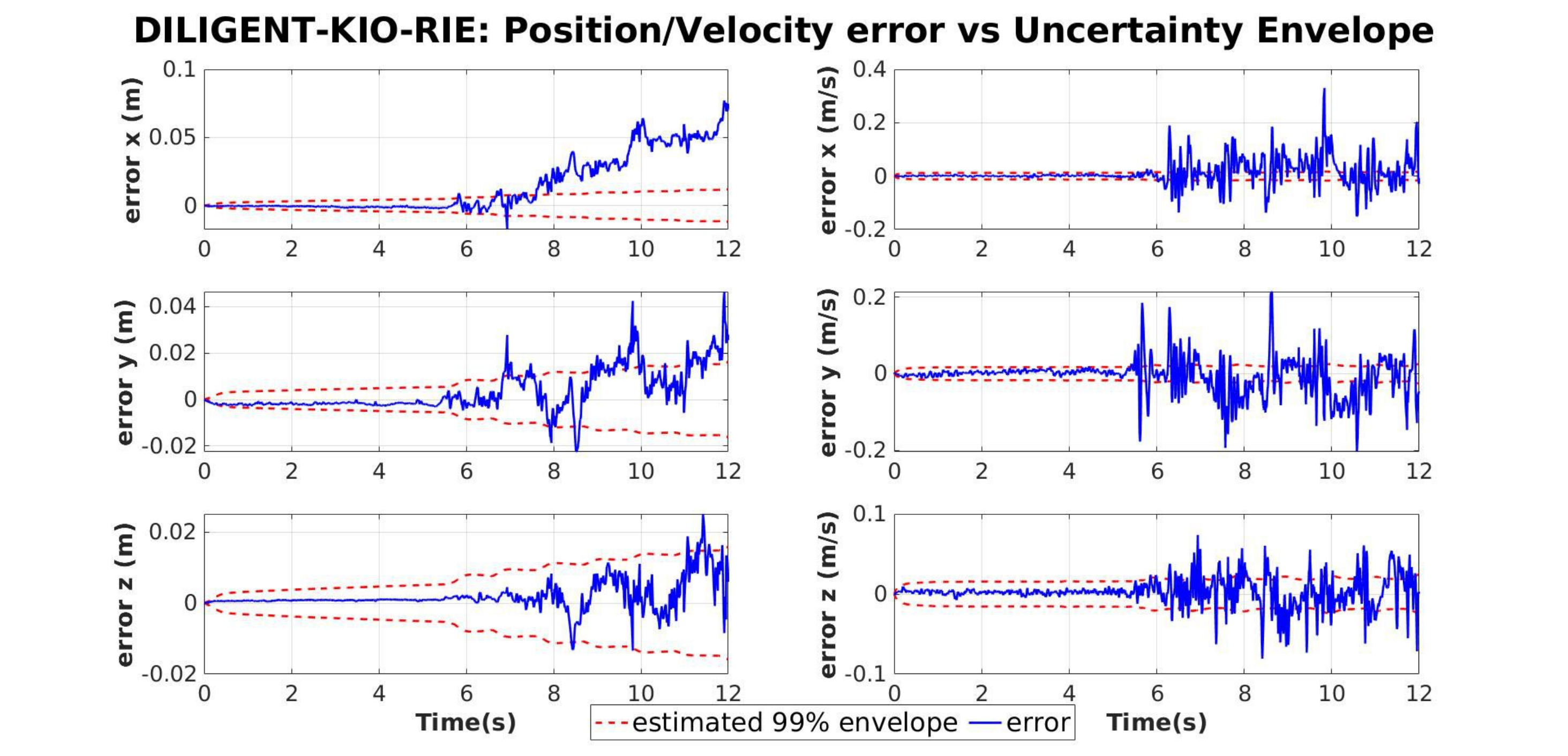}
	\end{subfigure}
\begin{subfigure}{\textwidth}
		\centering
\includegraphics[scale=0.45]{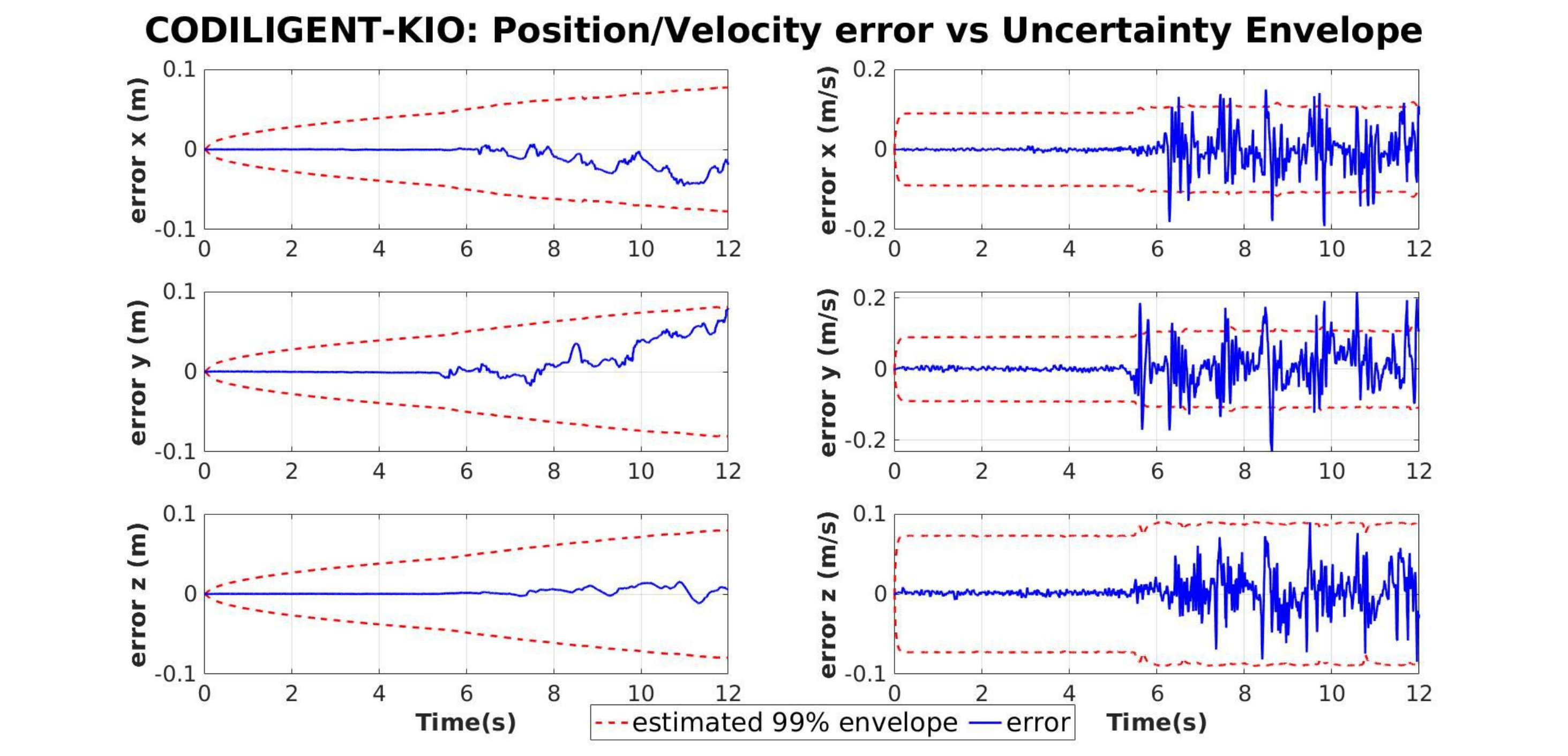}
	\end{subfigure}
\begin{subfigure}{\textwidth}
		\centering
\includegraphics[scale=0.45]{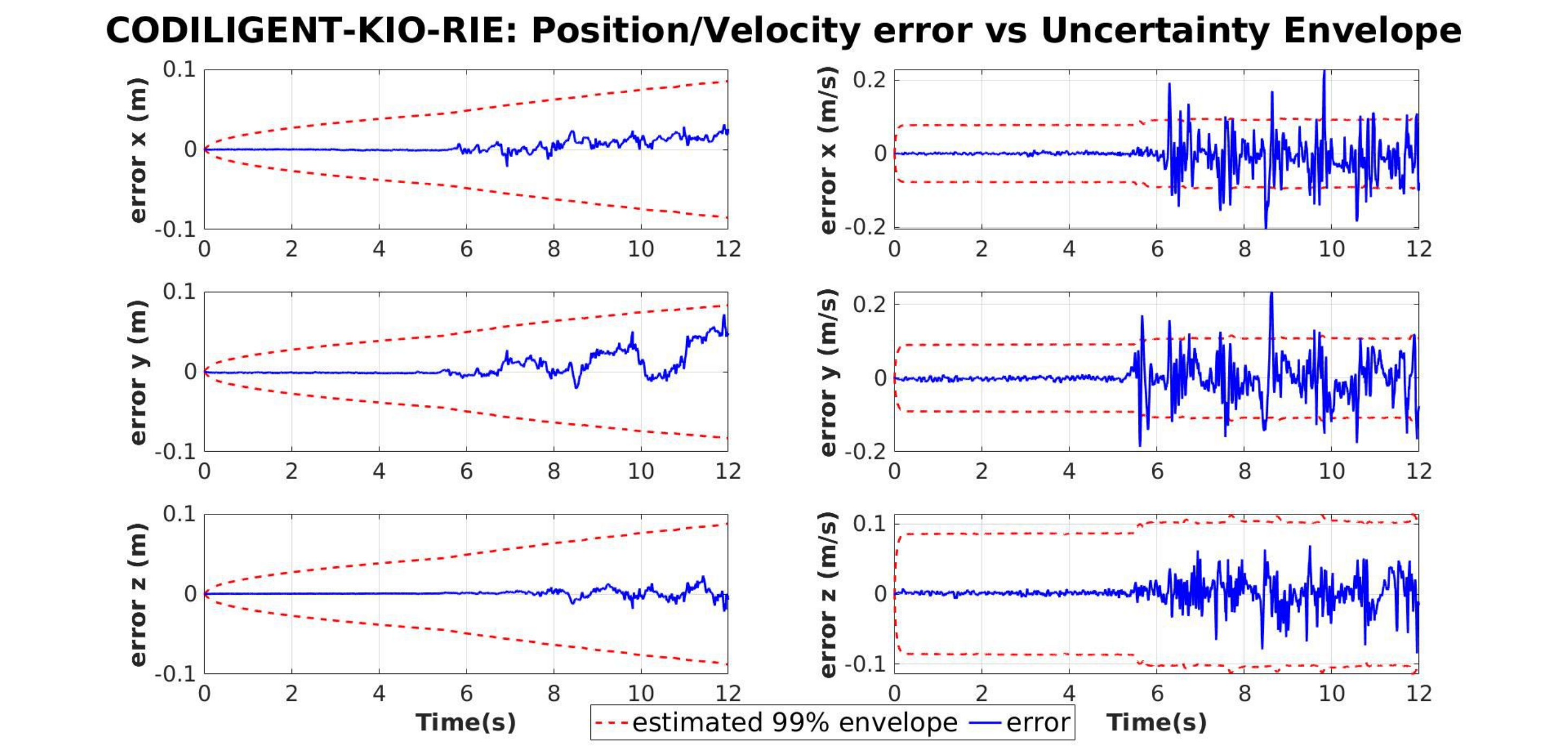}	
\end{subfigure}	
\caption{Comparison of evolution of position and velocity errors (blue lines) respectively along with 99$\%$ estimated uncertainty envelope (red lines) for state-of-the-art methods.}
\label{fig:chap:diligent-variants-comparison}
\end{figure}

The plots in Figure \ref{fig:chap:diligent-variants-comparison} show the error (in blue) computed using the invariant trajectory error definitions from Vicon and estimator trajectories in the unobservable position directions and observable velocity directions.
They also show the 99$\%$ estimated uncertainty envelope (in red) obtained from the estimated state error covariance matrices computed by the filters.

The uncertainty envelope is bound to increase over time in the unobservable directions, while it converges to track the error in the observable directions.
It can be seen that for DILIGENT-KIO and DILIGENT-KIO-RIE, the errors in both the unobservable position direction and observable velocity directions do not remain within the 99$\%$ estimated uncertainty envelope of the filter.
This shows that both these filters become overconfident in their estimation.
InvEKF and OCEKF try to cope considerably well with the observable directions, leading to the error of the base velocity and position in x- and z-directions being bounded within the envelope. 
However, it must be noted that InvEKF used here does not incorporate rotation constraints imposed by the flat foot of the humanoid robot, nevertheless, this estimator offers more reliable error statistics than the flat-foot filter counterparts, owing to its properties.
OCEKF also performs comparably better but requires an explicit computation of the linearization point based on the null space of the observability matrix.

The continuous-discrete variants of DILIGENT-KIO, which uses continuous system dynamics and discrete observations combine the autonomous propagation through non-linear error dynamics from the InvEKF with the non-invariant observations of the DILIGENT-KIO, leading to consistent tracking of error statistics.
The continuous system dynamics is discretized using a zero-order hold with time period $\Delta t$ for discrete-time implementation. 

The better performance of CODILIGENT-KIO-RIE than CODILIGENT-KIO is related to the linearized error propagation matrices. When IMU biases are not considered within the state, RIE variant matrices depend only on the gravity vector making it time-invariant but the propagation matrices of the LIE variant depend on IMU measurements making it time-varying.
But this is not true anymore when IMU biases are considered, in which case in the direction of bias errors, the CODILIGENT-KIO-RIE becomes state-dependent due to the action of an adjoint matrix that is required to transform the bias errors in the local frame onto the global frame.
CODILIGENT-KIO remains invariant in these directions since the bias errors are already expressed in the local frames.  
If the biases are negligible, then we have a time-invariant propagation matrix for the former variant which offers better overall performance.

CODILIGENT-KIO and CODILIGENT-KIO-RIE suffer from a slightly higher absolute trajectory error (ATE) and relative pose error (RPE) than their fully discrete counterparts.
Although the CODILIGENT-KIO-RIE does not provide higher accuracy estimates in comparison with DILIGENT-KIO, it provides reliable error statistics which might lead to avoiding divergence of the filter in the long term.
To improve the trust in DILIGENT-KIO and DILIGENT-KIO-RIE, then a thorough observability analysis is required to investigate the null-space of the observability matrix for imposing constraints on the choice of linearization points.

The estimators ranked with respect to consistency (reliable uncertainty estimation) from most reliable to least reliable are CODILIGENT-KIO-RIE, CODILIGENT-KIO, OCEKF, InvEKF, DILIGENT-KIO-RIE, and DILIGENT-KIO.
The estimators ranked with respect to error metrics (ATE and RPE) from most reliable to least reliable are DILIGENT-KIO-RIE, DILIGENT-KIO, CODILIGENT-KIO-RIE, InvEKF, CODILIGENT-KIO, and OCEKF.

It is clear from this comparison that, as the practitioner, we must pose the following questions towards reliable filter design: \emph{1) how does the representation choice for state and observations affect the filter design? 2) how does the filter design vary with the choice of error? 3) how does the time-representation of the system dynamics affect the design?}
Based on these choices, the linearized error system may or may not be independent of the trajectory of the system state.
Further, a well-chosen matrix Lie group representation  allows for a design \cite{barrau2017invariant} with which, a wide range of nonlinear problems can lead to linear error equations through a correct parameterization of the error. 
Proper choices for these questions often helps in formulating an estimator with strong convergence and consistency properties. 
We notice that, the filters designed on matrix Lie groups that exploit a group-affine system dynamics structure tend to perform better as consistent estimators than the standard EKF and OCEKF counterparts. 
This information can clearly aid in the future improvements of the proposed methods, particularly in achieving autonomous error propagation and update for the proposed discrete filter DILIGENT-KIO.
\looseness=-1

\section{Conclusion}
\label{sec:chapter05:conclusion}
This chapter presented a proprioceptive floating base estimation that was proposed using extended Kalman filtering on matrix Lie groups by considering the evolution of the state and the observations over distinct Lie groups.
This proprioceptive estimator is further demonstrated to be easily extended for an absolute humanoid localization with the inclusion of measurements obtained from the cameras.
The proposed filter was shown to perform better, in terms of convergence, than the observability-constrained quaternion-based extended Kalman filter (flat foot filter).
The latter is also a discrete EKF on Lie groups, however differs from the proposed estimator in terms of uncertainty propagation.
It must be noted that the theory of autonomous error propagation (\cite{barrau2017invariant, hartley2020contact}) and update (\cite{phogat2020discrete}) or observability-based rules (\cite{bloesch2013state, huang2010observability}) were not \emph{explicitly} exploited within this framework. 
This could \emph{potentially} lead to inconsistencies in the filter.
Nevertheless, the proposed estimator exhibits strong convergence properties with a large basin of attraction. 
Different variants of the proposed filter based on the choice of the error and time-representation for the system dynamics are also investigated.

It is important to point out where this filter is lacking which might help future research directions.
Firstly, a thorough observability and consistency analysis is lacking. 
Such analysis becomes even more important for filters whose linearized system matrices are state-dependent or input-dependent in order to analyze the permanent trajectories around which this EKF behaves as a stable observer.
One alternative that provides an implicitly consistent filter can be possible by extending DILIGENT-KIO to handle autonomous error evolution while handling invariant observations evolving in matrix Lie groups by combining the recent works presented by \cite{qin2020novel} and \cite{phogat2020discrete, phogat2020invariant}.

More rigorous handling of how the kinematics errors are propagated through the serial chain of the lower limbs can be considered using the concept of compounding uncertainties of rigid body transformations.
Such an approach might allow accounting for joint modeling errors such as backlash for tuning the kinematic noise parameters (\cite{su1992manipulation}).

When using the filter with exteroceptive sensors,  an outlier rejection method may be incorporated to robustify the estimator performance.
Further, in order to prevent drifts due to IMU integration in the absence of contacts, the DILIGENT-KIO filter equations can be modified to consider IMU preintegration within the filtering framework through constrained EKF approaches.

Finally, instead of relying on constant prediction models and dynamic covariance scaling for the evolution of the feet states, efforts need to be made to reformulate the filter equations by considering a hybrid dynamical system and accounting for a Saltation matrix-based jump maps (\cite{kong2021salted}) for a proper error-propagation between contact making and breaking events.

\chapter{Human Motion Estimation using Wearable Technologies}
\label{chap:human-motion}

Unlike robotic systems, when working with human agents, we do not have access to direct sensor measurements of their internal joint configuration. i.e. joint positions and velocities.
Motion tracking algorithms are employed to achieve the reconstruction of body movements from a set of sensor measurements that potentially describe the kinematics of the human body parts (usually called as \emph{segments} or \emph{links}).
These algorithms aim to find the joint configuration of the human model given the set of kinematic measurements called as \emph{targets} which are ideally composed of task space measurements such as link poses and 6D velocities (or twists). 
In this chapter, we present an approach for kinematic-free joint state estimation and floating base estimation through the use of multiple distributed inertial sensors and force-torque measurements.
We extend the work of \cite{rapetti2020model} for human motion estimation with the inclusion of contact detection and a floating base estimation module to achieve whole-body human motion reconstruction. 

The problem of human motion estimation from a set of distributed inertial sensors and sensorized shoes is described in Section \ref{sec:chap:hbe:motion-est} along with a proposed system architecture.
We then proceed to describe each sub-block of the proposed system architecture in detail across Section \ref{sec:chap:human-motion:dynIK}, Section \ref{sec:chap:human-motion:cop-contact} and Section \ref{sec:chap:human-motion:ekf-base}.
Finally experimental results validating the use-case of the proposed architecture for robot and human motion experiments are described in Section \ref{sec:chap:human-motion:experiments} with concluding remarks for the chapter made in Section \ref{sec:chap:human-motion:conclusion}.

\section{Motion Estimation using Distributed IMUs and Sensorized Shoes}
\label{sec:chap:hbe:motion-est}
In Section \ref{sec:modeling:ik}, we described the general problem of inverse kinematics assuming the availability of full kinematic data from the target measurements.
In this section, we describe the problem of motion estimation specifically for a human wearing a sensorized suit composed of distributed IMUs and a pair of sensorized shoes.
While the distributed IMUs are used for motion reconstruction, the sensorized shoes which measure contact wrenches are used for contact detection.

The target measurements that are available through state-of-the-art wearable technologies like a sensor suit are constituted by orientation and angular velocity measurements from a set of IMUs attached to various parts in a distributed manner over the body (\cite{roetenberg2009xsens}). A few among many challenges involved in the reconstruction of full human motion from a set of sparsely distributed IMUs attached to the body are,  \looseness=-1
\begin{itemize}
    \item the modeling of the human body as a free-floating, articulated, multi-body system which requires reliable estimation of the floating base pose and velocity, and
    \item the availability of only partial target measurements from the distributed IMUs.
\end{itemize}
Consider an IMU $S_i$ attached to one of the links $L_i$ of the body.
The sensor $S_i$ might have its own estimation algorithms running internally and may use a choice of inertial frame $A_{S_i}$ distinct from a desired inertial frame $A$ we would like to define, with respect to which we want to express the measurements from all the sensors. 
The target orientations $\Rot{A}{{L_i}}$ can be expressed using the absolute orientation measurements $\yRot{{A_{S_i}}}{{S_i}}$ from the IMU as $\Rot{A}{{L_i}} = \Rot{A}{{A_{S_i}}}\ \yRot{{A_{S_i}}}{{S_i}} \ \Rot{{S_i}}{{L_i}}$ and the target angular velocities can be expressed using the gyroscope measurements as $\omegaRightTriv{A}{{L_i}} = \Rot{A}{{L_i}} \yGyro{A}{{L_i}}$, where $\Rot{{S_i}}{{L_i}}$ defines the rotation of the sensor link's frame with respect to sensor frame and $\Rot{A}{{A_{S_i}}}$ is a calibration matrix used to express the measurements of sensor $S_i$ in a common inertial frame $A$.
Thus, the problem of motion estimation requires to solve for the floating base pose $\Transform{A}{B}$ and velocity $\twistMixedTriv{A}{B}$ of the human along with their joint configuration $\jointPos$ and velocity $\jointVel$ using a set of absolute target orientations $\Rot{A}{{L_i}}$ and target angular velocities $\omegaRightTriv{A}{{L_i}}$ obtained from each of the sparsely distributed IMUs, as seen in Eq. \eqref{chap:human-motion:ik-problem}.


The reduced \emph{pose target tuple} $\mathbf{x}(t)$ and the \emph{velocity vector} $\mathbf{v}(t)$ from Eq. \eqref{eq:ik:tuples} then become,
\begin{equation}
\mathbf{x}(t) \triangleq 
\begin{pmatrix} 
\Pos{A}{{B}}(t) \\ \Rot{A}{{B}}(t) \\ \Rot{A}{{L_1}}(t) \\ \vdots \\ \Rot{A}{{L_{n_L}}}(t)
\end{pmatrix}, \quad
\mathbf{v}(t) \triangleq 
\begin{bmatrix} 
\oDot{A}{{B}}(t) \\ \omegaRightTriv{A}{{B}}(t) \\  \omegaRightTriv{A}{{L_1}}(t) \\ \vdots \\ \omegaRightTriv{A}{{L_{n_L}}}(t)
\end{bmatrix}.
\end{equation}
It must be noted that the base position $\Pos{A}{{B}}(t)$ and linear velocity $\oDot{A}{{B}}(t)$ are not directly measured but are passed as quantities estimated from the previous time-instant. \looseness=-1

\subsubsection{Assumptions}
It can be noticed that a few assumptions have been implicitly made regarding the orientation and angular velocity measurements in the problem formulation of motion estimation using distributed IMUs.
We make these assumptions to simplify the development of the human motion estimation problem. \looseness=-1
\begin{assumption}[Extrinsic calibration of links to IMUs]
\label{assumption:chap:human-motion:extrinsic-imus}
The extrinsic calibration of the IMU, with respect to the segment or the link it is rigidly attached to, has already been performed implying that the rotation $\Rot{{L_i}}{{S_i}}$ between the link frame $L_i$ and the IMU frame $S_i$ is constant and known. \looseness=-1
\end{assumption}

\cite{latella2019human} uses an offline estimation procedure for determining the pose of the IMU sensors with respect to the body reference frame.
This estimation procedure gathers known body pose, linear and angular acceleration, and angular velocity of the body with respect to the inertial frame and proper acceleration measured by IMU sensor as measurements within an over-determined system of linear equations and solves such a system in the least-squares sense to obtain the IMU sensor poses with respect to the body frame.

\begin{assumption}[Inertial Frame for Multiple IMUs]
\label{assumption:chap:human-motion:inertial-frame-imus}
The complete set of distributed IMUs are calibrated to express the absolute orientation measurements $\Rot{A}{{L_i}}$ with respect to the same inertial frame $A$. \looseness=-1
\end{assumption}

\subsubsection{Multi-Sensor Calibration}
For the sake of completeness, we briefly describe the problem that needs to be handled before solving human motion estimation if Assumptions \ref{assumption:chap:human-motion:extrinsic-imus} and \ref{assumption:chap:human-motion:inertial-frame-imus} are not considered.

The target orientations obtained from the IMU measurements are described as, \looseness=-1
$$ 
\Rot{A}{{L_i}} = \Rot{A}{{A_{S_i}}}\ \Rot{{A_{S_i}}}{{S_i}} \ \Rot{{S_i}}{{L_i}},
$$
where, $\Rot{A}{{A_{S_i}}}$ and $\Rot{{S_i}}{{L_i}}$ are either unknown or uncertain rotations, while $\Rot{{A_{S_i}}}{{S_i}}$ and $\Rot{A}{{L_i}}$ are rotations which can be either measured or computed by enforcing some model constraints on the human body, such as known joint configurations of the body through which link poses with respect to the inertial frame $A$ can be obtained.
Thus, the calibration procedure involves computing the unknown rotations using the known quantities through a $\A \X  = \Y \B$ type calibration procedure with unknowns $\{\X, \Y\}$ and knowns $\{\A, \B\}$ (\cite{shah2012overview}).
$$ 
\Rot{{L_i}}{A} \Rot{A}{{A_{S_i}}} = \Rot{{L_i}}{{S_i}} \Rot{{S_i}}{{A_{S_i}}}.
$$

It must be emphasized that if the measurement set also includes linear accelerometer measurements from the IMUs, then special care must be taken to extend this extrinsic calibration to account for transformation matrices evolving in $\SE{3}$ (instead of only rotations), to also properly consider the effects of Coriolis terms while expressing the accelerometer measurements in the desired frame.

\subsection{System Description}
\begin{figure}[!h]
\centering
\includegraphics[scale=0.5]{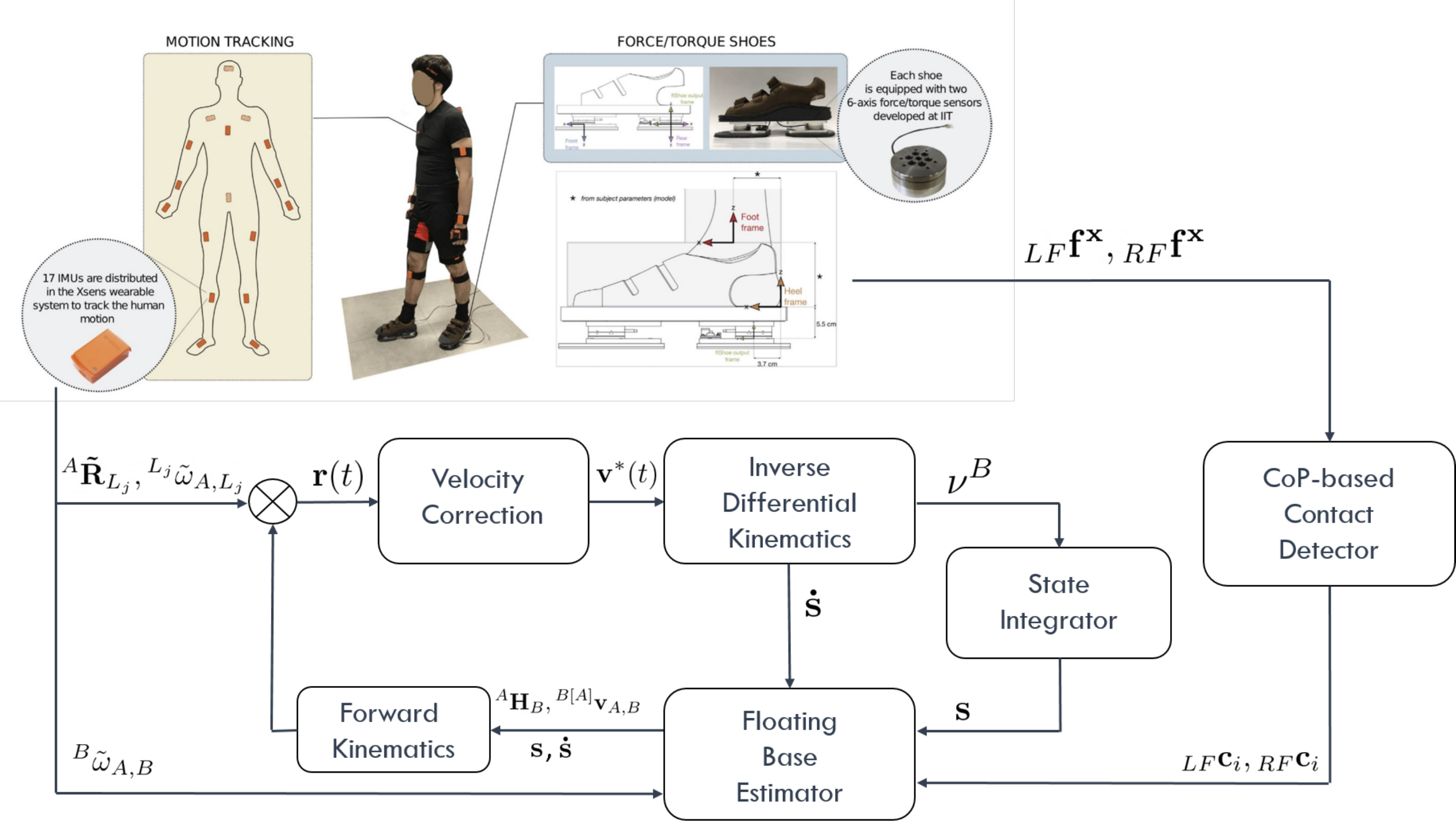}
\caption{Data flow diagram for human motion estimation.}
\label{fig:chap:human-motion:system-block-diagram}
\end{figure}

Instead of tackling the whole-body motion estimation through a monolithic non-linear optimization problem, we decompose it into two cascading stages of system state configuration estimation through a dynamical inverse kinematics approach followed by base pose and velocity correction through an extended Kalman filter on Lie groups.
This decomposition is done to handle the floating base estimation within a prediction-correction framework and be able to incorporate contact-aided kinematic measurements without the need for imposing constraints, while remaining computationally light-weight.
The overall system architecture for the human motion estimation using sparsely distributed IMUs and sensorized shoes is composed of three main blocks as shown in Figure \ref{fig:chap:human-motion:system-block-diagram}, \looseness=-1
\begin{itemize}
    \item Dynamical inverse kinematics optimization,
    \item Center of pressure based contact detection, and
    \item Extended Kalman Filter on Lie groups for floating base estimation.
\end{itemize}

The dynamical IK optimization block is used to compute the overall system configuration, which is then used as input to the EKF block along with foot contact states inferred from the contact detection module.
The EKF block estimates the floating base pose and velocity estimates which are subsequently passed back to the dynamical IK block as feedback. \looseness=-1

The dynamical optimization approach is rather different from the commonly used  instantaneous nonlinear optimization methods (described earlier in Eq. \eqref{chap:human-motion:nlopt-prob}) as,
\begin{equation*}
\begin{split} 
    \underset{\mathbf{q}(t_\kcurr), \nu(t_\kcurr)}{\text{minimize}} & \qquad \norm{\mathbf{K}_q\ \mathbf{r}_\mathbf{q}(\mathbf{q}(t_\kcurr), \mathbf{x}(t_\kcurr))} _2  + \norm{\mathbf{K}_\nu\ \mathbf{r}_\nu(\nu(t_\kcurr), \mathbf{v}(t_\kcurr))} _2 \\
    \text{subject to} & \qquad \mathbf{A} \begin{bmatrix} \jointPos(t_\kcurr) \\ \jointVel(t_\kcurr) \end{bmatrix} \leq \mathbf{b},
\end{split}
\end{equation*}
since the former does not aim to satisfy the kinematics system (described by Eq. \eqref{chap:human-motion:state-configuration}) as, 
\begin{equation*}
\begin{split}
    &\mathbf{x}(t) = h(\mathbf{q}(t)) \\
    &\mathbf{v}(t) = \mathbf{J}(\mathbf{q}(t)) \nu(t).
\end{split}
\end{equation*}
at every time-instant.
It rather aims to drive the state configuration $(\mathbf{q}(t), \nu(t))$ allowing the forward kinematics and differential kinematics of the system to converge towards the incoming target measurements in multiple time steps.
The contact detection block uses the contact wrench measurements from the sensorized shoes and a rectangular foot approximation for the foot geometry to infer the contact states of candidate contact points on the foot.
The EKF block handles filtering over matrix Lie groups employing right-invariant, left-invariant, and non-invariant observations in order to estimate the floating base pose and velocity along with the position of the candidate contact points with respect to an inertial frame.
The EKF block is activated after the dynamical Inverse Kinematics block has converged below a certain error threshold for a reliable state initialization, which is crucial for filter convergence.
Each of these blocks is discussed in detail in the subsequent sections. \looseness=-1

\section{Dynamical Inverse Kinematics Optimization}
\label{sec:chap:human-motion:dynIK}
In this subsection, we recall the dynamical IK optimization approach developed by \cite{rapetti2020model} which forms an important building block of the overall motion estimation pipeline.
The dynamical IK optimization draws inspiration from control theory and is constructed as a regulator of a velocity error computed through inverse differential kinematics and a system state integrator.
The system velocity $\nu(t)$ is the control input driving the defined residual errors towards zero.
The resulting error system, written in a general form as $\mathbf{\dot{e}} + k \mathbf{e} = \mathbf{0}$ is guaranteed to stabilize towards zero, since the solution of such differential equation is a decaying exponential function (\cite{olfati2001nonlinear}), allowing the system state to correspond to the target measurements with the rate of convergence depending on the choice of tunable gains.
The overall structure of Dynamical IK consists of three main steps, \looseness=-1
\begin{itemize}
    \item firstly the measured velocity $\mathbf{v}(t)$ is corrected using the forward kinematics error $\mathbf{r}(t)$ to produce an updated velocity vector $\mathbf{v}^{*}(t)$,
    \item the corrected velocity vector $\mathbf{v}^{*}(t)$ is then passed through the inverse differential kinematics module to compute the system velocity $\nu(t)$,
    \item and finally, the velocity $\nu(t)$ is integrated to obtain the system configuration $\mathbf{q}(t)$. \looseness=-1
\end{itemize}
These three steps establish a closed-loop regulator on the model-based forward kinematics, thereby, driving the estimated system configuration to the true configuration.

\subsection{Velocity Correction}
In order to define the corrections for the measured velocity vector $\mathbf{v}(t)$, we need to define residual errors between the measured pose target vector $\mathbf{x}(t)$ and the current system configuration $\mathbf{q}(t)$.
For the definition of errors concerning orientations, a rotation matrix parametrization is chosen and the corresponding residual error is obtained using a logarithm map on the rotation error, while the position errors are defined in the vector space.
Thus, collecting the pose target residuals in a residual vector $\mathbf{r}(\mathbf{q}(t), \mathbf{x}(t))$ gives us, \looseness=-1
\begin{equation}
\label{eq:chap:human-motion:dynik-pose-residual}
\mathbf{r}(\mathbf{q}(t), \mathbf{x}(t)) = 
\begin{bmatrix} 
\Pos{A}{{B}}(t) - h^p_{B}(\mathbf{q}(t)) \\ \gvee{\SO{3}}{(h^o_{O_B}(\mathbf{q}(t)))^T\ \Rot{A}{{B}}(t)} \\
\gvee{\SO{3}}{(h^o_{O_1}(\mathbf{q}(t)))^T\ \Rot{A}{{{L_1}}}(t)} \\ 
\vdots 
\\ 
\gvee{\SO{3}}{(h^o_{O_{n_L}}(\mathbf{q}(t)))^T\ \Rot{A}{{{L_{n_L}}}}(t)}
\end{bmatrix}.
\end{equation}
The velocity target residuals are collected in a velocity residual vector $\mathbf{u}(\mathbf{q}(t), \mathbf{x}(t))$ as,
\begin{equation}
\label{eq:chap:human-motion:dynik-vel-residual}
\mathbf{u}(\mathbf{q}(t), \mathbf{x}(t)) = \mathbf{v(t)} - \mathbf{J}(\mathbf{q}(t)) \nu(t).
\end{equation}
The Eqs. \eqref{eq:chap:human-motion:dynik-pose-residual} and \eqref{eq:chap:human-motion:dynik-vel-residual} can be seen as a dynamical system with system velocity $\nu(t)$ being the control input driving the residuals $\mathbf{r}(t)$ and $\mathbf{u}(t)$ to zero.
Considering a system,
\begin{equation}
\label{eq:chap:human-motion:dynik-err-system}
    \mathbf{u}(\mathbf{q}(t), \mathbf{x}(t)) + \mathbf{K}\ \mathbf{r}(\mathbf{q}(t), \mathbf{x}(t)) = \Zeros{(6 + 3n_L)}{1},
\end{equation}
with $\mathbf{K} \in \R^{(6 + 3n_L) \times (6 + 3n_L)}$ being a positive definite diagonal matrix. 
Then, $(\mathbf{r}, \mathbf{u}) = (\mathbf{0}, \mathbf{0})$ denotes an (almost) globally asymptotic stable equilibrium point for the system (see \cite[Corollary 1]{rapetti2020model}).
When the residual vectors are decomposed into linear and angular parts, the linear part of such a system follows a first-order autonomous linear system and for positive gains, the equilibrium point $(\mathbf{0}, \mathbf{0})$ is globally asymptotically stable for such a system.
For the angular part, the almost globally asymptotic stability of the equilibrium point $(\mathbf{0}, \mathbf{0})$ can be shown through an analysis using the Lyapunov function $V = \frac{1}{2} \text{tr}\left(\I{3} - (h^o_{O_B}(\mathbf{q}(t)))^T\ \Rot{A}{{B}}(t) \right)$ (see \cite{olfati2001nonlinear}).
The almost globally asymptotic stability is because the convergence of rotation error is not ensured for a particular initial condition for which we have $V = 2$ for any link, which would mean that target remains rotated at $\pi$ radians with respect to the link.
This corner case is not encountered in practice due to measurement noise and round-off errors.
Thus, the errors converge to zero for (almost) any initial condition on the system configuration, $\mathbf{q}(t_0)$ and the rate of convergence depends on the choice of the diagonal elements in matrix $\mathbf{K}$.

On substituting Eq. \eqref{eq:chap:human-motion:dynik-vel-residual} in Eq. \eqref{eq:chap:human-motion:dynik-err-system}, and expressing in discrete-time for a discrete-time implementation we have,
\begin{equation}
\label{eq:chap:human-motion:dynik-input}
     \mathbf{J}(\mathbf{q}(t_{k-1})) \nu(t_k) = \mathbf{v(t_k)} + \mathbf{K}\ \mathbf{r}(\mathbf{q}(t_{k-1}), \mathbf{x}(t_k)).
\end{equation}
Thus, we can define a corrected velocity vector $\mathbf{v}^{*}(t_k) = \mathbf{v}(t_k) + \mathbf{K}\ \mathbf{r}(\mathbf{q}(t_{k-1}), \mathbf{x}(t_k))$, which will be used in the inverse differential kinematics to solve for the system velocity.

\subsection{Inverse Differential Kinematics}
The inverse differential kinematics problems solves for the system velocity $\nu(t)$ given a set of task space velocities (or target velocities).
From the Eq. \eqref{eq:chap:human-motion:dynik-input}, it is necessary to solve the inverse differential kinematics using the corrected velocity vector $\mathbf{v}^{*}(t_k)$ in order to obtain the control input $\nu(t_k)$. \looseness=-1

The most common approach to solve Eq. \eqref{eq:chap:human-motion:dynik-input} is to use a regularized, weighted pseudo-inverse of the Jacobian matrix to compute the system velocity as,
$$
\nu_k = (\J_{k-1}^T\ \mathbf{W}\ \J_{k-1} + \lambda \I{})^{-1}\J_{k-1}^T\  \mathbf{W}\ \mathbf{v}_{k}^{*},
$$
where, $\mathbf{W}$ and $\lambda \I{}$ are weighting and regularization matrix respectively to weight the different measurements and dampen the least-squares problem to avoid singularities.

It is, however, preferable to account for model constraints such as the limits on the joint positions and velocities while solving for the system velocity $\nu(t_k)$.
In such cases, a Quadratic Programming (QP) solver is preferred to solve Eq. \eqref{eq:chap:human-motion:dynik-input} since model constraints can be accounted for naturally within the QP problem.
The QP optimization problem is then written as, \looseness=-1
\begin{equation}
\label{chap:human-motion:dynik-qp-prob}
\begin{split} 
    \underset{\nu(t_k)}{\text{minimize}} & \qquad \norm{\mathbf{v}^{*}(t_k) - \mathbf{J}(\mathbf{q}(t_{k-1})) \nu(t_k)} _2  \\
    \text{subject to} & \qquad \mathbf{G}\  \jointVel(t_k) \leq \mathbf{g},
\end{split}
\end{equation}
It can be seen that the joint position constraints of the form $\mathbf{A} \jointPos(t_k) \leq \mathbf{b}^q$ cannot be directly applied to the QP problem.
A strategy to convert the joint position constraints into joint velocity constraints is proposed by \cite{rapetti2020model} which allows to include also joint position limits within the inverse differential velocity kinematics problem.

\subsection{State Integration}
Starting from an initial configuration $\mathbf{q}(t_0)$, the state configuration $\mathbf{q}(t_k)$ can be obtained by integrating the system velocity solution $\nu(t_k)$ over time.
The base position and the joint positions evolving over the Euclidean vector space can be obtained by a straightforward application of numerical integration methods such as Euler integration, Tustin's, or Runge-Kutta (RK4) methods (\cite{davis2007methods}).
The base orientation is obtained by integrating the solved angular velocity $\omegaRightTriv{A}{B}$ over the space of rotations using a Lie-Euler integration with the sampling period $\Delta T = t_{k} - t_{k-1}$,
$$
\Rot{A}{B}(t_k) = \gexphat{\SO{3}}(\omegaRightTriv{A}{B}(t_k) \Delta T)\ \Rot{A}{B}(t_{k-1}).
$$
Alternative choices for the integration of the rotation matrix can use Baumgarte stabilization on $\SO{3}$ (\cite{gros2015baumgarte}) or a higher-order Lie group integration methods such as Runge-Kutta Munthe-Kaas methods (\cite{munthe1998runge}). \looseness=-1

Thus, the continuous feedback of the system configuration forces the residual errors $(\mathbf{r}, \mathbf{u})$ to tend towards zero and the rate of convergence depends on the choice of the gains $\mathbf{K}$.
When the residual error falls below a certain threshold, the floating base estimation module is activated.
Before moving on to describe the EKF for floating base estimation, we will first describe the contact detection strategy that will become useful during the development of the base estimation block. \looseness=-1

\section{Center of Pressure based Contact Detection}
\label{sec:chap:human-motion:cop-contact}

Instead of relying on the contact state of the foot based on the thresholding of 6D contact wrench acting on it, we rely on the contact states of candidate contact points chosen from a simplified geometry of the foot.
This will allow us to slightly relax the rigid contact assumption for the foot and allow correcting for any position displacements due to foot rotations.
For this purpose, we first decompose a 6D contact wrench into contact normal forces acting at the vertices of a rectangular approximation of the foot.
These vertices are chosen as the candidate contact points and if the force at the vertices exceeds a certain threshold over a period of time then we infer the vertex to be in rigid contact with the environment.
A Schmitt Trigger-based thresholding is used to be robust to noisy values. \looseness=-1

\begin{figure}[!h]
\centering
\includegraphics[scale=0.4]{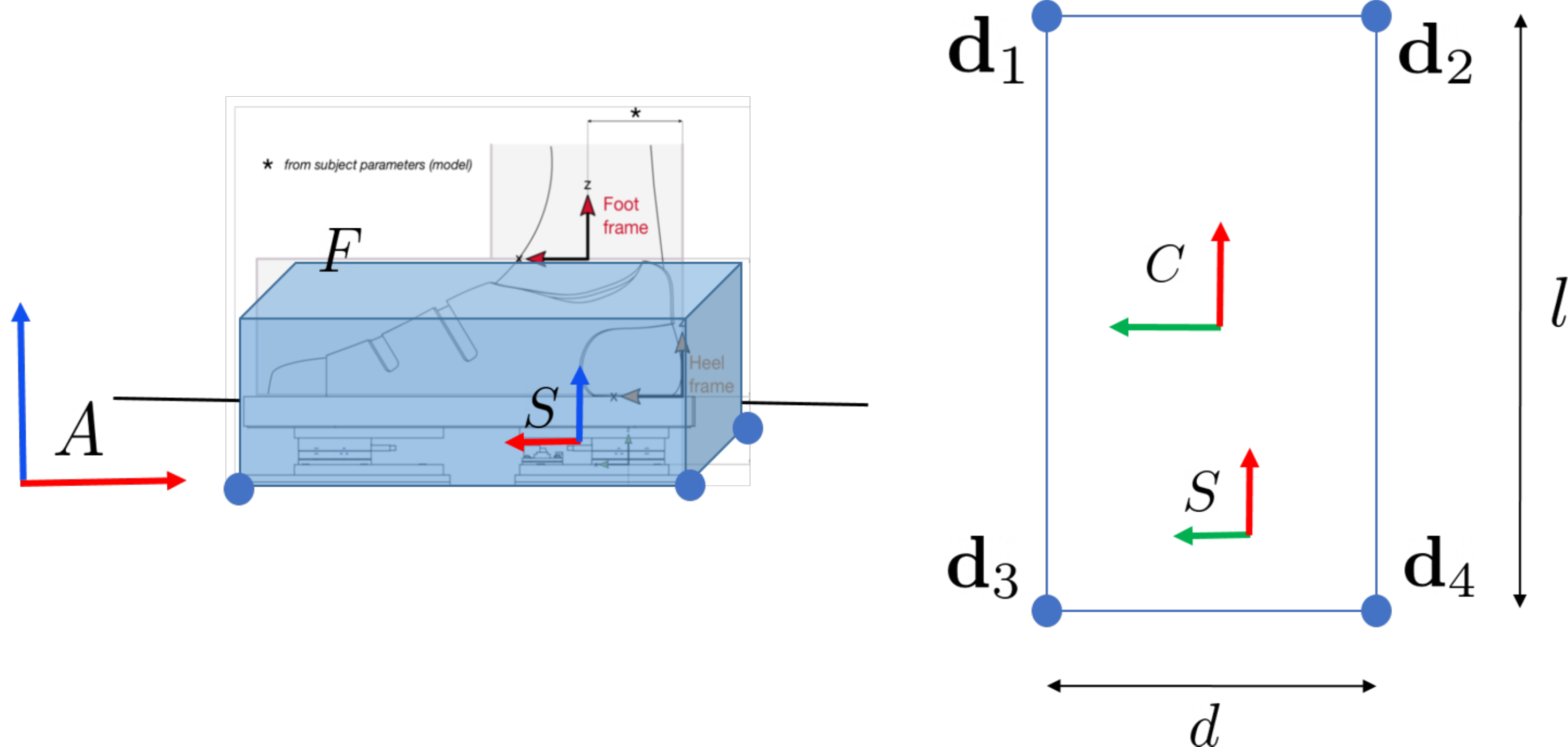}
\caption{Rectangular approximation of the foot geometry}
\label{fig:chap:human-motion:foot-geometery}
\end{figure}

For the contact normal force decomposition, we follow the approach presented by \cite{dafarra2020predictive}.
Consider the sole of foot to have a simplified rectangular geometry with length $l$ and width $d$ as shown in the Figure \ref{fig:chap:human-motion:foot-geometery}.
The sole is assumed to have a planar contact with the ground.
The coordinate frame $C$ associated with the sole is placed at the center of the sole surface, with the $x$-axis pointing forward and parallel to the side with length $l$, while $z$-axis points upwards and perpendicular to the surface of the foot-sole.
The coordinates of the vertices in the frame $C$ are given by,
\begin{equation}
    \mathbf{d}_1 = \begin{bmatrix} \frac{l}{2}\\ \frac{d}{2}\\ 0 \end{bmatrix}, \
    \mathbf{d}_2 = \begin{bmatrix} \frac{l}{2}\\ -\frac{d}{2}\\ 0 \end{bmatrix}, \
    \mathbf{d}_3 = \begin{bmatrix} -\frac{l}{2}\\ \frac{d}{2}\\ 0 \end{bmatrix}, \
    \mathbf{d}_4 = \begin{bmatrix} -\frac{l}{2}\\ -\frac{d}{2}\\ 0 \end{bmatrix}.
\end{equation}

A contact wrench $\wrenchExt{} = \begin{bmatrix}{\forceExt{}}^T & {\torqueExt{}}^T \end{bmatrix}^T \in \R^6$ acting on frame $C$ is related to the 3D forces $\force{}_i \in \R^3,\ i = \{1,2, 3, 4\}$ at the vertices through the following conditions,
\begin{align}
   & \force{}_1 + \force{}_2 + \force{}_3 + \force{}_4 = \forceExt{}, \\
   & \sum \mathbf{d}_i \times \force{}_i = \torqueExt{}.
\end{align}
From the equations above, the contact normal forces can be extracted to correspond to the total contact normal force and the contact tangential torques applied along $x$- and $y$-axes.
\begin{align}
\label{eq:chap:human-motion:cop-normal-force}
   & {f_1}_z + {f_2}_z + {f_3}_z + {f_4}_z = f^\mathbf{x}_z, \\
\label{eq:chap:human-motion:cop-tau-x}
   & \frac{d}{2}({f_1}_z + {f_3}_z) - \frac{d}{2}({f_2}_z + {f_4}_z) = \tau^\mathbf{x}_x, \\
\label{eq:chap:human-motion:cop-tau-y}   
   & \frac{l}{2}({f_3}_z + {f_4}_z) - \frac{l}{2}({f_1}_z + {f_2}_z) = \tau^\mathbf{x}_y.
\end{align}

From Eqs. \eqref{eq:chap:human-motion:cop-tau-x} and \eqref{eq:chap:human-motion:cop-tau-y}, we have,
\begin{align}
\label{eq:chap:human-motion:cop-f1}   
   & {f_1}_z = {f_4}_z +\ \frac{\tau^\mathbf{x}_x}{d} -\ \frac{\tau^\mathbf{x}_y}{l}, \\
\label{eq:chap:human-motion:cop-f3}
   & {f_3}_z = {f_2}_z +\ \frac{\tau^\mathbf{x}_x}{d} +\ \frac{\tau^\mathbf{x}_y}{l}.
\end{align}

On substituting ${f_2}_z = f^\mathbf{x}_z - {f_1}_z - {f_3}_z - {f_4}_z$ and Eq. \eqref{eq:chap:human-motion:cop-f1} in Eq. \eqref{eq:chap:human-motion:cop-f3}, we get,
\begin{align}
   & {f_3}_z = \frac{f^\mathbf{x}_z}{2} -\ {f_4}_z +\ \frac{\tau^x_y}{l}.
\end{align}

${f_3}_z$ is now expressed as a function of ${f_4}_z$.
We can express ${f_1}_z$ and ${f_2}_z$ also as a function of ${f_4}_z$.
In summary, we have an infinite solutions for the contact normal forces at the vertices depending on the choice of ${f_4}_z$, \looseness=-1
\begin{align}
    & {f_1}_z = \frac{\tau^\mathbf{x}_x}{d} +\ {f_4}_z -\ \frac{\tau^\mathbf{x}_y}{l}, \\
    & {f_2}_z = \frac{f^\mathbf{x}_z}{2} -\ {f_4}_z -\ \frac{\tau^\mathbf{x}_x}{d}, \\
    & {f_3}_z = \frac{f^\mathbf{x}_z}{2} -\ {f_4}_z +\ \frac{\tau^\mathbf{x}_y}{l}.
\end{align}

The infinite solution set shrinks by imposing a positivity constraint on the contact normal forces, ${f_i}_z \geq 0$, leading to the following inequalities,
\begin{align}
\label{eq:chap:human-motion:f4-ineq}
\begin{split}
    & {f_4}_z \geq 0, \\
    & {f_4}_z \geq - \frac{\tau^\mathbf{x}_x}{d} + \frac{\tau^\mathbf{x}_y}{l}, \\
    & {f_4}_z \leq \frac{f^\mathbf{x}_z}{2} + \frac{\tau^\mathbf{x}_y}{l}, \\
    & {f_4}_z \leq \frac{f^\mathbf{x}_z}{2} - \frac{\tau^\mathbf{x}_x}{d}.
\end{split}
\end{align}

Let us define $\alpha_i = {{f_i}_z}/{f^\mathbf{x}_z}$ as the ratio of the normal force at the vertex to the total contact normal force.
Further, the Center of Pressure (CoP) in the frame $C$ is given by,
\begin{align}
    & \text{CoP}_x = - \frac{\tau^\mathbf{x}_y}{f^\mathbf{x}_z},\quad \text{CoP}_y = \frac{\tau^\mathbf{x}_x}{f^\mathbf{x}_z}.
\end{align}
Thus, on dividing the inequalities in Eq. \eqref{eq:chap:human-motion:f4-ineq} by $f^\mathbf{x}_z$, they can be written as,
\begin{align}
\label{eq:chap:human-motion:alpha4-ineq}
\begin{split}
    & \alpha_4 \geq 0, \\
    & \alpha_4 \geq - \frac{\text{CoP}_y}{d} - \frac{\text{CoP}_x}{l}, \\
    & \alpha_4 \leq \frac{1}{2} - \frac{\text{CoP}_x}{l}, \\
    & \alpha_4 \leq \frac{1}{2} - \frac{\text{CoP}_y}{d}.
\end{split}
\end{align}
Having $\alpha_i < 0$ leads to a negative normal force on the vertex which would physically represent the suction of the vertex into the ground.
Having $\alpha_i > 1$ would cause a normal force on some other corner to be negative.
These situations can be avoided by introducing bounds on $\alpha_4$, implying $\alpha_4 \in [0, 1]$.
The CoP is constrained to be within the support polygon of the foot meaning the CoP always lies within the rectangular geometry of the sole,
$$
\text{CoP}_x \in [-{l}/{2},\ {l}/{2}], \quad \text{CoP}_y \in [-{d}/{2},\ {d}/{2}].
$$
We may further bound $\alpha_4$ depending on the position of the CoP,
\begin{equation}
    \max\left(0, - \frac{\text{CoP}_y}{d} - \frac{\text{CoP}_x}{l} \right) \leq \alpha_4 \leq \min\left(\frac{1}{2} - \frac{\text{CoP}_y}{d}, \frac{1}{2} - \frac{\text{CoP}_x}{l} \right),
\end{equation}
meaning that the two bounds will coincide at 1 when the CoP lies on the vertex position $\mathbf{d}_4$. \looseness=-1
Additionally, $\alpha_4$ needs to be zero when the CoP lies on the opposite vertex position $\mathbf{d}_1$.
With the choice of $\alpha_4$ to equal the average of the CoP bounds, we have a suitable solution for the ratios of contact normal force decomposition,
\begin{align}
\label{eq:chap:human-motion:alpha4-ineq}
\begin{split}
    & \alpha_4 = \frac{\left(  \max\left(0, - \frac{\text{CoP}_y}{d} - \frac{\text{CoP}_x}{l} \right) + \min\left(\frac{1}{2} - \frac{\text{CoP}_y}{d}, \frac{1}{2} - \frac{\text{CoP}_x}{l} \right) \right)}{2}, \\
    & \alpha_1 = \alpha_4 + \frac{\text{CoP}_x}{l} + \frac{\text{CoP}_y}{d}, \\
    & \alpha_2 = -\alpha_4 + \frac{1}{2} - \frac{\text{CoP}_y}{d}, \\
    & \alpha_3 = -\alpha_4 + \frac{1}{2} - \frac{\text{CoP}_x}{l}.
\end{split}
\end{align}
With this choice of ratios, $\alpha_i$ all equal to $1/4$ when the CoP lies in the center of the foot and the ratios are adjusted appropriately as the CoP moves around the rectangular surface of the approximated foot. 
Such a decomposition and a chosen threshold value for vertex contact detection allow the inference of the contact status of candidate points and the corner in strongest contact with the environment.


\section{Extended Kalman Filter for Base Estimation}
\label{sec:chap:human-motion:ekf-base}

We employ an EKF over matrix Lie groups with right-invariant, left-invariant, and non-invariant observations for estimating the base pose and velocity.
This block acts as a replacement for the sub-module of the state integrator block that was meant to integrate the base link velocity to obtain the base pose.
The EKF block also incorporates contact information and system configuration to obtain an estimate of the floating base state.

\subsection{State Representation}
The EKF block maintains its own internal state and the results of the Dynamical IK block are considered as measurements passed to this block.
We wish to estimate the position $\Pos{A}{B}$, orientation $\Rot{A}{B}$, linear velocity  $\oDot{A}{B}$, and angular velocity $\omegaLeftTriv{A}{B}$ of the base link $B$ in the inertial frame $A$. 
Additionally we also consider the position of candidate contact points on the feet $\Pos{A}{{F_{1,\dots,n_f}}}$ and orientation of the feet $\Rot{A}{F}$, where $F = \{{LF}, {RF} \}$ in the state. 
As seen in the previous section, we approximate the foot links to have a simplified bounding box collision geometry for which the candidate contact points are chosen as the vertices of the bottom face of the bounding box.
Hence the information about each foot incorporated into the internal state of the EKF consists of four vertex positions of the foot ($n_f = 4$) along with a rotation of the foot.

For the sake of readability, we introduce some shorthand notation for the rest of this chapter.
The tuple of state variables $(\Pos{A}{B},\Rot{A}{B}, \oDot{A}{B}, \Pos{A}{{F_{1,\dots,n_f}}})$ or $(\mathbf{p}, \mathbf{R}, \mathbf{v}, \mathbf{d}_{F_{1 \dots n_f}})$ are encapsulated within $\SEk{N+2}{3}$ matrix Lie group (similar to \cite{hartley2020contact}). 
The left-trivialized angular velocity of the base link $\omegaLeftTriv{A}{B}$ denoted as $\omega$ forms a translation group $\T{3}$, while feet rotations $\Rot{A}{F}$ or $\mathbf{Z}_F$ evolve over the group of rotations $\SO{3}$. 
Together, a state space element $\X \in \mathscr{M}$ is represented by a composite matrix Lie group, $\SEk{N+2}{3} \times \T{3} \times \SO{3}^2$
We simplify the derivation in subsequent sections by considering only one foot with only one contact point per foot and dropping the suffix $F_{1, \dots, {n_f}}$ and $F$ for the feet positions $\mathbf{d}_{F_{1 \dots n_f}}$ and orientation $\mathbf{Z}_F$ respectively.

A state space element $\X \in \mathscr{M}$ keeping only one foot contact point and one foot orientation is then given by, \looseness=-1
\begin{align}
\label{eq:chap:human-motion:ekf-state-repr}
\X_1 &=
\begin{bmatrix}
\mathbf{R} & \mathbf{p} & \mathbf{v} & \mathbf{d}   \\
\Zeros{1}{3} & 1 & 0 & 0  \\
\Zeros{1}{3} & 0 & 1 & 0   \\
\Zeros{1}{3} & 0 & 0 & 1   
\end{bmatrix}, & \X_2 &=
\begin{bmatrix}
\I{3} & \omega    \\
\Zeros{1}{3} & 1 
\end{bmatrix}, & \X = \text{blkdiag}(\X_1, \X_2, \Z).
\end{align}
In tuple representation, we can denote the state space element as, $\X \triangleq (\mathbf{p}, \mathbf{R}, \mathbf{v}, \mathbf{d},\omega, \mathbf{Z})_\mathscr{M}$.
The inverse of the element $\X$ can be written as,
$$
\X^{-1} = \text{blkdiag}(\X_1^{-1}, \X_2^{-1}, \Z^T),
$$
where, 
$$
\X_1^{-1} = \begin{bmatrix}
	\Rot{}{}^T &  -\Rot{}{}^T\ \mathbf{p} &  -\Rot{}{}^T\ \mathbf{v} &  -\Rot{}{}^T\ \mathbf{d} \\
	\Zeros{1}{3} & 1 & 0 & 0  \\
	\Zeros{1}{3} & 0 & 1 & 0   \\
	\Zeros{1}{3} & 0 & 0 & 1
\end{bmatrix},  \quad
\X_2^{-1} = \begin{bmatrix}
	\I{3} & -\omega    \\
	\Zeros{1}{3} & 1 
\end{bmatrix}
$$
The hat operator $\ghat{\mathscr{M}}{\err}:  \R^{9+3{n_f}+3+3} \to \mathfrak{m}$ with $n_f = 1$ and $N = 1$ becomes,
\begin{align}
\label{eq:chap:human-motion:ekf-state-hat}
& \ghat{\SEk{N+2}{3}}{\err_1} =
\begin{bmatrix}
S(\err_\mathbf{R}) & \err_\mathbf{p} & \err_\mathbf{v} & \err_\mathbf{d}   \\
\Zeros{1}{3} & 0 & 0 & 0  \\
\Zeros{1}{3} & 0 & 0 & 0   \\
\Zeros{1}{3} & 0 & 0 & 0   
\end{bmatrix}, \quad \ghat{\T{3}}{\err_2} =
\begin{bmatrix}
\Zero{3} & \err_\omega    \\
\Zeros{1}{3} & 0 
\end{bmatrix}, \\ 
& \ghat{\mathscr{M}}{\err} = \text{blkdiag}(\ghat{\SEk{N+2}{3}}{\err_1},\  \ghat{\T{3}}{\err_2},\  S(\err_\mathbf{Z})),
\end{align}
with, $\err_1 = \text{vec}(\err_\mathbf{p}, \err_\mathbf{R}, \err_\mathbf{v}, \err_\mathbf{d})$ and the vector $\err \triangleq \text{vec}(\err_1, \err_\omega, \err_\mathbf{Z})$.
The exponential mapping of the state space expressed in a tuple representation is given by, \looseness=-1
\begin{equation}
\label{eq:chap:human-motion:ekf-state-exp-map}
\begin{split}
\gexphat{\mathscr{M}}\left(\err\right) &= \big(\J\left(\err_\mathbf{R}\right)\ \err_\mathbf{p},\ \Exp\left(\err_\mathbf{R}\right),\ \J\left(\err_\mathbf{R}\right)\ \err_\mathbf{v}, \\
& \quad \ \ \ \J\left(\err_{\mathbf{R}}\right)\ \err_{\mathbf{d}},\ \err_\omega,\ 
\Exp\left(\err_{\mathbf{Z}}\right)\big)_{\mathscr{M}}.
\end{split}
\end{equation}
where, $\Exp(.) \triangleq\ \gexphat{\SO{3}}(.)$ is the exponential map of $\SO{3}$ and $\J(.) \triangleq \gljac{\SO{3}}(.)$ is the left Jacobian of $\SO{3}$.
The adjoint matrix of the considered state space $\mathscr{M}$ is given as,
\begin{align}
\label{eq:chap:human-motion:ekf-state-adj}
\begin{split}
\gadj{\X_1} =
\begin{bmatrix}
\Rot{}{} & S(\mathbf{p}) \Rot{}{} & \Zero{3}  & \Zero{3} \\
\Zero{3} & \Rot{}{} & \Zero{3}  & \Zero{3} \\
\Zero{3} &  S(\mathbf{v}) \Rot{}{} & \Rot{}{} & \Zero{3} \\
\Zero{3} &  S(\mathbf{d}) \Rot{}{} & \Zero{3} & \Rot{}{}
\end{bmatrix}, \quad \gadj{\X} = \text{blkdiag}(\gadj{\X_1},\ \I{3},\ \Z).
\end{split}
\end{align}

The left Jacobian of the matrix Lie group $\mathscr{M}$ is obtained as a block-diagonal form of the left Jacobians of the constituting matrix Lie groups given as,
\begin{align}
\label{eq:chap:human-motion:ekf-state-ljac}
\begin{split}
\gljac{\mathscr{M}}(\err) = \text{blkdiag} \left( \gljac{\SEk{N+2}{3}}\left(\err_1\right), \; \I{3},\ \J\left(\err_{\mathbf{Z}}\right) \right).
\end{split}
\end{align}
The left Jacobian for the translation group of angular velocity is the identity matrix, since the underlying representation of the translation group is the Euclidean vector space.

\subsection{System Dynamics}
The continuous system dynamics evolving on the group with state $\X \in \mathscr{M}$ is given by the dynamical system, $\frac{d}{dt} \X = f(\X, \mathbf{u}) + \X \ghat{\mathscr{M}}{\textbf{w}}$, where $f$ is the deterministic dynamics, $\mathbf{u} \in \R^m$ is the exogenous control input and  $\textbf{w} \in \mathbb{R}^p$  is a continuous white noise, with covariance $\Q$.
The system dynamics is formulated using rigid body kinematics with constant motion models for the base link along with rigid contact models for the contact points and the foot orientations.
The continuous system dynamics is then given by,
\begin{equation}
	\label{eq:chap:human-motion:ekf-sys-dyn}
	\begin{split}
	\mathbf{\dot{p}} = \mathbf{v}, & \qquad  \mathbf{\dot{R}} = \mathbf{R} \; S(\omega), \\
	\mathbf{\dot{v}} = \mathbf{R} \; (-\noiseLinVel{B}), & \qquad \mathbf{\dot{d}} =  \mathbf{R}\; (-{\Rot{B}{F}}(\encoders)\ \noiseLinVel{F}), \\
	\dot{\omega} =  -\noiseAngVel{B}, & \qquad \mathbf{\dot{Z}} =  \mathbf{Z} \; S(-\noiseAngVel{F}),
	\end{split}
\end{equation}
with the left trivialized noise vector defined as,
\begin{equation}
 \begin{aligned}
  \label{eq:chap:human-motion:ekf-sys-left-triv-noise}
  & \mathbf{w}  \footnotesize{ \;= \text{vec}\big( \Zeros{3}{1}, \; \Zeros{3}{1}, \; -\noiseLinVel{B}, \;  -{\Rot{B}{F}}(\encoders)\;\noiseLinVel{F}, \; -\noiseAngVel{B},\; -\noiseAngVel{F}\big) },
\end{aligned}
\end{equation}
and the prediction noise covariance matrix $\mathbf{Q}_c = \expectation{\mathbf{w} \mathbf{w}^T}$. Here, we have used ${\Rot{B}{F}}(\encoders)$ which denotes the rotation of the foot frame with respect to the base link computed through the forward kinematics map using the joint position inputs.
The system dynamics defined in Eq. \eqref{eq:chap:human-motion:ekf-sys-dyn} does not obey the group-affine property. \looseness=-1

\subsubsection{Linearized Error Dynamics}
The right invariant error of type 2, $\eta^R = \Xhat \X^{-1}$ (see Def. \ref{def:right-invariant-error-type2}) can be expressed  in the tuple form as,
$$
\left(\mathbf{\hat{p}} - \mathbf{\hat{R}}\mathbf{R}^T\;\mathbf{p},\ \mathbf{\hat{R}}\;\mathbf{R}^T,\  
\mathbf{\hat{v}} - \mathbf{\hat{R}}\mathbf{R}^T\;\mathbf{v}, \
\mathbf{\hat{d}} - \mathbf{\hat{R}}\mathbf{R}^T\;\mathbf{d}, \
\hat{\omega} - \omega, \
\mathbf{\hat{Z}}\mathbf{Z}^T
\right)_\mathscr{M}.
$$

The error dynamics can be obtained by simply taking the time derivatives of the right invariant error and expressing them as a function of $\err$ by approximating $\eta^R = \I{} + \ghat{\mathscr{M}}{\err}$.
Consider the evolution equations of the base rotations, $\mathbf{\dot{R}} = \mathbf{R} \; S(\omega)$, the base rotation error dynamics can be obtained as,
\begin{align*}
	\frac{d}{dt}(\mathbf{\hat{R}}\;\mathbf{R}^T) &=  \frac{d}{dt}(\mathbf{\hat{R}})\;\mathbf{R}^T + \mathbf{\hat{R}}\; \frac{d}{dt}(\mathbf{R}^T) \\
	&= \mathbf{\hat{R}}\; S(\hat{\omega})\mathbf{R}^T + \mathbf{\hat{R}}\;\left(\mathbf{R} \; S(\omega)\right)^T \\
	&= \mathbf{\hat{R}}\; S(\hat{\omega})\mathbf{R}^T - \mathbf{\hat{R}}\; S(\omega)\mathbf{R}^T \\
	&= \mathbf{\hat{R}}\; S(\hat{\omega} - \omega)\mathbf{R}^T \\
	&= \mathbf{\hat{R}}\; S(\err_\omega)\mathbf{\hat{R}}^T (\I{3} + S(\err_\mathbf{R})) \\
	&\approx S(\mathbf{\hat{R}}\; \err_\omega)
\end{align*}
where, we have used $\X^{-1} = \Xhat^{-1}\; \eta^R $ and substituted a first-order approximation for the exponential map of the error $\err$, while neglecting the cross-product of the error terms.
The error dynamics for the base position, base velocity and the foot position can be obtained as,
\begin{align*}
	\frac{d}{dt}(\mathbf{\hat{p}} - \mathbf{\hat{R}}\mathbf{R}^T\;\mathbf{p}) &=  \mathbf{\dot{p}} - \frac{d}{dt}(\mathbf{\hat{R}}\;\mathbf{R}^T) \mathbf{\hat{p}} - \mathbf{\hat{R}}\;\mathbf{R}^T  \mathbf{\dot{\hat{p}}} \\
	&= \mathbf{v} - S(\hat{\mathbf{R}} \err_\omega) \hat{\mathbf{p}} - (\I{3} + S(\err_\mathbf{R})) \hat{\mathbf{v}} \\
	&= (\hat{\mathbf{v}} + S(\err_\mathbf{R}) \hat{\mathbf{v}} + \err_\mathbf{v} ) - S(\hat{\mathbf{R}} \err_\omega) \hat{\mathbf{p}} - \hat{\mathbf{v}} - S(\err_\mathbf{R}) \hat{\mathbf{v}} \\
	&= S(\mathbf{\hat{p}})\mathbf{\hat{R}}\; \err_\omega + \err_\mathbf{v}.
\end{align*}
\begin{align*}
	\frac{d}{dt}(\mathbf{\hat{v}} - \mathbf{\hat{R}}\mathbf{R}^T\;\mathbf{v}) &=  \mathbf{\dot{v}} - \frac{d}{dt}(\mathbf{\hat{R}}\;\mathbf{R}^T) \mathbf{\hat{v}} - \mathbf{\hat{R}}\;\mathbf{R}^T  \mathbf{\dot{\hat{v}}} \\
	&= \mathbf{R} \; (-\noiseLinVel{B}) - S(\hat{\mathbf{R}} \err_\omega) \hat{\mathbf{v}}  \\
	&\approx S(\mathbf{\hat{v}})\mathbf{\hat{R}}\; \err_\omega - \hat{\mathbf{R}} \; \noiseLinVel{B}.
\end{align*}
\begin{align*}
	\frac{d}{dt}(\mathbf{\hat{d}} - \mathbf{\hat{R}}\mathbf{R}^T\;\mathbf{d}) &=  \mathbf{\dot{d}} - \frac{d}{dt}(\mathbf{\hat{R}}\;\mathbf{R}^T) \mathbf{\hat{d}} - \mathbf{\hat{R}}\;\mathbf{R}^T  \mathbf{\dot{\hat{d}}} \\
	&= \mathbf{R}\; (-{\Rot{B}{F}}(\encoders)\ \noiseLinVel{F}) - S(\hat{\mathbf{R}} \err_\omega) \hat{\mathbf{d}}  \\
	&\approx S(\mathbf{\hat{d}})\mathbf{\hat{R}}\; \err_\omega - \hat{\mathbf{R}}\; {\Rot{B}{F}}(\encoders)\ \noiseLinVel{F}.
\end{align*}
We can compute the error dynamics for the base angular velocity and the foot orientation in a straightforward manner following a similar procedure of considering a first-order approximation of the exponential map and neglecting the cross-product terms of error and noise terms. \looseness=-1

Thus, the overall error dynamics can then be summarized as,
\begin{equation}
	\label{eq:chap:human-motion:ekf-sys-errordynamics}
	\begin{split}
	&\frac{d}{dt}(\mathbf{\hat{p}} - \mathbf{\hat{R}}\mathbf{R}^T\;\mathbf{p}) = S(\mathbf{\hat{p}}) \mathbf{\hat{R}}\; \err_\omega + \err_\mathbf{v}, \\
	&\frac{d}{dt}(\mathbf{\hat{R}}\;\mathbf{R}^T) = S(\mathbf{\hat{R}}\; \err_\omega), \\
	&\frac{d}{dt}(\mathbf{\hat{v}} - \mathbf{\hat{R}}\mathbf{R}^T\;\mathbf{v}) = S(\mathbf{\hat{v}}) \mathbf{\hat{R}}\; \err_\omega - \mathbf{\hat{R}}\;\noiseLinVel{B}, \\
	&\frac{d}{dt}(\mathbf{\hat{d}} - \mathbf{\hat{R}}\mathbf{R}^T\;\mathbf{d}) = S(\mathbf{\hat{d}}) \mathbf{\hat{R}}\; \err_\omega - \mathbf{\hat{R}}\;\Rot{B}{F}(\encoders)\;\noiseLinVel{F}, \\
	&\frac{d}{dt}(\hat{\omega} - \omega) = -\noiseAngVel{B},\\
	&\frac{d}{dt}(\mathbf{\hat{Z}}\mathbf{Z}^T) = - S(\mathbf{\hat{Z}}\; \noiseAngVel{F}).
	\end{split}
\end{equation}

Using the log-linearity property, the linearized error dynamics and covariance propagation equation then become, \looseness=-1
\begin{equation}
	\label{eq:chap:human-motion:ekf-sys-lin-err-prop}
	\begin{split}
	&\dot{\err} = \mathbf{F}_c\ \err + \gadj{\Xhat}\ \mathbf{w}, \\
	& \dot{\cov}  = \mathbf{F}_c\; \cov + \cov\; \mathbf{F}_c^T + \mathbf{\hat{Q}}_c ,
	\end{split}
\end{equation}
where the continuous-time, linearized error propagation matrix $\mathbf{F}_c$ and the prediction noise covariance matrix $\mathbf{\hat{Q}}_c$ are given as, \looseness=-1
\begin{equation}
	\label{eq:chap:human-motion:ekf-err-prop-matrices}
	\begin{split}
	&\mathbf{F}_c\ = \begin{bmatrix}
	\Zero{3} & \Zero{3} & \I{3} & \Zero{3} & S(\mathbf{\hat{p}}) \mathbf{\hat{R}} & \Zero{3} \\
	\Zero{3} & \Zero{3} & \Zero{3} & \Zero{3} & \mathbf{\hat{R}} & \Zero{3} \\
	\Zero{3} & \Zero{3} & \Zero{3} & \Zero{3} & S(\mathbf{\hat{v}}) \mathbf{\hat{R}} & \Zero{3} \\
	\Zero{3} & \Zero{3} & \Zero{3} & \Zero{3} & S(\mathbf{\hat{d}}) \mathbf{\hat{R}} & \Zero{3} \\
	\Zero{3} & \Zero{3} & \Zero{3} & \Zero{3} & \Zero{3} & \Zero{3} \\
	\Zero{3} & \Zero{3} & \Zero{3} & \Zero{3} & \Zero{3} & \Zero{3} \\
	\end{bmatrix}, \\
	& \mathbf{\hat{Q}}_c  = \gadj{\Xhat}\;\mathbf{Q}_c\;\gadj{\Xhat}^T.
	\end{split}
\end{equation}
A discretization of the continuous-time quantities is performed assuming a zero-order hold with sampling time $\Delta T$ gives the discrete system dynamics as, \looseness=-1
\begin{equation}
	\label{eq:chap:human-motion:ekf-disc-dyn}
	\begin{split}
	\mathbf{\hat{p}}_\knext = \; \mathbf{\hat{p}}_\kcurr + \;  \mathbf{\hat{v}}_{\kcurr} \; \Delta T, & \quad  \quad \mathbf{\hat{R}}_{\knext} = \; \mathbf{\hat{R}}_{\kcurr} \; \gexphat{\SO{3}}\big( \hat{\omega}_k \; \Delta T \big), \\
	\mathbf{\hat{v}}_{\knext} = \; \mathbf{\hat{v}}_{\kcurr}, &  \quad \quad \quad \quad \mathbf{\hat{d}}_{\knext} = \; \mathbf{\hat{d}}_{\kcurr},\\
	\hat{\omega}_{\knext} = \; \hat{\omega}_{\kcurr}, & \quad \quad \quad \quad \mathbf{\hat{Z}}_{\knext} = \; \mathbf{\hat{Z}}_{\kcurr}.
	\end{split}
\end{equation}
A first-order approximation for the Ricatti equations leads to,
\begin{equation}
	\label{eq:chap:human-motion:ekf-disc-cov-prop}
	\begin{split}
	&\cov_{\knext} = \mathbf{F}_{\kcurr} \; \cov_{\kcurr}\; \mathbf{F}_{\kcurr}^T + \mathbf{Q}_{\kcurr}, \\
	&\mathbf{F}_{\kcurr} = \exp(\mathbf{F}_c \Delta T) \approx  \I{p} + \mathbf{F}_c \Delta T,\\
	&\Q_{\kcurr} = \mathbf{F}_{\kcurr}\; \mathbf{\hat{Q}}_c\; \mathbf{F}_{\kcurr}^T\; \Delta T.
	\end{split}
\end{equation}

\subsection{Right Invariant Observations}
We consider measurements with right-invariant observation structure,  $\mathbf{z}_{\kcurr} = \X_{\kcurr}^{-1}\ \mathbf{b}\ +\ \mathbf{n}_{\kcurr} \in \R^q$ which lead to a time-invariant measurement model Jacobian (see Section \ref{sec:chapter02:invekf-matrix-lie-groups}) and follows the filter update, \looseness=-1 
\begin{equation}
	\label{eq:chap:human-motion:ekf-rie-update}
\Xhat_{\kcurr}^{+} = \gexphat{\mathscr{M}} \left(\mathbf{K}_\kcurr \left(\Xhat_{\kcurr}\ \mathbf{z}_{\kcurr} - \mathbf{b}\right) \right)\ \Xhat_{\kcurr},
\end{equation}
where, $\mathbf{K}$ is the Kalman gain and $\mathbf{n}_{\kcurr}$ is the noise associated with the observation and $\mathbf{b}$ is a constant vector.
The measurement model Jacobian is obtained through a first-order approximation of the non-linear error update, \looseness=-1
$$\errG^{R+}_{\kcurr} = \gexphat{\mathscr{M}} \left(\mathbf{K}_\kcurr \left( \errG^{R}_{\kcurr}\ \mathbf{b}\ -\ \mathbf{b} + \Xhat_{\kcurr}\mathbf{n}_{\kcurr} \ \right) \right) \errG^{R}_{\kcurr}.$$ \looseness=-1

We consider measurement updates from relative contact point positions, null foot velocity-aided left trivialized base velocity computations as the right invariant observations which are functions of joint positions' and velocities' measurements.

\subsubsection{Relative  candidate  contact point position}
\label{subsec:chap:human-motion:ekf-base-rel-pos-meas}
The joint positions $\encoders = \jointPos + \encoderNoise$ obtained from the integration of estimated joint velocities are assumed to be affected by white Gaussian noise $\encoderNoise$ and are used to determine the relative  candidate  contact point positions with respect to the base link using forward kinematics.
The forward kinematics measurement $h_p(\encoders)$ expressed as a function of states is given as, 
$$ 
h_p(\X) = \mathbf{\hat{R}}^T(\mathbf{\hat{d}} - \mathbf{\hat{p}}) + \relativeJacobianLeftTrivLinIn{F}{B}(\encoders) \; \encoderNoise \in \mathbb{R}^3.
$$
In matrix form, the measurement has a right invariant observation structure $\mathbf{z}_p = \X^{-1} \mathbf{b}_p + \mathbf{n}_p$,
$$
\begin{bmatrix}
	h_p(\encoders) \\ 1 \\ 0 \\ -1 \\ \Zeros{3}{1} \\ 0 \\ \Zeros{3}{1} 
\end{bmatrix}
=\ \X^{-1} 
\begin{bmatrix}
	\Zeros{3}{1} \\ 1 \\ 0 \\ -1 \\ \Zeros{3}{1} \\ 0 \\ \Zeros{3}{1}
\end{bmatrix}\ +\ 
\begin{bmatrix}
	\relativeJacobianLeftTrivLinIn{F}{B}(\encoders) \; \encoderNoise \\ 0 \\ 0 \\ 0 \\ \Zeros{3}{1} \\ 0 \\ \Zeros{3}{1}
\end{bmatrix}.
$$
With such an observation structure, the innovation term $\Xhat_{\kcurr}\ \mathbf{z}_{\kcurr} - \mathbf{b}$ in Eq. \eqref{eq:chap:human-motion:ekf-rie-update} depends only on the invariant error. 
A reduced dimensional gain $\mathbf{K}_\kcurr^r$ can be computed by applying an auxiliary selection matrix $\Pi_p = \begin{bmatrix} \I{3} & \Zeros{3}{10}
\end{bmatrix}$ that selects only the non-zero elements from the full innovation vector in such a way that,
$$\mathbf{K}_\kcurr \left(\Xhat_{\kcurr}\ \mathbf{z}_{\kcurr} - \mathbf{b}\right) = \mathbf{K}_\kcurr^r \Pi_p \Xhat_{\kcurr}\ \mathbf{z}_{\kcurr}.$$
The measurement model Jacobian and the measurement noise covariance can be computed through the linearization of the error-update equations as already seen in Eq. \eqref{eq:chap02:invekf-linearized-update-rie},
\begin{align*}
\begin{split}
\err^{R+}_{\kcurr} &= \ \err^{R}_{\kcurr} + \mathbf{K}_\kcurr^r\ \Pi_p \left(\ghat{G}{\err^{R}_{\kcurr}}\ \mathbf{b} + \Xhat_{\kcurr}\mathbf{n}_{\kcurr}\right) \\
&= \ \err^{R}_{\kcurr} + \mathbf{K}_\kcurr^r\ \Pi_p \left(\begin{bmatrix}
	\err_\mathbf{p} - \err_\mathbf{d} \\ 1 \\ 0 \\ -1 \\ \Zeros{3}{1} \\ 0 \\ \Zeros{3}{1}
\end{bmatrix} + 
\begin{bmatrix}
	\hat{\Rot{}{}}\ \relativeJacobianLeftTrivLinIn{F}{B}(\encoders) \; \encoderNoise \\ 0 \\ 0 \\ 0 \\ \Zeros{3}{1} \\ 0 \\ \Zeros{3}{1}
\end{bmatrix} 
\right) \\
&= \ \err^{R}_{\kcurr} - \mathbf{K}_\kcurr^r\  \left(\begin{bmatrix}
	-\I{3} & \Zero{3} & \Zero{3} & \I3 & \Zero{3} & \Zero{3}
\end{bmatrix} \err^{R}_{\kcurr} - 
	\hat{\Rot{}{}}\ \relativeJacobianLeftTrivLinIn{F}{B}(\encoders)\ \encoderNoise
\right).
\end{split}
\end{align*}

The right-invariant observation $\mathbf{z}_p$, constant vector $\mathbf{b}_p$, the measurement model Jacobian $\mathbf{H}_p$ and the measurement noise covariance matrix $\mathbf{N}_p$ associated with the relative position measurements can then be written as, \looseness=-1
\begin{equation}
\label{eq:chap:human-motion:ekf-relposmeas}
	\begin{split}
	& \mathbf{z}_p^T = \begin{bmatrix}
	h_p^T(\encoders) & 1 & 0 & -1 & \Zeros{1}{3} & 0 & \Zeros{1}{3}
	\end{bmatrix},\\
	& \mathbf{b}_p^T = \begin{bmatrix}
	\Zeros{1}{3} & 1 & 0 & -1 & \Zeros{1}{3} & 0 & \Zeros{1}{3}
	\end{bmatrix},\\
	& \Pi_p = \begin{bmatrix} \I{3} & \Zeros{3}{10}
	\end{bmatrix}, \\
	& \mathbf{H}_p = \begin{bmatrix}
	-\I{3} & \Zero{3} & \Zero{3} & \I3 & \Zero{3} & \Zero{3}
	\end{bmatrix}, \\
	& \mathbf{N}_p = \mathbf{\hat{R}} \; \relativeJacobianLeftTrivLinIn{F}{B}(\encoders) \; \text{Cov}(\encoderNoise) \; \relativeJacobianLeftTrivLinIn{F}{B}^T(\encoders) \; \mathbf{\hat{R}}^T.
	\end{split}
\end{equation}

\subsubsection{Zero-velocity Update (ZUPT) aided Left Trivialized Base Velocity}
As already seen in Section \ref{sec:chap:loosely-coupled:swa-legged-odom}, with a rigid contact assumption, the null stance foot velocity can be associated with the base velocity using the configuration dependent Jacobian matrix.
We use a left-trivialized base velocity measurement computed using the joint velocity measurements through the free-floating Jacobian of the candidate contact point as a right-invariant observation for the filter.
This observation $h_v(\encoders, \encoderSpeeds) = -  \big(\Xtwist{F}{B}^{-1} \; \relativeJacobianLeftTrivIn{B}{F}(\encoders) \; \encoderSpeeds\big)_{\text{lin}} \in \mathbb{R}^3$ directly relates to the linear velocity state $\mathbf{v}$.
The linear part of the base velocity measurement in the form of $\X_{\kcurr}^{-1}\ \mathbf{b}\ +\ \mathbf{n}_{\kcurr}$ is then described by,
$$
\begin{bmatrix}
	h_v(\encoders, \encoderSpeeds) \\ 0 \\ -1 \\ 0 \\ \Zeros{3}{1} \\ 0 \\ \Zeros{3}{1}
\end{bmatrix}
=\ \X^{-1} 
\begin{bmatrix}
	\Zeros{3}{1} \\ 0 \\ -1 \\ 0 \\ \Zeros{3}{1} \\ 0 \\ \Zeros{3}{1}
\end{bmatrix}\ +\ 
\begin{bmatrix}
	\fkNoiseLinVel{B} \\ 0 \\ 0 \\ 0 \\ \Zeros{3}{1} \\ 0 \\ \Zeros{3}{1}
\end{bmatrix}.
$$
Following the procedure similar to Section \ref{subsec:chap:human-motion:ekf-base-rel-pos-meas}, we can obtain the relevant quantities required for the filter computations as follows,
\begin{equation}
	\label{eq:chap:human-motion:ekf-ZUPTlin}
	\begin{split}
	& \mathbf{z}_v^T = \begin{bmatrix}
	h_v(\encoders, \encoderSpeeds) & 0 & -1 & 0 & \Zeros{1}{3} & 0 & \Zeros{1}{3}
	\end{bmatrix},\\
	& \mathbf{b}_v^T = \begin{bmatrix}
	\Zeros{1}{3} & 0 & -1 & 0 & \Zeros{1}{3} & 0 & \Zeros{1}{3}
	\end{bmatrix},\\
	& \Pi_v = \begin{bmatrix} \I{3} & \Zeros{3}{10}
	\end{bmatrix}, \\
	& \mathbf{H}_v = \begin{bmatrix}
	\Zero{3} & \Zero{3} & \I{3} & \Zero{3} & \Zero{3} & \Zero{3} 
	\end{bmatrix}, \\
	& \mathbf{N}_v = \mathbf{\hat{R}} \; \text{Cov}(\fkNoiseLinVel{B})  \mathbf{\hat{R}}^T.
	\end{split}
\end{equation}
Similarly, the angular part of the left-trivialized base velocity measurement 
$$
h_\omega(\encoders, \encoderSpeeds) = -  \big(\Xtwist{F}{B}^{-1} \; \relativeJacobianLeftTrivIn{B}{F}(\encoders) \; \encoderSpeeds\big)_{\text{ang}} \in \mathbb{R}^3,
$$
 relating to the state variable $\omega$ can be expressed in the matrix form as,
$$
\begin{bmatrix}
	\Zeros{3}{1} \\ 0 \\ 0 \\ 0 \\ h_\omega(\encoders, \encoderSpeeds) \\ -1 \\ \Zeros{3}{1} 
\end{bmatrix}
=\ \X^{-1} 
\begin{bmatrix}
	\Zeros{3}{1} \\ 0 \\ 0 \\ 0 \\  \Zeros{3}{1} \\ -1 \\ \Zeros{3}{1}
\end{bmatrix}\ +\ 
\begin{bmatrix}
	\Zeros{3}{1} \\ 0 \\ 0 \\ 0 \\ \fkNoiseAngVel{B} \\ 0 \\ \Zeros{3}{1}
\end{bmatrix},
$$
and the following quantities relevant for filter update are obtained as,
\begin{equation}
\label{eq:chap:human-motion:ekf-ZUPTang}
	\begin{split}
	& \mathbf{z}_\omega^T = \begin{bmatrix}
	\Zeros{1}{3} & 0 & 0 & 0 & h_\omega(\encoders, \encoderSpeeds) & -1 & \Zeros{1}{3}
	\end{bmatrix},\\
	& \mathbf{b}_\omega^T = \begin{bmatrix}
	\Zeros{1}{3} & 0 & 0 & 0 & \Zeros{1}{3} & -1 & \Zeros{1}{3}
	\end{bmatrix},\\
	& \Pi_\omega = \begin{bmatrix} \Zeros{3}{6} & \I{3} & \Zeros{3}{4}
	\end{bmatrix}, \\
	& \mathbf{H}_\omega = \begin{bmatrix}
	\Zero{3} & \Zero{3} & \Zero{3} & \Zero{3} & \I{3} & \Zero{3} 
	\end{bmatrix}, \\
	& \mathbf{N}_\omega = \text{Cov}(\fkNoiseAngVel{B}).
	\end{split}
\end{equation}

\subsection{Left Invariant Observations}
Similar to right invariant observations, we also consider measurements having a left-invariant observation structure,  $\mathbf{z}_{\kcurr} = \X_{\kcurr}\ \mathbf{b}\ +\ \mathbf{n}_{\kcurr}$ which also lead to a time-invariant measurement model Jacobian obtained through a first-order approximation of the error update equation,
$$
\errG^{L+}_{\kcurr} = \errG^{L}_{\kcurr}\ \gexphat{\mathscr{M}} \left(\mathbf{K}_\kcurr \left( \left(\errG^{L}_{\kcurr}\right)^{-1}\mathbf{b}\ -\ \mathbf{b} + \Xhat^{-1}_{\kcurr}\mathbf{n}_{\kcurr} \ \right) \right).
$$ 
The state update follows
$\Xhat_{\kcurr}^{+} = \Xhat_{\kcurr}\ \gexphat{\mathscr{M}} \left(\mathbf{K}_\kcurr \left(\Xhat^{-1}_{\kcurr}\ \mathbf{z}_{\kcurr} - \mathbf{b}\right) \right)$. 
However, since we have considered right-invariant error $\errG^R$ in our filter design, in order to incorporate the updates from the left-invariant observations, it is necessary to transform the right invariant error to be expressed as the left-invariant error (\cite{hartley2020contact}).
The switching from the right invariant error to the left invariant error can be done using the adjoint map.
\begin{equation}
	\label{eq:chap:human-motion:ekf-err-switch}
	\begin{split}
	& \errG^R =\ \Xhat \X^{-1} =\ \Xhat \X^{-1} \Xhat \Xhat^{-1} =\ \Xhat \errG^L \Xhat^{-1}, \\
	& \gexphat{\mathscr{M}}(\err^R) = \Xhat \gexphat{\mathscr{M}}(\err^L)  \Xhat^{-1} =\ \gexphat{\mathscr{M}}(\gadj{\Xhat}\ \err^L), \\
	& \err^R =\ \gadj{\Xhat}\ \err^L.
	\end{split}
\end{equation}
Similarly, we have $\err^L =\ \gadj{{\Xhat^{-1}}}\ \err^R$. This implies a switching between the covariance of right- and left-invariant errors as,
\begin{equation}
	\label{eq:chap:human-motion:ekf-cov-switch}
	\begin{split}
	& \cov^L = \gadj{{\Xhat^{-1}}}\ \cov^R\ \gadj{{\Xhat^{-1}}}^T, \\
	& \cov^R = \gadj{\Xhat}\ \cov^L\ \gadj{\Xhat}^T.    
	\end{split}
\end{equation}

Once the covariance is transformed from $\cov^R$ to $\cov^L$ thus reflecting the left-invariant errors, the filter updates for the mean and covariance can be followed to incorporate corrections from the left-invariant observations by steps described in Section \ref{sec:chapter02:invekf-matrix-lie-groups}.
Once these updates are incorporated, we may transform the covariance $\cov^L$ back to $\cov^R$ to reflect the right-invariant errors.

We consider measurement updates from terrain height measurements obtained from a known map and gyroscope measurements obtained from an IMU collocated on the base link as left-invariant observations since they directly relate to some quantities in the state-space.

\subsubsection{Terrain Height Updates}
For candidate points that are actively in contact with the environment, the height measurement from a known map is used to update the filter states. 
High covariance values are associated with $(d_x, d_y)$ coordinates while map covariance values is set for the height $d_z$.
The measurement $h_d(m)$ which is a function of known height map $m$ relates to the state variable $\mathbf{d}$.
The terrain height measurement when expressed in the form of a left invariant observation $\X_{\kcurr}\ \mathbf{b}\ +\ \mathbf{n}_{\kcurr}$ takes the form,
$$
\begin{bmatrix}
	h_d(m) \\ 0 \\ 0 \\ 1 \\ \Zeros{3}{1} \\ 0 \\ \Zeros{3}{1}
\end{bmatrix}
=\ \X
\begin{bmatrix}
	\Zeros{3}{1} \\ 0 \\ 0 \\ 1 \\ \Zeros{3}{1} \\ 0 \\ \Zeros{3}{1}
\end{bmatrix}\ +\ 
\begin{bmatrix}
	\mathbf{w}_\mathbf{d} \\ 0 \\ 0 \\ 0 \\ \Zeros{3}{1} \\ 0 \\ \Zeros{3}{1}
\end{bmatrix}.
$$
Similar to the right-invariant observations, since the innovation vector contains non-zero elements only in the first three rows, we may use an auxiliary selection matrix $\Pi_d = \begin{bmatrix} \I{3} & \Zeros{3}{10}
\end{bmatrix}$ to obtained a reduced dimensional gain $\mathbf{K}_\kcurr^r$.
The measurement model Jacobian and the measurement noise covariance can be computed through the linearization of the left invariant error-update equations as seen in Eq. \eqref{eq:chap02:invekf-linearized-update-lie},
\begin{align*}
	\begin{split}
		\err^{L+}_{\kcurr} &= \ \err^L_{\kcurr} + \mathbf{K}_\kcurr^r\ \Pi_d \left(-\ghat{G}{\err^{L}_{\kcurr}}\ \mathbf{b} + \Xhat^{-1}_{\kcurr}\mathbf{n}_{t_\kcurr}\right) \\
		&= \ \err^{L}_{\kcurr} + \mathbf{K}_\kcurr^r\ \Pi_p \left(\begin{bmatrix}
			 - \err_\mathbf{d} \\ 0 \\ 0 \\ 1 \\ \Zeros{3}{1} \\ 0 \\ \Zeros{3}{1}
		\end{bmatrix} + 
		\begin{bmatrix}
			\hat{\Rot{}{}}^T\ \mathbf{w}_\mathbf{d} \\ 0 \\ 0 \\ 0 \\ \Zeros{3}{1} \\ 0 \\ \Zeros{3}{1}
		\end{bmatrix} 
		\right) \\
		&= \ \err^{L}_{\kcurr} - \mathbf{K}_\kcurr^r\  \left(\begin{bmatrix}
			\Zero{3} & \Zero{3} & \Zero{3} & \I{3} & \Zero{3} & \Zero{3}
		\end{bmatrix} \err^{R}_{\kcurr} - 
		\hat{\Rot{}{}}^T\ \mathbf{w}_\mathbf{d}
		\right).
	\end{split}
\end{align*}

The quantities required for the filter update can then be summarized as,
\begin{equation}
\label{eq:chap:human-motion:ekf-terrain-height}
	\begin{split}
	& \mathbf{b}_d^T = \begin{bmatrix}
	\Zeros{1}{3} & 0 & 0 & 1 & \Zeros{1}{3} & 0 & \Zeros{1}{3}
	\end{bmatrix},\\
	& \Pi_d = \begin{bmatrix}
		\I{3} & \Zeros{3}{10}
	\end{bmatrix},\\
	& \mathbf{H}_d = \begin{bmatrix}
	\Zero{3} & \Zero{3} & \Zero{3} & \I{3} & \Zero{3} & \Zero{3} 
	\end{bmatrix}, \\
	& \mathbf{N}_d = \Rot{}{}^T\text{Cov}(\mathbf{w}_\mathbf{d})\Rot{}{} .
	\end{split}
\end{equation}

\subsubsection{Base Collocated Gyroscope}
One approach of considering the IMU on the pelvis link affects the system dynamics of the considered filter and requires to augment the state representation with IMU biases, leading to an approach similar to what was presented in Section \ref{sec:chap:cd-ekf-rie}.

A simpler approach can be taken by considering solely the gyroscope measurements from the pelvis IMU while disregarding any biases and formulating a left-invariant observation to be considered in a straightforward manner into the current filter design.
Considering that the IMU is rigidly attached to the pelvis link and the rotation $\Rot{B}{{B_{\text{IMU}}}}$ between the pelvis link and the pelvis IMU is known, we have the measurement ,
$$ h_g(\tilde{\omega}) = \Rot{B}{{B_{\text{IMU}}}} \yGyro{A}{{B_{\text{IMU}}}} \in \mathbb{R}^3, $$
that has a left-invariant observation structure, 
$$
\begin{bmatrix}
	\Zeros{3}{1} \\ 0 \\ 0 \\ 0 \\ h_g(\tilde{\omega}) \\ 1 \\ \Zeros{3}{1} 
\end{bmatrix}
=\ \X
\begin{bmatrix}
	\Zeros{3}{1} \\ 0 \\ 0 \\ 0 \\  \Zeros{3}{1} \\ 1 \\ \Zeros{3}{1}
\end{bmatrix}\ +\ 
\begin{bmatrix}
	\Zeros{3}{1} \\ 0 \\ 0 \\ 0 \\ \noiseGyro{B} \\ 0 \\ \Zeros{3}{1}
\end{bmatrix},
$$
leading to the quantities relevant for filter update as,
\begin{equation}
\label{eq:chap:human-motion:ekf-basegyro}
	\begin{split}
	& \mathbf{z}_g^T = \begin{bmatrix}
	\Zeros{1}{3} & 0 & 0 & 0 & h_g(\tilde{\omega}) & 1 & \Zeros{1}{3}
	\end{bmatrix},\\
	& \mathbf{b}_g^T = \begin{bmatrix}
	\Zeros{1}{3} & 0 & 0 & 0 & \Zeros{1}{3} & 1 & \Zeros{1}{3}
	\end{bmatrix},\\
	& \Pi_\omega = \begin{bmatrix} \Zeros{3}{6} & \I{3} & \Zeros{3}{4} \end{bmatrix}, \\
	& \mathbf{H}_g = \begin{bmatrix}
	\Zero{3} & \Zero{3} & \Zero{3} & \Zero{3} & \I{3} & \Zero{3}
	\end{bmatrix}, \\
	& \mathbf{N}_g = \text{Cov}(\noiseGyro{B}).
	\end{split}
\end{equation}

\subsection{Non Invariant Observations}
These measurements have the form $\Z = h (\X) \;\gexphat{G^\prime}(\mathbf{n})$  evolving over a distinct matrix Lie group $G^\prime$.
The choice of right invariant error leads to the innovation term 
$$
\mathbf{\tilde{z}} = \glogvee{G^\prime}\big(h^{-1}(\Xhat)\; h(\gexp{\mathscr{M}}(-\err) \Xhat)\big),
$$ 
and the measurement model Jacobian as $\mathbf{H} = -\frac{\partial}{\partial \err}\; \glogvee{G^\prime}\big(h^{-1}(\Xhat)\; h(\gexp{\mathscr{M}}(-\err)\;\Xhat)\big)\bigg|_{\err = 0}$.
The Kalman gain is computed as $ \mathbf{K}_\knext = \cov_\kpred\;\mathbf{H}_\knext^T\;\left(\mathbf{H}_\knext\;\cov_\kpred\;\mathbf{H}_\knext^T +\; \mathbf{N}_\knext\right)^{-1}$ and is used to determine the state update in the tangent space as $\mathbf{m}^{-}_\knext = \mathbf{K}_\knext\; \mathbf{\tilde{z}}_\knext $, which is further used for the state reparametrization $\Xhat_\knext = \gexphat{\mathscr{M}}\big(\mathbf{m}^{-}_\knext\big) \Xhat_\kpred$. 
The state covariance is updated as $\cov_\knext = \J^l_G(\mathbf{m}^{-}_\knext)\;(\I{p} - \mathbf{K}_\knext\;\mathbf{H}_\knext)\cov_\kpred\;\J^l_G(\mathbf{m}^{-}_\knext)^T$.

We consider relative orientations of the foot link and a terrain height update from a known map and contact plane orientation updates as non-invariant observations.
It must be noted that the non-invariant observation for the terrain height is used as a computationally flexible alternative for the left-invariant observation described in the previous section. \looseness=-1

\subsubsection{Relative foot link rotation}
The joint positions $\encoders$ obtained from the integration of estimated joint velocities $\encoderSpeeds$ is passed through the forward kinematics map to determine the relative foot orientations. 
The measurement model for the measurement $h_R(\encoders)$ of foot orientation relative to base link is expressed as,\looseness=-1
$$
 h_R(\X)= \mathbf{R}^T\;\mathbf{Z}\; \gexphat{\SO{3}}\big(\relativeJacobianLeftTrivAng{B}{F}(\encoders) \; \encoderNoise\big) \in \SO{3}.
$$
In order to obtain the measurement model Jacobian, we need to compute,
$$
h^{-1}(\Xhat)\; h(\gexp{\mathscr{M}}(-\err) \Xhat).
$$
The inverse of measurement model is obtained as $h^{-1}(\Xhat) = \mathbf{\hat{Z}}^T\mathbf{\hat{R}}$ and the measurement model of the perturbed state is computed as,
\begin{align*}
h(\gexp{\mathscr{M}}(-\err) \Xhat) &= \left(\mathbf{\hat{R}} - S(\err_\mathbf{R}) \mathbf{\hat{R}}\right)^T \left(\mathbf{\hat{Z}} - S(\err_\mathbf{Z}) \mathbf{\hat{Z}}\right) \\
&= \left(\mathbf{\hat{R}}^T + \mathbf{\hat{R}}^T S(\err_\mathbf{R})\right) \left(\mathbf{\hat{Z}} - S(\err_\mathbf{Z}) \mathbf{\hat{Z}}\right) \\
&\approx \mathbf{\hat{R}}^T\mathbf{\hat{Z}} - \mathbf{\hat{R}}^T\ S(\err_\mathbf{Z}) \mathbf{\hat{Z}} + \mathbf{\hat{R}}^T\ S(\err_\mathbf{R}) \mathbf{\hat{Z}}.
\end{align*}
On composing the two rotations obtained from $h^{-1}(\Xhat)$ and $h(\gexp{\mathscr{M}}(-\err) \Xhat)$, we get
$$
h^{-1}(\Xhat)\; h(\gexp{\mathscr{M}}(-\err) \Xhat) = \I{3} - S(\mathbf{\hat{Z}}^T\err_\mathbf{Z}) + S(\mathbf{\hat{Z}}^T\err_\mathbf{R}).
$$
On applying the $\glogvee{\SO{3}}$ operator, we have,
$$
\glogvee{\SO{3}}\left(h^{-1}(\Xhat)\; h(\gexp{\mathscr{M}}(-\err) \Xhat)\right) = \hat{\mathbf{Z}}^T\; \err_\mathbf{R} -\hat{\mathbf{Z}}^T\; \err_\mathbf{Z}.
$$
The measurement model Jacobian can be obtained by computing the negative of partial derivatives of the above equation.
The filter quantities relevant for measurement update can thus be written as,
\begin{equation}
	\label{eq:eq:chap:human-motion:ekf-relrotmeas}
	\begin{split}
	& \mathbf{H}_R = \begin{bmatrix}
	\Zero{3} & -\mathbf{\hat{Z}}^T & \Zero{3} & \Zero{3} & \Zero{3} & \mathbf{\hat{Z}}^T
	\end{bmatrix}, \\
	& \mathbf{N}_R = \; \relativeJacobianLeftTrivAng{B}{F}(\encoders) \; \text{Cov}(\encoderNoise) \; {\relativeJacobianLeftTrivAng{B}{F}}^T(\encoders).
	\end{split}
\end{equation}

\subsubsection{Terrain Height Update as a Non-Invariant Observation}
As already noticed, incorporating a left-invariant observation within a Right-Invariant EKF structure requires a transformation of the covariance from the global to local frame through the adjoint matrix.
Although such a transformation is quite useful to incorporate multiple sources of information, this operation might be computationally expensive in the case of a high-dimensional state representation, such as ours.
This is much more evident, due to the consideration of the vertex positions of the feet within the state resulting in a large number of matrix-vector (for residuals) and matrix-matrix multiplication operations (for covariance).
In this case, we may alternately choose to represent this measurement as a non-invariant observation which will lessen the computational burden on the filter.
The measurement model for the terrain height measurement is expressed as,
$$
h_d(\X) =\mathbf{d} + \mathbf{w}_\mathbf{d} \in \T{3}.
$$
We compute the following quantity using the first-order approximation of the exponential map as,
\begin{align*}
h^{-1}(\Xhat)\; h(\gexp{\mathscr{M}}(-\err) \Xhat) &= \begin{bmatrix}
	\I{3} & -\hat{\mathbf{d}} \\ \Zeros{1}{3} & 1 
\end{bmatrix}\; \begin{bmatrix}
\I{3} & \hat{\mathbf{d}} + S(\hat{\mathbf{d}})\err_{\Rot{}{}}  - \err_{\mathbf{d}} \\ \Zeros{1}{3} & 1 
\end{bmatrix} \\
& = \begin{bmatrix}
	\I{3} & S(\hat{\mathbf{d}})\err_{\Rot{}{}}  - \err_{\mathbf{d}} \\ \Zeros{1}{3} & 1 
\end{bmatrix}.
\end{align*}

On applying the $\glogvee{\T{3}}$ operator, we have,
$$
\glogvee{\T{3}}\left(h^{-1}(\Xhat)\; h(\gexp{\mathscr{M}}(-\err) \Xhat)\right) = S(\hat{\mathbf{d}})\err_{\Rot{}{}}  - \err_{\mathbf{d}}.
$$
The filter quantities relevant for measurement update can be written as,
\begin{equation}
	\label{eq:eq:chap:human-motion:ekf-relrotmeas}
	\begin{split}
	& \mathbf{H}_d = \begin{bmatrix}
	\Zero{3} & -S(\mathbf{\hat{d}}) & \Zero{3} & \I{3} & \Zero{3} & \Zero{3}
	\end{bmatrix}, \\
	& \mathbf{N}_d = \; \text{Cov}(\mathbf{w}_d).
	\end{split}
\end{equation}

\subsubsection{Contact Plane Orientation Update}
In addition to the terrain height updates, another source of information that may help to better constrain the estimation problem is the contact plane orientation.
In cases where, the foot is in complete planar rigid contact with the environment, we may enable the contact plane orientation update.
The contact plane orientation can either be obtained by fitting a plane to the vertices of the foot or through active perception.
We relate the contact plane orientation directly to the foot orientation $\Z$ and this leads to the measurement model,
$$
h_c(\X) = \mathbf{\Z}\; \gexphat{\SO{3}}(\mathbf{w}_c) \in \SO{3}
$$
In order to obtain the measurement model Jacobian, we need to compute the partial derivatives after applying the logarithm operator  to 
$$
h^{-1}(\Xhat)\; h(\gexp{\mathscr{M}}(-\err) \Xhat),
$$ 
which in turn can be obtained using first-order approximation of the exponential map as,
\begin{align*}
	\begin{split}
		h^{-1}(\Xhat)\; h(\gexp{\mathscr{M}}(-\err) \Xhat) &\approx \hat{\mathbf{Z}}^T\; \left(\left(\I{3} - S(\err_\mathbf{Z})\right) \hat{\mathbf{Z}} \right) \\
		& = \hat{\mathbf{Z}}^T\; \left(\hat{\mathbf{Z}} - S(\err_\mathbf{Z}) \hat{\mathbf{Z}} \right) \\
		&= \I{3} - \hat{\mathbf{Z}}^T  S(\err_\mathbf{Z}) \hat{\mathbf{Z}} \\
		&= \I{3} +   S(-\hat{\mathbf{Z}}^T\; \err_\mathbf{Z}).
	\end{split}
\end{align*}
On applying the $\glogvee{\SO{3}}$ operator, we have,
$$
\glogvee{\SO{3}}\left(h^{-1}(\Xhat)\; h(\gexp{\mathscr{M}}(-\err) \Xhat)\right) = -\hat{\mathbf{Z}}^T\; \err_\mathbf{Z}.
$$
The filter quantities relevant for measurement update can be written as,
\begin{equation}
	\label{eq:eq:chap:human-motion:ekf-relrotmeas}
	\begin{split}
	& \mathbf{H}_c = \begin{bmatrix}
	\Zero{3} & \Zero{3} & \Zero{3} & \Zero{3} & \Zero{3} & \mathbf{\hat{Z}}^T
	\end{bmatrix}, \\
	& \mathbf{N}_c = \; \text{Cov}(\mathbf{w}_c).
	\end{split}
\end{equation}

Thus, the EKF is constructed using constant system dynamics as the prediction model and corrective updates obtained from relative forward kinematics as a function of joint position and velocities estimated by the dynamical IK optimization, base collocated gyroscope measurement and terrain information in order to obtain estimates of the base link state and the foot pose.



\section{Experimental Results}
\label{sec:chap:human-motion:experiments}

To validate the entire pipeline of the kinematic-free joint state and floating base estimation, we first rely on data simulated from real-world walking experiments conducted on a robotic platform.
This preliminary validation demonstrates the use of the presented method also for robots modeled with the framework of multi-rigid body systems.
We then demonstrate the application of the proposed methodology for human motion estimation for in-place walking, squatting, and in-place swinging experiments. \looseness=-1

\subsection{Robot Experiments}
We validate the proposed method for a position-controlled walking experiment conducted on the robot.
The robot is made to walk $0.5$ meters in the Vicon room equipped with reflective markers on its base link for obtaining the ground truth reference trajectory.

With the help of the ground-truth trajectory of the base link $\Transform{A}{B}$ and encoder measurements $\encoders, \encoderSpeeds$, we obtain the measurements for link rotations $\Rtilde{A}{L}$ and link angular velocities $\yGyro{A}{L}$ through forward kinematics which is then used as inputs for the Dynamical Inverse Kinematics block.
The contact wrench measurements acting on the robot's feet are obtained from a whole-body dynamics estimation algorithm described in \cite{nori2015icub}.
The dimensions for the rectangular approximation of the foot are determined from a bounding box of the foot mesh retrieved from the URDF model.
The contact wrenches and the rectangular foot approximation are used to compute the normal forces at the vertices which are then used as inputs to a Schmitt Trigger thresholding based contact detector.
The Schmitt trigger is tuned with make and break
thresholds as $30$ N and $15$ N respectively with stable switching time parameters as $0.01$
seconds for both making and breaking contacts at each vertex. 
The joint states computed from the IK and the contact states inferred from the thresholding are then passed onto the base estimator.
The base estimator is activated for operation only after the errors between the target measurements and the internal state of the IK block fall below an error threshold of $10^{-4}$ \si{\radian}. 
This error threshold is reached within a few iterations depending on the choice of gains used for the IK.
The base estimator is enabled to force terrain height measurement updates corresponding to a flat floor, while the contact plane orientation update is disabled for this experiment.
Noise parameters  used for the base estimator are tabulated in Table \ref{table:hbe:noise-parameters}.
The priors deviations are chosen as the same as Table \ref{table:diligent-kio:prior-parameters}. \looseness=-1

\begin{table}
\centering
						\begin{tabular}{|c|c|}
						\hline 
							Sensor & noise std dev. \\
								\hline	
							Gyroscope & 0.01 \si{\radian \per \second} \\
							Contact foot lin. velocity & $10^{-3}$ \si{\meter\per\second}\\
							Contact foot ang. velocity & $10^{-3}$ \si{\radian \per \second} \\ 
							Base lin. velocity prediction & $10$ \si{\meter \per \second} \\	
							Base ang. velocity prediction & $10$ \si{\radian \per \second} \\ 
							Terrain height & $0.03$ \si{\meter} \\
							Joint position noise &  $0.00872$ \si{\radian}\\ \hline
						\end{tabular}
						\caption{Noise parameters used for the base estimator.}
	\label{table:hbe:noise-parameters}
\end{table}

\begin{figure}[!t]
	\begin{subfigure}{\textwidth}
		\centering
\includegraphics[scale=0.53]{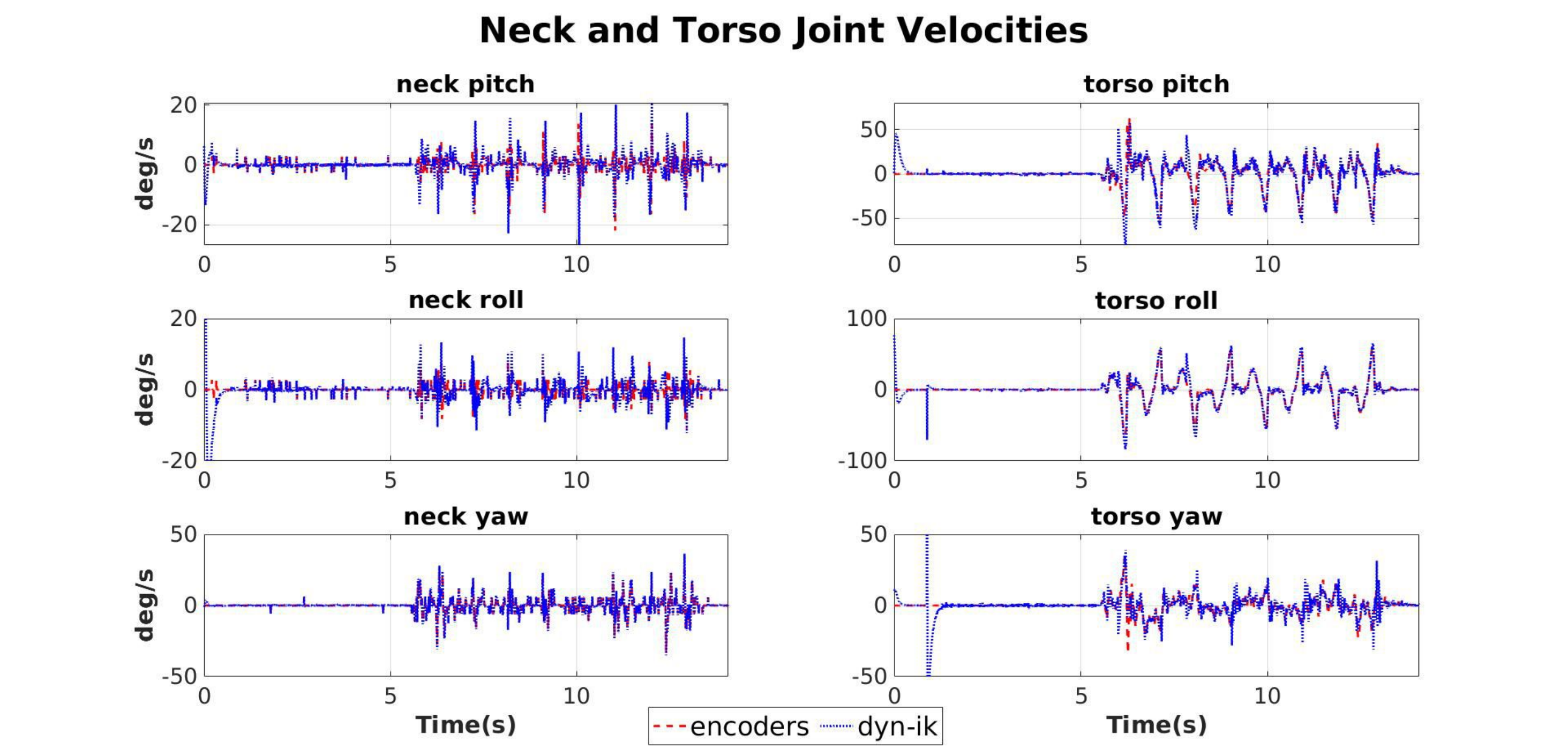}
\label{fig:chap:hbe:robot-jvel-upper-body}
	\end{subfigure}
\begin{subfigure}{\textwidth}
		\centering
\includegraphics[scale=0.53]{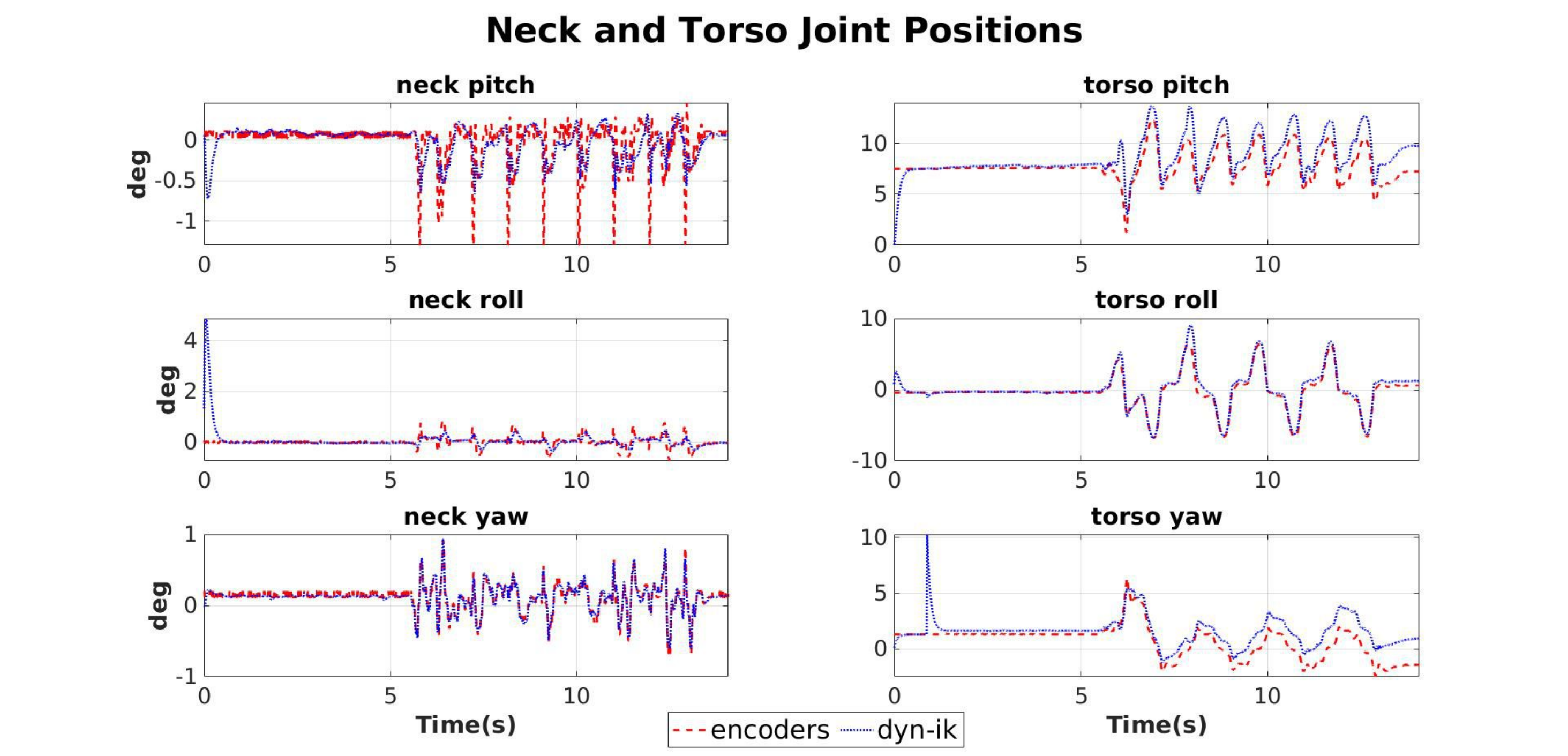}
\label{fig:chap:hbe:robot-jpos-upper-body}
	\end{subfigure}	
\caption{Upper body joint velocities and positions from Dynamical Inverse Kinematics block  in comparison with joint encoder measurements for the robot walking experiment.}
\label{fig:chap:hbe:robot-joint-states-upper-body}
\end{figure}

\begin{figure}[!t]
	\begin{subfigure}{\textwidth}
		\centering
\includegraphics[scale=0.54]{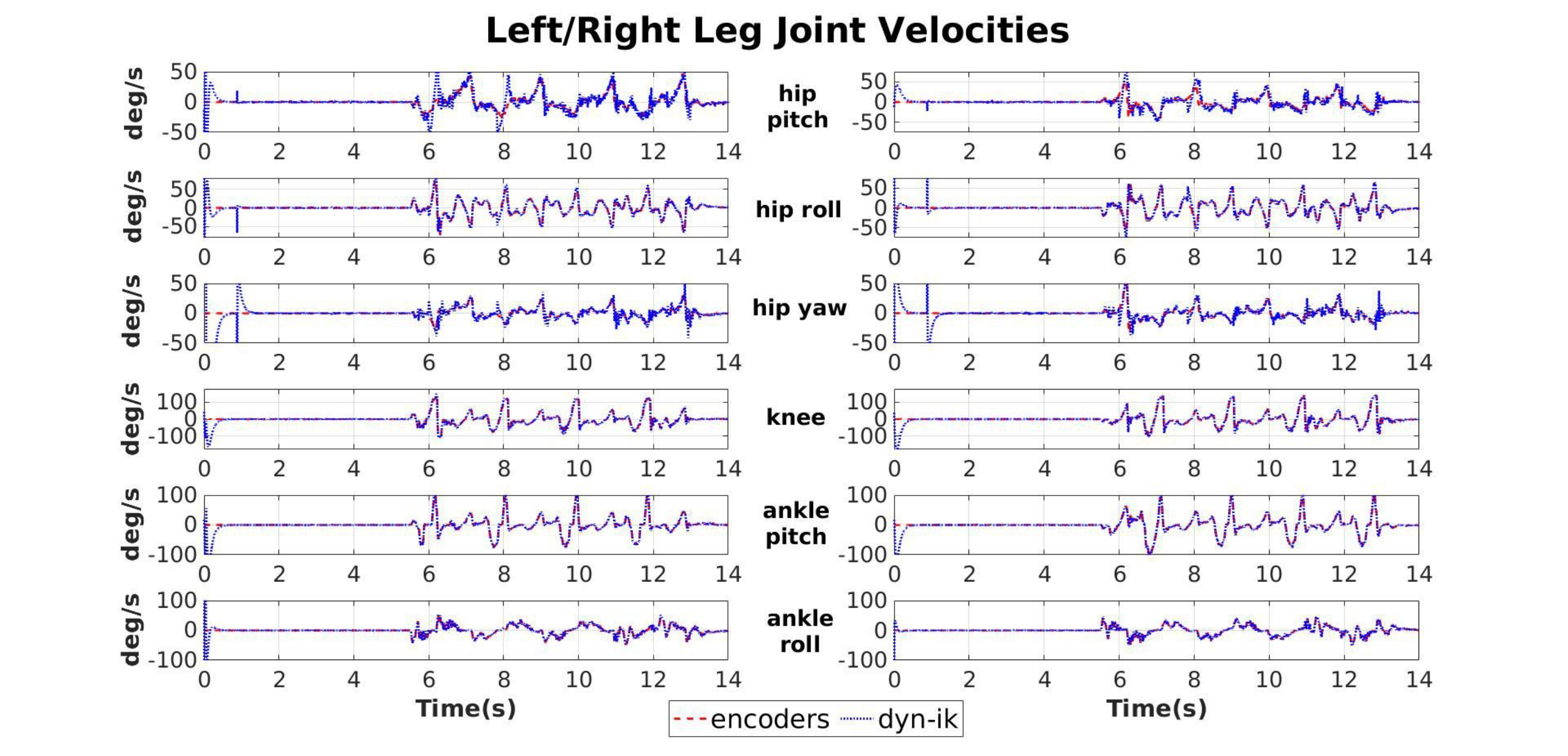}
\label{fig:chap:hbe:robot-jvel-leg}
	\end{subfigure}
\begin{subfigure}{\textwidth}
		\centering
\includegraphics[scale=0.54]{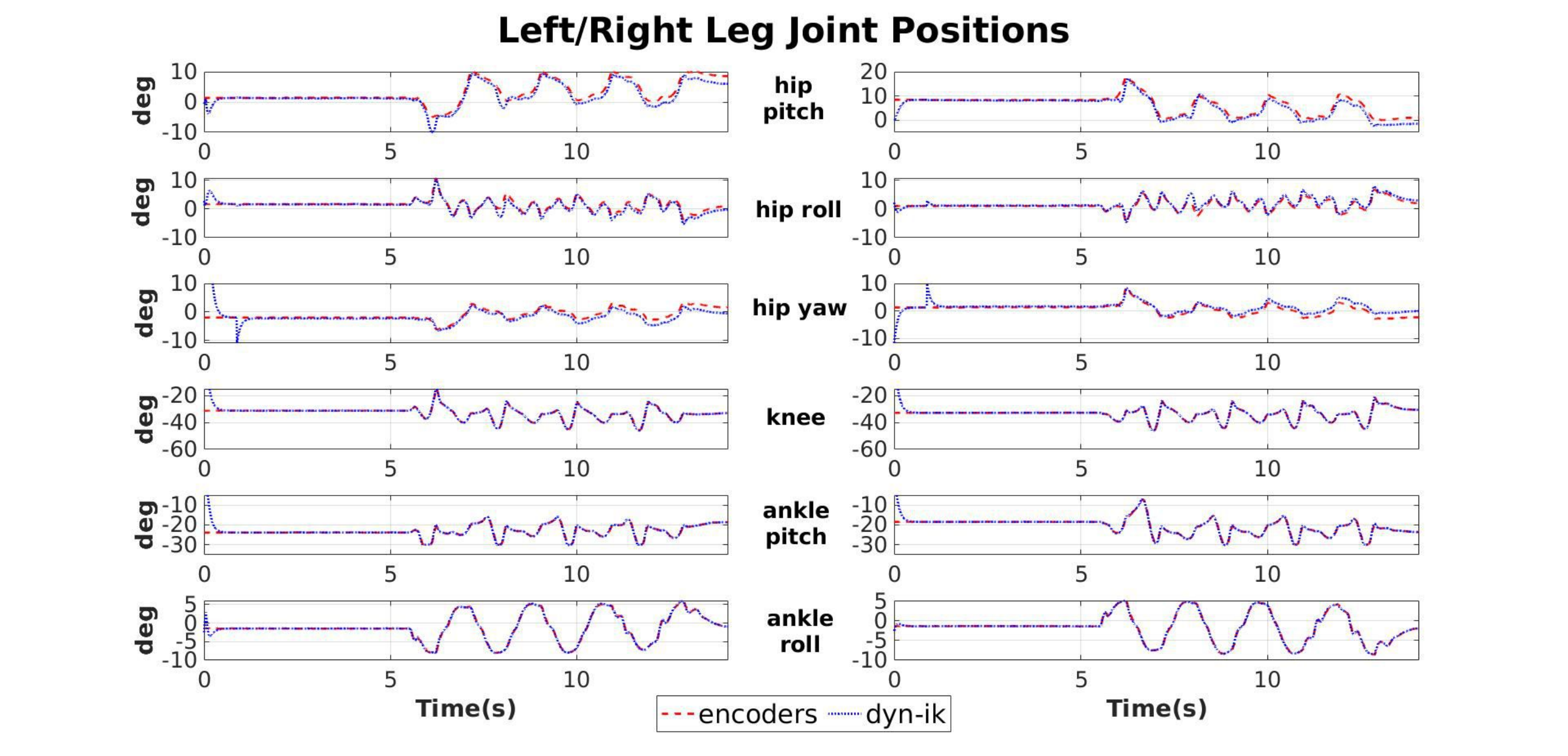}
\label{fig:chap:hbe:robot-jpos-leg}
	\end{subfigure}	
\caption{Left and right leg joint velocities and positions from Dynamical Inverse Kinematics block  in comparison with joint encoder measurements for the robot walking experiment.}
\label{fig:chap:hbe:robot-joint-states-leg}
\end{figure}

Figures \ref{fig:chap:hbe:robot-joint-states-upper-body} and \ref{fig:chap:hbe:robot-joint-states-leg} show the joint state evolution for the joints on the upper body and legs of the robot in comparison with the encoder measurements.
The joint states of the arms are omitted to be shown since they do not affect the base estimation.
Proper reconstruction of a corresponding joint state relies on the availability of measurements from the links connected by the joints.
A regularized, weighted pseudo-inverse is used to solve the inverse differential kinematics.
For the first few iterations, there exists a huge error between the estimated joint states and the encoder measurements, nevertheless, the joint states soon converge towards the encoder measurements, since the dynamical IK closes the loop using the error in forward kinematics as feedback.

\begin{figure}[!t]
	\begin{subfigure}{\textwidth}
		\centering
\includegraphics[scale=0.2, width=\textwidth]{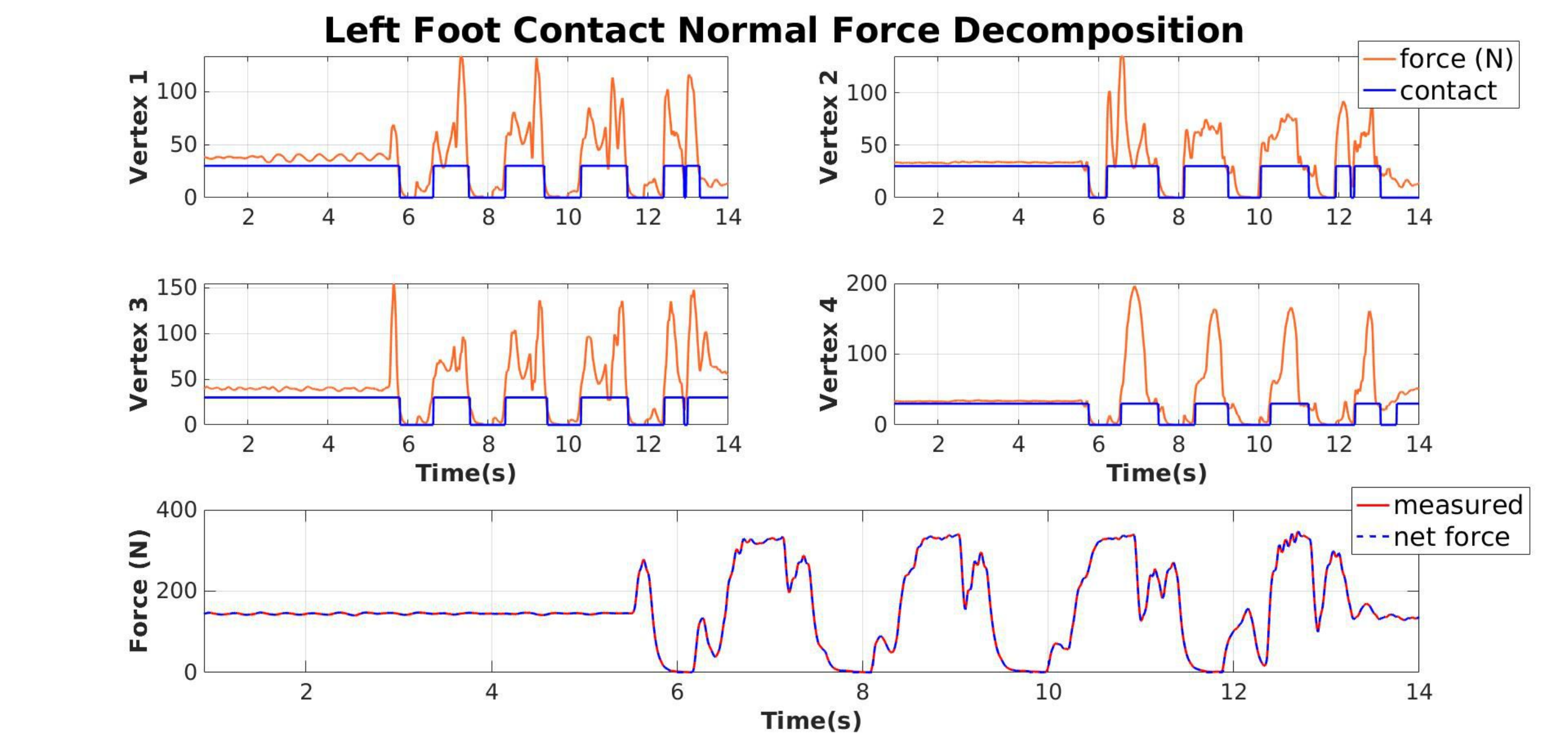}
	\end{subfigure}
\begin{subfigure}{\textwidth}
		\centering
\includegraphics[scale=0.2, width=\textwidth]{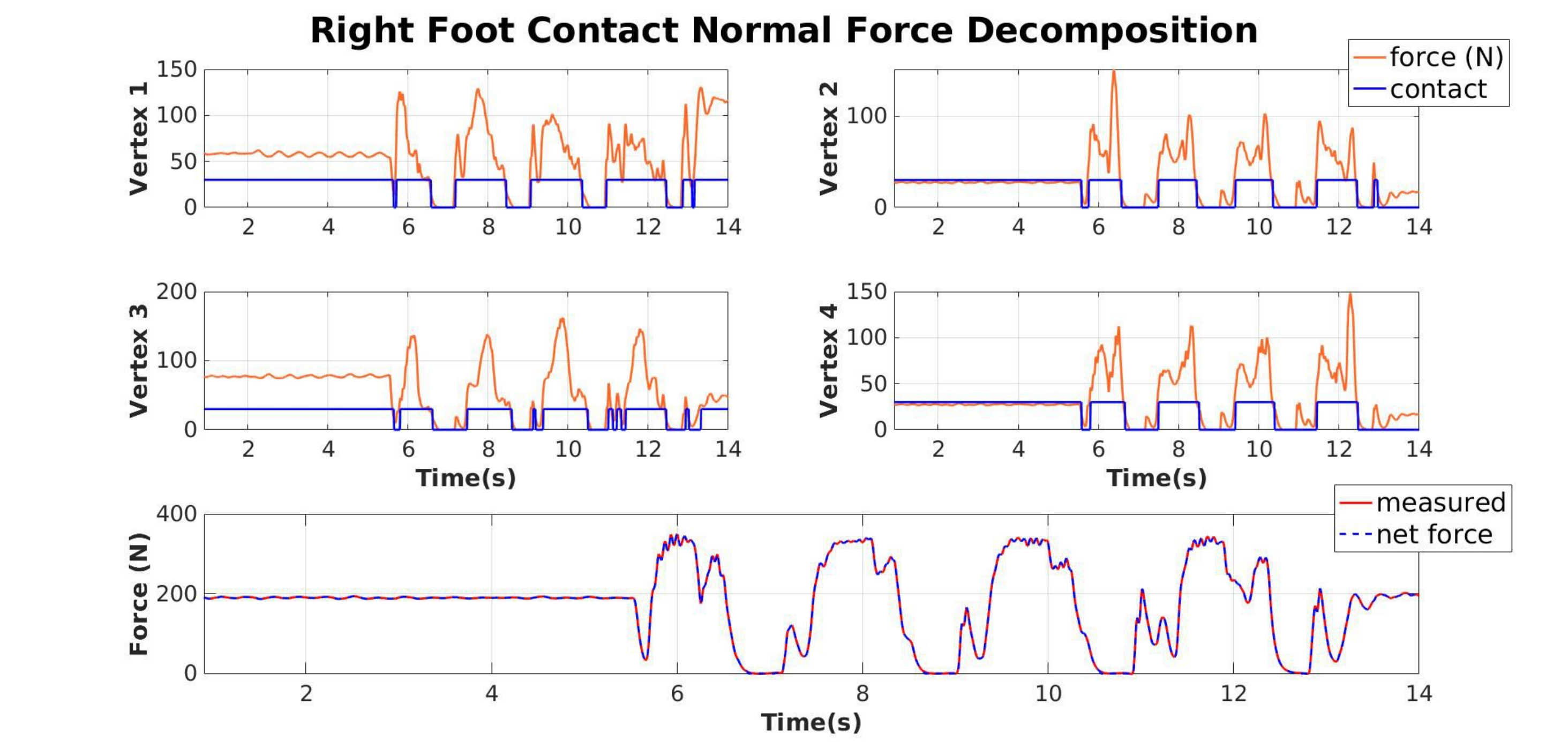}
	\end{subfigure}	
\caption{Contact wrench decomposition to contact normal forces at the vertices for left and right rectangular foot for the robot walking experiment.}
\label{fig:chap:hbe:robot-foot}
\end{figure}

\begin{figure}[!h]
\centering
	\begin{subfigure}{0.22\textwidth}
		\centering
\includegraphics[scale=0.125]{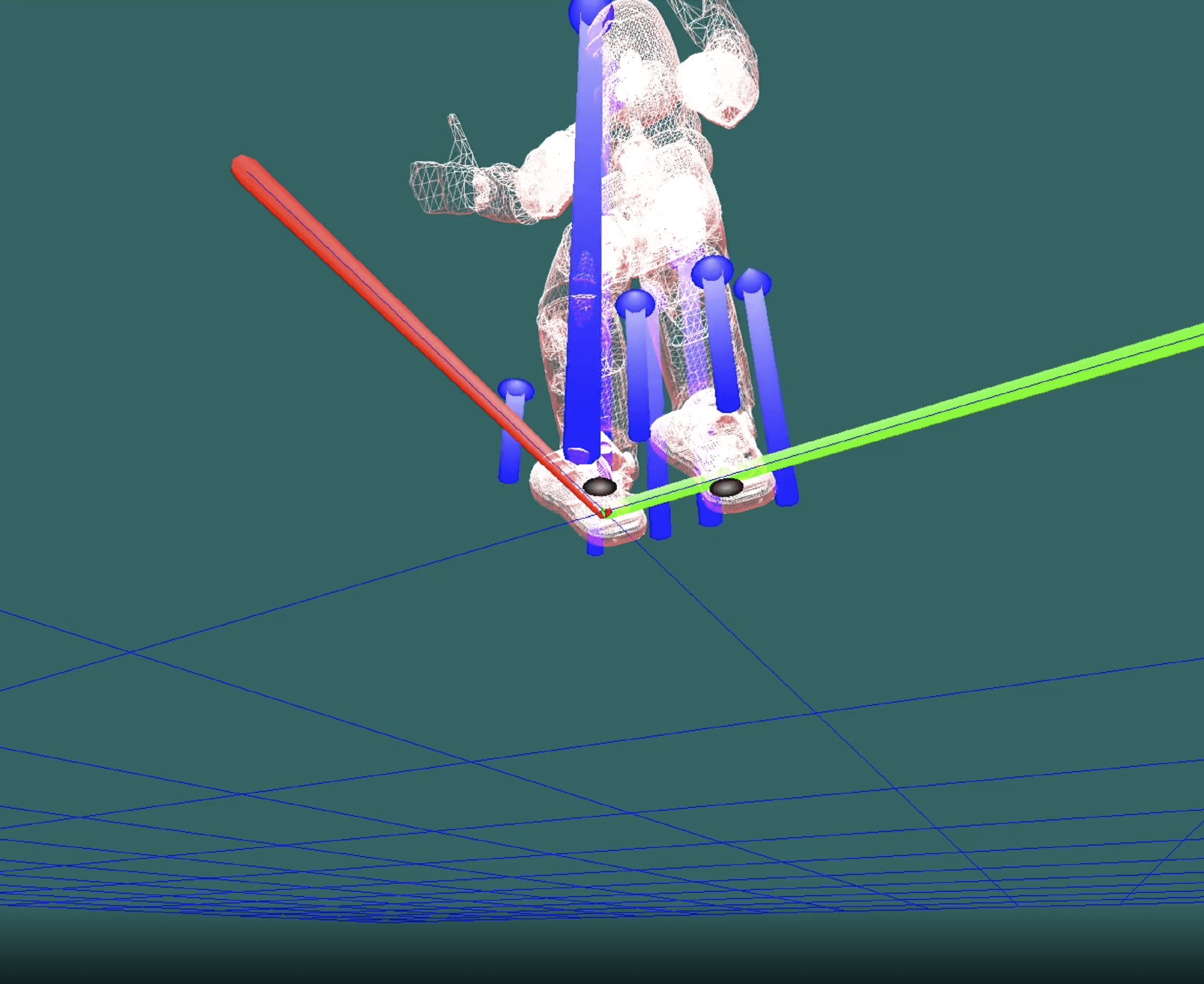}
	\end{subfigure}
\begin{subfigure}{0.22\textwidth}
		\centering
\includegraphics[scale=0.125]{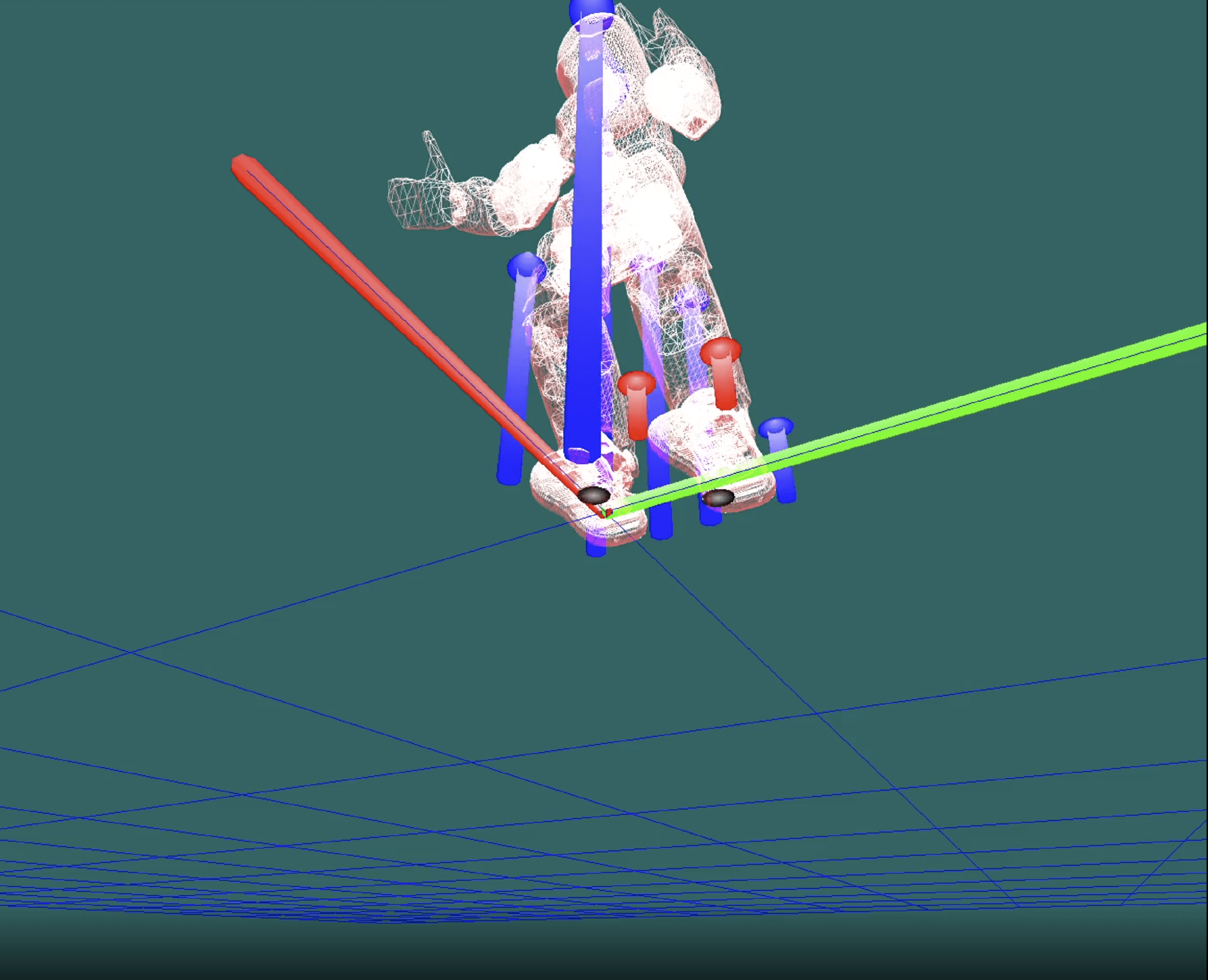}
	\end{subfigure}	
	\begin{subfigure}{0.22\textwidth}
		\centering
\includegraphics[scale=0.125]{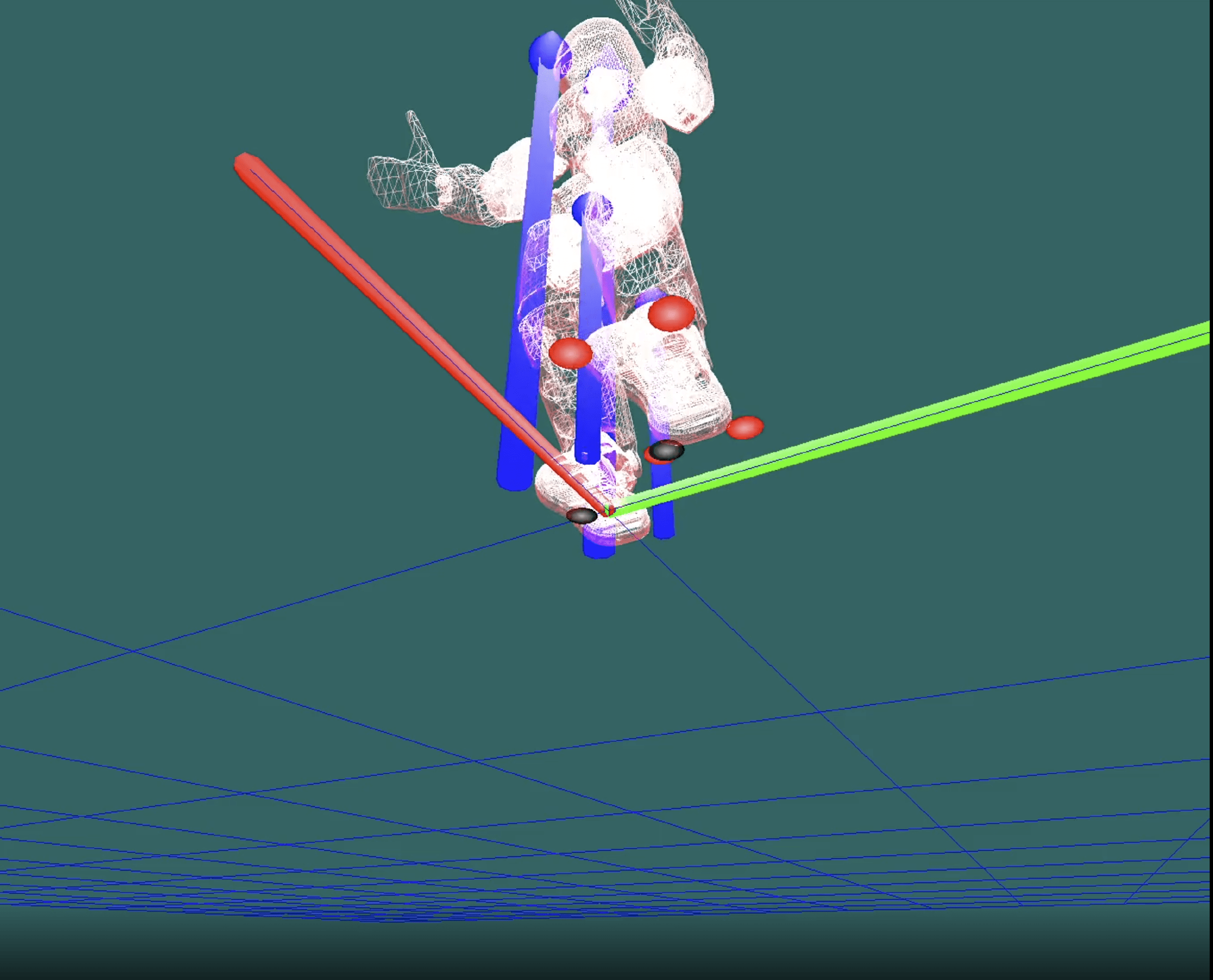}
	\end{subfigure}
\begin{subfigure}{0.22\textwidth}
		\centering
\includegraphics[scale=0.125]{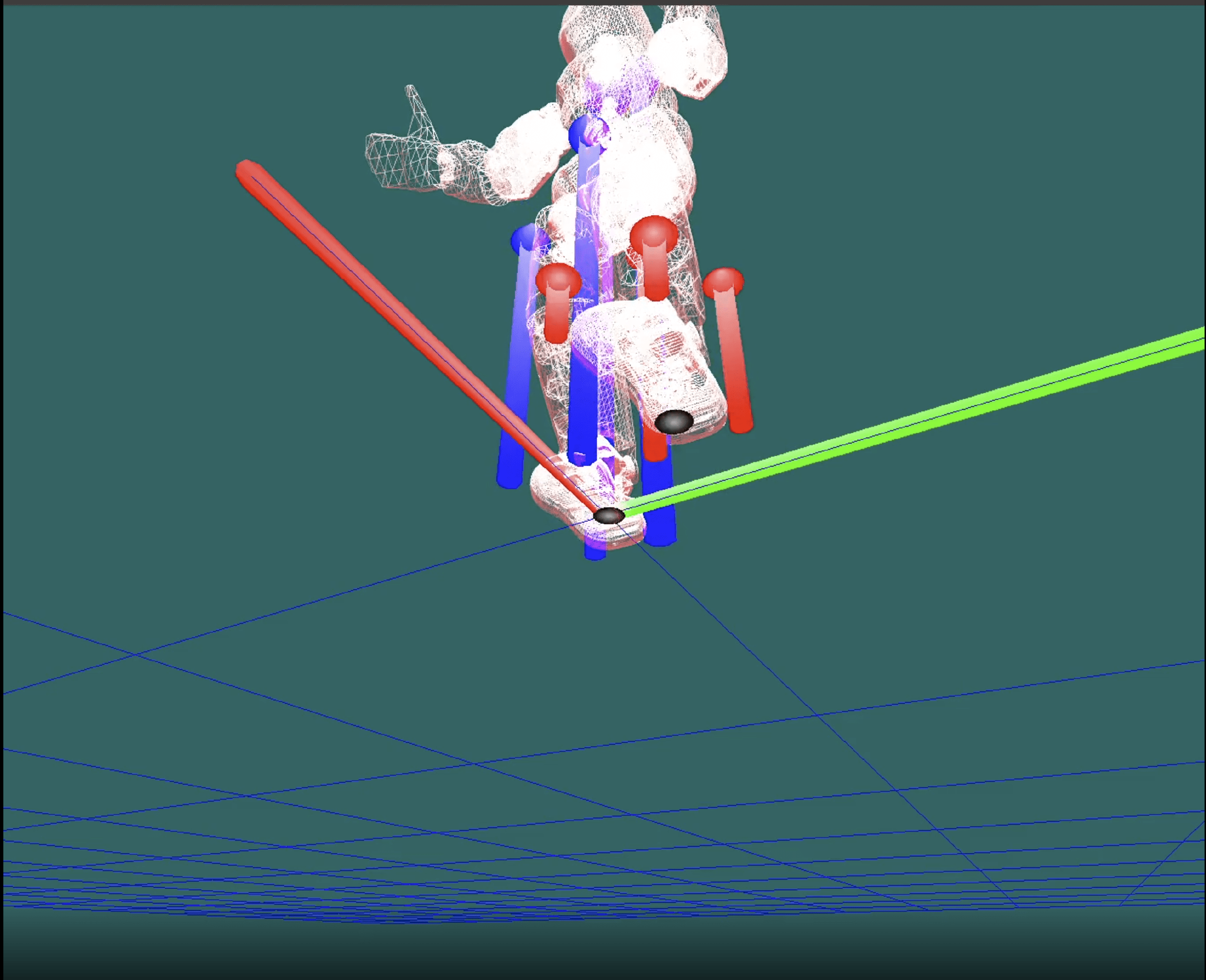}
	\end{subfigure}
	\begin{subfigure}{0.22\textwidth}
		\centering
\includegraphics[scale=0.125]{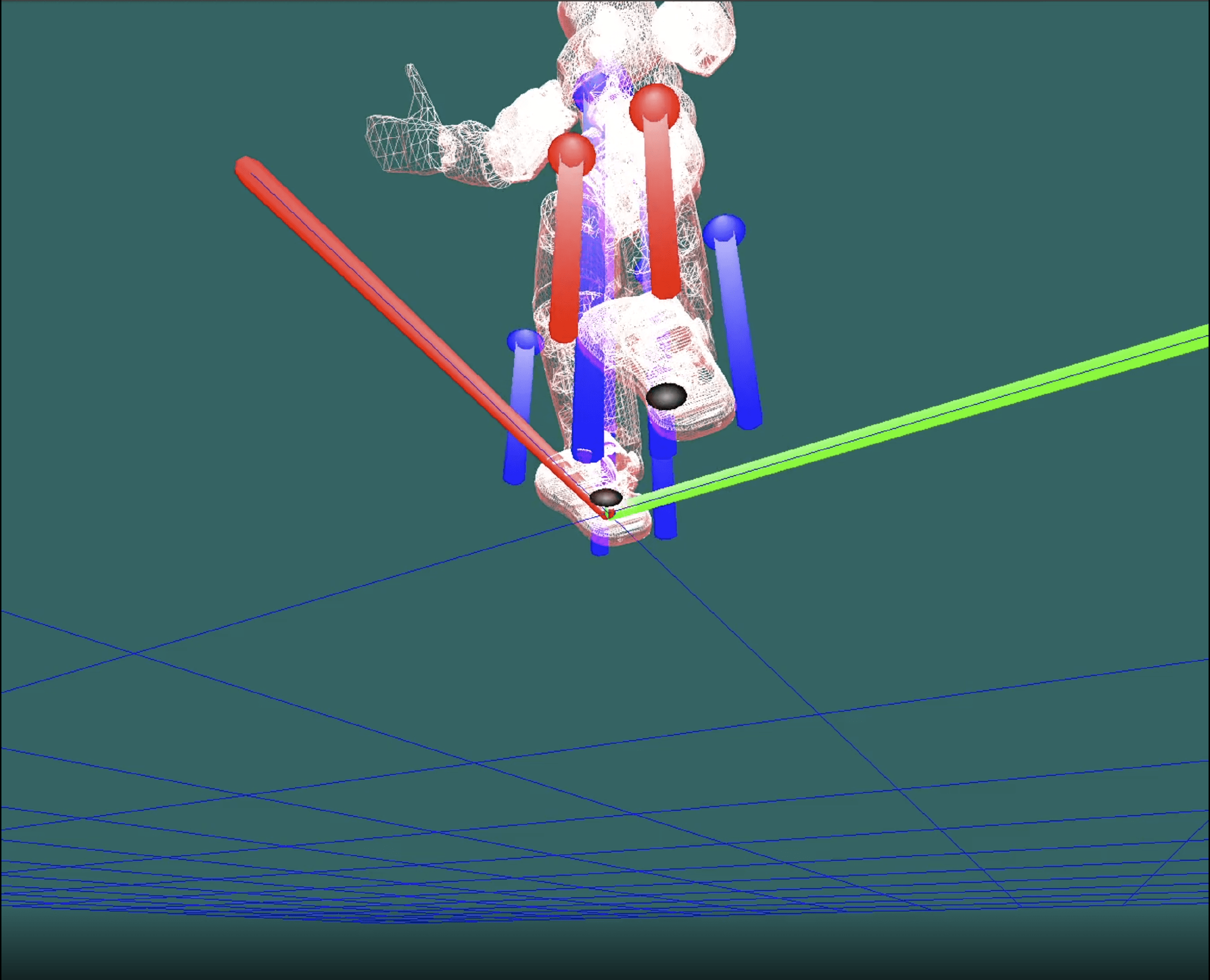}
	\end{subfigure}
\begin{subfigure}{0.22\textwidth}
		\centering
\includegraphics[scale=0.125]{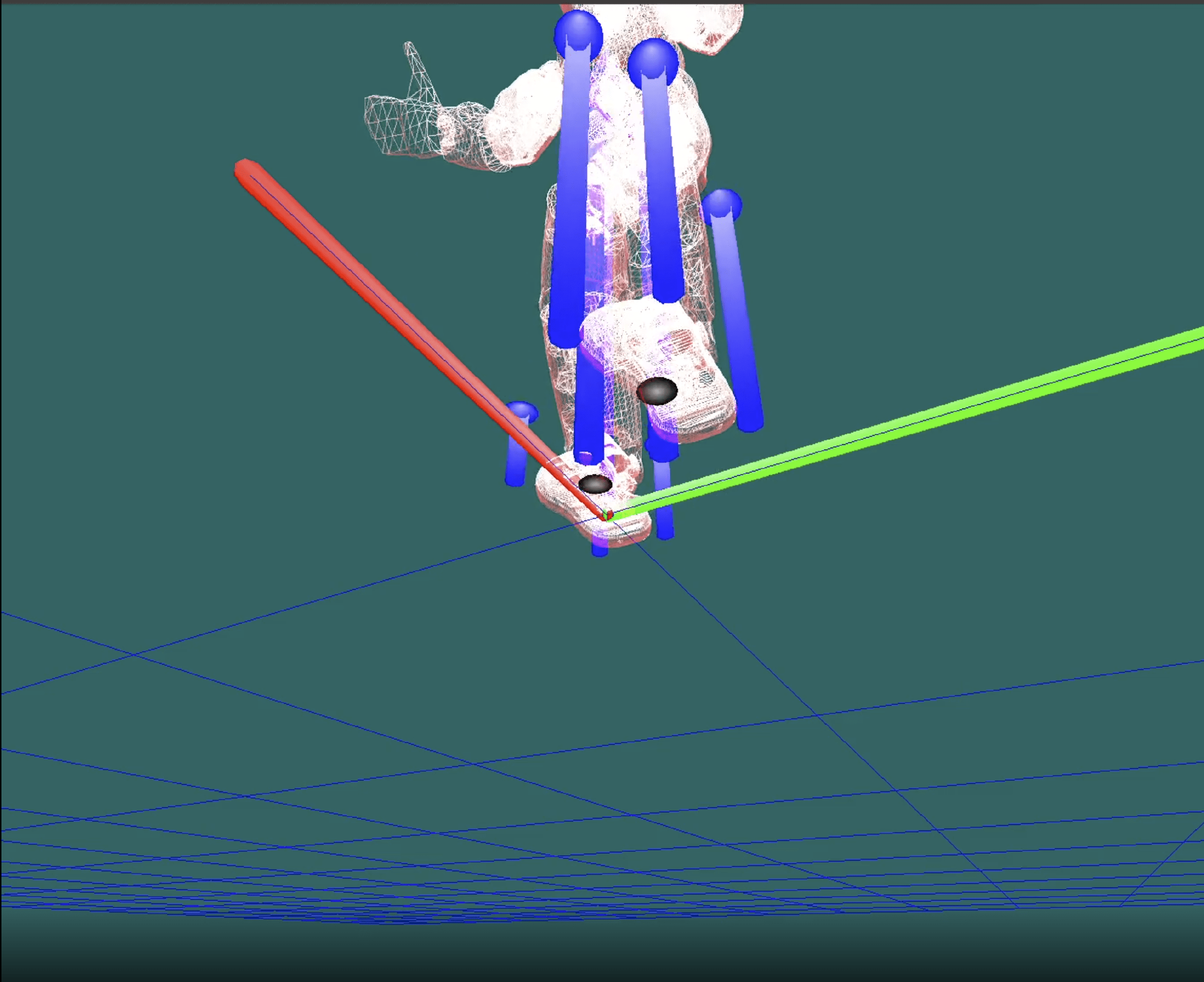}
	\end{subfigure}	
	\begin{subfigure}{0.22\textwidth}
		\centering
\includegraphics[scale=0.125]{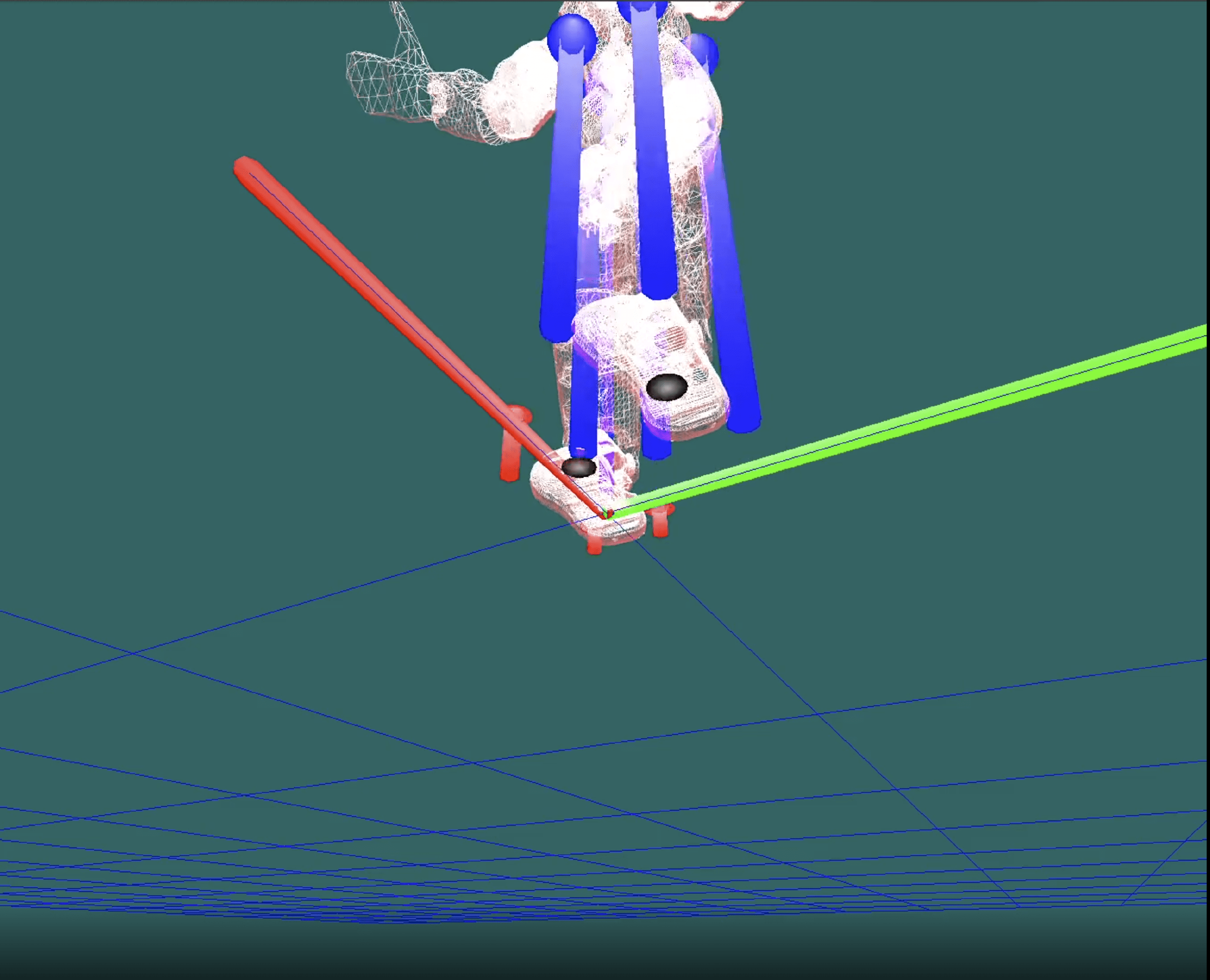}
	\end{subfigure}
\begin{subfigure}{0.22\textwidth}
		\centering
\includegraphics[scale=0.125]{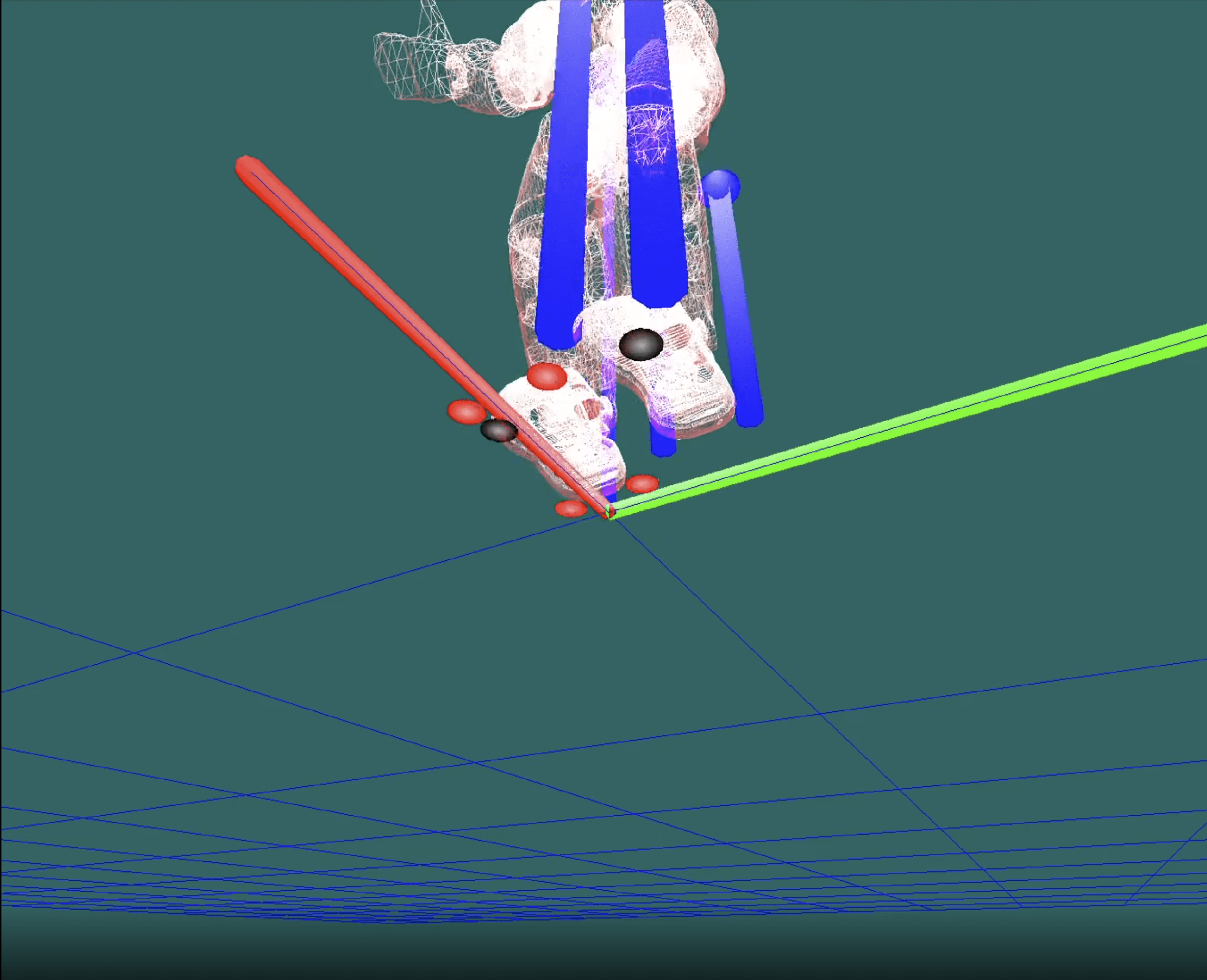}
	\end{subfigure}	
		\begin{subfigure}{0.22\textwidth}
		\centering
\includegraphics[scale=0.125]{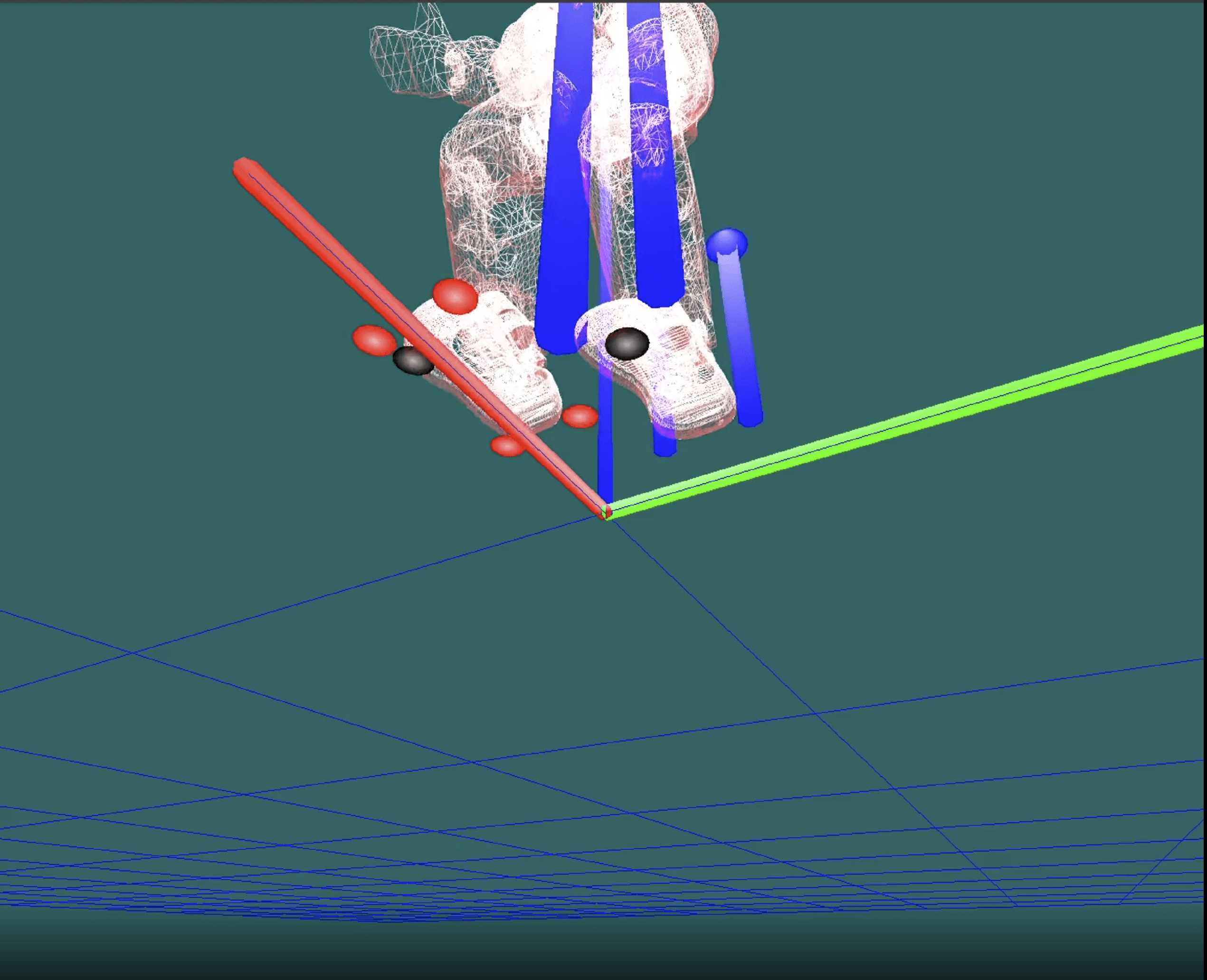}
	\end{subfigure}
\begin{subfigure}{0.22\textwidth}
		\centering
\includegraphics[scale=0.125]{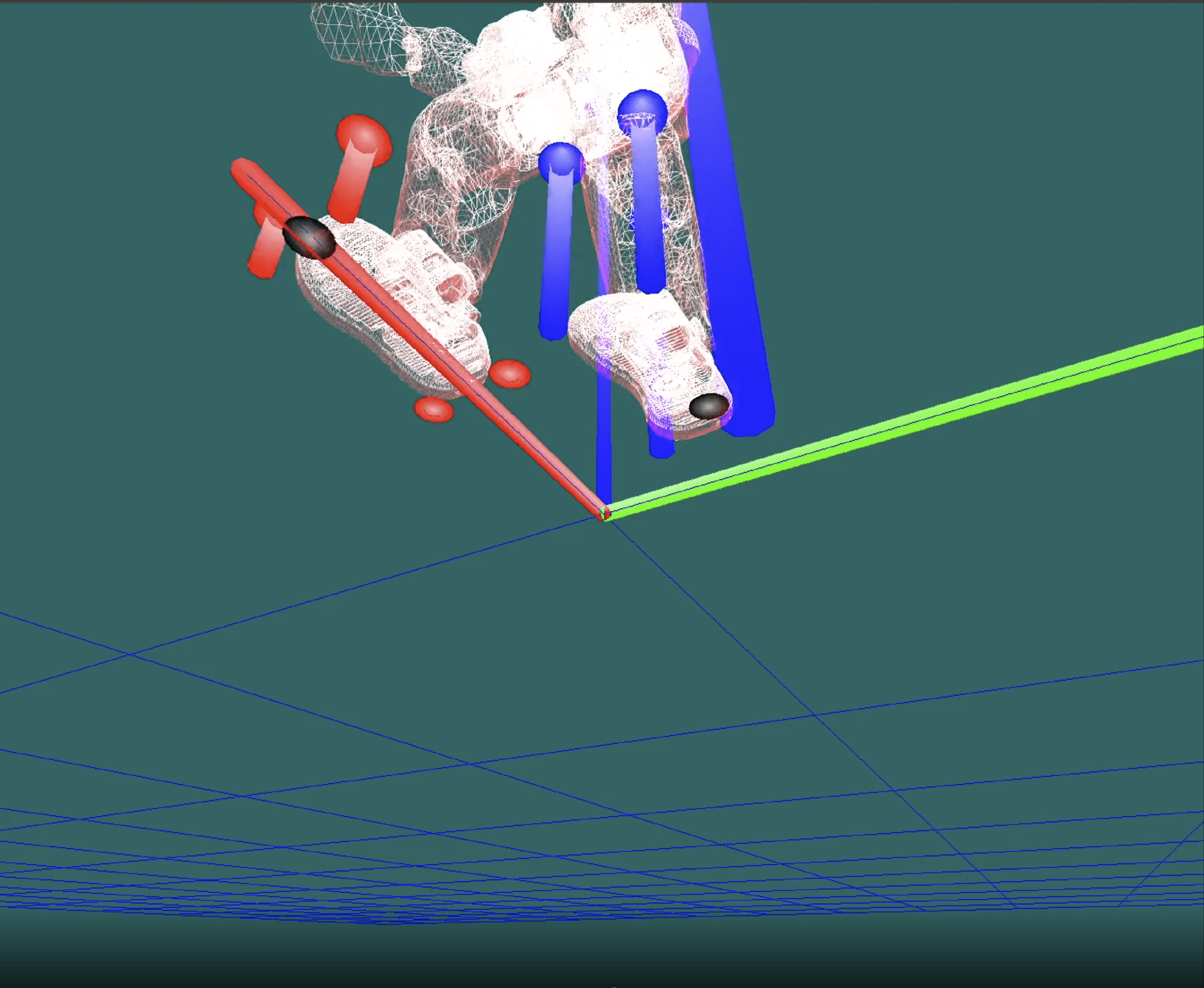}
	\end{subfigure}	
	\begin{subfigure}{0.22\textwidth}
		\centering
\includegraphics[scale=0.125]{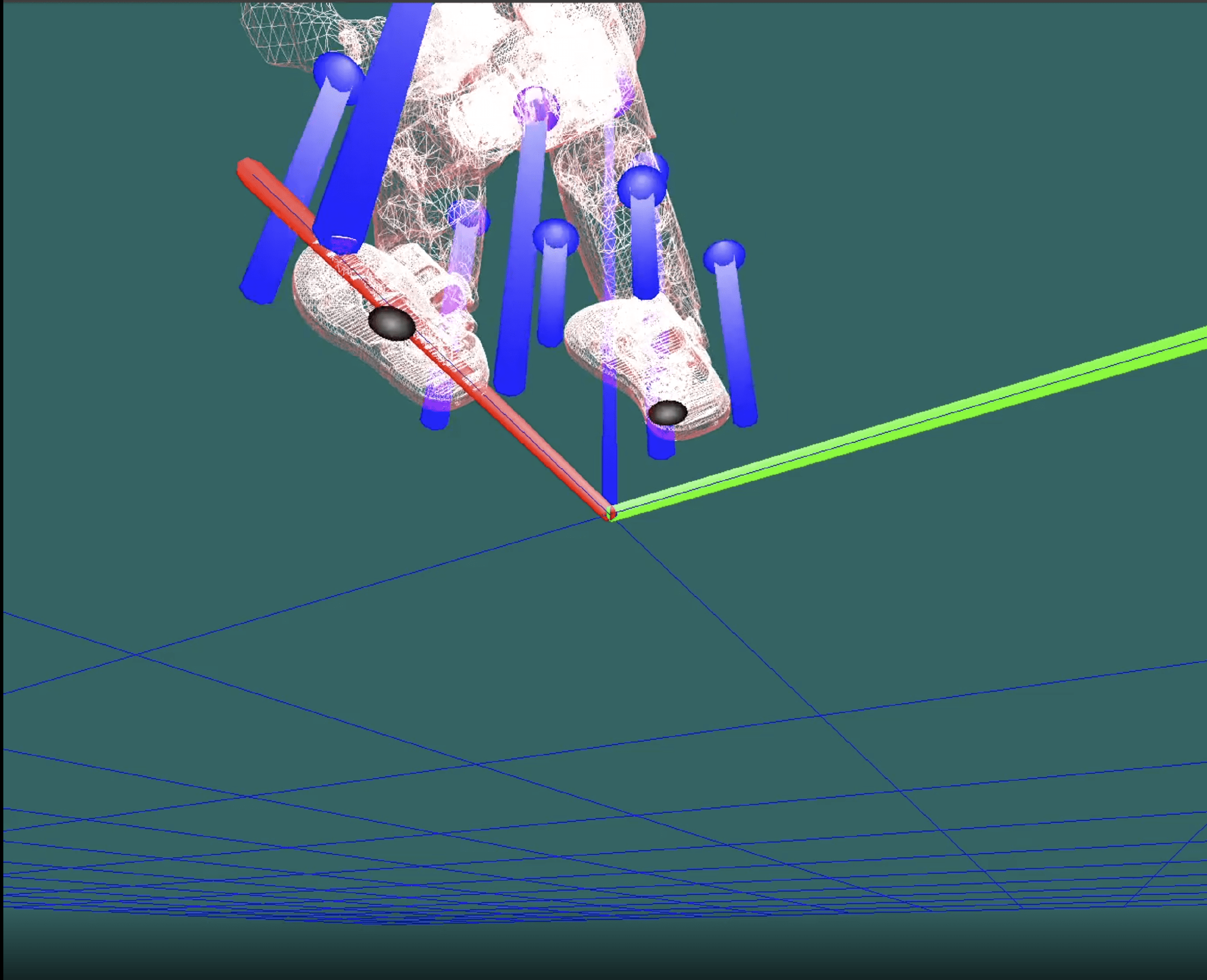}
	\end{subfigure}
\begin{subfigure}{0.22\textwidth}
		\centering
\includegraphics[scale=0.125]{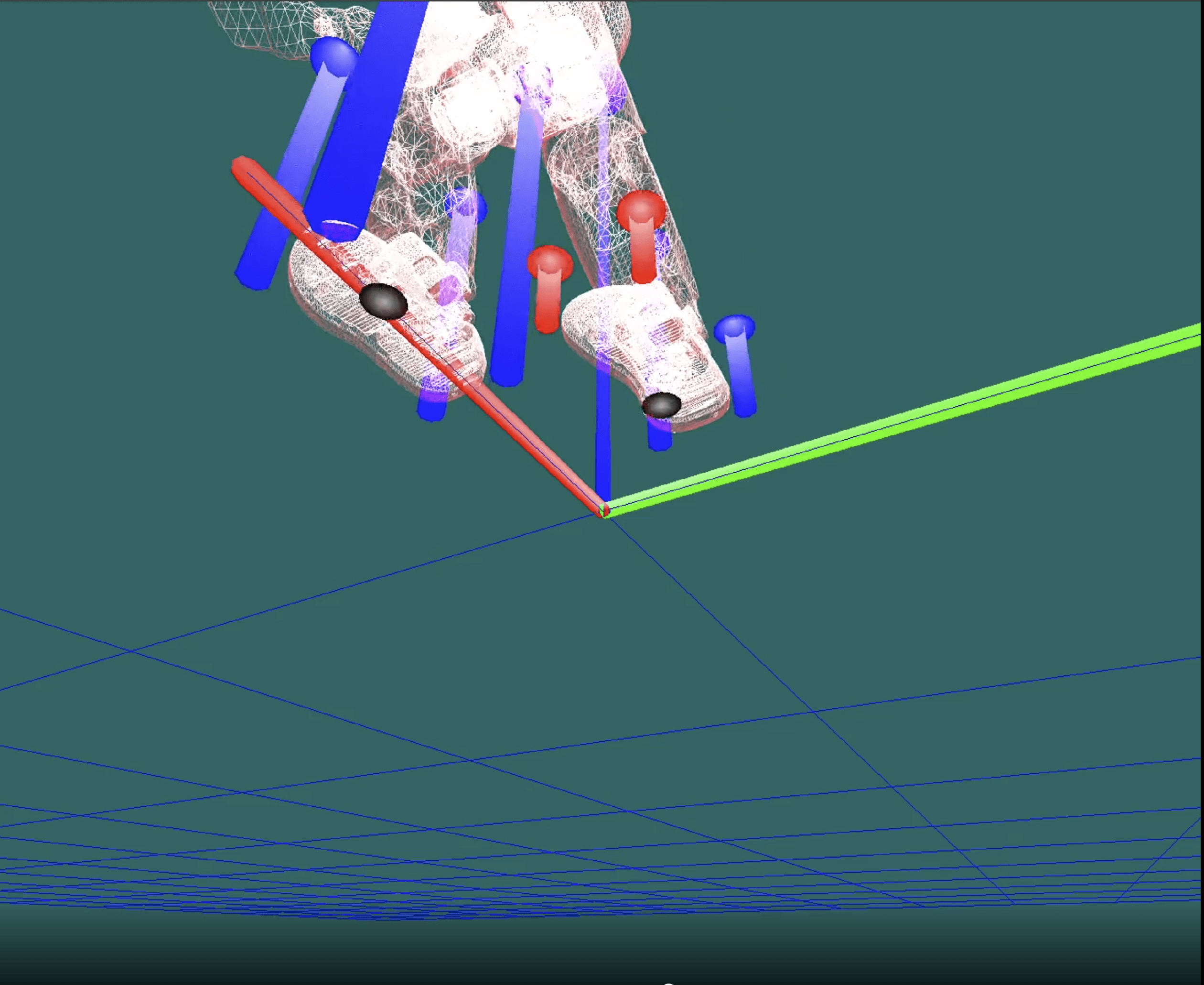}
	\end{subfigure}	
\caption{Center of Pressure (CoP) and contact normal forces evolution for robot walking experiment. CoP shown as black dot evolves within the support polygon affecting the normal forces at the vertices of the foot. Blue arrows depict forces of vertices in contact, while red arrows depict forces of vertices not considered to be in contact. Red dots denote the loss of contact.}
\label{fig:chap:hbe:robot-cop-evol}
\end{figure}

The contact wrench decomposition into normal forces at the vertices is shown in Figure \ref{fig:chap:hbe:robot-foot} for the left and the right foot.
It can be seen that the sum of the vertex forces is equivalent to the overall measured contact normal force. 
Figure \ref{fig:chap:hbe:robot-cop-evol} shows the evolution of the center of pressure of the feet in contact with the environment as the robot walks forward.
When the robot is stationary and standing on both its feet, the center of pressure depicted as a \emph{black} dot lies close to the center of the foot and the magnitude of normal forces at the vertices are higher than the chosen thresholds.
The vertices are inferred to be in contact for such a situation, depicted by the \emph{blue} arrows, where the length of the arrow represents the magnitude of the force.
It can be seen that as the CoP approaches the boundary of the support polygon, which is a rectangle in our case, the normal forces at the vertices are reduced, and depending on the chosen threshold values, these vertices are inferred to have lost contact with the environment, shown as \emph{red} arrows.
As the foot enters the stance phase, the contact wrenches tend to zero and the CoP exits the support polygon and the vertex normal forces are chosen as zero. 
To account for noise in the contact wrenches, a zero threshold is also chosen to infer that the CoP is not within the support polygon.

\begin{figure}[!h]
	\begin{subfigure}{\textwidth}
		\centering
\includegraphics[scale=0.2, width=\textwidth]{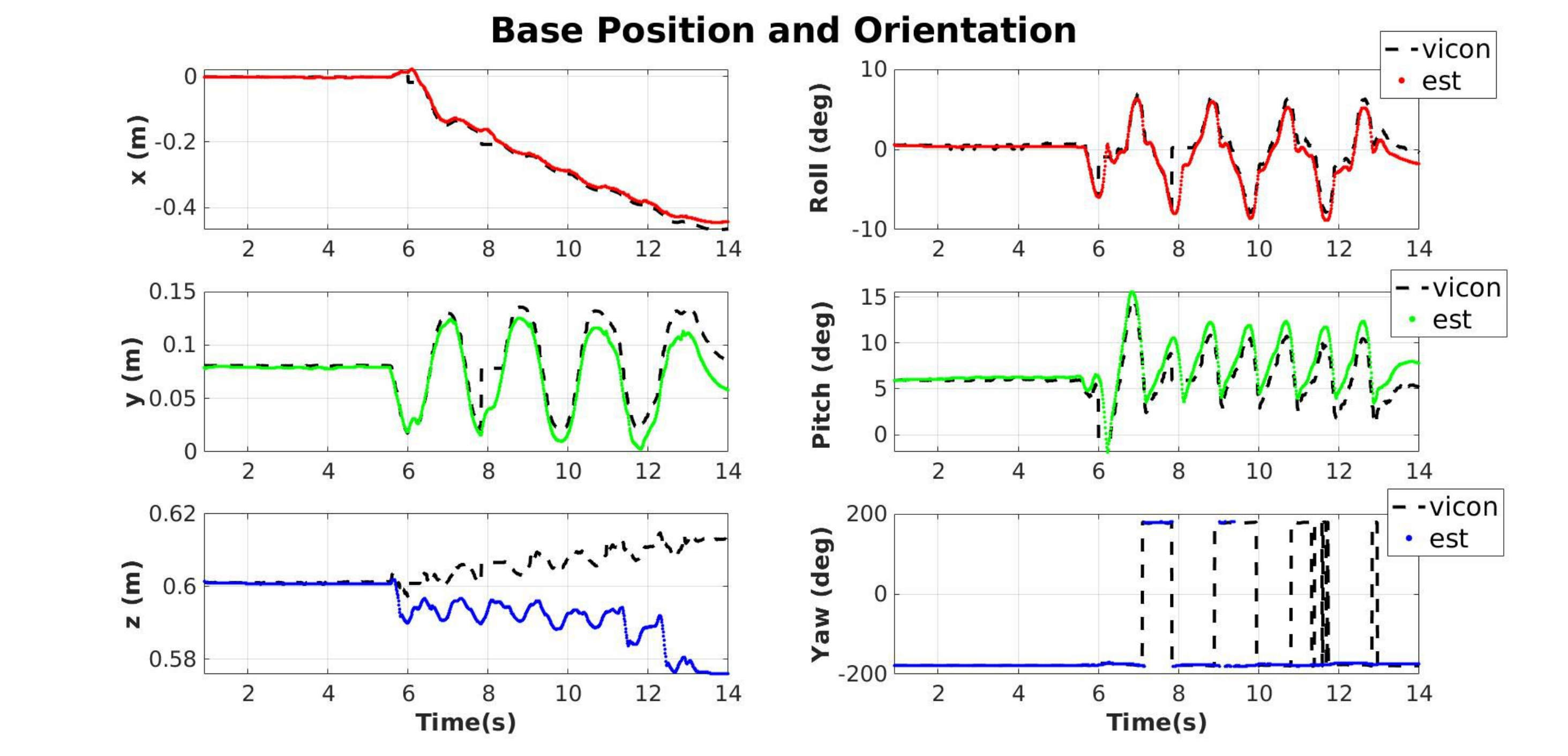}
	\end{subfigure}
\begin{subfigure}{\textwidth}
		\centering
\includegraphics[scale=0.2, width=\textwidth]{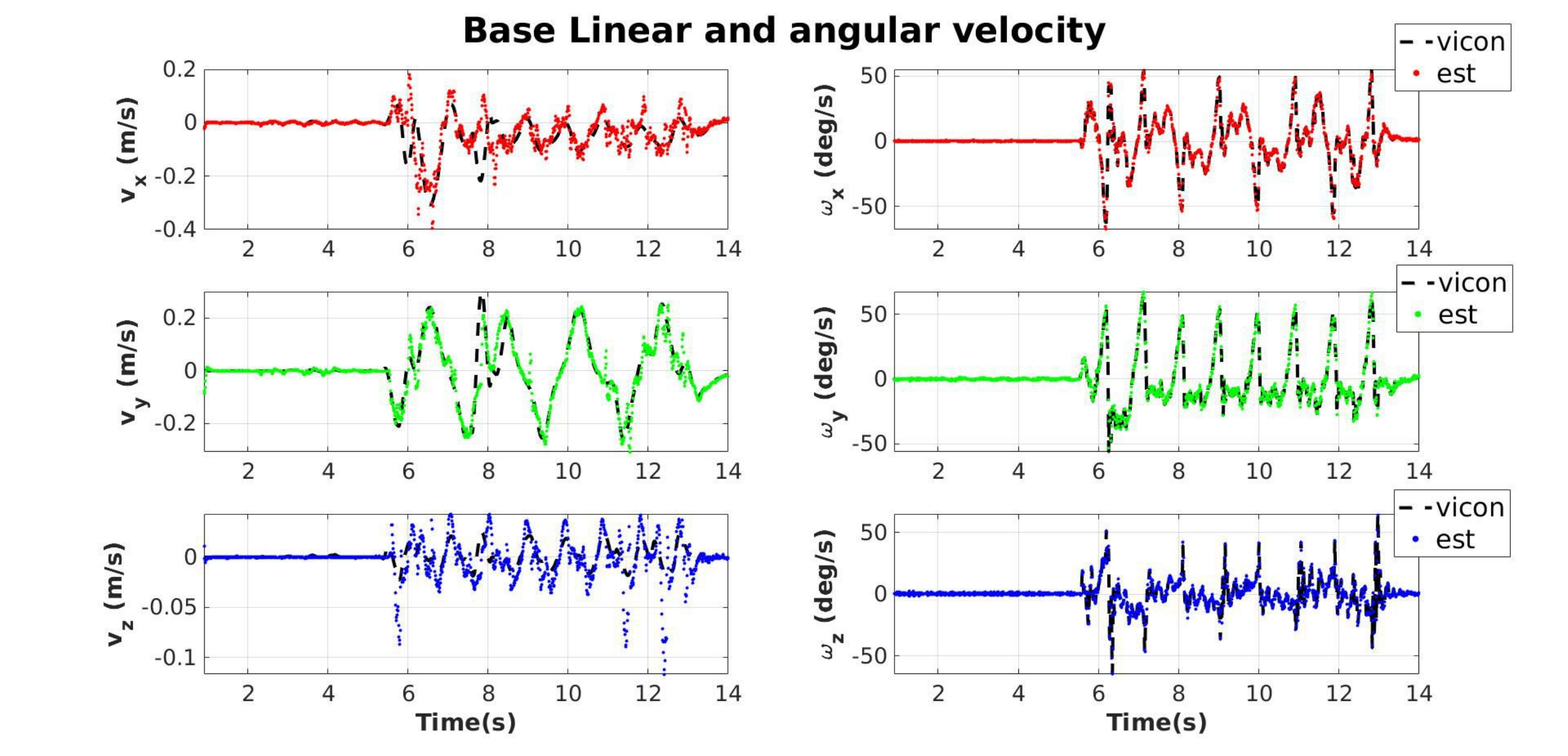}
	\end{subfigure}	
\caption{Base pose and velocity estimates for robot walking experiment using the outputs from contact detection and dynamical inverse kinematics block.}
\label{fig:chap:hbe:robot-base-state}
\end{figure}

Figure \ref{fig:chap:hbe:robot-base-state} shows the evolution of the base state.
It can be seen that the estimated states closely track the ground truth reference trajectory.
The linear velocity estimates suffer from peaks at intermittent time instants which is a result of high norm joint velocities obtained through the pseudo-inverse.
These peaks can be considerably reduced by considering alternative methods for solving the inverse differential kinematics.
Due to the imposition of the terrain height updates, the position estimate in the z-direction retraces itself to the same value every time a contact is made.
This is because we choose a constant value for the height and never update it.
The position estimates in the z-direction could be improved through a dynamic updating of the terrain height either through the average height of all the contact points or by an active perception of the terrain. 
The slight drift in the base pitch angle towards the end of the experiment is highly related to the drifts in the joint angles along the torso and hip joints.
The absolute trajectory error and the relative pose error in the right-invariant sense are shown in Table \ref{table:hbe:walking-base-errors} for the $0.5 \si{\meter}$ walking experiment.
It can be seen that the values corresponding to the rotation, position, and velocity errors show a comparable evaluation in comparison with the results presented in Chapter \ref{chap:floating-base-diligent-kio}.\looseness=-1

\begin{table}[ht]
\centering 
\begin{tabular}[width=\textwidth]{c c c c c}
\hline
\multicolumn{5}{c}{Walking $0.5$ \si{\meter} } \\
 												\hline
\multicolumn{3}{c}{ATE} & \multicolumn{2}{c}{RPE}   \\
\hline
rot [$\si{\degree}$] & pos [$\si{\meter}$] & vel [$\si{\meter/\second}$] & rot [$\si{\degree}$] & pos [$\si{\meter}$] \\
\hline
3.1215 & 0.033 & 0.0714 & 2.6958 & 0.0228 \\
\hline
\end{tabular}
\caption{Right invariant absolute trajectory error and relative pose error for the walking experiment using the dynamical inverse kinematics and EKF base estimator. \looseness=-1}
\label{table:hbe:walking-base-errors} 
\end{table}
\subsection{Human Experiments}

In this section, we demonstrate the application of the proposed method for human motion estimation during in-place walking, squatting, and in-place swinging experiments.
\emph{Subject 08}, from the list of subjects whose URDF models are available in the \href{https://github.com/robotology/human-gazebo}{\textbf{robotology/human-gazebo}} project repository, is considered as the test subject.
We use a reduced DoF model consisting of $46$ DoFs.
While the actual reduced DoF model consists of $48$ DoFs, we do not consider the ball joints on the foot that connects the foot link and the toe link (see Section \ref{sec:chap03:modeling-human}). 
Instead, we consider a single rigid body for the complete foot without the articulation for the toe.


The subject is equipped with the XSens motion capture suit with measurements streaming from 17 IMUs attached on several the links of the body.
During the experiments, due to the unavailability of the sensorized shoes, AMTI force plates are used for measuring the six-axis ground reaction wrenches.
Since an accurate positioning of the subject's feet on the AMTI force plate is not available, instead of transforming the wrenches measured in the AMTI coordinate system into the foot coordinate system, we consider the entirety of each AMTI force plate as each foot sole of the subject.
This allows for a more reliable measurement of the Center of Pressure quantity than that obtained from an inaccurately transformed wrench.

We do not have a ground-truth reference system for evaluating the performance of the proposed method quantitatively.
Thus, in the following sub-sections, we only demonstrate a qualitative evaluation and discussion by visualizing the reconstructed human motion. 
We also present a comparison of the estimates obtained from the proposed method with those measured by the XSens motion capture suit.
However, it has been observed through experience that the base pose measurements obtained from the XSens suit also suffer from slipping and sliding issues from time to time.

The Schmitt trigger is tuned with contact make and break thresholds as $65$ N and $45$ N respectively with stable switching time parameters as $0.02$ seconds for both making and breaking contacts at each vertex.
The base estimator uses noise parameters similar to those listed in Table \ref{table:hbe:noise-parameters}.
The base-estimator is enabled with terrain height updates and contact plane orientation updates.
The contact plane orientation is imposed by setting the orientation of the foot with respect to the inertial frame as identity, every time all the four vertices of the foot are inferred to be in contact with the environment. \looseness=-1

\subsubsection{In-place walking}
\begin{figure}[!h]
\centering
\begin{subfigure}{\textwidth}
\centering
\includegraphics[scale=0.45]{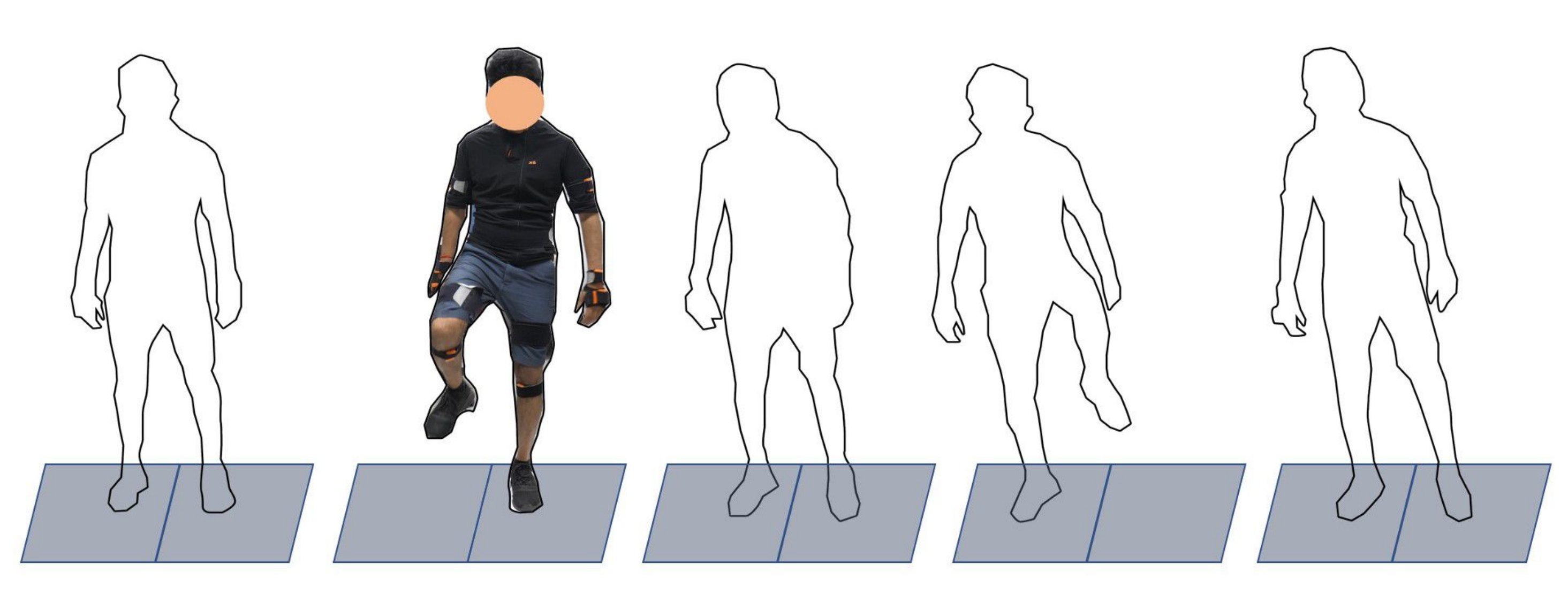}
\end{subfigure}
\begin{subfigure}{0.22\textwidth}
\includegraphics[scale=0.3]{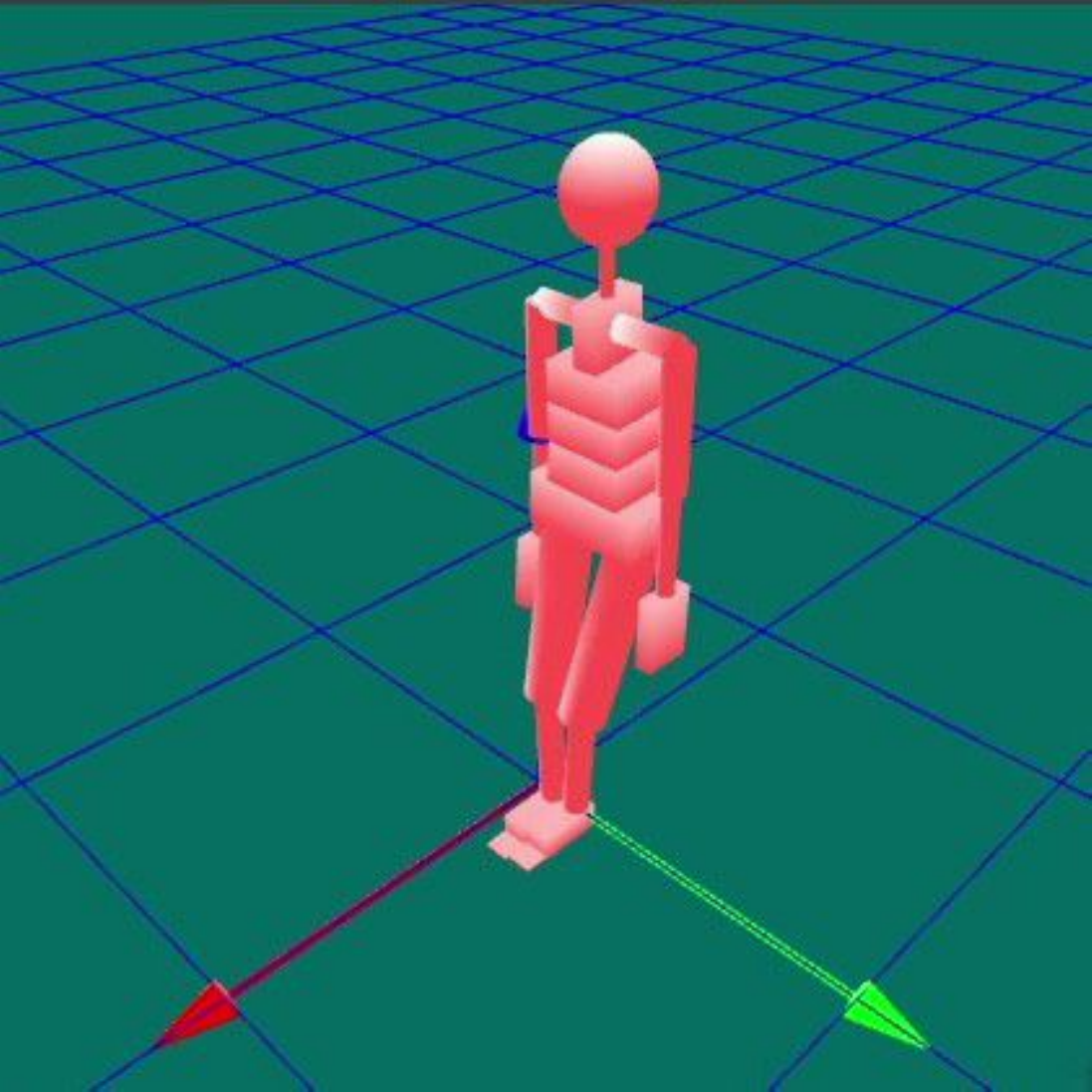}
\end{subfigure}
\begin{subfigure}{0.22\textwidth}
\includegraphics[scale=0.3]{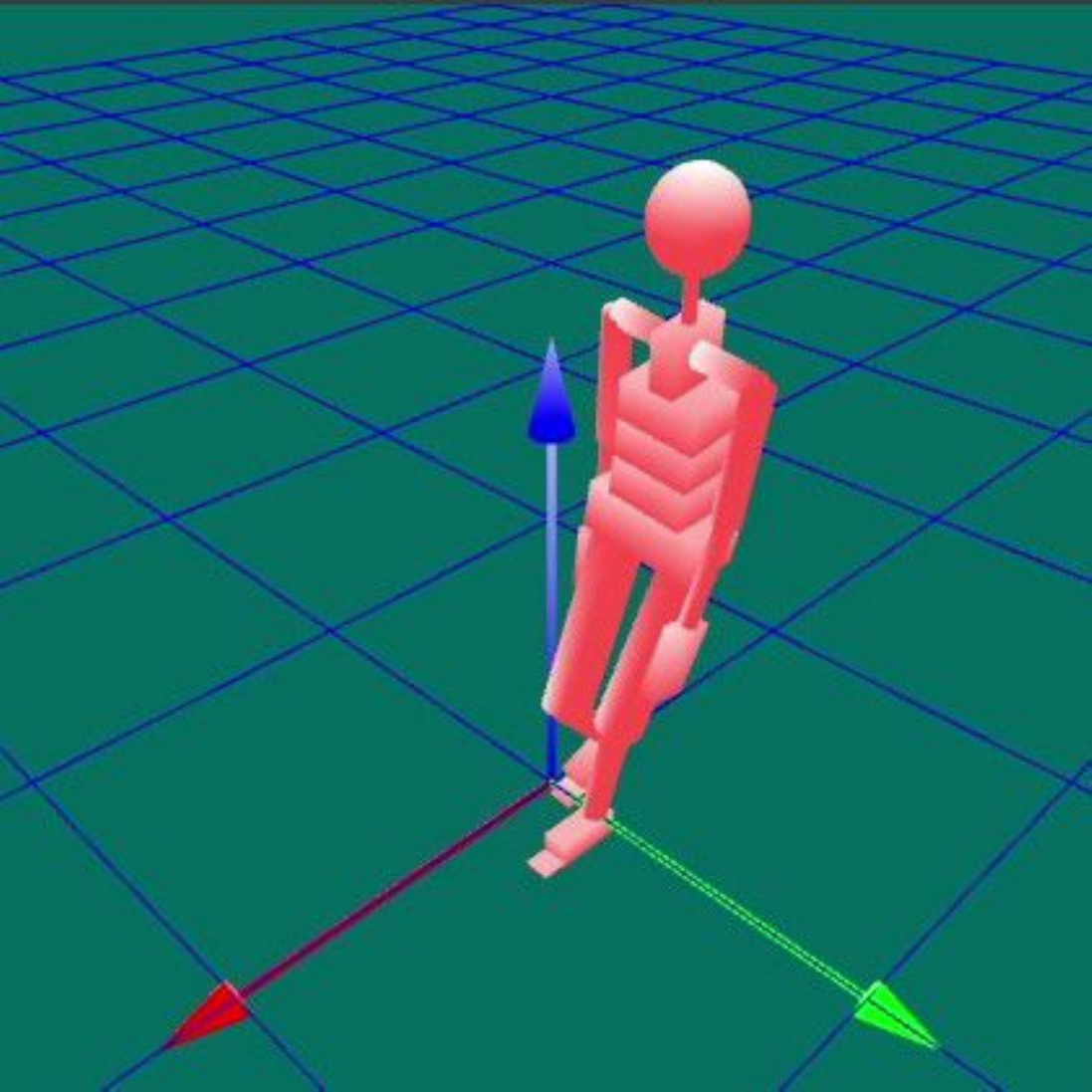}
\end{subfigure}
\begin{subfigure}{0.22\textwidth}
\includegraphics[scale=0.3]{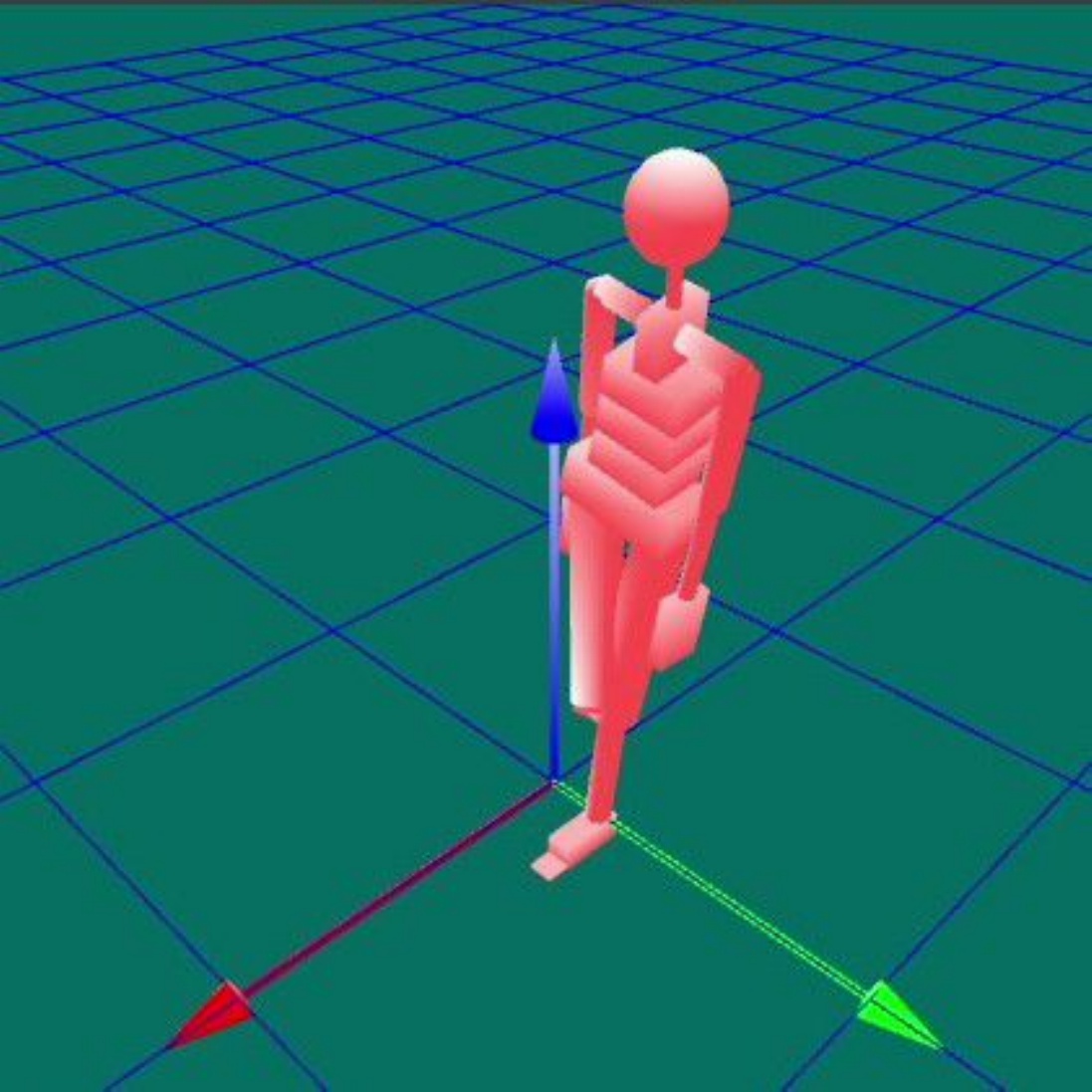}
\end{subfigure}
\begin{subfigure}{0.22\textwidth}
\includegraphics[scale=0.3]{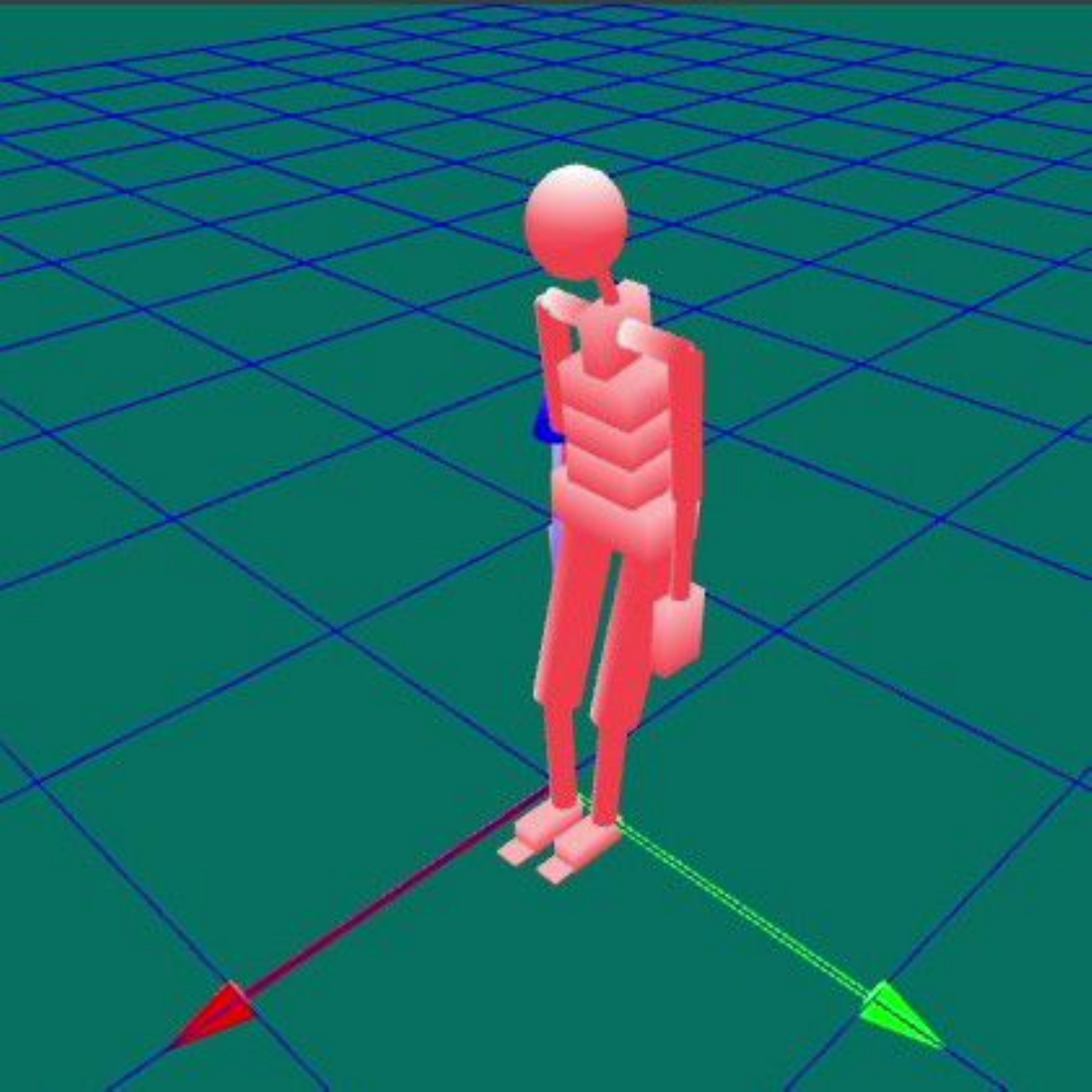}
\end{subfigure}
\begin{subfigure}{0.22\textwidth}
\includegraphics[scale=0.3]{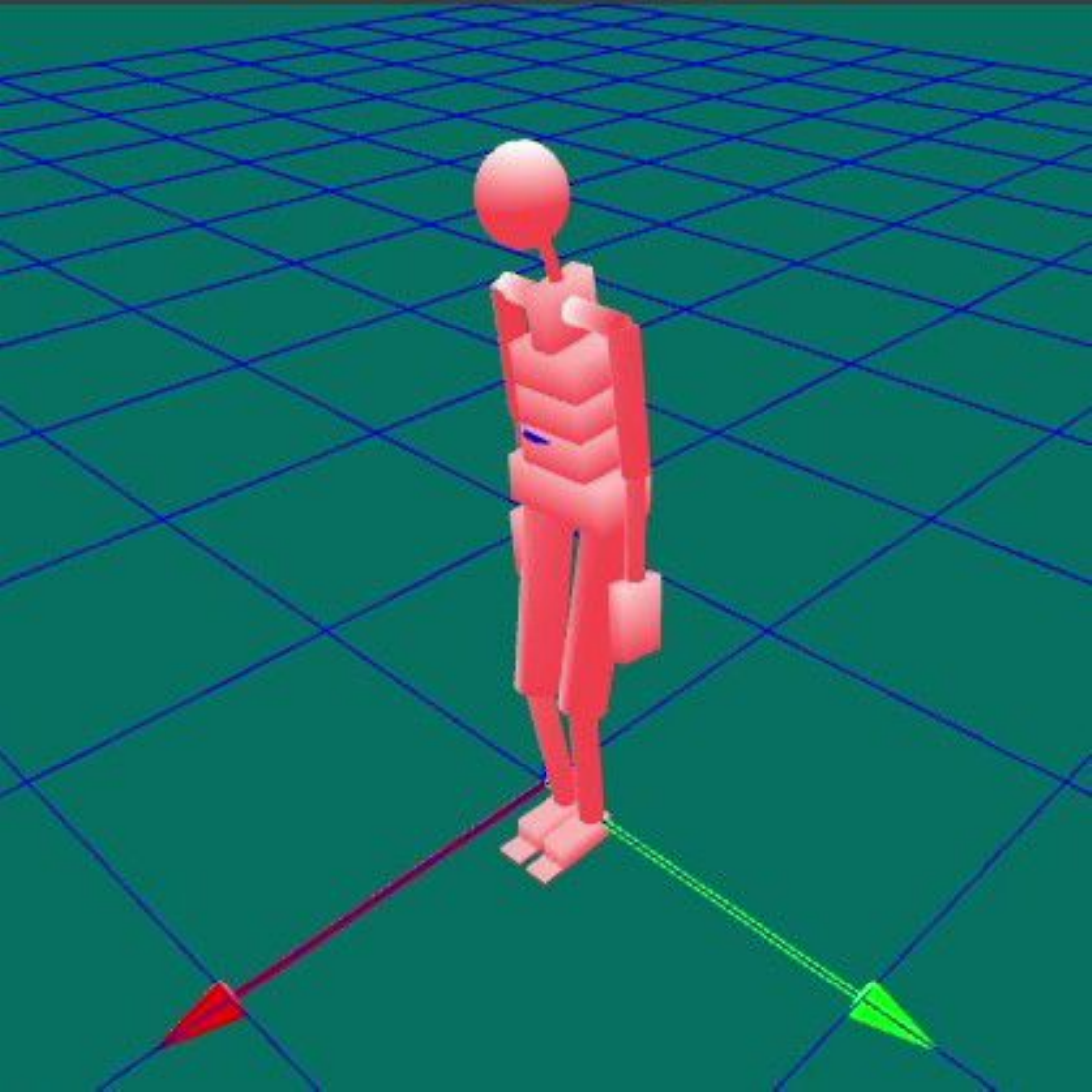}
\end{subfigure}
\begin{subfigure}{0.22\textwidth}
\includegraphics[scale=0.3]{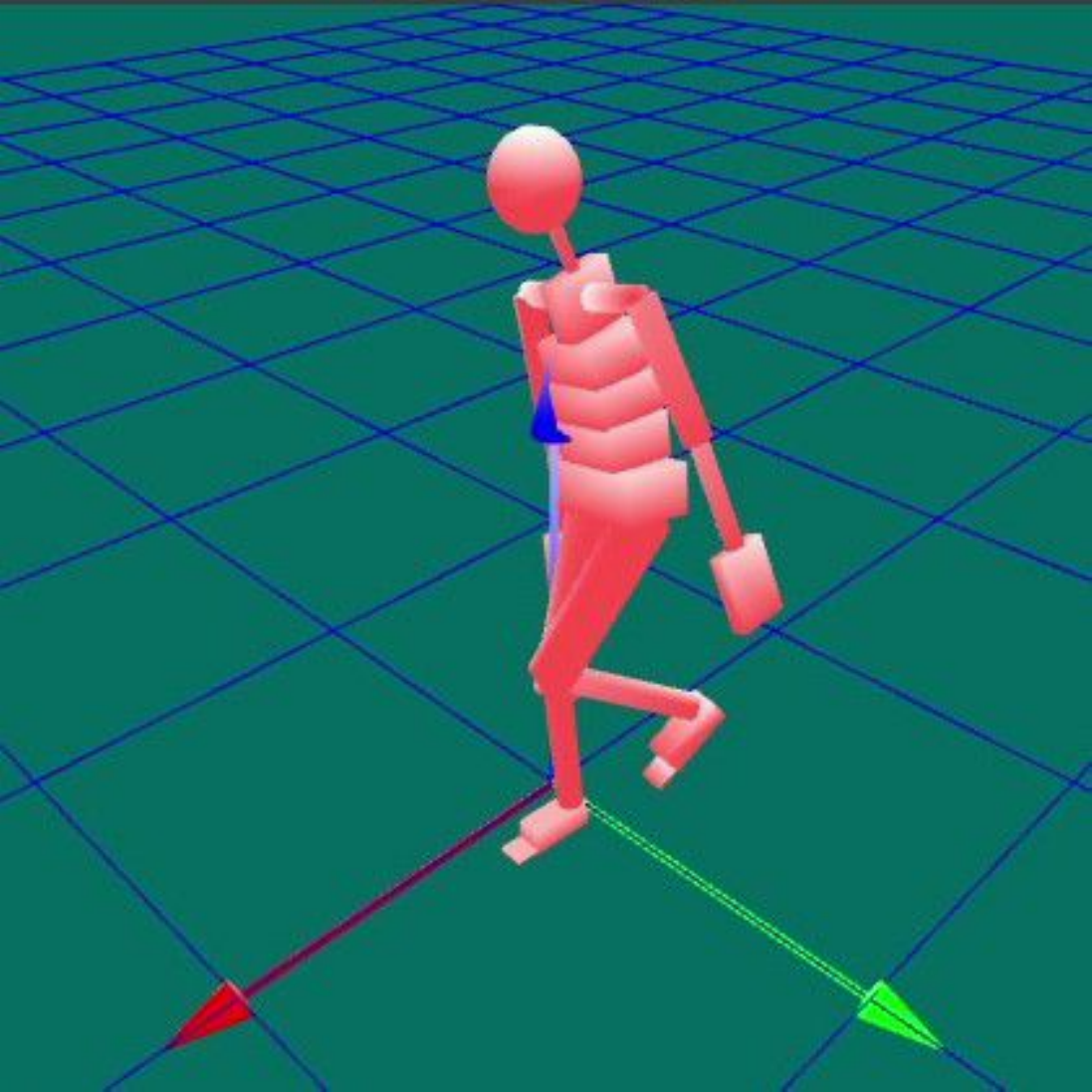}
\end{subfigure}
\begin{subfigure}{0.22\textwidth}
\includegraphics[scale=0.3]{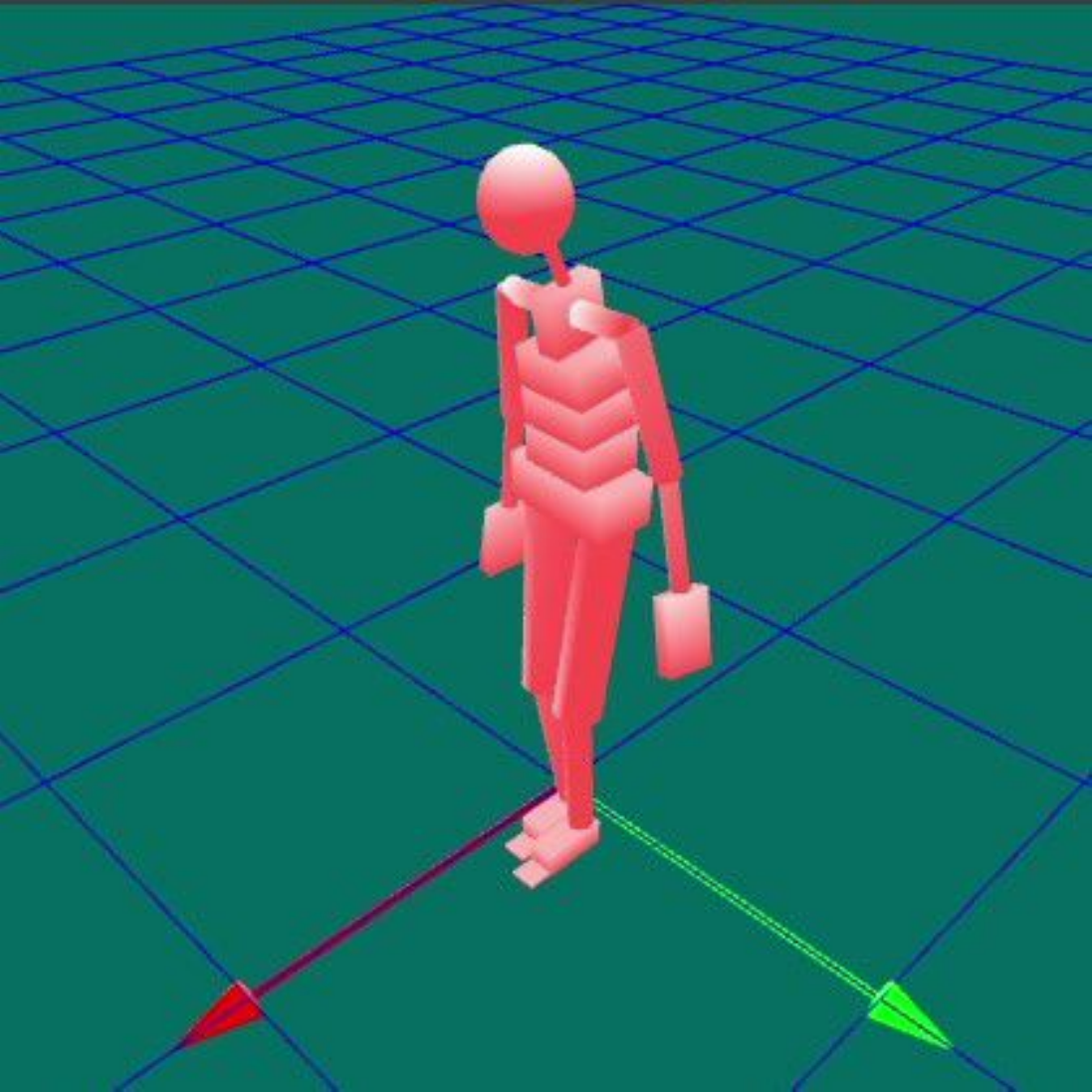}
\end{subfigure}
\begin{subfigure}{0.22\textwidth}
\includegraphics[scale=0.3]{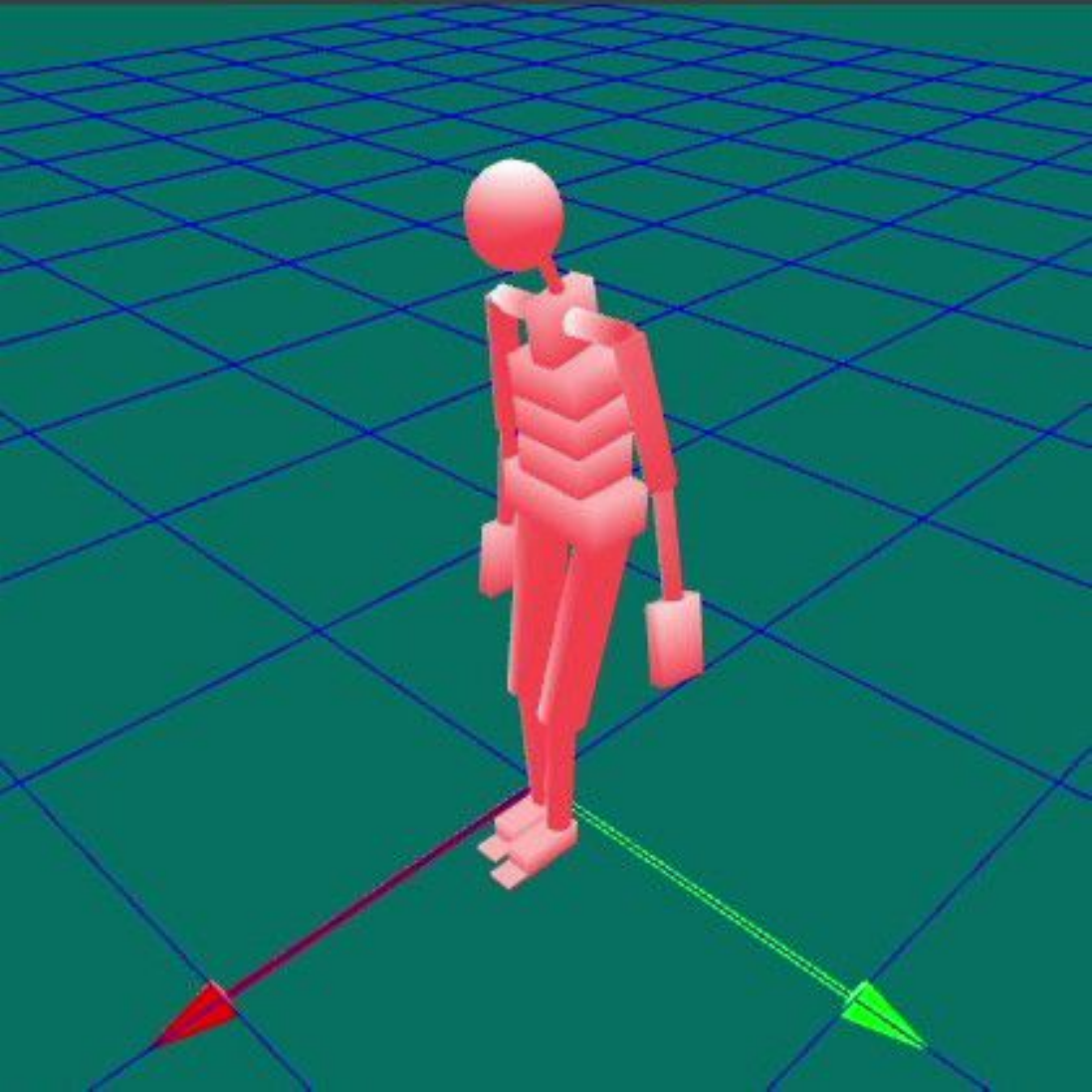}
\end{subfigure}
\caption{Human motion reconstruction for in-place walking experiment.}
\label{fig:chap:hbe:human-in-place-walk-evol}
\end{figure}

In this experiment, the subject initially stands in a stationary configuration with each of their foot placed over each AMTI force plate.
The arms remain close to an N-pose configuration and the subject begins to raise their right knee upward until the right thigh is approximately parallel to the ground surface.
Then, the subject brings down the left leg entering into a double support phase before repeating the same knee raise with their left leg.
These actions are repeated in a manner representing a dynamic in-place walking motion as depicted in Figure \ref{fig:chap:hbe:human-in-place-walk-evol}.
The reconstructed human motion is also depicted in Figure \ref{fig:chap:hbe:human-in-place-walk-evol}.
The human motion seems to be reasonably reconstructed with the joint states recovered from the dynamical inverse kinematics and the base state estimated by the EKF.

However, some irregularities in the motion can be emphasized.
Since joint velocity and position limits are not considered while solving for the joint states, the recovered human motion sometimes does not remain close to the actual subject motion.
The recovered joint configurations sometimes happen to be very close to configurations that might result in self-collision of the links.
Such unfeasible configurations are due to the fact that the joint position limits are not explicitly considered within the inverse differential kinematics for finding the joint velocities. 
The joint limits are considered only at the integrator level for saturating the joint positions integrated from joint velocities.
Hence, the pseudo-inverse based IK solver produces high norm velocities without considering the joint limits that render the overall configuration sometimes physically unfeasible. 
The inverse differential kinematics is solved using regularized, weighted pseudo-inverse resulting in high magnitude joint velocities incurs peaks in the base velocity estimation.

\begin{figure}[!h]
\centering
	\begin{subfigure}{0.9\textwidth}
		\centering
\includegraphics[scale=0.05, width=\textwidth]{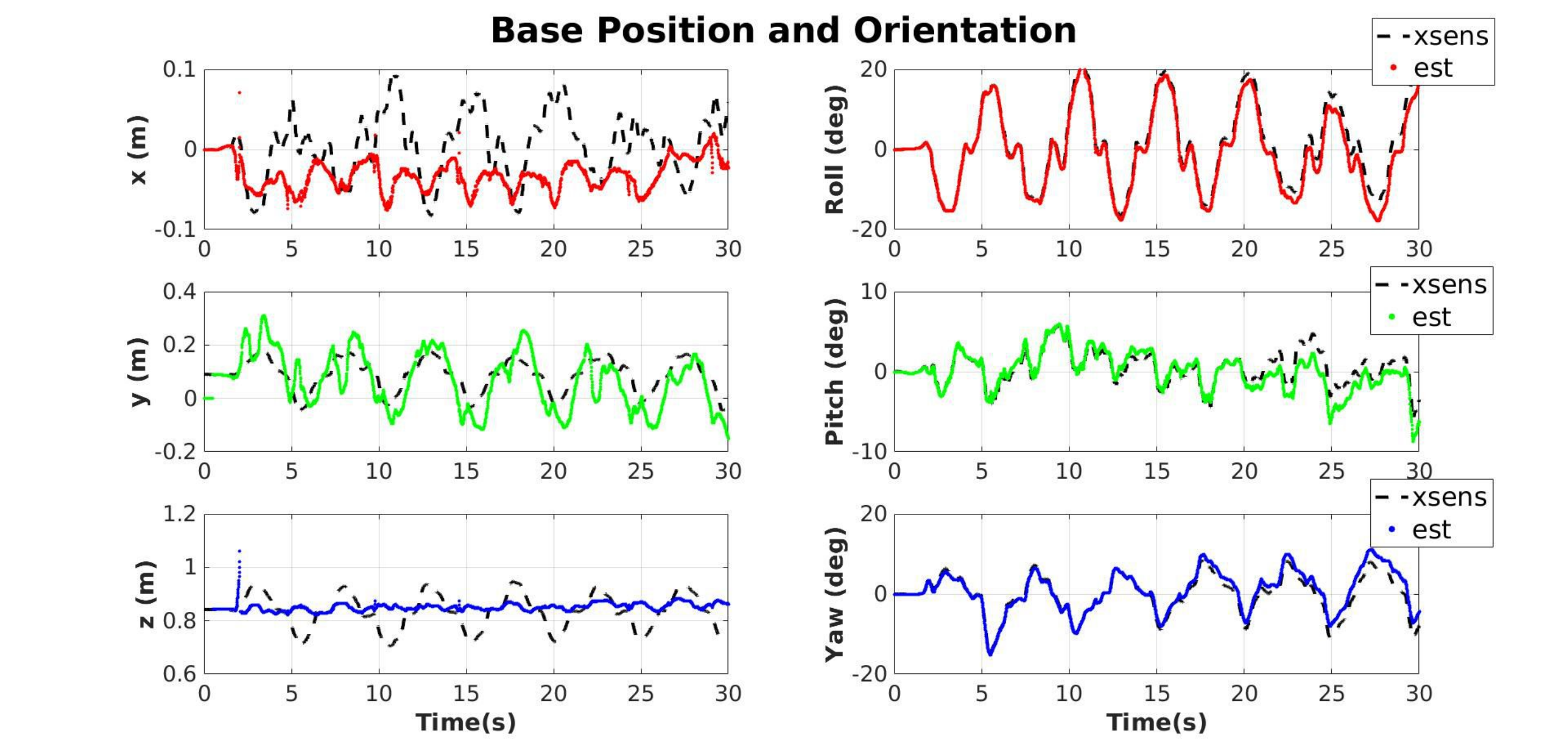}
	\end{subfigure}
\begin{subfigure}{0.9\textwidth}
		\centering
\includegraphics[scale=0.05, width=\textwidth]{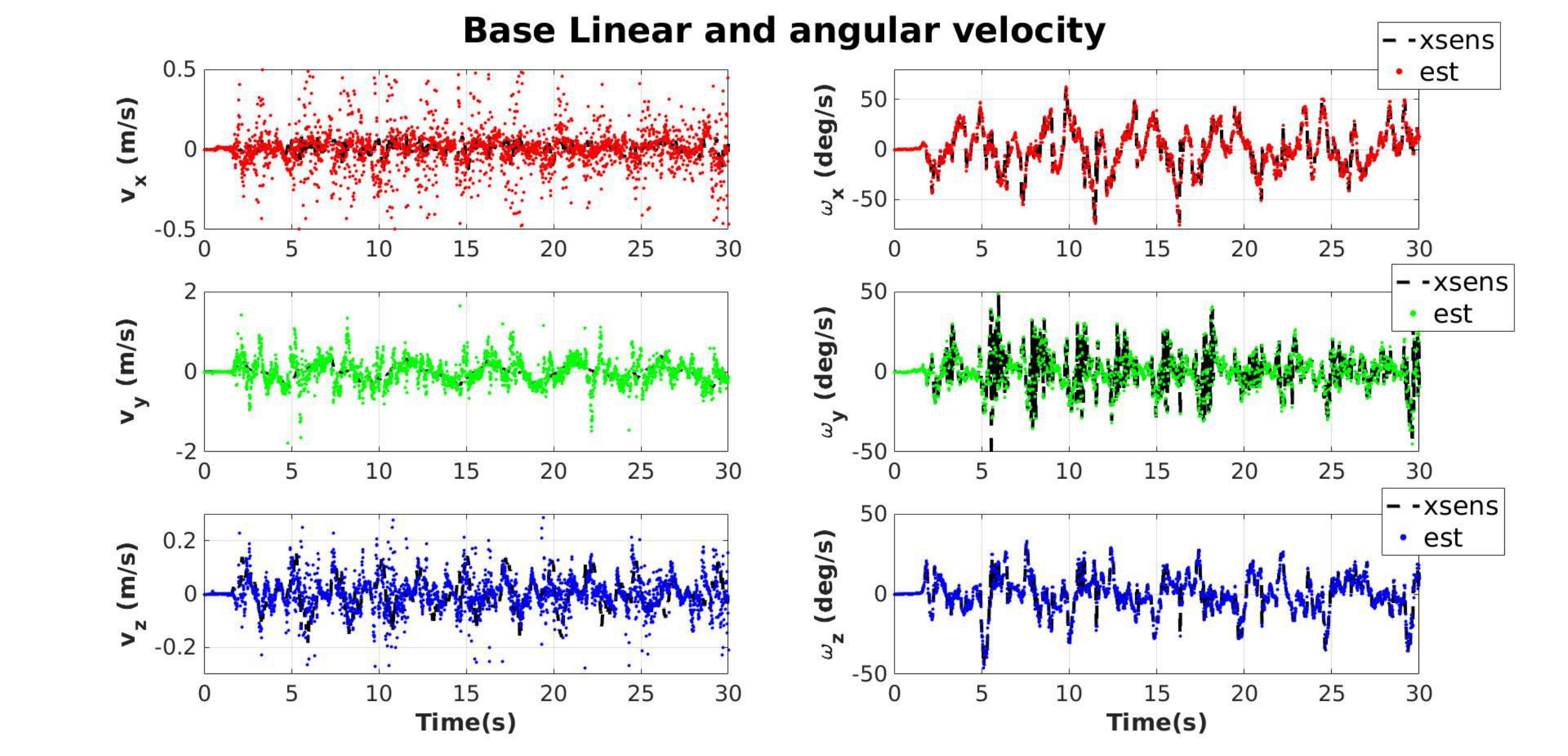}
	\end{subfigure}	
\begin{subfigure}{0.9\textwidth}
		\centering
\includegraphics[scale=0.05, width=\textwidth]{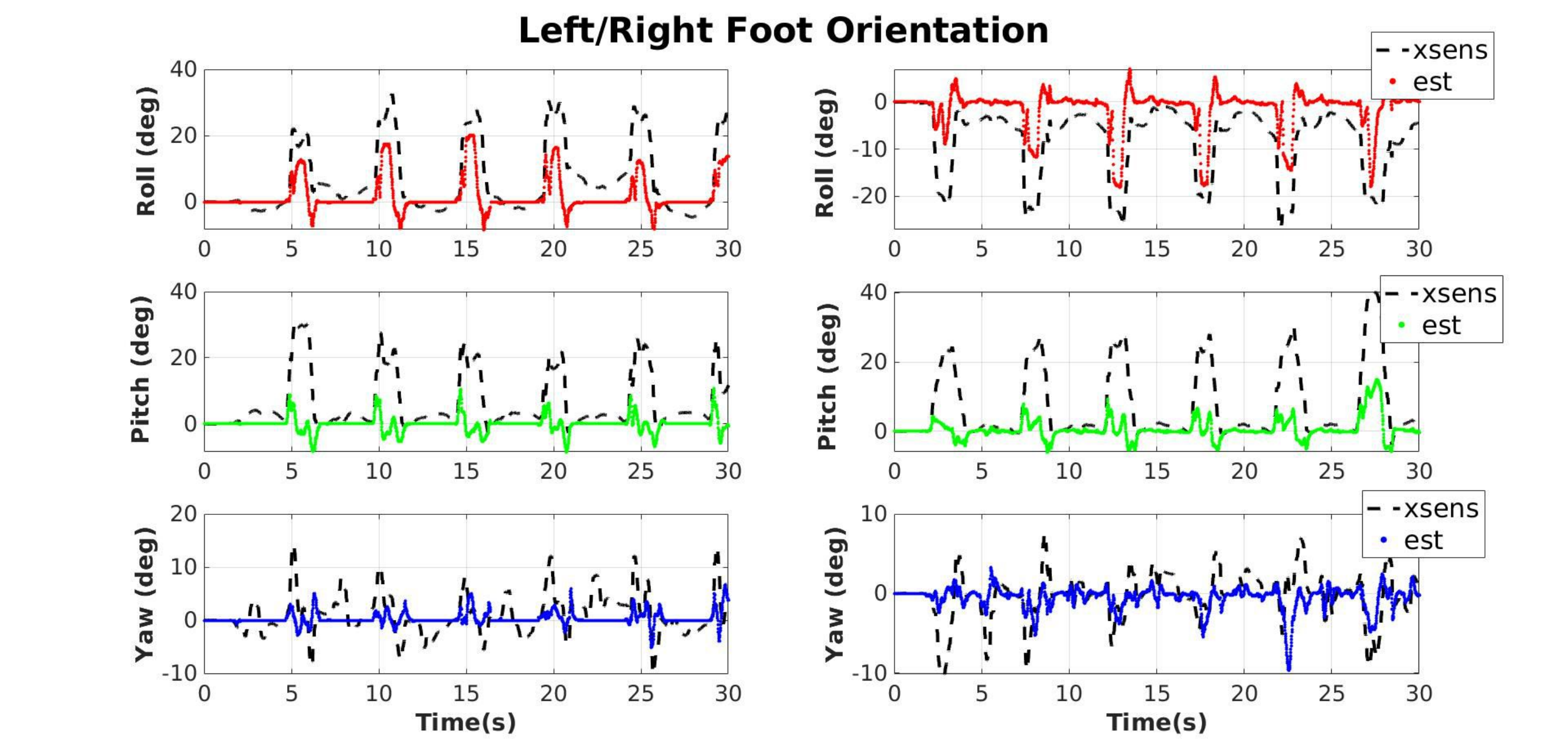}
	\end{subfigure}	
\caption{Base pose and velocity estimates along with feet rotation estimates for human in-place-walking experiment.}
\label{fig:chap:hbe:human-in-place-walk-base-state}
\end{figure}

Figure \ref{fig:chap:hbe:human-in-place-walk-base-state} shows the comparison of the base link states and foot rotations between the proposed method and the XSens motion capture system.
The noisy nature of the recovered joint velocities is reflected in the noisy estimates of base linear velocity. 
However, this is partly also due to the considered constant prediction model in the EKF which does not fully capture the actual evolution of the base velocity.
The feet rotations estimated by the EKF remain suitably close to those measured by the XSens motion capture system with respect to its motion profile.
The small increase in magnitude of the foot orientations is suitably captured during the swing phase of the foot while they are forced to zero during contact, given that the measurement updates a flat contact plane orientation is enforced.
During the in-place walking motion, the center of mass is shifted from one foot to the other and the CoM projected to the contact plane remains within the support polygon of the support foot during single support.
This implies that the base position follows should follow a sinusoidal trajectory along the $\mathbf{y}$-direction while maintaining the base height $\mathbf{z}$ and forward position $\mathbf{x}$ within nominal bounds every time the subject lifts the thigh and makes foot contact with the ground while lifting the other thigh. 
Such a trajectory is clearly observed from the estimated base position in Figure \ref{fig:chap:hbe:human-in-place-walk-base-state}.

\subsubsection{Squatting}
\begin{figure}[!h]
\centering
\begin{subfigure}{\textwidth}
\centering
\includegraphics[scale=0.45]{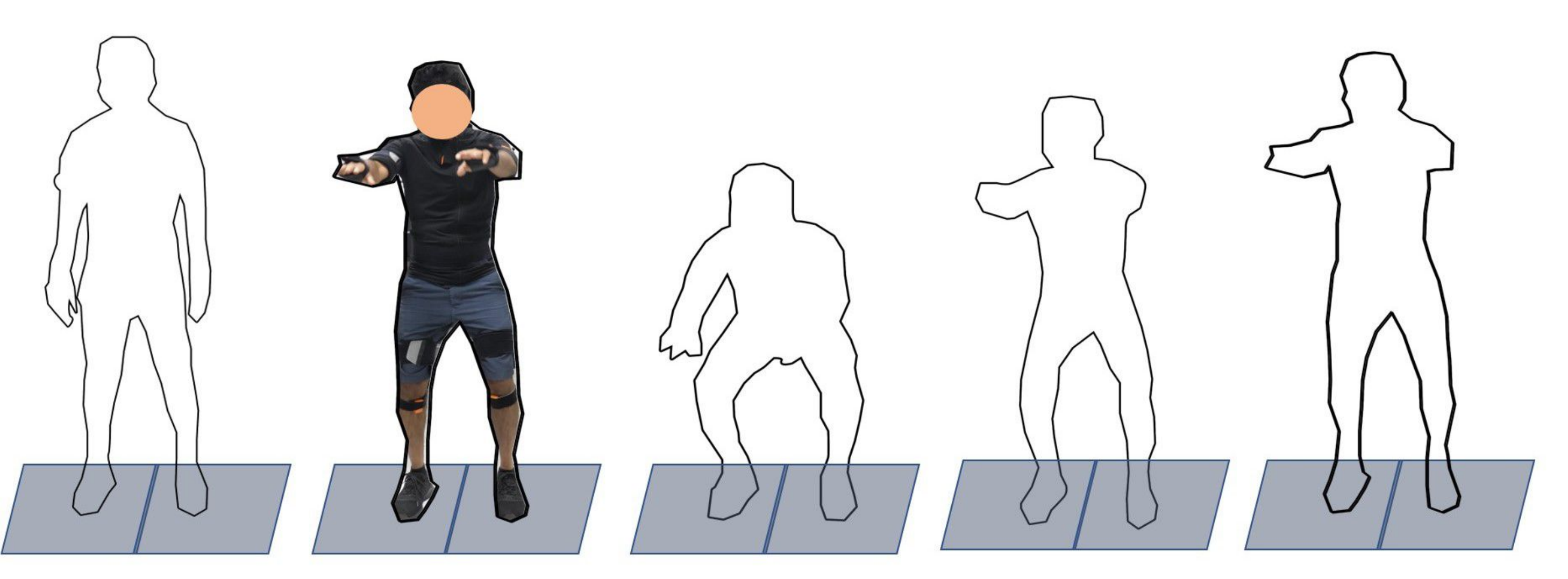}
\end{subfigure}
\begin{subfigure}{0.22\textwidth}
\includegraphics[scale=0.3]{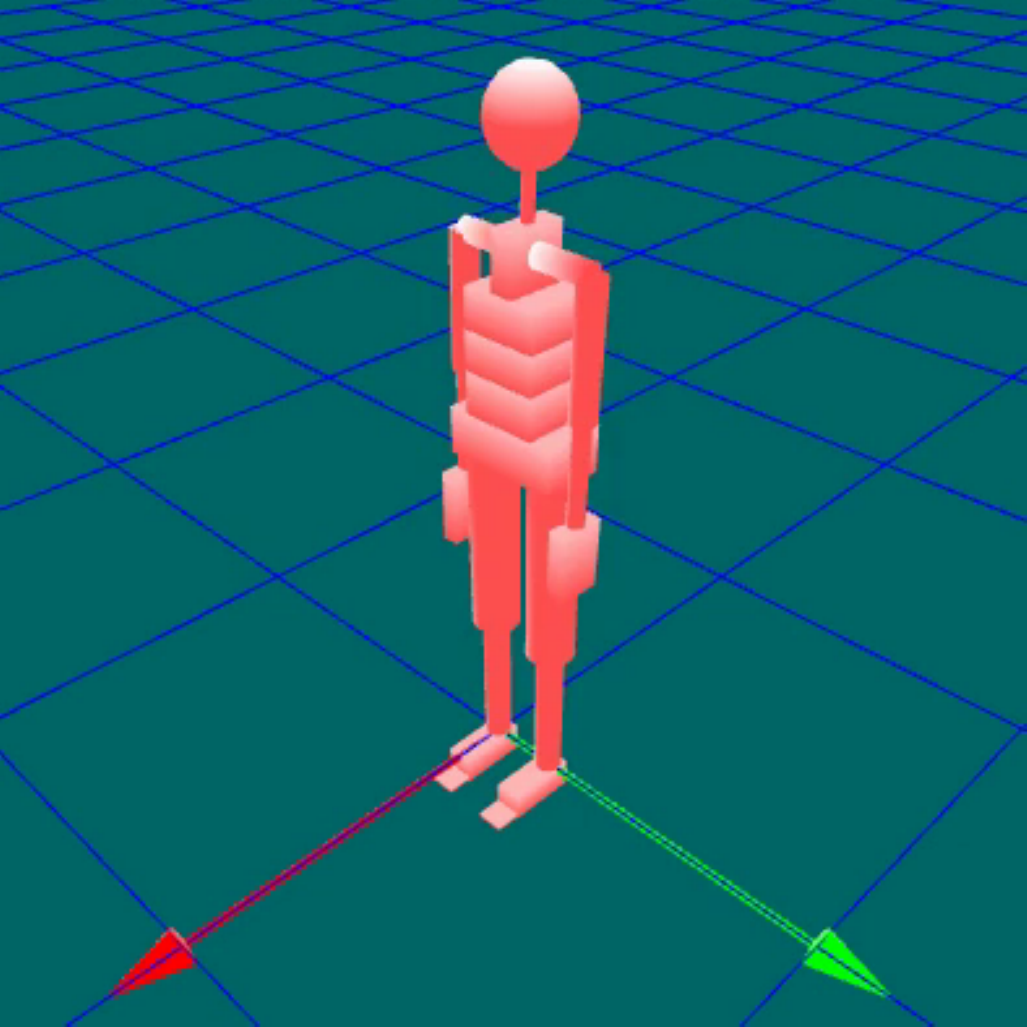}
\end{subfigure}
\begin{subfigure}{0.22\textwidth}
\includegraphics[scale=0.3]{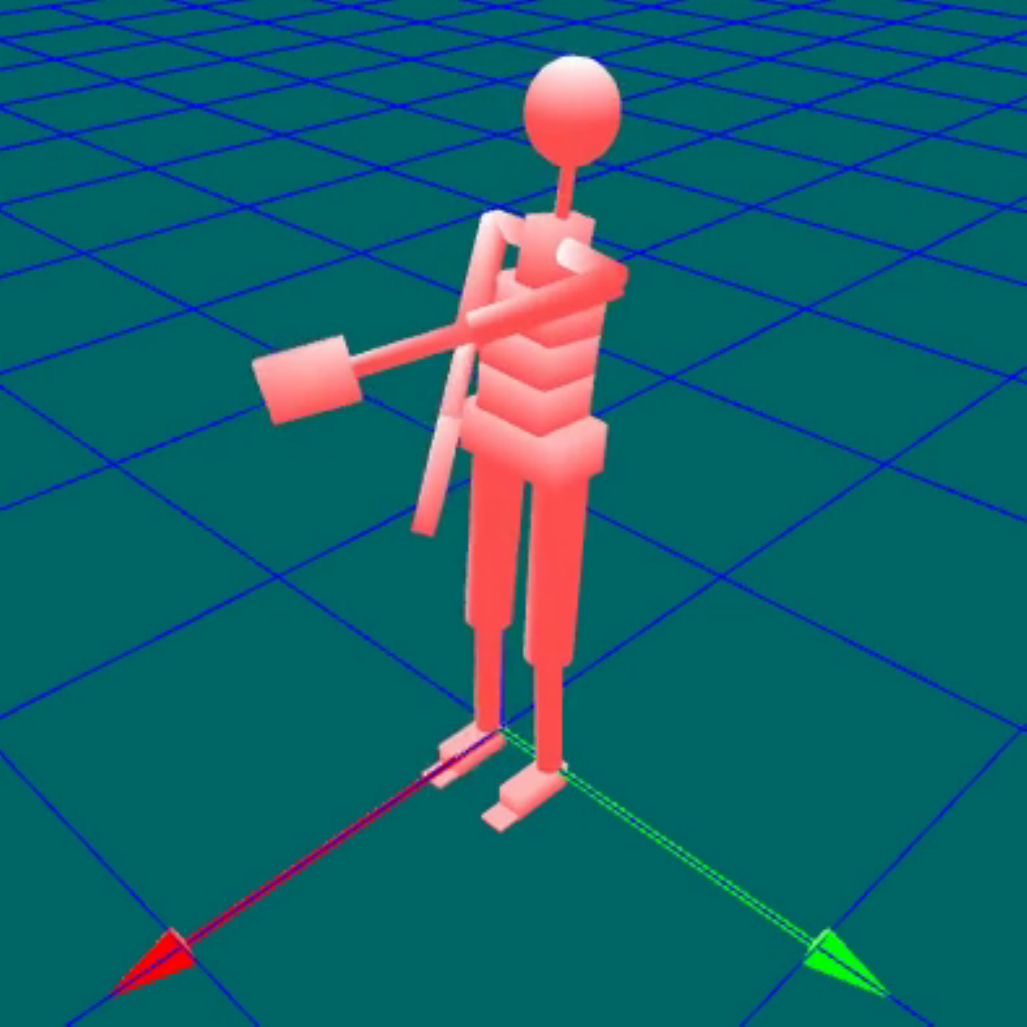}
\end{subfigure}
\begin{subfigure}{0.22\textwidth}
\includegraphics[scale=0.3]{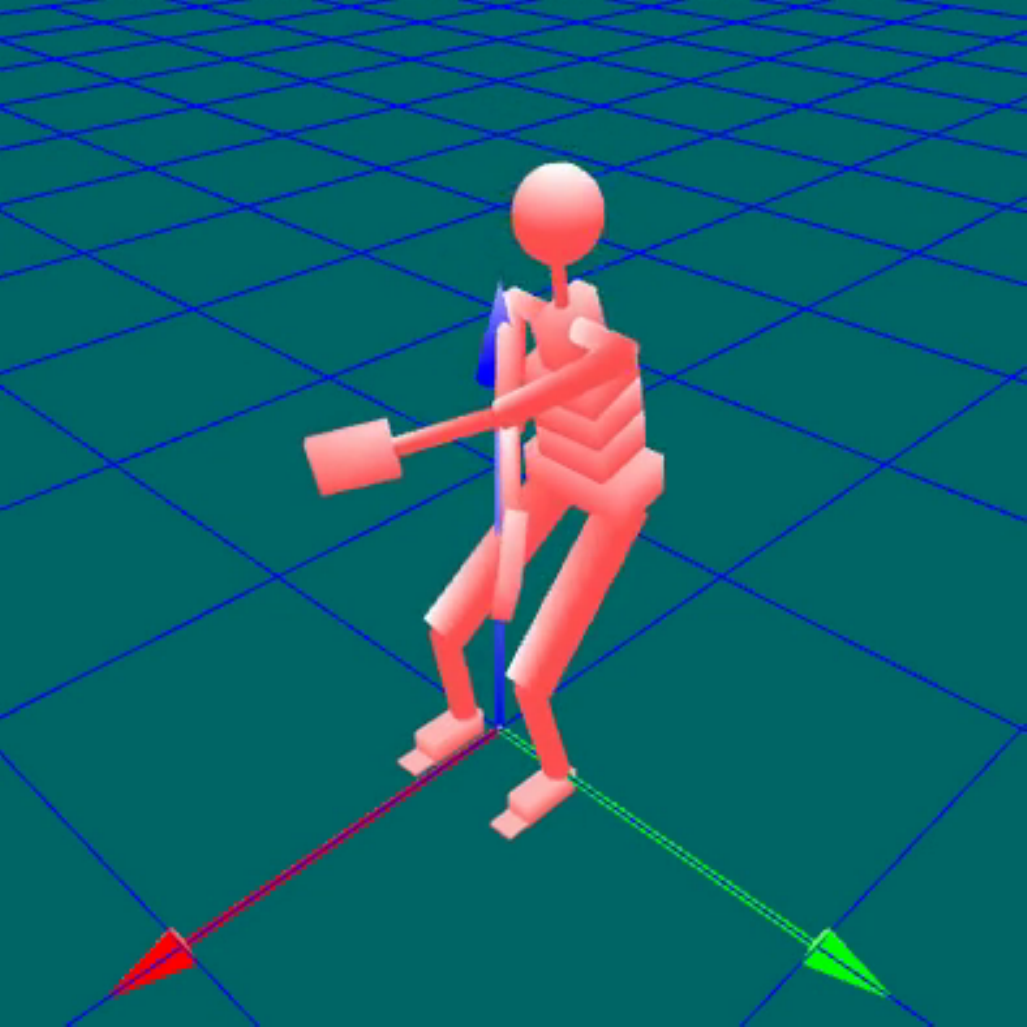}
\end{subfigure}
\begin{subfigure}{0.22\textwidth}
\includegraphics[scale=0.3]{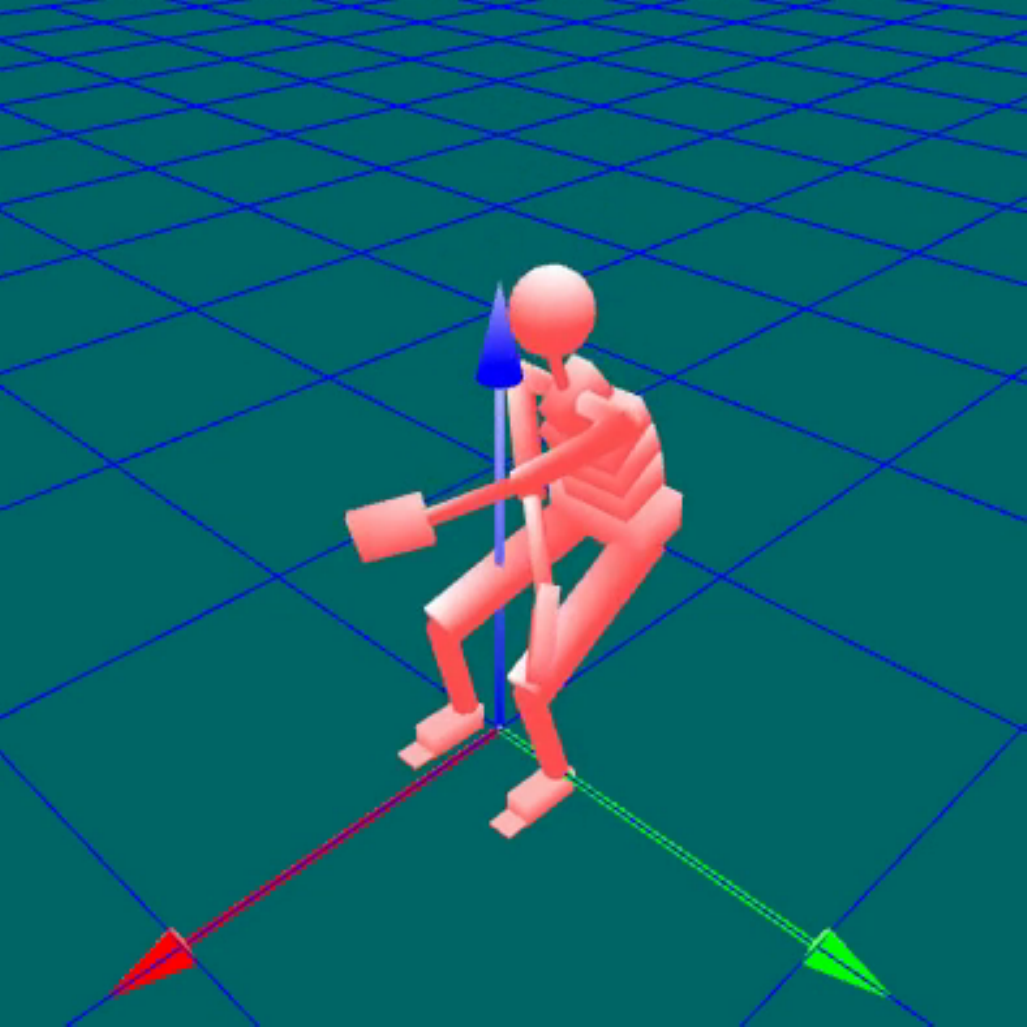}
\end{subfigure}
\begin{subfigure}{0.22\textwidth}
\includegraphics[scale=0.3]{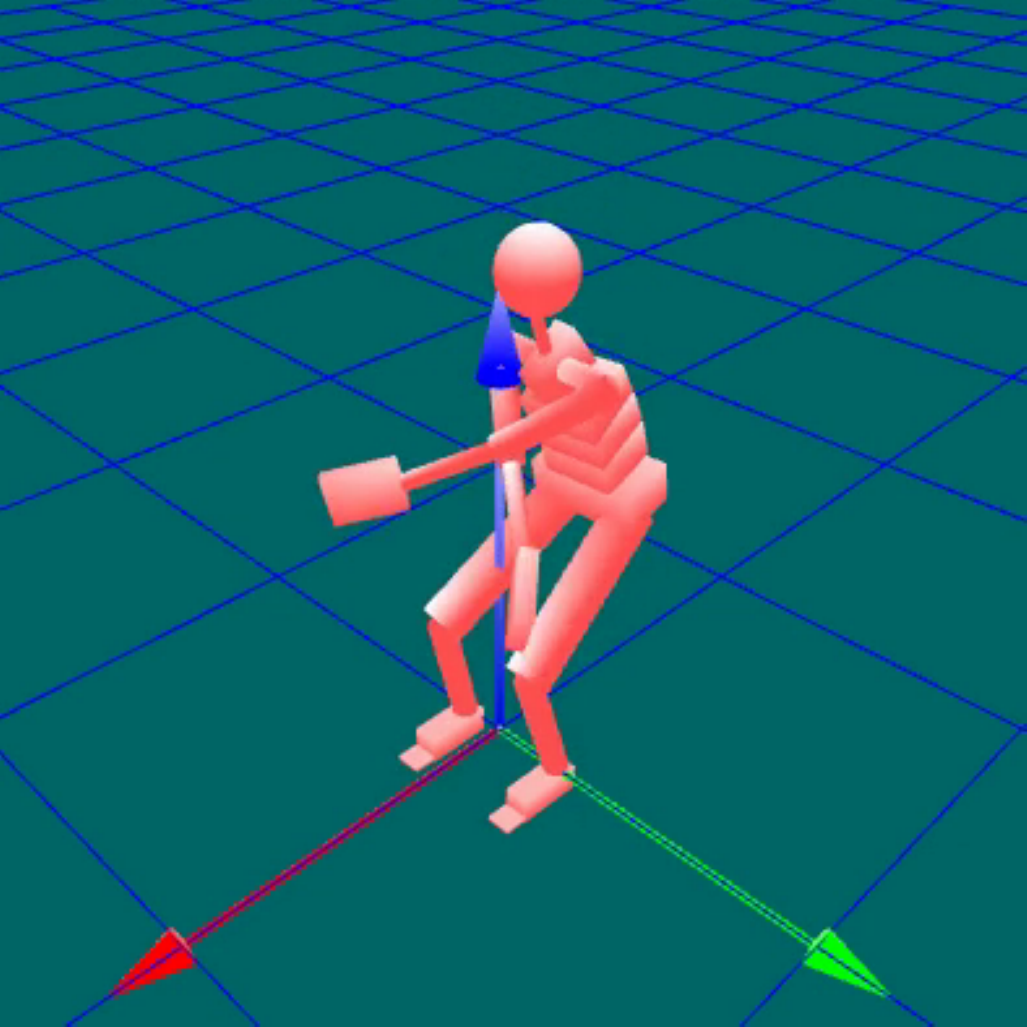}
\end{subfigure}
\begin{subfigure}{0.22\textwidth}
\includegraphics[scale=0.3]{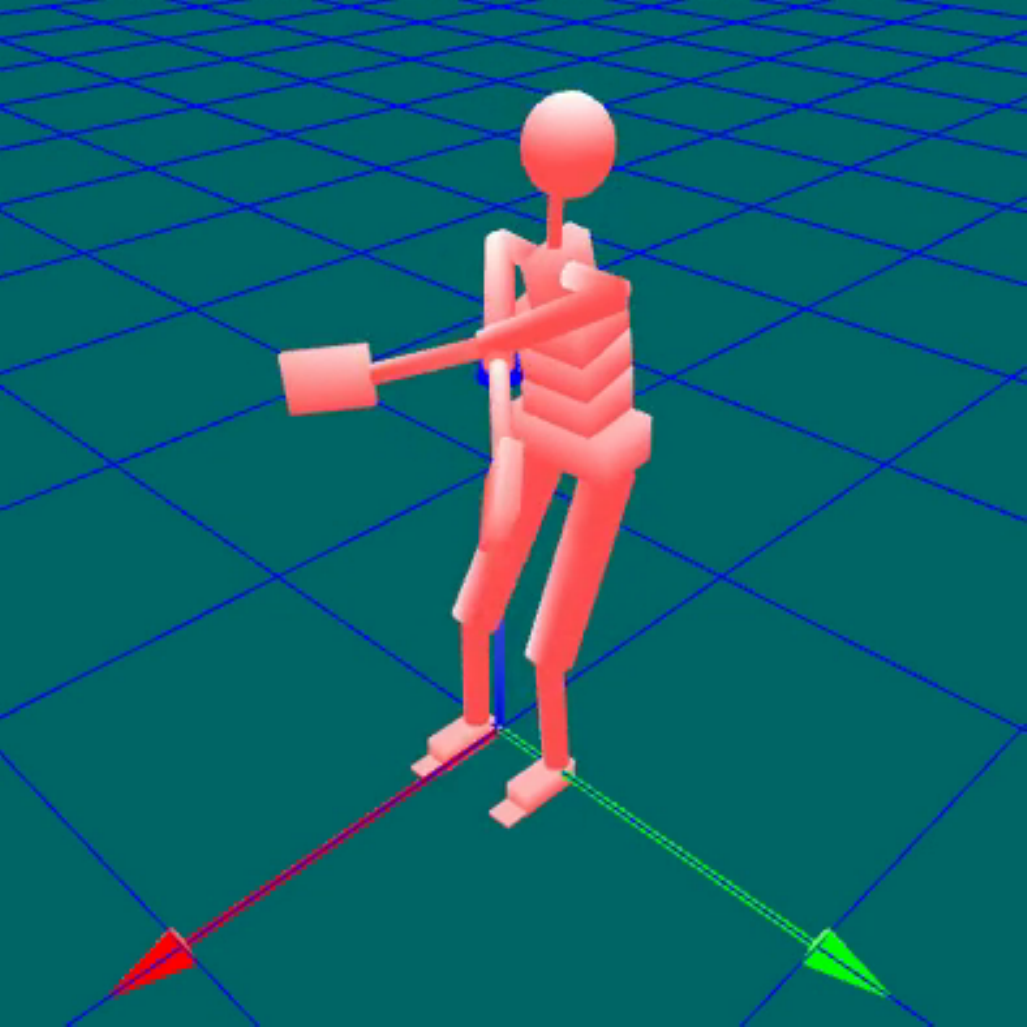}
\end{subfigure}
\begin{subfigure}{0.22\textwidth}
\includegraphics[scale=0.3]{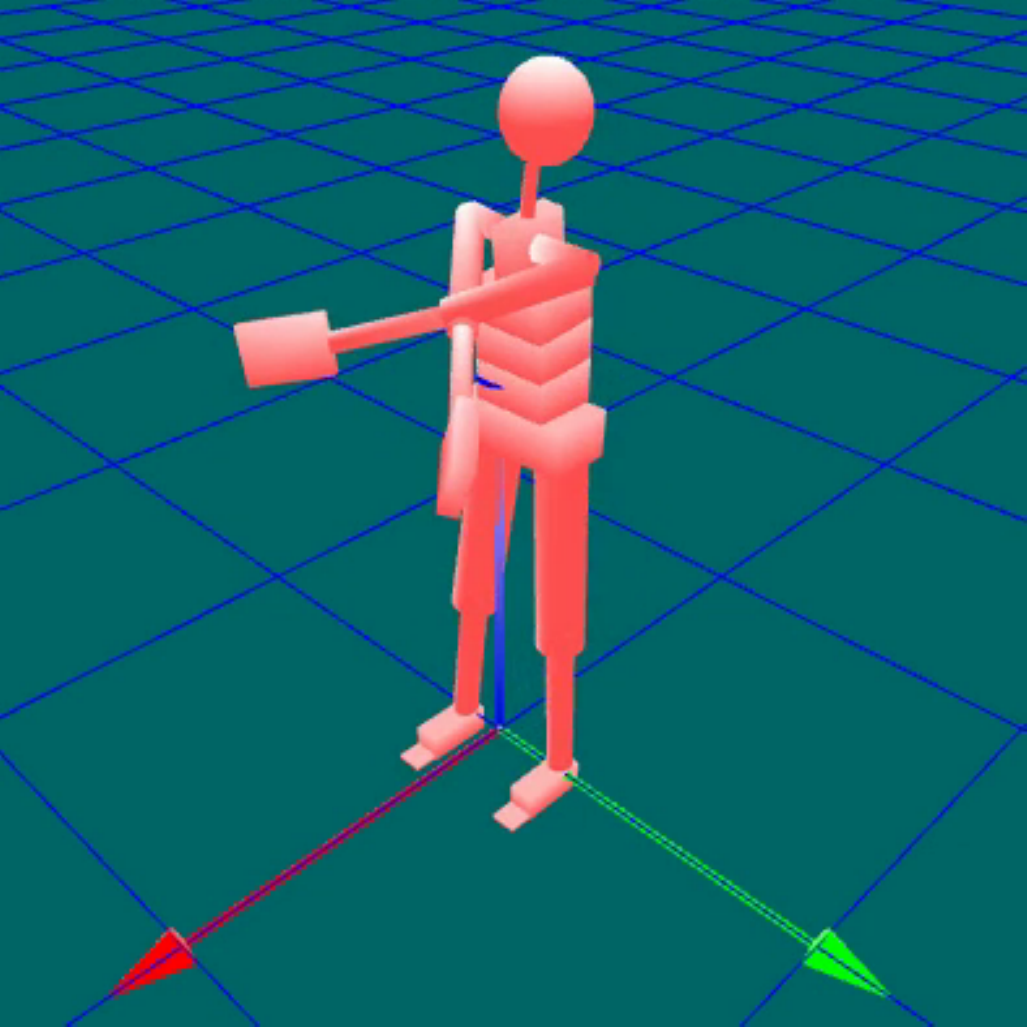}
\end{subfigure}
\begin{subfigure}{0.22\textwidth}
\includegraphics[scale=0.3]{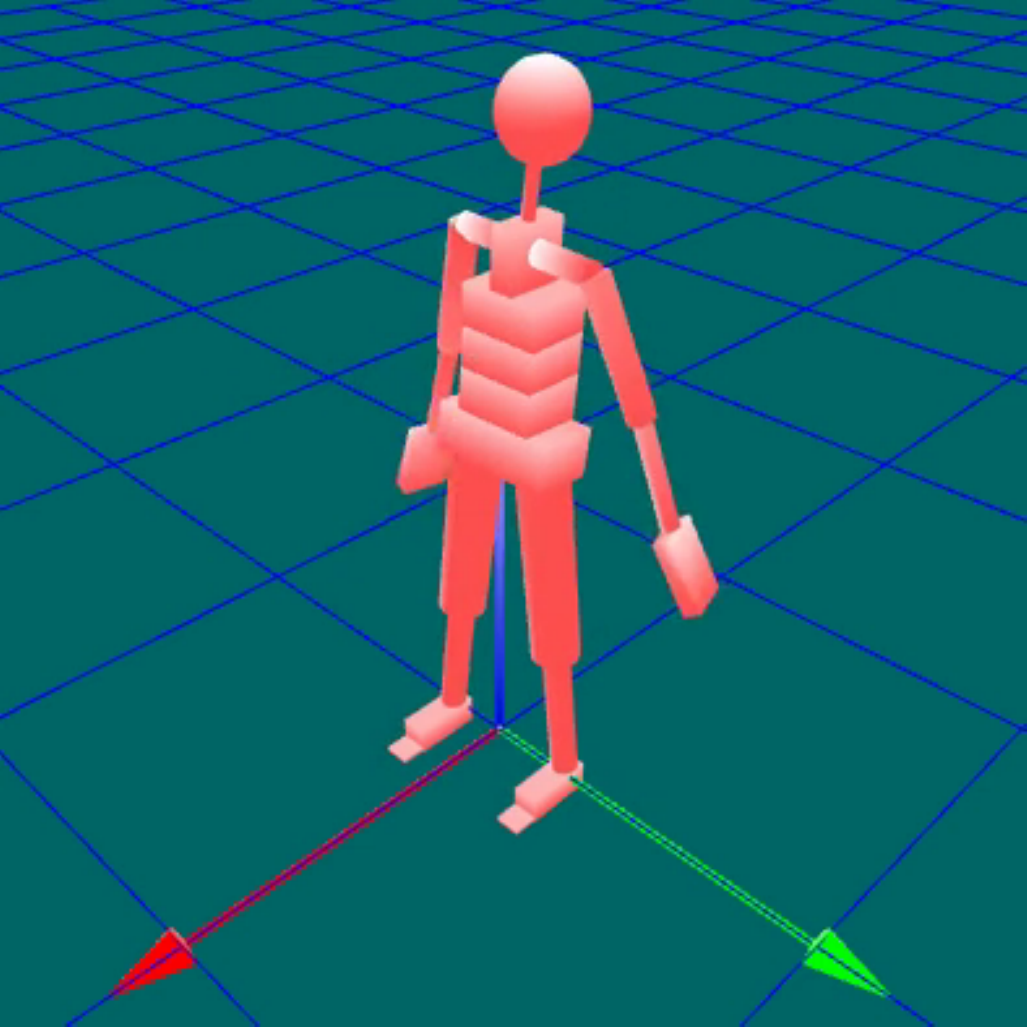}
\end{subfigure}
\caption{Human motion reconstruction for squatting experiment.}
\label{fig:chap:hbe:human-squat-evol}
\end{figure}

For the squatting experiment, the subject performs repeated squatting motions starting from an N-pose stationary configuration.
The squatting motion is performed with the arms stretched in front of the subject.
The whole sequence of actual and reconstructed motion is depicted in Figure \ref{fig:chap:hbe:human-squat-evol}. \looseness=-1

\begin{figure}[!h]
	\centering
\begin{subfigure}{0.7\textwidth}
	\centering
	\includegraphics[scale=0.05, width=\textwidth]{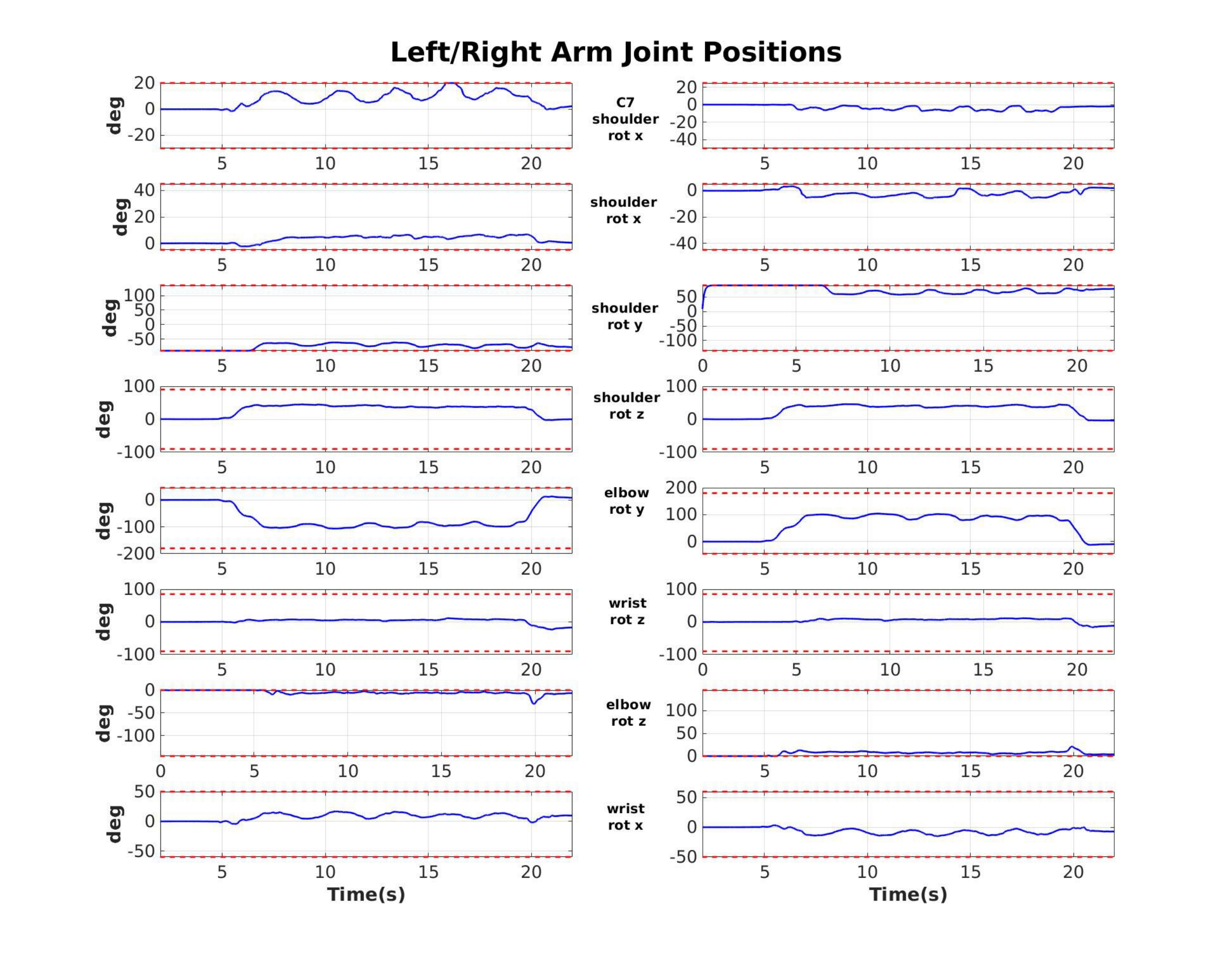}
	\caption{Joint positions of left and right arm estimated by the Dynamical IK shown as a blue line for the squatting experiment. Red dotted lines indicate the joint limits obtained from the URDF model.}
	\label{fig:chap:hbe:human-squat-arm-joints}
\end{subfigure}
\begin{subfigure}{0.7\textwidth}
	\centering
	\includegraphics[scale=0.05, width=\textwidth]{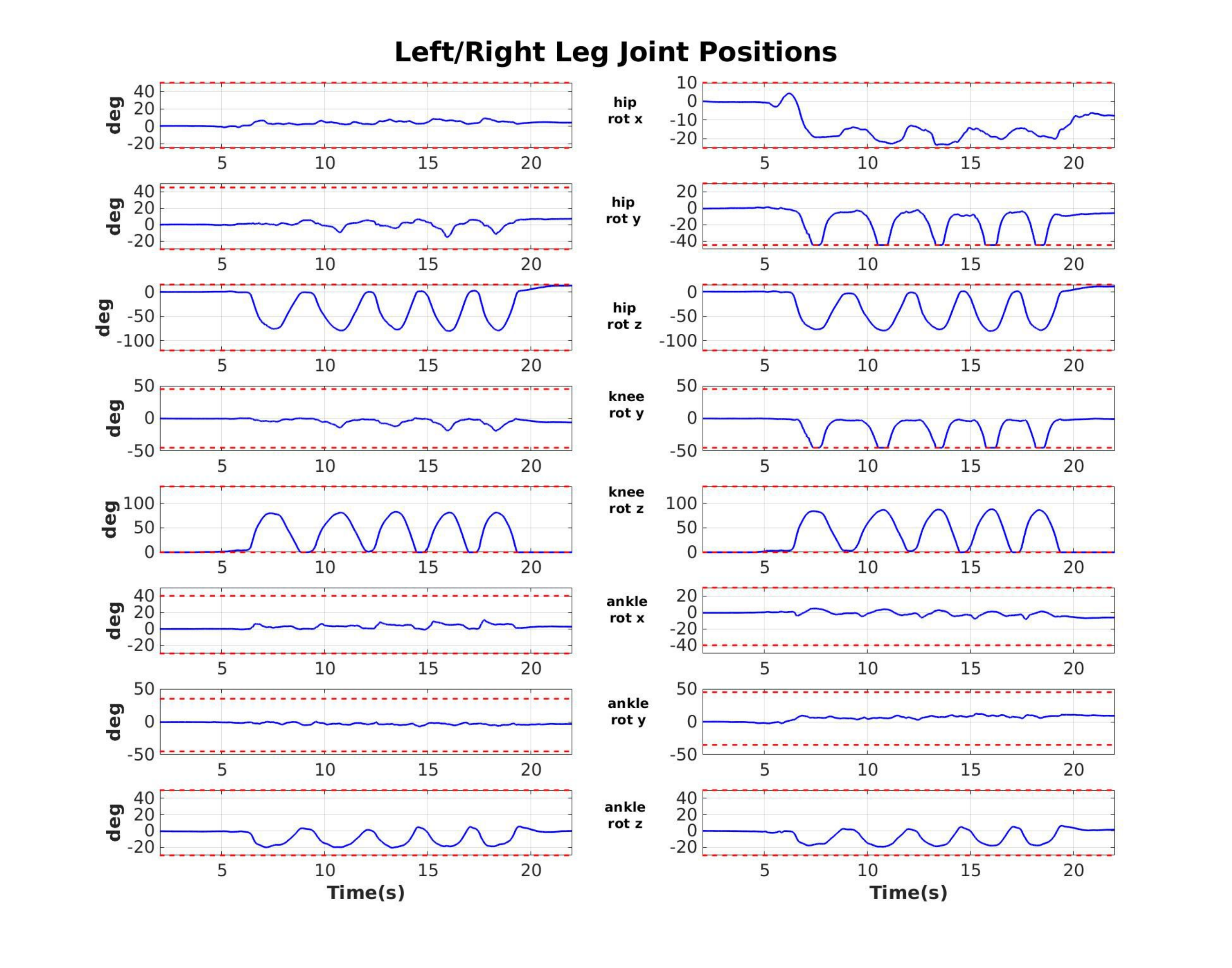}
	\caption{Joint positions of left and right leg estimated by the Dynamical IK shown as a blue line for the squatting experiment. Red dotted lines indicate the joint limits obtained from the URDF model. There is a clear asymmetry in the hip, knee and ankle \emph{RotY} joints and hip \emph{RotX} joints of the left and right leg, which are the crucial joints to reconstruct the squatting motion. \looseness=-1}
	\label{fig:chap:hbe:human-squat-leg-joints}
\end{subfigure}
\end{figure}

The overall motion is fairly reconstructed in comparison with the actual motion.
The stretching of the arms is properly recovered through the shoulder joints only for the left arm, this is reflected through the joint velocity profiles for the \emph{jRightShoulderC7} joint which remains not as excited as the \emph{jLeftShoulderC7} joint as seen in Figure \ref{fig:chap:hbe:human-squat-arm-joints}, there by failing to achieve a symmetric motion of the arms. \looseness=-1

\begin{figure}[!h]
	\centering
	\begin{subfigure}{0.9\textwidth}
		\centering
		\includegraphics[scale=0.05, width=\textwidth]{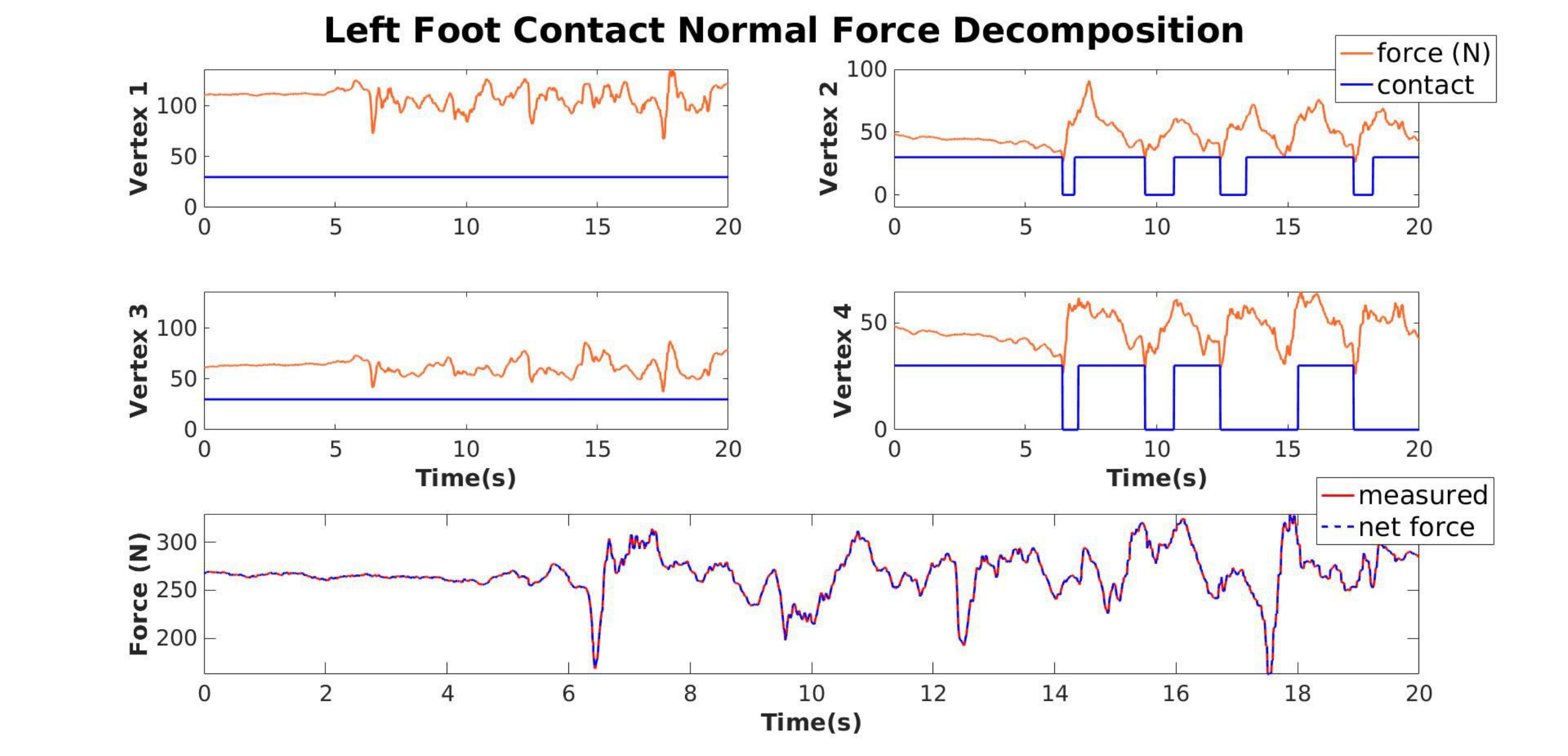}
		\caption{Wrench decomposition into contact normal forces for vertex contact detection of the left foot}
		\label{fig:chap:hbe:human-squat-lfoot}
	\end{subfigure}
	\begin{subfigure}{0.9\textwidth}
		\centering
		\includegraphics[scale=0.05, width=\textwidth]{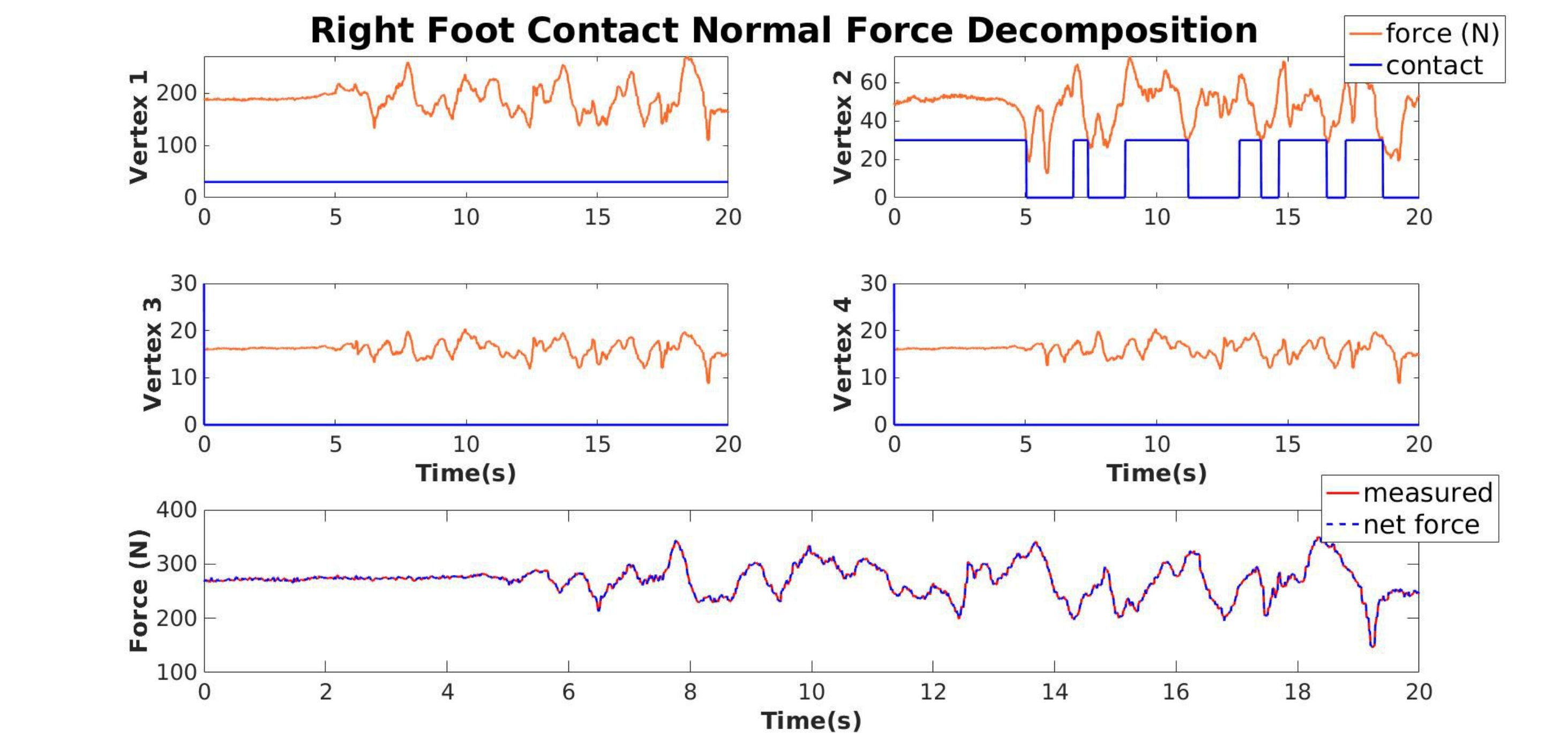}
		\caption{Wrench decomposition into contact normal forces for vertex contact detection of the right foot}
		\label{fig:chap:hbe:human-squat-rfoot}
	\end{subfigure}
\end{figure}

As the subject squats, a forward-leaning motion is implied leading to the center of pressure moving forward along the support foot polygon and resting close to the toes of the subject.
This is indicated by very high magnitudes of forces up to $120$ \si{\newton} at one of the upper vertices of the left foot and up to $180$ \si{\newton} for the right foot while the forces at other vertices remain close to $50$ \si{\newton} as seen in Figures and \ref{fig:chap:hbe:human-squat-lfoot} and \ref{fig:chap:hbe:human-squat-rfoot}..
The weight of the subject is $55.2$ \si{\kilogram} which would imply a total of $270$ \si{\newton} on each foot when the weight is evenly distributed which is observed by summing up the contact normal forces which is evident in Figures and \ref{fig:chap:hbe:human-squat-lfoot} and \ref{fig:chap:hbe:human-squat-rfoot}.
A force distribution with peak force near the toes  near the squatting motion does not allow to activate the contact plane orientation update to better constrain the motion , since this update is enforced only when all the vertices are inferred to be in contact.
Near a full squat configuration, some of the vertices of the feet are not inferred to be in contact given the tuned Schmitt trigger thresholds.

\begin{figure}[!h]
\centering
	\begin{subfigure}{0.9\textwidth}
		\centering
\includegraphics[scale=0.05, width=\textwidth]{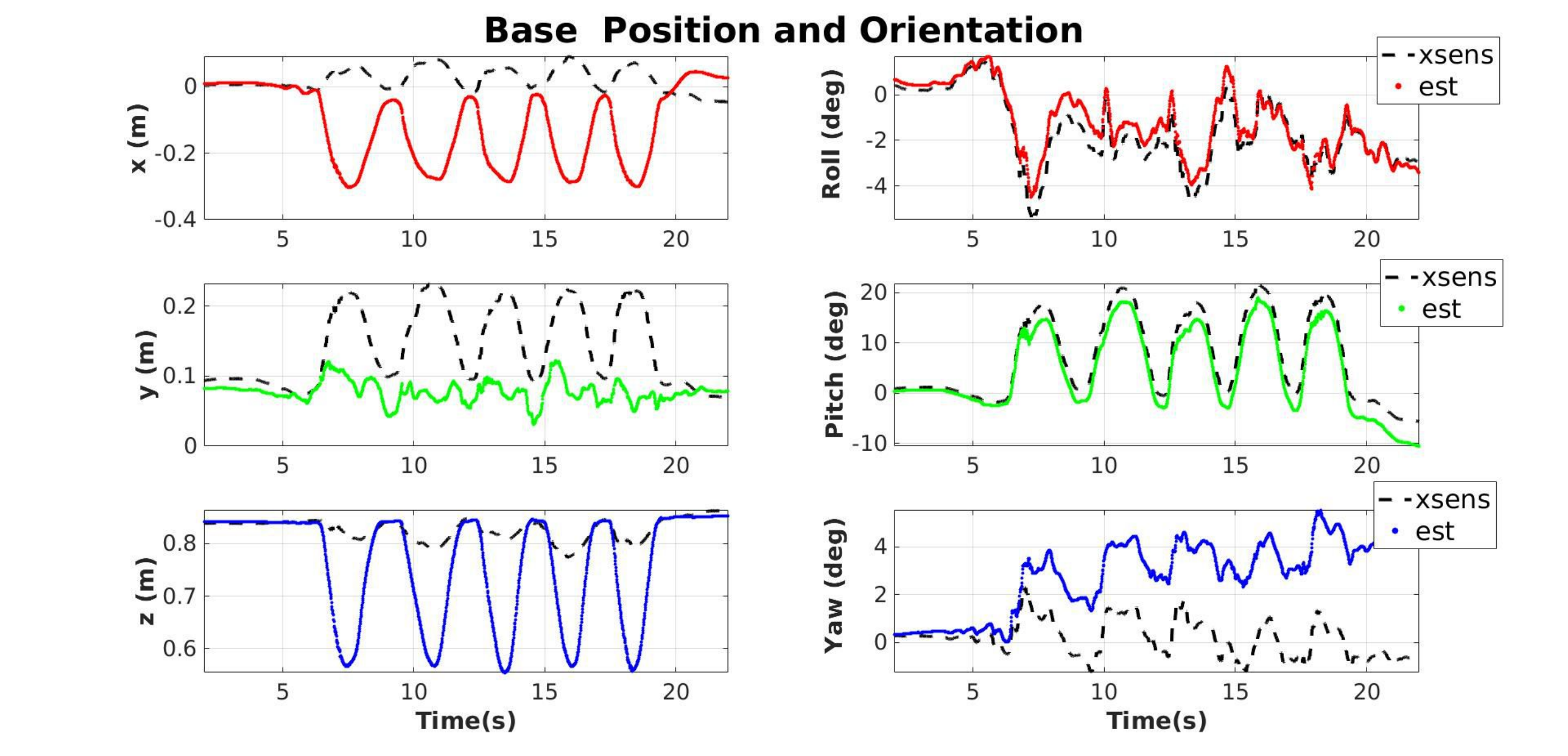}
	\end{subfigure}
\begin{subfigure}{0.9\textwidth}
		\centering
\includegraphics[scale=0.05, width=\textwidth]{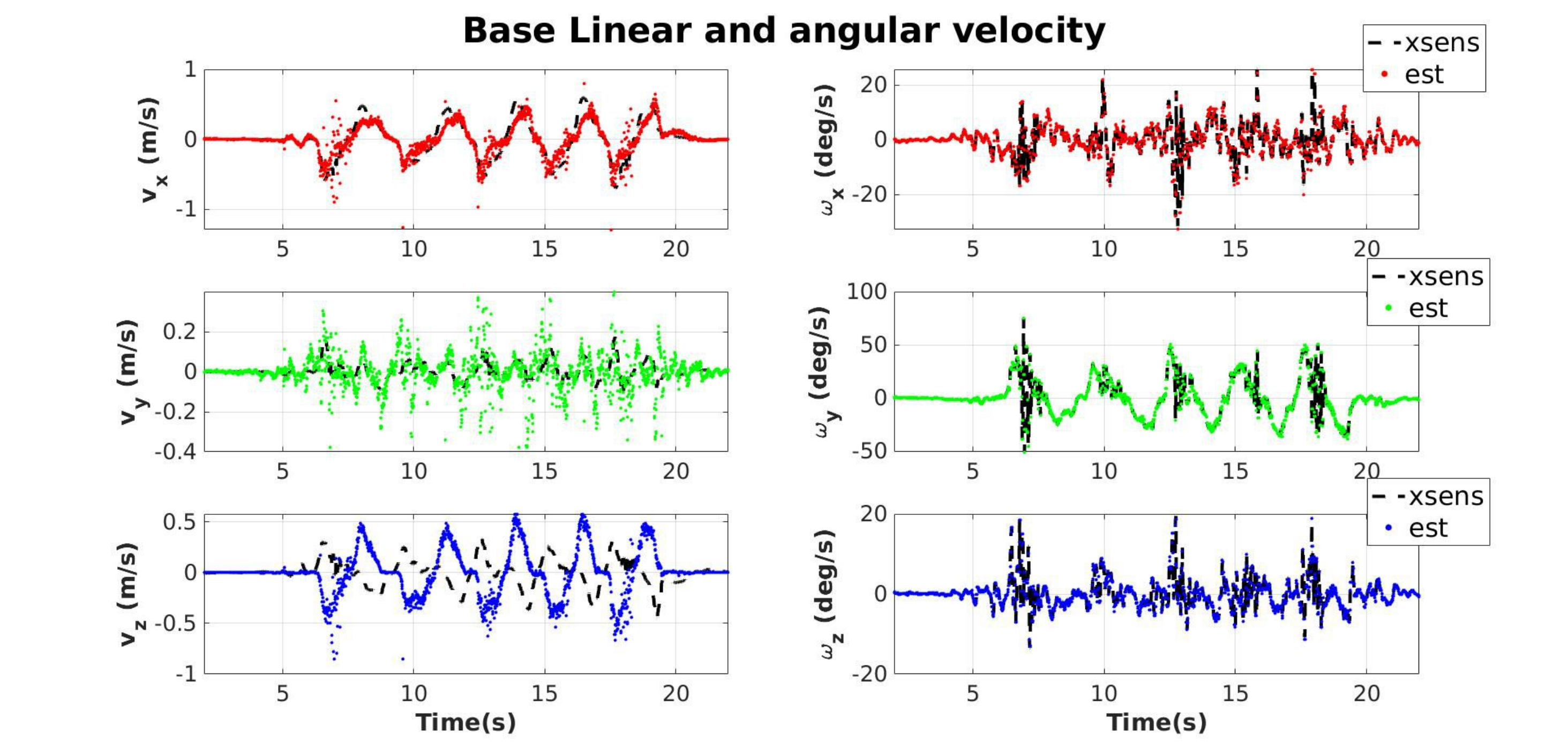}
	\end{subfigure}	
\begin{subfigure}{0.9\textwidth}
		\centering
\includegraphics[scale=0.05, width=\textwidth]{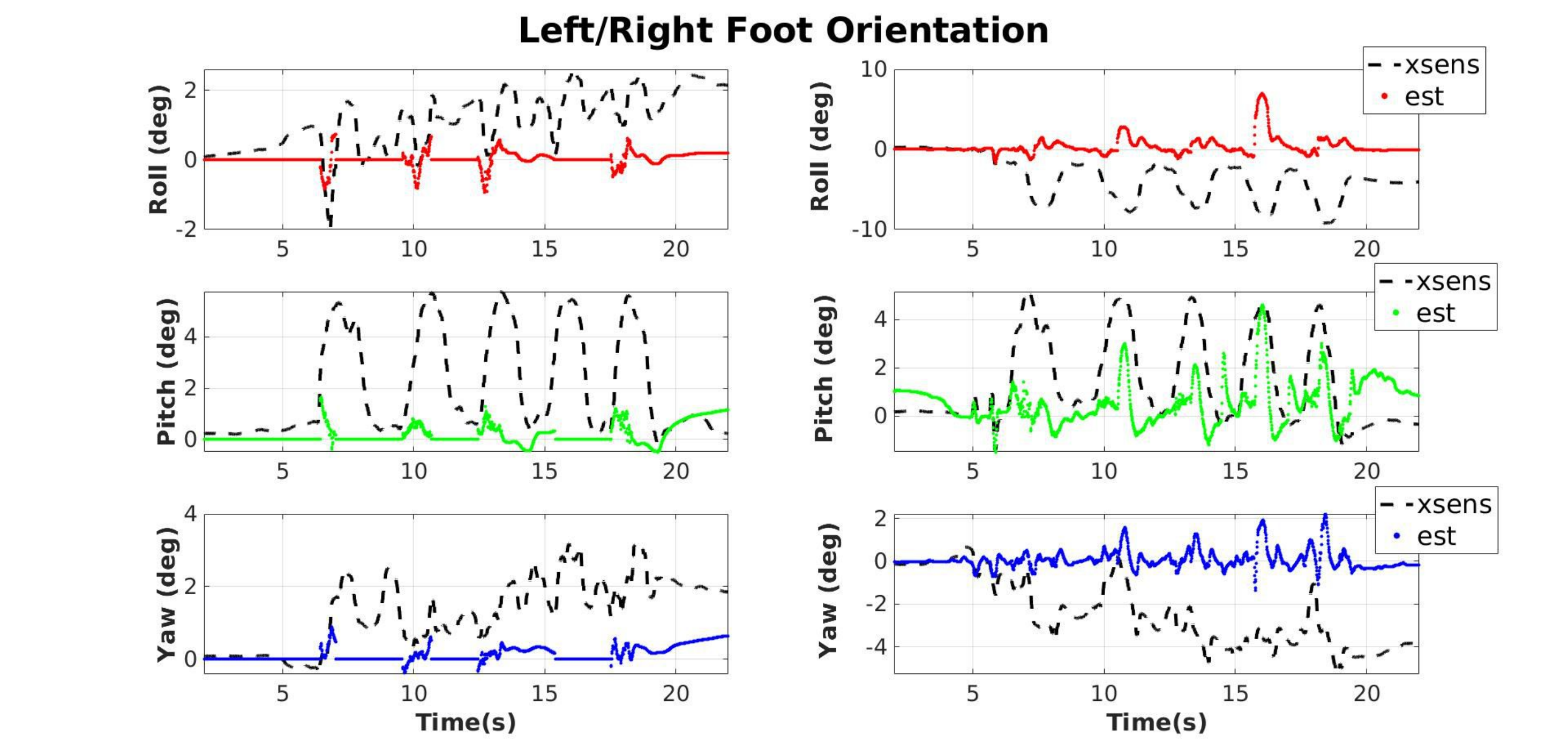}
	\end{subfigure}	
\caption{Base pose and velocity and feet rotation estimates for human squatting experiment.}
\label{fig:chap:hbe:human-squat-base-state}
\end{figure}

This is clearly evident in the foot orientation estimates depicted in Figure \ref{fig:chap:hbe:human-squat-base-state}, where at time instants when the contact plane orientation update is not activated, the foot orientation slightly changes from zero, for instance during $t=10$ seconds.
The contact plane orientation update is enforced every time the subject is near upright configuration during which all the vertices of the left foot are inferred to be in contact in Figure \ref{fig:chap:hbe:human-squat-lfoot}.
It is also clear that this orientation update is never applied to the right foot leading to slightly higher magnitude of its orientation.
This phenomenon along with the clear asymmetry in the hip, knee and ankle \emph{RotY} joints and hip \emph{RotX} joints of the left and right leg as seen in Figure \ref{fig:chap:hbe:human-squat-leg-joints}, which are the crucial joints to reconstruct the squatting motion are the reasons for the noticeable yaw drifts in the base orientation seen in Figure \ref{fig:chap:hbe:human-squat-base-state}.
This is also the cause for the slipping-like motion observed also in this experiment as noticed in the reconstructed configurations during the initial and end phase, where the human subject has shifted along the negative y-direction slightly. \looseness=-1

However, it can be seen from the evolution of the base pose and velocity as shown in Figure \ref{fig:chap:hbe:human-squat-base-state} in comparison with the XSens measured quantities that, the position trajectories estimated by the proposed method follow the actual base motion more reliably than the XSens base trajectory.
As the human subject does the squatting motion, the knee joints are bent in such a way that the base height is reduced considerably and the base link moves backward to accommodate the bending motion.
This is clearly seen in the base position trajectory through a reduction of the base height by $20 \si{\centi\meter}$ and corresponding decrease in the base position $\mathbf{x}$ upto $30 \si{\centi\meter}$ implying a backward motion, every time the subject enters into a full squat configuration from the upright configuration.
This nominal motion  is not reconstructed by the Xsens. \looseness=-1

\subsubsection{In-place Swinging}
\begin{figure}[!h]
\centering
\begin{subfigure}{\textwidth}
\centering
\includegraphics[scale=0.45]{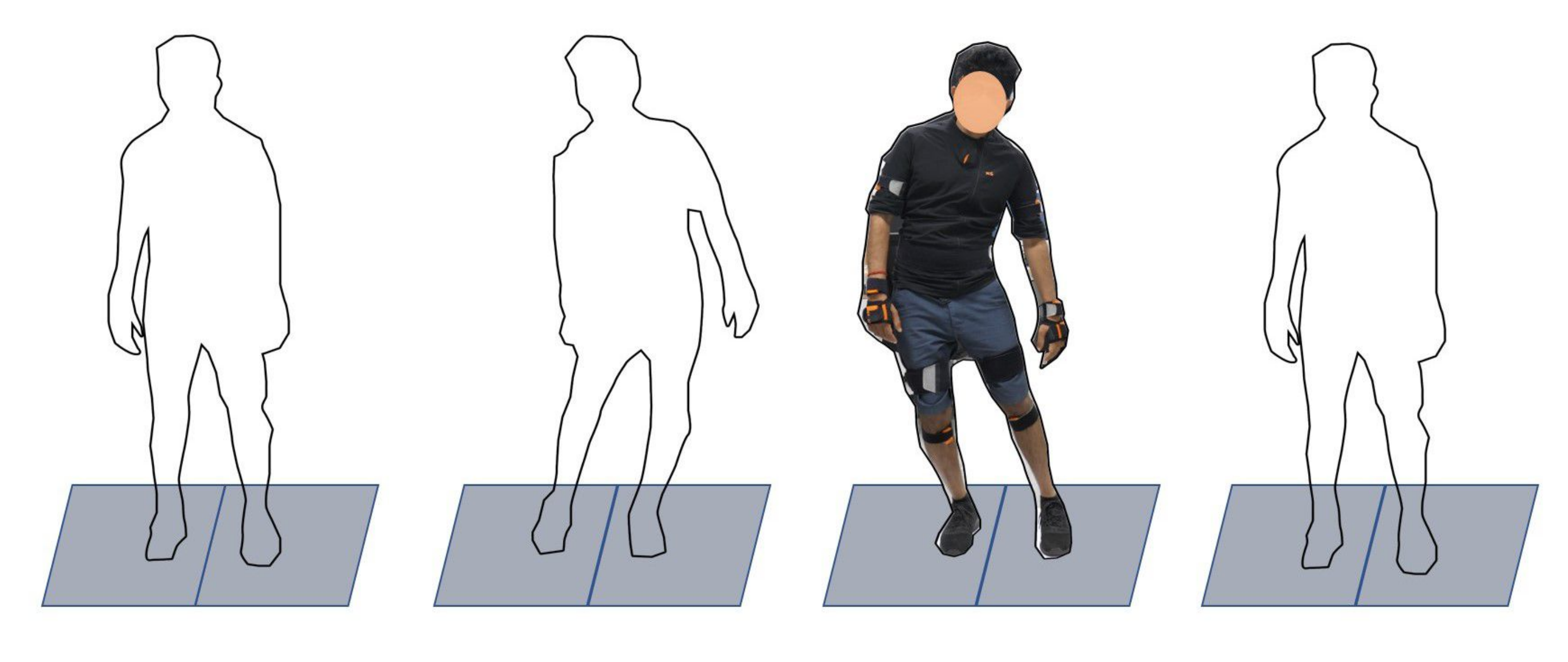}
\end{subfigure}
\begin{subfigure}{0.2\textwidth}
\includegraphics[scale=0.275]{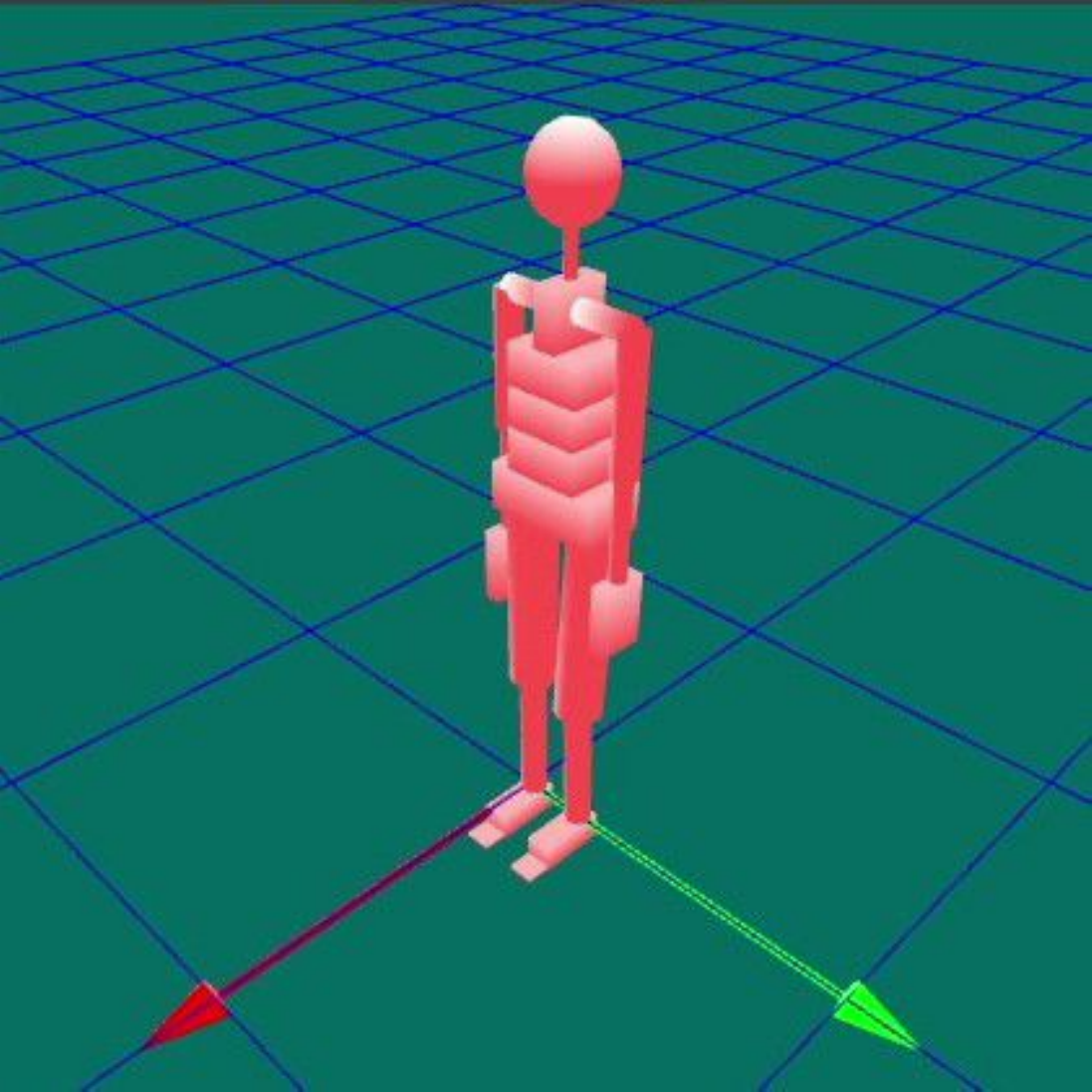}
\end{subfigure}
\begin{subfigure}{0.2\textwidth}
\includegraphics[scale=0.275]{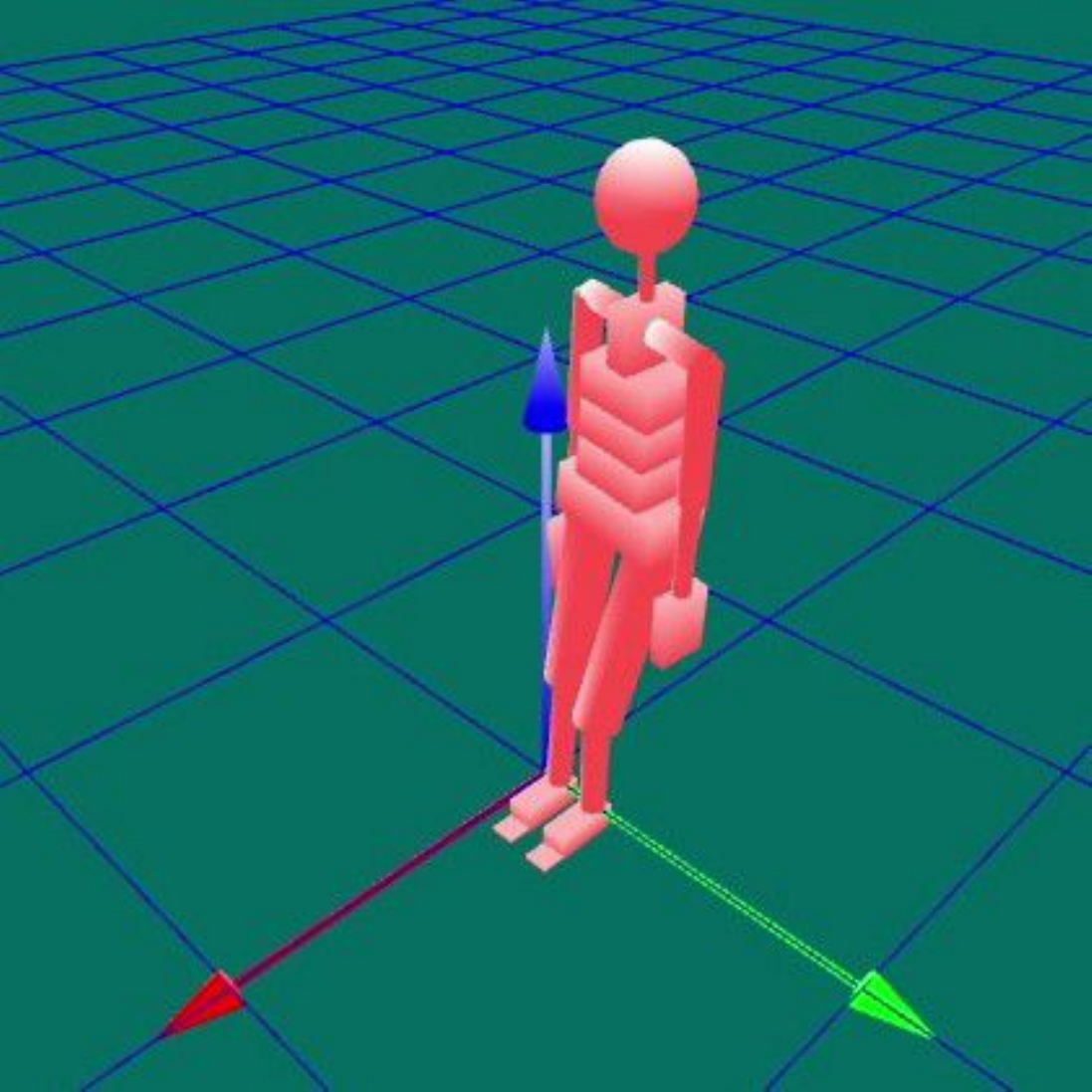}
\end{subfigure}
\begin{subfigure}{0.2 \textwidth}
\includegraphics[scale=0.275]{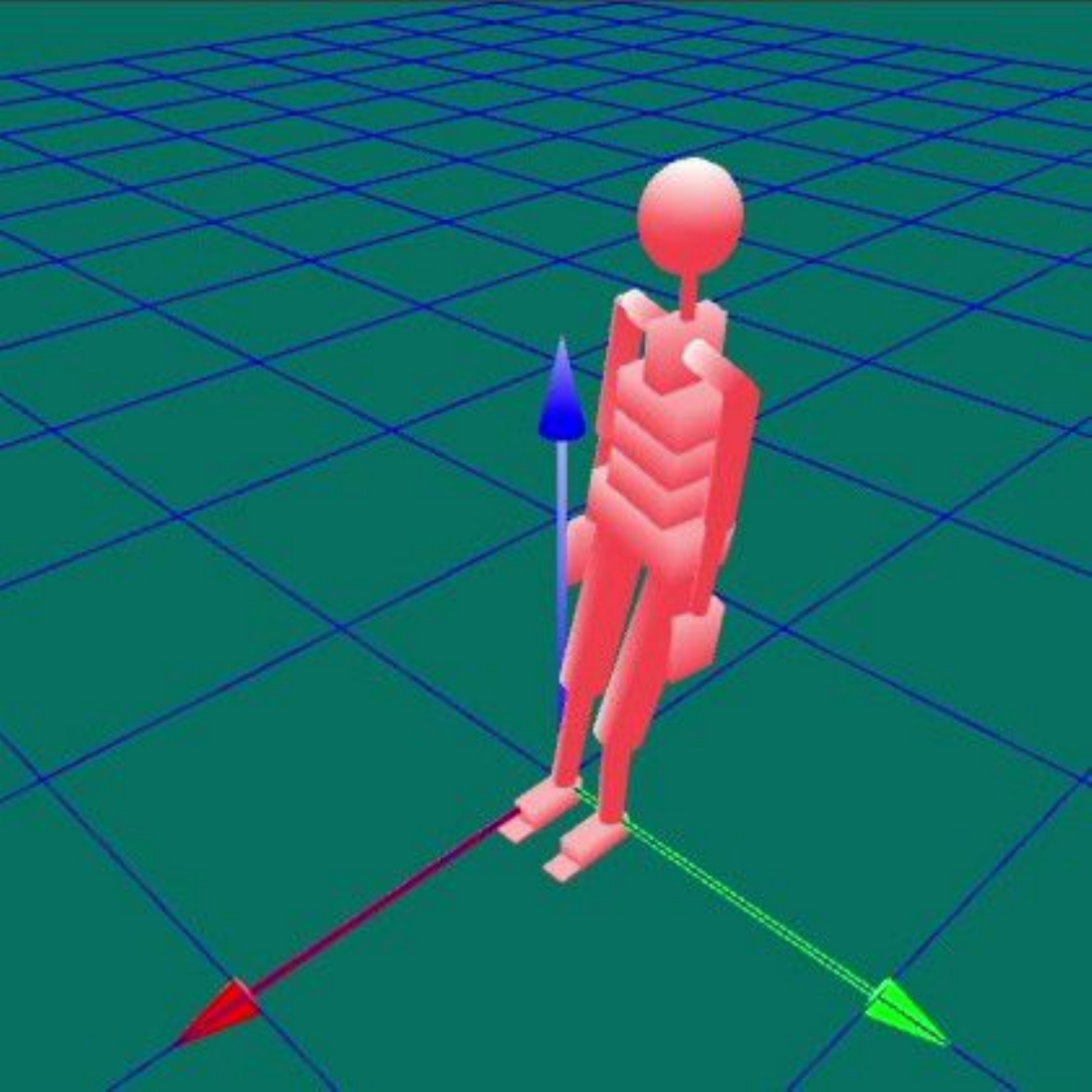}
\end{subfigure}
\begin{subfigure}{0.2\textwidth}
\includegraphics[scale=0.275]{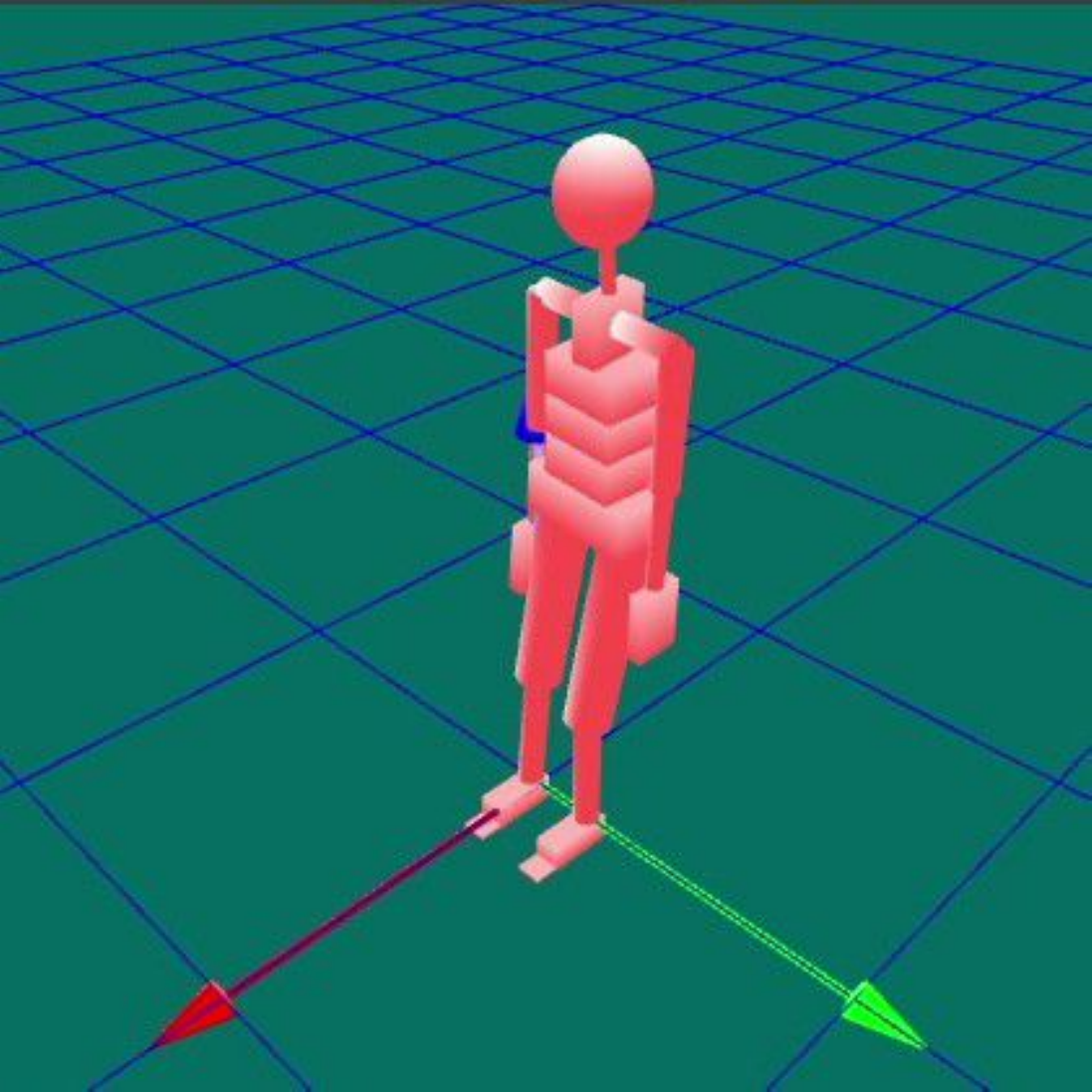}
\end{subfigure}
\begin{subfigure}{0.2\textwidth}
\includegraphics[scale=0.275]{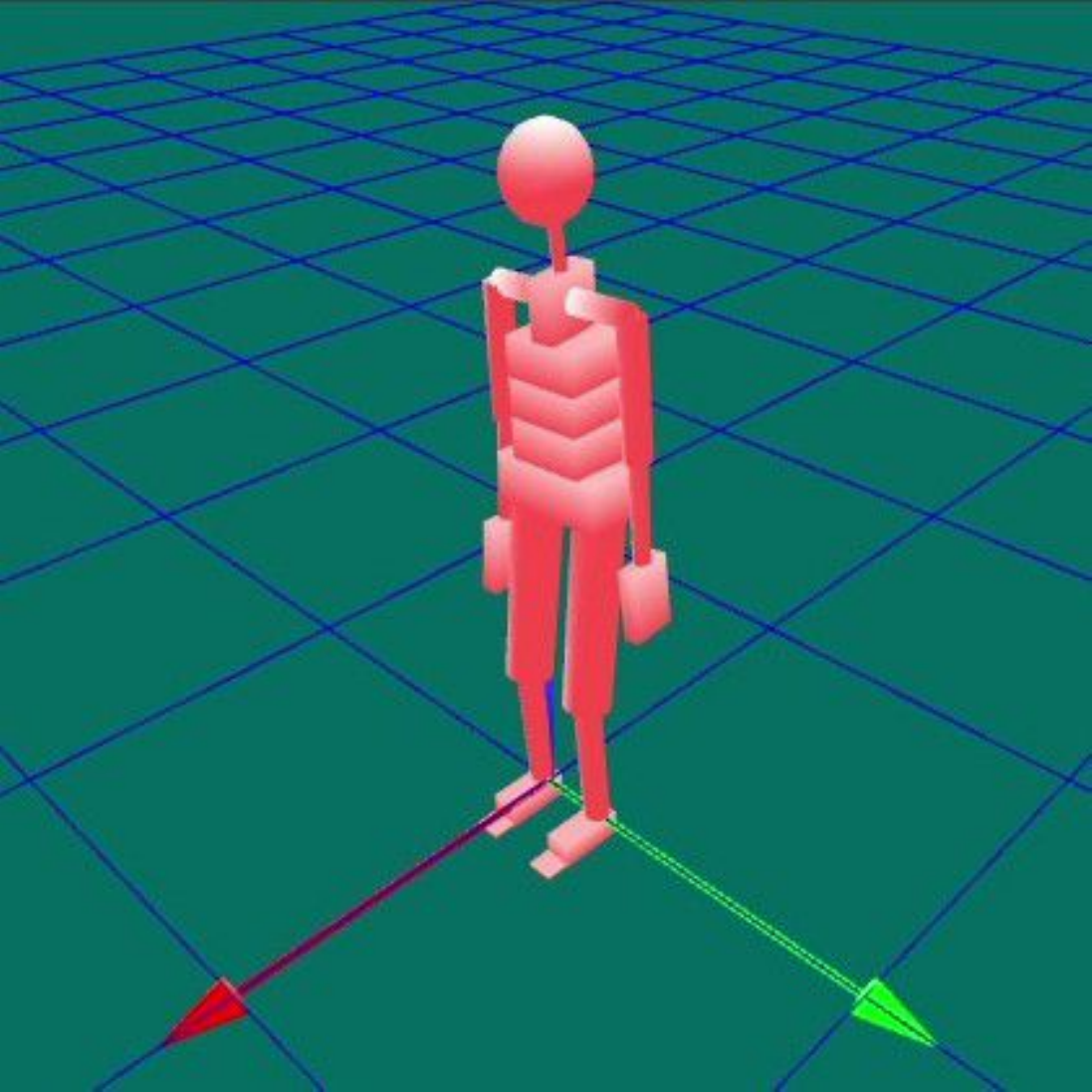}
\end{subfigure}
\begin{subfigure}{0.2\textwidth}
\includegraphics[scale=0.275]{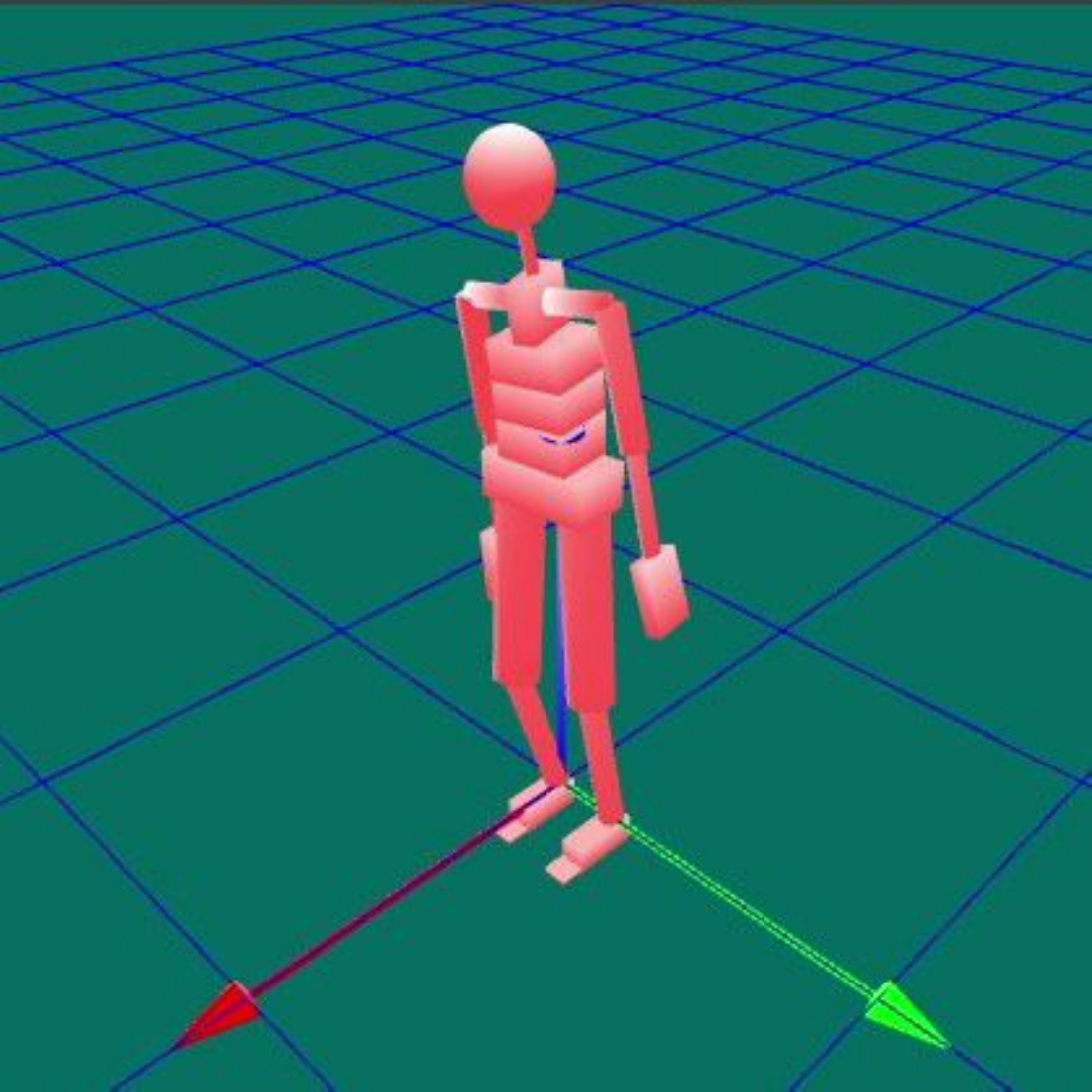}
\end{subfigure}
\begin{subfigure}{0.2\textwidth}
\includegraphics[scale=0.275]{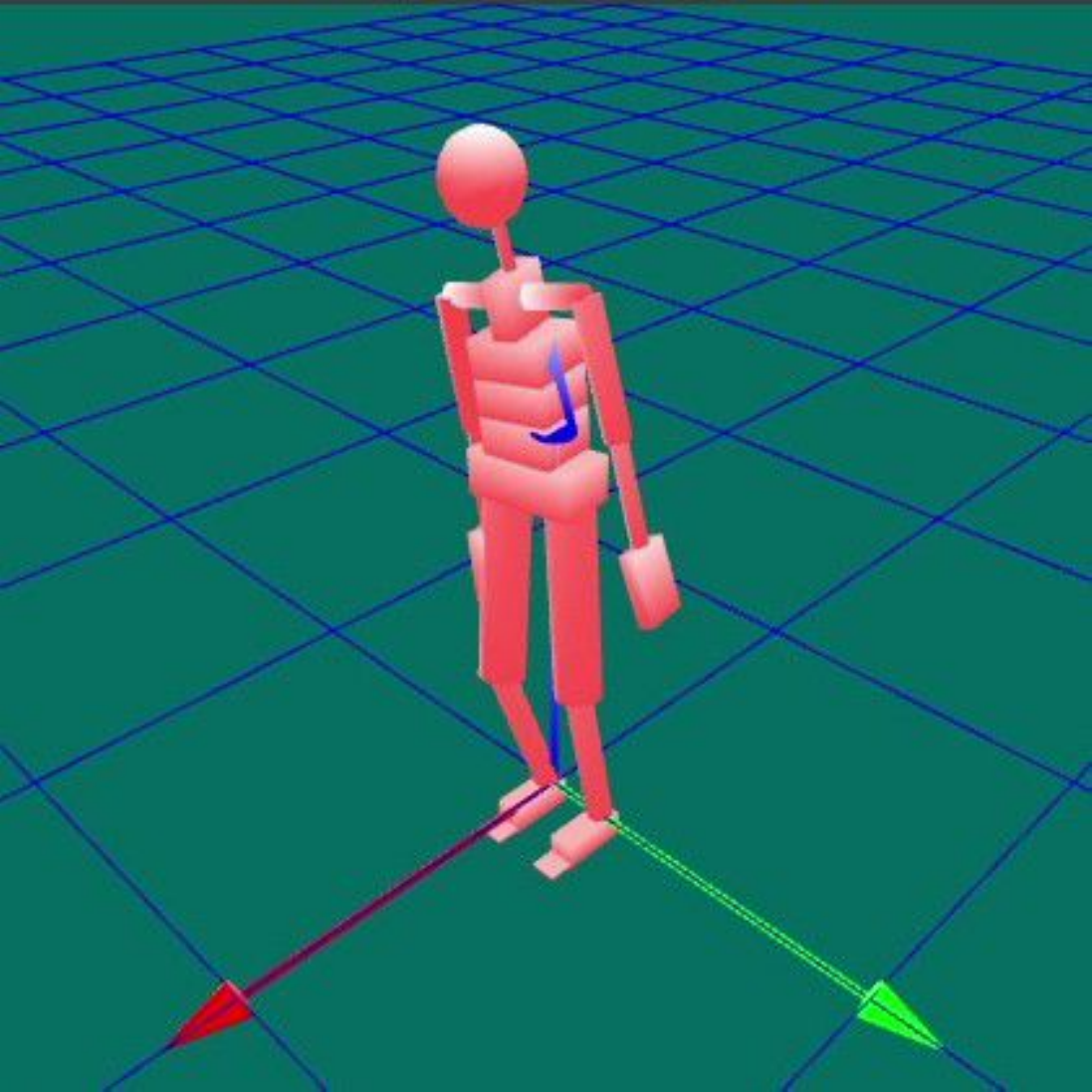}
\end{subfigure}
\begin{subfigure}{0.2\textwidth}
\includegraphics[scale=0.275]{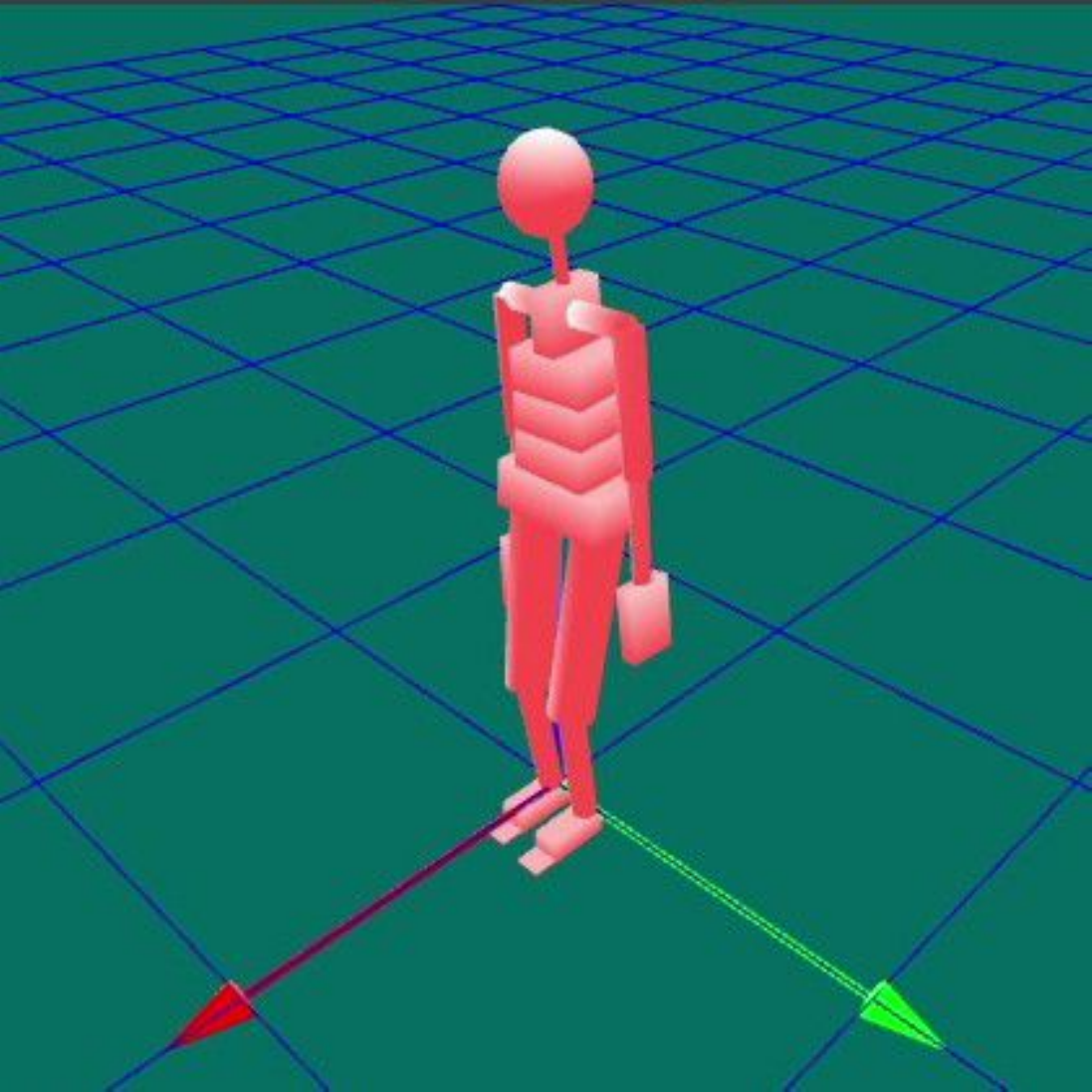}
\end{subfigure}
\begin{subfigure}{0.2\textwidth}
\includegraphics[scale=0.275]{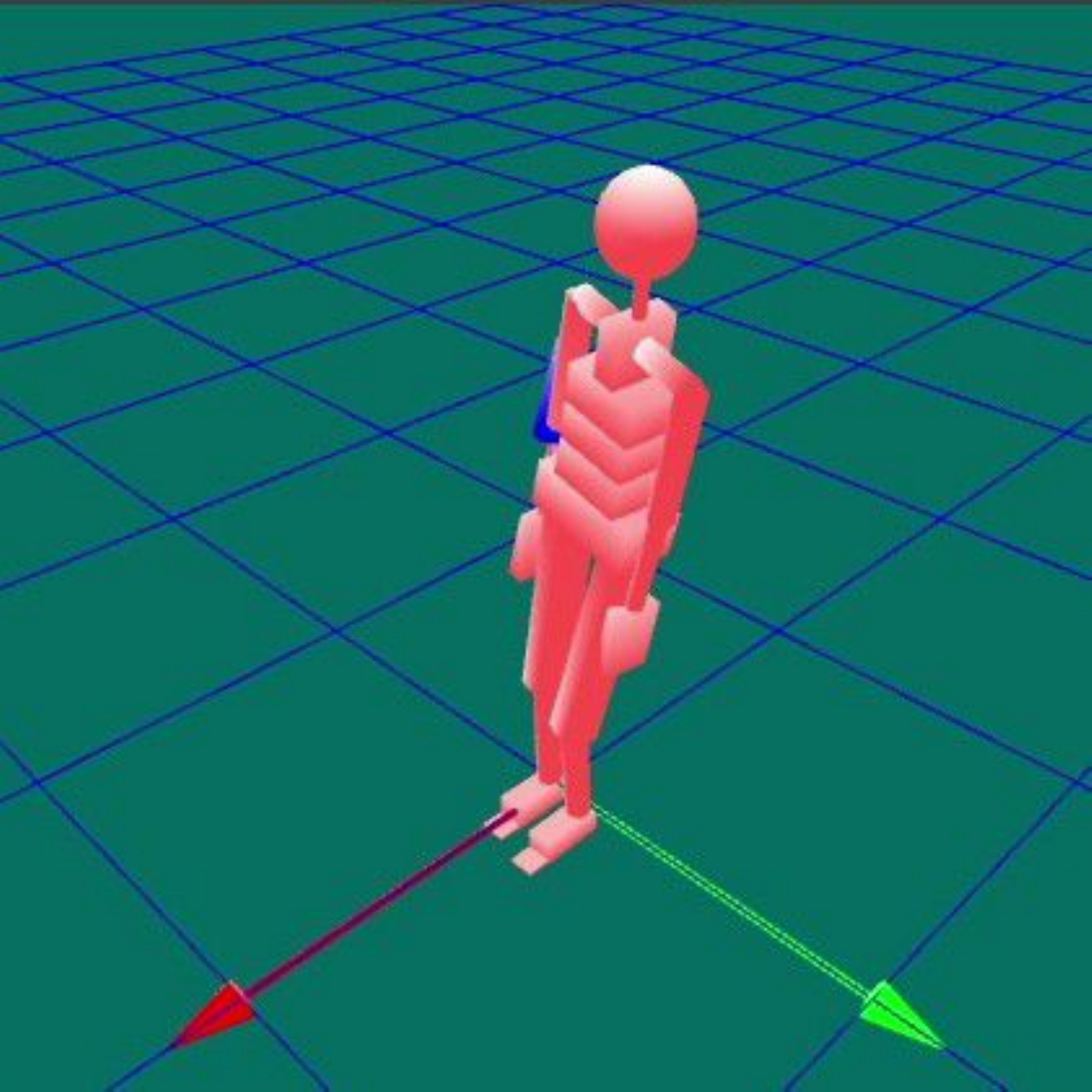}
\end{subfigure}
\begin{subfigure}{0.2\textwidth}
\includegraphics[scale=0.275]{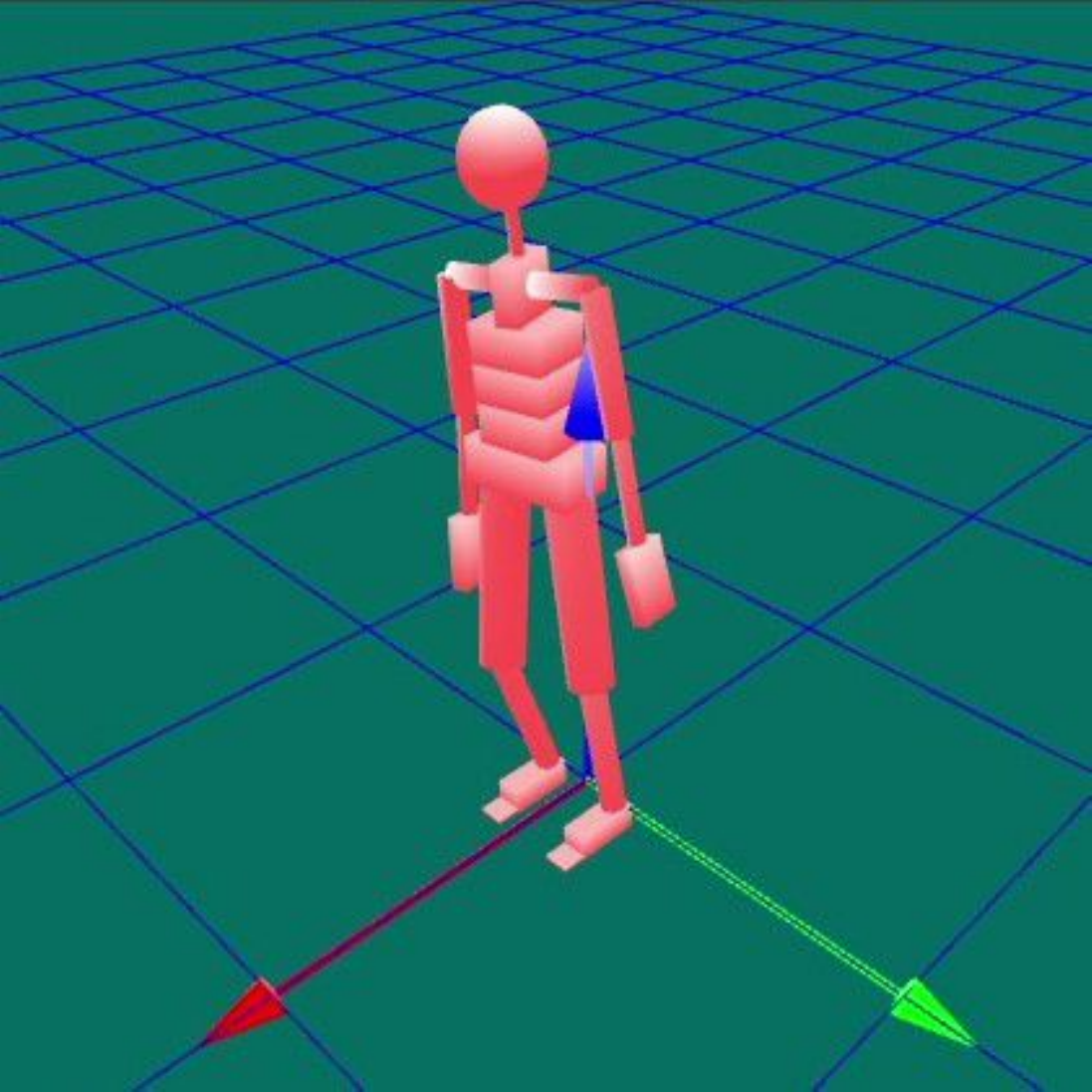}
\end{subfigure}
\begin{subfigure}{0.2\textwidth}
\includegraphics[scale=0.275]{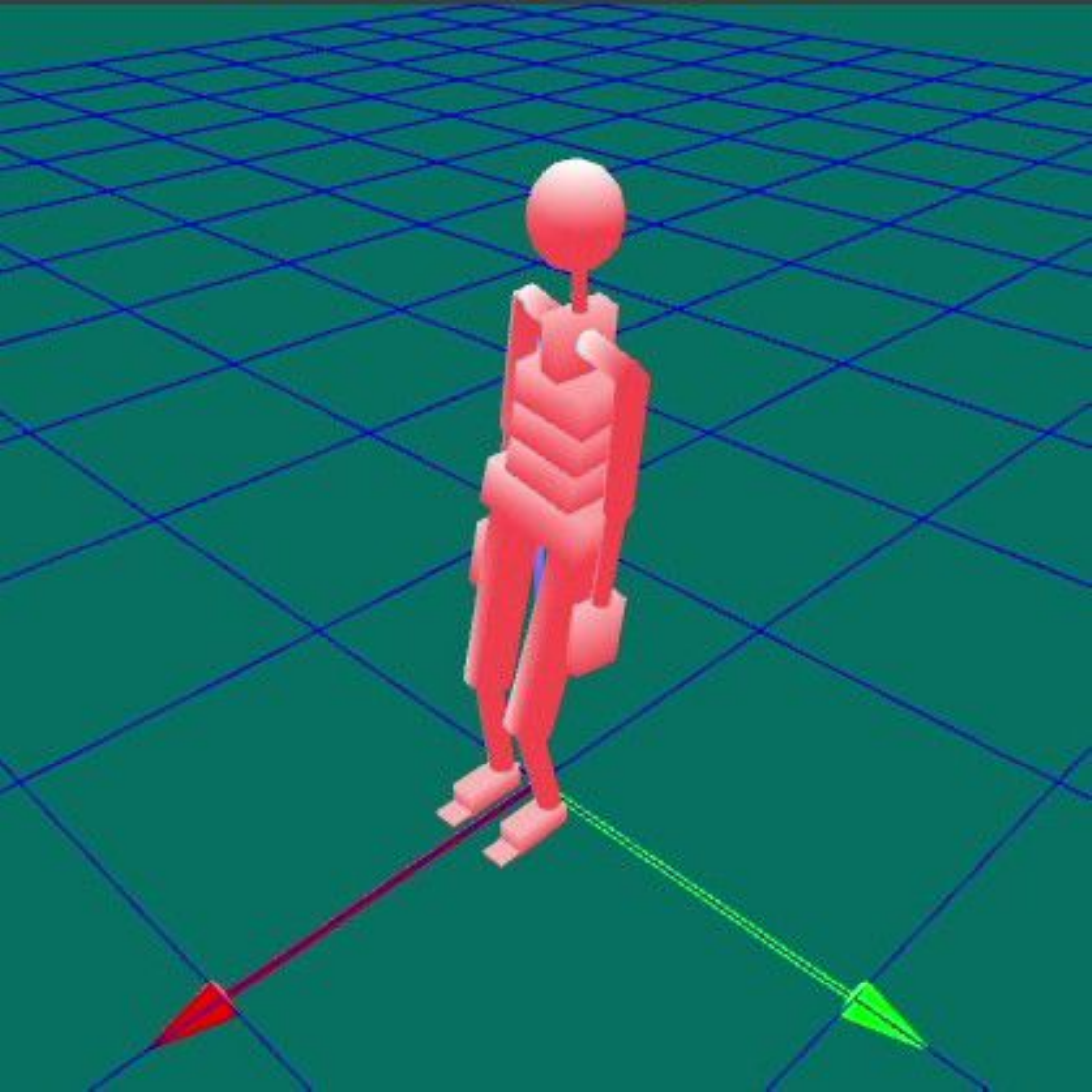}
\end{subfigure}
\begin{subfigure}{0.2\textwidth}
\includegraphics[scale=0.275]{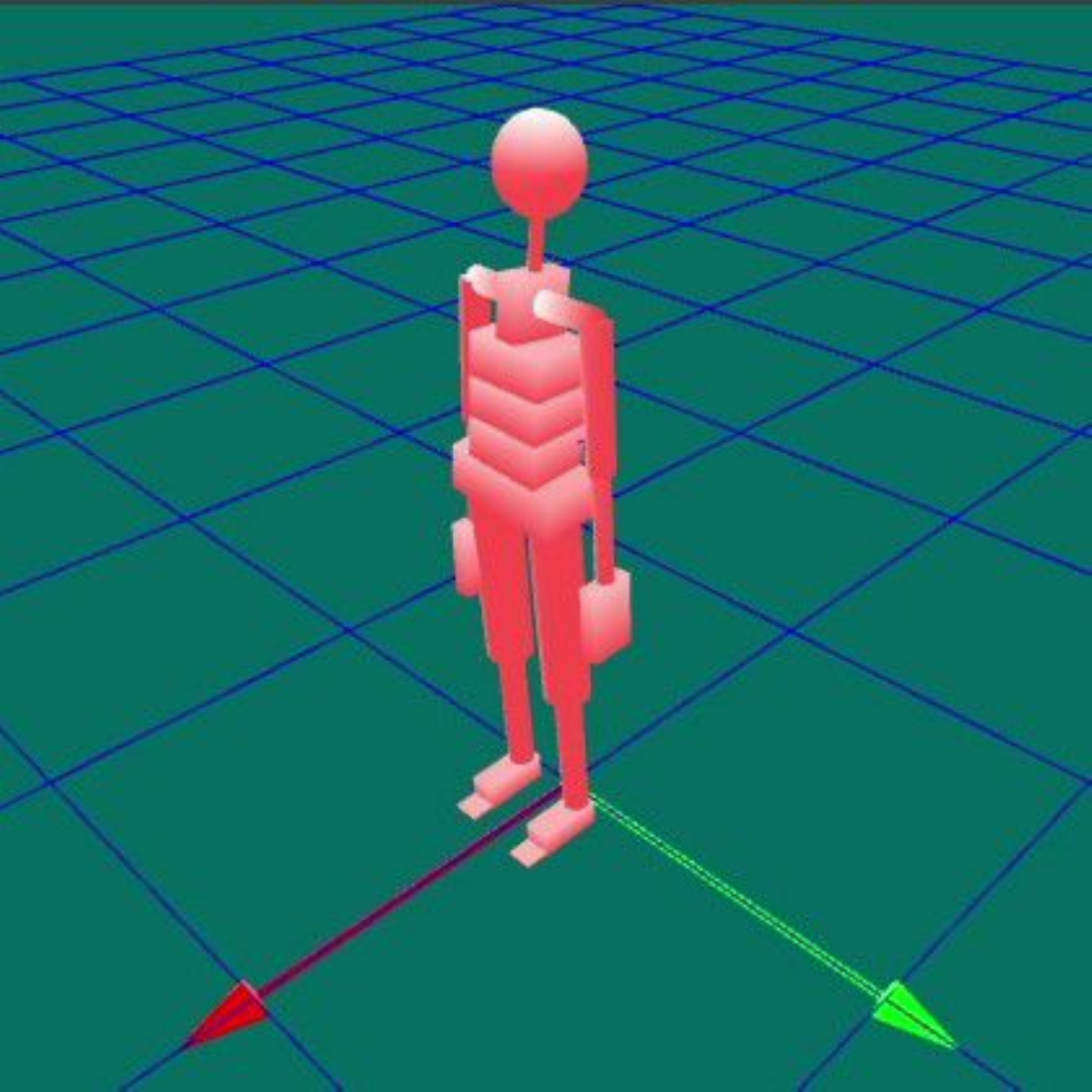}
\end{subfigure}
\caption{Human motion reconstruction for in-place swinging experiment.}
\label{fig:chap:hbe:human-swing-evol}
\end{figure}

\begin{figure}[!h]
\centering
	\begin{subfigure}{0.9\textwidth}
		\centering
\includegraphics[scale=0.05, width=\textwidth]{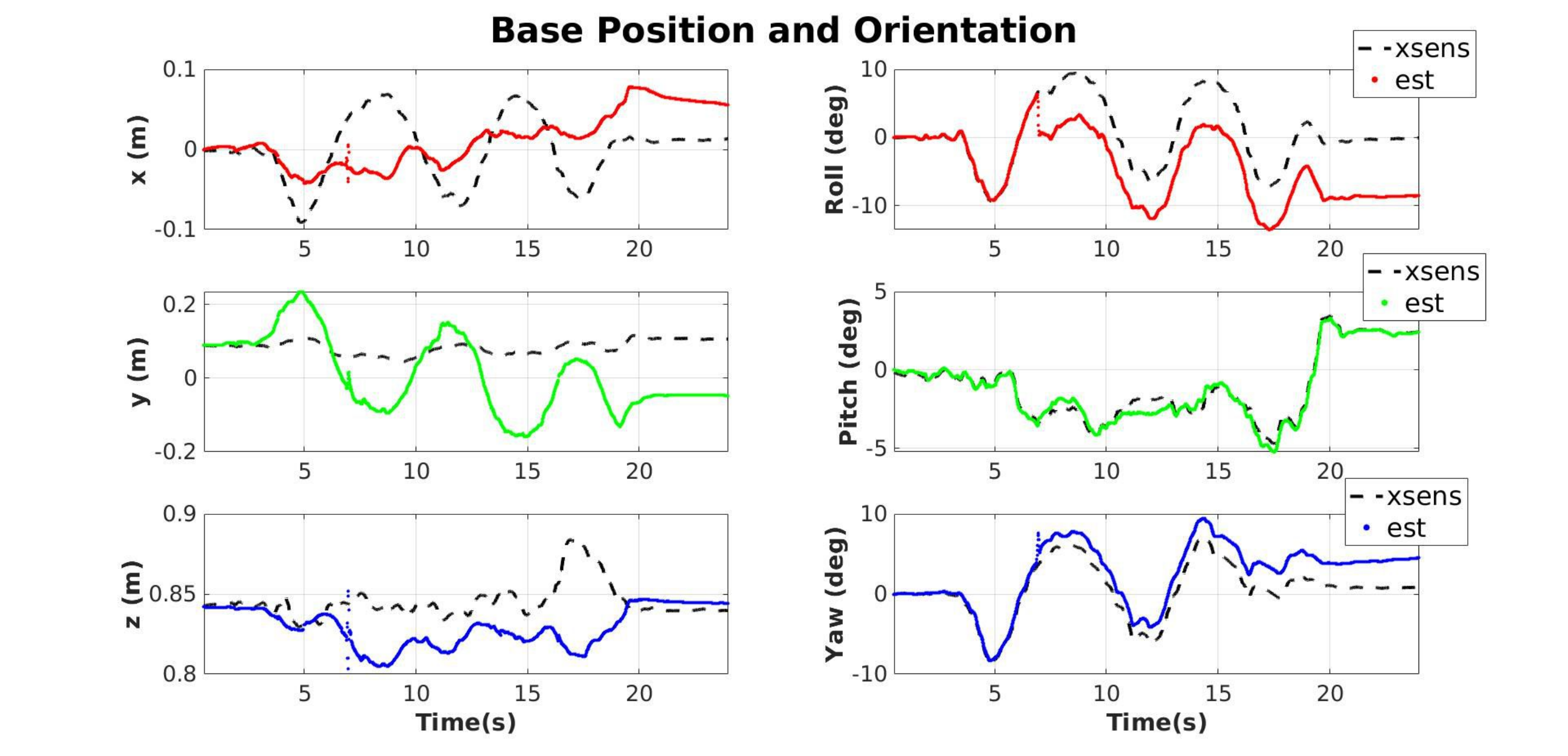}
	\end{subfigure}
\begin{subfigure}{0.9\textwidth}
		\centering
\includegraphics[scale=0.05, width=\textwidth]{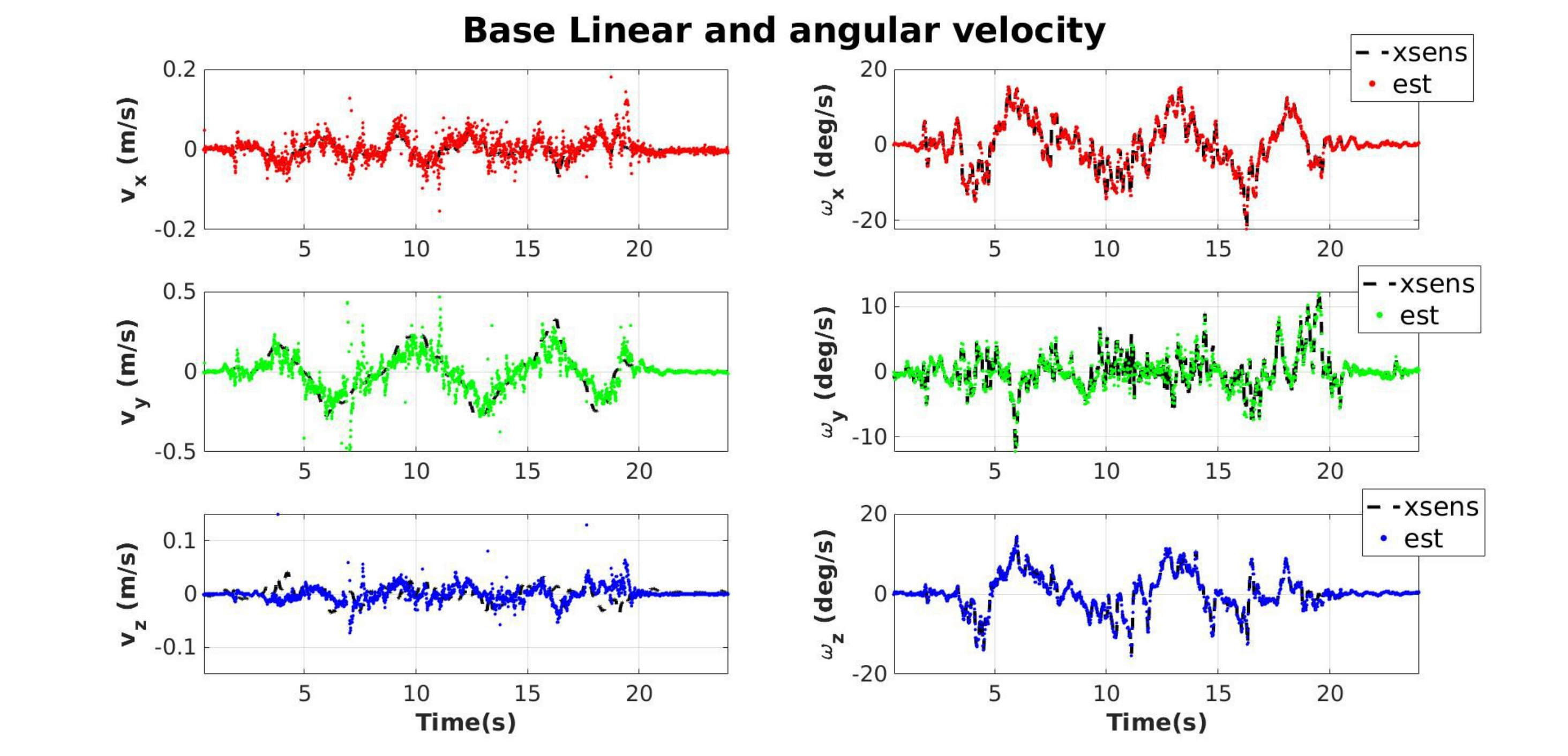}
	\end{subfigure}	
\begin{subfigure}{0.9\textwidth}
		\centering
\includegraphics[scale=0.05, width=\textwidth]{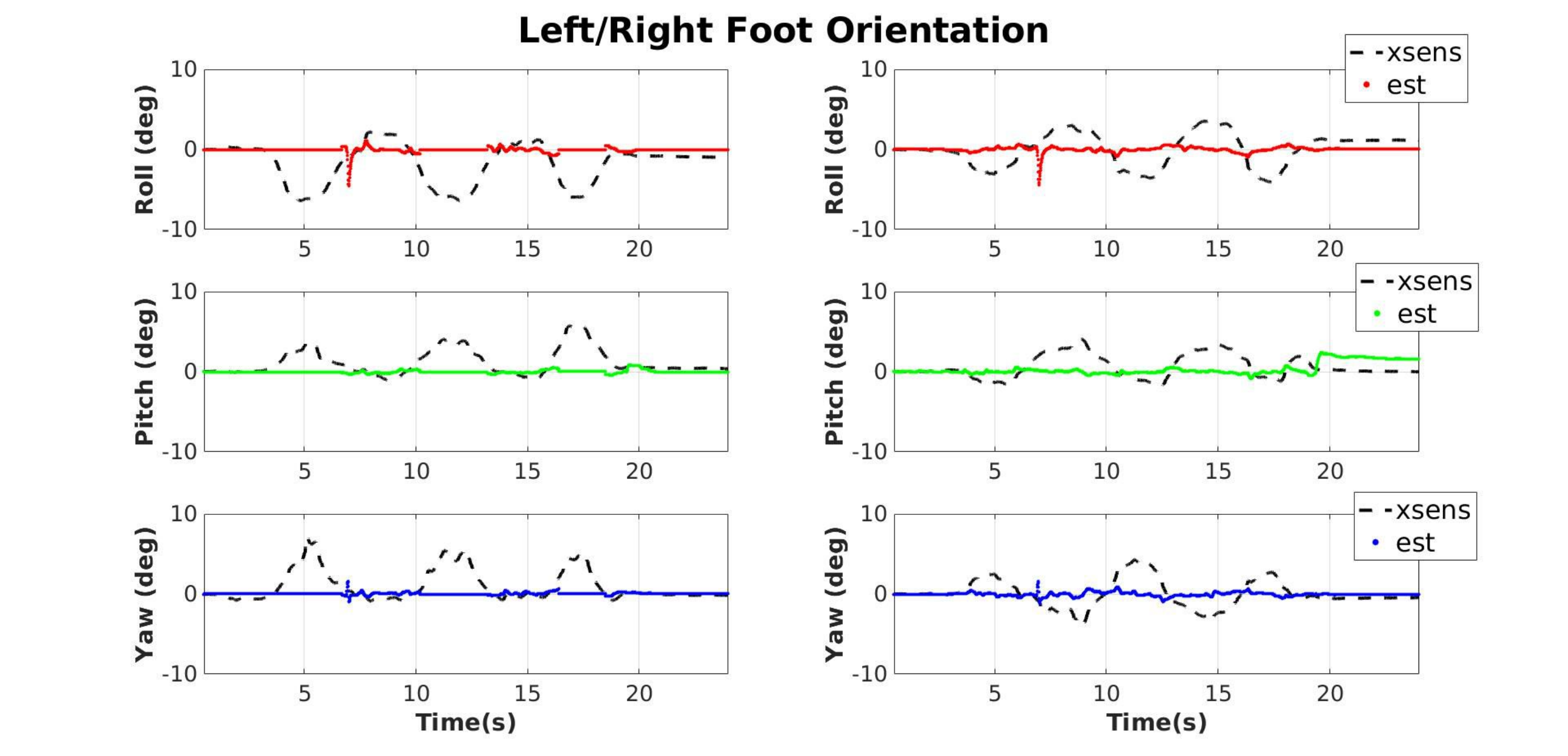}
	\end{subfigure}	
\caption{Base pose and velocity estimates along with feet rotation estimates for human  in-place swinging experiment.}
\label{fig:chap:hbe:human-swing-base-state}
\end{figure}

Similar to the CoM sinusoidal trajectory tracking experiment conducted on the robot in Chapter \ref{chap:floating-base-diligent-kio}, the human subject performs an in-place swinging experiment.
In this experiment, the subject maintains rigid foot contact with the AMTI force plates while swinging only their Pelvis link by shifting their center of mass to stay over one foot to the other.
The sequence of actual subject motion and the reconstructed motion is depicted in Figure \ref{fig:chap:hbe:human-swing-evol} and the comparison with XSens measured quantities are shown in Figure \ref{fig:chap:hbe:human-swing-base-state}.

\begin{figure}[!h]
	\centering
	\includegraphics[scale=0.05, width=\textwidth]{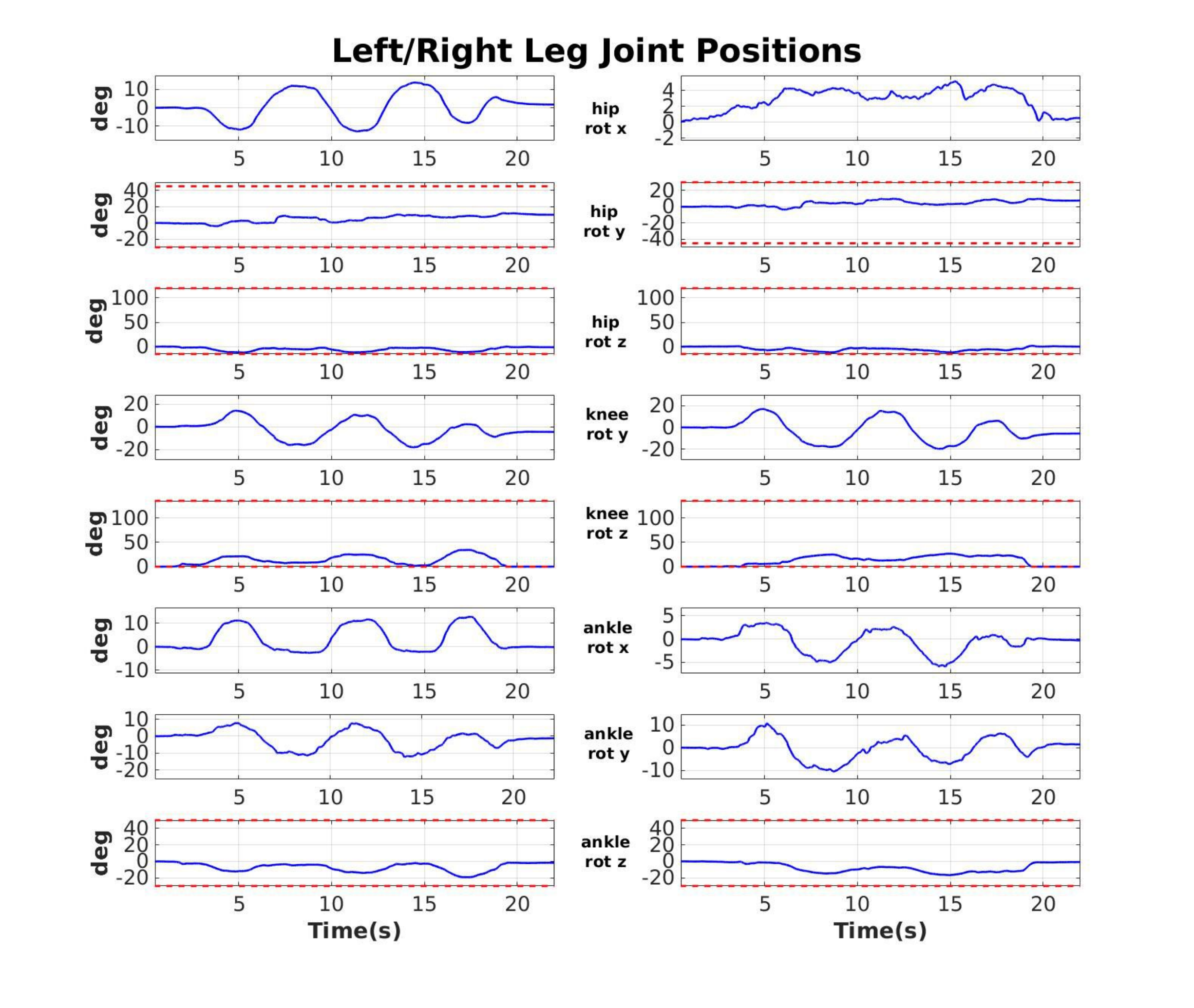}
	\caption{Joint positions of left and right leg estimated by the Dynamical IK shown as a blue line for the in-place swinging experiment. Red dotted lines indicate the joint limits obtained from the URDF model. There is a clear asymmetry in the hip and ankle \emph{RotX} joints of left and right leg. \looseness=-1}
	\label{fig:chap:hbe:human-swing-leg-joints}
\end{figure}

It can be noticed that the contact plane orientation update enforces the foot orientation to not change throughout the experiments which is the nominal orientation of the foot during the swinging experiment.
The Xsens measured foot orientation depicts a considerable rotation of the foot which is not observed during the experiment.
This implies that the whole body will be rotated according to the rotated foot when considering the XSens trajectory. 
The main joints that need to be excited during such a motion are the \emph{RotX} joints of the hip and the ankle in order shift the CoM between the feet while keeping them fixed owing to the constraints introduced by the full planar surface contact.
The reconstructed motion however suffers from the problem of sliding resulting in a displacement of the subject position in their final state. 
This is noticeable as drifts in the base position estimates as seen in Figure \ref{fig:chap:hbe:human-swing-base-state}.
This is clearly due to an asymmetry in the estimated joint positions of the left and right leg. \looseness=-1

As it can be noticed in Figure \ref{fig:chap:hbe:human-swing-leg-joints}, the \emph{RotX} joints of the hip and the ankle are not equally excited as those in the left leg which are used to reconstruct the base motion, thereby causing the final position of the base to be shifted in the $y$-direction.
Similarly, high magnitudes of change is observed in the \emph{RotZ} joints of the knee and the ankle which is physically unrealistic for the human.
These issues in the estimated joint states that are fed as inputs to the base estimation algorithm causes the overall base state to diverge.
This can be fixed by considering constraints on the limits of \emph{RotZ} joints of the knee and ankle to zero and allowing a full range of motion on the \emph{RotX} joints of the hip and ankle.
Clearly, higher accuracy estimates can be obtained by replacing the pseudo-inverse based IK solution with a QP based solution which allow for such constraints in a straight-forward manner.
\looseness=-1
\section{Conclusion}
\label{sec:chap:human-motion:conclusion}
This chapter presented a cascaded approach for the joint and base state estimation of a rigid multibody system without relying on kinematic sensor measurements such as encoders.
The presented methodology uses a dynamical optimization approach for joint state estimation and filtering over Lie groups for base estimation. 
The dynamical optimization allows for constant-time computations while retaining very fast convergence to the true states and its combination with the EKF remains computationally efficient alternative for whole-body state estimation.
The overall methodology of the dynamical IK and the EKF for the base estimation relies on error definitions and uncertainty representations using the theory of matrix Lie groups.
A whole-body human motion estimation was achieved using a set of distributed inertial sensors and force-torque measurements obtained from sensorized shoes.
In particular, this method is suitable for achieving proprioceptive whole-body kinematics estimation in the absence of position sensors.
The presented approach was also demonstrated for its application of robot state estimation.
Such a method can be used for building low-cost prototypes of the motion capture suits which can be suitably customized for several human-robot interaction experiments. \looseness=-1

Several improvements can be suggested for the proposed method.
An immediate improvement will account for the use of a QP solver for the inverse differential kinematics accounting for the joint limits of the model within the problem formulation.
Extending the Kalman filtering towards a constrained Kalman filter to account for the friction cone constraints might considerably reduce the slipping or sliding of the subject.
The CoP based contact detection can be further improved to also account for kinematic quantities such as vertex acceleration, velocity, and height along with the forces through a probabilistic Bayes filter, leading to a more precise inference of contact states.
Suitable prediction models can be obtained through motion prediction algorithms constructed using recurrent neural networks or differentiable filter architectures.





\chapter*{Epilogue}
\addcontentsline{toc}{chapter}{Epilogue}

Although, scientists working on humanoid robots acknowledge the critical requirement of state estimation algorithms for augmenting their feedback controllers, very little effort has been taken in the community towards the investigation of rigorous state estimation methodologies that deal with the several complexities of the system under consideration.
Even now, many humanoid robots tend to rely either only on legged odometry or planned trajectories as feedback for implementing locomotion controllers, which may be suitably sufficient for the application under consideration but also constrains and shadows the benefits that a reliable state estimation strategy might offer.
Only recently, owing to the fast-paced commercial development of legged robots, the interest in employing theoretically rigorous strategies for state estimation has surged.
This pattern is also starting to trickle into the humanoid community. \looseness=-1

In this context, this thesis attempted to investigate the theory of matrix Lie groups for the state estimation of rigid multibody systems in the subtopic of kinematic estimation for human motion and humanoid locomotion.
The notion of spatial uncertainties is represented using concentrated Gaussian distributions over matrix Lie groups. 
Such a notion is used to construct simple averaging estimators and extended Kalman filters for the state estimation problems. \looseness=-1

Chapter \ref{chap:floating-base-swa} presented a loosely-coupled sensor fusion approach for the estimation of the floating base state of a humanoid robot.
This approach relied on proprioceptive measurements from encoders, contact wrenches, and a base-collocated inertial measurement unit.
The problem of how to average on matrix Lie groups to obtain a reasonable estimate of several samples of complex objects such as poses and rotations was studied in the context of floating base estimation. \looseness=-1

A tightly-coupled sensor fusion approach was suggested in Chapter \ref{chap:floating-base-diligent-kio}, again for the floating base estimation of the humanoid robot relying on a similar measurement set, specifically designed to handle the flat-foot constraints of a humanoid robot.
This estimator design is suitable for applications in which both the state and measurements evolve over distinct matrix Lie groups.
The question of how the choice of error affects the filter equations was also investigated in this chapter.
It can be noticed that the non-commutative nature of the matrix Lie groups leads to different choices of error.
This modeling choice is important to be considered in the EKF design which may or may not lead to a linearized system with desirable properties.
The resulting linearized systems affect the accuracy and the stability properties of the estimator.

In Chapter \ref{chap:human-motion}, we extend our pursuit of state estimation for humanoid locomotion towards whole-body human motion estimation.
A kinematic sensor-free approach for joint state and base state estimation is presented for reconstructing the whole-body human motion simplified using rigid body models and primitive geometries.
The problem of translating measurements from a set of distributed inertial sensors and sensorized shoes into a whole-body motion estimation is achieved through a combination of dynamical optimization-based inverse kinematics, a center of pressure-based contact detection, and filtering over matrix Lie groups.
This model-based approach is suitable for any highly articulated mechanical system modeled using the theory of rigid multibody systems.

As remarked in Chapter \ref{chap:intro}, the investigations were driven towards achieving a reliable human-humanoid collaboration.
To make the state estimation infrastructure for humanoid robots proposed in this thesis holistic, it is important to augment the proposed methodologies with three important modules.
A requirement for an elaborate \emph{perception system} is necessary to complement the contact-aided kinematic-inertial odometry systems to reduce the overall drift in the state estimates for long-term operation and autonomy.
This may involve tracking visual features using geometric primitives or using learned features through convolutional neural networks.
Secondly, a precise \emph{contact detection} strategy combining both dynamic and kinematic quantities to infer the contact state of the end-effectors will not only benefit the state estimation for the robot but also enable advanced disturbance-rejection planning for humanoid control allowing the robot to locomote compliant terrains effectively.
A significant research consideration could be to investigate how to properly incorporate the effect of contacts and impacts within the prediction models of the filtering algorithms and how uncertainties can be efficiently propagated during such discontinuous events through a well-defined jump map.
This requires a study in the intersection of hybrid dynamical systems and impact-invariant state estimation.
The final module is the \emph{local terrain mapping} through active perception which complements both contact detection and floating base estimation of the robot.
Local terrain maps obtained either through probabilistic grid maps or sparse Gaussian process regression methods in a streaming setting, using proprioceptive or exteroceptive measurements or a combination of both, could effectively improve the estimation algorithms for the robot.
The precision of contact detection can be improved with the knowledge of the distance of the foot with respect to the terrain. \looseness=-1

Concerning the human motion estimation, we presented a methodology that relied on a partly deterministic approach for joint state estimation and a partly probabilistic approach for the base estimation considering information from a large set of sensors.
To democratize such a system, it is necessary to rely on a reduced count of low-cost inertial sensors.
This might require a fully probabilistic approach for both joint state estimation and base estimation properly accounting for the noise characteristics of the low-cost sensors.
Further, we considered rotation measurements from the motion capture suit which are obtained as a result of fusing accelerometer, gyroscope, and magnetometer measurements.
These systems, not only pertained to but generally are operated in indoor environments where they are subject to ambient magnetic disturbances affecting the magnetometer measurements leading to an undesired drift in the fused rotations.
It is thus necessary to investigate the handling of the effects of such magnetic disturbances or the effect of not considering the magnetometer measurements within the overall human motion estimation method.
Further, for quick operational deployment of such systems, it is also necessary to investigate rapid multi-sensor calibration algorithms.
Finally, the estimation strategy presented in this thesis requires the complementarity of inertial sensing for motion estimation and force-torque sensing for contact state inference.
To make the estimation strategy self-contained to the use of only distributed inertial sensors, a possible research direction would require the investigation of contact state inference using the inertial sensors equipped at the end-effectors. \looseness=-1

Building such elaborate architectures and strategies for achieving a reliable state estimation for reconstructing human motion and augmenting humanoid locomotion is vital for the advancements in physical human-robot interactions and collaborations.

\begin{appendices} 



\chapter{Examples of Matrix Lie Groups} 
In this appendix, we go through the definitions of the operators of matrix Lie groups such as group of rotations $\SO{3}$,  group of rigid-body transformations $\SE{3}$, group of extended poses $\SEk{2}{3}$ and group of translations $\T{n}$, which are relevant for the derivations within the scope of this thesis. \looseness=-1

\label{appendix:chapter02:examples-matrix-lie-groups}

\section[Group of 3D Rotations]{Group of 3D Rotations: $\SO{3}$}
The group of rotations are the set of $3 \times 3$ orthonormal matrices with positive unit determinant,
\begin{equation}
\label{eq:chap02:examples-so3-definition}
    \SO{3} = \left\{ \Rot{}{} \subset \textup{GL}(3, \R) \in \R^{3 \times 3} \mid \Rot{}{} \Rot{}{}^T\ =\ \Rot{}{}^T \Rot{}{}\ =\ \I{3},\ \det{\Rot{}{}} = +1  \right\}.
\end{equation}
$\textup{GL}(3, \R)$ denotes the general linear group of $3 \times 3$ invertible matrix in the domain of real numbers, whose binary operation is an ordinary matrix multiplication.

\subsubsection{Lie Algebra}
Consider ${\phi} \triangleq \text{vec}\left(\phi_1, \phi_2, \phi_3\right) \in \mathbb{R}^3$, whose axis-angle representation can be defined with an angle $\theta =\norm{{\phi}}$ and an axis $\mathbf{a} = {\phi}/\theta$.
The hat operator $\ghat{\SO{3}}{.}$ or $S(.)$ then transports the vector $\phi$ to the Lie algebra of $\SO{3}$,
\begin{align}
\label{eq:chap02:examples-so3-hat}
\begin{split}
\ghat{\SO{3}}{{\phi}} = S\left({\phi}\right) =
\begin{bmatrix}
0 & -\phi_3 & \phi_2 \\
\phi_3 & 0 & -\phi_1 \\
-\phi_2 & \phi_1 & 0
\end{bmatrix} \in \so{3} \subset \mathbb{R}^{3 \times 3}.
\end{split}
\end{align}

\subsubsection{Exponential map}
The exponential map of $\SO{3}$ can be obtained from the expansion of the matrix exponential series and a suitable substitution with the definitions of sine and cosine series expansions of real numbers, in this case that of $\theta$.
It is also the same as the Rodrigues' rotation formula.
The exponential map of $\SO{3}$ transports elements from the $\so{3}$ space of skew symmetric matrices to produce a rotation element in $\SO{3}$,
\begin{align}
\label{eq:chap02:examples-so3-expmap}
    \begin{split}
        \gexphat{\SO{3}}\left({\phi}\right) &= \sum_{k = 0}^\infty \frac{(S(\theta \mathbf{a}))^k}{k !}\\
        &= \I{3}\ + \frac{\sin\theta}{\theta}\ S\left({\phi}\right)\ + \frac{1 - \cos\theta}{\theta^2}\ S\left({\phi}\right)^2 \\
        &= \cos\theta\ \I{3}\ +\ \sin\theta\ S\left(\mathbf{a}\right)\ +\ \left(1 - \cos\theta\right)\  \mathbf{a} \mathbf{a}^T.
    \end{split} 
\end{align}
The final equivalence is obtained using the following identity of skew-symmetric matrices, $\big(S(\phi)\big)^2 = \phi \phi^T - \theta \I{3}$.

\subsubsection{Logarithm map}
The logarithm map of $\SO{3}$ followed by the vee operator that transports the elements of $\SO{3}$ to vector representing the Lie algebra $\so{3}$ can be computed as,
\begin{align}
\label{eq:chap02:examples-so3-logmap}
    \begin{split}
        & \glogvee{\SO{3}}\left(\Rot{}{}\right) = \gvee{\SO{3}}{\frac{\theta}{2 \sin\theta}(\Rot{}{} - \Rot{}{}^T)},  \\
        & \cos\theta = \frac{\text{tr}\left(\Rot{}{}\right) - 1}{2}.
    \end{split}
\end{align}
If $\Rot{}{} = \I{3} \implies \theta = 0$, then $\glogvee{\SO{3}}(\Rot{}{})$ is an undetermined system which has infinite solutions. In this case, the axis corresponding to the skew-symmetric matrix can be chosen arbitrarily. \looseness=-1

\subsubsection{Adjoint}
The adjoint matrix operator of $\SO{3}$ can be obtained from the definition of the adjoint in Eq. \eqref{eq:chap02-liegroups-adjoint-repr},
$$\gadj{\Rot{}{}}{\phi} = \gvee{\SO{3}}{\Rot{}{} S\left({\phi} \right)\Rot{}{}^T} = \gvee{\SO{3})}{S\left(\Rot{}{} {\phi}\right)} = \Rot{}{} {\phi}, $$
\begin{equation}
\label{eq:chap02:examples-so3-adj}
    \gadj{\Rot{}{}} = \Rot{}{}.
\end{equation}

\subsubsection{Jacobians}
The Jacobians of $\SO{3}$ admit closed-form solutions \cite[Section 7.1.5, Page 233]{barfoot2017state}, \looseness=-1
\begin{align}
    \begin{split}
    \label{eq:chap02:examples-so3-ljac}
         \gljac{\SO{3}}\left({\phi}\right) &=  \sum_{k=0}^\infty \frac{1}{(k+1)!}(S(\phi))^k \\
        & = \frac{\sin\theta}{\theta} \I{3} + \left(1 -\frac{\sin\theta}{\theta}\right) \mathbf{a} \mathbf{a}^T + \frac{1 - \cos \theta}{\theta} S(\mathbf{a}) \\
        & \approx \I{3} + \frac{1}{2} S(\phi),
    \end{split}
\end{align}
\begin{align}
    \begin{split}
    \label{eq:chap02:examples-so3-ljacinv}
         ({\gljac{\SO{3}}}\left({\phi}\right))^{-1} &=  \frac{\theta}{2} \cot \frac{\theta}{2} \I{3} + \left(1 -  \frac{\theta}{2} \cot \frac{\theta}{2}\right) \mathbf{a} \mathbf{a}^T - \frac{\theta}{2} S(\mathbf{a}),
    \end{split}
\end{align}
\begin{align}
\begin{split}
    \label{eq:chap02:examples-so3-rjac}
         \grjac{\SO{3}}\left({\phi}\right) &= \sum_{k=0}^\infty \frac{1}{(k+1)!}(-S(\phi))^k  \\
        &= \frac{\sin\theta}{\theta} \I{3} + \left(1 -\frac{\sin\theta}{\theta}\right) \mathbf{a} \mathbf{a}^T - \frac{1 - \cos \theta}{\theta} S(\mathbf{a}) \\
        & \approx \I{3} - \frac{1}{2} S(\phi),
    \end{split}
\end{align}
\begin{align}
    \begin{split}
    \label{eq:chap02:examples-so3-rjacinv}
        ({\grjac{\SO{3}}}\left({\phi}\right))^{-1} &= \frac{\theta}{2} \cot \frac{\theta}{2} \I{3} + \left(1 -  \frac{\theta}{2} \cot \frac{\theta}{2}\right) \mathbf{a} \mathbf{a}^T + \frac{\theta}{2} S(\mathbf{a}) .
    \end{split}
\end{align}
It must be noted that the matrices $({\grjac{\SO{3}}}\left({\phi}\right))^{-1}$ and $({\gljac{\SO{3}}}\left({\phi}\right))^{-1}$ are singular at $\theta = 2 \pi m$ with $m$ a non-zero integer due to the presence of $\cot(\frac{\theta}{2})$ functions.
Several relationships can be drawn given the definitions, such as,
\begin{align}
      & \gljac{\SO{3}}\left(-{\phi}\right) = \grjac{\SO{3}}\left({\phi}\right), \\
      &  \gljac{\SO{3}}\left({\phi}\right)  = \gadj{\Rot{}{}}\ \grjac{\SO{3}}\left({\phi}\right) = \Rot{}{}\ \grjac{\SO{3}}\left({\phi}\right), \\
      & \gljac{\SO{3}}\left({\phi}\right) = {\grjac{\SO{3}}}^T\left({\phi}\right) \quad (\text{due to skew-symmetric property}), \\
      & ({\gljac{\SO{3}}}\left({\phi}\right))^{-1} = ({\grjac{\SO{3}}}\left({\phi}\right))^{-T} \quad (\text{due to skew-symmetric property}).
\end{align}

\section[Group of Direct Spatial Isometries]{Group of Direct Spatial Isometries: $\SE{3}$}
The group of rigid body transformations or direct spatial isometries are constructed with a rotation matrix and a translation vector in semi-direct product with the rotation matrix.
These are also called as the set of homogeneous transformation matrices,
\begin{equation}
\label{eq:chap02:examples-se3-definition}
\SE{3} = \left\{ \Transform{}{} = \begin{bmatrix}
\Rot{}{} & \mathbf{p} \\ \Zeros{1}{3} & 1
\end{bmatrix} \subset \textup{GL}(4, \R) \in \R^{4 \times 4} \mid \Rot{}{} \in \SO{3}, \mathbf{p} \in \R^3  \right\}.
\end{equation}

\subsubsection{Lie Algebra}
Given a vector $\twist = \begin{bmatrix}
\rho \\ \phi
\end{bmatrix} \in \R^6$, the Lie algebra of $\SE{3}$ is given by,
\begin{align}
\label{eq:chap02:examples-se3-hat}
\ghat{\SE{3}}{\twist} = \begin{bmatrix}
S(\phi) & \rho \\ \Zeros{1}{3} & 0
\end{bmatrix} \in \se{3} \subset \R^{4 \times 4}.
\end{align}

\subsubsection{Exponential map}
The exponential map of $\SE{3}$ can be obtained from the series expansion of the matrix exponential of the Lie algebra element $\se{3}$,
\begin{align}
\label{eq:chap02:examples-se3-expmap}
    \begin{split}
        \gexphat{\SE{3}}\left({\twist}\right) &= \sum_{k = 0}^\infty \frac{(\ghat{\SE{3}}{\twist})^k}{k !}\\
        &= \sum_{k = 0}^\infty \frac{1}{k !} \begin{bmatrix}
        S(\phi) & \rho \\ \Zeros{1}{3} & 0
        \end{bmatrix}^k \\
        &= \begin{bmatrix}
        \sum_{k = 0}^\infty \frac{(S(\phi))^k}{k !} & \left(\sum_{k = 0}^\infty \frac{(S(\phi))^k}{(k+1)!}\right) \rho \\ \Zeros{1}{3} & 1 
        \end{bmatrix} \\
        &= \begin{bmatrix}
        \gexphat{\SO{3}}\left({\phi}\right) & \gljac{\SO{3}}\left({\phi}\right) \rho \\ \Zeros{1}{3} & 1 
        \end{bmatrix}.
    \end{split} 
\end{align}
As a consequence of the expansion, the exponential map of $\SE{3}$ is directly related to the exponential map and the left Jacobian of $\SO{3}$.
The coupling effect between the rotation and the translation elements due to the semi-direct product is captured by the occurrence of the left Jacobian matrix of $\SO{3}$.
If one can imagine a rigid body making a sharp turn of 90 degrees, the effect that the Jacobian has on this turning motion will make the motion more elastic due to the inherent continuum motion in $\SE{3}$.
The left Jacobian will not appear in the case of $\SO{3} \times \R^3$, where the rotation and the translations are decoupled causing the motion to be a chained roto-translation (See \cite[Example 7]{sola2018micro}). \looseness=-1

\subsubsection{Logarithm map}
Consider  $\phi = \glogvee{\SO{3}}\left(\Rot{}{}\right)$.  
The logarithm of $\SE{3}$ followed by the vee operator is given as,
\begin{align}
\label{eq:chap02:examples-se3-logmap}
& \glogvee{\SE{3}}\left(\Transform{}{}\right) =  \begin{bmatrix}
({\gljac{\SO{3}}}(\phi))^{-1} \rho \\
\phi
\end{bmatrix}.
\end{align}

\subsubsection{Adjoint}
The structure of the adjoint matrix operator of $\SE{3}$ depends on the serialization of the vectors related to the tangent space. 
In our case, $\twist = \text{vec}(\rho, \phi) \in \R^6$, this leads to the adjoint matrix operator as,
\begin{align}
\label{eq:chap02:examples-se3-adj}
\gadj{\Transform{}{}} = \begin{bmatrix}
\Rot{}{} & S(\mathbf{p}) \Rot{}{} \\ \Zero{3} & \Rot{}{}
\end{bmatrix} \in \R^{6 \times 6}.
\end{align}
For more details, refer to Section \ref{sec:chap03:rigid-body-adjoint-matrix}.

Consider $\A = \glogvee{\SE{3}}\left(\Transform{}{}\right)$, then the small adjoint matrix is given as,
$$
\gsmalladj{\A} = \begin{bmatrix}
         S(\phi) & S(\rho) \\ \Zero{3} & S(\phi)
         \end{bmatrix} \in \R^{6 \times 6}.
$$

\subsubsection{Jacobians}
Consider ${\grjac{\SO{3}}} \triangleq {\grjac{\SO{3}}}\left({\phi}\right)$ and ${\gljac{\SO{3}}} \triangleq {\gljac{\SO{3}}}\left({\phi}\right)$. The Jacobians of $\SE{3}$ are defined as, \looseness=-1
\begin{align}
    \begin{split}
    \label{eq:chap02:examples-se3-ljac}
         \gljac{\SE{3}}\left({\twist}\right) &=  \sum_{k=0}^\infty \frac{1}{(k+1)!}\left(\begin{bmatrix}
         S(\phi) & S(\rho) \\ \Zero{3} & S(\phi)
         \end{bmatrix}\right)^k = \begin{bmatrix}
         {\gljac{\SO{3}}} & \Q_l \\ \Zero{3} & {\gljac{\SO{3}}}
         \end{bmatrix},
    \end{split}
\end{align}
where,
\begin{align}
    \begin{split}
        \Q_l &\triangleq \Q_l(\twist) = \sum_{n=0}^\infty\ \sum_{m=0}^\infty \frac{1}{(n+m+2)!} (S(\phi))^n \ S(\rho)\  S(\phi)^m \\
        &= \frac{1}{2} S(\rho) + \left(\frac{\theta - \sin \theta}{\theta^3}\right) \left(S(\phi)\ S(\rho) + S(\rho)\ S(\phi) + S(\phi)\ S(\rho)\ S(\phi)\right)  \\
        & \qquad + \left(\frac{\theta^2 + 2 \cos \theta - 2}{2\theta^4}\right) \left(S(\phi)\ S(\phi)\ S(\rho) + S(\rho)\ S(\phi)\ S(\phi)\ - 3\ S(\phi)\ S(\rho)\ S(\phi)\right) \\
        & \qquad +  \left(\frac{2 \theta - 3 \sin \theta + \theta \cos \theta}{2\theta^5}\right) \left(S(\phi)\ S(\rho)\ S(\phi)\ S(\phi) + S(\phi)\ S(\phi)\ S(\rho)\ S(\phi)\right),
    \end{split}
\end{align}
\begin{align}
    \begin{split}
    \label{eq:chap02:examples-se3-ljacinv}
        ({\gljac{\SE{3}}}\left({\twist}\right))^{-1} =  \begin{bmatrix}
        ({\gljac{\SO{3}}})^{-1} & -({\gljac{\SO{3}}})^{-1} {\Q_l}^{-1} ({\gljac{\SO{3}}})^{-1} \\
        \Zero{3} & ({\gljac{\SO{3}}})^{-1}
        \end{bmatrix},
    \end{split}
\end{align}
\begin{align}
\begin{split}
    \label{eq:chap02:examples-se3-rjac}
         \grjac{\SE{3}}\left({\twist}\right) &= \sum_{k=0}^\infty \frac{1}{(k+1)!}\left(-\begin{bmatrix}
         S(\phi) & S(\rho) \\ \Zero{3} & S(\phi)
         \end{bmatrix}\right)^k \\
         &= \begin{bmatrix}
         {\grjac{\SO{3}}} & \Q_r \\ \Zero{3} & {\grjac{\SO{3}}}
         \end{bmatrix},
    \end{split}
\end{align}
\begin{align}
    \begin{split}
    \label{eq:chap02:examples-se3-rjacinv}
        ({\grjac{\SE{3}}}\left({\twist}\right))^{-1} =  \begin{bmatrix}
        ({\grjac{\SO{3}}})^{-1} & -({\grjac{\SO{3}}})^{-1} {\Q_r}^{-1} ({\grjac{\SO{3}}})^{-1} \\
        \Zero{3} & ({\grjac{\SO{3}}})^{-1}
        \end{bmatrix}.
    \end{split}
\end{align}
It must be noted that $ \Q_r \triangleq \Q_r(\twist) = \Q_l(-\twist)$.
Many other interesting details about the Jacobians of $\SO{3}$ and $\SE{3}$ along with closed-form expressions for Jacobian inverse, alternative formulations using series expansion of the small adjoint matrices are discussed in detail in \cite[Chapter 7]{barfoot2017state}.
These details could be quite useful for carrying out higher-order numerical integration on $\SO{3}$ and $\SE{3}$, for example Runge-Kutta Munthe-Kaas (RKMK4) based integration methods on Lie groups.

\section[Group of Double-Direct Spatial Isometries]{Group of Double-Direct Spatial Isometries: $\SEk{2}{3}$}
The group of double direct spatial isometries are constructed with a rotation matrix, and two translation  vectors in semi-direct product with the rotation matrix.
These groups are well-suited for a direct state representation for an inertial measurement unit for the estimation of orientation $\Rot{}{}$, position $\mathbf{p}$ and velocity $\mathbf{v}$ of a rigid body in space.
These are called as the set of extended poses by \cite{barrau2015non},
\begin{equation}
\label{eq:chap02:examples-se23-definition}
\SEk{2}{3} = \left\{ \X = \begin{bmatrix}
\Rot{}{} & \mathbf{p} & \mathbf{v} \\ \Zeros{1}{3} & 1 & 0  \\ \Zeros{1}{3} & 0 & 1
\end{bmatrix} \subset \textup{GL}(5, \R) \in \R^{5 \times 5} \mid \Rot{}{} \in \SO{3}, \mathbf{p}, \mathbf{v} \in \R^3  \right\}.
\end{equation}
\subsubsection{Lie Algebra}
Given a vector $\tau = \text{vec}(\rho, \phi, \psi) \in \R^9$, the Lie algebra of $\SEk{2}{3}$ is given by,
\begin{align}
\label{eq:chap02:examples-se23-hat}
\ghat{\SE{3}}{\twist} = \begin{bmatrix}
S(\phi) & \rho & \psi \\ \Zeros{1}{3} & 0 & 0 \\ \Zeros{1}{3} & 0 & 0
\end{bmatrix} \in \sek{2}{3} \subset \R^{5 \times 5}.
\end{align}
\subsubsection{Exponential map}
The exponential map of $\SEk{2}{3}$ is similar to that of $\SE{3}$,
\begin{align}
\label{eq:chap02:examples-se23-expmap}
 \gexphat{\SEk{2}{3}}\left({\tau}\right) = \begin{bmatrix}
        \gexphat{\SO{3}}\left({\phi}\right) & \gljac{\SO{3}}\left({\phi}\right) \rho & \gljac{\SO{3}}\left({\phi}\right) \psi  \\ 
        \Zeros{1}{3} & 1 & 0 \\
        \Zeros{1}{3} & 0 & 1
        \end{bmatrix}.
\end{align}
\subsubsection{Logarithm map}
Consider  $\phi = \glogvee{\SO{3}}\left(\Rot{}{}\right)$.  
The logarithm of $\SEk{2}{3}$ followed by the vee operator is given as, \looseness=-1
\begin{align}
\label{eq:chap02:examples-se23-logmap}
& \glogvee{\SEk{2}{3}}\left(\X\right) =  \begin{bmatrix}
({\gljac{\SO{3}}}(\phi))^{-1} \rho \\
\phi \\
({\gljac{\SO{3}}}(\phi))^{-1} \psi
\end{bmatrix}.
\end{align}

\subsubsection{Adjoint}
Given the choice of the vector serialization in $\tau$, the adjoint matrix operator as,
\begin{align}
\label{eq:chap02:examples-se23-adj}
\gadj{\X} = \begin{bmatrix}
\Rot{}{} & S(\mathbf{p}) \Rot{}{} & \Zero{3}\\ 
\Zero{3} & \Rot{}{} & \Zero{3} \\
\Zero{3} & S(\mathbf{p}) \Rot{}{} &  \Rot{}{} 
\end{bmatrix} \in \R^{9 \times 9}.
\end{align}
Note that, if the angular part $\phi$ was placed on the top in the vector $\tau$ followed by the linear parts, then the adjoint matrix will follow a triangular structure, which might be useful for computational gains while performing matrix inversions. 
This is true also for the Jacobian computations for the Lie groups and its associated inverses.
However, in this thesis, we will follow the linear-angular serialization.

\subsubsection{Jacobians}
Consider ${\grjac{\SO{3}}} \triangleq {\grjac{\SO{3}}}\left({\phi}\right)$, ${\gljac{\SO{3}}} \triangleq {\gljac{\SO{3}}}\left({\phi}\right)$, $\mathbf{v}_1 = \text{vec}(
\rho, \phi)$ and  $\mathbf{v}_2 =\text{vec}(\psi, \phi)$. The Jacobians of $\SEk{2}{3}$ are given as,
\begin{align}
    \label{eq:chap02:examples-se23-ljac}
        & \gljac{\SEk{2}{3}}\left({\tau}\right) =  \begin{bmatrix}
        {\gljac{\SO{3}}} & \Q_l(\mathbf{v}_1) & \Zero{3} \\
        \Zero{3} & {\gljac{\SO{3}}} & \Zero{3} \\
         \Zero{3} & \Q_l(\mathbf{v}_2) & {\gljac{\SO{3}}}
        \end{bmatrix}, \\
    \label{eq:chap02:examples-se23-rjac}
        & \grjac{\SEk{2}{3}}\left({\tau}\right) =  \begin{bmatrix}
        {\grjac{\SO{3}}} & \Q_r(\mathbf{v}_1) & \Zero{3} \\
        \Zero{3} & {\grjac{\SO{3}}} & \Zero{3} \\
         \Zero{3} & \Q_r(\mathbf{v}_2) & {\grjac{\SO{3}}}
        \end{bmatrix}.
\end{align}
Since, the structure of Jacobians of $\SEk{2}{3}$ are not block-triangular like those of $\SE{3}$, for computation of inverses, we apply a brute force matrix inversion.
With the alternative serialization of angular-linear, it could be possible to exploit the block matrix inversion lemmas. \looseness=-1

It must be noted that the group of double-direct spatial isometries $\SEk{2}{3}$ can be extended to $k$-direct spatial isometries $\SEk{k}{3}$ in a straightforward manner.
In fact, when $k=1$, the group of $n$-direct spatial isometries reduces to group of rigid body transformations $\SEk{1}{3} \sim \SE{3}$, when $k=2$ we get the group of extended poses and with aribtrary number of vectors we have the so-called SLAM manifold $\SEk{k}{3} \sim \text{SLAM}_k(3)$ (\cite{mahony2017geometric}).

\section[Group of Translations]{Group of Translations: $\T{n}$}
The translations group represent the Euclidean vector space embedded as  matrix Lie groups, \looseness=-1
\begin{equation}
\label{eq:chap02:examples-tn-definition}
\T{n} = \left\{ \mathbf{T} = \begin{bmatrix}
\I{n} & \mathbf{t} \\ \Zeros{1}{n} & 1
\end{bmatrix} \subset \textup{GL}(n+1, \R) \in \R^{n+1 \times n+1} \mid \mathbf{t} \in \R^n  \right\}.
\end{equation}
It can be seen as the vectors living in semi-direct product with identity matrices. 
For intuition. when considering three-dimensional spaces, if the rotation is identity, we only have a translation in a geometric sense.
This idea can be extended to an arbitrary number of dimensions.
Effectively, without the context of matrix Lie groups, they are simply a group of vectors with addition as the bi-linear operation. \looseness=-1

\subsubsection{Lie Algebra}
The Lie algebra is given as,
\begin{align}
\label{eq:chap02:examples-tn-hat}
\ghat{\T{n}}{\mathbf{t}} = \begin{bmatrix}
\Zero{n} & \mathbf{t} \\ \Zeros{1}{n} & 0
\end{bmatrix}  \subset \R^{n+1 \times n+1}.
\end{align}
\subsubsection{Exponential map}
The exponential map is trivial due to the presence of the identity matrix, \looseness=-1
\begin{align}
\label{eq:chap02:examples-tn-expmap}
 \gexphat{\T{n}}\left(\mathbf{t}\right) = \begin{bmatrix}
\I{n} & \mathbf{t} \\ \Zeros{1}{n} & 1
\end{bmatrix}.
\end{align}
\subsubsection{Logarithm map}
The logarithm map is simply,
\begin{align}
\label{eq:chap02:examples-tn-logmap}
\glogvee{\T{n}}(\mathbf{T}) = \mathbf{t}.
\end{align}
\subsubsection{Adjoint and Jacobians}
The adjoint matrix operator and the Jacobians of the translation group are all identity matrices,
\begin{align}
\label{eq:chap02:examples-tn-adj} 
        & \gadj{\mathbf{T}} = \I{n}, \\
\label{eq:chap02:examples-tn-ljac}
        & \gljac{\T{n}}\left({\mathbf{t}}\right) =  \I{n},\\
\label{eq:chap02:examples-tn-rjac}
        & \grjac{\T{n}}\left({\mathbf{t}}\right) =  \I{n}.
\end{align}

Table \ref{table:matrixliegroups} summarizes the tuple representation, composition operation along with inverse and identity elements of $\SO{3}$, $\SE{3}$, $\SEk{2}{3}$ and $\T{n}$ matrix Lie groups.
\begin{sidewaystable}[p]
	\caption{Matrix Lie groups depicting the group of rotations, poses, extended poses, and translations, respectively. \looseness=-1 }
	\label{table:matrixliegroups}
	\centering
		\begin{tabular}{|c|c|c|c|c|c|c|c|}
			\hline & & & & & & & \\
			Lie Group $G$ & Dim. $p$ & $\X \in G$ & $\err \in \R^p$ & $\ghat{G}{\err} \in \mathfrak{g}$ & $\X_1 \circ \X_2 \in G$ & $\X^{-1}  \in G$ & $\X_{I} \in G$ \\
			& & & & & & & \\
			\hline & & & & & & & \\
			$\SO{3}$  & $3$ & \scalebox{0.7}{ $\Rot{}{}$ } & \scalebox{0.7}{$\omega$} & \scalebox{0.7}{$S(\omega)$} & \scalebox{0.7}{$\Rot{}{}_1 \Rot{}{}_2$} & \scalebox{0.7}{$\Rot{}{}^T$} & \scalebox{0.7}{$\I{3}$} \\
			& & & & & & & \\
			\hline & & & & & & & \\
		$\SE{3}$  & $6$ & \scalebox{0.7}{$(\mathbf{p}, \Rot{}{}) = \begin{bmatrix}
			\Rot{}{} & \mathbf{p} \\ \Zeros{1}{3} & 1
			\end{bmatrix}$} &  \scalebox{0.7}{$\begin{bmatrix}
			\bm{v} \\ \omega
			\end{bmatrix}$} & \scalebox{0.7}{$\begin{bmatrix}
				S(\omega) & \bm{v} \\\Zeros{1}{3} & 0
			\end{bmatrix}$} & \scalebox{0.7}{$(\Rot{}{}_1 \mathbf{p}_2 + \mathbf{p}_1, \Rot{}{}_1 \Rot{}{}_2)$} & \scalebox{0.7}{$(-\Rot{}{}^T \mathbf{p}, \Rot{}{}^T)$} & \scalebox{0.7}{$(\Zeros{3}{1}, \I{3})$} \\
			& & & & & & & \\
			\hline  & & & & & & & \\
			$\SEk{2}{3}$  & $9$ & \scalebox{0.7}{$(\mathbf{p}, \Rot{}{}, \bm{v}) = \begin{bmatrix}
				\Rot{}{} & \mathbf{p} & \bm{v} \\ \Zeros{1}{3} & 1 & 0 \\ \Zeros{1}{3} & 0 & 1
				\end{bmatrix} $} &  \scalebox{0.7}{$\begin{bmatrix}
				\bm{v} \\ \omega \\ \mathbf{a}
				\end{bmatrix}$} & \scalebox{0.7}{$\begin{bmatrix}
				S(\omega) & \bm{v} & \mathbf{a} \\ \Zeros{1}{3} & 0 & 0 \\  \Zeros{1}{3} & 0 & 0
				\end{bmatrix}$} & \scalebox{0.7}{$(\Rot{}{}_1 \mathbf{p}_2 + \mathbf{p}_1, \Rot{}{}_1 \Rot{}{}_2, \Rot{}{}_1 \bm{v}_2 + \bm{v}_1)$} & \scalebox{0.7}{$(-\Rot{}{}^T \mathbf{p}, \Rot{}{}^T, -\Rot{}{}^T \bm{v})$} & \scalebox{0.7}{$(\Zeros{3}{1}, \I{3}, \Zeros{3}{1})$} \\
				& & & & & & & \\
		    \hline  & & & & & & & \\
			$\T{n}$  & $n$ & \scalebox{0.7}{$ \mathbf{T} = \begin{bmatrix}
				\I{n} & \mathbf{t} \\ \Zeros{1}{n} & 1
				\end{bmatrix}$} &  \scalebox{0.7}{$\mathbf{t}$} & \scalebox{0.7}{$\begin{bmatrix}
				\Zero{n} & \mathbf{t} \\ \Zeros{1}{n} & 0
				\end{bmatrix}$} & \scalebox{0.7}{$ \mathbf{t}_1 + \mathbf{t}_2$} & \scalebox{0.7}{$-\mathbf{t}$} & \scalebox{0.7}{$\Zeros{n}{1}$} \\
				& & & & & & & \\
			\hline
		\end{tabular}
\end{sidewaystable}

\chapter{Supplementary Filter Derivations}

\section{Left Trivialized Motion Model Jacobian of DILIGENT-KIO}
\label{appendix:chapter:left-triv-motion-model-jacobian-diligent-kio}
Given the definition of the left-trivialized motion model in Eq. \eqref{eq:chap:diligent-kio-left-triv-motion-model}, the  motion model as a function of state estimate $\Xhat$ perturbed by vector $\err$ on the left is given as,
\begin{equation}
    \Omega(\Xhat\ \gexphat{\mathscr{M}}({\err}), \mathbf{u}) = 
    \begin{bmatrix} 
       \Omega_1(\Xhat\ \gexphat{\mathscr{M}}({\err}), \mathbf{u}) \\
       \Omega_2(\Xhat\ \gexphat{\mathscr{M}}({\err}), \mathbf{u}) \\
       \Omega_3(\Xhat\ \gexphat{\mathscr{M}}({\err}), \mathbf{u}) \\
       \Zeros{18}{1}
    \end{bmatrix},
\end{equation}
where, after applying a first order approximation of the exponential map the individual non-zero terms can be derived as, \looseness=-1
\begin{align*}
\Omega_1(\Xhat\ \gexphat{\mathscr{M}}({\err}), \mathbf{u}) &=  (\mathbf{\hat{R}} +\ \mathbf{\hat{R}}\ S(\err_\mathbf{R}))^T (\mathbf{\hat{v}} + \mathbf{\hat{R}}\ \err_\mathbf{v}) \Delta T  \\
& \quad + \frac{1}{2}(\yAcc{A}{B} - \biasAccHat - \err_\biasAcc) \Delta T^2 + \frac{1}{2}(\mathbf{\hat{R}} + \mathbf{\hat{R}}\ S(\err_\mathbf{R}))^T \gravity{A} \Delta T^2 \\
&= (\mathbf{\hat{R}}^T -\ S(\err_\mathbf{R})\mathbf{\hat{R}}^T) (\mathbf{\hat{v}} + \mathbf{\hat{R}}\ \err_\mathbf{v} + \frac{1}{2} \gravity{A} \Delta T) \Delta T + \frac{1}{2}(\yAcc{A}{B} - \biasAccHat - \err_\biasAcc) \Delta T^2 \\
&\approx \Omega_1(\Xhat) + \err_\mathbf{v} \Delta T - S(\err_\mathbf{R})\big(\mathbf{\hat{R}}^T (\mathbf{\hat{v}} + \frac{1}{2} \gravity{A} \Delta T) \Delta T\big) - \frac{1}{2} \err_\biasAcc \Delta T^2 \\
&= \Omega_1(\Xhat) + \err_\mathbf{v} \Delta T + S(\mathbf{\hat{R}}^T (\mathbf{\hat{v}} \Delta T + \frac{1}{2} \gravity{A} \Delta T^2))\err_\mathbf{R} - \frac{1}{2} \err_\biasAcc \Delta T^2,
\end{align*}
\begin{align*}
\Omega_2(\Xhat\ \gexphat{\mathscr{M}}({\err}), \mathbf{u}) &=  (\yGyroBar{A}{B} - \err_\biasGyro) \Delta T \\
&= \Omega_2(\Xhat) - \err_\biasGyro \Delta T, \quad \text{and}
\end{align*}
\begin{align*}
\Omega_3(\Xhat\ \gexphat{\mathscr{M}}({\err}), \mathbf{u}) &= (\yAcc{A}{B} - \biasAccHat - \err_\biasAcc) \Delta T + (\mathbf{\hat{R}} + \mathbf{\hat{R}}\  S(\err_\mathbf{R}))^T \gravity{A} \Delta T \\
&= (\yAcc{A}{B} - \biasAccHat) \Delta T - \err_\biasAcc \Delta T + \mathbf{\hat{R}}^T \gravity{A} \Delta T -\  S(\err_\mathbf{R}))^T \mathbf{\hat{R}}^T \gravity{A} \Delta T \\
&= \Omega_3(\Xhat) + S(\mathbf{\hat{R}}^T \gravity{A} \Delta T) \err_\mathbf{R} - \err_\biasAcc \Delta T.
\end{align*}

Computing partial derivatives of $\Omega_1$ with respect to the error-state variables gives,
\begin{align*}
\frac{\partial}{\partial \err_\mathbf{R}} \Omega_1 &= S(\mathbf{\hat{R}}^T (\mathbf{\hat{v}} \Delta T + \frac{1}{2}\ \gravity{A}\ \Delta T^2)),\\
\frac{\partial}{\partial \err_\mathbf{v}} \Omega_1 &= \I{3}\ \Delta T, \\
\frac{\partial}{\partial \err_\biasAcc} \Omega_1 &= -\frac{1}{2}\I{3}\ \Delta T^2.
\end{align*}

Computing partial derivatives of $\Omega_2$ with respect to the error-state variables gives,
\begin{align*}
\frac{\partial}{\partial \err_\biasGyro} \Omega_2 &= -\I{3}\ \Delta T.
\end{align*}

Computing partial derivatives of $\Omega_3$ with respect to the error-state variables gives,
\begin{align*}
\frac{\partial}{\partial \err_\mathbf{R}} \Omega_3 &= S(\mathbf{\hat{R}}^T \gravity{A}\ \Delta T),\\
\frac{\partial}{\partial \err_\biasAcc} \Omega_3 &= -\I{3}\ \Delta T.
\end{align*}

\section{Left Trivialized Motion Model Jacobian of DILIGENT-KIO-RIE}
\label{appendix:chapter:left-triv-motion-model-jacobian-diligent-kio-rie}
The left-trivialized motion model $\Omega(\gexphat{\mathscr{M}}({\err})\ \Xhat, \mathbf{u})$ as a function state estimate $\Xhat$ perturbed by vector $\err$ on the left is given as,
\begin{equation}
    \Omega(\gexphat{\mathscr{M}}({\err})\ \Xhat, \mathbf{u}) = 
    \begin{bmatrix} 
       \Omega_1(\gexphat{\mathscr{M}}({\err})\ \Xhat, \mathbf{u}) \\
       \Omega_2(\gexphat{\mathscr{M}}({\err})\ \Xhat, \mathbf{u}) \\
       \Omega_3(\gexphat{\mathscr{M}}({\err})\ \Xhat, \mathbf{u}) \\
       \Zeros{18}{1}
    \end{bmatrix},
\end{equation}
where, after applying a first order approximation of the exponential map the individual non-zero terms can be derived as,
\begin{align*}
\Omega_1(\gexphat{\mathscr{M}}({\err})\ \Xhat, \mathbf{u}) &=  (\mathbf{\hat{R}} + S(\err_\mathbf{R})\ \mathbf{\hat{R}})^T (\mathbf{\hat{v}} + S(\mathbf{\hat{v}})\ \err_\mathbf{R} + \err_\mathbf{v}) \Delta T  \\
& + \frac{1}{2}(\yAcc{A}{B} - \biasAccHat - \err_\biasAcc) \Delta T^2 + \frac{1}{2}(\mathbf{\hat{R}} + S(\err_\mathbf{R})\ \mathbf{\hat{R}})^T \gravity{A} \Delta T^2,
\end{align*}
\begin{align*}
\Omega_2(\gexphat{\mathscr{M}}({\err})\ \Xhat, \mathbf{u}) &=  (\yGyroBar{A}{B} - \err_\biasGyro) \Delta T,  \quad \text{and}
\end{align*}
\begin{align*}
\Omega_3(\gexphat{\mathscr{M}}({\err})\ \Xhat, \mathbf{u}) &= (\yAcc{A}{B} - \biasAccHat - \err_\biasAcc) \Delta T^2 + (\mathbf{\hat{R}} + S(\err_\mathbf{R})\ \mathbf{\hat{R}})^T \gravity{A} \Delta T.
\end{align*}
Simplifying $\Omega_1$ to be expressed linearly with respect to the error vector $\err$ in order to compute the partial derivatives lead to, 
\begin{align*}
\frac{\partial}{\partial \err} \Omega_1 &= \frac{\partial}{\partial \err}\left( \left(-\mathbf{\hat{R}}^T S(\mathbf{\hat{v}}) \err_\mathbf{R}  + \mathbf{\hat{R}}^T  \err_\mathbf{v}  - \mathbf{\hat{R}}^T S(\err_\mathbf{R}) \mathbf{\hat{v}}  - \frac{1}{2} (\err_\biasAcc  +  \mathbf{\hat{R}}^T S(\err_\mathbf{R})\ \gravity{A} )\Delta T \right)\Delta T\right),
\end{align*}
which can be re-written as partial derivatives with respect to error terms along each direction,
\begin{align*}
\frac{\partial}{\partial \err_\mathbf{R}} \Omega_1 &= \frac{\partial}{\partial \err_\mathbf{R}}(-\mathbf{\hat{R}}^T S(\mathbf{\hat{v}}) \err_\mathbf{R} \Delta T - \mathbf{\hat{R}}^T\ S(\err_\mathbf{R}) \mathbf{\hat{R}} \mathbf{\hat{R}}^T (\mathbf{\hat{v}} + \frac{1}{2}\gravity{A} \Delta T) \Delta T) \\
&= \frac{\partial}{\partial \err_\mathbf{R}}(- \mathbf{\hat{R}}^T S(\mathbf{\hat{v}}) \err_\mathbf{R} \Delta T + \mathbf{\hat{R}}^T S(\mathbf{\hat{v}}) \mathbf{\hat{R}} \mathbf{\hat{R}}^T \err_\mathbf{R} \Delta T + \frac{1}{2} \mathbf{\hat{R}}^T S(\gravity{A}) \mathbf{\hat{R}} \mathbf{\hat{R}}^T \err_\mathbf{R} \Delta T^2) \\
&= \frac{1}{2} \mathbf{\hat{R}}^T S(\gravity{A}) \Delta T^2, \\
\frac{\partial}{\partial \err_\mathbf{v}} \Omega_1 &= \mathbf{\hat{R}}^T \Delta T, \\
\frac{\partial}{\partial \err_\biasAcc} \Omega_1 &= -\frac{1}{2}\I{3} \Delta T^2.
\end{align*}
It must be noted that we have used the following identity which is a consequence of the adjoint property of $\SO{3}$ in the above simplification. Given $\forall \mathbf{u}, \mathbf{v} \in \R^3, \Rot{}{} \in \SO{3}$,
\begin{align*}
{\Rot{}{}}^T S(\mathbf{u}) \mathbf{v} &= {\Rot{}{}}^T S(\mathbf{u}) \Rot{}{} {\Rot{}{}}^T \mathbf{v} \\
&= S({\Rot{}{}}^T \mathbf{u}) {\Rot{}{}}^T \mathbf{v} \\
&= - S({\Rot{}{}}^T \mathbf{v}) {\Rot{}{}}^T \mathbf{u} \\
&= - {\Rot{}{}}^T S(\mathbf{v}) \Rot{}{} {\Rot{}{}}^T \mathbf{u} \\
&= - {\Rot{}{}}^T S(\mathbf{v}) \mathbf{u}.
\end{align*}
The partial derivatives of  $\Omega_2$ with respect to the error terms $\err$ is given by, $$\frac{\partial}{\partial \err_\biasAcc} \Omega_2 = -\I{3} \Delta T.$$
Simplifying $\Omega_3$ to be expressed linearly with respect to the error vector $\err$ and computing the partial derivatives results in,
\begin{align*}
\frac{\partial}{\partial \err} \Omega_3 &= \frac{\partial}{\partial \err}(-\err_\biasAcc \Delta T  -  \mathbf{\hat{R}}^T S(\err_\mathbf{R})\ \gravity{A} \Delta T ) \\
&= \frac{\partial}{\partial \err}(-\err_\biasAcc \Delta T +   \mathbf{\hat{R}}^T S(\gravity{A}) \err_\mathbf{R} \Delta T ),
\end{align*}
resulting in the individual terms,
\begin{align*}
\frac{\partial}{\partial \err_\mathbf{R}} \Omega_3 &= \mathbf{\hat{R}}^T S(\gravity{A}) \Delta T, \\
\frac{\partial}{\partial \err_\biasAcc} \Omega_3 &= -\frac{1}{2}\I{3} \Delta T^2.
\end{align*}

\section{Measurement model Jacobian of DILIGENT-KIO-RIE}
\label{appendix:chapter:meas-model-jacobian-diligent-kio-rie}
Let's rewrite the relative forward kinematics measurement model as defined in Eq. \eqref{eq:chap:diligent-kio-meas-model},
$$
h(\Xhat) = \begin{bmatrix}
   h_R(\Xhat) & h_p(\Xhat) \\ \Zeros{1}{3} & 1
\end{bmatrix} = (h_p(\Xhat), h_R(\Xhat))_{\SE{3}} = (\mathbf{\hat{R}}^T(\mathbf{\hat{d}} - \mathbf{\hat{p}}),\ \mathbf{\hat{R}}^T\mathbf{\hat{Z}})_{\SE{3}}.
$$
The inverse is given as,
$$
h^{-1}(\Xhat) = \big(h_p(\Xhat),\  h_R(\Xhat)\big)^{-1}_{\SE{3}} = (\mathbf{\hat{Z}}^T(\mathbf{\hat{p}} - \mathbf{\hat{d}}),\ \mathbf{\hat{Z}}^T\mathbf{\hat{R}})_{\SE{3}}.
$$
A perturbation in the vector space of the state representation induces the changes in the measurement model after a first-order approximation as,
\begin{align*}
h(\gexphat{\mathscr{M}}({\err})\ \Xhat) &= \big(\mathbf{\hat{R}}^T(\mathbf{\hat{d}} - \mathbf{\hat{p}}) +  \mathbf{\hat{R}}S(\mathbf{\hat{d}})(\err_\mathbf{R} - \err_\mathbf{Z}) + \mathbf{\hat{R}}^T\ (\err_\mathbf{d} - \err_\mathbf{p}), \\
& \qquad \qquad \mathbf{\hat{R}}^T\mathbf{\hat{Z}} -  \mathbf{\hat{R}}^T S(\err_\mathbf{R})\mathbf{\hat{Z}} + \mathbf{\hat{R}}^T S(\err_\mathbf{Z}) \mathbf{\hat{Z}}  \big)_{\SE{3}}.
\end{align*}
This leads to the computation of the term $h^{-1}(\Xhat)\ h(\gexphat{\mathscr{M}}({\err})\ \Xhat)$ through a straightforward composition of transformation matrices represented in a tuple as,
\begin{align*}
\left(
\mathbf{\hat{Z}}^T S(\mathbf{\hat{d}}) (\err_\mathbf{R} - \err_\mathbf{Z}) + \mathbf{\hat{Z}}^T (
\err_\mathbf{d} - \err_\mathbf{p}),\ 
\I{3} - S(\mathbf{\hat{Z}}^T \err_\mathbf{R}) + S(\mathbf{\hat{Z}}^T \err_\mathbf{Z})
\right)_{\SE{3}}.
\end{align*}
Applying the logarithm using the approximation $\log(\X) \approx \X - \I{l}\ \forall \X \in \R^{l \times l}$ followed by the vee operator as described in Section \ref{sec:chap:diligent-kio-meas-model} leads to linear equations with respect to the error terms for which computing the partial derivatives then become straightforward.

\section{Non-linear error dynamics of CODILIGENT-KIO-RIE}
\label{appendix:chapter:nonlinear-err-dyn-cdekf-rie}

Given the definition of right invariant error in Eq. \eqref{eq:eq:chap:cdekf-sys-rie} used for the CODILIGENT-KIO-RIE, in this section, we provide a detailed derivation of the error dynamics.

\paragraph{Derivative of $\eta^R_\Rot{}{}$:}
From Eq. \eqref{eq:eq:chap:cdekf-sys-dynamics}, we have,
\begin{align*}
&\Rdot{}{} = \mathbf{R}\ S({\yGyro{A}{B}} - \biasGyro - \noiseGyro{B}), \\    
&\mathbf{\dot{\hat{R}}} = \mathbf{\hat{R}}\ S({\yGyro{A}{B}} - \biasGyroHat),
\end{align*}
and the true rotation $\mathbf{R} = \eta^R_\Rot{}{}\ \mathbf{\hat{R}} \approx \mathbf{\hat{R}} + S(\err_\mathbf{R})\ \mathbf{\hat{R}}$ using a first-order approximation of the exponential map.
Using these definitions and neglecting the cross-product of infinitesimal error and noise terms, the derivative of rotation error can be computed as,
\begin{align*}
    \dot{\eta}^R_\Rot{}{} &= \dot{\Rot{}{}} \mathbf{\hat{R}}^T + \mathbf{R} \mathbf{\dot{\hat{R}}}^T \\
    &= \Rot{}{} \ S({\yGyro{A}{B}} - \biasGyro - \noiseGyro{B})\  \mathbf{\hat{R}}^T + \mathbf{R} (\mathbf{\hat{R}}\ S({\yGyro{A}{B}} - \biasGyroHat))^T \\
    &= (\mathbf{\hat{R}} + S(\err_\mathbf{R})\ \mathbf{\hat{R}}) \ S({\yGyro{A}{B}} - \biasGyro - \noiseGyro{B})\  \mathbf{\hat{R}}^T  - (\mathbf{\hat{R}} + S(\err_\mathbf{R})\ \mathbf{\hat{R}}) \ S({\yGyro{A}{B}} - \biasGyroHat)\  \mathbf{\hat{R}}^T \\
    &\approx \mathbf{\hat{R}}S({\yGyro{A}{B}} - \biasGyro - \noiseGyro{B})\  \mathbf{\hat{R}}^T - \mathbf{\hat{R}}S({\yGyro{A}{B}} - \biasGyroHat)\  \mathbf{\hat{R}}^T \\
    &=  \mathbf{\hat{R}}\ S(-\err_\biasGyro - \noiseGyro{B}) \mathbf{\hat{R}}^T \\
    &= S(-\mathbf{\hat{R}} \err_\biasGyro -\mathbf{\hat{R}} \noiseGyro{B}) \\
    &= S(\hat{\Rot{}{}} (-\err_\biasGyro - \noiseGyro{B}) ).
\end{align*}

\paragraph{Derivative of $\eta^R_\mathbf{p}$:}
From Eq. \eqref{eq:eq:chap:cdekf-sys-dynamics}, we have,
\begin{align*}
&\mathbf{\dot{p}} = \mathbf{v}, \\    
&\mathbf{\dot{\hat{p}}} = \mathbf{\hat{v}}.
\end{align*}
We will use  $\mathbf{R} = \mathbf{\hat{R}} + S(\err_\mathbf{R})\ \mathbf{\hat{R}}$ and $\mathbf{v} = \mathbf{\hat{v}} + S(\err_\mathbf{R})\mathbf{\hat{v}} + \err_\mathbf{v}$.
Using these definitions and neglecting the cross-product of infinitesimal error and noise terms, the derivative of position error can be computed as, \looseness=-1
\begin{align*}
    \dot{\eta}^R_\mathbf{p} &= \mathbf{\dot{p}} - \dot{\eta}_\mathbf{R} \mathbf{\hat{p}} - \eta_\mathbf{R}  \mathbf{\dot{\hat{p}}} \\
    &= \mathbf{v} - S(-\mathbf{\hat{R}} \err_\biasGyro -\mathbf{\hat{R}} \noiseGyro{B}) \mathbf{\hat{p}} - \Rot{}{} \mathbf{\hat{R}}^T \mathbf{\hat{v}} \\
    &= (\mathbf{\hat{v}} + S(\err_\mathbf{R})\mathbf{\hat{v}} + \err_\mathbf{v}) - S(-\mathbf{\hat{R}} \err_\biasGyro -\mathbf{\hat{R}} \noiseGyro{B}) \mathbf{\hat{p}} - (\mathbf{\hat{R}} + S(\err_\mathbf{R})\ \mathbf{\hat{R}}) \mathbf{\hat{R}}^T \mathbf{\hat{v}} \\
    &= \err_\mathbf{v} - S(-\mathbf{\hat{R}} \err_\biasGyro -\mathbf{\hat{R}} \noiseGyro{B}) \mathbf{\hat{p}} \\
    &= \err_\mathbf{v} - S(\mathbf{\hat{p}}) \mathbf{\hat{R}} (\err_\biasGyro + \noiseGyro{B}).
\end{align*}

\paragraph{Derivative of $\eta^R_\mathbf{v}$:}
From Eq. \eqref{eq:eq:chap:cdekf-sys-dynamics}, we have,
\begin{align*}
&\mathbf{\dot{v}} = \mathbf{R} (\yAcc{A}{B} - \biasAcc - \noiseAcc{B}) + \gravity{A}, \\    
&\mathbf{\dot{\hat{v}}} = \mathbf{\hat{R}}(\yAcc{A}{B} - \biasAccHat) + \gravity{A}.
\end{align*}
The derivative of linear velocity error can be computed as,
\begin{align*}
    \dot{\eta}^R_\mathbf{v} &= \mathbf{\dot{v}} - \dot{\eta}^R_\mathbf{R} \mathbf{\hat{v}} - \eta_\mathbf{R}  \mathbf{\dot{\hat{v}}} \\
    &= \mathbf{R} (\yAcc{A}{B} - \biasAcc - \noiseAcc{B}) + \gravity{A}  - S(-\mathbf{\hat{R}} \err_\biasGyro -\mathbf{\hat{R}} \noiseGyro{B}) \mathbf{\hat{v}} - \Rot{}{} \mathbf{\hat{R}}^T (\mathbf{\hat{R}}(\yAcc{A}{B} - \biasAccHat) + \gravity{A}) \\
    &= \mathbf{R}(- \err_\biasAcc - \noiseAcc{B}) + \gravity{A} - S(-\mathbf{\hat{R}} \err_\biasGyro -\mathbf{\hat{R}} \noiseGyro{B}) \mathbf{\hat{v}} - \Rot{}{} \mathbf{\hat{R}}^T \gravity{A} \\
    &= (\mathbf{\hat{R}} + S(\err_\mathbf{R})\ \mathbf{\hat{R}})(- \err_\biasAcc - \noiseAcc{B}) + \gravity{A} - S(-\mathbf{\hat{R}} \err_\biasGyro -\mathbf{\hat{R}} \noiseGyro{B}) \mathbf{\hat{v}} - (\mathbf{\hat{R}} + S(\err_\mathbf{R})\ \mathbf{\hat{R}}) \mathbf{\hat{R}}^T \gravity{A} \\
    &\approx \mathbf{\hat{R}}(- \err_\biasAcc - \noiseAcc{B}) + S(\mathbf{\hat{v}})\mathbf{\hat{R}} (-\err_\biasGyro - \noiseGyro{B}) + S(\gravity{A}) \err_\mathbf{R} \\
    &= S(\gravity{A}) \err_\mathbf{R} - S(\mathbf{\hat{v}})\hat{\Rot{}{}}\big(\err_\biasGyro + \noiseGyro{B}\big) -\hat{\Rot{}{}}\big(\err_\biasAcc + \noiseAcc{B}\big).
\end{align*}

\paragraph{Derivative of $\eta^R_\mathbf{Z}$:}
From Eq. \eqref{eq:eq:chap:cdekf-sys-dynamics}, we have,
\begin{align*}
&\mathbf{\dot{Z}} = \mathbf{Z}\ S(-\noiseAngVel{F}), \\    
&\mathbf{\dot{\hat{Z}}} = \Zero{3}.
\end{align*}
The derivative of foot orientation error can be computed as,
\begin{align*}
    \dot{\eta}^R_\mathbf{Z} &= \dot{\mathbf{Z}} \mathbf{\hat{Z}}^T + \mathbf{Z} \mathbf{\dot{\hat{Z}}}^T \\
    &= \mathbf{Z}\ S(-\noiseAngVel{F}) \mathbf{\hat{Z}}^T \\
    &= (\mathbf{\hat{Z}} + S(\err_\mathbf{Z})\ \mathbf{\hat{Z}}) S(-\noiseAngVel{F}) \mathbf{\hat{Z}}^T \\
    &\approx  \mathbf{\hat{Z}} \ S(-\noiseAngVel{F}) \mathbf{\hat{Z}}^T \\
    &= S(- \mathbf{\hat{Z}} \noiseAngVel{F}).
\end{align*}

\paragraph{Derivative of $\eta^R_\mathbf{d}$:}
From Eq. \eqref{eq:eq:chap:cdekf-sys-dynamics}, we have,
\begin{align*}
&\mathbf{\dot{d}} = -\mathbf{Z}\ \noiseLinVel{F}, \\    
&\mathbf{\dot{\hat{d}}} = \Zeros{3}{1}.
\end{align*}
The derivative of foot position error can be computed as,
\begin{align*}
    \dot{\eta}^R_\mathbf{d} &=  \mathbf{\dot{d}} - \dot{\eta}^R_\mathbf{Z} \mathbf{\hat{d}} - \eta_\mathbf{Z}  \mathbf{\dot{\hat{d}}} \\
    &=-\mathbf{Z}\ \noiseLinVel{F}  - S(- \mathbf{\hat{Z}} \noiseAngVel{F})  \mathbf{\hat{d}} \\
    & \approx -\mathbf{\hat{Z}}\ \noiseLinVel{F}  - S(\mathbf{\hat{d}}) \mathbf{\hat{Z}} \noiseAngVel{F}.
\end{align*}
\paragraph{Derivative of $\eta^R_\bias$:} The derivative of bias errors can be computed in a straightforward manner.

\section{Non-linear error dynamics of CODILIGENT-KIO}
\label{appendix:chapter:nonlinear-err-dyn-cdekf-lie}

Given the definition of left invariant error in Eq. \eqref{eq:eq:chap:cdekf-sys-lie} used for the CODILIGENT-KIO, in this section, we provide a detailed derivation of the error dynamics.

\paragraph{Derivative of $\eta^L_\Rot{}{}$:}
From Eq. \eqref{eq:eq:chap:cdekf-sys-dynamics}, we have,
\begin{align*}
&\Rdot{}{} = \mathbf{R}\ S({\yGyro{A}{B}} - \biasGyro - \noiseGyro{B}), \\    
&\mathbf{\dot{\hat{R}}} = \mathbf{\hat{R}}\ S({\yGyro{A}{B}} - \biasGyroHat),
\end{align*}
and the true rotation $\mathbf{R} = \mathbf{\hat{R}}\ \eta^L_\Rot{}{} \approx \mathbf{\hat{R}} + \mathbf{\hat{R}} S(\err_\mathbf{R})$ using a first-order approximation of the exponential map.
Using these definitions and neglecting the cross-product of infinitesimal error and noise terms, the derivative of rotation error can be computed as,
\begin{align*}
    \dot{\eta}^L_\Rot{}{} &=   \mathbf{\dot{\hat{R}}}^T\ \mathbf{R} + \mathbf{\hat{R}}^T\ \dot{\Rot{}{}}\\
    &= (\mathbf{\hat{R}}\ S({\yGyro{A}{B}} - \biasGyroHat))^T\ \mathbf{R} + \mathbf{\hat{R}}^T\ (\mathbf{R}\ S({\yGyro{A}{B}} - \biasGyro - \noiseGyro{B}))\\
    &= - S({\yGyro{A}{B}} - \biasGyroHat)\ \mathbf{\hat{R}}^T  \mathbf{R} + \mathbf{\hat{R}}^T\ \mathbf{R}\ S({\yGyro{A}{B}} - \biasGyro - \noiseGyro{B})\\
    &= - S({\yGyro{A}{B}} - \biasGyroHat)\ \mathbf{\hat{R}}^T  (\mathbf{\hat{R}} + \mathbf{\hat{R}} S(\err_\mathbf{R})) + \mathbf{\hat{R}}^T\ (\mathbf{\hat{R}} + \mathbf{\hat{R}} S(\err_\mathbf{R}))\ S({\yGyro{A}{B}} - \biasGyro - \noiseGyro{B})\\
    &= S(- \err_\biasGyro - \noiseGyro{B}).
\end{align*}

\paragraph{Derivative of $\eta^L_\mathbf{p}$:}
From Eq. \eqref{eq:eq:chap:cdekf-sys-dynamics}, we have,
\begin{align*}
&\mathbf{\dot{p}} = \mathbf{v}, \\    
&\mathbf{\dot{\hat{p}}} = \mathbf{\hat{v}}.
\end{align*}
We will use  $\mathbf{R} = \mathbf{\hat{R}} +  \mathbf{\hat{R}}\ S(\err_\mathbf{R})$, $\mathbf{p} = \mathbf{\hat{p}} + \mathbf{\hat{R}} \err_\mathbf{p}$ and $\mathbf{v} = \mathbf{\hat{v}} + \mathbf{\hat{R}} \err_\mathbf{v}$.
Using these definitions and neglecting the cross-product of infinitesimal error and noise terms, the derivative of position error can be computed as, \looseness=-1
\begin{align*}
    \dot{\eta}^L_\mathbf{p} &= \mathbf{\dot{\hat{R}}}^T(\mathbf{p} - \mathbf{\hat{p}}) + \mathbf{\hat{R}}^T \mathbf{\dot{p}} - \mathbf{\hat{R}}^T \mathbf{\dot{\hat{p}}} \\
    &= (\mathbf{\hat{R}}\ S({\yGyro{A}{B}} - \biasGyroHat))^T (\mathbf{p} - \mathbf{\hat{p}}) + \mathbf{\hat{R}}^T \mathbf{v} - \mathbf{\hat{R}}^T \mathbf{{\hat{v}}} \\
    &= (\mathbf{\hat{R}}\ S({\yGyro{A}{B}} - \biasGyroHat))^T (\mathbf{\hat{p}} + \mathbf{\hat{R}} \err_\mathbf{p} - \mathbf{\hat{p}}) + \mathbf{\hat{R}}^T (\mathbf{\hat{v}} + \mathbf{\hat{R}} \err_\mathbf{v}) - \mathbf{\hat{R}}^T \mathbf{{\hat{v}}} \\
    &=  - S({\yGyro{A}{B}} - \biasGyroHat)\ \mathbf{\hat{R}}^T ( \mathbf{\hat{R}} \err_\mathbf{p}) +  \err_\mathbf{v} \\
    &=  - S({\yGyro{A}{B}} - \biasGyroHat) \err_\mathbf{p} +  \err_\mathbf{v}.
\end{align*}

\paragraph{Derivative of $\eta^L_\mathbf{v}$:}
We have,
\begin{align*}
&\mathbf{\dot{v}} = \mathbf{R} (\yAcc{A}{B} - \biasAcc - \noiseAcc{B}) + \gravity{A}, \\    
&\mathbf{\dot{\hat{v}}} = \mathbf{\hat{R}}(\yAcc{A}{B} - \biasAccHat) + \gravity{A}.
\end{align*}
The derivative of linear velocity error can be computed as,
\begin{align*}
    \dot{\eta}^L_\mathbf{v} &= \mathbf{\dot{\hat{R}}}^T(\mathbf{v} - \mathbf{\hat{v}}) + \mathbf{\hat{R}}^T \mathbf{\dot{v}} - \mathbf{\hat{R}}^T \mathbf{\dot{\hat{v}}} \\
    &= - S({\yGyro{A}{B}} - \biasGyroHat)\ \mathbf{\hat{R}}^T (\mathbf{v} - \mathbf{\hat{v}}) + \mathbf{\hat{R}}^T (\mathbf{R} (\yAcc{A}{B} - \biasAcc - \noiseAcc{B}) + \gravity{A}) + \\
    & \qquad \qquad \qquad \mathbf{\hat{R}}^T (\mathbf{\hat{R}}(\yAcc{A}{B} - \biasAccHat) + \gravity{A}) \\
    &= - S({\yGyro{A}{B}} - \biasGyroHat) \err_\mathbf{v} + (\yAcc{A}{B} - \biasAcc - \noiseAcc{B}) + S(\err_\mathbf{R}) (\yAcc{A}{B} - \biasAcc - \noiseAcc{B}) - (\yAcc{A}{B} - \biasAccHat) \\
    &\approx - S({\yGyro{A}{B}} - \biasGyroHat) \err_\mathbf{v} - S(\yAcc{A}{B} - \biasAccHat)  \err_\mathbf{R} -\err_\biasAcc -\noiseAcc{B}.
\end{align*}

\paragraph{Derivative of $\eta^L_\mathbf{Z}$:}
From Eq. \eqref{eq:eq:chap:cdekf-sys-dynamics}, we have,
\begin{align*}
&\mathbf{\dot{Z}} = \mathbf{Z}\ S(-\noiseAngVel{F}), \\    
&\mathbf{\dot{\hat{Z}}} = \Zero{3}.
\end{align*}
Neglecting the cross-product of infinitesimal terms, the derivative of foot orientation error can be computed as,
\begin{align*}
    \dot{\eta}^L_\mathbf{Z} &= \mathbf{\dot{\hat{Z}}}^T\ \mathbf{Z} + \mathbf{\hat{Z}}^T\ \dot{\mathbf{Z}}\\
    &= \mathbf{\hat{Z}}^T \mathbf{Z}\ S(-\noiseAngVel{F}) \\
    &\approx  S(-\noiseAngVel{F}).
\end{align*}

\paragraph{Derivative of $\eta^L_\mathbf{d}$:}
We have,
\begin{align*}
&\mathbf{\dot{d}} = -\mathbf{Z}\ \noiseLinVel{F}, \\    
&\mathbf{\dot{\hat{d}}} = \Zeros{3}{1}.
\end{align*}
Neglecting the cross-product of infinitesimal terms, the derivative of foot position error can be computed as,
\begin{align*}
    \dot{\eta}^L_\mathbf{d} &=  \mathbf{\dot{\hat{Z}}}^T(\mathbf{d} - \mathbf{\hat{d}}) + \mathbf{\hat{Z}}^T \mathbf{\dot{d}} - \mathbf{\hat{Z}}^T \mathbf{\dot{\hat{d}}} \\
    &= S(-\noiseAngVel{F}) (\mathbf{\hat{Z}} \err_\mathbf{d}) -  \mathbf{\hat{Z}}^T \mathbf{Z}\ \noiseLinVel{F} \\
    &\approx -\noiseLinVel{F}.
\end{align*}
\paragraph{Derivative of $\eta^L_\bias$:} The derivative of bias errors can be computed in a straightforward manner.

\end{appendices}

\begin{spacing}{0.9}


\bibliographystyle{plainnat} 

\cleardoublepage
\bibliography{references} 

\begin{thebibliography}{192}
\providecommand{\natexlab}[1]{#1}
\providecommand{\url}[1]{\texttt{#1}}
\expandafter\ifx\csname urlstyle\endcsname\relax
  \providecommand{\doi}[1]{doi: #1}\else
  \providecommand{\doi}{doi: \begingroup \urlstyle{rm}\Url}\fi

\bibitem[Aristidou et~al.(2018)Aristidou, Lasenby, Chrysanthou, and
  Shamir]{aristidou2018inverse}
Andreas Aristidou, Joan Lasenby, Yiorgos Chrysanthou, and Ariel Shamir.
\newblock Inverse kinematics techniques in computer graphics: A survey.
\newblock In \emph{Computer Graphics Forum}, volume~37, pages 35--58. Wiley
  Online Library, 2018.

\bibitem[Bae et~al.(2017)Bae, Oh, Jeong, and Oh]{bae2017new}
Hyoin Bae, Jaesung Oh, Hyobin Jeong, and Jun-Ho Oh.
\newblock A new state estimation framework for humanoids based on a moving
  horizon estimator.
\newblock \emph{IFAC-PapersOnLine}, 50\penalty0 (1):\penalty0 3793--3799, 2017.

\bibitem[Bailly et~al.(2019)Bailly, Carpentier, Benallegue, Watier, and
  Sou{\`e}res]{bailly2019estimating}
Fran{\c{c}}ois Bailly, Justin Carpentier, Mehdi Benallegue, Bruno Watier, and
  Philippe Sou{\`e}res.
\newblock Estimating the center of mass and the angular momentum derivative for
  legged locomotion—a recursive approach.
\newblock \emph{IEEE Robotics and Automation Letters}, 4\penalty0 (4):\penalty0
  4155--4162, 2019.

\bibitem[Barfoot(2017)]{barfoot2017state}
Timothy~D Barfoot.
\newblock \emph{State estimation for robotics}.
\newblock Cambridge University Press, 2017.

\bibitem[Barfoot and Furgale(2014)]{barfoot2014associating}
Timothy~D. Barfoot and Paul~T. Furgale.
\newblock Associating uncertainty with three-dimensional poses for use in
  estimation problems.
\newblock 30\penalty0 (3):\penalty0 679--693, jun 2014.
\newblock \doi{10.1109/tro.2014.2298059}.

\bibitem[Barrau(2015)]{barrau2015non}
Axel Barrau.
\newblock \emph{Non-linear state error based extended Kalman filters with
  applications to navigation}.
\newblock PhD thesis, Mines Paristech, 2015.

\bibitem[Barrau and Bonnabel(2017)]{barrau2017invariant}
Axel Barrau and Silvere Bonnabel.
\newblock The invariant extended kalman filter as a stable observer.
\newblock 62\penalty0 (4):\penalty0 1797--1812, apr 2017.
\newblock \doi{10.1109/tac.2016.2594085}.

\bibitem[Benallegue et~al.(2015)Benallegue, Mifsud, and
  Lamiraux]{benallegue2015fusion}
Mehdi Benallegue, Alexis Mifsud, and Florent Lamiraux.
\newblock Fusion of force-torque sensors, inertial measurements units and
  proprioception for a humanoid kinematics-dynamics observation.
\newblock In \emph{2015 IEEE-RAS 15th International Conference on Humanoid
  Robots (Humanoids)}, pages 664--669. IEEE, 2015.

\bibitem[Benallegue et~al.(2020)Benallegue, Cisneros, Benallegue, Chitour,
  Morisawa, and Kanehiro]{benallegue2020lyapunov}
Mehdi Benallegue, Rafael Cisneros, Abdelaziz Benallegue, Yacine Chitour,
  Mitsuharu Morisawa, and Fumio Kanehiro.
\newblock Lyapunov-stable orientation estimator for humanoid robots.
\newblock \emph{IEEE Robotics and Automation Letters}, 5\penalty0 (4):\penalty0
  6371--6378, 2020.
\newblock \doi{10.1109/LRA.2020.3013854}.

\bibitem[Blanco(2010)]{blanco2010tutorial}
Jose-Luis Blanco.
\newblock A tutorial on se (3) transformation parameterizations and on-manifold
  optimization.
\newblock \emph{University of Malaga, Tech. Rep}, 3:\penalty0 6, 2010.

\bibitem[Bloesch(2017)]{bloesch2017state}
Michael Bloesch.
\newblock \emph{State estimation for legged robots-kinematics, inertial
  sensing, and computer vision}.
\newblock PhD thesis, ETH Zurich, 2017.

\bibitem[Bloesch et~al.(2013)Bloesch, Hutter, Hoepflinger, Leutenegger,
  Gehring, Remy, and Siegwart]{bloesch2013state}
Michael Bloesch, Marco Hutter, Mark~A Hoepflinger, Stefan Leutenegger,
  Christian Gehring, C~David Remy, and Roland Siegwart.
\newblock State estimation for legged robotsconsistent fusion of leg kinematics
  and imu.
\newblock \emph{Robotics}, 17:\penalty0 17--24, 2013.

\bibitem[Bloesch et~al.(2014)Bloesch, Omari, Fankhauser, Sommer, Gehring,
  Hwangbo, Hoepflinger, Hutter, and Siegwart]{bloesch2014fusion}
Michael Bloesch, Sammy Omari, P{\'e}ter Fankhauser, Hannes Sommer, Christian
  Gehring, Jemin Hwangbo, Mark~A Hoepflinger, Marco Hutter, and Roland
  Siegwart.
\newblock Fusion of optical flow and inertial measurements for robust egomotion
  estimation.
\newblock In \emph{2014 IEEE/RSJ International Conference on Intelligent Robots
  and Systems}, pages 3102--3107. IEEE, 2014.

\bibitem[Bogo et~al.(2016)Bogo, Kanazawa, Lassner, Gehler, Romero, and
  Black]{bogo2016keep}
Federica Bogo, Angjoo Kanazawa, Christoph Lassner, Peter Gehler, Javier Romero,
  and Michael~J Black.
\newblock Keep it smpl: Automatic estimation of 3d human pose and shape from a
  single image.
\newblock In \emph{European conference on computer vision}, pages 561--578.
  Springer, 2016.

\bibitem[Bonnabel et~al.(2009)Bonnabel, Martin, and Rouchon]{bonnabel2009non}
Silvere Bonnabel, Philippe Martin, and Pierre Rouchon.
\newblock Non-linear symmetry-preserving observers on lie groups.
\newblock \emph{IEEE Transactions on Automatic Control}, 54\penalty0
  (7):\penalty0 1709--1713, 2009.

\bibitem[Bourmaud(2015)]{bourmaud2015estimation}
Guillaume Bourmaud.
\newblock \emph{Estimation de param{\`e}tres {\'e}voluant sur des groupes de
  Lie: Application {\`a} la cartographie et {\`a} la localisation d'une
  cam{\'e}ra monoculaire}.
\newblock PhD thesis, Universit{\'e} de Bordeaux, 2015.

\bibitem[Bourmaud et~al.(2013)Bourmaud, Mégret, Giremus, and
  Berthoumieu]{bourmaud2013discrete}
Guillaume Bourmaud, Rémi Mégret, Audrey Giremus, and Yannick Berthoumieu.
\newblock Discrete extended kalman filter on lie groups.
\newblock In \emph{21st European Signal Processing Conference (EUSIPCO 2013)},
  pages 1--5, 2013.

\bibitem[Bourmaud et~al.(2015)Bourmaud, M{\'e}gret, Arnaudon, and
  Giremus]{bourmaud2015continuous}
Guillaume Bourmaud, R{\'e}mi M{\'e}gret, Marc Arnaudon, and Audrey Giremus.
\newblock Continuous-discrete extended kalman filter on matrix lie groups using
  concentrated gaussian distributions.
\newblock \emph{Journal of Mathematical Imaging and Vision}, 51\penalty0
  (1):\penalty0 209--228, 2015.

\bibitem[Bresson et~al.(2017)Bresson, Alsayed, Yu, and
  Glaser]{bresson2017simultaneous}
Guillaume Bresson, Zayed Alsayed, Li~Yu, and S{\'e}bastien Glaser.
\newblock Simultaneous localization and mapping: A survey of current trends in
  autonomous driving.
\newblock \emph{IEEE Transactions on Intelligent Vehicles}, 2\penalty0
  (3):\penalty0 194--220, 2017.

\bibitem[Brossard et~al.(2017)Brossard, Bonnabel, and
  Condomines]{brossard2017unscented}
Martin Brossard, Silvere Bonnabel, and Jean-Philippe Condomines.
\newblock Unscented kalman filtering on lie groups.
\newblock In \emph{2017 IEEE/RSJ International Conference on Intelligent Robots
  and Systems (IROS)}, pages 2485--2491. IEEE, 2017.

\bibitem[Brossard et~al.(2018)Brossard, Bonnabel, and
  Barrau]{brossard2018unscented}
Martin Brossard, Silvere Bonnabel, and Axel Barrau.
\newblock Unscented kalman filter on lie groups for visual inertial odometry.
\newblock In \emph{2018 IEEE/RSJ International Conference on Intelligent Robots
  and Systems (IROS)}, pages 649--655. IEEE, 2018.

\bibitem[Cadena et~al.(2016)Cadena, Carlone, Carrillo, Latif, Scaramuzza,
  Neira, Reid, and Leonard]{cadena2016past}
Cesar Cadena, Luca Carlone, Henry Carrillo, Yasir Latif, Davide Scaramuzza,
  Jos{\'e} Neira, Ian Reid, and John~J Leonard.
\newblock Past, present, and future of simultaneous localization and mapping:
  Toward the robust-perception age.
\newblock \emph{IEEE Transactions on robotics}, 32\penalty0 (6):\penalty0
  1309--1332, 2016.

\bibitem[Camurri et~al.(2017)Camurri, Fallon, Bazeille, Radulescu, Barasuol,
  Caldwell, and Semini]{camurri2017probabilistic}
Marco Camurri, Maurice Fallon, St{\'e}phane Bazeille, Andreea Radulescu, Victor
  Barasuol, Darwin~G Caldwell, and Claudio Semini.
\newblock Probabilistic contact estimation and impact detection for state
  estimation of quadruped robots.
\newblock \emph{IEEE Robotics and Automation Letters}, 2\penalty0 (2):\penalty0
  1023--1030, 2017.

\bibitem[Camurri et~al.(2020)Camurri, Ramezani, Nobili, and
  Fallon]{camurri2020pronto}
Marco Camurri, Milad Ramezani, Simona Nobili, and Maurice Fallon.
\newblock Pronto: A multi-sensor state estimator for legged robots in
  real-world scenarios.
\newblock \emph{Frontiers in Robotics and AI}, 7:\penalty0 68, 2020.

\bibitem[Cao et~al.(2019)Cao, Hidalgo, Simon, Wei, and Sheikh]{cao2019openpose}
Zhe Cao, Gines Hidalgo, Tomas Simon, Shih-En Wei, and Yaser Sheikh.
\newblock Openpose: realtime multi-person 2d pose estimation using part
  affinity fields.
\newblock \emph{IEEE transactions on pattern analysis and machine
  intelligence}, 43\penalty0 (1):\penalty0 172--186, 2019.

\bibitem[Carpentier et~al.(2015)Carpentier, Benallegue, Mansard, and
  Laumond]{carpentier2015kinematics}
Justin Carpentier, Mehdi Benallegue, Nicolas Mansard, and Jean-Paul Laumond.
\newblock A kinematics-dynamics based estimator of the center of mass position
  for anthropomorphic system—a complementary filtering approach.
\newblock In \emph{2015 IEEE-RAS 15th International Conference on Humanoid
  Robots (Humanoids)}, pages 1121--1126. IEEE, 2015.

\bibitem[Carpentier et~al.(2016)Carpentier, Benallegue, Mansard, and
  Laumond]{carpentier2016center}
Justin Carpentier, Mehdi Benallegue, Nicolas Mansard, and Jean-Paul Laumond.
\newblock Center-of-mass estimation for a polyarticulated system in contact—a
  spectral approach.
\newblock \emph{IEEE Transactions on Robotics}, 32\penalty0 (4):\penalty0
  810--822, 2016.

\bibitem[Cashbaugh and Kitts(2018)]{cashbaugh2018automatic}
Jasmine Cashbaugh and Christopher Kitts.
\newblock Automatic calculation of a transformation matrix between two frames.
\newblock \emph{IEEE Access}, 6:\penalty0 9614--9622, 2018.

\bibitem[{\'C}esi{\'c} et~al.(2016){\'C}esi{\'c}, Joukov, Petrovi{\'c}, and
  Kuli{\'c}]{cesic2016full}
Josip {\'C}esi{\'c}, Vladimir Joukov, Ivan Petrovi{\'c}, and Dana Kuli{\'c}.
\newblock Full body human motion estimation on lie groups using 3d marker
  position measurements.
\newblock In \emph{2016 IEEE-RAS 16th International Conference on Humanoid
  Robots (Humanoids)}, pages 826--833. IEEE, 2016.

\bibitem[Chatfield(1997)]{chatfield1997fundamentals}
Averil~Burton Chatfield.
\newblock \emph{Fundamentals of high accuracy inertial navigation}, volume 174.
\newblock Aiaa, 1997.

\bibitem[Chauchat et~al.(2018)Chauchat, Barrau, and
  Bonnabel]{chauchat2018invariant}
Paul Chauchat, Axel Barrau, and Silvere Bonnabel.
\newblock Invariant smoothing on lie groups.
\newblock In \emph{2018 IEEE/RSJ International Conference on Intelligent Robots
  and Systems (IROS)}, pages 1703--1710. IEEE, 2018.

\bibitem[Chavez et~al.(2016)Chavez, Traversaro, Pucci, and
  Nori]{chavez2016model}
Francisco Javier~Andrade Chavez, Silvio Traversaro, Daniele Pucci, and
  Francesco Nori.
\newblock Model based in situ calibration of six axis force torque sensors.
\newblock In \emph{2016 IEEE-RAS 16th International Conference on Humanoid
  Robots (Humanoids)}, pages 422--427. IEEE, 2016.

\bibitem[Chirikjian(2009)]{chirikjian2009stochastic}
Gregory~S Chirikjian.
\newblock \emph{Stochastic Models, Information Theory, and Lie Groups, Volume
  1: Classical Results and Geometric Methods}.
\newblock Springer Science \& Business Media, 2009.

\bibitem[Chirikjian(2011)]{chirikjian2011stochastic}
Gregory~S Chirikjian.
\newblock \emph{Stochastic models, information theory, and Lie groups, volume
  2: Analytic methods and modern applications}, volume~2.
\newblock Springer Science \& Business Media, 2011.

\bibitem[Chitta et~al.(2007)Chitta, Vemaza, Geykhman, and
  Lee]{chitta2007proprioceptive}
Sachin Chitta, Paul Vemaza, Roman Geykhman, and Daniel~D Lee.
\newblock Proprioceptive localilzatilon for a quadrupedal robot on known
  terrain.
\newblock In \emph{Proceedings 2007 IEEE International Conference on Robotics
  and Automation}, pages 4582--4587. IEEE, 2007.

\bibitem[Crassidis et~al.(2007)Crassidis, Markley, and
  Cheng]{crassidis2007survey}
John~L Crassidis, F~Landis Markley, and Yang Cheng.
\newblock Survey of nonlinear attitude estimation methods.
\newblock \emph{Journal of guidance, control, and dynamics}, 30\penalty0
  (1):\penalty0 12--28, 2007.

\bibitem[Dafarra(2020)]{dafarra2020predictive}
Stefano Dafarra.
\newblock Predictive whole-body control of humanoid robot locomotion.
\newblock \emph{arXiv preprint arXiv:2004.07699}, 2020.

\bibitem[Dafarra et~al.(2016)Dafarra, Romano, and Nori]{dafarra2016torque}
Stefano Dafarra, Francesco Romano, and Francesco Nori.
\newblock Torque-controlled stepping-strategy push recovery: Design and
  implementation on the icub humanoid robot.
\newblock In \emph{2016 IEEE-RAS 16th International Conference on Humanoid
  Robots (Humanoids)}, pages 152--157. IEEE, 2016.

\bibitem[Darvish et~al.(2019)Darvish, Tirupachuri, Romualdi, Rapetti, Ferigo,
  Chavez, and Pucci]{darvish2019whole}
Kourosh Darvish, Yeshasvi Tirupachuri, Giulio Romualdi, Lorenzo Rapetti, Diego
  Ferigo, Francisco Javier~Andrade Chavez, and Daniele Pucci.
\newblock Whole-body geometric retargeting for humanoid robots.
\newblock In \emph{2019 IEEE-RAS 19th International Conference on Humanoid
  Robots (Humanoids)}, pages 679--686. IEEE, 2019.

\bibitem[Davis and Rabinowitz(2007)]{davis2007methods}
Philip~J Davis and Philip Rabinowitz.
\newblock \emph{Methods of numerical integration}.
\newblock Courier Corporation, 2007.

\bibitem[Dellaert et~al.(2017)Dellaert, Kaess, et~al.]{dellaert2017factor}
Frank Dellaert, Michael Kaess, et~al.
\newblock Factor graphs for robot perception.
\newblock \emph{Foundations and Trends{\textregistered} in Robotics},
  6\penalty0 (1-2):\penalty0 1--139, 2017.

\bibitem[Ding et~al.(2021)Ding, Yang, Zhang, Xu, and Gao]{ding2021vid}
Ziming Ding, Tiankai Yang, Kunyi Zhang, Chao Xu, and Fei Gao.
\newblock Vid-fusion: Robust visual-inertial-dynamics odometry for accurate
  external force estimation.
\newblock In \emph{2021 IEEE International Conference on Robotics and
  Automation (ICRA)}, pages 14469--14475. IEEE, 2021.

\bibitem[Eckenhoff et~al.(2019)Eckenhoff, Yang, Geneva, and
  Huang]{eckenhoff2019tightly}
Kevin Eckenhoff, Yulin Yang, Patrick Geneva, and Guoquan Huang.
\newblock Tightly-coupled visual-inertial localization and 3-d rigid-body
  target tracking.
\newblock \emph{IEEE Robotics and Automation Letters}, 4\penalty0 (2):\penalty0
  1541--1548, 2019.

\bibitem[Eckenhoff et~al.(2020)Eckenhoff, Geneva, and Huang]{eckenhoff2020high}
Kevin Eckenhoff, Patrick Geneva, and Guoquan Huang.
\newblock High-accuracy preintegration for visual-inertial navigation.
\newblock In \emph{Algorithmic Foundations of Robotics XII}, pages 48--63.
  Springer, 2020.

\bibitem[Englsberger et~al.(2018)Englsberger, Mesesan, Werner, and
  Ott]{englsberger2018torque}
Johannes Englsberger, George Mesesan, Alexander Werner, and Christian Ott.
\newblock Torque-based dynamic walking-a long way from simulation to
  experiment.
\newblock In \emph{2018 IEEE International Conference on Robotics and
  Automation (ICRA)}, pages 440--447. IEEE, 2018.

\bibitem[Fallon et~al.(2014)Fallon, Antone, Roy, and Teller]{fallon2014drift}
Maurice~F Fallon, Matthew Antone, Nicholas Roy, and Seth Teller.
\newblock Drift-free humanoid state estimation fusing kinematic, inertial and
  lidar sensing.
\newblock In \emph{2014 IEEE-RAS International Conference on Humanoid Robots},
  pages 112--119. IEEE, 2014.

\bibitem[Fang et~al.(2018)Fang, Tian, Wang, Zhou, and Haile]{fang2018nonlinear}
Huazhen Fang, Ning Tian, Yebin Wang, MengChu Zhou, and Mulugeta~A. Haile.
\newblock Nonlinear bayesian estimation: from kalman filtering to a broader
  horizon.
\newblock 5\penalty0 (2):\penalty0 401--417, mar 2018.
\newblock \doi{10.1109/jas.2017.7510808}.

\bibitem[Fankhauser et~al.(2018)Fankhauser, Bloesch, and
  Hutter]{fankhauser2018probabilistic}
P{\'e}ter Fankhauser, Michael Bloesch, and Marco Hutter.
\newblock Probabilistic terrain mapping for mobile robots with uncertain
  localization.
\newblock \emph{IEEE Robotics and Automation Letters}, 3\penalty0 (4):\penalty0
  3019--3026, 2018.

\bibitem[Featherstone(2014)]{featherstone2014rigid}
Roy Featherstone.
\newblock \emph{Rigid body dynamics algorithms}.
\newblock Springer, 2014.

\bibitem[Fern{\'a}ndez-Madrigal(2012)]{fernandez2012simultaneous}
Juan-Antonio Fern{\'a}ndez-Madrigal.
\newblock \emph{Simultaneous Localization and Mapping for Mobile Robots:
  Introduction and Methods: Introduction and Methods}.
\newblock IGI global, 2012.

\bibitem[Filippeschi et~al.(2017)Filippeschi, Schmitz, Miezal, Bleser,
  Ruffaldi, and Stricker]{filippeschi2017survey}
Alessandro Filippeschi, Norbert Schmitz, Markus Miezal, Gabriele Bleser,
  Emanuele Ruffaldi, and Didier Stricker.
\newblock Survey of motion tracking methods based on inertial sensors: A focus
  on upper limb human motion.
\newblock \emph{Sensors}, 17\penalty0 (6):\penalty0 1257, 2017.

\bibitem[Fink and Semini(2020)]{fink2020proprioceptive}
Geoff Fink and Claudio Semini.
\newblock Proprioceptive sensor fusion for quadruped robot state estimation.
\newblock In \emph{2020 IEEE/RSJ International Conference on Intelligent Robots
  and Systems (IROS)}, pages 10914--10920. IEEE, 2020.

\bibitem[Flayols et~al.(2017)Flayols, Del~Prete, Wensing, Mifsud, Benallegue,
  and Stasse]{flayols2017experimental}
Thomas Flayols, Andrea Del~Prete, Patrick Wensing, Alexis Mifsud, Mehdi
  Benallegue, and Olivier Stasse.
\newblock Experimental evaluation of simple estimators for humanoid robots.
\newblock In \emph{2017 IEEE-RAS 17th International Conference on Humanoid
  Robotics (Humanoids)}, pages 889--895. IEEE, 2017.

\bibitem[Forster et~al.(2016)Forster, Carlone, Dellaert, and
  Scaramuzza]{forster2016manifold}
Christian Forster, Luca Carlone, Frank Dellaert, and Davide Scaramuzza.
\newblock On-manifold preintegration for real-time visual--inertial odometry.
\newblock \emph{IEEE Transactions on Robotics}, 33\penalty0 (1):\penalty0
  1--21, 2016.

\bibitem[Fourmy et~al.(2019)Fourmy, Atchuthan, Mansard, Sola, and
  Flayols]{fourmy2019absolute}
Mederic Fourmy, Dinesh Atchuthan, Nicolas Mansard, Joan Sola, and Thomas
  Flayols.
\newblock Absolute humanoid localization and mapping based on imu lie group and
  fiducial markers.
\newblock In \emph{2019 IEEE-RAS 19th International Conference on Humanoid
  Robots (Humanoids)}, pages 237--243. IEEE, 2019.

\bibitem[Fourmy et~al.(2021)Fourmy, Flayols, Mansard, and
  Sol{\`a}]{fourmy2021contact}
Mederic Fourmy, Thomas Flayols, Nicolas Mansard, and Joan Sol{\`a}.
\newblock Contact forces pre-integration for the whole body estimation of
  legged robots.
\newblock In \emph{2021 IEEE International Conference on Robotics and
  Automation-ICRA}, 2021.

\bibitem[Ganapathi et~al.(2010)Ganapathi, Plagemann, Koller, and
  Thrun]{ganapathi2010real}
Varun Ganapathi, Christian Plagemann, Daphne Koller, and Sebastian Thrun.
\newblock Real time motion capture using a single time-of-flight camera.
\newblock In \emph{2010 IEEE Computer Society Conference on Computer Vision and
  Pattern Recognition}, pages 755--762. IEEE, 2010.

\bibitem[Gros et~al.(2015)Gros, Zanon, and Diehl]{gros2015baumgarte}
Sebastien Gros, Marion Zanon, and Moritz Diehl.
\newblock Baumgarte stabilisation over the so (3) rotation group for control.
\newblock In \emph{2015 54th IEEE Conference on Decision and Control (CDC)},
  pages 620--625. IEEE, 2015.

\bibitem[Guerra-Filho(2005)]{guerra2005optical}
Gutemberg Guerra-Filho.
\newblock Optical motion capture: Theory and implementation.
\newblock \emph{RITA}, 12\penalty0 (2):\penalty0 61--90, 2005.

\bibitem[Hall(2013)]{hall2013lie}
Brian~C Hall.
\newblock Lie groups, lie algebras, and representations.
\newblock In \emph{Quantum Theory for Mathematicians}, pages 333--366.
  Springer, 2013.

\bibitem[Hanavan~Jr(1964)]{hanavan1964mathematical}
Ernest~P Hanavan~Jr.
\newblock A mathematical model of the human body.
\newblock Technical report, Air Force Aerospace Medical Research Lab
  Wright-patterson AFB OH, 1964.

\bibitem[Hartley(2019)]{hartley2019contact}
Matthew Hartley.
\newblock \emph{Contact-aided state estimation on lie groups for legged robot
  mapping and control}.
\newblock PhD thesis, 2019.

\bibitem[Hartley et~al.(2013)Hartley, Trumpf, Dai, and Li]{hartley2013rotation}
Richard Hartley, Jochen Trumpf, Yuchao Dai, and Hongdong Li.
\newblock Rotation averaging.
\newblock \emph{International journal of computer vision}, 103\penalty0
  (3):\penalty0 267--305, 2013.

\bibitem[Hartley et~al.(2018{\natexlab{a}})Hartley, Jadidi, Gan, Huang,
  Grizzle, and Eustice]{hartley2018hybrid}
Ross Hartley, Maani~Ghaffari Jadidi, Lu~Gan, Jiunn-Kai Huang, Jessy~W Grizzle,
  and Ryan~M Eustice.
\newblock Hybrid contact preintegration for visual-inertial-contact state
  estimation using factor graphs.
\newblock In \emph{2018 IEEE/RSJ International Conference on Intelligent Robots
  and Systems (IROS)}, pages 3783--3790. IEEE, 2018{\natexlab{a}}.

\bibitem[Hartley et~al.(2018{\natexlab{b}})Hartley, Mangelson, Gan, Jadidi,
  Walls, Eustice, and Grizzle]{hartley2018legged}
Ross Hartley, Josh Mangelson, Lu~Gan, Maani~Ghaffari Jadidi, Jeffrey~M Walls,
  Ryan~M Eustice, and Jessy~W Grizzle.
\newblock Legged robot state-estimation through combined forward kinematic and
  preintegrated contact factors.
\newblock In \emph{2018 IEEE International Conference on Robotics and
  Automation (ICRA)}, pages 4422--4429. IEEE, 2018{\natexlab{b}}.

\bibitem[Hartley et~al.(2020)Hartley, Ghaffari, Eustice, and
  Grizzle]{hartley2020contact}
Ross Hartley, Maani Ghaffari, Ryan~M Eustice, and Jessy~W Grizzle.
\newblock Contact-aided invariant extended kalman filtering for robot state
  estimation.
\newblock 39\penalty0 (4):\penalty0 402--430, jan 2020.
\newblock \doi{10.1177/0278364919894385}.

\bibitem[Hertzberg et~al.(2013)Hertzberg, Wagner, Frese, and
  Schr{\"o}der]{hertzberg2013integrating}
Christoph Hertzberg, Ren{\'e} Wagner, Udo Frese, and Lutz Schr{\"o}der.
\newblock Integrating generic sensor fusion algorithms with sound state
  representations through encapsulation of manifolds.
\newblock \emph{Information Fusion}, 14\penalty0 (1):\penalty0 57--77, 2013.

\bibitem[Hornung et~al.(2010)Hornung, Wurm, and Bennewitz]{hornung2010humanoid}
Armin Hornung, Kai~M Wurm, and Maren Bennewitz.
\newblock Humanoid robot localization in complex indoor environments.
\newblock In \emph{2010 IEEE/RSJ International Conference on Intelligent Robots
  and Systems}, pages 1690--1695. IEEE, 2010.

\bibitem[Huai et~al.(2021)Huai, Lin, Zhuang, and Shi]{huai2021consistent}
Jianzhu Huai, Yukai Lin, Yuan Zhuang, and Min Shi.
\newblock Consistent right-invariant fixed-lag smoother with application to
  visual inertial slam.
\newblock \emph{arXiv preprint arXiv:2102.08596}, 2021.

\bibitem[Huai and Huang(2018)]{huai2018robocentric}
Zheng Huai and Guoquan Huang.
\newblock Robocentric visual-inertial odometry.
\newblock In \emph{2018 IEEE/RSJ International Conference on Intelligent Robots
  and Systems (IROS)}, pages 6319--6326. IEEE, 2018.

\bibitem[Huang(2019)]{huang2019visual}
Guoquan Huang.
\newblock Visual-inertial navigation: A concise review.
\newblock In \emph{2019 international conference on robotics and automation
  (ICRA)}, pages 9572--9582. IEEE, 2019.

\bibitem[Huang et~al.(2009)Huang, Mourikis, and Roumeliotis]{huang2009first}
Guoquan~P Huang, Anastasios~I Mourikis, and Stergios~I Roumeliotis.
\newblock A first-estimates jacobian ekf for improving slam consistency.
\newblock In \emph{Experimental Robotics}, pages 373--382. Springer, 2009.

\bibitem[Huang et~al.(2010)Huang, Mourikis, and
  Roumeliotis]{huang2010observability}
Guoquan~P Huang, Anastasios~I Mourikis, and Stergios~I Roumeliotis.
\newblock Observability-based rules for designing consistent ekf slam
  estimators.
\newblock \emph{The International Journal of Robotics Research}, 29\penalty0
  (5):\penalty0 502--528, 2010.

\bibitem[Huang et~al.(2015)Huang, Xu, Mohammed, and Shu]{huang2015posture}
Jian Huang, Wenxia Xu, Samer Mohammed, and Zhen Shu.
\newblock Posture estimation and human support using wearable sensors and
  walking-aid robot.
\newblock \emph{Robotics and Autonomous Systems}, 73:\penalty0 24--43, 2015.

\bibitem[Joukov et~al.(2017)Joukov, {\'C}esi{\'c}, Westermann, Markovi{\'c},
  Kuli{\'c}, and Petrovi{\'c}]{joukov2017human}
Vladimir Joukov, Josip {\'C}esi{\'c}, Kevin Westermann, Ivan Markovi{\'c}, Dana
  Kuli{\'c}, and Ivan Petrovi{\'c}.
\newblock Human motion estimation on lie groups using imu measurements.
\newblock In \emph{2017 IEEE/RSJ International Conference on Intelligent Robots
  and Systems (IROS)}, pages 1965--1972. IEEE, 2017.

\bibitem[Joukov et~al.(2019)Joukov, {\'C}esi{\'c}, Westermann, Markovi{\'c},
  Petrovi{\'c}, and Kuli{\'c}]{joukov2019estimation}
Vladimir Joukov, Josip {\'C}esi{\'c}, Kevin Westermann, Ivan Markovi{\'c}, Ivan
  Petrovi{\'c}, and Dana Kuli{\'c}.
\newblock Estimation and observability analysis of human motion on lie groups.
\newblock \emph{IEEE transactions on cybernetics}, 50\penalty0 (3):\penalty0
  1321--1332, 2019.

\bibitem[Karatsidis et~al.(2017)Karatsidis, Bellusci, Schepers, De~Zee,
  Andersen, and Veltink]{karatsidis2017estimation}
Angelos Karatsidis, Giovanni Bellusci, H~Martin Schepers, Mark De~Zee,
  Michael~S Andersen, and Peter~H Veltink.
\newblock Estimation of ground reaction forces and moments during gait using
  only inertial motion capture.
\newblock \emph{Sensors}, 17\penalty0 (1):\penalty0 75, 2017.

\bibitem[Karcher(1977)]{karcher1977riemannian}
Hermann Karcher.
\newblock Riemannian center of mass and mollifier smoothing.
\newblock \emph{Communications on pure and applied mathematics}, 30\penalty0
  (5):\penalty0 509--541, 1977.

\bibitem[Kok et~al.(2014)Kok, Hol, and Sch{\"o}n]{kok2014optimization}
Manon Kok, Jeroen~D Hol, and Thomas~B Sch{\"o}n.
\newblock An optimization-based approach to human body motion capture using
  inertial sensors.
\newblock \emph{IFAC Proceedings Volumes}, 47\penalty0 (3):\penalty0 79--85,
  2014.

\bibitem[Kok et~al.(2017)Kok, Hol, and Sch{\"o}n]{kok2017using}
Manon Kok, Jeroen~D Hol, and Thomas~B Sch{\"o}n.
\newblock Using inertial sensors for position and orientation estimation.
\newblock \emph{arXiv preprint arXiv:1704.06053}, 2017.

\bibitem[Kong et~al.(2021)Kong, Payne, Council, and Johnson]{kong2021salted}
Nathan~J Kong, J~Joe Payne, George Council, and Aaron~M Johnson.
\newblock The salted kalman filter: Kalman filtering on hybrid dynamical
  systems.
\newblock \emph{Automatica}, 131:\penalty0 109752, 2021.

\bibitem[Kyrkjeb{\o}(2018)]{kyrkjebo2018inertial}
Erik Kyrkjeb{\o}.
\newblock Inertial human motion estimation for physical human-robot interaction
  using an interaction velocity update to reduce drift.
\newblock In \emph{Companion of the 2018 ACM/IEEE International Conference on
  Human-Robot Interaction}, pages 163--164, 2018.

\bibitem[Latella(2019)]{latella2019human}
Claudia Latella.
\newblock Human whole-body dynamics estimation for enhancing physical
  human-robot interaction.
\newblock \emph{arXiv preprint arXiv:1912.01136}, 2019.

\bibitem[Latella et~al.(2020)Latella, Rapetti, Tirupachuri, Darvish,
  Traversaro, and Pucci]{latella2020human}
Claudia Latella, Lorenzo Rapetti, Yeshasvi Tirupachuri, Kourosh Darvish, Silvio
  Traversaro, and Daniele Pucci.
\newblock The human wearable perception system: a human-robot collaboration
  application.
\newblock 2020.

\bibitem[Lee(2018)]{lee2018bayesian}
Taeyoung Lee.
\newblock Bayesian attitude estimation with the matrix fisher distribution on
  so (3).
\newblock \emph{IEEE Transactions on Automatic Control}, 63\penalty0
  (10):\penalty0 3377--3392, 2018.

\bibitem[Lee et~al.(2008)Lee, Leok, and McClamroch]{lee2008global}
Taeyoung Lee, Melvin Leok, and N~Harris McClamroch.
\newblock Global symplectic uncertainty propagation on so (3).
\newblock In \emph{2008 47th IEEE Conference on Decision and Control}, pages
  61--66. IEEE, 2008.

\bibitem[Li et~al.(2021)Li, Liu, Wang, Zhao, Zhang, Qiu, Zhou, Cai, Ni, and
  Cangelosi]{li2021real}
Jie Li, Xiao~Feng Liu, Zhelong Wang, Hongyu Zhao, Tingting Zhang, Sen Qiu,
  Xu~Zhou, Huili Cai, Rong~Rong Ni, and Angelo Cangelosi.
\newblock Real-time human motion capture based on wearable inertial sensor
  networks.
\newblock \emph{IEEE Internet of Things Journal}, 2021.

\bibitem[Lim et~al.(2021)Lim, Yu, Kim, Byun, Kwon, Park, and
  Myung]{lim2021walk}
Hyunjun Lim, Byeongho Yu, Yeeun Kim, Joowoong Byun, Soonpyo Kwon, Haewon Park,
  and Hyun Myung.
\newblock Walk-vio: Walking-motion-adaptive leg kinematic constraint
  visual-inertial odometry for quadruped robots.
\newblock \emph{arXiv preprint arXiv:2111.15164}, 2021.

\bibitem[Lin et~al.(2006{\natexlab{a}})Lin, Komsuoglu, and
  Koditschek]{lin2006sensor}
Pei-Chun Lin, Haldun Komsuoglu, and Daniel~E Koditschek.
\newblock Sensor data fusion for body state estimation in a hexapod robot with
  dynamical gaits.
\newblock \emph{IEEE Transactions on Robotics}, 22\penalty0 (5):\penalty0
  932--943, 2006{\natexlab{a}}.

\bibitem[Lin et~al.(2006{\natexlab{b}})Lin, Komsuoḡlu, and
  Koditschek]{lin2006legged}
Pei-Chun Lin, Haldun Komsuoḡlu, and Daniel~E Koditschek.
\newblock Legged odometry from body pose in a hexapod robot.
\newblock In \emph{Experimental Robotics IX}, pages 439--448. Springer,
  2006{\natexlab{b}}.

\bibitem[Long et~al.(2012)Long, Wolfe, Mashner, and Chirikjian]{long2012banana}
Andrew Long, Kevin Wolfe, Michael Mashner, and Gregory Chirikjian.
\newblock The banana distribution is gaussian: A localization study with
  exponential coordinates.
\newblock Robotics: Science and Systems Foundation, jul 2012.
\newblock \doi{10.15607/rss.2012.viii.034}.

\bibitem[Loper et~al.(2015)Loper, Mahmood, Romero, Pons-Moll, and
  Black]{loper2015smpl}
Matthew Loper, Naureen Mahmood, Javier Romero, Gerard Pons-Moll, and Michael~J
  Black.
\newblock Smpl: A skinned multi-person linear model.
\newblock \emph{ACM transactions on graphics (TOG)}, 34\penalty0 (6):\penalty0
  1--16, 2015.

\bibitem[Lopez-Nava and Munoz-Melendez(2016)]{lopez2016wearable}
Irvin~Hussein Lopez-Nava and Angelica Munoz-Melendez.
\newblock Wearable inertial sensors for human motion analysis: A review.
\newblock \emph{IEEE Sensors Journal}, 16\penalty0 (22):\penalty0 7821--7834,
  2016.

\bibitem[Lupton and Sukkarieh(2011)]{lupton2011visual}
Todd Lupton and Salah Sukkarieh.
\newblock Visual-inertial-aided navigation for high-dynamic motion in built
  environments without initial conditions.
\newblock \emph{IEEE Transactions on Robotics}, 28\penalty0 (1):\penalty0
  61--76, 2011.

\bibitem[Lynch and Park(2017)]{lynch2017modern}
Kevin~M Lynch and Frank~C Park.
\newblock \emph{Modern robotics}.
\newblock Cambridge University Press, 2017.

\bibitem[Lynen et~al.(2013)Lynen, Achtelik, Weiss, Chli, and
  Siegwart]{lynen2013robust}
Simon Lynen, Markus~W Achtelik, Stephan Weiss, Margarita Chli, and Roland
  Siegwart.
\newblock A robust and modular multi-sensor fusion approach applied to mav
  navigation.
\newblock In \emph{2013 IEEE/RSJ international conference on intelligent robots
  and systems}, pages 3923--3929. IEEE, 2013.

\bibitem[Mahony and Hamel(2017)]{mahony2017geometric}
Robert Mahony and Tarek Hamel.
\newblock A geometric nonlinear observer for simultaneous localisation and
  mapping.
\newblock In \emph{2017 IEEE 56th Annual Conference on Decision and Control
  (CDC)}, pages 2408--2415. IEEE, 2017.

\bibitem[Mahony et~al.(2008)Mahony, Hamel, and Pflimlin]{mahony2008nonlinear}
Robert Mahony, Tarek Hamel, and Jean-Michel Pflimlin.
\newblock Nonlinear complementary filters on the special orthogonal group.
\newblock \emph{IEEE Transactions on automatic control}, 53\penalty0
  (5):\penalty0 1203--1218, 2008.

\bibitem[Manton(2004)]{manton2004globally}
Jonathan~H Manton.
\newblock A globally convergent numerical algorithm for computing the centre of
  mass on compact lie groups.
\newblock In \emph{ICARCV 2004 8th Control, Automation, Robotics and Vision
  Conference, 2004.}, volume~3, pages 2211--2216. IEEE, 2004.

\bibitem[Masuya and Ayusawa(2020)]{masuya2020review}
Ken Masuya and Ko~Ayusawa.
\newblock A review of state estimation of humanoid robot targeting the center
  of mass, base kinematics, and external wrench.
\newblock \emph{Advanced Robotics}, 34\penalty0 (21-22):\penalty0 1380--1389,
  2020.

\bibitem[Masuya and Sugihara(2015)]{masuya2015dead}
Ken Masuya and Tomomichi Sugihara.
\newblock Dead reckoning for biped robots that suffers less from foot contact
  condition based on anchoring pivot estimation.
\newblock \emph{Advanced Robotics}, 29\penalty0 (12):\penalty0 785--799, 2015.

\bibitem[McGrath and Stirling(2020)]{mcgrath2020body}
Timothy McGrath and Leia Stirling.
\newblock Body-worn imu human skeletal pose estimation using a factor
  graph-based optimization framework.
\newblock \emph{Sensors}, 20\penalty0 (23):\penalty0 6887, 2020.

\bibitem[Metta et~al.(2008)Metta, Sandini, Vernon, Natale, and
  Nori]{metta2008icub}
Giorgio Metta, Giulio Sandini, David Vernon, Lorenzo Natale, and Francesco
  Nori.
\newblock The icub humanoid robot: an open platform for research in embodied
  cognition.
\newblock In \emph{Proceedings of the 8th workshop on performance metrics for
  intelligent systems}, pages 50--56, 2008.

\bibitem[Miezal et~al.(2016)Miezal, Taetz, and Bleser]{miezal2016inertial}
Markus Miezal, Bertram Taetz, and Gabriele Bleser.
\newblock On inertial body tracking in the presence of model calibration
  errors.
\newblock \emph{Sensors}, 16\penalty0 (7):\penalty0 1132, 2016.

\bibitem[Miezal et~al.(2017)Miezal, Taetz, and Bleser]{miezal2017real}
Markus Miezal, Bertram Taetz, and Gabriele Bleser.
\newblock Real-time inertial lower body kinematics and ground contact
  estimation at anatomical foot points for agile human locomotion.
\newblock In \emph{2017 IEEE international conference on robotics and
  automation (ICRA)}, pages 3256--3263. IEEE, 2017.

\bibitem[Mifsud et~al.(2015)Mifsud, Benallegue, and
  Lamiraux]{mifsud2015estimation}
Alexis Mifsud, Mehdi Benallegue, and Florent Lamiraux.
\newblock Estimation of contact forces and floating base kinematics of a
  humanoid robot using only inertial measurement units.
\newblock In \emph{2015 IEEE/RSJ International Conference on Intelligent Robots
  and Systems (IROS)}, pages 3374--3379. IEEE, 2015.

\bibitem[Mitjans et~al.(2021)Mitjans, Theofanidis, Collimore, Disney, Levine,
  Awad, and Tron]{mitjans2021visual}
Marc Mitjans, Michail Theofanidis, Ashley~N Collimore, Madelaine~L Disney,
  David~M Levine, Louis~N Awad, and Roberto Tron.
\newblock Visual-inertial filtering for human walking quantification.
\newblock In \emph{2021 IEEE International Conference on Robotics and
  Automation (ICRA)}, pages 13510--13516. IEEE, 2021.

\bibitem[Monzani et~al.(2000)Monzani, Baerlocher, Boulic, and
  Thalmann]{monzani2000using}
Jean-S{\'e}bastien Monzani, Paolo Baerlocher, Ronan Boulic, and Daniel
  Thalmann.
\newblock Using an intermediate skeleton and inverse kinematics for motion
  retargeting.
\newblock In \emph{Computer Graphics Forum}, volume~19, pages 11--19. Wiley
  Online Library, 2000.

\bibitem[Mourikis and Roumeliotis(2007)]{mourikis2007multi}
Anastasios~I Mourikis and Stergios~I Roumeliotis.
\newblock A multi-state constraint kalman filter for vision-aided inertial
  navigation.
\newblock In \emph{Proceedings 2007 IEEE International Conference on Robotics
  and Automation}, pages 3565--3572. IEEE, 2007.

\bibitem[M{\"u}ller(2018)]{muller2018screw}
Andreas M{\"u}ller.
\newblock Screw and lie group theory in multibody kinematics.
\newblock \emph{Multibody System Dynamics}, 43\penalty0 (1):\penalty0 37--70,
  2018.

\bibitem[Munthe-Kaas(1998)]{munthe1998runge}
Hans Munthe-Kaas.
\newblock Runge-kutta methods on lie groups.
\newblock \emph{BIT Numerical Mathematics}, 38\penalty0 (1):\penalty0 92--111,
  1998.

\bibitem[Murray et~al.(2017)Murray, Li, and Sastry]{murray2017mathematical}
Richard~M Murray, Zexiang Li, and S~Shankar Sastry.
\newblock \emph{A mathematical introduction to robotic manipulation}.
\newblock CRC press, 2017.

\bibitem[Natale et~al.(2017)Natale, Bartolozzi, Pucci, Wykowska, and
  Metta]{natale2017icub}
Lorenzo Natale, Chiara Bartolozzi, Daniele Pucci, Agnieszka Wykowska, and
  Giorgio Metta.
\newblock icub: The not-yet-finished story of building a robot child.
\newblock \emph{Science Robotics}, 2\penalty0 (13):\penalty0 eaaq1026, 2017.

\bibitem[Natale et~al.(2021)Natale, Bartolozzi, Nori, Sandini, and
  Metta]{natale2021icub}
Lorenzo Natale, Chiara Bartolozzi, Francesco Nori, Giulio Sandini, and Giorgio
  Metta.
\newblock icub.
\newblock \emph{arXiv preprint arXiv:2105.02313}, 2021.

\bibitem[Nisar et~al.(2019)Nisar, Foehn, Falanga, and
  Scaramuzza]{nisar2019vimo}
Barza Nisar, Philipp Foehn, Davide Falanga, and Davide Scaramuzza.
\newblock Vimo: Simultaneous visual inertial model-based odometry and force
  estimation.
\newblock \emph{IEEE Robotics and Automation Letters}, 4\penalty0 (3):\penalty0
  2785--2792, 2019.

\bibitem[Nobili et~al.(2017)Nobili, Camurri, Barasuol, Focchi, Caldwell,
  Semini, and Fallon]{nobili2017heterogeneous}
Simona Nobili, Marco Camurri, Victor Barasuol, Michele Focchi, Darwin Caldwell,
  Claudio Semini, and Maurice~F Fallon.
\newblock Heterogeneous sensor fusion for accurate state estimation of dynamic
  legged robots.
\newblock 2017.

\bibitem[Nori et~al.(2015{\natexlab{a}})Nori, Kuppuswamy, and
  Traversaro]{nori2015simultaneous}
Francesco Nori, Naveen Kuppuswamy, and Silvio Traversaro.
\newblock Simultaneous state and dynamics estimation in articulated structures.
\newblock In \emph{2015 IEEE/RSJ International Conference on Intelligent Robots
  and Systems (IROS)}, pages 3380--3386. IEEE, 2015{\natexlab{a}}.

\bibitem[Nori et~al.(2015{\natexlab{b}})Nori, Traversaro, Eljaik, Romano,
  Del~Prete, and Pucci]{nori2015icub}
Francesco Nori, Silvio Traversaro, Jorhabib Eljaik, Francesco Romano, Andrea
  Del~Prete, and Daniele Pucci.
\newblock icub whole-body control through force regulation on rigid
  non-coplanar contacts.
\newblock \emph{Frontiers in Robotics and AI}, 2:\penalty0 6,
  2015{\natexlab{b}}.

\bibitem[Olfati-Saber(2001)]{olfati2001nonlinear}
Reza Olfati-Saber.
\newblock \emph{Nonlinear control of underactuated mechanical systems with
  application to robotics and aerospace vehicles}.
\newblock PhD thesis, Massachusetts Institute of Technology, 2001.

\bibitem[Oriolo et~al.(2016)Oriolo, Paolillo, Rosa, and
  Vendittelli]{oriolo2016humanoid}
Giuseppe Oriolo, Antonio Paolillo, Lorenzo Rosa, and Marilena Vendittelli.
\newblock Humanoid odometric localization integrating kinematic, inertial and
  visual information.
\newblock \emph{Autonomous Robots}, 40\penalty0 (5):\penalty0 867--879, 2016.

\bibitem[Park et~al.(1995)Park, Bobrow, and Ploen]{park1995lie}
Frank~C Park, James~E Bobrow, and Scott~R Ploen.
\newblock A lie group formulation of robot dynamics.
\newblock \emph{The International journal of robotics research}, 14\penalty0
  (6):\penalty0 609--618, 1995.

\bibitem[Park(1991)]{park1991optimal}
Frank~Chongwoo Park.
\newblock \emph{The optimal kinematic design of mechanisms}.
\newblock Harvard University, 1991.

\bibitem[Parmiggiani et~al.(2012)Parmiggiani, Maggiali, Natale, Nori, Schmitz,
  Tsagarakis, Victor, Becchi, Sandini, and Metta]{parmiggiani2012design}
Alberto Parmiggiani, Marco Maggiali, Lorenzo Natale, Francesco Nori, Alexander
  Schmitz, Nikos Tsagarakis, Jose~Santos Victor, Francesco Becchi, Giulio
  Sandini, and Giorgio Metta.
\newblock The design of the icub humanoid robot.
\newblock \emph{International journal of humanoid robotics}, 9\penalty0
  (04):\penalty0 1250027, 2012.

\bibitem[Pham et~al.(2018)Pham, Yang, and Sheng]{pham2018sensor}
Minh Pham, Dan Yang, and Weihua Sheng.
\newblock A sensor fusion approach to indoor human localization based on
  environmental and wearable sensors.
\newblock \emph{IEEE Transactions on Automation Science and Engineering},
  16\penalty0 (1):\penalty0 339--350, 2018.

\bibitem[Phogat and Chang(2020{\natexlab{a}})]{phogat2020discrete}
Karmvir~Singh Phogat and Dong~Eui Chang.
\newblock Discrete-time invariant extended kalman filter on matrix lie groups.
\newblock 30\penalty0 (12):\penalty0 4449--4462, jun 2020{\natexlab{a}}.
\newblock \doi{10.1002/rnc.4984}.

\bibitem[Phogat and Chang(2020{\natexlab{b}})]{phogat2020invariant}
Karmvir~Singh Phogat and Dong~Eui Chang.
\newblock Invariant extended kalman filter on matrix lie groups.
\newblock 114:\penalty0 108812, apr 2020{\natexlab{b}}.
\newblock \doi{10.1016/j.automatica.2020.108812}.

\bibitem[Piperakis(2019)]{piperakis2019robust}
Stylianos Piperakis.
\newblock \emph{Robust nonlinear state estimation for humanoid robots}.
\newblock PhD thesis,
  $\Pi$$\alpha$$\nu$$\varepsilon$$\pi$$\iota$$\sigma$$\tau$$\acute{\eta}$$\mu$$\iota$o
  K$\rho$$\acute{\eta}$$\tau$$\eta$$\varsigma$.
  $\Sigma$$\chi$o$\lambda$$\acute{\eta}$
  $\Theta$$\varepsilon$$\tau$$\iota$$\kappa$$\acute{\omega}$$\nu$
  $\kappa$$\alpha$$\iota$
  T$\varepsilon$$\chi$$\nu$o$\lambda$o$\gamma$$\iota$$\kappa$$\acute{\omega}$$\nu$
  E$\pi$$\iota$$\sigma$$\tau$$\eta$$\mu$$\acute{\omega}$$\nu$.
  T$\mu$$\acute{\eta}$$\mu$$\alpha$~…, 2019.

\bibitem[Piperakis and Trahanias(2016)]{piperakis2016non}
Stylianos Piperakis and Panos Trahanias.
\newblock Non-linear zmp based state estimation for humanoid robot locomotion.
\newblock In \emph{2016 IEEE-RAS 16th International Conference on Humanoid
  Robots (Humanoids)}, pages 202--209. IEEE, 2016.

\bibitem[Piperakis et~al.(2018)Piperakis, Koskinopoulou, and
  Trahanias]{piperakis2018nonlinear}
Stylianos Piperakis, Maria Koskinopoulou, and Panos Trahanias.
\newblock Nonlinear state estimation for humanoid robot walking.
\newblock \emph{IEEE Robotics and Automation Letters}, 3\penalty0 (4):\penalty0
  3347--3354, 2018.

\bibitem[Piperakis et~al.(2019)Piperakis, Kanoulas, Tsagarakis, and
  Trahanias]{piperakis2019outlier}
Stylianos Piperakis, Dimitrios Kanoulas, Nikos~G Tsagarakis, and Panos
  Trahanias.
\newblock Outlier-robust state estimation for humanoid robots.
\newblock In \emph{2019 IEEE/RSJ International Conference on Intelligent Robots
  and Systems (IROS)}, pages 706--713. IEEE, 2019.

\bibitem[Pucci et~al.(2016)Pucci, Romano, Traversaro, and
  Nori]{pucci2016highly}
Daniele Pucci, Francesco Romano, Silvio Traversaro, and Francesco Nori.
\newblock Highly dynamic balancing via force control.
\newblock In \emph{2016 IEEE-RAS 16th International Conference on Humanoid
  Robots (Humanoids)}, pages 141--141. IEEE, 2016.

\bibitem[Qin et~al.(2020)Qin, Yu, Chen, Huang, Fu, Ming, and Tao]{qin2020novel}
Mingyue Qin, Zhangguo Yu, Xuechao Chen, Qiang Huang, Chenglong Fu, Aiguo Ming,
  and Chunjing Tao.
\newblock A novel foot contact probability estimator for biped robot state
  estimation.
\newblock In \emph{2020 IEEE International Conference on Mechatronics and
  Automation (ICMA)}, pages 1901--1906. IEEE, 2020.

\bibitem[Qin et~al.(2018)Qin, Li, and Shen]{qin2018vins}
Tong Qin, Peiliang Li, and Shaojie Shen.
\newblock Vins-mono: A robust and versatile monocular visual-inertial state
  estimator.
\newblock \emph{IEEE Transactions on Robotics}, 34\penalty0 (4):\penalty0
  1004--1020, 2018.

\bibitem[Qiu et~al.(2016)Qiu, Wang, Zhao, and Hu]{qiu2016using}
Sen Qiu, Zhelong Wang, Hongyu Zhao, and Huosheng Hu.
\newblock Using distributed wearable sensors to measure and evaluate human
  lower limb motions.
\newblock \emph{IEEE Transactions on Instrumentation and Measurement},
  65\penalty0 (4):\penalty0 939--950, 2016.

\bibitem[Raghavan et~al.(2018)Raghavan, Kanoulas, Zhou, Caldwell, and
  Tsagarakis]{raghavan2018study}
Vignesh~Sushrutha Raghavan, Dimitrios Kanoulas, Chengxu Zhou, Darwin~G
  Caldwell, and Nikos~G Tsagarakis.
\newblock A study on low-drift state estimation for humanoid locomotion, using
  lidar and kinematic-inertial data fusion.
\newblock In \emph{2018 IEEE-RAS 18th International Conference on Humanoid
  Robots (Humanoids)}, pages 1--8. IEEE, 2018.

\bibitem[Ramadoss et~al.(2021)Ramadoss, Romualdi, Dafarra, Andrade~Chavez,
  Traversaro, and Pucci]{ramadoss2021diligent}
Prashanth Ramadoss, Giulio Romualdi, Stefano Dafarra, Francisco~Javier
  Andrade~Chavez, Silvio Traversaro, and Daniele Pucci.
\newblock Diligent-kio: A proprioceptive base estimator for humanoid robots
  using extended kalman filtering on matrix lie groups.
\newblock In \emph{2021 IEEE International Conference on Robotics and
  Automation (ICRA)}, pages 2904--2910, 2021.
\newblock \doi{10.1109/ICRA48506.2021.9561248}.

\bibitem[Rapetti et~al.(2020)Rapetti, Tirupachuri, Darvish, Dafarra, Nava,
  Latella, and Pucci]{rapetti2020model}
Lorenzo Rapetti, Yeshasvi Tirupachuri, Kourosh Darvish, Stefano Dafarra,
  Gabriele Nava, Claudia Latella, and Daniele Pucci.
\newblock Model-based real-time motion tracking using dynamical inverse
  kinematics.
\newblock \emph{Algorithms}, 13\penalty0 (10):\penalty0 266, 2020.

\bibitem[Rapetti et~al.(2021)Rapetti, Tirupachuri, Ranavolo, Kawakami,
  Yoshiike, and Pucci]{rapetti2021shared}
Lorenzo Rapetti, Yeshasvi Tirupachuri, Alberto Ranavolo, Tomohiro Kawakami,
  Takahide Yoshiike, and Daniele Pucci.
\newblock Shared control of robot-robot collaborative lifting with agent
  postural and force ergonomic optimization.
\newblock In \emph{2021 IEEE International Conference on Robotics and
  Automation (ICRA)}, pages 9840--9847, 2021.
\newblock \doi{10.1109/ICRA48506.2021.9561623}.

\bibitem[Roetenberg et~al.(2009)Roetenberg, Luinge, and
  Slycke]{roetenberg2009xsens}
Daniel Roetenberg, Henk Luinge, and Per Slycke.
\newblock Xsens mvn: Full 6dof human motion tracking using miniature inertial
  sensors.
\newblock \emph{Xsens Motion Technologies BV, Tech. Rep}, 1, 2009.

\bibitem[Romualdi et~al.(2020)Romualdi, Dafarra, Hu, Ramadoss, Chavez,
  Traversaro, and Pucci]{romualdi2020benchmarking}
Giulio Romualdi, Stefano Dafarra, Yue Hu, Prashanth Ramadoss, Francisco
  Javier~Andrade Chavez, Silvio Traversaro, and Daniele Pucci.
\newblock A benchmarking of dcm-based architectures for position, velocity and
  torque-controlled humanoid robots.
\newblock \emph{International Journal of Humanoid Robotics}, 17\penalty0
  (01):\penalty0 1950034, 2020.

\bibitem[Rotella et~al.(2014)Rotella, Bloesch, Righetti, and
  Schaal]{rotella2014state}
Nicholas Rotella, Michael Bloesch, Ludovic Righetti, and Stefan Schaal.
\newblock State estimation for a humanoid robot.
\newblock In \emph{2014 {IEEE}/{RSJ} International Conference on Intelligent
  Robots and Systems}. {IEEE}, sep 2014.
\newblock \doi{10.1109/iros.2014.6942674}.

\bibitem[S{\"a}rkk{\"a}(2013)]{sarkka2013bayesian}
Simo S{\"a}rkk{\"a}.
\newblock \emph{Bayesian filtering and smoothing}.
\newblock Number~3. Cambridge University Press, 2013.

\bibitem[Schubert et~al.(2018)Schubert, Goll, Demmel, Usenko, St{\"u}ckler, and
  Cremers]{schubert2018tum}
David Schubert, Thore Goll, Nikolaus Demmel, Vladyslav Usenko, J{\"o}rg
  St{\"u}ckler, and Daniel Cremers.
\newblock The tum vi benchmark for evaluating visual-inertial odometry.
\newblock In \emph{2018 IEEE/RSJ International Conference on Intelligent Robots
  and Systems (IROS)}, pages 1680--1687. IEEE, 2018.

\bibitem[Schwendner et~al.(2014)Schwendner, Joyeux, and
  Kirchner]{schwendner2014using}
Jakob Schwendner, Sylvain Joyeux, and Frank Kirchner.
\newblock Using embodied data for localization and mapping.
\newblock \emph{Journal of Field Robotics}, 31\penalty0 (2):\penalty0 263--295,
  2014.

\bibitem[Seth et~al.(2011)Seth, Sherman, Reinbolt, and Delp]{seth2011opensim}
Ajay Seth, Michael Sherman, Jeffrey~A Reinbolt, and Scott~L Delp.
\newblock Opensim: a musculoskeletal modeling and simulation framework for in
  silico investigations and exchange.
\newblock \emph{Procedia Iutam}, 2:\penalty0 212--232, 2011.

\bibitem[Shah et~al.(2012)Shah, Eastman, and Hong]{shah2012overview}
Mili Shah, Roger~D Eastman, and Tsai Hong.
\newblock An overview of robot-sensor calibration methods for evaluation of
  perception systems.
\newblock In \emph{Proceedings of the Workshop on Performance Metrics for
  Intelligent Systems}, pages 15--20, 2012.

\bibitem[Shen et~al.(2015)Shen, Michael, and Kumar]{shen2015tightly}
Shaojie Shen, Nathan Michael, and Vijay Kumar.
\newblock Tightly-coupled monocular visual-inertial fusion for autonomous
  flight of rotorcraft mavs.
\newblock In \emph{2015 IEEE International Conference on Robotics and
  Automation (ICRA)}, pages 5303--5310. IEEE, 2015.

\bibitem[Siciliano et~al.(2010)Siciliano, Sciavicco, Villani, and
  Oriolo]{siciliano2010robotics}
Bruno Siciliano, Lorenzo Sciavicco, Luigi Villani, and Giuseppe Oriolo.
\newblock \emph{Robotics: modelling, planning and control}.
\newblock Springer Science \& Business Media, 2010.

\bibitem[Simon(2006)]{simon2006optimal}
Dan Simon.
\newblock \emph{Optimal state estimation: Kalman, H infinity, and nonlinear
  approaches}.
\newblock John Wiley \& Sons, 2006.

\bibitem[Sola et~al.(2018)Sola, Deray, and Atchuthan]{sola2018micro}
Joan Sola, Jeremie Deray, and Dinesh Atchuthan.
\newblock A micro lie theory for state estimation in robotics.
\newblock \emph{arXiv preprint arXiv:1812.01537}, 2018.

\bibitem[Sorrentino et~al.(2020)Sorrentino, Andrade~Chavez, Latella, Fiorio,
  Traversaro, Rapetti, Tirupachuri, Guedelha, Maggiali, Dussoni,
  et~al.]{sorrentino2020novel}
Ines Sorrentino, Francisco~Javier Andrade~Chavez, Claudia Latella, Luca Fiorio,
  Silvio Traversaro, Lorenzo Rapetti, Yeshasvi Tirupachuri, Nuno Guedelha,
  Marco Maggiali, Simeone Dussoni, et~al.
\newblock A novel sensorised insole for sensing feet pressure distributions.
\newblock \emph{Sensors}, 20\penalty0 (3):\penalty0 747, 2020.

\bibitem[Stasse et~al.(2006)Stasse, Davison, Sellaouti, and
  Yokoi]{stasse2006real}
Olivier Stasse, Andrew~J Davison, Ramzi Sellaouti, and Kazuhito Yokoi.
\newblock Real-time 3d slam for humanoid robot considering pattern generator
  information.
\newblock In \emph{2006 IEEE/RSJ International Conference on Intelligent Robots
  and Systems}, pages 348--355. IEEE, 2006.

\bibitem[Strasdat et~al.(2010)Strasdat, Montiel, and Davison]{strasdat2010real}
Hauke Strasdat, JMM Montiel, and Andrew~J Davison.
\newblock Real-time monocular slam: Why filter?
\newblock In \emph{2010 IEEE International Conference on Robotics and
  Automation}, pages 2657--2664. IEEE, 2010.

\bibitem[Sturm et~al.(2012)Sturm, Burgard, and Cremers]{sturm2012evaluating}
J{\"u}rgen Sturm, Wolfram Burgard, and Daniel Cremers.
\newblock Evaluating egomotion and structure-from-motion approaches using the
  tum rgb-d benchmark.
\newblock In \emph{Proc. of the Workshop on Color-Depth Camera Fusion in
  Robotics at the IEEE/RJS International Conference on Intelligent Robot
  Systems (IROS)}, 2012.

\bibitem[Su and Lee(1992)]{su1992manipulation}
Shun-Feng Su and C.S.G. Lee.
\newblock Manipulation and propagation of uncertainty and verification of
  applicability of actions in assembly tasks.
\newblock 22\penalty0 (6):\penalty0 1376--1389, 1992.
\newblock \doi{10.1109/21.199463}.

\bibitem[Sy et~al.(2020{\natexlab{a}})Sy, Lovell, and
  Redmond]{sy2020estimatinga}
Luke Sy, Nigel~H Lovell, and Stephen~J Redmond.
\newblock Estimating lower limb kinematics using a lie group constrained ekf
  and a reduced wearable imu count.
\newblock In \emph{2020 8th IEEE RAS/EMBS International Conference for
  Biomedical Robotics and Biomechatronics (BioRob)}, pages 310--315. IEEE,
  2020{\natexlab{a}}.

\bibitem[Sy et~al.(2020{\natexlab{b}})Sy, Raitor, Del~Rosario, Khamis, Kark,
  Lovell, and Redmond]{sy2020estimating}
Luke Sy, Michael Raitor, Michael Del~Rosario, Heba Khamis, Lauren Kark, Nigel~H
  Lovell, and Stephen~J Redmond.
\newblock Estimating lower limb kinematics using a reduced wearable sensor
  count.
\newblock \emph{IEEE Transactions on Biomedical Engineering}, 68\penalty0
  (4):\penalty0 1293--1304, 2020{\natexlab{b}}.

\bibitem[Tagliapietra et~al.(2018)Tagliapietra, Modenese, Ceseracciu,
  Mazz{\`a}, and Reggiani]{tagliapietra2018validation}
Luca Tagliapietra, L~Modenese, E~Ceseracciu, C~Mazz{\`a}, and M~Reggiani.
\newblock Validation of a model-based inverse kinematics approach based on
  wearable inertial sensors.
\newblock \emph{Computer methods in biomechanics and biomedical engineering},
  21\penalty0 (16):\penalty0 834--844, 2018.

\bibitem[Tao et~al.(2012)Tao, Liu, Zheng, and Feng]{tao2012gait}
Weijun Tao, Tao Liu, Rencheng Zheng, and Hutian Feng.
\newblock Gait analysis using wearable sensors.
\newblock \emph{Sensors}, 12\penalty0 (2):\penalty0 2255--2283, 2012.

\bibitem[Teng et~al.(2021)Teng, Mueller, and Sreenath]{teng2021legged}
Sangli Teng, Mark~Wilfried Mueller, and Koushil Sreenath.
\newblock Legged robot state estimation in slippery environments using
  invariant extended kalman filter with velocity update.
\newblock In \emph{2021 IEEE International Conference on Robotics and
  Automation (ICRA)}, pages 3104--3110, 2021.
\newblock \doi{10.1109/ICRA48506.2021.9561313}.

\bibitem[Thrun(2002)]{thrun2002probabilistic}
Sebastian Thrun.
\newblock Probabilistic robotics.
\newblock \emph{Communications of the ACM}, 45\penalty0 (3):\penalty0 52--57,
  2002.

\bibitem[Tirupachuri(2020)]{tirupachuri2020enabling}
Yeshasvi Tirupachuri.
\newblock Enabling human-robot collaboration via holistic human perception and
  partner-aware control.
\newblock \emph{arXiv preprint arXiv:2004.10847}, 2020.

\bibitem[Tirupachuri et~al.(2021)Tirupachuri, Ramadoss, Rapetti, Latella,
  Darvish, Traversaro, and Pucci]{tirupachuri2021online}
Yeshasvi Tirupachuri, Prashanth Ramadoss, Lorenzo Rapetti, Claudia Latella,
  Kourosh Darvish, Silvio Traversaro, and Daniele Pucci.
\newblock Online non-collocated estimation of payload and articular stress for
  real-time human ergonomy assessment.
\newblock \emph{IEEE Access}, 9:\penalty0 123260--123279, 2021.

\bibitem[Titterton et~al.(2004)Titterton, Weston, and
  Weston]{titterton2004strapdown}
David Titterton, John~L Weston, and John Weston.
\newblock \emph{Strapdown inertial navigation technology}, volume~17.
\newblock IET, 2004.

\bibitem[Traversaro(2017)]{traversaro2017modelling}
S~Traversaro.
\newblock \emph{Modelling, estimation and identification of humanoid robots
  dynamics}.
\newblock PhD thesis, Ph. D. dissertation, Italian Institute of Technology, 4
  2017, https~…, 2017.

\bibitem[Traversaro and Saccon(2019)]{traversaro2019multibody}
S~Traversaro and Alessandro Saccon.
\newblock Multibody dynamics notation (version 2).
\newblock 2019.

\bibitem[Varin and Kuindersma(2018)]{varin2018constrained}
Patrick Varin and Scott Kuindersma.
\newblock A constrained kalman filter for rigid body systems with frictional
  contact.
\newblock In \emph{WAFR}, pages 474--490, 2018.

\bibitem[Vigne et~al.(2019)Vigne, El~Khoury, Di~Meglio, and
  Petit]{vigne2019state}
Matthieu Vigne, Antonio El~Khoury, Florent Di~Meglio, and Nicolas Petit.
\newblock State estimation for a legged robot with multiple flexibilities using
  imu s: A kinematic approach.
\newblock \emph{IEEE Robotics and Automation Letters}, 5\penalty0 (1):\penalty0
  195--202, 2019.

\bibitem[Von~Marcard et~al.(2017)Von~Marcard, Rosenhahn, Black, and
  Pons-Moll]{von2017sparse}
Timo Von~Marcard, Bodo Rosenhahn, Michael~J Black, and Gerard Pons-Moll.
\newblock Sparse inertial poser: Automatic 3d human pose estimation from sparse
  imus.
\newblock In \emph{Computer Graphics Forum}, volume~36, pages 349--360. Wiley
  Online Library, 2017.

\bibitem[Waldron and Schmiedeler(2016)]{waldron2016kinematics}
Kenneth~J. Waldron and James Schmiedeler.
\newblock \emph{Kinematics}, pages 11--36.
\newblock Springer International Publishing, Cham, 2016.
\newblock ISBN 978-3-319-32552-1.
\newblock \doi{10.1007/978-3-319-32552-1_2}.
\newblock URL \url{https://doi.org/10.1007/978-3-319-32552-1_2}.

\bibitem[Wang and Chirikjian(2008)]{wang2008nonparametric}
Yunfeng Wang and Gregory~S. Chirikjian.
\newblock Nonparametric second-order theory of error propagation on motion
  groups.
\newblock 27\penalty0 (11-12):\penalty0 1258--1273, nov 2008.
\newblock \doi{10.1177/0278364908097583}.

\bibitem[Wang and Chirikjian(2006)]{wang2006error}
Yunfeng Wang and G.S. Chirikjian.
\newblock Error propagation on the euclidean group with applications to
  manipulator kinematics.
\newblock 22\penalty0 (4):\penalty0 591--602, aug 2006.
\newblock \doi{10.1109/tro.2006.878978}.

\bibitem[Weiss and Siegwart(2011)]{weiss2011real}
Stephan Weiss and Roland Siegwart.
\newblock Real-time metric state estimation for modular vision-inertial
  systems.
\newblock In \emph{2011 IEEE international conference on robotics and
  automation}, pages 4531--4537. IEEE, 2011.

\bibitem[Weisstein(2021{\natexlab{a}})]{weisstein2021convergentseries}
Eric~W. Weisstein.
\newblock Convergent series. from mathworld--a wolfram web resource.,
  2021{\natexlab{a}}.
\newblock URL \url{https://mathworld.wolfram.com/ConvergentSeries.html}.
\newblock Last visited on 2022-01-05.

\bibitem[Weisstein(2021{\natexlab{b}})]{weisstein2021pointsettopology}
Eric~W. Weisstein.
\newblock Point-set topology. from mathworld--a wolfram web resource.,
  2021{\natexlab{b}}.
\newblock URL \url{https://mathworld.wolfram.com/Point-SetTopology.html}.
\newblock Last visited on 2022-01-05.

\bibitem[Weisstein(2021{\natexlab{c}})]{weisstein2021taylorseries}
Eric~W. Weisstein.
\newblock Taylor series. from mathworld--a wolfram web resource.,
  2021{\natexlab{c}}.
\newblock URL \url{https://mathworld.wolfram.com/TaylorSeries.html}.
\newblock Last visited on 2022-01-05.

\bibitem[Winter(2009)]{winter2009biomechanics}
David~A Winter.
\newblock \emph{Biomechanics and motor control of human movement}.
\newblock John Wiley \& Sons, 2009.

\bibitem[Wisth et~al.(2019)Wisth, Camurri, and Fallon]{wisth2019robust}
David Wisth, Marco Camurri, and Maurice Fallon.
\newblock Robust legged robot state estimation using factor graph optimization.
\newblock \emph{IEEE Robotics and Automation Letters}, 4\penalty0 (4):\penalty0
  4507--4514, 2019.

\bibitem[Wisth et~al.(2020)Wisth, Camurri, and Fallon]{wisth2020preintegrated}
David Wisth, Marco Camurri, and Maurice Fallon.
\newblock Preintegrated velocity bias estimation to overcome contact
  nonlinearities in legged robot odometry.
\newblock In \emph{2020 IEEE International Conference on Robotics and
  Automation (ICRA)}, pages 392--398. IEEE, 2020.

\bibitem[Wisth et~al.(2021{\natexlab{a}})Wisth, Camurri, Das, and
  Fallon]{wisth2021unified}
David Wisth, Marco Camurri, Sandipan Das, and Maurice Fallon.
\newblock Unified multi-modal landmark tracking for tightly coupled
  lidar-visual-inertial odometry.
\newblock \emph{IEEE Robotics and Automation Letters}, 6\penalty0 (2):\penalty0
  1004--1011, 2021{\natexlab{a}}.

\bibitem[Wisth et~al.(2021{\natexlab{b}})Wisth, Camurri, and
  Fallon]{wisth2021vilens}
David Wisth, Marco Camurri, and Maurice Fallon.
\newblock Vilens: Visual, inertial, lidar, and leg odometry for all-terrain
  legged robots.
\newblock \emph{arXiv preprint arXiv:2107.07243}, 2021{\natexlab{b}}.

\bibitem[Wolfe and Mashner(2011)]{wolfe2011bayesian}
Kevin~C Wolfe and Michael Mashner.
\newblock Bayesian fusion on lie groups.
\newblock \emph{Journal of Algebraic Statistics}, 2\penalty0 (1), 2011.

\bibitem[Wong et~al.(2015)Wong, Zhang, Lo, and Yang]{wong2015wearable}
Charence Wong, Zhi-Qiang Zhang, Benny Lo, and Guang-Zhong Yang.
\newblock Wearable sensing for solid biomechanics: A review.
\newblock \emph{IEEE Sensors Journal}, 15\penalty0 (5):\penalty0 2747--2760,
  2015.

\bibitem[Wu et~al.(2017)Wu, Zhang, Su, Huang, and Dissanayake]{wu2017invariant}
Kanzhi Wu, Teng Zhang, Daobilige Su, Shoudong Huang, and Gamini Dissanayake.
\newblock An invariant-ekf vins algorithm for improving consistency.
\newblock In \emph{2017 IEEE/RSJ International Conference on Intelligent Robots
  and Systems (IROS)}, pages 1578--1585. IEEE, 2017.

\bibitem[Xie et~al.(2021)Xie, Wang, Iqbal, Guo, Fidler, and
  Shkurti]{xie2021physics}
Kevin Xie, Tingwu Wang, Umar Iqbal, Yunrong Guo, Sanja Fidler, and Florian
  Shkurti.
\newblock Physics-based human motion estimation and synthesis from videos.
\newblock In \emph{Proceedings of the IEEE/CVF International Conference on
  Computer Vision}, pages 11532--11541, 2021.

\bibitem[Xinjilefu et~al.(2014)Xinjilefu, Feng, Huang, and
  Atkeson]{xinjilefu2014decoupled}
X~Xinjilefu, Siyuan Feng, Weiwei Huang, and Christopher~G Atkeson.
\newblock Decoupled state estimation for humanoids using full-body dynamics.
\newblock In \emph{2014 IEEE International Conference on Robotics and
  Automation (ICRA)}, pages 195--201. IEEE, 2014.

\bibitem[Xinjilefu et~al.(2015)Xinjilefu, Feng, and
  Atkeson]{xinjilefu2015center}
X~Xinjilefu, Siyuan Feng, and Christopher~G Atkeson.
\newblock Center of mass estimator for humanoids and its application in
  modelling error compensation, fall detection and prevention.
\newblock In \emph{2015 IEEE-RAS 15th International Conference on Humanoid
  Robots (Humanoids)}, pages 67--73. IEEE, 2015.

\bibitem[Yan et~al.(2017)Yan, Li, Li, and Zhang]{yan2017wearable}
Xuzhong Yan, Heng Li, Angus~R Li, and Hong Zhang.
\newblock Wearable imu-based real-time motion warning system for construction
  workers' musculoskeletal disorders prevention.
\newblock \emph{Automation in Construction}, 74:\penalty0 2--11, 2017.

\bibitem[Yang et~al.(2019)Yang, Kumar, Gu, Zhang, Travers, and
  Choset]{yang2019state}
Shuo Yang, Hans Kumar, Zhaoyuan Gu, Xiangyuan Zhang, Matthew Travers, and Howie
  Choset.
\newblock State estimation for legged robots using contact-centric leg
  odometry.
\newblock \emph{arXiv preprint arXiv:1911.05176}, 2019.

\bibitem[Zhang and Scaramuzza(2018)]{zhang2018tutorial}
Zichao Zhang and Davide Scaramuzza.
\newblock A tutorial on quantitative trajectory evaluation for visual
  (-inertial) odometry.
\newblock In \emph{2018 IEEE/RSJ International Conference on Intelligent Robots
  and Systems (IROS)}, pages 7244--7251. IEEE, 2018.

\bibitem[Zhu and Zhou(2004)]{zhu2004real}
Rong Zhu and Zhaoying Zhou.
\newblock A real-time articulated human motion tracking using tri-axis
  inertial/magnetic sensors package.
\newblock \emph{IEEE Transactions on Neural systems and rehabilitation
  engineering}, 12\penalty0 (2):\penalty0 295--302, 2004.

\bibitem[Zillner et~al.(2020)Zillner, Bisset, Milano, Curry, S{\"o}derg{\aa}rd,
  Tuikka, et~al.]{zillner2020strategic}
Sonja Zillner, David Bisset, Michela Milano, Edward Curry, Caj
  S{\"o}derg{\aa}rd, Tuomo Tuikka, et~al.
\newblock Strategic research, innovation and deployment agenda: Ai, data and
  robotics partnership.
\newblock 2020.

\end{thebibliography}



\end{spacing}

\printthesisindex 

\end{document}